\documentclass[a4paper, 11pt, oneside]{Thesis}  
\graphicspath{Figures/}  
\pdfoutput=1

\usepackage[square, numbers, comma, sort&compress]{natbib}  
\usepackage{verbatim}  
\usepackage{vector}  
\hypersetup{urlcolor=[rgb]{0, 0.234, 0.44}, colorlinks=true}  
\usepackage{epigraph}
\usepackage[T1]{fontenc}
\usepackage{mathrsfs}
\usepackage{bm}
\usepackage{tikz}
\usepackage{animate}
\usepackage{adjustbox}
\usepackage{pifont}
\usepackage{xcolor}
\usepackage{tabularx}
\usepackage{multirow}
\usepackage{rotating}
\usepackage{algorithm}
\usepackage{algpseudocode}
\usepackage{cancel}
\usepackage{caption}

\usepackage{bibentry}
\nobibliography*  

\DeclareMathOperator*{\argmax}{arg\,max}

\begin{document}
\frontmatter      
\title  {Bayesian Deep Learning for Graphs}
\authors  {\texorpdfstring{}
            {Federico Errica}
            {Federico Errica}
            }
\addresses  {\groupname\\\deptname\\\univname}  
\date       {\today}
\subject    {}

\maketitle

\setstretch{1.}  

\fancyhead{}  
\rhead{\thepage}  
\lhead{}  

\pagestyle{fancy}  

\pagestyle{empty}  

\null\vfill
\textit{``O frati'', dissi ``che per cento milia} \\
\textit{perigli siete giunti a l’occidente,} \\
\textit{a questa tanto picciola vigilia} \\

\textit{d’i nostri sensi ch’è del rimanente,} \\
\textit{non vogliate negar l’esperienza,} \\
\textit{di retro al sol, del mondo sanza gente.} \\

\textit{Considerate la vostra semenza:} \\
\textit{fatti non foste a viver come bruti,} \\
\textit{ma per seguir virtute e canoscenza''.} \\

\textit{Li miei compagni fec’io sì aguti,} \\
\textit{con questa orazion picciola, al cammino,} \\
\textit{che a pena poscia li avrei ritenuti;} \\

\textit{e volta nostra poppa nel mattino,} \\
\textit{de’ remi facemmo ali al folle volo,} \\
\textit{sempre acquistando dal lato mancino.}

\begin{flushright}
Dante Alighieri - Divina Commedia - Inferno, Canto XXVI
\end{flushright}

\vfill\vfill\vfill\vfill\vfill\vfill\null
\clearpage  

\newcommand{\st}{s.t. }
\newcommand{\ie}{i.e., }
\newcommand{\eg}{e.g., }
\newcommand{\quotes}[1]{``#1''}
\newcommand{\revise}[1]{\textcolor{red}{#1}}
\newcommand{\norevise}[1]{#1}
\newcommand{\aka}{a.k.a. }
\newcommand{\wrt}{w.r.t.}

\setstretch{1.3}  

\acknowledgements{
\addtocontents{toc}{\vspace{1em}}  

Words could never begin to describe how grateful I am to my supervisors, Alessio Micheli and Davide Bacciu. Throughout this PhD I feel I have grown in so many personal and professional ways thanks to you, and if I have ever learned something it was because of the time you took to share your knowledge with me, to provide support when I doubted myself, and to answer my countless emails at ungodly hours. You are the reason why I chose to pursue a career in research, and in these three years, despite the pandemic, you have never made me feel alone. The only pressure I had was the one I put on myself, because you have always trusted me. Today, all these memories bring me joy, purpose, and strength.

Alessio, Davide, you are the best example of men of science I could have asked for, and I will always be indebted to you for that. I hope that in these few years you perceived the profound respect I have for you, just as I hope to have been a dedicated student, a reliable colleague, and maybe even a good friend. In the years to come, and in every scientific work, I shall apply everything you taught me at the best of my abilities.

I wish to thank the international reviewers of this thesis, Mark Coates and Shirui Pan, for their valuable comments and constructive criticism, which made me question what I thought I already knew and, therefore, gave me yet another opportunity to learn from their distinguished scientific expertise. Many thanks to my internal committee members, Luca Oneto and Roberto Grossi, for the invaluable suggestions during these years, as well as to the PhD coordinators Paolo Ferragina and Antonio Brogi for their intense commitment.

During this PhD I had the chance to work side-by-side with wonderful colleagues. I would like to start by thanking Marco, with whom I shared the immense pleasure (and hardship!) of working at all hours of the day. You always reminded me that good science does not depend on the venue, and I will never forget your calm in the most desperate of deadlines. Without you, my PhD would have been half as enjoyable as it was.

A warm thank you goes to Daniele, for our endless discussions on how the hell Bayesian methods work. I am forever grateful for the time you spent teaching me what you knew and for the precious tea breaks in the kitchen that always became small machine learning workshops.

To Antonio and Andrea, thank you for your enthusiasm while working together at the intersection of our research fields, and for our beer-based evenings at the Orzo Bruno. I will not forget you, if you know what I mean.

I also had the dumb luck to get to know Fabrizio, Bora, Guillaume, Vassilis, Ola, Sebastian, Ludovic, Fabio, Lazar, Pushkar, Anton, Cameron, Daniyar and Martin, among others, during my internship at Facebook London. You made me understand how important and enriching it is to make connections and share our stories. There, I also met Pasquale, who gave me the opportunity to virtually visit UCL and whom I sincerely thank for our profound discussions around the similarities and differences of neural link predictors and deep graph networks.

The practical applications shown in this dissertation stem from two distinct collaborations. Allow me to deeply thank Marco, Roberto, and Raffaello of the VARIAMOLS group at the University of Trento, for their patience when I continuously got lost in the mysteries of molecular dynamics and for giving me the first chance to work on real world problems. Let me also thank Giacomo, Francesco, and Fabio at CNR, as we managed to improve robustness in malware classification problems against nasty intra-procedural code obfuscation techniques.

I wish to whole-heartedly thank all the people of the Computational Intelligence and Machine Learning group (we have a wonderful, brand-new website as well as 3D logos!), because each single one of you gave me something to think about at some point. You have been my source of inspiration so many times while having our frequent chats at the sofas. Keep up with the good work and let's meet at the next ESANN! \\ It was an honor to be part of this amazing group.

While the kitchen was still open, I had the pleasure of sharing refreshing moments with professors, students, and post-docs alike. A word of acknowledgment also goes to them.

I have to say these years have been tough under many aspects, especially because of the pandemic. But having all of you around, physically or virtually, made the journey light and worthwhile.

To me, this PhD has been a great privilege.

\subsection*{Ringraziamenti Personali}

Tengo a ringraziare dal profondo del cuore i miei genitori, Marina e Giampaolo, per il loro amore, fiducia e supporto incondizionati, indispensabili per superare tutte le difficolt\`{a} della vita. Mamma, Babbo, spero di diventare la met\`{a} delle persone meravigliose che siete. Questa tesi \`{e} dedicata a voi.

Sono fermamente convinto che questo lavoro sia anche il frutto dell'affetto profondo di Licia e Piero, che da troppo tempo se ne sono andati ma la cui memoria resta indelebile.

Un ringraziamento particolare va a Martina, che ha sempre creduto in me durante questo percorso \quotes{infernale}. Nei momenti pi\`{u} difficili mi hai sempre spronato a dare il massimo, e non smetter\`{o} mai di essertene grato.

A Irene, Mattia e Giorgio, compagni di percorso e di ufficio, che pi\`{u} che colleghi sono diventati cari amici, dico grazie per la vostra gentilezza, generosit\`{a} e bont\`{a} d'animo.

In tutti questi anni, in periodi pi\`{u} o meno delicati, ho sempre potuto contare sui \quotes{cavalieri del polo}, Giacomo, Iacopo con la I, Jacopo con la J e Nicola, i miei amici di sempre e pi\`{u} intimi confidenti. Un semplice grazie non baster\`{a} mai per delle persone importanti come voi, ma a dirla tutta non basterebbero neppure le parole, considerato anche che non sono mai stato mai bravo in queste cose. Siete e resterete sempre il mio porto sicuro.

Ho lasciato l'ultimo spazio di queste poche pagine ai ragazzi di \quotes{Info Proposte Alcoliche (I.P.A.)}, Daniele, Lorenzo, Marco, Matteo, Michele, Thomas e Tommaso, i miei compagni di universit\`{a}. Con voi ho condiviso i momenti pi\`{u} belli e fuori di testa della mia decade pisana. I ricordi si susseguono incessantemente, e ogni volta mi ci scappa una risata. Conoscervi \`{e} stata una delle pi\`{u} grandi fortune della mia vita.
}
\clearpage  

\addtotoc{Abstract}  
\abstract{
\addtocontents{toc}{\vspace{1em}}  
\rule{1.\columnwidth}{0.3pt}
The adaptive processing of structured data is a long-standing research topic in machine learning that investigates how to automatically learn a mapping from a structured input to outputs of various nature. Recently, there has been an increasing interest in the adaptive processing of graphs, which led to the development of different neural network-based methodologies. In this thesis, we take a different route and develop a Bayesian Deep Learning framework for the adaptive processing of graphs. The dissertation begins with a review of the foundational principles over which most of the methods in the field are built, and the discussion is complemented with a thorough study on graph classification reproducibility issues. We then proceed to bridge the basic ideas of deep learning for graphs with the Bayesian world, by building our deep architectures in an incremental fashion. The theoretical framework allows us to consider graphs with both discrete and continuous edge features, and it produces unsupervised embeddings rich enough to reach the state of the art on a number of classification tasks. We later discover that our approach is also amenable to a Bayesian nonparametric extension, which automatizes the choice of almost all model's hyper-parameters. Real-world applications are incorporated into the discussion to demonstrate the efficacy of deep learning for graphs. The first one concerns the prediction of information-theoretic quantities useful in molecular simulations, a problem tackled with supervised neural models for graphs. After that, we exploit our Bayesian models to solve a malware-classification task in such a way that the prediction is robust to intra-procedural code obfuscation techniques. We conclude the dissertation with our attempt to blend the best of the neural and Bayesian worlds together. The resulting hybrid model is able to predict multimodal distributions conditioned on input graphs, with the consequent ability to model stochasticity and uncertainty better than most works in the literature. Overall, we aim to provide a Bayesian perspective into the articulated research field of deep learning for graphs.
\rule{1.\columnwidth}{0.3pt}
}


\pagestyle{fancy}  

\lhead{\emph{Contents}}  
\hypersetup{linkcolor=black}
\tableofcontents  



\setstretch{1.5}  
\clearpage  
\lhead{\emph{Abbreviations}}  
\newcommand{\ML}{machine learning}
\newcommand{\CG}{coarse-grained}
\newcommand{\T}{\mathcal{T}}

\definecolor{applegreen}{rgb}{0.55, 0.71, 0.0}
\definecolor{ao(english)}{rgb}{0.0, 0.5, 0.0}
\definecolor{darkpastelred}{rgb}{0.76, 0.23, 0.13}

\newcommand{\present}{\textcolor{ao(english)}{\ding{51}}}
\newcommand{\absent}{\textcolor{darkpastelred}{\ding{55}}}

\newcommand{\R}[1]{{\mathbb{R}^{#1}}}

\newcommand{\hypotheses}{{\mathcal{H}}}
\newcommand{\dataset}{{\mathcal{D}}}
\newcommand{\Vset}[1]{{\mathcal{V}_{#1}}}
\newcommand{\Eset}[1]{{\mathcal{E}_{#1}}}
\newcommand{\Xset}[1]{{\mathcal{X}_{#1}}}
\newcommand{\Aset}[1]{{\mathcal{A}_{#1}}}

\newcommand{\uv}{{uv}}
\newcommand{\vu}{{vu}}
\newcommand{\puv}{{(u,v)}}
\newcommand{\pvu}{{(v,u)}}
\newcommand{\suv}{{\{u,v\}}}
\newcommand{\svu}{{\{v,u\}}}

\newcommand{\e}[1]{{e_{#1}}}
\newcommand{\ebold}[1]{{\bm{e}_{#1}}}

\newcommand{\x}[1]{{x_{#1}}}
\newcommand{\boldx}[1]{{\bm{x}_{#1}}}
\newcommand{\h}[1]{{h_{#1}}}
\newcommand{\boldh}[1]{{\bm{h}_{#1}}}
\newcommand{\boldhell}[2]{{\bm{h}^{#2}_{#1}}}

\newcommand{\lprev}{{\ell-1}}
\newcommand{\lcurr}{{\ell}}
\newcommand{\lnext}{{\ell+1}}
\newcommand{\N}[2]{{\mathcal{N}^{#2}_{#1}}}
\newcommand{\qN}[2]{{\mathbf{q}^{#2}_{\N{#1}{}}}}
\newcommand{\boldq}[1]{{\bm{q}_{#1}}}
\newcommand{\boldqell}[2]{{\bm{q}^{#2}_{#1}}}

\newcommand{\E}[2]{{\mathbb{E}_{#1}[#2]}}
\newcommand{\Q}[2]{{Q_{#1}^{#2}}}
\newcommand{\boldQ}[2]{{\boldsymbol{Q}_{#1}^{#2}}}
\newcommand{\pmf}{p.m.f.}
\newcommand{\cdf}{c.d.f.}
\newcommand{\pdf}{p.d.f.}
\newcommand{\iid}{i.i.d.}
\newcommand{\wlogity}{w.l.o.g.}
\newcommand{\likelihood}{{\mathcal{L}}}
\newcommand{\DP}[1]{DP{#1}}

\newcommand{\iCGMM}{\textsc{iCGMM}}
\newcommand{\CGMM}{\textsc{CGMM}}
\newcommand{\ECGMM}{\textsc{E-CGMM}}
\newcommand{\GMDN}{\textsc{GMDN}}

\newcommand{\MDN}{\textsc{MDN}}
\newcommand{\DGN}{\textsc{DGN}}
\newcommand{\RAND}{\textsc{RAND}}
\newcommand{\HIST}{\textsc{HIST}}

\newcommand{\Baseline}{\textsc{Baseline}} 
\newcommand{\DGCNN}{\textsc{DGCNN}} 
\newcommand{\DiffPool}{\textsc{DiffPool}}
\newcommand{\ECC}{\textsc{ECC}}
\newcommand{\DGI}{\textsc{DGI}}
\newcommand{\GIN}{\textsc{GIN}} 
\newcommand{\GraphSAGE}{\textsc{GraphSAGE}}

\listofsymbols{ll}  
{

AI & Artificial Intelligence \\
\aka & Also Known As \\
BA & Barab\'asi-Albert \\
BNP & Bayesian Nonparametric \\
CDE & Conditional Density Estimation \\
\cdf{} & Cumulative Density Function \\
CE & Cross-Entropy \\
CG & Call Graph \\
CGMM & Contextual Graph Markov Model \\
CPU & Central Processing Unit \\
CRP & Chinese Restaurant Process \\
CRF & Chinese Restaurant Franchise \\
CV & Cross Validation \\
DAG & Directed Acyclic Graph \\
DBGN & Deep Bayesian Graph Network \\
DGI & Deep Graph Infomax \\
DGGN & Deep Generative Graph Network \\
DGN & Deep Graph Network \\
DNGN & Deep Neural Graph Network \\
DOAG & Directed Ordered Acyclic Graph \\
DPAG & Directed Positional Acyclic Graph \\
DP & Dirichlet Process \\
e.g. & Exempli Gratia \\
ECC & Edge Conditioned Convolution \\
E-CGMM & Extended Contextual Graph Markov Model \\
EM & Expectation Maximization \\
ER & Erd\H{o}s-R\'enyi \\
GAE & Graph Auto-Encoder \\
GCN & Graph Convolutional Network \\
GG-NN & Gated Graph Neural Network \\
GIN & Graph Isomorphism Network \\
GMDN & Graph Mixture Density Network \\
GNN & Graph Neural Network \\
GPU & Graphics Processing Unit \\
GraphESN & Graph Echo State Network \\
GRL & Graph Representation Learning \\
HDP & Hierarchical Dirichlet Process \\
HMM & Hidden Markov Model \\
iCGMM & Infinite Contextual Graph Markov Model \\
i.e. & Id Est \\
\iid{} & Independent and Identically Distributed \\
LSTM & Long Short-Term Memory \\
MAE & Mean Absolute Error \\
MCMC & Markov Chain Monte Carlo \\
MDN & Mixture Density Network \\
MLP & Multi-Layer Perceptron \\
ML & Machine Learning \\
MLE & Maximum Likelihood Estimation \\
MPNN & Message Passing Neural Network \\
MSE & Mean Squared Error \\
NLP & Natural Language Processing \\
\pdf{} & Probability Density Function \\
\pmf{} & Probability Mass Function \\
ReLU & Rectifier Linear Unit \\
R-GCN & Relational Graph Convolutional Network \\
RNN & Recurrent Neural Network \\
SD & Structured Data \\
SDL & Structured-Data Learning \\
SGD & Stochastic Gradient Descent \\
SIR & Susceptible-Infectious-Recovered \\
SOM & Self-Organizing Map \\
SP & Switching Parent \\
SVM & Support Vector Machine \\
WL & Weisfeiler-Lehman \\
\wlogity{} & Without Loss of Generality \\
\wrt{} & With Respect To
}




\setstretch{1.3}  

\pagestyle{empty}  
\dedicatory{This thesis is dedicated to my parents \\ and to Licia and Piero, wherever you are.}

\addtocontents{toc}{\vspace{2em}}  


\mainmatter	  
\lhead{}  
\pagestyle{fancy}  


\chapter{Introduction}
\label{chapter:introduction}
\epigraph{\textit{Ahi quanto a dir qual era è cosa dura \\ esta selva selvaggia e aspra e forte \\ che nel pensier rinova la paura! \\~\\ Tant’è amara che poco è più morte; \\ ma per trattar del ben ch’i’ vi trovai, \\ dirò de l’altre cose ch’i’ v’ho scorte.}}{\textit{Inferno - Canto I}}

\section{Motivations}
Abstraction and compositionality are the indispensable principles that we, as humans, avail ourselves of in order to organize, compress, and comprehend the immense amount of information perceived at every instant of our lives. In its broader sense, abstraction is the process of simplifying some aspects of a complex system that are unnecessary to our original purpose. It grants us the ability to find similar patterns and connections between relatively distant ideas or, for instance, research fields. Compositionality, on the other hand, is the tendency to design and understand systems as made of smaller but entangled sub-components. 
That is to say, most dynamics of the real world are arguably best modeled in relational terms, by considering entities that interact with each other: bees organize themselves in a hierarchy of roles, each of which is crucial for the survival of the colony; in classical physics, the movement of planets can be explained by their mutual interactions through gravity; in chemistry, the disposition of atoms in space, together with their chemical bonds, contribute to the characterization of the properties of the molecule under consideration.

In Computer Science, a data structure is a collection of values that adheres to the principles of abstraction and compositionality and helps us to efficiently organize the information. Depending on our needs different structures are feasible, such as sequences and trees, but hereinafter we will be concerned with the notion of graphs. 
A graph is a data structure composed of entities freely interacting with each other, so it should come as no surprise that graphs are used in the most disparate problems, from chemistry, physics, and mathematics to linguistics and network science.

Many a time, the combinatorial nature of such structured problems makes it hard to find exact solutions with classical algorithms. In these circumstances, a viable option may reside in \ML{} techniques. The adaptive processing of Structured Data (SD) is indeed a longstanding research topic of \ML{} \cite{bacciu_gentle_2020}, whose goal is to learn a mapping from the input structure to the desired output. Over the years, researchers have developed a plethora of specialized methods to process sequences and trees \cite{sperduti_supervised_1997,hochreiter_long_1997,frasconi_general_1998} relying on their structural regularities, but it was not until recently, after the advent of deep learning, that considerable interest was devoted to the study of adaptive methodologies for graphs. Encouraged by the abrupt availability of graph data and new hardware devices, as well as stimulated by the open research challenges ahead, researchers began exploring the many facets of what is now called Deep Learning for Graphs. This auspicious research direction is characterized by a local and iterative processing of the structured datum, which favors efficiency over combinatorial complexities and is functional to the spreading of information through the graph's entities. 
The approach also enables automatic features' extraction to solve a task with no human intervention.

Recent research on deep learning for graphs has been very prolific and intense, especially as regards neural networks. The reasons are fairly straightforward: neural networks are incredibly flexible, they can be implemented on hardware accelerators, and we know how to propagate the error signal through very deep architectures. 
At the same time, such productivity has come at the price of a certain forgetfulness, if not lack of appropriate referencing, of pioneering and consolidated methodologies. Troubling trends on the reproducibility of experiments and the robustness of evaluation protocols immediately followed, generating confusion and ambiguities across the whole literature. 

In addition, it could be argued that the Bayesian research direction for deep learning on graphs, \ie statistical methods modeling the probability distribution of graphs, has been abundantly overlooked in spite of the advantages that the probabilistic approaches usually bring to the table, \eg expressing causal relationships in the data, incorporating prior information in the process, modeling uncertainty, and building unsupervised embeddings from the posterior distribution. Perhaps, part of the general hesitation is caused by the difficulty of defining deep and end-to-end trainable architectures.

\section{Objectives}
Starting from a unified, objective, and high-level overview of deep learning for graphs, the main ambition of this thesis is to develop a fully probabilistic framework that embraces the most distinctive traits of the field in a Bayesian context. The cross-pollination of ideas between the neural and Bayesian worlds will naturally emerge throughout the manuscript, for the sake of a mathematical formalization rooted in simplicity, efficiency, and empirical efficacy. To move towards our goal, we will first have to understand and review the basic principles that guide the development of most deep learning architectures for graphs. Similarly, we shall attempt at mitigating the reproducibility issues that would make our empirical analyses inconsistent with other works in the literature. With solid ground below our feet, we will then devise deep, unsupervised, and probabilistic models for graphs, called Deep Bayesian Graph Networks, that approximate the data distribution through latent 
factors. As a by-product of the knowledge gained, we shall additionally investigate how to deal with graph-related uncertainty by mixing neural and probabilistic components, therefore concluding this dissertation with both worlds working in close liaison.

\section{Contributions}
In view of the objectives outlined above, our main contributions can be ascribed to an introductory review of the building blocks that are peculiar to deep learning for graphs, accompanied by a rigorous and standardized evaluation that will allow the theoretical and practical analysis of probabilistic and hybrid models. We complement the discussion with some examples of applications highlighting the advantages of the adaptive processing of structured data.

\paragraph*{Unified Review \cite{bacciu_gentle_2020}} The analysis of a large body of literature, alongside the foundational works of the field, revealed that there exist elemental principles that govern how structured information is usually processed. We believed that the creation of a high-level description of such basic concepts, rather than the systematic analysis of the recent advances, would have benefited both beginners and experts. In this sense, the discourse adopts a top-down organization, in which details are presented only after the key notions have been given. Additionally, we provide a uniform mathematical notation under which different models are compared, to show how sometimes there are subtle but meaningful technical differences in the definition of the main operations. Finally, we systematically organize a consistent number of works according to their major characteristics, \eg how they propagate information, the choice of the number of layers, and their nature.

\paragraph*{Fair and Robust Empirical Re-evaluation \cite{errica_fair_2020}} As anticipated, the large stream of recent works has caused severe issues in reproducibility and standardization of the experimental settings. To alleviate this, we propose a robust re-evaluation of various models across several graph classification benchmarks. Starting from a report of the specific issues of each paper under examination, we proceeded to run more than 47000 experiments to fairly compare all models under the same controlled environment. The incorporation of a structure-agnostic baseline in the process led to the discovery that, in some cases, said baseline performs better than most of these deep learning models for graphs.

\paragraph*{Basic Deep Probabilistic Framework for Graphs \cite{bacciu_contextual_2018,bacciu_probabilistic_2020}} The first methodological contribution of the thesis is the Contextual Graph Markov Model, our attempt at building a fully probabilistic framework for deep learning on graphs. Borrowing ideas from pioneering works, the construction of the deep architecture is incremental, with each layer being trained after another, and the embeddings generation is completely unsupervised. We provide a probabilistic implementation of the neighborhood aggregation mechanisms that operate under the hood, as well as closed-form update equations that guarantee convergence to a local minima of the likelihood landscape. We empirical evaluate our approach on classification benchmarks, and we discover that the unsupervised construction of representations for the graph and its individual entities is surprisingly rich, with subsequent classification performances that are quite close to the state of the art. Furthermore, we analyze the behavior of the model across different layers, showing that depth is of paramount importance to achieve a better generalization.

\paragraph*{Architectural Extension of the Framework \cite{atzeni_modeling_2021}} The Contextual Graph Markov Model can deal with discrete edge information, but as soon as we have more articulated edge features it becomes tricky how to incorporate them into the mathematical formulation. Instead of resorting to hand-crafted discretization techniques, we choose to learn a discretization mapping via an architectural extension of the model. In particular, an additional Bayesian network captures the latent discrete 
factors responsible for the generation of edge features, and such factors are then incorporated into the original model at the next layer of the architecture. We empirically show that this mechanism is able not only to improve performances over basic edge discretization techniques, but it also boosts classification accuracy whenever edge features are not available. 

\paragraph*{Bridging Graph Learning and Bayesian Nonparametrics} The automatic selection of hyper-parameters is an intriguing research topic that finds elegant and mathematically sound solutions in the Bayesian nonparametric literature. These methods support the selection of the \quotes{right} number of latent factors, or clusters, to use in a Bayesian network. For this reason, we apply a Bayesian nonparametric treatment to each layer of the Contextual Graph Markov Model, motivated by the need to automatize as much as possible the choice of its most important hyper-parameter. We also develop a faster but approximated version of the alghoritm that scales to larger graphs, without losing predictive accuracy on the empirical classification tasks considered. Our method, born from the cross-fertilization of ideas belonging to relatively distant fields, reduces by more than 90\% the size of the unsupervised graph embeddings, thus saving a great amount of computational resources for the supervised classifier built on top of said embeddings.

\paragraph*{Hybrid Approach to Uncertainty Modeling \cite{errica_graph_2021}} The neural and probabilistic techniques for graph learning undoubtedly have complementary advantages, namely the flexibility of neural networks and the ability of Bayesian networks to naturally handle uncertainty via probability distributions. We realized that some problems, for instance the prediction of stochastic epidemic outcomes in a social network, could not be handled by the current models in the literature. For this reason, we developed the Graph Mixture Density Network, a fairly extensible framework to output multimodal distributions conditioned on input graphs. We provide evidence that previous deep learning approaches for graphs produce unsatisfactory results in the aforementioned contexts, whereas our proposal can express its uncertainty about the plausible continuous value(s) to predict, adding a degree of trustworthiness to the process.

\paragraph*{Applications \cite{errica_deep_2021,errica_robust_2021}} Throughout the manuscript, we take advantage of two practical real-world problems to support our claims about the importance of the methodologies discussed. We will present an application of deep learning for graphs to the field of molecular biosciences, where the goal is to approximate the prediction of a molecule's information-theoretic quantity in a fraction of the time required by the original algorithm. If successful, the learned model would enable a quasi-exhaustive exploration of the output space, due to the combinatorial nature of the problem. \\ The other application concerns malware classification of software that is subject of obfuscation techniques, in particular those that do not change the topology of the associated \quotes{call graph}. We will show that our probabilistic models are able to perform very well on a classification task where a structure-agnostic baseline dramatically fails, and such models are also robust to those software obfuscations.

\clearpage
\section{Thesis' Outline}
The thesis is organized into 5 more chapters.

In Chapter \ref{chapter:background}, we first review the basic definitions of probability, Bayesian learning, and the models we will take inspiration throughout the rest of this work. Then, we will talk about the formal definition of graphs, thus initiating the reader to the most used mathematical notation. Finally, we will shortly summarized related approaches for the adaptive processing of graphs that do not directly belong to deep learning.

In Chapter \ref{chapter:gentle-introduction}, we introduce the basic principles of machine learning for graphs, regardless of the nature of the models, let them be neural, probabilistic, or hybrid. This broad overview is integrated by an empirical fair comparison of models under the same graph classification settings, in the attempt to partially take back control of the situation, made unstable by the recent wave of (re-)discovery. We conclude the chapter with an application from the field of molecular biosciences.

In Chapter \ref{chapter:dbgn}, we present the main methodological contributions of this thesis, which fall under the name of Deep Bayesian Graph Networks. The exposition is organized in such a way that new techniques can be seen as extensions of previous ones, and many parallelism are made with the basic notions of Chapter \ref{chapter:gentle-introduction}. For each of the models presented, we will show variegated empirical analyses in support of the benchmark results. At the end of the chapter, we apply the developed models to a real-world malware classification task.

In Chapter \ref{chapter:hybrid}, we take the best of the neural and probabilistic worlds and design a hybrid model, called Graph Mixture Density Network, to output multimodal distributions conditioned on arbitrary input graphs. The empirical evaluation on synthetic random graphs and real-world chemical tasks is meant to show that, for some problems, the \quotes{standard} approach to deep learning for graphs fails at producing the correct output.

In Chapter \ref{chapter:conclusions}, we add summarizing thoughts to our dissertation, discussing open problems and future research directions.  

\clearpage
\section{Origin of the Chapters}
Most of the results dispensed in this thesis have already been presented at conferences and/or published at journals. The list below is the outcome of hard but much pleasant work with a number of co-authors, who gave us the opportunity to collaborate at the cross-road of different research fields.

\paragraph*{Chapter \ref{chapter:gentle-introduction}}
\begin{itemize}
    \item Sections \ref{sec:contextual-processing} to \ref{sec:dl4g}: \\~\\ \bibentry{bacciu_gentle_2020}
    \item Section \ref{sec:scholarship-issues}: \\~\\ \bibentry{errica_fair_2020}
    \item Section \ref{sec:application-molecular}: \\~\\ \bibentry{errica_deep_2021}
\end{itemize}

\paragraph*{Chapter \ref{chapter:dbgn}}
\begin{itemize}
    \item Section \ref{sec:cgmm}: \\~\\ \bibentry{bacciu_contextual_2018} \\~\\ \bibentry{bacciu_probabilistic_2020}
    \item Section \ref{sec:ecgmm}: \\~\\ \bibentry{atzeni_modeling_2021}
    \item Section \ref{sec:icgmm} reports unpublished work, currently under review.
    \item Section \ref{sec:application-security}: \\~\\ \bibentry{errica_robust_2021}
\end{itemize}

\paragraph*{Chapter \ref{chapter:hybrid}}
\begin{itemize}
    \item All Sections: \\~\\ \bibentry{errica_graph_2021}
\end{itemize}

\chapter{Preliminaries}
\label{chapter:background}
\epigraph{\textit{E poi che la sua mano a la mia puose \\ con lieto volto, ond’io mi confortai, \\ mi mise dentro a le segrete cose.}}{\textit{Inferno - Canto III}}

In this chapter, we shall delineate basic definitions and techniques that will be used throughout the rest of the manuscript. In doing so, it is assumed that the reader is familiar with linear algebra and the fundamental \ML{} concepts such as supervised and unsupervised learning, multi-layer perceptrons, hidden units,  and activation functions, to name a few. We shall begin with a probability refresher and an introduction to some probabilistic modeling techniques. Special attention is devoted to mixture models, as the probabilistic models developed in this thesis inherit many of their characteristics. Then, we move to more advanced topics, such as ensity networks for flat data, which borrow ideas from both neural and probabilistic worlds, and Bayesian non-parametric mixture models, where the model's complexity grows with the data. \\
We shall continue with a much needed discourse about the multifaceted nature of graphs, introducing standard definitions and mentioning the challenges that \ML{} models have to face when handling this kind of structured data: these include the presence of cycles and the absence of a known ordering of the graph entities. Also, we describe particular instances of graphs with a more rigid structure, for which learning models are known and well-studied. Then, we discuss random graphs and the process to generate them: we will necessitate synthetic datasets to carry out some of our experiments. To conclude the chapter, we provide a brief summary of different research directions that are complementary to the topics presented in this manuscript. \clearpage

\section{Probabilistic modeling}
We now review the principles of probability theory and some probabilistic modeling techniques that are especially relevant for this thesis. The reader can refer to \cite{bishop_pattern_2006,barber_bayesian_2012} for a complete treatment of these topics.

\subsection{Probability Refresher}
\subsubsection{Basic Definitions}
Probability theory provides us with the mathematical tools to rigorously formalize our intuition of uncertainty and randomness \cite{bruni_models_2017}. To this aim, we first introduce the set of all possible outcomes of an experiment with the symbol $\Omega$ (the \textit{sample space}) and the set of events of interest that may occur as $\mathscr{A}\subseteq \mathscr{P}(\Omega)$, where $\mathscr{P}(\cdot)$ denotes the powerset operator; in particular, we require $\mathscr{A}$ to be a $\sigma$-algebra (or $\sigma$-field).
\begin{definition}[$\bm{\sigma}$\textbf{-algebra}]
Let $\Omega$ be the set of possible outcomes and consider the set of events $\mathscr{A}\subseteq \mathscr{P}(\Omega)$. Then, $\mathscr{A}$ is a $\sigma$-algebra if the following holds:
\begin{enumerate}
\item $\emptyset \in \mathscr{A}$ (accounting for the impossible event)
\item $\forall A \in \mathscr{A} \implies (\Omega/A) \in \mathscr{A}$  (closure under complement)
\item $\forall\{A_n\}_{n \in \mathbb{N}} \subseteq \mathscr{A} \implies \bigcup_{i \in \mathbb{N}} A_i \in \mathscr{A}$ (closure under countable union).
\end{enumerate}
\end{definition}
For instance, if we wanted to toss a coin, we would have a sample space $\Omega=\{\text{head,tail}\}$ and $\mathscr{A} = \mathscr{P}(\Omega) = \{\{\emptyset\}, \{\text{head}\}, \{\text{tail}\}, \{\text{head, tail}\}\}$.  Instead, if we had considered an experiment made of two coin tosses, a single outcome could have been $\omega=\{\text{head, head}\} \in \Omega$.

In order to assign a number to a subset of events, the reason for which will become clear in a moment, we need the notion of a measure over sets.
\begin{definition}[\textbf{Measure}]
Let $\mathscr{A}$ be a $\sigma$-algebra defined over $\Omega$. A function $f:$  $\mathscr{A}\rightarrow [0,+\infty]$ is a measure on $(\Omega,\mathscr{A})$ whenever:
\begin{enumerate}
\item $\forall A \in \mathscr{A} \implies f(A) \ge 0$ (non-negativity)
\item $f(\emptyset) = 0$ (null empty set)
\item For all countable collections $\{A_n\}_{n \in \mathbb{N}} \in \mathscr{A}$ of disjoint sets it holds  $f(\bigcup_{i \in \mathbb{N}} A_i) = \sum_{i \in \mathbb{N}} f(A_i)$ ($\sigma$-additivity).
\end{enumerate}
\end{definition}
Therefore, a measure is a function that takes a set of elementary outcomes and outputs a real number. Intuitively, this number may be regarded as the \quotes{size} of that set. Moreover, the pair $(\Omega,\mathscr{A})$ is called a measurable (or Borel) space. At this point, we explit these concepts to define what a probability is.
\begin{definition}[\textbf{Probability}]
A probability is a measure $P$ on $(\Omega,\mathscr{A})$ where $P(\Omega)=1$. Moreover, the tuple $(\Omega, \mathscr{A},P)$ is called a \textbf{probability space}.
\end{definition}
Continuing with the coin flip example, we can imagine a fair coin toss experiment where the probabilities are: $P(\{\emptyset\})=0$, $P(\{\text{head}\})=\frac{1}{2}$, $P(\{\text{tail}\})=\frac{1}{2}$, and $P(\{\text{head,tail}\})=1.$

Throughout the following sections and chapters, we will frequently encounter the notion of random variable. Informally, a random variable is a variable whose values are associated with a probability of occurrence.
\begin{definition}[\textbf{Random Variable}]
Given a probability space $(\Omega, \mathscr{A},P)$, a random variable is a measurable function $X:\Omega \rightarrow E$ \st $\{\omega \in \Omega \mid X(\omega) \in E \} \in \mathscr{A}$. \end{definition}
A random variable $X$ can be \textbf{discrete} or \textbf{continuous}, depending on the nature of its image $E$. To model the coin toss experiment, we can construct a discrete random variable $X$ such that $E=\{0,1\}$, $X(\text{head}) = 0$, $X(\text{tail}) = 1$, and use the notation $P(X=\text{head}) = \frac{1}{2}$ and $P(X=\text{tail}) =\frac{1}{2}$ to convey the same information as above. In general, different outcomes may be assigned the same discrete value in $E$. For example, if we toss a six-face dice two times and compute the sum of the numbers on the faces, we can have 36 possible outcomes (the size of the sample space) but only 11 different results (the size of the discrete set $E$).

Closely related to the concept of random variable is the notion of stochastic (or random) process. A stochastic process allows to mathematically model the behaviour of complex systems by considering families of random variables indexed by an appropriate set.
\begin{definition}[\textbf{Stochastic Process}] Given a probability space $(\Omega, \mathscr{A},P)$ and a set $T$, a stochastic process refers to a family of random variables $\{X_t\}_{t \in T}$. The values that each random variable $X_t$ can take are called \textit{states}.
\end{definition}

Any random variable is completely characterised by its Cumulative Distribution Function (or probability law), which describes the probability that the value assumed by the random variable is smaller than a given parameter.
\begin{definition}[\textbf{Cumulative Distribution Function} (\cdf{})]
The cumulative distribution function $F$ of a random variable $X$ is defined as $F(X\leq x) = P(\{\omega \in \Omega \mid X(\omega) \leq x \})$, which is abbreviated as $P(X\leq x)$.
\end{definition}
For discrete random variables, we can also consider the tabular \textbf{probability mass function} (\pmf{}) $p_X(x) = P(X=x)$, whereas
in the case of continuous random variables we define the \textbf{probability density function} (\pdf) as $f_X(x) = \frac{d}{dx}F_X(x)$. When clear from the context, we shall use the notation $P(X=x)$ (or $P(x)$ for short) in place of $p_X(x)$ or $f_X(x)$. Also, the \textbf{support} of a probability distribution is, the set of values for which it returns a non-zero probability of occurrence.

Many's the time we are interested in the probability of more than one event occurring. We formalize this case with a set of random variables, each capturing the occurrence of a specific event.  Accordingly, we call the \textbf{joint probability distribution} $P(X_1=x_1, X_2=x_2,\dots, X_n=x_n)$ the multivariate distribution that represents this particular stochastic process. In case a set of $n$ random variables is said to be \textbf{mutually independent}, the joint probability decomposes into the product of the single terms, \ie $\prod_{i=1}^n P(X_i=x_i)$. This is because knowing about a given $X_i$ does not change our uncertainty about another random variable and viceversa. Moreover, when a set of random variables is mutually independent and each variable has the same probability distribution, the variables are said to be \textbf{independently and identically distributed} (\iid). On the contrary, when the realization of an event $Y=y$  has an effect on the occurrence of other random variables, we talk about \textbf{conditional probabilities}, denoted by $P(X_1=x_1, X_2=x_2,\dots, X_n=x_n\mid Y=y)$. Like mutual independence, random variables are \textbf{conditionally independent} if it holds $P(X_1=x_1, X_2=x_2,\dots, X_n=x_n\mid Y=y)=\prod_{i=1}^n P(X_i=x_i|Y=y)$. \\ We are now ready to define the crucial rules that will be extensively used in the following.
\begin{definition}[\textbf{Sum Rule}]
Given a random variable $X$, the sum of probabilities over all its values must sum to 1, \ie $\sum_x P(X=x) = 1$.
\end{definition}
\begin{definition}[\textbf{Product Rule}, \aka Chain Rule]
The joint distribution of $n$ variables can always be rewritten as
\begin{align}
& P(X_1=x_1, X_2=x_2,\dots, X_n=x_n)= \nonumber \\
& \prod_{i=1}^n P(X_i=x_i|\bigcap_{j=1}^{i-1} X_j=x_j). \nonumber
\end{align}
\end{definition}
\begin{definition}[\textbf{Marginalization}]
Combining the sum and product rules, and given two random variables $X$ and $Y$ \wlogity, we can obtain the marginal probabilities of $X$ and $Y$ as follows:
\begin{align}
& P(X=x) = \sum_y P(X=x,Y=y) \nonumber \\
& P(Y=y) = \sum_x P(Y=y, X=x) \nonumber
\end{align}
\end{definition}

Oftentimes, we are also interested in computing the average value that $X$ assumes, which is called the expected value of $X$.
\begin{definition}[\textbf{Expected Value}]
Given a random variable $X$ defined over a probability space $(\Omega, \mathscr{A},P)$, the expected value of $X$ is defined as
\begin{align}
& \E{}{X} = \int_{\omega \in \Omega} X(\omega) dP(\omega), \nonumber
\end{align}
where the Lebesgue integral is taken with respect to the measure $P$. For a discrete random variable $X$ with a finite number of attainable states and a known \pmf{}, the above equation simplifies to:
\begin{align}
& \E{}{X} = \sum_x x P(X=x). \nonumber
\end{align}
Instead, for continuous random variables whose distribution $P$ has a \pdf{}, we can write
\begin{align}
& \E{}{X} = \int_\R{} x f_X(x) dx. \nonumber
\end{align}
\end{definition}
In the remainder of this thesis, whenever we want to make explicit the distribution over which we are computing the expectation of a certain variable, we will use the notation $\E{x \sim P}{X}$. In other words, we compute the expectation with respect to the values of $x$ sampled from the distribution $P$.

Finally, we introduce the notion of (discrete-time) Markov Chain to later discuss about inference algorithms.
\begin{definition}[\textbf{Markov Chain}]
Let $(\Omega, \mathscr{A},P)$ be a probability space, and consider a stochastic process $\{X_t\}_{t \in T}$ where $T$ is a totally ordered set. Then, $\{X_t\}_{t \in T}$ is a Markov Chain whenever, for all sequences $t_0 < \dots < t_n < t_{n+1}$ and for all states $x_0,\dots,x_n,x_{n+1}$ the 1-st order \textbf{Markov property} holds:
\begin{align*}
    P(X_{n+1}=x_{n+1}\mid X_n=x_n,\dots,X_0=x_0) = P(X_{n+1}=x_{n+1}\mid X_n=x_n).
\end{align*}
We can easily generalize this definition to the \textbf{k-th order} Markov property
\begin{align*}
    P(x_{n+1}\mid X_n=x_n,\dots,X_0=x_0) = P(X_{n+1}=x_{n+1}\mid X_n=x_n,\dots,X_{n-k}=x_{n-k}),
\end{align*}
giving rise to a \textbf{higher-order} Markov Chain.
\end{definition}

When a process satisfies the Markov property, we say it is \textit{Markovian}.

\subsubsection{Useful Distributions}
As most of this work will be devoted to the development of deep and probabilistic models for graphs, it is useful to briefly introduce the distributions that model discrete and continuous data features.

\begin{figure}[ht]
\begin{subfigure}
  \centering
  \includegraphics[width=0.49\textwidth]{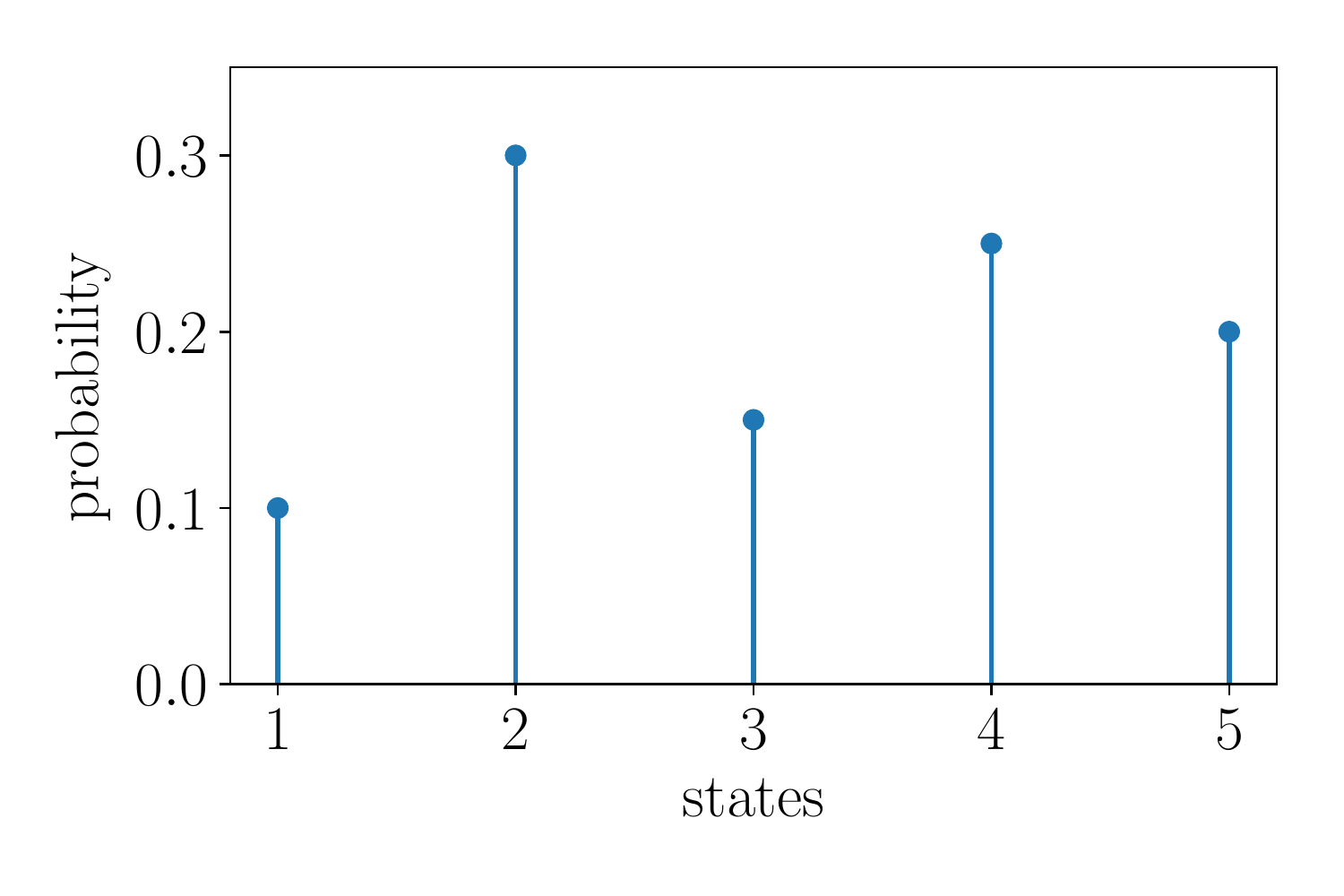}
\end{subfigure}%
\begin{subfigure}
  \centering
  \includegraphics[width=0.49\textwidth]{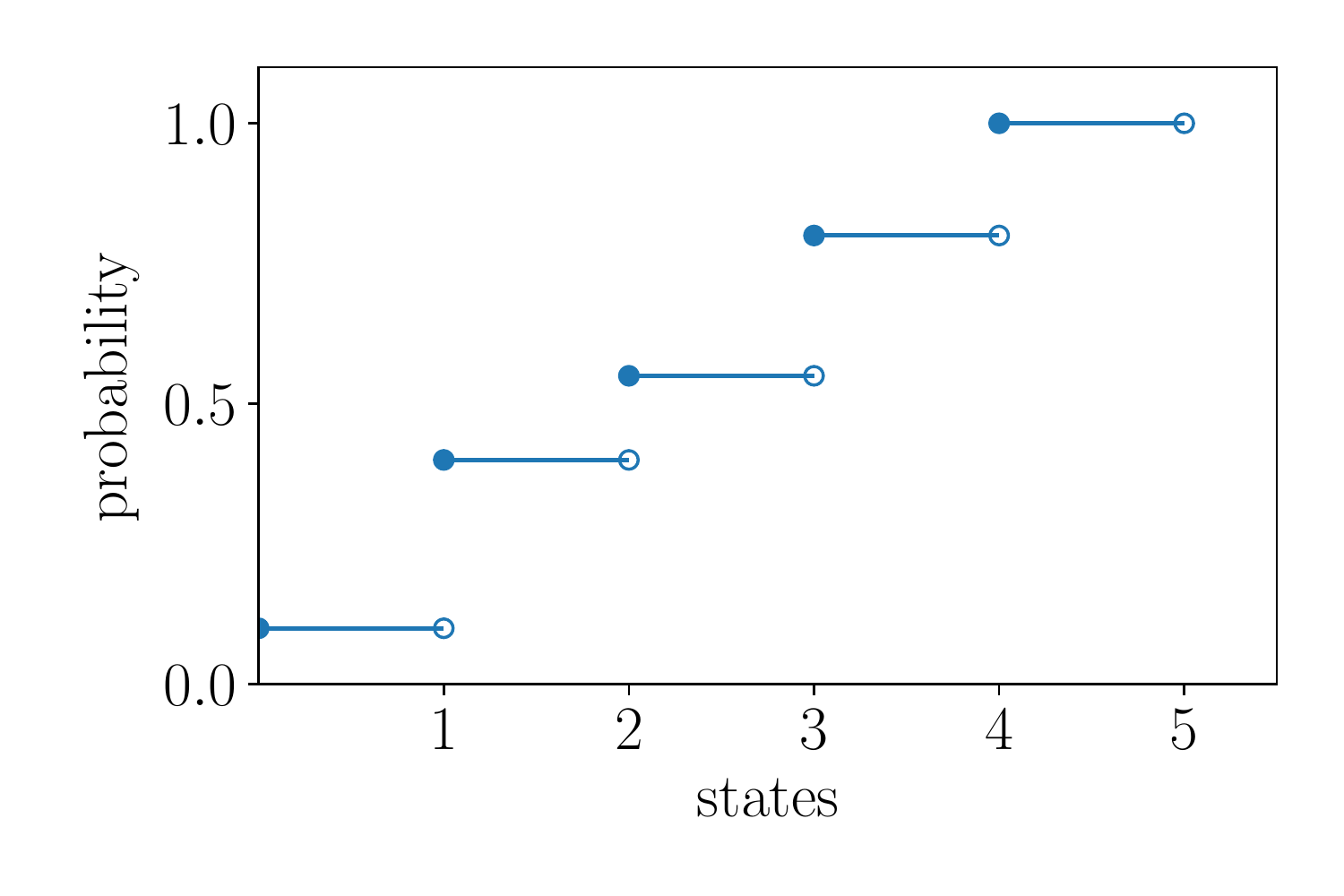}
\end{subfigure}
\caption{We present a possible realization of a categorical distribution through its probability mass function (left) and cumulative distribution function (right). Empty dots symbolize non-smooth jumps to the next probability value.}
\label{fig:categorical}
\end{figure}

\paragraph*{Categorical Distribution.}
The categorical distribution is a discrete probability distribution working on a finite set of values of size $C$. It is often used by discrete random variables that model $C$ different possible outcomes, and it can conveniently be represented as a real vector of size $C$ whose entries sum to 1.  The parameters of the distributions are given by the $C$ different probabilities $p_i$ that constitute said vector. Simply put, the \pmf{} of this distribution writes
\begin{align*}
& p(X=i) = p_i \ \ \forall i \in \{1,\dots,C\}.
\end{align*}
Figure \ref{fig:categorical} depicts the \pmf{} and \cdf{} of a categorical distribution.

\begin{figure}[ht]
\begin{subfigure}
  \centering
  \includegraphics[width=0.49\textwidth]{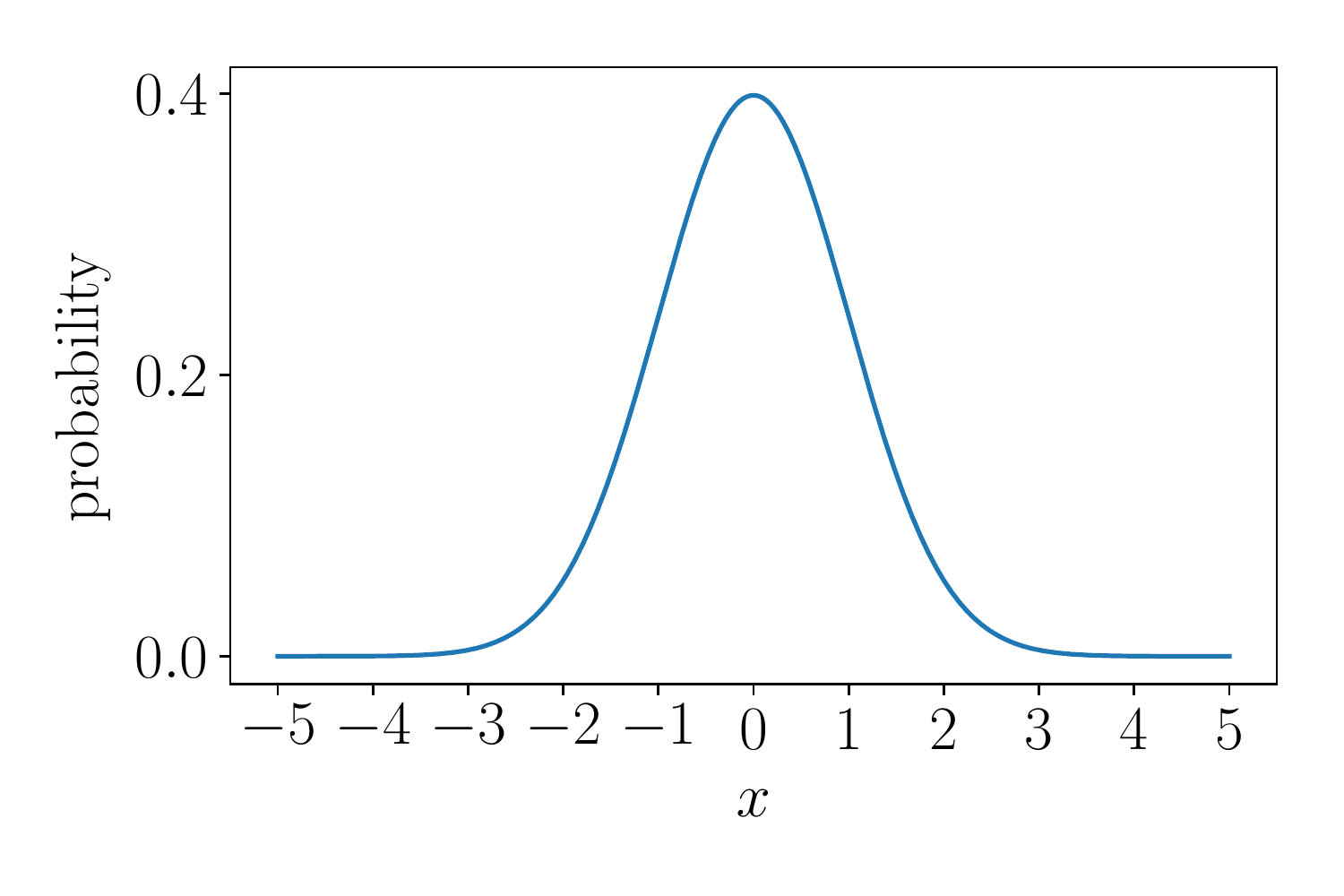}
\end{subfigure}%
\begin{subfigure}
  \centering
  \includegraphics[width=0.49\textwidth]{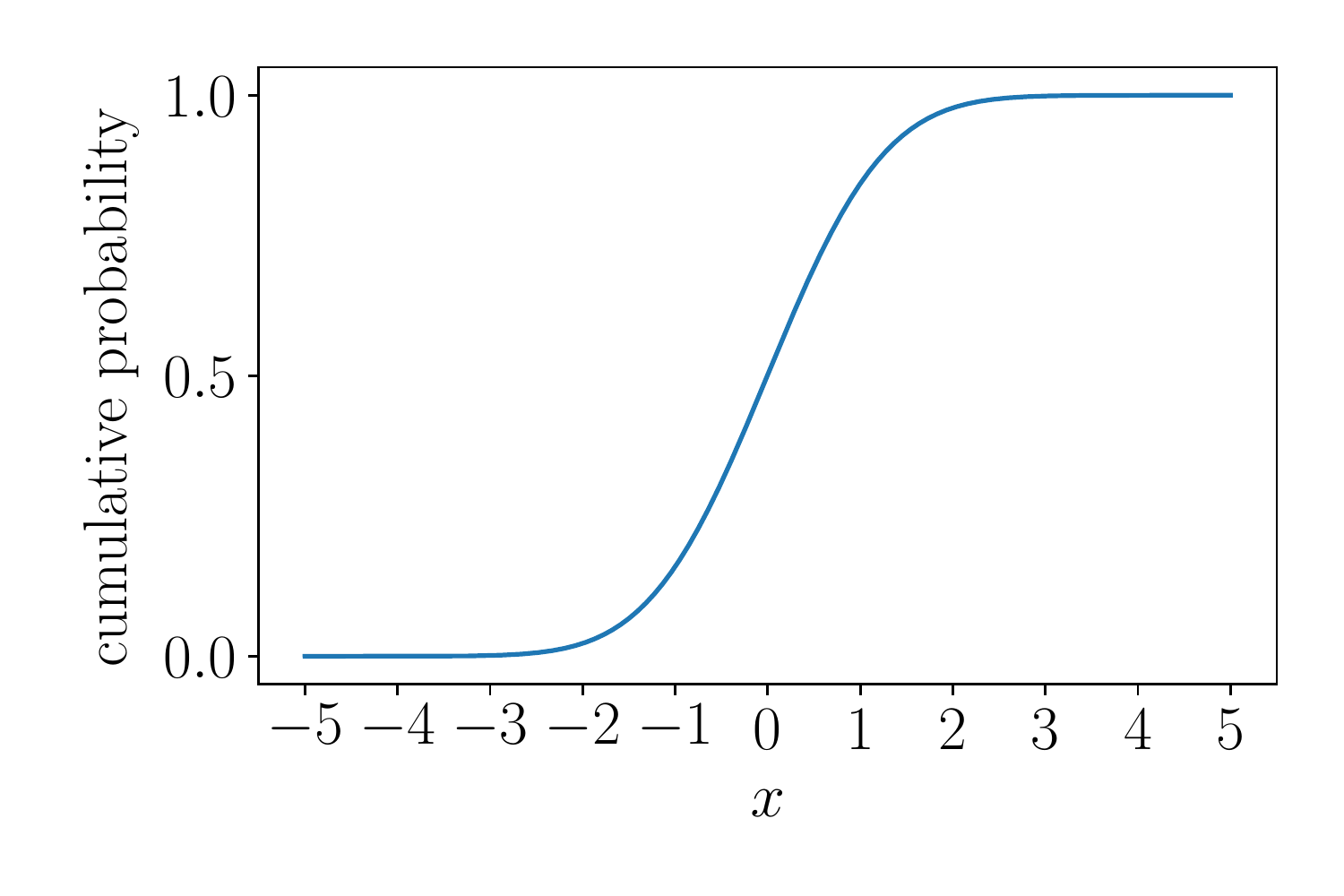}
\end{subfigure}
\centering
\begin{subfigure}
  \centering
  \includegraphics[width=0.49\textwidth]{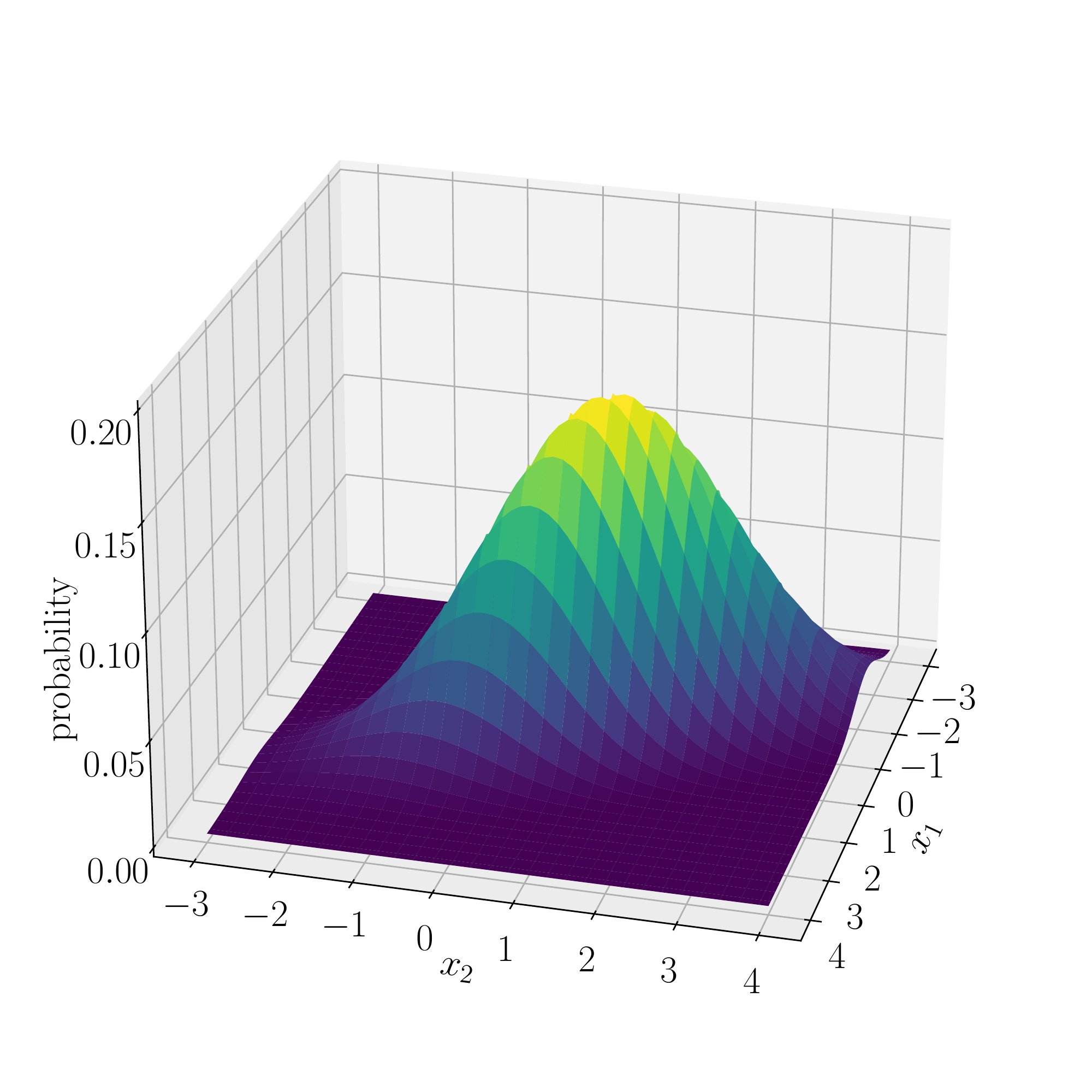}
\end{subfigure}
\caption{We depict the probability density function (left) and cumulative distribution function (right) of a univariate Gaussian distribution with mean 0 and variance 1. In addition, we plot the \pdf{} of a bivariate Gaussian with $\bm{\mu}=[0,1]$ and $\bm{\Sigma}=
\begin{pmatrix}
1 & -1 \\
-1 & 2
\end{pmatrix}$.}
\label{fig:univariate-gaussian}
\end{figure}

\paragraph*{Gaussian Distribution.}
The Gaussian distribution is a continuous probability distribution whose support is $\R{}$. For a single random variable (\textbf{univariate} case), the parameters of the distribution are just the mean value $\mu$ and the variance $\sigma^2$, and the probability density function is defined as
\begin{align*}
P(x\mid \mu, \sigma^2) = \frac{1}{\sigma\sqrt{2\pi}} e^{-\frac{1}{2}(\frac{x-\mu}{\sigma})^2}.
\end{align*}Generalizing the distribution to $n$ random variables (\textbf{multivariate} case), the parameters that define the distribution are a vector of $n$ means $\bm{\mu}$ and an $n\times n$ covariance matrix $\bm{\Sigma}$. The \pmf{} of a multivariate distribution then becomes a function that takes an input $\boldx{} \in \R{n}$ and returns a probability score:
\begin{align*}
P(\boldx{}\mid \bm{\mu}, \bm{\Sigma}) = \frac{1}{\sqrt{(2\pi)^n det(\bm{\Sigma}})} e^{-\frac{1}{2}(\boldx{}-\bm{\mu})^T \bm{\Sigma}^{-1} (\boldx{}-\bm{\mu})},
\end{align*}
where $det(\cdot)$ is the determinant of a matrix. A multivariate Gaussian distribution is said to be \textbf{isotropic} when the covariance matrix is diagonal, meaning the random variables under consideration are independent. In this case, the total number of parameters becomes $2n$ rather than $n + n^2$. Figure \ref{fig:univariate-gaussian} visualizes instances of the univariate and bivariate Gaussian distributions.

\begin{figure}[ht]
\begin{subfigure}
  \centering
  \includegraphics[width=0.49\textwidth]{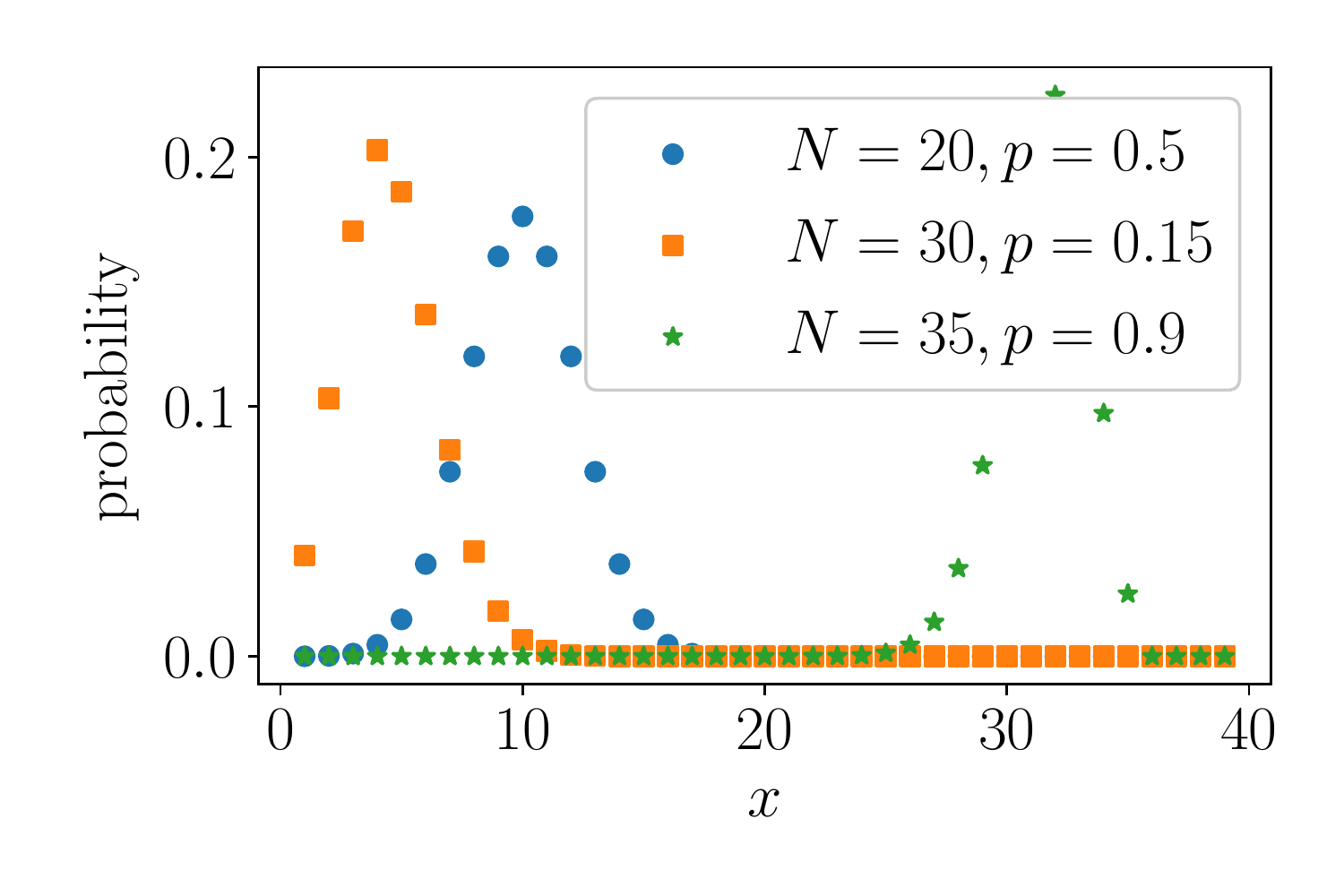}
\end{subfigure}%
\begin{subfigure}
  \centering
  \includegraphics[width=0.49\textwidth]{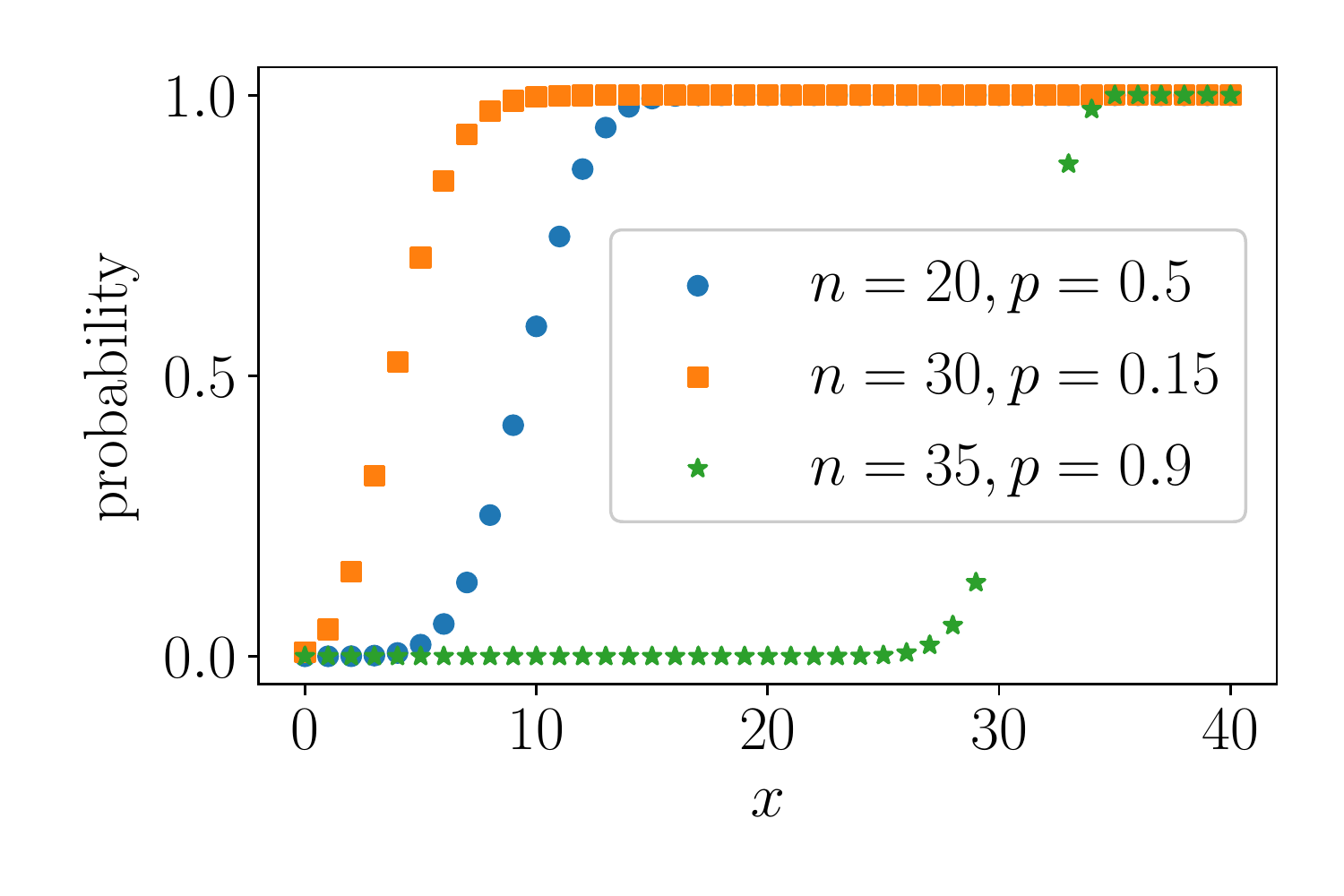}
\end{subfigure}
\caption{We plot different binomial distributions (left) and their \cdf{} (right) for different choices of $n$ and $p$.}
\label{fig:binomial}
\end{figure}

\paragraph*{Binomial Distribution.} The binomial distribution is another discrete probability distribution that accounts for the number of successes in a sequence of $n$ experiments, each of which has a probability $p$ of success and $1-p$ of failure. Its support is the set $\{0,\dots,n\}$ and the \pmf{} is computed as
\begin{align*}
& P(x\mid n,p) = \binom{n}{x} p^x(1-p)^{n-x} \\
& \text{where} \ \ \binom{n}{x} = \frac{n!}{k!(n-k)!}.
\end{align*}Usually, the single experiment is called a Bernoulli trial, whereas the entire sequence of outcomes is a Bernoulli process. 
The reference to Bernoulli comes from the fact that, when $n=1$, the distribution simplifies to a Bernoulli distribution (not shown here). We conclude by showing the \pmf{} and \cdf{} of different binomial distributions in Figure \ref{fig:binomial}.

\begin{figure}[ht]
\begin{subfigure}
  \centering
  \includegraphics[width=0.49\textwidth]{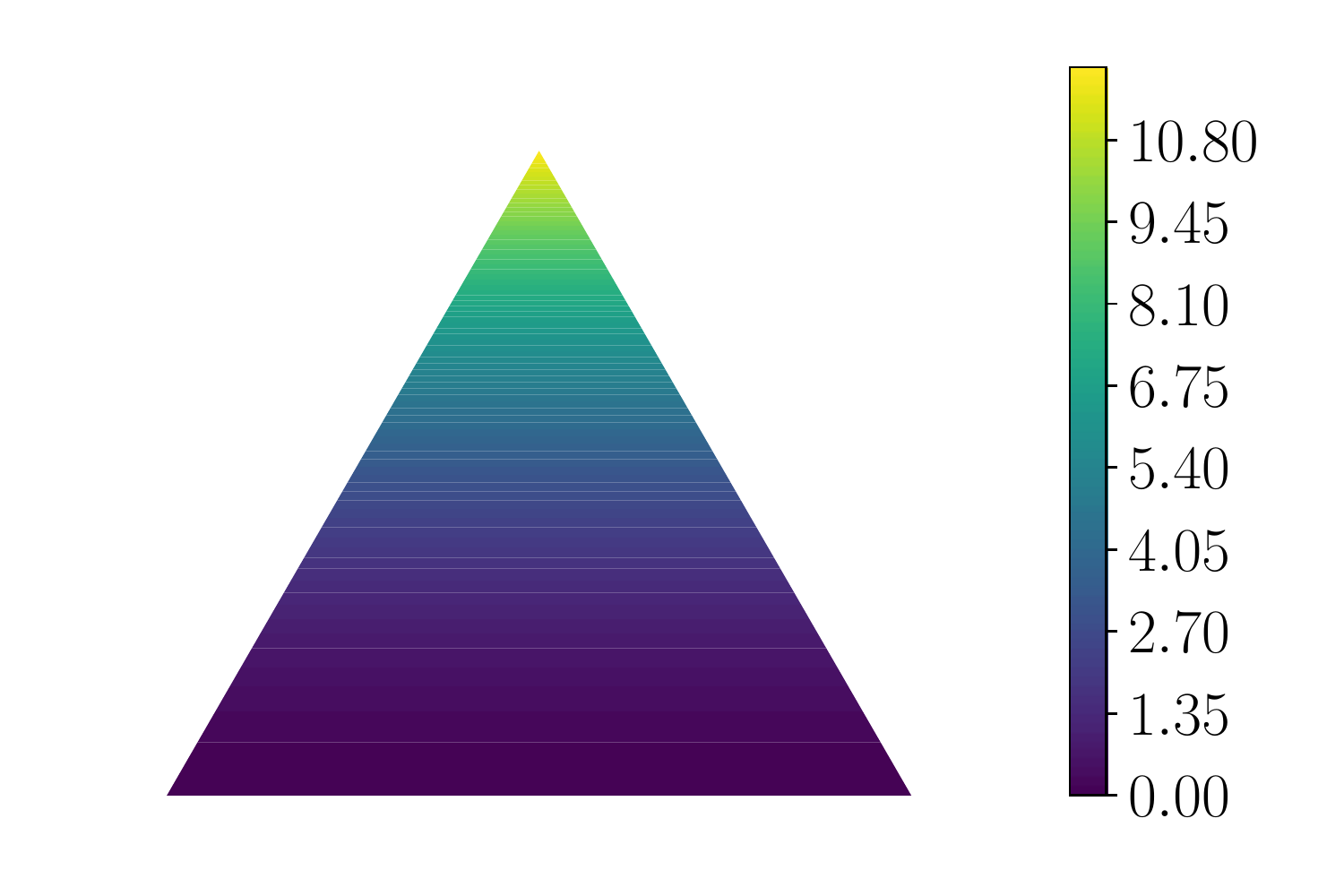}
\end{subfigure}%
\begin{subfigure}
  \centering
  \includegraphics[width=0.49\textwidth]{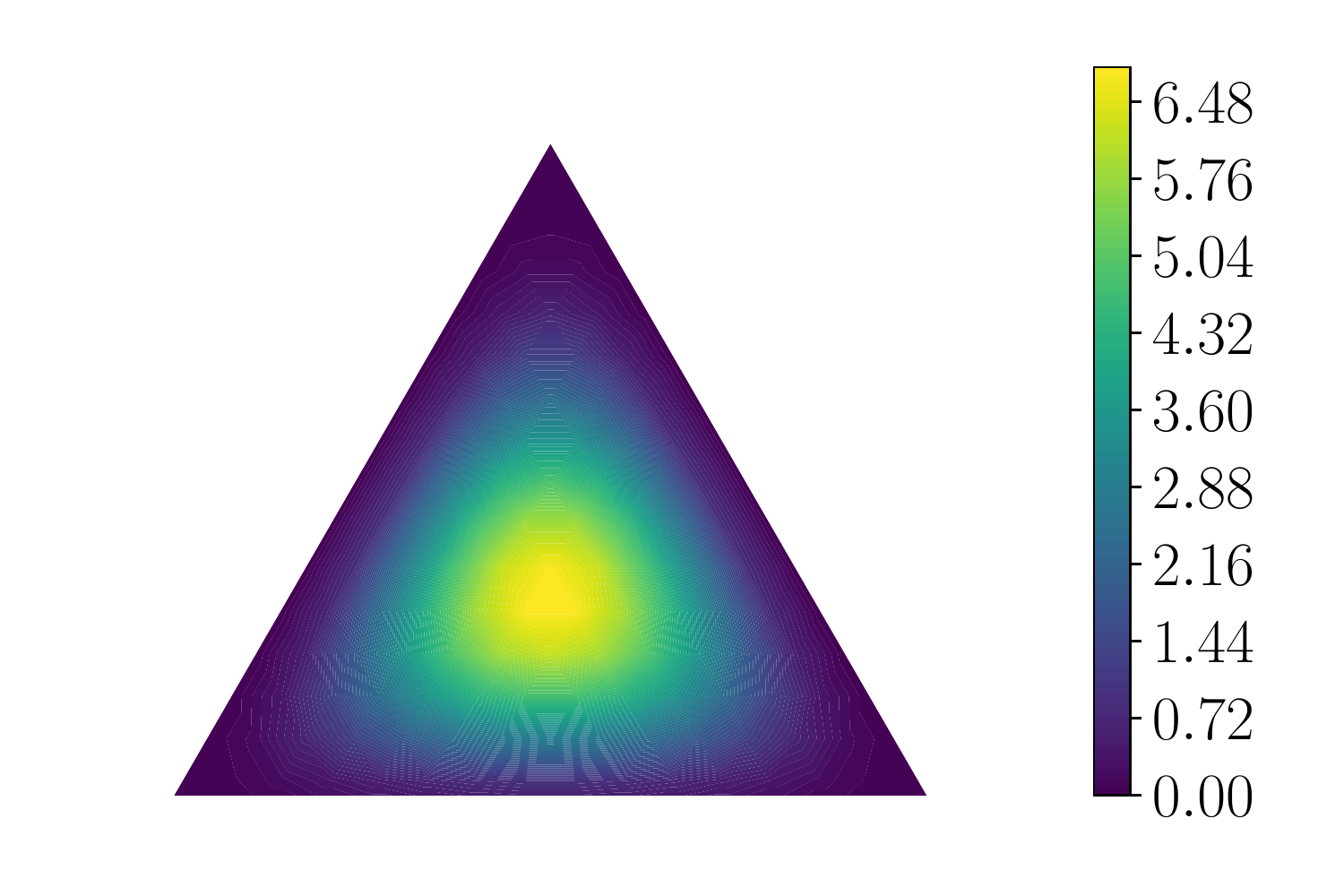}
\end{subfigure}
\caption{Two Dirichlet distributions with $\bm{\alpha}=[1,1,3]$ (left) and $\bm{\alpha}=[3,3,3]$ (right). Each side of the triangles denotes a different input component $x_i \in [0,1]$.}
\label{fig:dirichlet}
\end{figure}

\paragraph*{Dirichlet Distribution.} The Dirichlet distribution is a multivariate continuous distribution parametrized by a fixed vector $\bm{\alpha}$ of $C$ real values called concentration parameters. It can be seen as a distribution over distributions, because the sampling process outputs a vector $\bm{p}$, belonging to the standard $C-1$ simplex, which might be used to parametrize a categorical distribution. The \pdf{} of the Dirichlet distribution $D(\bm{\alpha})$ is defined as
\begin{align*}
    & P(\boldx{}|\bm{\alpha}) =\frac{1}{B(\bm{\alpha})}\prod_{i=1}^{C}x_i^{\alpha_i - 1} \\
    & \text{where} \ \ B(\bm{\alpha}) = \frac{\prod_{i=1}^C \Gamma(\alpha_i)}{\Gamma(\sum_{i=1}^C\alpha_i)}, \Gamma(\alpha_i) = (\alpha_i -1)!.
\end{align*}
The Dirichlet distribution is particularly important in Bayesian statistics, being a \textbf{conjugate} distribution for discrete distributions such as the categorical. If a distribution $P$ is the conjugate of a distribution $Q$, it means that multiplying the \pdf{} of $P$ and $Q$ will result in a distribution whose \pdf{} belongs to the same family of distributions as $P$. This greatly simplifies the math and provides closed-form solutions in Bayesian learning, where the Dirichlet distribution is used to embed \textbf{prior knowledge} in the learning framework. Finally, we depict two examples of Dirichlet distributions in Figure \ref{fig:dirichlet}.

\paragraph*{Normal-Gamma Distribution.} The last distribution we introduce is the Normal-Gamma distribution, a continuous probability distribution that is conjugate of the univariate Gaussian distribution. It has four parameters that act as prior knowledge: $\mu$, the empirical mean of the data; $\lambda$, which is inversely proportional to the prior variance of the data (\ie our belief about the robustness of the empirical mean); $a$ and $b$, whose ratio $t=\frac{b}{a}$ controls the expected variance of the data captured by the univariate Gaussian. Assuming a Gaussian random variable $X$ and a Gamma random variable $T$, the \pdf{} of a Normal-Gamma distribution is computed as
\begin{align*}
    & X \sim \mathcal{N}(\mu, 1/(\lambda T))\\
    & T \sim \text{Gamma}(a, b) \ \text{whose \pdf{} is } f(t\mid a,b) = \frac{b^{a}t^{a-1}e^{-bt}}{\Gamma(a)}, \Gamma(a)=(a-1)! \\
    & P(x,t|\mu,\lambda,a,b) = \frac{b^a\sqrt{\lambda}}{\Gamma(a)\sqrt{2\pi}}t^{a-\frac{1}{2}}e^{-bt}\exp(-\frac{\lambda t(x-\mu)^2}{2}).
\end{align*}

\subsubsection{Learning as an Inference Problem}
The goal of a generic \ML{} task is to find a suitable choice of the model's parameters in order to optimize a pre-defined objective function. Likewise, in a probabilistic setting where we need to capture the underlying distribution of the data, one chooses a family of parametrized distributions assuming it is flexible enough to mimic the true (unknown) data distribution. Therefore, the learning task can be framed as an \textbf{inference} problem in which we adjust our beliefs about the parameters of the selected family of distributions, \eg a Gaussian or a Categorical. Before we outline the basics of the probabilistic learning framework adopted in this thesis, we shall introduce the Bayes' Rule, which lies down the foundations of \textbf{Bayesian learning}.
\begin{definition}[\textbf{Bayes' Rule}]
Given a discrete hypotheses space $\hypotheses$ and a set of observations $\dataset{}$ (both of which can be modeled as random processes), for each hypothesis $h_i \in \hypotheses$ it holds
\begin{align*}
& P(h_i \mid \dataset{}) = \frac{P(\dataset{} \mid h_i)P(h_i)}{P(\dataset{})} = \frac{P(\dataset{} \mid h_i)P(h_i)}{\sum_j P(\dataset{} \mid h_j)P(h_j)}
\end{align*}where $P(h)$ is called the \textbf{prior} probability of $h$, $P(\dataset{} \mid h)$ is the \textbf{likelihood} that the data has been generated by a certain hypothesis (with corresponding parameters  $\bm{\theta}$), and $P(h\mid \dataset{})$ is the posterior probability that the hypothesis is correct given the data and out prior beliefs about that hypothesis. Therefore, under the Bayes' rule there is a trade-off between our prior beliefs and the evidence coming from the data.
\end{definition}
Given a set of random variables $\bm{X}=\{X_1,\dots,X_n\}$ called \textbf{observations} that formalize our knowledge about \quotes{the world}, the prediction for a new data point in a Bayesian learning setting can be written as
\begin{align*}
P(\bm{X}=\bm{x} \mid \dataset{}) = \sum_i P(\bm{X}=\bm{x} \mid \dataset{}, h_i)P(h_i \mid \dataset{}) = \sum_i \underbrace{P(\bm{X}=\bm{x} \mid h_i)}_{\substack{\text{hypothesis} \\ \text{prediction}}}\underbrace{P(h_i \mid \dataset{})}_{\substack{\text{posterior} \\ \text{weighting}}},
\end{align*}
where we used the product rule, marginalization, and assumed that $\bm{X}$ does not depend on the data when hypothesis $h_i$ holds. Unfortunately, very often the space of the hypotheses is infinite, and the exact computation of the posterior distribution of $\bm{X}$ becomes intractable. To address this practical issue, we can look for the single most likely hypothesis given our data. This is called Maximum A Posteriori (MAP) inference:
\begin{align*}
& h_{MAP} = \argmax_{h \in \hypotheses} P(h \mid \dataset{})  \overset{{\substack{\text{Bayes'} \\ \text{Rule} \\}}}{=}  \argmax_{h \in \hypotheses} \frac{P(\dataset{} \mid h)P(h)}{P(\dataset{})} = \argmax_{h \in \hypotheses} P(\dataset{} \mid h)P(h),
\end{align*}
noting that we ignored the contribution of $P(\dataset{})$ in the maximization because constant. When assuming a \textbf{uniform} prior distribution over the choice of $\hypotheses$, meaning $P(h)$ is the same everywhere, we obtain the \textbf{Maximum Likelihood Estimation} (MLE) objective:
\begin{align*}
& h_{MLE} = \argmax_{h \in \hypotheses} P(\dataset{} \mid h) = \argmax_{h \in \hypotheses} \likelihood(\bm{\theta}_h \mid \dataset{}).
\end{align*}
Most of the techniques presented in this thesis will be based on the MLE objective, whereas a restricted number of them will adopt MAP inference.

\subsubsection{Bayesian Networks}
\begin{figure}[ht]
\centering
\resizebox{0.4\textwidth}{!}{\tikzset{every picture/.style={line width=0.75pt}} 

\begin{tikzpicture}[x=0.75pt,y=0.75pt,yscale=-1,xscale=1]

\draw [line width=1.5]    (170,106.23) -- (142.43,178.5) ;
\draw [shift={(141,182.23)}, rotate = 290.89] [fill={rgb, 255:red, 0; green, 0; blue, 0 }  ][line width=0.08]  [draw opacity=0] (11.61,-5.58) -- (0,0) -- (11.61,5.58) -- cycle    ;
\draw [line width=1.5]    (170,106.23) -- (259.41,179.69) ;
\draw [shift={(262.5,182.23)}, rotate = 219.41] [fill={rgb, 255:red, 0; green, 0; blue, 0 }  ][line width=0.08]  [draw opacity=0] (11.61,-5.58) -- (0,0) -- (11.61,5.58) -- cycle    ;
\draw [line width=1.5]    (170,106.23) -- (81.08,180.18) ;
\draw [shift={(78,182.73)}, rotate = 320.26] [fill={rgb, 255:red, 0; green, 0; blue, 0 }  ][line width=0.08]  [draw opacity=0] (11.61,-5.58) -- (0,0) -- (11.61,5.58) -- cycle    ;
\draw  [fill={rgb, 255:red, 31; green, 119; blue, 180 }  ,fill opacity=1 ][line width=1.5]  (53,207.73) .. controls (53,193.93) and (64.19,182.73) .. (78,182.73) .. controls (91.81,182.73) and (103,193.93) .. (103,207.73) .. controls (103,221.54) and (91.81,232.73) .. (78,232.73) .. controls (64.19,232.73) and (53,221.54) .. (53,207.73) -- cycle ;
\draw  [fill={rgb, 255:red, 255; green, 255; blue, 255 }  ,fill opacity=1 ][line width=1.5]  (145,106.23) .. controls (145,92.43) and (156.19,81.23) .. (170,81.23) .. controls (183.81,81.23) and (195,92.43) .. (195,106.23) .. controls (195,120.04) and (183.81,131.23) .. (170,131.23) .. controls (156.19,131.23) and (145,120.04) .. (145,106.23) -- cycle ;
\draw  [fill={rgb, 255:red, 31; green, 119; blue, 180 }  ,fill opacity=1 ][line width=1.5]  (116,207.23) .. controls (116,193.43) and (127.19,182.23) .. (141,182.23) .. controls (154.81,182.23) and (166,193.43) .. (166,207.23) .. controls (166,221.04) and (154.81,232.23) .. (141,232.23) .. controls (127.19,232.23) and (116,221.04) .. (116,207.23) -- cycle ;
\draw  [fill={rgb, 255:red, 31; green, 119; blue, 180 }  ,fill opacity=1 ][line width=1.5]  (237.5,207.23) .. controls (237.5,193.43) and (248.69,182.23) .. (262.5,182.23) .. controls (276.31,182.23) and (287.5,193.43) .. (287.5,207.23) .. controls (287.5,221.04) and (276.31,232.23) .. (262.5,232.23) .. controls (248.69,232.23) and (237.5,221.04) .. (237.5,207.23) -- cycle ;

\draw (170,106.23) node  [font=\Large]  {$Q$};
\draw (78,207.73) node  [font=\Large,color={rgb, 255:red, 255; green, 255; blue, 255 }  ,opacity=1 ]  {$X_{1}$};
\draw (141,207.23) node  [font=\Large,color={rgb, 255:red, 255; green, 255; blue, 255 }  ,opacity=1 ]  {$X_{2}$};
\draw (262.5,207.23) node  [font=\Large,color={rgb, 255:red, 255; green, 255; blue, 255 }  ,opacity=1 ]  {$X_{n}$};
\draw (202.14,207.9) node  [font=\huge] [align=left] {$\displaystyle \dotsc $};

\end{tikzpicture}}
\caption{A Bayesian network with latent (white) and observed (blue) variables.}
\label{fig:bayesian-network}
\end{figure}
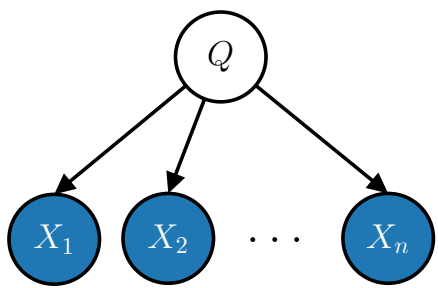

When modeling the world with a set of random variables, it is natural to make assumptions about the 
relationships between those. One way to graphically express the conditional dependencies between such variables is to use a Bayesian network. In this graphical representation, an istance of which is illustrated in Figure \ref{fig:bayesian-network}, nodes represent variables and edges convey the conditional independence information. We distinguish between \textbf{observed} variables that can always be inspected (\ie the observations $\dataset{}$ contain information about the values of these variables)  and \textbf{latent} or \textbf{hidden} variables whose values have to be inferred from the data. By definition, a Bayesian network allows us to decompose the joint probability distribution of all variables following a straightforward rule:
\begin{align*}
    P(X_1\dots,X_n) = \prod_i^n P(X_i\mid pa(X_i)),
\end{align*}where the term $pa(X_i)$ refers to the set of nodes that have an edge pointing to $X_i$ (the \quotes{parents} of $X_i$). Note that in a Bayesian network a variable cannot depend, directly or indirectly, on itself. In turn, this simplifies the math and enables tractable solutions in many cases.

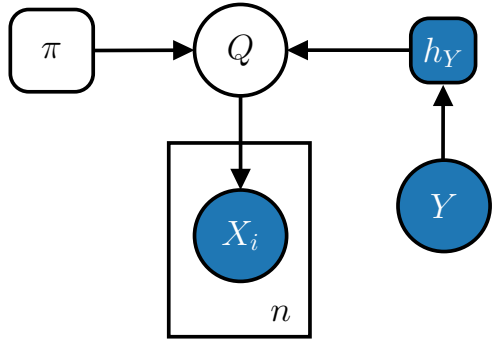
\begin{figure}[ht]
\centering
\resizebox{0.45\textwidth}{!}{\tikzset{every picture/.style={line width=0.75pt}} 

\begin{tikzpicture}[x=0.75pt,y=0.75pt,yscale=-1,xscale=1]

\draw [line width=1.5]    (146,93.23) -- (146,165.73) ;
\draw [shift={(146,169.73)}, rotate = 270] [fill={rgb, 255:red, 0; green, 0; blue, 0 }  ][line width=0.08]  [draw opacity=0] (11.61,-5.58) -- (0,0) -- (11.61,5.58) -- cycle    ;
\draw [line width=1.5]    (43.5,93.5) -- (117,93.25) ;
\draw [shift={(121,93.23)}, rotate = 539.8] [fill={rgb, 255:red, 0; green, 0; blue, 0 }  ][line width=0.08]  [draw opacity=0] (11.61,-5.58) -- (0,0) -- (11.61,5.58) -- cycle    ;
\draw  [fill={rgb, 255:red, 31; green, 119; blue, 180 }  ,fill opacity=1 ][line width=1.5]  (121,194.73) .. controls (121,180.93) and (132.19,169.73) .. (146,169.73) .. controls (159.81,169.73) and (171,180.93) .. (171,194.73) .. controls (171,208.54) and (159.81,219.73) .. (146,219.73) .. controls (132.19,219.73) and (121,208.54) .. (121,194.73) -- cycle ;
\draw  [fill={rgb, 255:red, 255; green, 255; blue, 255 }  ,fill opacity=1 ][line width=1.5]  (21,81) .. controls (21,75.48) and (25.48,71) .. (31,71) -- (56,71) .. controls (61.52,71) and (66,75.48) .. (66,81) -- (66,106) .. controls (66,111.52) and (61.52,116) .. (56,116) -- (31,116) .. controls (25.48,116) and (21,111.52) .. (21,106) -- cycle ;
\draw  [fill={rgb, 255:red, 255; green, 255; blue, 255 }  ,fill opacity=1 ][line width=1.5]  (121,93.23) .. controls (121,79.43) and (132.19,68.23) .. (146,68.23) .. controls (159.81,68.23) and (171,79.43) .. (171,93.23) .. controls (171,107.04) and (159.81,118.23) .. (146,118.23) .. controls (132.19,118.23) and (121,107.04) .. (121,93.23) -- cycle ;
\draw  [line width=1.5]  (107,144) -- (183.67,144) -- (183.67,250) -- (107,250) -- cycle ;
\draw [line width=1.5]    (256.67,93.23) -- (175,93.23) ;
\draw [shift={(171,93.23)}, rotate = 360] [fill={rgb, 255:red, 0; green, 0; blue, 0 }  ][line width=0.08]  [draw opacity=0] (11.61,-5.58) -- (0,0) -- (11.61,5.58) -- cycle    ;
\draw [line width=1.5]    (255.87,179.2) -- (255.99,116.2) ;
\draw [shift={(256,112.2)}, rotate = 450.11] [fill={rgb, 255:red, 0; green, 0; blue, 0 }  ][line width=0.08]  [draw opacity=0] (11.61,-5.58) -- (0,0) -- (11.61,5.58) -- cycle    ;
\draw  [fill={rgb, 255:red, 31; green, 119; blue, 180 }  ,fill opacity=1 ][line width=1.5]  (231.47,178.4) .. controls (231.47,164.59) and (242.66,153.4) .. (256.47,153.4) .. controls (270.27,153.4) and (281.47,164.59) .. (281.47,178.4) .. controls (281.47,192.21) and (270.27,203.4) .. (256.47,203.4) .. controls (242.66,203.4) and (231.47,192.21) .. (231.47,178.4) -- cycle ;
\draw  [fill={rgb, 255:red, 31; green, 119; blue, 180 }  ,fill opacity=1 ][line width=1.5]  (238.42,85.82) .. controls (238.42,80.29) and (242.89,75.82) .. (248.42,75.82) -- (263.58,75.82) .. controls (269.11,75.82) and (273.58,80.29) .. (273.58,85.82) -- (273.58,100.98) .. controls (273.58,106.51) and (269.11,110.98) .. (263.58,110.98) -- (248.42,110.98) .. controls (242.89,110.98) and (238.42,106.51) .. (238.42,100.98) -- cycle ;

\draw (146,93.23) node  [font=\Large]  {$Q$};
\draw (43.5,93.5) node  [font=\Large] [align=left] {$\displaystyle \pi $};
\draw (146,194.73) node  [font=\Large,color={rgb, 255:red, 255; green, 255; blue, 255 }  ,opacity=1 ]  {$X_{i}$};
\draw (256.47,178.4) node  [font=\Large,color={rgb, 255:red, 255; green, 255; blue, 255 }  ,opacity=1 ]  {$Y$};
\draw (256,94.4) node  [font=\Large,color={rgb, 255:red, 255; green, 255; blue, 255 }  ,opacity=1 ]  {$h_{Y}$};
\draw (167.84,237.35) node  [font=\Large] [align=left] {$\displaystyle n$};

\end{tikzpicture}}
\caption{Plate notation for the Bayesian network of Figure \ref{fig:bayesian-network}, with the addition of prior hyper-parameters $\pi$ and an intermediate node $h_Y$ obtained deterministically from the observed value of $Y$.}
\label{fig:plate-notation}
\end{figure}

To graphically represent a set of random variables in a compact way, we adopt the so-called \textbf{plate notation} of Figure \ref{fig:plate-notation}, where a plate (the box) symbolises replication of conditional relations and encompassed variables for a number of times indicated by the letter in the corner. Still, when clear from the context, we may drop the box and simply use a subscript to denote the identity of the variables. With slight abuse of notation, we also use white squares to represent hyper-parameters and blue squares for intermediate deterministic computations.

\subsubsection{The Expectation-Maximization Algorithm}
\label{subsec:em}
A widely adopted tool to train a probabilistic model with latent variables $\bm{Z}$ is the Expectation-Maximization (EM) algorithm \cite{dempster_maximum_1977}. The key insight is that, rather than maximizing the \quotes{incomplete} likelihood of the observed data $P(\bm{X}\mid \bm{\theta})$, we focus on the \textbf{complete likelihood} of the data $P(\bm{X},\bm{Z}\mid \bm{\theta})$. It is a two-step iterative procedure that maximizes the likelihood of the data, in which the parameters of the model at time step $t$, namely $\bm{\theta}^{(t)}$, are updated (M-step) using the \textbf{current estimate} for the values of the hidden variables $\bm{Z}$ (E-step). This resulting objective is simpler to maximize, since it does not involve marginalization over all variables in $\bm{Z}$. 
Formally, the EM algorithm involves the following steps:
\begin{enumerate}
\item initialize the parameters $\bm{\theta}^{(1)}$ at random
\item (E-step) compute the expected value of the complete log-likelihood w.r.t. $\bm{\theta}^{(t)}$
\begin{align*}
Q_{EM}(\bm{\theta}\mid \bm{\theta}^{(t)}) = \E{\bm{Z}\mid \bm{X},\bm{\theta}^{(t)}}{\log P(\bm{X},\bm{Z}\mid \bm{\theta}^{(t)})}
\end{align*}
\item (M-step) find the parameters that maximize the previous quantity
\begin{align*}
\bm{\theta}^{(t+1)} = \argmax_{\bm{\theta}}Q_{EM}(\bm{\theta}\mid \bm{\theta}^{(t)})
\end{align*}
\item Repeat steps 2 and 3 until the complete log-likelihood stops increasing.
\end{enumerate}
It can be shown that the quantity computed by the E-step is a lower-bound of the true likelihood \cite{dempster_maximum_1977}. Therefore, the EM algorithm can converge monotonically to a local minimum of the initial objective. In some cases, it is possible to compute the M-step solutions in closed-form, this obtaining the maximum improvement over $Q_{EM}$. Whenever this is not possible, common optimization algorithms may be used such as Stochastic Gradient Descent (SGD) \cite{robbins_stochastic_1951}; as long as the value of the lower-bound increases, convergence is still guaranteed. In this case, we talk about \textbf{Generalized Expectation Maximization} (GEM) \cite{dempster_maximum_1977}. Finally, note that EM can be easily extended to deal with MAP estimation, by adding the contribution of the log-prior to the quantity $E(\bm{\theta}\mid \bm{\theta}^{(t)})$ to maximize.

Knowledge about the values of hidden variables can be modeled via \textbf{indicator random variables}. An indicator variable is a binary variable that is 1 when, \eg another random variable is in a certain state, and 0 otherwise. For instance, given a categorical variable $Q$ with $C$ states, an indicator variable $Z_c$ may be described as follows:
\begin{align*}
    Z_c=
    \begin{cases}
      1, & \text{if}\ Q=c \ \text{with probability } p_c \\
      0, & \text{otherwise} \ \text{with probability } (1-p_c) \\
    \end{cases}
   \ \ \ \ \text{where } p_c = P(Q=c).
\end{align*}
As we will see, indicator variables will be invaluable to define the complete log-likelihood of mixture models.

\subsubsection{Gibbs Sampling}
Whenever we shall encounter a joint probability distribution that is difficult to formalize or sample from but the conditional probability of the individual variables is easier to compute, we will depend on Markov Chain Monte Carlo (MCMC) algorithms. An MCMC algorithm that is often used in Bayesian inference is \textbf{Gibbs sampling} \cite{geman_stochastic_1984}, which works by sampling the value of a single variable at a time before moving to the next one. By iterating this process over and over, eventually the sampled values should approximate the original joint distribution. Also, when some of the variables are observed, their values are never updated. Given $n$ random variables $\bm{X}=\{X_1,\dots,X_n\}$ from a joint probability distribution $P(X_1=x_1,\dots,X_n=x_n$), we can obtain $k$ samples from $\bm{X}$ using the Gibbs sampler:
\begin{enumerate}
    \item Initialize the sample as $\bm{X}^{(0)} = (x_1^{(0)},\dots,x_n^{(0)})$
    \item When creating the $i+1$-th sample, for each component $j$ (in order), update its value by sampling from the known conditional probability using the following formula
    \begin{align*}
        x_j^{(t+1)} \sim P(x_j^{(t+1)}|x_1^{(t+1)},\dots,x_{j-1}^{(t+1)},x_{j+1}^{(t)},\dots,x_n^{(t)})
    \end{align*}
    \item Repeat the previous step $k$ times in order to obtain $k$ distinct samples.
\end{enumerate}
In general, we assume there is a \textbf{burn-in period} of variable length in which samples do not accurately represent the joint distribution, so they are discarded.
With Gibbs sampling, we are able to infer the posterior distribution of the model's parameters conditioned on the data. Still, to do so, we need to assume some family of distributions for the different conditional probabilities; in this sense, a Bayesian network is an excellent companion for Gibbs sampling, as its joint distribution is specified by means of a set of conditional distributions.

\subsection{Mixture Models}
\label{subsec:mixture-model-basic}
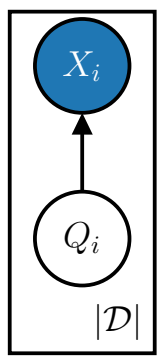
\begin{figure}[ht]
\centering
\resizebox{0.15\textwidth}{!}{\tikzset{every picture/.style={line width=0.75pt}} 

\begin{tikzpicture}[x=0.75pt,y=0.75pt,yscale=-1,xscale=1]

\draw [line width=1.5]    (189,190.23) -- (189,127.23) ;
\draw [shift={(189,123.23)}, rotate = 450] [fill={rgb, 255:red, 0; green, 0; blue, 0 }  ][line width=0.08]  [draw opacity=0] (11.61,-5.58) -- (0,0) -- (11.61,5.58) -- cycle    ;
\draw  [fill={rgb, 255:red, 255; green, 255; blue, 255 }  ,fill opacity=1 ][line width=1.5]  (164,190.23) .. controls (164,176.43) and (175.19,165.23) .. (189,165.23) .. controls (202.81,165.23) and (214,176.43) .. (214,190.23) .. controls (214,204.04) and (202.81,215.23) .. (189,215.23) .. controls (175.19,215.23) and (164,204.04) .. (164,190.23) -- cycle ;
\draw  [fill={rgb, 255:red, 31; green, 119; blue, 180 }  ,fill opacity=1 ][line width=1.5]  (164,98.23) .. controls (164,84.43) and (175.19,73.23) .. (189,73.23) .. controls (202.81,73.23) and (214,84.43) .. (214,98.23) .. controls (214,112.04) and (202.81,123.23) .. (189,123.23) .. controls (175.19,123.23) and (164,112.04) .. (164,98.23) -- cycle ;
\draw  [line width=1.5]  (150.67,69.33) -- (227.33,69.33) -- (227.33,251.33) -- (150.67,251.33) -- cycle ;

\draw (189,190.23) node  [font=\Large]  {$Q_{i}$};
\draw (189,98.23) node  [font=\Large,color={rgb, 255:red, 255; green, 255; blue, 255 }  ,opacity=1 ]  {$X_{i}$};
\draw (207.84,234.67) node  [font=\Large] [align=left] {$\displaystyle |\mathcal{D} |$};

\end{tikzpicture}}
\caption{The probabilistic graphical representation for a mixture model. The latent variable $Q$ assigns each data point to one of $C$ different sub-populations (or \textbf{clusters}).}
\label{fig:mixture-model-graphical}
\end{figure}

We shall now introduce the concept of mixture of distributions, which will be central to this thesis thanks to its simplicity and flexibility. Let us assume we have a population of samples, \ie our data, in which there exist $C$ sub-populations we would like to model. In other words, we would like to associate each data point with one of the $C$ sub-populations; this process is usually referred to as \textbf{clustering}, and we can represent it with the graphical model of Figure \ref{fig:mixture-model-graphical}.
In particular, we assume that there is a latent 
factor, represented by the categorical variable $Q$ with $C$ states, that is responsible for the generation of the data points. Knowledge of the probability distributions of $P(Q)$ and $P(X\mid Q)$ would allow us to generate new samples by first sampling the cluster $c$ from $P(Q)$ and then drawing the data point $x$ from $P(X|Q=c)$. This is a mechanism known as \textbf{ancestral sampling}.

Formally, we can model the distribution of our population by introducing the latent variables via marginalization, assuming our samples are independent and identically distributed:
\begin{align*}
    \likelihood(\bm{\theta}\mid \dataset{}) = P(\dataset{} \mid \bm{\theta}) = \prod_{i=1}^{|\dataset{}|}P(X_i = x_i) = \prod_{i=1}^{|\dataset{}|}\sum_{j=1}^C P(X_i = x_i \mid Q_i=j,\bm{\theta})P(Q_i=j \mid \bm{\theta}),
\end{align*}
where $\bm{\theta}$ are the parameters of the mixture model that we have to learn. From now on, we will refer to $P(X\mid Q)$ as the \textbf{emission distribution}. 

Usually, we have no guarantees that the data will be organized in $C$ different sub-populations. At the same time, the larger $C$ is the more flexible our approximation will be. Hence, the parameter $C$, which controls the number of distributions to mix, has to be therefore treated as an hyper-parameter.

To train a mixture model that maximizes the likelihood of our data, let us adopt the EM framework and define the \textit{complete log-likelihood} of the data. The usual way to do this is to first assume we know the assignment of each data point to its cluster. Therefore, let us introduce the indicator variables $\bm{Z}=\{Z_{11},\dots,Z_{|\dataset{}|C}\}$ where a generic variable $Z_{ij}$ has value 1 when the data point $i$ is assigned to cluster $j$. We can take advantage of $\bm{Z}$ to write the complete log-likelihood
\begin{align*}
    \likelihood_c(\bm{\theta}\mid \bm{Z},\dataset{}) & = \prod_{i=1}^{|\dataset{}|}\sum_{j=1}^C z_{ij} P(X_i = x_i \mid Q_i=j,\bm{\theta})P(Q_i=j \mid \bm{\theta}) \\
    & = \prod_{i=1}^{|\dataset{}|}\prod_{j=1}^C \big(P(X_i = x_i \mid Q_i=j,\bm{\theta})P(Q_i=j \mid \bm{\theta})\big)^{z_{ij}}.
\end{align*}
It is possible to obtain the last identity by noting that $z_{ij}$ nullifies the contributions of other clusters. To compute the E-step, we now need to apply the logarithm to the above quantity, obtaining
\begin{align*}
    \log \likelihood_c(\bm{\theta}\mid \bm{Z},\dataset{}) = \sum_{i=1}^{|\dataset{}|}\sum_{j=1}^C z_{ij}\log\big(P(X_i = x_i \mid Q_i=j,\bm{\theta})P(Q_i=j \mid \bm{\theta})\big).
\end{align*}
Assuming we have some parameters $\bm{\theta}^{(t)}$, the quantity to maximize at each EM iteration can be computed as
\begin{align*}
    & Q_{EM}(\bm{\theta}\mid \bm{\theta}^{(t)}) = \E{\bm{Z}\mid \dataset{},\bm{\theta}^{(t)}}{\log P(\dataset{},\bm{Z}\mid \bm{\theta}^{(t)})} = \\
    & = \sum_{i=1}^{|\dataset{}|}\sum_{j=1}^C \E{\bm{Z}\mid \bm{X},\bm{\theta}^{(t)}}{z_{ij}\mid \dataset{},\bm{\theta}^{(t)}}\log\big(P(X_i = x_i \mid Q_i=j,\bm{\theta})P(Q_i=j \mid \bm{\theta})\big).
\end{align*}
Depending on the nature of the data, the family of the emission distributions $P(X \mid Q)$ will be different. For instance, discrete data may require a categorical emission, whereas continuous data could be modeled by a Gaussian distribution. In the former case, it is possible to show that the optimal emission distribution is given by
\begin{align}
& P^{(t+1)}(X = k | Q = j) = \frac{\sum_i \delta(x_i, k) \E{}{z_{ij}\mid \dataset{},\bm{\theta}^{(t)}}}{\sum_{j'=1}^C\sum_i \delta(x_i, j') \E{}{z_{ij}\mid \dataset{},\bm{\theta}^{(t)}}}, \nonumber
\end{align}
with $\delta(\cdot,\cdot)$ being the Kronecker delta function. In practice, the above term amounts to compute the fraction of points belonging to a certain cluster $j$ that have label $k$, each weighted by the expected probability (according to our current parameters) that the point will be in cluster $j$. This can be seen by noticing that the expected value of an indicator variable is its associated probability.

On the other hand, for a Gaussian emission, the sufficient statistics for the $j$-th mixture component are
\begin{align}
& \mu^{(t+1)}_j = \frac{\sum_i x_i \E{}{z_{ij}\mid \dataset{},\bm{\theta}^{(t)}}}{\sum_i \E{}{z_{ij}\mid \dataset{},\bm{\theta}^{(t)}}}, \nonumber \\
& \sigma^{(t+1)}_j = \sqrt{\frac{\sum_i \E{}{z_{ij}\mid \dataset{},\bm{\theta}^{(t)}}(x_i - \mu^{(t)}_j)^2}{\sum_i \E{}{z_{ij}\mid \dataset{},\bm{\theta}^{(t)}}}}. \nonumber
\end{align}

Finally, the prior probability $P(Q)$ is updated as follows:
\begin{align}
& P^{(t+1)}(Q = j) = \frac{\sum_i \E{}{z_{ij}\mid \dataset{},\bm{\theta}^{(t)}}}{\sum_i \sum_{j'=1}^C \E{}{z_{ij'}\mid \dataset{},\bm{\theta}^{(t)}}}. \nonumber
\end{align}

For a complete treatment of the multivariate Gaussian case and more, the reader is referred to \cite{bishop_pattern_2006,barber_bayesian_2012}. We conclude this part with an example of a three-component mixture model fitting a population with three well-separated clusters. Figure \ref{fig:gaussian-mixture-model} shows the outcome of fitting a Gaussian mixture model via maximum likelihood estimation. We observe how the three components adapt their mean (the diamond symbol) and covariance matrices (the ellipses) to capture the original data distribution. These concepts will be extended later on in this thesis to accommodate the complex domain of graphs.
\begin{figure}[ht]
    \begin{center}
        \hspace{-1cm}
        \animategraphics[poster=last, autoplay,loop,keepaspectratio,width=0.7\textwidth,trim=13 11 10 10 ]{1}{Figures/Chapter2/gmm_animation}{1}{20}
    \end{center}
    \caption{Fitting a three-component bivariate Gaussian mixture model on the data samples (crosses). We see that there are three different subpopulations in the data, whose distributions have been well-approximated by the components of the mixture model. The digital readers can click on this figure to start an animation.}
    \label{fig:gaussian-mixture-model}
\end{figure}

\clearpage
\subsection{Mixture Density Networks}
\label{subsec:mdn}
\begin{figure}[ht]
\centering
\begin{subfigure}
\centering
\resizebox{0.49\textwidth}{!}{\input{Figures/Chapter2/mdn_function_plot}}
\end{subfigure}
\begin{subfigure}
\centering
\includegraphics[width=0.49\textwidth]{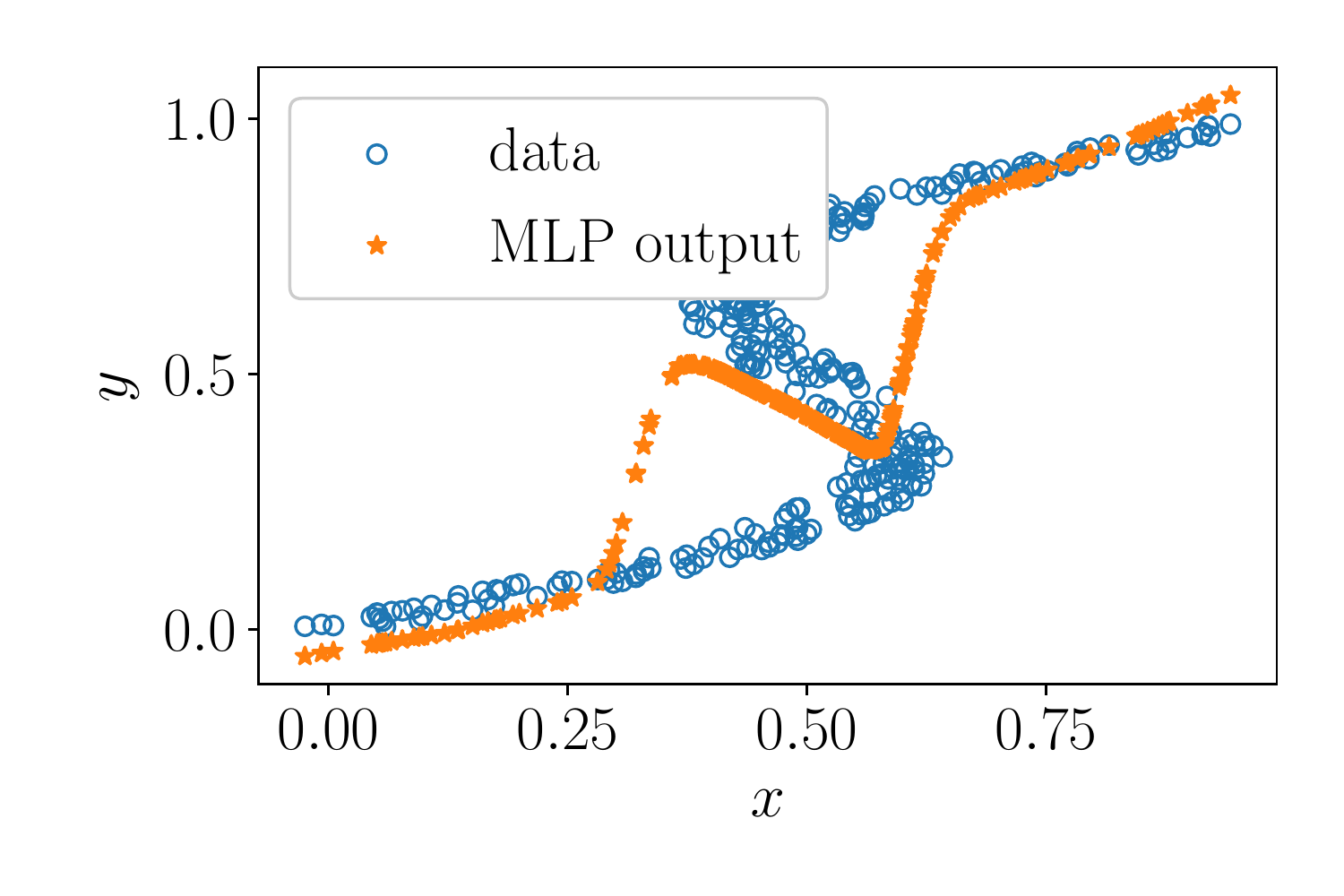}
\end{subfigure}
\caption{We sketch an example of a common regression dataset (left) in which we assume a unimodal Gaussian noise around each output value. Instead, on the right we show an example of an inverse problem, where $y$ can be multi-valued for each $x$ and an MLP fails at capturing the data distribution which cannot be expressed by means of a function.}
\label{fig:mdn-function-plot}
\end{figure}

In supervised \ML{} tasks, we care about approximating the distribution of a target $\bm{y}$ conditioned on an input $\boldx{}$, \ie $P(\bm{y}\mid \boldx{})$. As regards regression problems, the usual underlying assumption is that the output $\bm{y}$ we observe might be noisy, and we let the distribution of such noise be Gaussian. Under these conditions, it is possible to show that the \textbf{function} that computes the expected conditional value of $\bm{y}$ given $\boldx{}$, \ie $\E{\bm{x},\bm{y}\sim \dataset{}}{\bm{y}\mid \boldx{}}$, corresponds to the optimal solution of our regression problem \cite{bishop_mixture_1994}. This is intuitively represented by the left hand-side of Figure \ref{fig:mdn-function-plot}. Put differently, as long as the conditional distribution of the output is \textbf{unimodal}, we can train function approximators such as a Multi-Layer Perceptron (MLP) \cite{rosenblatt_perceptron_1957,rumelhart_learning_1986} to solve the task.

Now let us imagine that the unimodality assumption does not hold anymore. In particular, given an input vector $\boldx{}$, there can be more than one plausible values for the output $\bm{y}$. This is true for the so-called \textit{inverse} problems, such as robot inverse kinematics and stochastic simulations, where the mapping can be multi-valued. As we show in the toy example, adapted from \cite{bishop_mixture_1994}, of Figure \ref{fig:mdn-function-plot} (right), an MLP cannot express uncertainty about the possible values of $\bm{y}$ given $\boldx{}$, due to the fact that neural networks are \textit{function} approximators. Nonetheless, we could use the mixture models from the previous section to capture the (possibly \textbf{multimodal}) distribution $P(\bm{y}\mid \boldx{})$ via maximum likelihood estimation. The crucial difference with respect to the classical formalization of mixture models is the supervised nature of the task. We call this kind of problems \textbf{Conditional Density Estimation} (CDE) tasks.

\begin{figure}[ht]
\centering
\resizebox{0.28\textwidth}{!}{\tikzset{every picture/.style={line width=0.75pt}} 

\begin{tikzpicture}[x=0.75pt,y=0.75pt,yscale=-1,xscale=1]

\draw [line width=1.5]    (292.8,210) -- (230.69,137.78) ;
\draw [shift={(228.08,134.75)}, rotate = 409.3] [fill={rgb, 255:red, 0; green, 0; blue, 0 }  ][line width=0.08]  [draw opacity=0] (11.61,-5.58) -- (0,0) -- (11.61,5.58) -- cycle    ;
\draw [line width=1.5]    (292.8,210) -- (238,210.22) ;
\draw [shift={(234,210.23)}, rotate = 359.77] [fill={rgb, 255:red, 0; green, 0; blue, 0 }  ][line width=0.08]  [draw opacity=0] (11.61,-5.58) -- (0,0) -- (11.61,5.58) -- cycle    ;
\draw [line width=1.5]    (209,210.23) -- (209,147.23) ;
\draw [shift={(209,143.23)}, rotate = 450] [fill={rgb, 255:red, 0; green, 0; blue, 0 }  ][line width=0.08]  [draw opacity=0] (11.61,-5.58) -- (0,0) -- (11.61,5.58) -- cycle    ;
\draw  [fill={rgb, 255:red, 255; green, 255; blue, 255 }  ,fill opacity=1 ][line width=1.5]  (184,210.23) .. controls (184,196.43) and (195.19,185.23) .. (209,185.23) .. controls (222.81,185.23) and (234,196.43) .. (234,210.23) .. controls (234,224.04) and (222.81,235.23) .. (209,235.23) .. controls (195.19,235.23) and (184,224.04) .. (184,210.23) -- cycle ;
\draw  [fill={rgb, 255:red, 31; green, 119; blue, 180 }  ,fill opacity=1 ][line width=1.5]  (184,118.23) .. controls (184,104.43) and (195.19,93.23) .. (209,93.23) .. controls (222.81,93.23) and (234,104.43) .. (234,118.23) .. controls (234,132.04) and (222.81,143.23) .. (209,143.23) .. controls (195.19,143.23) and (184,132.04) .. (184,118.23) -- cycle ;
\draw  [line width=1.5]  (170.67,89.33) -- (326.34,89.33) -- (326.34,339.19) -- (170.67,339.19) -- cycle ;
\draw [line width=1.5]    (209,294.23) -- (293.88,294.19) -- (293.52,231.73) ;
\draw [shift={(293.5,227.73)}, rotate = 449.68] [fill={rgb, 255:red, 0; green, 0; blue, 0 }  ][line width=0.08]  [draw opacity=0] (11.61,-5.58) -- (0,0) -- (11.61,5.58) -- cycle    ;
\draw  [fill={rgb, 255:red, 31; green, 119; blue, 180 }  ,fill opacity=1 ][line width=1.5]  (184,294.23) .. controls (184,280.43) and (195.19,269.23) .. (209,269.23) .. controls (222.81,269.23) and (234,280.43) .. (234,294.23) .. controls (234,308.04) and (222.81,319.23) .. (209,319.23) .. controls (195.19,319.23) and (184,308.04) .. (184,294.23) -- cycle ;
\draw  [fill={rgb, 255:red, 31; green, 119; blue, 180 }  ,fill opacity=1 ][line width=1.5]  (275.22,202.42) .. controls (275.22,196.89) and (279.69,192.42) .. (285.22,192.42) -- (300.38,192.42) .. controls (305.91,192.42) and (310.38,196.89) .. (310.38,202.42) -- (310.38,217.58) .. controls (310.38,223.11) and (305.91,227.58) .. (300.38,227.58) -- (285.22,227.58) .. controls (279.69,227.58) and (275.22,223.11) .. (275.22,217.58) -- cycle ;

\draw (209,210.23) node  [font=\Large]  {$Q_{i}$};
\draw (209,118.23) node  [font=\Large,color={rgb, 255:red, 255; green, 255; blue, 255 }  ,opacity=1 ]  {$Y_{i}$};
\draw (302.76,323.31) node  [font=\Large] [align=left] {$\displaystyle |\mathcal{D} |$};
\draw (209,294.23) node  [font=\Large,color={rgb, 255:red, 255; green, 255; blue, 255 }  ,opacity=1 ]  {$X_{i}$};
\draw (293.2,211.8) node  [font=\Large,color={rgb, 255:red, 255; green, 255; blue, 255 }  ,opacity=1 ]  {$\boldsymbol{z}_{i}$};

\end{tikzpicture}}
    \caption{We sketch the graphical model for a Mixture Density Network model. Notice how the value of the variable $X_i$ undergoes a deterministic transformation into the hidden representation $\bm{z}_i$.}\label{fig:mdn-graphical-model}
\end{figure}
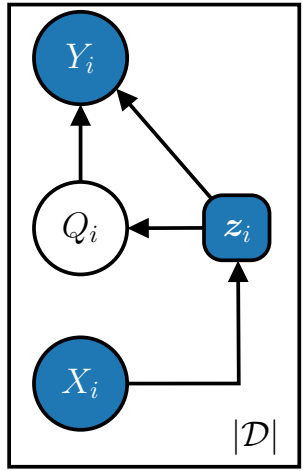

Therefore, the idea behind the Mixture Density Network (MDN) model \cite{bishop_mixture_1994} is to compute multimodal output distributions conditioned on arbitrary \textbf{flat} input data. Since a fully probabilistic formulation with closed-form solutions is difficult in the general case, MDN minimizes the negative log-likelihood of the data using backpropagation. We take advantage of the extended plate notation introduced before and represent the Mixture Density Network as a graphical model in Figure \ref{fig:mdn-graphical-model}.

The first step of the process is to encode the input $\boldx{}$ into a hidden representation $\bm{z}$. This can be achieved, for instance, with any neural network of choice. Then, the first $C$ components of the hidden representation, namely $\bm{z}^{Q}$ are used to compute the \textit{conditional} mixing weights for the $C$ output distributions:
\begin{align*}
    P(Q=j|\boldx{}) = \frac{e^{z^{Q}_j}}{\sum_{j'}^Ce^{z^{Q}_{j'}}}.
\end{align*}
The other components of the $\bm{z}$ vector are used to compute the sufficient statistics of the $C$ output distributions. For a continuous output $y$, this amounts to having $2C$ parameters in $\bm{z}$ associated with the means ($\bm{z}^{\mu}$) and standard deviations ($\bm{z}^{\sigma}$) of univariate Gaussians. On a practical note, the variance can be kept stricly positive by applying an exponential transformation to the related components of $\bm{z}$, whereas under/overflow numerical errors can be mitigated with the \quotes{exp-normalise trick}
\begin{align*}
    & P(Q=j|\boldx{}) = \frac{e^{z^{Q}_j - b}}{\sum_{j'}^Ce^{z^{Q}_{j'} - b}}, \ \ b = \max_{j'} z^{Q}_{j'}.
\end{align*}
Similarly to mixture models, the objective to maximize is the log-likelihood of the \iid{} samples:
\begin{align*}
    \log P(\bm{y}\mid \bm{x}) = \sum_j^C P(y\mid z_j^{\mu},z_j^{\sigma})P(Q=j\mid \bm{x}).
\end{align*}

To show that MDNs can solve the conditional density estimation problem for the toy dataset of Figure \ref{fig:mdn-function-plot}, we train a three-component MDN on that dataset and visualize both the values of the mixing weights $P(Q\mid x)$ and those of the means $z^\mu_i$ (Figure \ref{fig:mixture-density-network-example} - top). To observe whether the learned probability distribution captures the uncertainty associated with each data point $x$, we also draw the probability contour plot of the model as well as the mean value of the most likely conditional mode (Figure \ref{fig:mixture-density-network-example} - bottom); the latter is selected by simple inspection of the mixing weights. We can see how a MDN is able to solve the conditional density estimation problem, by providing a multimodal distribution for each input $x$ that can be appreciated by looking at the contour plot.
\begin{figure}[h!]
    \begin{center}
        \animategraphics[poster=last, autoplay,loop,keepaspectratio,width=1\textwidth,trim=13 13 5 10 ]{1}{Figures/Chapter2/mdn_animation}{1}{15}
    \end{center}
    \caption{Fitting a three-component Mixture Density Network model on the toy dataset of Figure \ref{fig:mdn-function-plot}, as well as the value for the mixing weights and Gaussian means when the value of the scalar input $x$ varies. The model can correctly capture the data distribution by properly mixing the different Gaussians. The digital readers can click on this figure to start an animation.}
    \label{fig:mixture-density-network-example}
\end{figure}

\clearpage
\subsection{Bayesian Nonparametric Mixture Models}
One of the most recurrent questions that \ML{} practitioners ask when using mixture models is \quotes{how many components should we use?}. In this section, we introduce techniques that \textbf{automatically} answer this question by exploiting the available data. The field of Bayesian Nonparametric (BNP) methods deals with statistical models that are, in fact, \textit{not parametric}: the parameters of the model \textbf{grow/shrink with the data}, and as such the model cannot be simply formalized by means of a fixed number of parameters. Of course, this does not mean that we shall not make any assumption about the underlying data distribution, as we still have hyper-parameters to tune and distribution families to choose. The BNP field has been extensively studied \cite{hoppe_polya_1984,sethuraman_constructive_1994,neal_markov_2000,teh_hdp_2006,orbanz_bayesian_2010,teh_dirichlet_2010,gershman_tutorial_2012} so, in the interest of conciseness, we shall focus on the most relevant and intuitive concepts that will be needed in the following. We will provide a high-level summary of a Dirichlet Processes (DPs), its basic properties, and ways to represent DP-based models, before discussing how we can perform inference in (Hierarchical) \textbf{\DP{} mixture models}.

Let us start the discussion with the fundamental notion of \textbf{exchangeability} \cite{orbanz_bayesian_2010}. Informally, this notion specifies that the order in which a sequence of $n$ identically distributed observations $\{X_1,\dots, X_n\}$ appear does not influence the joint probability distribution $P(X_1,\dots, X_n)$.
\begin{definition}[\textbf{Exchangeability}]
Let  $\{X_1,\dots
, X_n\}$ be a set of $n$ random variables, each defined on the same probability space, and let $P$ be their joint distribution. These variables are said to be exchangeable if, for any permutation $\sigma$ of \{1,\dots,n\} it holds
\begin{align*}
P(X_1=x_1,\dots,X_n=x_n) = P(X_1=x_{\sigma(1)},\dots,X_n=x_{\sigma(n)}).
\end{align*}
\end{definition}
Notice that this does not imply the observations are \iid{}; on the contrary, \iid{} variables are implicitly exchangeable. The notion of exchangeability is very important in the Bayesian/non-Bayesian debate about modeling the parameters as random variables. Because of De Finetti's Theorem \cite{funzione_finetti_1931}, under the exchangeability assumption there exists a random variable $\bm{\theta}$ such that $P(X_1,\dots
, X_n) = \int P(\bm{\theta})\prod_{i=1}^n P(X_i\mid \bm{\theta})d\bm{\theta}$. Therefore, for exchangeable sequences, there exists a Bayesian model whose parameters are in fact a \textbf{random variable}. 

For the rest of the section, we shall follow \cite{teh_dirichlet_2010} and assume a setting where observations are exchangeable. Without going into needless technicalities, we define a \textbf{Dirichlet Process} \cite{ferguson_bayesian_1973} as a distribution over other probability distributions. We parametrize a \DP{} using a \textbf{base distribution} $G_0$, which is the expected value of the process, and a \textbf{concentration} (or scaling) parameter $\alpha_0$ that controls how close \DP{} realizations are to $G_0$. Moreover, draws from DP($G_0, \alpha_0$) almost surely form a discrete distribution, even when $G_0$ is continuous. 

Intuitively, to generate new data using a \DP{}, one first draws a distribution $G$ from the \DP{} and then uses that distribution to independently sample values for $X_1,X_2,\dots$. Nevertheless, the harsh truth is that, in practice, one cannot directly sample a distribution from a \DP{}, as doing otherwise would require an infinite amount of information to represent the \DP{}. Consequently, over the years researchers developed different ways to draw samples from a \DP{} \cite{orbanz_bayesian_2010}. Some of them, like the Chinese Restaurant Process (CRP) \cite{aldous_exchangeability_1985} define the \DP{} \textbf{implicitly}. Others describe a random draw, rather than the distribution, in an \textbf{explicit} matter, \eg the Stick-Breaking construction  \cite{sethuraman_constructive_1994}. Finally, it is possible to take the limit of a \textbf{finite and parametric} model to obtain a nonparametric one, an instance of which is the finite mixture model that can be converted into a \DP{} mixture model \cite{teh_hdp_2006}. In the following, we shall focus on the Stick-breaking representation for its convenient implementation.

\subsubsection{The Stick-Breaking Construction}
\label{subsubsec:stick-breaking}
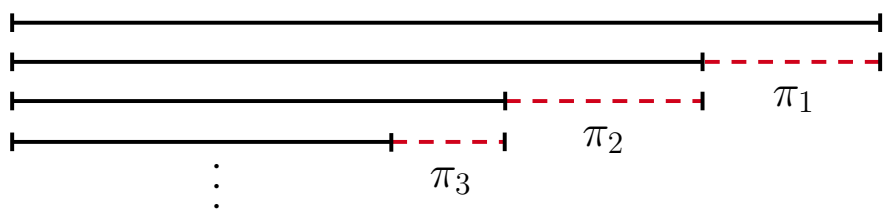
\begin{figure}[ht]
\centering
\resizebox{0.8\textwidth}{!}{\tikzset{every picture/.style={line width=0.75pt}} 

\begin{tikzpicture}[x=0.75pt,y=0.75pt,yscale=-1,xscale=1]

\draw [color={rgb, 255:red, 208; green, 2; blue, 27 }  ,draw opacity=1 ][line width=1.5]  [dash pattern={on 5.63pt off 4.5pt}]  (293.03,130.86) -- (350.18,130.86) ;
\draw [color={rgb, 255:red, 208; green, 2; blue, 27 }  ,draw opacity=1 ][line width=1.5]  [dash pattern={on 5.63pt off 4.5pt}]  (350.6,110) -- (450.13,110) ;
\draw [color={rgb, 255:red, 208; green, 2; blue, 27 }  ,draw opacity=1 ][line width=1.5]  [dash pattern={on 5.63pt off 4.5pt}]  (451.46,90) -- (540.13,90) ;
\draw [line width=1.5]    (100.42,90) -- (450.41,90) ;
\draw [line width=1.5]    (100.42,110) -- (350.7,110) ;
\draw [line width=1.5]    (100.42,85.23) -- (100.42,94.73) ;
\draw [line width=1.5]    (100.42,105.23) -- (100.42,114.73) ;
\draw [line width=1.5]    (540.27,85.23) -- (540.27,94.73) ;
\draw [line width=1.5]    (450.27,105.23) -- (450.27,114.73) ;
\draw [line width=1.5]    (100.43,130.86) -- (292.03,130.86) ;
\draw [line width=1.5]    (100.43,126.08) -- (100.43,135.58) ;
\draw [line width=1.5]    (350.02,126.08) -- (350.02,135.58) ;
\draw [line width=1.5]    (450.27,85.23) -- (450.27,94.73) ;
\draw [line width=1.5]    (350.27,105.23) -- (350.27,114.73) ;
\draw [line width=1.5]    (292.56,126.36) -- (292.56,135.86) ;
\draw [line width=1.5]    (100.42,70) -- (540.33,70) ;
\draw [line width=1.5]    (100.42,65.23) -- (100.42,74.73) ;
\draw [line width=1.5]    (540.27,65.23) -- (540.27,74.73) ;

\draw (496.55,109.05) node  [font=\LARGE]  {$\pi _{1}$};
\draw (400.12,129.91) node  [font=\LARGE]  {$\pi _{2}$};
\draw (322.69,150.76) node  [font=\LARGE]  {$\pi _{3}$};
\draw (204.02,155.26) node  [font=\LARGE,rotate=-90.19]  {$\dotsc $};

\end{tikzpicture}}
    \caption{By recursively splitting a stick of length 1, we can obtain an infinite number of prior probabilities $\pi_i$.}\label{fig:stick-breaking}
\end{figure}

We now present the Stick-Breaking construction, an explicit method to represent a \DP{}. To start, imagine having a stick of length 1 and splitting it every time we need to create a new component in an infinite but discrete distribution (as in Figure \ref{fig:stick-breaking}). While each component is assigned to a piece of the stick, whose length represents the prior probability of that component, all the other infinitely many components are \textbf{represented by the unassigned portion of that stick}, that is, the remaining portion of the black stick in the figure.
Practically speaking, this allows us to implement a \DP{} mixture model on a computer, since we could create a potentially infinite number of mixtures but only a finite amount of them is stored in memory.

On a more formal note, the stick-breaking construction is based on sequences of independent (but not identically distributed) random variables, namely $\{\pi_k\}_{k\in \mathbb{N}}$ and $\{\bm{\theta}_k\}_{k\in \mathbb{N}}$:
\begin{align*}
	\begin{alignedat}{2}
		\pi'_k \mid \alpha_0,G_0 &\sim \text{Beta}(1, \alpha_0) &\qquad \qquad \bm{\theta}_k \mid \alpha_0, G_0 &\sim G_0
	\end{alignedat}
\end{align*}
where Beta$(a,b) = \frac{\Gamma(a)\Gamma(b)}{\Gamma(a+b)}$ is the Beta distribution. Defining a discrete probability distribution $G$ as
\begin{align*}
    \pi_k = \pi'_k \prod_{l=1}^{k-1}(1-\pi'_l) &\qquad \qquad G = \sum_{k=1}^{\infty} \pi_k \delta_{\bm{\theta}_k}
\end{align*}
where $\delta_{\bm{\theta}_k}$ is the \textbf{Dirac} distribution with all probability mass concentrated on $\bm{\theta}_k$, it can be shown \cite{sethuraman_constructive_1994} that G is distributed according to $\text{DP}(\alpha_0,G_0)$.

Because it holds $\sum_k \pi_k = 1$, we can interpret $\bm{\pi} = \{\pi_k\}_{k \in \mathbb{N}}$ as a discrete random variable whose support is $\mathbb{N}$. From now on, we shall identify the stick-breaking construction using the standard terminology $\bm{\pi} = \text{Stick}(\alpha_0)$ \cite{teh_hdp_2006}.

\subsubsection{Dirichlet Process Mixture Models}
\label{subsubsec:dpmm}
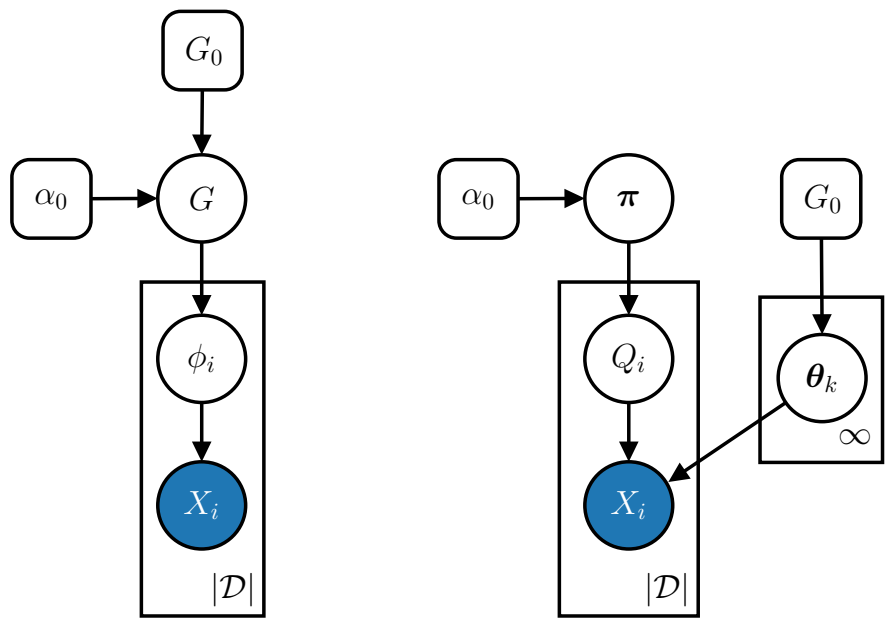
\begin{figure}[ht]
\centering
\resizebox{0.8\textwidth}{!}{\tikzset{every picture/.style={line width=0.75pt}} 

\begin{tikzpicture}[x=0.75pt,y=0.75pt,yscale=-1,xscale=1]

\draw [line width=1.5]    (519,238.23) -- (434.81,295.58) ;
\draw [shift={(431.5,297.83)}, rotate = 325.74] [fill={rgb, 255:red, 0; green, 0; blue, 0 }  ][line width=0.08]  [draw opacity=0] (11.61,-5.58) -- (0,0) -- (11.61,5.58) -- cycle    ;
\draw [line width=1.5]    (166,227.23) -- (166,282.23) ;
\draw [shift={(166,286.23)}, rotate = 270] [fill={rgb, 255:red, 0; green, 0; blue, 0 }  ][line width=0.08]  [draw opacity=0] (11.61,-5.58) -- (0,0) -- (11.61,5.58) -- cycle    ;
\draw [line width=1.5]    (166,135.23) -- (166,198.23) ;
\draw [shift={(166,202.23)}, rotate = 270] [fill={rgb, 255:red, 0; green, 0; blue, 0 }  ][line width=0.08]  [draw opacity=0] (11.61,-5.58) -- (0,0) -- (11.61,5.58) -- cycle    ;
\draw  [fill={rgb, 255:red, 255; green, 255; blue, 255 }  ,fill opacity=1 ][line width=1.5]  (141,227.23) .. controls (141,213.43) and (152.19,202.23) .. (166,202.23) .. controls (179.81,202.23) and (191,213.43) .. (191,227.23) .. controls (191,241.04) and (179.81,252.23) .. (166,252.23) .. controls (152.19,252.23) and (141,241.04) .. (141,227.23) -- cycle ;
\draw  [fill={rgb, 255:red, 255; green, 255; blue, 255 }  ,fill opacity=1 ][line width=1.5]  (141,135.23) .. controls (141,121.43) and (152.19,110.23) .. (166,110.23) .. controls (179.81,110.23) and (191,121.43) .. (191,135.23) .. controls (191,149.04) and (179.81,160.23) .. (166,160.23) .. controls (152.19,160.23) and (141,149.04) .. (141,135.23) -- cycle ;
\draw  [line width=1.5]  (131.67,183.17) -- (201.37,183.17) -- (201.37,374.63) -- (131.67,374.63) -- cycle ;
\draw  [fill={rgb, 255:red, 31; green, 119; blue, 180 }  ,fill opacity=1 ][line width=1.5]  (141,311.23) .. controls (141,297.43) and (152.19,286.23) .. (166,286.23) .. controls (179.81,286.23) and (191,297.43) .. (191,311.23) .. controls (191,325.04) and (179.81,336.23) .. (166,336.23) .. controls (152.19,336.23) and (141,325.04) .. (141,311.23) -- cycle ;
\draw [line width=1.5]    (80.5,135.5) -- (137,135.25) ;
\draw [shift={(141,135.23)}, rotate = 539.75] [fill={rgb, 255:red, 0; green, 0; blue, 0 }  ][line width=0.08]  [draw opacity=0] (11.61,-5.58) -- (0,0) -- (11.61,5.58) -- cycle    ;
\draw  [fill={rgb, 255:red, 255; green, 255; blue, 255 }  ,fill opacity=1 ][line width=1.5]  (58,123) .. controls (58,117.48) and (62.48,113) .. (68,113) -- (93,113) .. controls (98.52,113) and (103,117.48) .. (103,123) -- (103,148) .. controls (103,153.52) and (98.52,158) .. (93,158) -- (68,158) .. controls (62.48,158) and (58,153.52) .. (58,148) -- cycle ;
\draw [line width=1.5]    (166.5,57.5) -- (166.04,106.23) ;
\draw [shift={(166,110.23)}, rotate = 270.54] [fill={rgb, 255:red, 0; green, 0; blue, 0 }  ][line width=0.08]  [draw opacity=0] (11.61,-5.58) -- (0,0) -- (11.61,5.58) -- cycle    ;
\draw  [fill={rgb, 255:red, 255; green, 255; blue, 255 }  ,fill opacity=1 ][line width=1.5]  (144,38) .. controls (144,32.48) and (148.48,28) .. (154,28) -- (179,28) .. controls (184.52,28) and (189,32.48) .. (189,38) -- (189,63) .. controls (189,68.52) and (184.52,73) .. (179,73) -- (154,73) .. controls (148.48,73) and (144,68.52) .. (144,63) -- cycle ;
\draw [line width=1.5]    (409,227.23) -- (409,282.23) ;
\draw [shift={(409,286.23)}, rotate = 270] [fill={rgb, 255:red, 0; green, 0; blue, 0 }  ][line width=0.08]  [draw opacity=0] (11.61,-5.58) -- (0,0) -- (11.61,5.58) -- cycle    ;
\draw [line width=1.5]    (409,135.23) -- (409,198.23) ;
\draw [shift={(409,202.23)}, rotate = 270] [fill={rgb, 255:red, 0; green, 0; blue, 0 }  ][line width=0.08]  [draw opacity=0] (11.61,-5.58) -- (0,0) -- (11.61,5.58) -- cycle    ;
\draw  [fill={rgb, 255:red, 255; green, 255; blue, 255 }  ,fill opacity=1 ][line width=1.5]  (384,227.23) .. controls (384,213.43) and (395.19,202.23) .. (409,202.23) .. controls (422.81,202.23) and (434,213.43) .. (434,227.23) .. controls (434,241.04) and (422.81,252.23) .. (409,252.23) .. controls (395.19,252.23) and (384,241.04) .. (384,227.23) -- cycle ;
\draw  [fill={rgb, 255:red, 255; green, 255; blue, 255 }  ,fill opacity=1 ][line width=1.5]  (384,135.23) .. controls (384,121.43) and (395.19,110.23) .. (409,110.23) .. controls (422.81,110.23) and (434,121.43) .. (434,135.23) .. controls (434,149.04) and (422.81,160.23) .. (409,160.23) .. controls (395.19,160.23) and (384,149.04) .. (384,135.23) -- cycle ;
\draw  [line width=1.5]  (369.5,183.17) -- (448.5,183.17) -- (448.5,374.63) -- (369.5,374.63) -- cycle ;
\draw  [fill={rgb, 255:red, 31; green, 119; blue, 180 }  ,fill opacity=1 ][line width=1.5]  (384,311.23) .. controls (384,297.43) and (395.19,286.23) .. (409,286.23) .. controls (422.81,286.23) and (434,297.43) .. (434,311.23) .. controls (434,325.04) and (422.81,336.23) .. (409,336.23) .. controls (395.19,336.23) and (384,325.04) .. (384,311.23) -- cycle ;
\draw [line width=1.5]    (323.5,135.5) -- (380,135.25) ;
\draw [shift={(384,135.23)}, rotate = 539.75] [fill={rgb, 255:red, 0; green, 0; blue, 0 }  ][line width=0.08]  [draw opacity=0] (11.61,-5.58) -- (0,0) -- (11.61,5.58) -- cycle    ;
\draw  [fill={rgb, 255:red, 255; green, 255; blue, 255 }  ,fill opacity=1 ][line width=1.5]  (301,123) .. controls (301,117.48) and (305.48,113) .. (311,113) -- (336,113) .. controls (341.52,113) and (346,117.48) .. (346,123) -- (346,148) .. controls (346,153.52) and (341.52,158) .. (336,158) -- (311,158) .. controls (305.48,158) and (301,153.52) .. (301,148) -- cycle ;
\draw [line width=1.5]    (518.5,142.5) -- (518.97,209.23) ;
\draw [shift={(519,213.23)}, rotate = 269.59000000000003] [fill={rgb, 255:red, 0; green, 0; blue, 0 }  ][line width=0.08]  [draw opacity=0] (11.61,-5.58) -- (0,0) -- (11.61,5.58) -- cycle    ;
\draw  [fill={rgb, 255:red, 255; green, 255; blue, 255 }  ,fill opacity=1 ][line width=1.5]  (496,123) .. controls (496,117.48) and (500.48,113) .. (506,113) -- (531,113) .. controls (536.52,113) and (541,117.48) .. (541,123) -- (541,148) .. controls (541,153.52) and (536.52,158) .. (531,158) -- (506,158) .. controls (500.48,158) and (496,153.52) .. (496,148) -- cycle ;
\draw  [line width=1.5]  (483.67,191.83) -- (553.37,191.83) -- (553.37,286.59) -- (483.67,286.59) -- cycle ;
\draw  [fill={rgb, 255:red, 255; green, 255; blue, 255 }  ,fill opacity=1 ][line width=1.5]  (494,238.23) .. controls (494,224.43) and (505.19,213.23) .. (519,213.23) .. controls (532.81,213.23) and (544,224.43) .. (544,238.23) .. controls (544,252.04) and (532.81,263.23) .. (519,263.23) .. controls (505.19,263.23) and (494,252.04) .. (494,238.23) -- cycle ;

\draw (166,227.23) node  [font=\Large]  {$\phi _{i}$};
\draw (166,135.23) node  [font=\Large,color={rgb, 255:red, 0; green, 0; blue, 0 }  ,opacity=1 ]  {$G$};
\draw (183.81,359.46) node  [font=\Large] [align=left] {$\displaystyle |\mathcal{D} |$};
\draw (166,311.23) node  [font=\Large,color={rgb, 255:red, 255; green, 255; blue, 255 }  ,opacity=1 ]  {$X_{i}$};
\draw (80.5,135.5) node  [font=\Large] [align=left] {$\displaystyle \alpha _{0}$};
\draw (167.5,50.5) node  [font=\Large] [align=left] {$\displaystyle G_{0}$};
\draw (409,227.23) node  [font=\Large]  {$Q_{i}$};
\draw (409,135.23) node  [font=\Large,color={rgb, 255:red, 0; green, 0; blue, 0 }  ,opacity=1 ]  {$\boldsymbol{\pi }$};
\draw (430.81,359.46) node  [font=\Large] [align=left] {$\displaystyle |\mathcal{D} |$};
\draw (409,311.23) node  [font=\Large,color={rgb, 255:red, 255; green, 255; blue, 255 }  ,opacity=1 ]  {$X_{i}$};
\draw (323.5,135.5) node  [font=\Large] [align=left] {$\displaystyle \alpha _{0}$};
\draw (519.5,135.5) node  [font=\Large] [align=left] {$\displaystyle G_{0}$};
\draw (537.81,271.66) node  [font=\Large] [align=left] {$\displaystyle \infty $};
\draw (519,238.23) node  [font=\Large]  {$\boldsymbol{\theta }_{k}$};

\end{tikzpicture}}
    \caption{The Dirichlet Process Mixture Model can be graphically represented in more than one way. Here, we present its classical version (left) as well as the Stick-Breaking variant (right).}\label{fig:dpmm}
\end{figure}

A Dirichlet Process mixture model is a BNP model in which a DP acts as prior in an infinite mixture of distributions \cite{orbanz_bayesian_2010,gershman_tutorial_2012,teh_hdp_2006,teh_dirichlet_2010}. Assuming $G$ has distribution DP$(\alpha_0,G_0)$, we can formalize this model by writing
\begin{align*}
    & \phi_i \mid G \sim G \\
    & x_i \mid \phi_i \sim F(\phi_i),
\end{align*}
where we use the abstraction $F(\phi_i)$ to denote which emission distribution to use for the \textbf{factor} $\phi_i$. For instance, $\phi_i$ might take as a value the parameters of a mixture model component, whose emission distribution is represented as $F(\phi_i)$. The graphical model is visually represented on the left hand-side of Figure \ref{fig:dpmm}.
We can draw an alternative formalization using the stick-breaking construction we just introduced. First of all, let us assume that the factor $\phi_i$ of a data point $x_i$ can take values $\bm{\theta}_k$ with probability $\pi_k$, according to the formulation of $G$ in the stick-breaking construction. Then, we introduce the random variable $Q_i$, distributed as $\bm{\pi}$, with support over the set $\mathbb{N}$; the value $Q_i$ is interpreted as the index of an infinite mixture component. Whenever $Q_i=q_i$, the emission distribution for $x_i$ will be a $\bm{\theta}_{q_i}$ drawn from $G_0$. The graphical model is sketched in Figure \ref{fig:dpmm} (right hand-side) and formally defined below:
\begin{align*}
	\begin{alignedat}{2}
		\bm{\pi} \mid \alpha_0 &\sim \text{Stick}(\alpha_0) &\qquad \qquad q_i \mid \bm{\pi} &\sim \bm{\pi} \\
		\bm{\theta}_k \mid G_0 &\sim G_0 &\qquad \qquad x_i \mid q_i,\{\bm{\theta}\}_{k \in \mathbb{N}} &\sim F(\bm{\theta}_{q_i}).
	\end{alignedat}
\end{align*}

Due to the sheer complexity of the BNP treatment, we will defer details about inference (\ie learning) to the following chapters. However, for the sake of clarity, let us conclude this section by showing, in Figure \ref{fig:dp-mixture-model-example}, how a DP Gaussian mixture model \textbf{learns} the \quotes{right} number of clusters while fitting the data of a toy problem.
\begin{figure}[ht]
    \begin{center}
        \animategraphics[every=2,poster=last, autoplay,loop,keepaspectratio,width=1\textwidth,trim=13 13 7 10 ]{1}{Figures/Chapter2/dpgmm_animation}{1}{60}
    \end{center}
    \caption{We show the behaviour of a \DP{} Gaussian mixture model (variational inference implementation) that starts with 10 Gaussian components initialized using the kmeans algorithm \cite{macqueen_some_1967}. Iteration after iteration, the number of active components decreases until it reaches 3, the same number of clusters that generated the data. In other words, the model adapted its complexity to fit the underlying data distribution. The digital readers can click on this figure to start an animation.}
    \label{fig:dp-mixture-model-example}
\end{figure}

\subsubsection{Hierarchical Dirichlet Process Mixture Models}
\begin{figure}[ht]
\centering
\resizebox{0.8\textwidth}{!}{\input{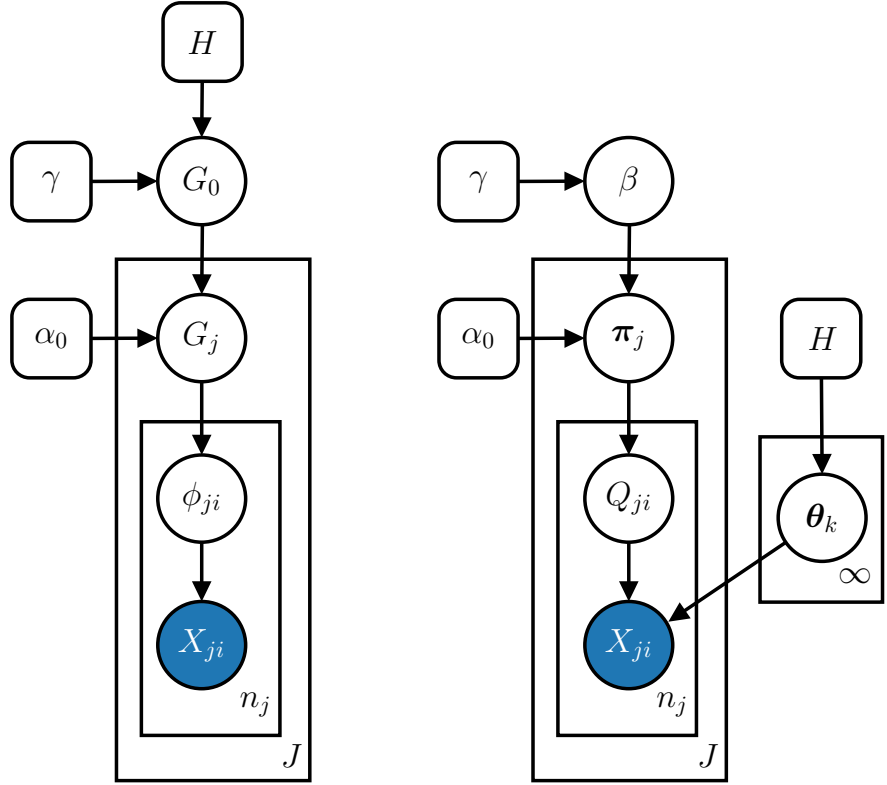}}
    \caption{We visualize the Hierarchical Dirichlet Process Mixture Model in both its classical version (left) and Stick-Breaking alternative (right).}\label{fig:hdpmm}
\end{figure}

In this thesis, we will work with a more complex version of a DP called Hierarchical Dirichlet Process (HDP) \cite{teh_hdp_2006}. In essence, an HDP adds a prior on the base distribution $G_0$ using another DP, which is parametrized by a concentration parameter $\gamma$ and a base distribution $H$. The rationale behind the HDP is given by the problem setting: the data points belong to $J$ \textbf{known} and related groups, and each group has its own DP mixture model \cite{teh_hdp_2006}. In addition, we want to \textit{induce dependencies} between these DP mixture models by sharing the parameters of the emission distributions, \ie the clusters are the same. As before, there exists a stick-breaking version of the HDP that we depict in Figure \ref{fig:hdpmm}.

We now define the formulation of the stick-breaking construction for an HDP. Recall that, for the $i$-th data point, the assignment to a group $j$ is known; we will use the subscript $x_{ji}$ to reflect this notion.
\begin{align*}
	\begin{alignedat}{2}
		\bm{\beta} \mid \gamma &\sim \text{Stick}(\gamma) & \\
		\bm{\pi_j} \mid \alpha_0,\bm{\beta} &\sim \text{DP}(\alpha_0,\bm{\beta}) &\qquad \qquad q_{ji} \mid \bm{\pi}_j &\sim \bm{\pi}_j \\
		\bm{\theta}_k \mid H &\sim H &\qquad \qquad x_{ji} \mid q_{ji},\{\bm{\theta}\}_{k \in \mathbb{N}} &\sim F(\bm{\theta}_{q_{ji}}).
	\end{alignedat}
\end{align*}
However, it still remains unclear how one can sample $\bm{\pi}_j$ from $\text{DP}(\alpha_0,\bm{\beta})$. In \cite{teh_hdp_2006}, the authors show that there exists a relationship between $\bm{\pi}_j$ and $\bm{\beta}$. In particular, it holds
\begin{align*}
	\pi'_{jk} \sim \text{Beta}\Big(\alpha_0\beta_k, \alpha_0\big(1 - \sum_{l=1}^k\beta_l \big)\Big) &\qquad \qquad \pi_{jk} = \pi'_{jk} \prod_{l=1}^{k-1}(1-\pi'_{jl}).
\end{align*}
All in all, these are the basic notions of the model upon which we shall build an infinite mixture model for graphs.

\paragraph*{Inference.} To perform inference in BNP mixture models, one usually relies on Markov Chain Monte Carlo (MCMC) algorithms \cite{neal_markov_2000,orbanz_bayesian_2010,gershman_tutorial_2012}. This is possible whenever we can efficiently integrate out the infinitely many \quotes{unused} latent components of the posterior distribution, so that only the values for the finite number of active components are to be inferred using known inference procedures. Recall that the idea of MCMC is to define a Markov chain on the hidden variables, and by drawing samples from this chain we \textbf{eventually} get a sample coming from the true posterior; the Gibbs sampling algorithm we already saw belongs to the family of MCMC methods. Generally speaking, it is not uncommon that MCMC algorithms require many draws before producing high-quality samples, though the burn-in period heavily depends on the chosen algorithm.

Notice that, due to the random variables being exchangeable, Gibbs sampling is particularly suitable for DP mixture models \cite{gershman_tutorial_2012}. Indeed, each observation in the dataset can be chosen as the last to use, since the order of appearance of the values does not matter in the joint distribution. This is why we will rely on Gibbs sampling for the BNP method developed in this thesis.

\section{Graph Basics}

In this section, we shall give a few but fundamental definitions, taken from graph theory \cite{bondy_graph_1976} and deep learning for graphs \cite{bacciu_gentle_2020}, which will be abundantly used throughout the rest of this work. Generally speaking, a graph is a highly flexible data structure whose entities of interest are connected to each other by some particular form of relationship. The way connections are organized is often called the \textbf{structure} (or \textbf{topology}) of the graph. It is this flexibility what makes it hard to learn from graphs, since we need to take into account the topological variability of each input sample.

\subsection{Fundamentals}
\label{subsec:fundamentals}

Let us commence with a more precise definition of what a graph is.
\begin{definition}[\textbf{Graph}]
A graph is a tuple $g = (\Vset{g},\Eset{g},\Xset{g},\Aset{g})$ where $\Vset{g}$ is the set of \textbf{vertices} (or \textbf{nodes} with slight abuse of terminology \cite{cormen_introduction_2009}) identifying the entities of interest, and $\Eset{g}$ is the set of \textbf{edges} (or \textbf{arcs}) that couple pairs of \textbf{adjacent} vertices. We will follow the notational convention that a vertex in a graph is identified by a natural number with symbols $u$ or $v$. Instead, $\Xset{g}$ is a function that takes a vertex $u \in \Vset{g}$ and maps it to a vector of \textbf{vertex features} (or attributes) $\boldx{u}$; similarly, $\Aset{g}$ maps edges to \textbf{edge features} $\bm{a}_\uv$. Note that the \textbf{size} of a graph $g$ corresponds to the cardinality of the vertex set, \ie $size(g) = |\Vset{g}|$.
\end{definition}

When vertex features are available, we require that \textit{each} vertex $u$ has its own feature vector, and the same must hold for each edge. In jargon, when vertex and/or edge features are used, the graph must be \quotes{uniformly labelled} \cite{frasconi_general_1998}. For many practical problems, is often the case that $\Xset{g}: \Vset{g}\rightarrow \R{d}, d \in \mathbb{N}$ and $\Aset{g}: \Eset{g}\rightarrow \R{d'}, d' \in \mathbb{N}$. Nonetheless, due to the nature of the models proposed in this thesis, most of the time we will consider discrete or continuous values \wlog{}. In case a graph has no vertex feature information, it suffices to consider an equivalent graph in which all vertices have the same dummy feature; clearly, the same goes whenever edge features are missing.

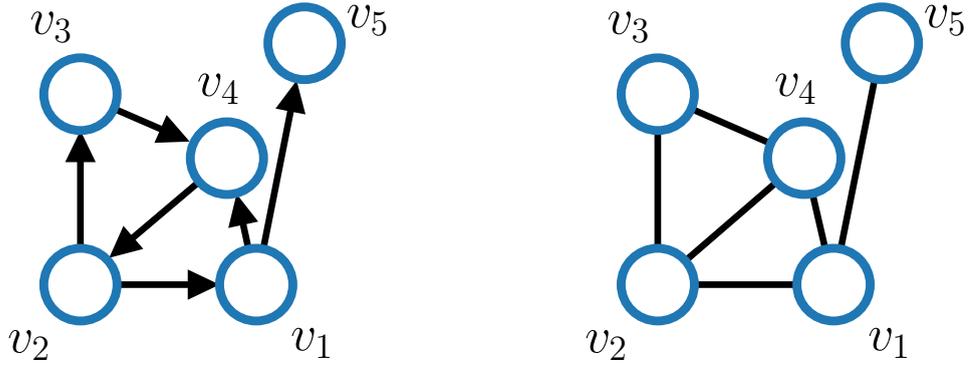
\begin{figure}[ht]
    \centering
    \resizebox{0.9\textwidth}{!}{\tikzset{every picture/.style={line width=0.75pt}} 

\begin{tikzpicture}[x=0.75pt,y=0.75pt,yscale=-1,xscale=1]

\draw [line width=2.25]    (214.74,146.39) -- (165.69,188.22) ;
\draw [shift={(161.88,191.47)}, rotate = 319.53999999999996] [fill={rgb, 255:red, 0; green, 0; blue, 0 }  ][line width=0.08]  [draw opacity=0] (14.29,-6.86) -- (0,0) -- (14.29,6.86) -- cycle    ;
\draw [line width=2.25]    (220.11,167.44) -- (227.88,203.57) ;
\draw [shift={(219.06,162.55)}, rotate = 77.87] [fill={rgb, 255:red, 0; green, 0; blue, 0 }  ][line width=0.08]  [draw opacity=0] (14.29,-6.86) -- (0,0) -- (14.29,6.86) -- cycle    ;
\draw [line width=2.25]    (149.01,203.57) -- (149.01,138.9) ;
\draw [shift={(149.01,133.9)}, rotate = 450] [fill={rgb, 255:red, 0; green, 0; blue, 0 }  ][line width=0.08]  [draw opacity=0] (14.29,-6.86) -- (0,0) -- (14.29,6.86) -- cycle    ;
\draw [line width=2.25]    (227.88,203.57) -- (245.05,116.19) ;
\draw [shift={(246.01,111.28)}, rotate = 461.11] [fill={rgb, 255:red, 0; green, 0; blue, 0 }  ][line width=0.08]  [draw opacity=0] (14.29,-6.86) -- (0,0) -- (14.29,6.86) -- cycle    ;
\draw [line width=2.25]    (149.01,117.47) -- (193.45,136.89) ;
\draw [shift={(198.03,138.89)}, rotate = 203.6] [fill={rgb, 255:red, 0; green, 0; blue, 0 }  ][line width=0.08]  [draw opacity=0] (14.29,-6.86) -- (0,0) -- (14.29,6.86) -- cycle    ;
\draw [line width=2.25]    (149.01,203.57) -- (206.45,203.57) ;
\draw [shift={(211.45,203.57)}, rotate = 180] [fill={rgb, 255:red, 0; green, 0; blue, 0 }  ][line width=0.08]  [draw opacity=0] (14.29,-6.86) -- (0,0) -- (14.29,6.86) -- cycle    ;
\draw  [color={rgb, 255:red, 31; green, 119; blue, 180 }  ,draw opacity=1 ][fill={rgb, 255:red, 255; green, 255; blue, 255 }  ,fill opacity=1 ][line width=3]  (132.58,203.57) .. controls (132.58,194.49) and (139.94,187.14) .. (149.01,187.14) .. controls (158.09,187.14) and (165.44,194.49) .. (165.44,203.57) .. controls (165.44,212.64) and (158.09,220) .. (149.01,220) .. controls (139.94,220) and (132.58,212.64) .. (132.58,203.57) -- cycle ;
\draw  [color={rgb, 255:red, 31; green, 119; blue, 180 }  ,draw opacity=1 ][fill={rgb, 255:red, 255; green, 255; blue, 255 }  ,fill opacity=1 ][line width=3]  (132.58,117.47) .. controls (132.58,108.4) and (139.94,101.04) .. (149.01,101.04) .. controls (158.09,101.04) and (165.44,108.4) .. (165.44,117.47) .. controls (165.44,126.55) and (158.09,133.9) .. (149.01,133.9) .. controls (139.94,133.9) and (132.58,126.55) .. (132.58,117.47) -- cycle ;
\draw  [color={rgb, 255:red, 31; green, 119; blue, 180 }  ,draw opacity=1 ][fill={rgb, 255:red, 255; green, 255; blue, 255 }  ,fill opacity=1 ][line width=3]  (233.14,94.47) .. controls (233.14,85.4) and (240.49,78.04) .. (249.57,78.04) .. controls (258.64,78.04) and (266,85.4) .. (266,94.47) .. controls (266,103.54) and (258.64,110.9) .. (249.57,110.9) .. controls (240.49,110.9) and (233.14,103.54) .. (233.14,94.47) -- cycle ;
\draw  [color={rgb, 255:red, 31; green, 119; blue, 180 }  ,draw opacity=1 ][fill={rgb, 255:red, 255; green, 255; blue, 255 }  ,fill opacity=1 ][line width=3]  (211.45,203.57) .. controls (211.45,194.49) and (218.81,187.14) .. (227.88,187.14) .. controls (236.96,187.14) and (244.31,194.49) .. (244.31,203.57) .. controls (244.31,212.64) and (236.96,220) .. (227.88,220) .. controls (218.81,220) and (211.45,212.64) .. (211.45,203.57) -- cycle ;
\draw  [color={rgb, 255:red, 31; green, 119; blue, 180 }  ,draw opacity=1 ][fill={rgb, 255:red, 255; green, 255; blue, 255 }  ,fill opacity=1 ][line width=3]  (198.31,146.39) .. controls (198.31,137.32) and (205.66,129.96) .. (214.74,129.96) .. controls (223.81,129.96) and (231.17,137.32) .. (231.17,146.39) .. controls (231.17,155.47) and (223.81,162.82) .. (214.74,162.82) .. controls (205.66,162.82) and (198.31,155.47) .. (198.31,146.39) -- cycle ;
\draw [line width=2.25]    (473.74,146.39) -- (411.79,200.29) ;
\draw [shift={(408.01,203.57)}, rotate = 318.98] [fill={rgb, 255:red, 0; green, 0; blue, 0 }  ][line width=0.08]  [draw opacity=0] (14.29,-6.86) -- (0,0) -- (14.29,6.86) -- cycle    ;
\draw [line width=2.25]    (474.86,151.26) -- (486.88,203.57) ;
\draw [shift={(473.74,146.39)}, rotate = 77.05] [fill={rgb, 255:red, 0; green, 0; blue, 0 }  ][line width=0.08]  [draw opacity=0] (14.29,-6.86) -- (0,0) -- (14.29,6.86) -- cycle    ;
\draw [line width=2.25]    (408.01,203.57) -- (408.01,122.47) ;
\draw [shift={(408.01,117.47)}, rotate = 450] [fill={rgb, 255:red, 0; green, 0; blue, 0 }  ][line width=0.08]  [draw opacity=0] (14.29,-6.86) -- (0,0) -- (14.29,6.86) -- cycle    ;
\draw [line width=2.25]    (486.88,203.57) -- (507.59,99.37) ;
\draw [shift={(508.57,94.47)}, rotate = 461.24] [fill={rgb, 255:red, 0; green, 0; blue, 0 }  ][line width=0.08]  [draw opacity=0] (14.29,-6.86) -- (0,0) -- (14.29,6.86) -- cycle    ;
\draw [line width=2.25]    (408.01,117.47) -- (469.16,144.38) ;
\draw [shift={(473.74,146.39)}, rotate = 203.75] [fill={rgb, 255:red, 0; green, 0; blue, 0 }  ][line width=0.08]  [draw opacity=0] (14.29,-6.86) -- (0,0) -- (14.29,6.86) -- cycle    ;
\draw [line width=2.25]    (408.01,203.57) -- (481.88,203.57) ;
\draw [shift={(486.88,203.57)}, rotate = 180] [fill={rgb, 255:red, 0; green, 0; blue, 0 }  ][line width=0.08]  [draw opacity=0] (14.29,-6.86) -- (0,0) -- (14.29,6.86) -- cycle    ;
\draw  [color={rgb, 255:red, 31; green, 119; blue, 180 }  ,draw opacity=1 ][fill={rgb, 255:red, 255; green, 255; blue, 255 }  ,fill opacity=1 ][line width=3]  (391.58,203.57) .. controls (391.58,194.49) and (398.94,187.14) .. (408.01,187.14) .. controls (417.09,187.14) and (424.44,194.49) .. (424.44,203.57) .. controls (424.44,212.64) and (417.09,220) .. (408.01,220) .. controls (398.94,220) and (391.58,212.64) .. (391.58,203.57) -- cycle ;
\draw  [color={rgb, 255:red, 31; green, 119; blue, 180 }  ,draw opacity=1 ][fill={rgb, 255:red, 255; green, 255; blue, 255 }  ,fill opacity=1 ][line width=3]  (391.58,117.47) .. controls (391.58,108.4) and (398.94,101.04) .. (408.01,101.04) .. controls (417.09,101.04) and (424.44,108.4) .. (424.44,117.47) .. controls (424.44,126.55) and (417.09,133.9) .. (408.01,133.9) .. controls (398.94,133.9) and (391.58,126.55) .. (391.58,117.47) -- cycle ;
\draw  [color={rgb, 255:red, 31; green, 119; blue, 180 }  ,draw opacity=1 ][fill={rgb, 255:red, 255; green, 255; blue, 255 }  ,fill opacity=1 ][line width=3]  (492.14,94.47) .. controls (492.14,85.4) and (499.49,78.04) .. (508.57,78.04) .. controls (517.64,78.04) and (525,85.4) .. (525,94.47) .. controls (525,103.54) and (517.64,110.9) .. (508.57,110.9) .. controls (499.49,110.9) and (492.14,103.54) .. (492.14,94.47) -- cycle ;
\draw  [color={rgb, 255:red, 31; green, 119; blue, 180 }  ,draw opacity=1 ][fill={rgb, 255:red, 255; green, 255; blue, 255 }  ,fill opacity=1 ][line width=3]  (470.45,203.57) .. controls (470.45,194.49) and (477.81,187.14) .. (486.88,187.14) .. controls (495.96,187.14) and (503.31,194.49) .. (503.31,203.57) .. controls (503.31,212.64) and (495.96,220) .. (486.88,220) .. controls (477.81,220) and (470.45,212.64) .. (470.45,203.57) -- cycle ;
\draw  [color={rgb, 255:red, 31; green, 119; blue, 180 }  ,draw opacity=1 ][fill={rgb, 255:red, 255; green, 255; blue, 255 }  ,fill opacity=1 ][line width=3]  (457.31,146.39) .. controls (457.31,137.32) and (464.66,129.96) .. (473.74,129.96) .. controls (482.81,129.96) and (490.17,137.32) .. (490.17,146.39) .. controls (490.17,155.47) and (482.81,162.82) .. (473.74,162.82) .. controls (464.66,162.82) and (457.31,155.47) .. (457.31,146.39) -- cycle ;

\draw (126,231) node  [font=\LARGE] [align=left] {$\displaystyle v_{2}$};
\draw (136,87) node  [font=\LARGE] [align=left] {$\displaystyle v_{3}$};
\draw (278,84) node  [font=\LARGE] [align=left] {$\displaystyle v_{5}$};
\draw (253,230) node  [font=\LARGE] [align=left] {$\displaystyle v_{1}$};
\draw (211,115) node  [font=\LARGE] [align=left] {$\displaystyle v_{4}$};
\draw (385,231) node  [font=\LARGE] [align=left] {$\displaystyle v_{2}$};
\draw (395,87) node  [font=\LARGE] [align=left] {$\displaystyle v_{3}$};
\draw (537,84) node  [font=\LARGE] [align=left] {$\displaystyle v_{5}$};
\draw (512,230) node  [font=\LARGE] [align=left] {$\displaystyle v_{1}$};
\draw (470,115) node  [font=\LARGE] [align=left] {$\displaystyle v_{4}$};

\end{tikzpicture}}
    \caption{On the left, we present a simple example of a directed graph of size five, whereas its undirected counterpart is shown on the right.}
    \label{fig:graph}
\end{figure}

To express the notion of \quotes{direction} in the connections, we can distinguish between \textit{directed} and \textit{undirected} graphs.
\begin{definition}[\textbf{\textbf{Directed/Undirected Graph}}]
A graph $g = (\Vset{g},\Eset{g},\Xset{g},\Aset{g})$ is said to be directed when the edges are ordered, pairs of vertices \ie $\Eset{g} \subseteq \{\puv \mid u,v \in \Vset{g}\}$, and the edges are \textbf{oriented} with \textbf{tail} $u$ and \textbf{head} $v$. On the contrary, when the vertex pairs are not ordered, \ie $\Eset{g} \subseteq \{\suv \mid u,v \in \Vset{g}\}$, we talk about undirected graphs and \textbf{non-oriented} edges. In general, an edge connecting $u$ and $v$ is said to be \textbf{incident} to both vertices.
\end{definition}
An undirected graph is useful when, \eg representing molecules and mutual social interactions. Instead, a directed graph can be used to model road networks or hyperlinks, where the direction of the edge conveys additional information. Figure \ref{fig:graph} depicts two directed and undirected graphs, where the direction is graphically characterized by the arrow symbol. To avoid confusion with the graphical notation of a random variable, we will place the id of the vertex \textit{outside} the respective circle; when clear from the context, we may omit the vertex id in favor of a cleaner visualization.

Another fundamental concept is that of degree of a vertex; we first provide its definition for directed graphs.
\begin{definition}[\textbf{Degree}]
Let $g$ be a directed graph. The \textbf{in-degree} of a vertex $u \in \Vset{g}$ is defined as the number of ordered edges with head $u$, that is, $indegree(u) = |\{v \mid \pvu \in \Eset{g}\}|$. In contrast, the \textbf{out-degree} of a vertex $u$ is given by the number of ordered edges with tail $u$, \ie $outdegree(u) = |\{v \mid \puv \in \Eset{g}\}|$.
\end{definition}
For most of the manuscript, we will implicitly work with the \textbf{in-degree} of a vertex $u$, denoting it as $deg(u)$ to simplify the notation. It follows that for undirected graphs there is no distinction between in-degree and out-degree.

As an alternative to $\Eset{g}$, the structural information of a graph can be encoded by its square adjacency matrix.
\begin{definition}[\textbf{\textbf{Adjacency Matrix}}]
The adjacency matrix of a graph $g$ is a binary square matrix $\mathbf{A} \in \{0,1\}^{|\Vset{g}| \times |\Vset{g}|}$ where each entry $A_\uv$ is 1 if an edge links $u$ and $v$ together and 0 otherwise. In the case of undirected graphs, their adjacency matrix is symmetric. Moreover, whenever edge features are scalars, we can encode this information in a \textbf{weighted adjacency matrix} with $A_\uv = a_\uv, a_\uv \in \R{}$.
\end{definition}

The out-degree of vertex $u$ can be extracted from the adjacency matrix by computing the sum of the values on the $u$-th row $A_{u,:}$, while the in-degree requires to sum over all values on the $u$-th column $A_{:,u}$. From the adjacency matrix, we can also construct the corresponding Laplacian matrix:

\begin{definition}[\textbf{Symmetric Normalized Laplacian Matrix}]
Let $g$ be a graph with adjacency matrix $A$, and let $D$ be the diagonal degree matrix with entries $D_{uu}=deg(u)$. Then, the symmetric normalized Laplacian matrix of $g$ is a square matrix defined as
\begin{align*}
    & L^{sym} = D^{-\frac{1}{2}}LD^{-\frac{1}{2}} = I - D^{-\frac{1}{2}}AD^{-\frac{1}{2}} \\
    & \text{where} \ \  L = D - A \ \ \text{is the unnormalized Laplacian matrix}.
\end{align*}
\end{definition}
The symmetric normalized Laplacian has a great deal of \quotes{good} properties that, especially in the case of undirected graphs, come in handy when studying its properties and \ML{} models via spectral graph theory (see section \ref{sec:spectral-graph-theory}).

We now introduce the concept of a cycle. Cycles are particularly insidious for deep learning models that aim at automatically extracting features from graphs; the reason will become clear in the next chapter.
\begin{definition}[\textbf{Cycle}]
Let us informally define a \textbf{path} as a sequence of distinct edges that allow us to move from vertex $u$ to $v$ in the corresponding graph. A graph cycle occurs when there exists a non-empty path from $u$ to itself with no repeated edges. If a graph contains no cycles, then it is called \textbf{acyclic}.
\end{definition}
Instances of cyclic graphs have already been provided in Figure \ref{fig:graph}. Moreover, a graph is called \textbf{connected} if there exists a path from each vertex to another. Strictly related to cycles, another fundamental challenge when learning from graphs is the absence of a \textbf{consistent} topological ordering of the vertices across the dataset.

\begin{figure}[ht]
    \centering
    \resizebox{0.8\textwidth}{!}{\tikzset{every picture/.style={line width=0.75pt}} 

\begin{tikzpicture}[x=0.75pt,y=0.75pt,yscale=-1,xscale=1]

\draw [line width=2.25]    (247.57,111.47) -- (205.45,161.86) ;
\draw [shift={(202.25,165.7)}, rotate = 309.89] [fill={rgb, 255:red, 0; green, 0; blue, 0 }  ][line width=0.08]  [draw opacity=0] (14.29,-6.86) -- (0,0) -- (14.29,6.86) -- cycle    ;
\draw [line width=2.25]    (135.01,110.47) -- (175.16,161.76) ;
\draw [shift={(178.25,165.7)}, rotate = 231.95] [fill={rgb, 255:red, 0; green, 0; blue, 0 }  ][line width=0.08]  [draw opacity=0] (14.29,-6.86) -- (0,0) -- (14.29,6.86) -- cycle    ;
\draw [line width=2.25]    (190.01,43.47) -- (149.06,92.27) ;
\draw [shift={(145.85,96.1)}, rotate = 310.01] [fill={rgb, 255:red, 0; green, 0; blue, 0 }  ][line width=0.08]  [draw opacity=0] (14.29,-6.86) -- (0,0) -- (14.29,6.86) -- cycle    ;
\draw [line width=2.25]    (190.01,43.47) -- (232.99,93.51) ;
\draw [shift={(236.25,97.3)}, rotate = 229.34] [fill={rgb, 255:red, 0; green, 0; blue, 0 }  ][line width=0.08]  [draw opacity=0] (14.29,-6.86) -- (0,0) -- (14.29,6.86) -- cycle    ;
\draw  [color={rgb, 255:red, 31; green, 119; blue, 180 }  ,draw opacity=1 ][fill={rgb, 255:red, 255; green, 255; blue, 255 }  ,fill opacity=1 ][line width=3]  (118.58,110.47) .. controls (118.58,101.4) and (125.94,94.04) .. (135.01,94.04) .. controls (144.09,94.04) and (151.44,101.4) .. (151.44,110.47) .. controls (151.44,119.54) and (144.09,126.9) .. (135.01,126.9) .. controls (125.94,126.9) and (118.58,119.54) .. (118.58,110.47) -- cycle ;
\draw  [color={rgb, 255:red, 31; green, 119; blue, 180 }  ,draw opacity=1 ][fill={rgb, 255:red, 255; green, 255; blue, 255 }  ,fill opacity=1 ][line width=3]  (173.58,43.47) .. controls (173.58,34.4) and (180.94,27.04) .. (190.01,27.04) .. controls (199.09,27.04) and (206.44,34.4) .. (206.44,43.47) .. controls (206.44,52.55) and (199.09,59.9) .. (190.01,59.9) .. controls (180.94,59.9) and (173.58,52.55) .. (173.58,43.47) -- cycle ;
\draw  [color={rgb, 255:red, 31; green, 119; blue, 180 }  ,draw opacity=1 ][fill={rgb, 255:red, 255; green, 255; blue, 255 }  ,fill opacity=1 ][line width=3]  (231.14,110.47) .. controls (231.14,101.4) and (238.49,94.04) .. (247.57,94.04) .. controls (256.64,94.04) and (264,101.4) .. (264,110.47) .. controls (264,119.54) and (256.64,126.9) .. (247.57,126.9) .. controls (238.49,126.9) and (231.14,119.54) .. (231.14,110.47) -- cycle ;
\draw  [color={rgb, 255:red, 31; green, 119; blue, 180 }  ,draw opacity=1 ][fill={rgb, 255:red, 255; green, 255; blue, 255 }  ,fill opacity=1 ][line width=3]  (173.58,179.47) .. controls (173.58,170.4) and (180.94,163.04) .. (190.01,163.04) .. controls (199.09,163.04) and (206.44,170.4) .. (206.44,179.47) .. controls (206.44,188.55) and (199.09,195.9) .. (190.01,195.9) .. controls (180.94,195.9) and (173.58,188.55) .. (173.58,179.47) -- cycle ;
\draw [line width=2.25]    (503.57,111.47) -- (461.45,161.86) ;
\draw [shift={(458.25,165.7)}, rotate = 309.89] [fill={rgb, 255:red, 0; green, 0; blue, 0 }  ][line width=0.08]  [draw opacity=0] (14.29,-6.86) -- (0,0) -- (14.29,6.86) -- cycle    ;
\draw [line width=2.25]    (391.01,110.47) -- (431.16,161.76) ;
\draw [shift={(434.25,165.7)}, rotate = 231.95] [fill={rgb, 255:red, 0; green, 0; blue, 0 }  ][line width=0.08]  [draw opacity=0] (14.29,-6.86) -- (0,0) -- (14.29,6.86) -- cycle    ;
\draw [line width=2.25]    (446.01,43.47) -- (405.06,92.27) ;
\draw [shift={(401.85,96.1)}, rotate = 310.01] [fill={rgb, 255:red, 0; green, 0; blue, 0 }  ][line width=0.08]  [draw opacity=0] (14.29,-6.86) -- (0,0) -- (14.29,6.86) -- cycle    ;
\draw [line width=2.25]    (446.01,43.47) -- (488.99,93.51) ;
\draw [shift={(492.25,97.3)}, rotate = 229.34] [fill={rgb, 255:red, 0; green, 0; blue, 0 }  ][line width=0.08]  [draw opacity=0] (14.29,-6.86) -- (0,0) -- (14.29,6.86) -- cycle    ;
\draw  [color={rgb, 255:red, 31; green, 119; blue, 180 }  ,draw opacity=1 ][fill={rgb, 255:red, 255; green, 255; blue, 255 }  ,fill opacity=1 ][line width=3]  (374.58,110.47) .. controls (374.58,101.4) and (381.94,94.04) .. (391.01,94.04) .. controls (400.09,94.04) and (407.44,101.4) .. (407.44,110.47) .. controls (407.44,119.54) and (400.09,126.9) .. (391.01,126.9) .. controls (381.94,126.9) and (374.58,119.54) .. (374.58,110.47) -- cycle ;
\draw  [color={rgb, 255:red, 31; green, 119; blue, 180 }  ,draw opacity=1 ][fill={rgb, 255:red, 255; green, 255; blue, 255 }  ,fill opacity=1 ][line width=3]  (429.58,43.47) .. controls (429.58,34.4) and (436.94,27.04) .. (446.01,27.04) .. controls (455.09,27.04) and (462.44,34.4) .. (462.44,43.47) .. controls (462.44,52.55) and (455.09,59.9) .. (446.01,59.9) .. controls (436.94,59.9) and (429.58,52.55) .. (429.58,43.47) -- cycle ;
\draw  [color={rgb, 255:red, 31; green, 119; blue, 180 }  ,draw opacity=1 ][fill={rgb, 255:red, 255; green, 255; blue, 255 }  ,fill opacity=1 ][line width=3]  (487.14,110.47) .. controls (487.14,101.4) and (494.49,94.04) .. (503.57,94.04) .. controls (512.64,94.04) and (520,101.4) .. (520,110.47) .. controls (520,119.54) and (512.64,126.9) .. (503.57,126.9) .. controls (494.49,126.9) and (487.14,119.54) .. (487.14,110.47) -- cycle ;
\draw  [color={rgb, 255:red, 31; green, 119; blue, 180 }  ,draw opacity=1 ][fill={rgb, 255:red, 255; green, 255; blue, 255 }  ,fill opacity=1 ][line width=3]  (429.58,179.47) .. controls (429.58,170.4) and (436.94,163.04) .. (446.01,163.04) .. controls (455.09,163.04) and (462.44,170.4) .. (462.44,179.47) .. controls (462.44,188.55) and (455.09,195.9) .. (446.01,195.9) .. controls (436.94,195.9) and (429.58,188.55) .. (429.58,179.47) -- cycle ;

\draw (160.84,54.28) node  [font=\LARGE] [align=left] {$\displaystyle 1$};
\draw (220.44,53.68) node  [font=\LARGE] [align=left] {$\displaystyle 2$};
\draw (143.04,142.48) node  [font=\LARGE] [align=left] {$\displaystyle 1$};
\draw (240.24,142.27) node  [font=\LARGE] [align=left] {$\displaystyle 1$};
\draw (415.51,54.08) node  [font=\LARGE] [align=left] {$\displaystyle 1$};
\draw (476.15,54.41) node  [font=\LARGE] [align=left] {$\displaystyle 2$};
\draw (397.64,141.14) node  [font=\LARGE] [align=left] {$\displaystyle 2$};
\draw (497.71,141.34) node  [font=\LARGE] [align=left] {$\displaystyle 1$};

\end{tikzpicture}}
    \caption{A Directed Ordered Acyclic Graph is sketched on the left, whereas a Directed Positional Acyclic Graph is on the right. For simplicity of exposition, we have defined total orders for outgoing edges only.}
    \label{fig:doag-dpag}
\end{figure}
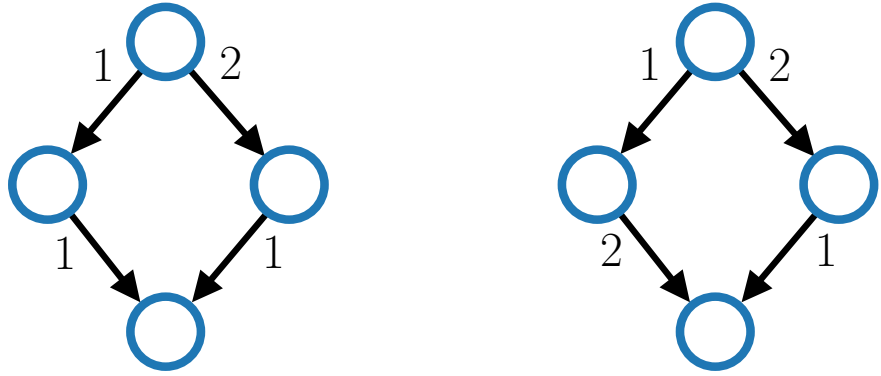

\begin{definition}[\textbf{Topological Ordering}]
A topological ordering of a \textbf{directed} graph $g$ is a total order over its vertices such that, for every oriented edge $\puv \in \Eset{g}$, $u$ comes before $v$ in such ordering.
\end{definition}
Crucially, there exists a topological ordering of the directed graph if and only if it is acyclic. In this case, we talk about Directed Acyclic Graphs (DAGs). In addition, we say a graph is \textbf{ordered} if there exists a total order on the edges incident to each vertex (\textbf{unordered} otherwise) \cite{frasconi_general_1998}. An example of an ordered graph is the Directed Ordered Acyclic Graph (DOAG) of Figure \ref{fig:doag-dpag} (left). Similarly, a \textbf{positional} graph is an ordered graph with bounded in-degree and out-degree for which there exist two injective functions mapping edges that enter or leave a vertex to a distinctive positive integer (\textbf{non-positional} otherwise). The main difference between ordered and positional graphs is that, in the latter, some positions are allowed to be absent. We sketch a Directed Positional Acyclic Graph (DPAG) on the right handside of Figure \ref{fig:doag-dpag}.

\paragraph*{Remark.} From now on, we will assume to work with the very general class of (un)directed, (a)cyclic and (non-)positional graphs. To cope with the methodologies presented here, whenever an undirected graph is provided as input it is simply transformed into its directed counterpart. In particular, every edge $\suv$ is converted into two \textit{distinct}, \textit{opposite} and \textit{oriented} arcs $\puv$ and $\pvu$; if an edge feature $\bm{a}_\uv$ is present in the original graph, this gets copied into both $\bm{a}_\uv$ and $\bm{a}_\vu$.

In the field of deep learning for graphs, another important notion is that of a vertex neighborhood. Intuitively, the neighborhood defines the \textbf{local view} of each vertex.

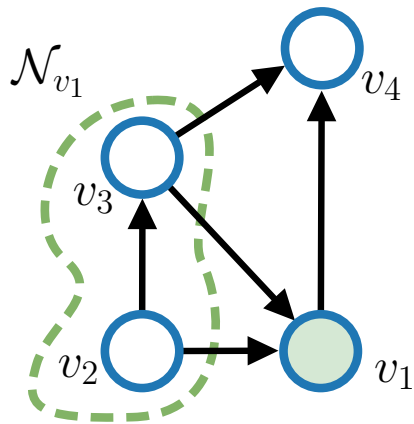
\begin{figure}[ht]
    \centering
    \resizebox{0.4\textwidth}{!}{\tikzset{every picture/.style={line width=0.75pt}} 

\begin{tikzpicture}[x=0.75pt,y=0.75pt,yscale=-1,xscale=1]

\draw  [color={rgb, 255:red, 130; green, 179; blue, 102 }  ,draw opacity=1 ][dash pattern={on 7.88pt off 4.5pt}][line width=3]  (144.21,56.38) .. controls (192.46,54.31) and (157.31,114.27) .. (166.96,132.19) .. controls (176.61,150.11) and (181.19,202.99) .. (134.86,204.32) .. controls (88.53,205.66) and (84.53,199.32) .. (90.33,179.58) .. controls (96.14,159.84) and (123.12,147.4) .. (103.19,130.66) .. controls (83.27,113.92) and (95.97,58.44) .. (144.21,56.38) -- cycle ;
\draw [line width=2.25]    (139.18,173.52) -- (139.32,107.32) ;
\draw [shift={(139.33,102.32)}, rotate = 450.12] [fill={rgb, 255:red, 0; green, 0; blue, 0 }  ][line width=0.08]  [draw opacity=0] (14.29,-6.86) -- (0,0) -- (14.29,6.86) -- cycle    ;
\draw [line width=2.25]    (206.59,155.66) -- (139.18,83.23) ;
\draw [shift={(210,159.32)}, rotate = 227.05] [fill={rgb, 255:red, 0; green, 0; blue, 0 }  ][line width=0.08]  [draw opacity=0] (14.29,-6.86) -- (0,0) -- (14.29,6.86) -- cycle    ;
\draw [line width=2.25]    (221.89,173.52) -- (222.32,56.99) ;
\draw [shift={(222.33,51.99)}, rotate = 450.21] [fill={rgb, 255:red, 0; green, 0; blue, 0 }  ][line width=0.08]  [draw opacity=0] (14.29,-6.86) -- (0,0) -- (14.29,6.86) -- cycle    ;
\draw [line width=2.25]    (139.18,83.23) -- (201.14,43.04) ;
\draw [shift={(205.33,40.32)}, rotate = 507.03] [fill={rgb, 255:red, 0; green, 0; blue, 0 }  ][line width=0.08]  [draw opacity=0] (14.29,-6.86) -- (0,0) -- (14.29,6.86) -- cycle    ;
\draw [line width=2.25]    (139.18,173.52) -- (197.33,173.34) ;
\draw [shift={(202.33,173.32)}, rotate = 539.8199999999999] [fill={rgb, 255:red, 0; green, 0; blue, 0 }  ][line width=0.08]  [draw opacity=0] (14.29,-6.86) -- (0,0) -- (14.29,6.86) -- cycle    ;
\draw  [color={rgb, 255:red, 31; green, 119; blue, 180 }  ,draw opacity=1 ][fill={rgb, 255:red, 255; green, 255; blue, 255 }  ,fill opacity=1 ][line width=3]  (121.95,173.52) .. controls (121.95,164.01) and (129.66,156.29) .. (139.18,156.29) .. controls (148.7,156.29) and (156.41,164.01) .. (156.41,173.52) .. controls (156.41,183.04) and (148.7,190.75) .. (139.18,190.75) .. controls (129.66,190.75) and (121.95,183.04) .. (121.95,173.52) -- cycle ;
\draw  [color={rgb, 255:red, 31; green, 119; blue, 180 }  ,draw opacity=1 ][fill={rgb, 255:red, 255; green, 255; blue, 255 }  ,fill opacity=1 ][line width=3]  (121.95,83.23) .. controls (121.95,73.72) and (129.66,66) .. (139.18,66) .. controls (148.7,66) and (156.41,73.72) .. (156.41,83.23) .. controls (156.41,92.75) and (148.7,100.47) .. (139.18,100.47) .. controls (129.66,100.47) and (121.95,92.75) .. (121.95,83.23) -- cycle ;
\draw  [color={rgb, 255:red, 31; green, 119; blue, 180 }  ,draw opacity=1 ][fill={rgb, 255:red, 255; green, 255; blue, 255 }  ,fill opacity=1 ][line width=3]  (205.35,32.23) .. controls (205.35,22.71) and (213.06,15) .. (222.58,15) .. controls (232.09,15) and (239.81,22.71) .. (239.81,32.23) .. controls (239.81,41.75) and (232.09,49.46) .. (222.58,49.46) .. controls (213.06,49.46) and (205.35,41.75) .. (205.35,32.23) -- cycle ;
\draw  [color={rgb, 255:red, 31; green, 119; blue, 180 }  ,draw opacity=1 ][fill={rgb, 255:red, 213; green, 232; blue, 212 }  ,fill opacity=1 ][line width=3]  (204.66,173.52) .. controls (204.66,164.01) and (212.37,156.29) .. (221.89,156.29) .. controls (231.41,156.29) and (239.12,164.01) .. (239.12,173.52) .. controls (239.12,183.04) and (231.41,190.75) .. (221.89,190.75) .. controls (212.37,190.75) and (204.66,183.04) .. (204.66,173.52) -- cycle ;

\draw (256.35,184.55) node  [font=\LARGE] [align=left] {$\displaystyle v_{1}$};
\draw (250.11,49.78) node  [font=\LARGE] [align=left] {$\displaystyle v_{4}$};
\draw (116.85,102.23) node  [font=\LARGE] [align=left] {$\displaystyle v_{3}$};
\draw (109.85,182.19) node  [font=\LARGE] [align=left] {$\displaystyle v_{2}$};
\draw (96.33,44.04) node  [font=\LARGE] [align=left] {$\displaystyle \mathcal{N}_{v_{1}}$};

\end{tikzpicture}}
    \caption{The neighborhood of vertex $v_1$ is here drawn as the set of vertices belonging to the dashed green region.}
    \label{fig:neighborhood}
\end{figure}

\begin{definition}[\textbf{Neighborhood}]
Given a directed graph $g$ and a vertex $u \in \Vset{g}$, the neighborhood of $u$ is the set of vertices connected to $u$ with an ordered edge:
\begin{align*}
    \N{u}{} = \{v \in \Vset{g} \mid \pvu \in \Eset{g}\}.
\end{align*}
The neighborhood of $u$ is \textbf{closed} if it always includes $u$ and \textbf{open} otherwise. Whenever the image of $\Aset{g}$ is the finite and discrete set $\{c_1,\dots,c_n\}$, we shall extend our notation to define the subset of those neighbors that are connected to $u$ with an arc labeled as $c_k$: $\N{u}{c_k} = \{v \in \N{u}{} \mid \bm{a}_\vu = c_k\}$.
\end{definition}
Figure \ref{fig:neighborhood} provides a simple visual depiction of a vertex neighborhood.

Furthermore, let us formally define when two graphs are structurally equivalent, which can be related to the expressive power of \ML{} models for structured data.

\begin{definition}[\textbf{Isomorphism}]
Two graphs $g'$ and $g''$ are isomorphic (ignoring their vertex and edge features) if there is a bijections $f:\Vset{g'} \rightarrow \Vset{g''}$ such that two vertices $u,v$ are adjacent if and only if $f(u),f(v)$ are adjacent.
\end{definition}

We conclude this first part by informally introducing the reader to the notion of \textbf{structural transductions}, and we refer to \cite{frasconi_general_1998} for a detailed mathematical treatment. A transduction is, in general, a binary relation between two spaces $\mathcal{U}$ and $\mathcal{Y}$, but we will restrict ourselves to functions $\T:\mathcal{U} \rightarrow \mathcal{Y}$. Now assume that both $\mathcal{U}$ and $\mathcal{Y}$ are \textit{structured} spaces, \eg they both represent the set of possible graphs, and define the skeleton of a graph $g$, namely $\text{skel}(g)$, as the graph obtained by discarding all vertex or edge labels. We say that a transduction $\T(\cdot)$ is \textbf{IO-isomorph} if it holds
\begin{align*}
    \text{skel}(\T(g)) = \text{skel}(g) \ \ \forall g \in \mathcal{U}.
\end{align*}
As we shall see in the next chapter, IO-isomorph transductions are central to most deep learning architectures for graphs.

\subsection{Instances of a Graph}
Depending on the constraints posed on the connections of a graph, we obtain very specific families of structures.

\begin{definition}[\textbf{Sequence}]
A sequence is a connected acyclic graph in which vertices are adjacent if and only if they are consecutive in the topological ordering induced on the graph.
\end{definition}

\begin{definition}[\textbf{Tree}]
A tree is a connected acyclic graph in which any two vertices are connected by exactly one path.
\end{definition}
Because of their flexibility, trees have been successfully used in natural language processing and chemistry to model, for example, syntactic dependencies in sentences and molecules. 

\begin{figure}[ht]
    \centering
    \resizebox{1\textwidth}{!}{\tikzset{every picture/.style={line width=0.75pt}} 

\begin{tikzpicture}[x=0.75pt,y=0.75pt,yscale=-1,xscale=1]

\draw [line width=2.25]    (391.01,173.47) -- (345.01,249.47) ;
\draw [line width=2.25]    (113.58,144.47) -- (23.01,144.47) ;
\draw [shift={(118.58,144.47)}, rotate = 180] [fill={rgb, 255:red, 0; green, 0; blue, 0 }  ][line width=0.08]  [draw opacity=0] (14.29,-6.86) -- (0,0) -- (14.29,6.86) -- cycle    ;
\draw [line width=2.25]    (226.14,144.47) -- (135.01,144.47) ;
\draw [shift={(231.14,144.47)}, rotate = 180] [fill={rgb, 255:red, 0; green, 0; blue, 0 }  ][line width=0.08]  [draw opacity=0] (14.29,-6.86) -- (0,0) -- (14.29,6.86) -- cycle    ;
\draw  [color={rgb, 255:red, 31; green, 119; blue, 180 }  ,draw opacity=1 ][fill={rgb, 255:red, 255; green, 255; blue, 255 }  ,fill opacity=1 ][line width=3]  (118.58,144.47) .. controls (118.58,135.4) and (125.94,128.04) .. (135.01,128.04) .. controls (144.09,128.04) and (151.44,135.4) .. (151.44,144.47) .. controls (151.44,153.54) and (144.09,160.9) .. (135.01,160.9) .. controls (125.94,160.9) and (118.58,153.54) .. (118.58,144.47) -- cycle ;
\draw  [color={rgb, 255:red, 31; green, 119; blue, 180 }  ,draw opacity=1 ][fill={rgb, 255:red, 255; green, 255; blue, 255 }  ,fill opacity=1 ][line width=3]  (231.14,144.47) .. controls (231.14,135.4) and (238.49,128.04) .. (247.57,128.04) .. controls (256.64,128.04) and (264,135.4) .. (264,144.47) .. controls (264,153.54) and (256.64,160.9) .. (247.57,160.9) .. controls (238.49,160.9) and (231.14,153.54) .. (231.14,144.47) -- cycle ;
\draw [line width=2.25]    (382.78,167.24) -- (446.01,248.47) ;
\draw [line width=2.25]    (391.01,173.47) -- (391.01,111.47) ;
\draw [line width=2.25]    (446.01,44.47) -- (391.01,111.47) ;
\draw [line width=2.25]    (446.01,44.47) -- (503.57,111.47) ;
\draw  [color={rgb, 255:red, 31; green, 119; blue, 180 }  ,draw opacity=1 ][fill={rgb, 255:red, 255; green, 255; blue, 255 }  ,fill opacity=1 ][line width=3]  (374.58,111.47) .. controls (374.58,102.4) and (381.94,95.04) .. (391.01,95.04) .. controls (400.09,95.04) and (407.44,102.4) .. (407.44,111.47) .. controls (407.44,120.54) and (400.09,127.9) .. (391.01,127.9) .. controls (381.94,127.9) and (374.58,120.54) .. (374.58,111.47) -- cycle ;
\draw  [color={rgb, 255:red, 31; green, 119; blue, 180 }  ,draw opacity=1 ][fill={rgb, 255:red, 255; green, 255; blue, 255 }  ,fill opacity=1 ][line width=3]  (429.58,44.47) .. controls (429.58,35.4) and (436.94,28.04) .. (446.01,28.04) .. controls (455.09,28.04) and (462.44,35.4) .. (462.44,44.47) .. controls (462.44,53.55) and (455.09,60.9) .. (446.01,60.9) .. controls (436.94,60.9) and (429.58,53.55) .. (429.58,44.47) -- cycle ;
\draw  [color={rgb, 255:red, 31; green, 119; blue, 180 }  ,draw opacity=1 ][fill={rgb, 255:red, 255; green, 255; blue, 255 }  ,fill opacity=1 ][line width=3]  (487.14,111.47) .. controls (487.14,102.4) and (494.49,95.04) .. (503.57,95.04) .. controls (512.64,95.04) and (520,102.4) .. (520,111.47) .. controls (520,120.54) and (512.64,127.9) .. (503.57,127.9) .. controls (494.49,127.9) and (487.14,120.54) .. (487.14,111.47) -- cycle ;
\draw  [color={rgb, 255:red, 31; green, 119; blue, 180 }  ,draw opacity=1 ][fill={rgb, 255:red, 255; green, 255; blue, 255 }  ,fill opacity=1 ][line width=3]  (374.58,173.47) .. controls (374.58,164.4) and (381.94,157.04) .. (391.01,157.04) .. controls (400.09,157.04) and (407.44,164.4) .. (407.44,173.47) .. controls (407.44,182.54) and (400.09,189.9) .. (391.01,189.9) .. controls (381.94,189.9) and (374.58,182.54) .. (374.58,173.47) -- cycle ;
\draw  [color={rgb, 255:red, 31; green, 119; blue, 180 }  ,draw opacity=1 ][fill={rgb, 255:red, 255; green, 255; blue, 255 }  ,fill opacity=1 ][line width=3]  (429.58,248.47) .. controls (429.58,239.4) and (436.94,232.04) .. (446.01,232.04) .. controls (455.09,232.04) and (462.44,239.4) .. (462.44,248.47) .. controls (462.44,257.55) and (455.09,264.9) .. (446.01,264.9) .. controls (436.94,264.9) and (429.58,257.55) .. (429.58,248.47) -- cycle ;
\draw  [color={rgb, 255:red, 31; green, 119; blue, 180 }  ,draw opacity=1 ][fill={rgb, 255:red, 255; green, 255; blue, 255 }  ,fill opacity=1 ][line width=3]  (6.58,144.47) .. controls (6.58,135.4) and (13.94,128.04) .. (23.01,128.04) .. controls (32.09,128.04) and (39.44,135.4) .. (39.44,144.47) .. controls (39.44,153.54) and (32.09,160.9) .. (23.01,160.9) .. controls (13.94,160.9) and (6.58,153.54) .. (6.58,144.47) -- cycle ;
\draw  [color={rgb, 255:red, 31; green, 119; blue, 180 }  ,draw opacity=1 ][fill={rgb, 255:red, 255; green, 255; blue, 255 }  ,fill opacity=1 ][line width=3]  (328.58,249.47) .. controls (328.58,240.4) and (335.94,233.04) .. (345.01,233.04) .. controls (354.09,233.04) and (361.44,240.4) .. (361.44,249.47) .. controls (361.44,258.55) and (354.09,265.9) .. (345.01,265.9) .. controls (335.94,265.9) and (328.58,258.55) .. (328.58,249.47) -- cycle ;

\draw (53.35,162.55) node  [font=\LARGE] [align=left] {$\displaystyle v_{1}$};
\draw (166.35,162.55) node  [font=\LARGE] [align=left] {$\displaystyle v_{2}$};
\draw (278.35,163.55) node  [font=\LARGE] [align=left] {$\displaystyle v_{3}$};
\draw (484.35,50.55) node  [font=\LARGE] [align=left] {$\displaystyle v_{1}$};
\draw (537.35,126.55) node  [font=\LARGE] [align=left] {$\displaystyle v_{3}$};
\draw (426.35,126.55) node  [font=\LARGE] [align=left] {$\displaystyle v_{2}$};
\draw (425.35,187.55) node  [font=\LARGE] [align=left] {$\displaystyle v_{4}$};
\draw (480.35,265.55) node  [font=\LARGE] [align=left] {$\displaystyle v_{6}$};
\draw (379.35,265.55) node  [font=\LARGE] [align=left] {$\displaystyle v_{5}$};

\end{tikzpicture}}
    \caption{We sketch an directed sequence on the left and an undirected tree on the right.}
    \label{fig:sequence-tree}
\end{figure}
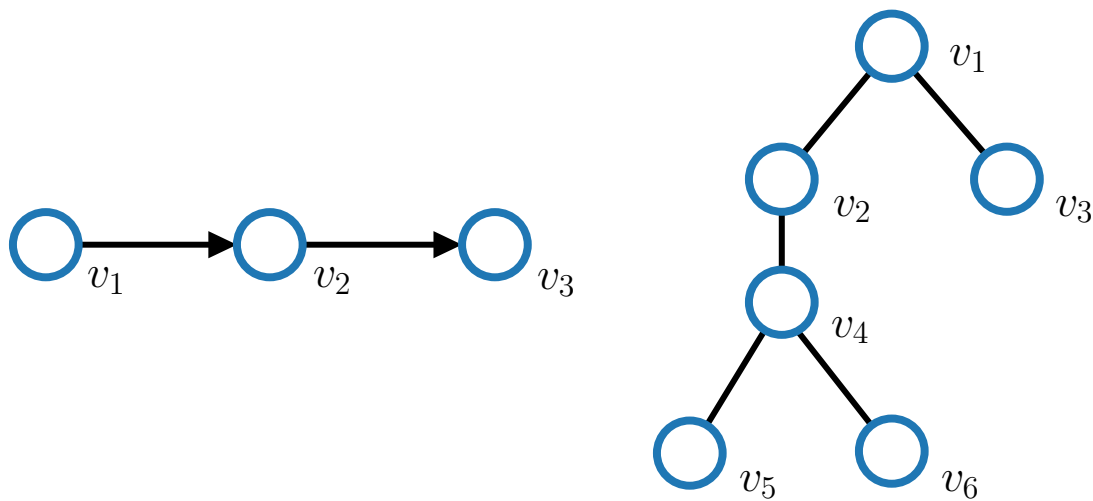

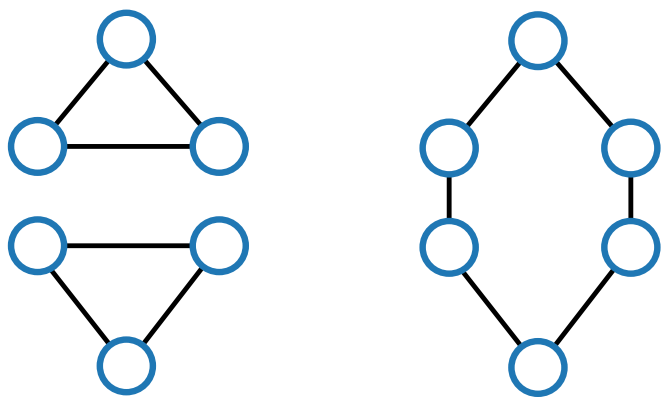
\begin{figure}[ht]
    \centering
    \resizebox{0.6\textwidth}{!}{\tikzset{every picture/.style={line width=0.75pt}} 

\begin{tikzpicture}[x=0.75pt,y=0.75pt,yscale=-1,xscale=1]

\draw [line width=2.25]    (502.57,111.47) -- (502.57,168.47) ;
\draw [shift={(502.57,173.47)}, rotate = 270] [fill={rgb, 255:red, 0; green, 0; blue, 0 }  ][line width=0.08]  [draw opacity=0] (14.29,-6.86) -- (0,0) -- (14.29,6.86) -- cycle    ;
\draw [line width=2.25]    (126.78,166.24) -- (186.94,243.53) ;
\draw [shift={(190.01,247.47)}, rotate = 232.1] [fill={rgb, 255:red, 0; green, 0; blue, 0 }  ][line width=0.08]  [draw opacity=0] (14.29,-6.86) -- (0,0) -- (14.29,6.86) -- cycle    ;
\draw [line width=2.25]    (247.57,110.47) -- (140.01,110.47) ;
\draw [shift={(135.01,110.47)}, rotate = 360] [fill={rgb, 255:red, 0; green, 0; blue, 0 }  ][line width=0.08]  [draw opacity=0] (14.29,-6.86) -- (0,0) -- (14.29,6.86) -- cycle    ;
\draw [line width=2.25]    (190.01,43.47) -- (138.19,106.61) ;
\draw [shift={(135.01,110.47)}, rotate = 309.38] [fill={rgb, 255:red, 0; green, 0; blue, 0 }  ][line width=0.08]  [draw opacity=0] (14.29,-6.86) -- (0,0) -- (14.29,6.86) -- cycle    ;
\draw [line width=2.25]    (190.01,43.47) -- (244.31,106.68) ;
\draw [shift={(247.57,110.47)}, rotate = 229.34] [fill={rgb, 255:red, 0; green, 0; blue, 0 }  ][line width=0.08]  [draw opacity=0] (14.29,-6.86) -- (0,0) -- (14.29,6.86) -- cycle    ;
\draw  [color={rgb, 255:red, 31; green, 119; blue, 180 }  ,draw opacity=1 ][fill={rgb, 255:red, 255; green, 255; blue, 255 }  ,fill opacity=1 ][line width=3]  (118.58,110.47) .. controls (118.58,101.4) and (125.94,94.04) .. (135.01,94.04) .. controls (144.09,94.04) and (151.44,101.4) .. (151.44,110.47) .. controls (151.44,119.54) and (144.09,126.9) .. (135.01,126.9) .. controls (125.94,126.9) and (118.58,119.54) .. (118.58,110.47) -- cycle ;
\draw  [color={rgb, 255:red, 31; green, 119; blue, 180 }  ,draw opacity=1 ][fill={rgb, 255:red, 255; green, 255; blue, 255 }  ,fill opacity=1 ][line width=3]  (173.58,43.47) .. controls (173.58,34.4) and (180.94,27.04) .. (190.01,27.04) .. controls (199.09,27.04) and (206.44,34.4) .. (206.44,43.47) .. controls (206.44,52.55) and (199.09,59.9) .. (190.01,59.9) .. controls (180.94,59.9) and (173.58,52.55) .. (173.58,43.47) -- cycle ;
\draw  [color={rgb, 255:red, 31; green, 119; blue, 180 }  ,draw opacity=1 ][fill={rgb, 255:red, 255; green, 255; blue, 255 }  ,fill opacity=1 ][line width=3]  (231.14,110.47) .. controls (231.14,101.4) and (238.49,94.04) .. (247.57,94.04) .. controls (256.64,94.04) and (264,101.4) .. (264,110.47) .. controls (264,119.54) and (256.64,126.9) .. (247.57,126.9) .. controls (238.49,126.9) and (231.14,119.54) .. (231.14,110.47) -- cycle ;
\draw [line width=2.25]    (247.57,172.47) -- (193.06,243.51) ;
\draw [shift={(190.01,247.47)}, rotate = 307.5] [fill={rgb, 255:red, 0; green, 0; blue, 0 }  ][line width=0.08]  [draw opacity=0] (14.29,-6.86) -- (0,0) -- (14.29,6.86) -- cycle    ;
\draw [line width=2.25]    (135.01,172.47) -- (242.57,172.47) ;
\draw [shift={(247.57,172.47)}, rotate = 180] [fill={rgb, 255:red, 0; green, 0; blue, 0 }  ][line width=0.08]  [draw opacity=0] (14.29,-6.86) -- (0,0) -- (14.29,6.86) -- cycle    ;
\draw  [color={rgb, 255:red, 31; green, 119; blue, 180 }  ,draw opacity=1 ][fill={rgb, 255:red, 255; green, 255; blue, 255 }  ,fill opacity=1 ][line width=3]  (118.58,172.47) .. controls (118.58,163.4) and (125.94,156.04) .. (135.01,156.04) .. controls (144.09,156.04) and (151.44,163.4) .. (151.44,172.47) .. controls (151.44,181.54) and (144.09,188.9) .. (135.01,188.9) .. controls (125.94,188.9) and (118.58,181.54) .. (118.58,172.47) -- cycle ;
\draw  [color={rgb, 255:red, 31; green, 119; blue, 180 }  ,draw opacity=1 ][fill={rgb, 255:red, 255; green, 255; blue, 255 }  ,fill opacity=1 ][line width=3]  (231.14,172.47) .. controls (231.14,163.4) and (238.49,156.04) .. (247.57,156.04) .. controls (256.64,156.04) and (264,163.4) .. (264,172.47) .. controls (264,181.54) and (256.64,188.9) .. (247.57,188.9) .. controls (238.49,188.9) and (231.14,181.54) .. (231.14,172.47) -- cycle ;
\draw  [color={rgb, 255:red, 31; green, 119; blue, 180 }  ,draw opacity=1 ][fill={rgb, 255:red, 255; green, 255; blue, 255 }  ,fill opacity=1 ][line width=3]  (173.58,247.47) .. controls (173.58,238.4) and (180.94,231.04) .. (190.01,231.04) .. controls (199.09,231.04) and (206.44,238.4) .. (206.44,247.47) .. controls (206.44,256.55) and (199.09,263.9) .. (190.01,263.9) .. controls (180.94,263.9) and (173.58,256.55) .. (173.58,247.47) -- cycle ;
\draw [line width=2.25]    (381.78,167.24) -- (441.94,244.53) ;
\draw [shift={(445.01,248.47)}, rotate = 232.1] [fill={rgb, 255:red, 0; green, 0; blue, 0 }  ][line width=0.08]  [draw opacity=0] (14.29,-6.86) -- (0,0) -- (14.29,6.86) -- cycle    ;
\draw [line width=2.25]    (390.01,173.47) -- (390.01,116.47) ;
\draw [shift={(390.01,111.47)}, rotate = 450] [fill={rgb, 255:red, 0; green, 0; blue, 0 }  ][line width=0.08]  [draw opacity=0] (14.29,-6.86) -- (0,0) -- (14.29,6.86) -- cycle    ;
\draw [line width=2.25]    (445.01,44.47) -- (393.19,107.61) ;
\draw [shift={(390.01,111.47)}, rotate = 309.38] [fill={rgb, 255:red, 0; green, 0; blue, 0 }  ][line width=0.08]  [draw opacity=0] (14.29,-6.86) -- (0,0) -- (14.29,6.86) -- cycle    ;
\draw [line width=2.25]    (445.01,44.47) -- (499.31,107.68) ;
\draw [shift={(502.57,111.47)}, rotate = 229.34] [fill={rgb, 255:red, 0; green, 0; blue, 0 }  ][line width=0.08]  [draw opacity=0] (14.29,-6.86) -- (0,0) -- (14.29,6.86) -- cycle    ;
\draw  [color={rgb, 255:red, 31; green, 119; blue, 180 }  ,draw opacity=1 ][fill={rgb, 255:red, 255; green, 255; blue, 255 }  ,fill opacity=1 ][line width=3]  (373.58,111.47) .. controls (373.58,102.4) and (380.94,95.04) .. (390.01,95.04) .. controls (399.09,95.04) and (406.44,102.4) .. (406.44,111.47) .. controls (406.44,120.54) and (399.09,127.9) .. (390.01,127.9) .. controls (380.94,127.9) and (373.58,120.54) .. (373.58,111.47) -- cycle ;
\draw  [color={rgb, 255:red, 31; green, 119; blue, 180 }  ,draw opacity=1 ][fill={rgb, 255:red, 255; green, 255; blue, 255 }  ,fill opacity=1 ][line width=3]  (428.58,44.47) .. controls (428.58,35.4) and (435.94,28.04) .. (445.01,28.04) .. controls (454.09,28.04) and (461.44,35.4) .. (461.44,44.47) .. controls (461.44,53.55) and (454.09,60.9) .. (445.01,60.9) .. controls (435.94,60.9) and (428.58,53.55) .. (428.58,44.47) -- cycle ;
\draw  [color={rgb, 255:red, 31; green, 119; blue, 180 }  ,draw opacity=1 ][fill={rgb, 255:red, 255; green, 255; blue, 255 }  ,fill opacity=1 ][line width=3]  (486.14,111.47) .. controls (486.14,102.4) and (493.49,95.04) .. (502.57,95.04) .. controls (511.64,95.04) and (519,102.4) .. (519,111.47) .. controls (519,120.54) and (511.64,127.9) .. (502.57,127.9) .. controls (493.49,127.9) and (486.14,120.54) .. (486.14,111.47) -- cycle ;
\draw [line width=2.25]    (502.57,173.47) -- (448.06,244.51) ;
\draw [shift={(445.01,248.47)}, rotate = 307.5] [fill={rgb, 255:red, 0; green, 0; blue, 0 }  ][line width=0.08]  [draw opacity=0] (14.29,-6.86) -- (0,0) -- (14.29,6.86) -- cycle    ;
\draw  [color={rgb, 255:red, 31; green, 119; blue, 180 }  ,draw opacity=1 ][fill={rgb, 255:red, 255; green, 255; blue, 255 }  ,fill opacity=1 ][line width=3]  (373.58,173.47) .. controls (373.58,164.4) and (380.94,157.04) .. (390.01,157.04) .. controls (399.09,157.04) and (406.44,164.4) .. (406.44,173.47) .. controls (406.44,182.54) and (399.09,189.9) .. (390.01,189.9) .. controls (380.94,189.9) and (373.58,182.54) .. (373.58,173.47) -- cycle ;
\draw  [color={rgb, 255:red, 31; green, 119; blue, 180 }  ,draw opacity=1 ][fill={rgb, 255:red, 255; green, 255; blue, 255 }  ,fill opacity=1 ][line width=3]  (486.14,173.47) .. controls (486.14,164.4) and (493.49,157.04) .. (502.57,157.04) .. controls (511.64,157.04) and (519,164.4) .. (519,173.47) .. controls (519,182.54) and (511.64,189.9) .. (502.57,189.9) .. controls (493.49,189.9) and (486.14,182.54) .. (486.14,173.47) -- cycle ;
\draw  [color={rgb, 255:red, 31; green, 119; blue, 180 }  ,draw opacity=1 ][fill={rgb, 255:red, 255; green, 255; blue, 255 }  ,fill opacity=1 ][line width=3]  (428.58,248.47) .. controls (428.58,239.4) and (435.94,232.04) .. (445.01,232.04) .. controls (454.09,232.04) and (461.44,239.4) .. (461.44,248.47) .. controls (461.44,257.55) and (454.09,264.9) .. (445.01,264.9) .. controls (435.94,264.9) and (428.58,257.55) .. (428.58,248.47) -- cycle ;

\end{tikzpicture}}
    \caption{Two instances of 2-regular graphs are presented. When ignoring vertex or edge features, it is impossible to distinguish these two graphs by the 1-dim WL test of graph isomorphism.}
    \label{fig:regular-graphs}
\end{figure}

There is a special class of graphs that, due to specific patterns in the structure, is usually impossible to be discriminated by current deep learning models for graphs as well as isomorphism testing algorithms such as the 1-dim Weisfeiler-Lehman (WL) test \cite{douglas_weisfeiler-lehman_2011}.
\begin{definition}[$\bm{k}$\textbf{-Regular Graph}]
A $k$-regular graph $g$ is one in which $deg(u)=k \ \forall u \in \Vset{g}$.
\end{definition}
As a way of example, Figure \ref{fig:regular-graphs} shows two 2-regular graphs that, however, are very different in nature. Indeed, the former is a disconnected graph whereas the second is connected, but from the point of view of their degree distributions they look identical. This is the reason why neither the 1-dim WL test nor most of the works on deep learning for graphs can distinguish these two graphs as non-isomorphic.

\subsection{Random Graphs}
A topic at the intersection of graph and probability theory is that of random graphs \cite{bollobas_random_2001}. This term generally refers to probability distributions defined over the discrete mathematical representation of graphs. These distributions are usually defined by means of a stochastic process over the creation of vertices and/or edges. Different random graph distributions give rise to graphs with peculiar characteristics. In what follows, we introduce two popular random graph distributions that are used to generate synthetic datasets in this thesis.

\begin{definition}[\textbf{Random Graph}]
A random graph is a graph $G$ of size $N$, where each pair of vertices is connected with probability $p$.\\
We use the capital letter $G$ because we are dealing with random processes, and refer to the family of random graphs with the term $G(N,p)$.
\end{definition}
In particular, the probability to obtain a \textit{particular} realization $g$ (without vertex/edge attributes) of G with $|\Eset{g}|=M$ is (for undirected connections):
\begin{align*}
    P(G(N,p)) = p^{M}(1-p)^{\binom{N}{2}-M}.
\end{align*}

\subsubsection*{The Erd\H{o}s-R\'enyi Model}
\begin{figure}[ht]
\begin{subfigure}
  \centering
  \includegraphics[width=0.49\textwidth]{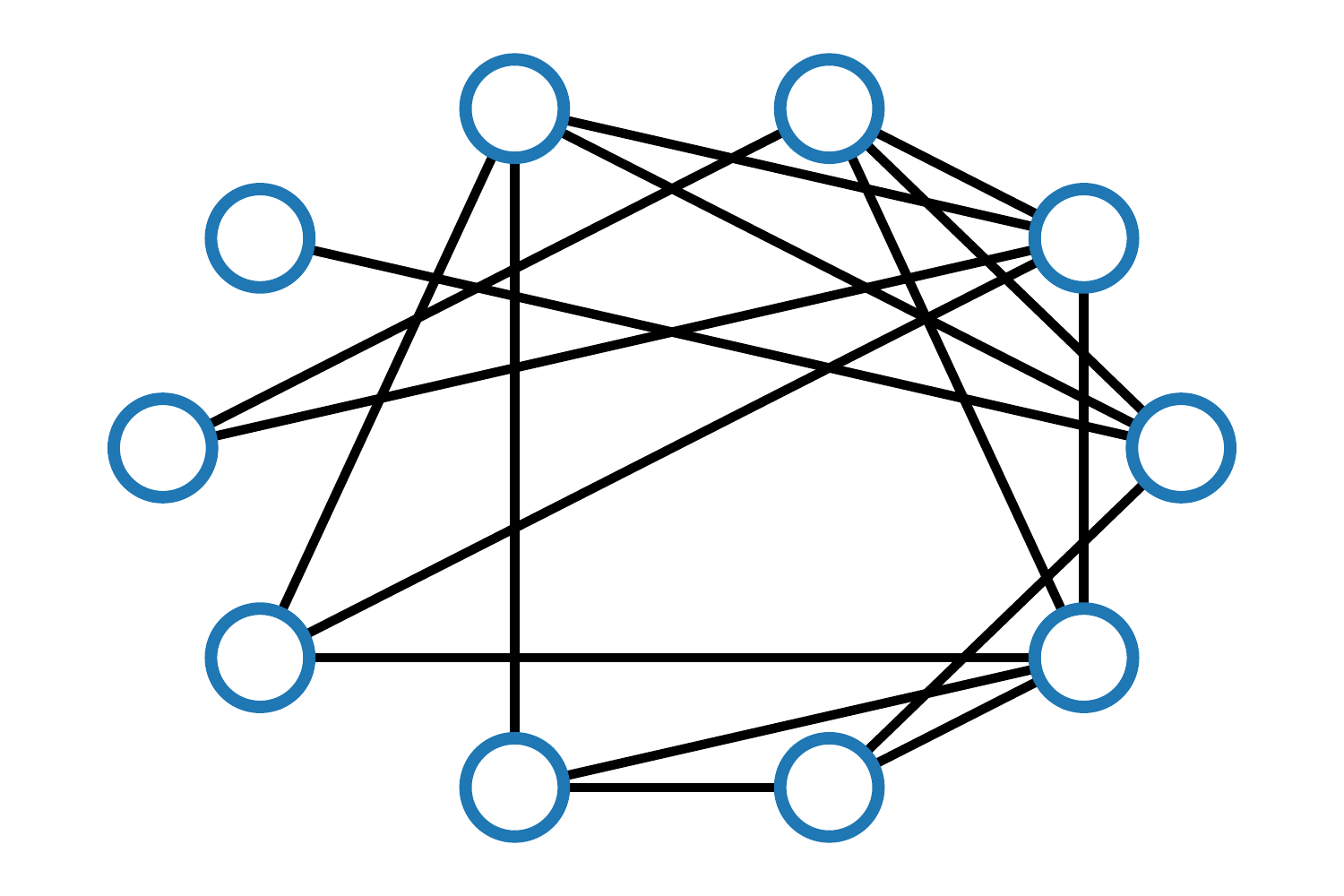}
\end{subfigure}%
\begin{subfigure}
  \centering
  \includegraphics[width=0.49\textwidth]{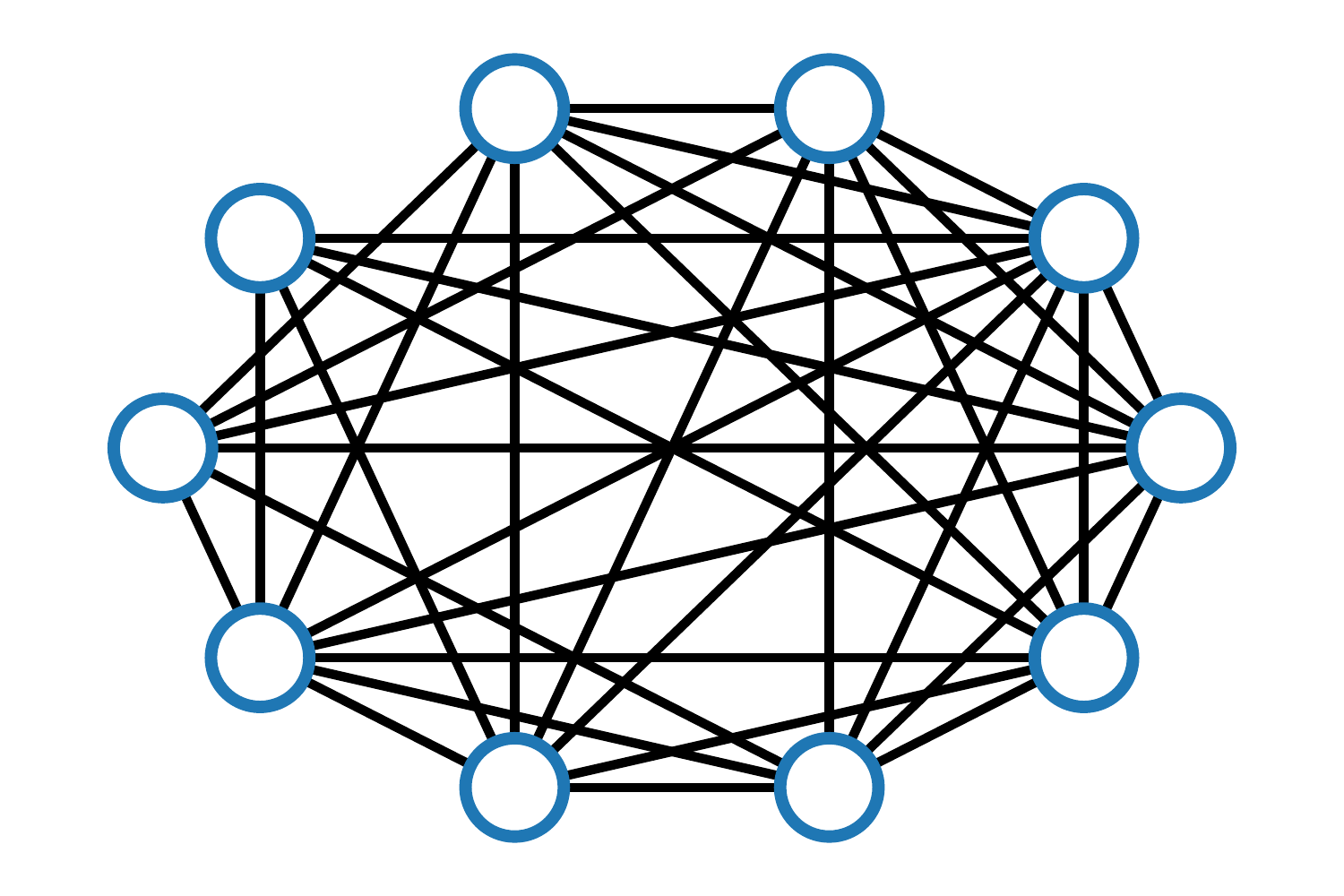}
\end{subfigure}
\caption{We generate two possible realizations of Erd\H{o}s-R\'enyi graphs of size $10$ with $p=0.3$ (left) and $p=0.8$ (right).}
\label{fig:ER-graphs}
\end{figure}

The Erd\H{o}s-R\'enyi (ER) model \cite{gilbert_random_1959,erdos_evolution_1960} is one of the pioneering works on random graphs. The realization of an undirected graph $g$ is obtained by considering all possible ordered pairs $\puv$, $u,v \in \Vset{g}$ and sampling an edge between them with probability $p$. The resulting distribution of the degree is \textbf{binomial} as follows:
\begin{align*}
    P(deg(u)=k) = \binom{N-1}{k}p^k(1-p)^{n-1-k}
\end{align*}
and a visualization of possible realizations of an ER graph are shown in Figure \ref{fig:ER-graphs} for different values of $p$.

\subsubsection*{The Barab\'asi-Albert Model}
\begin{figure}[ht]
\begin{subfigure}
  \centering
  \includegraphics[width=0.49\textwidth]{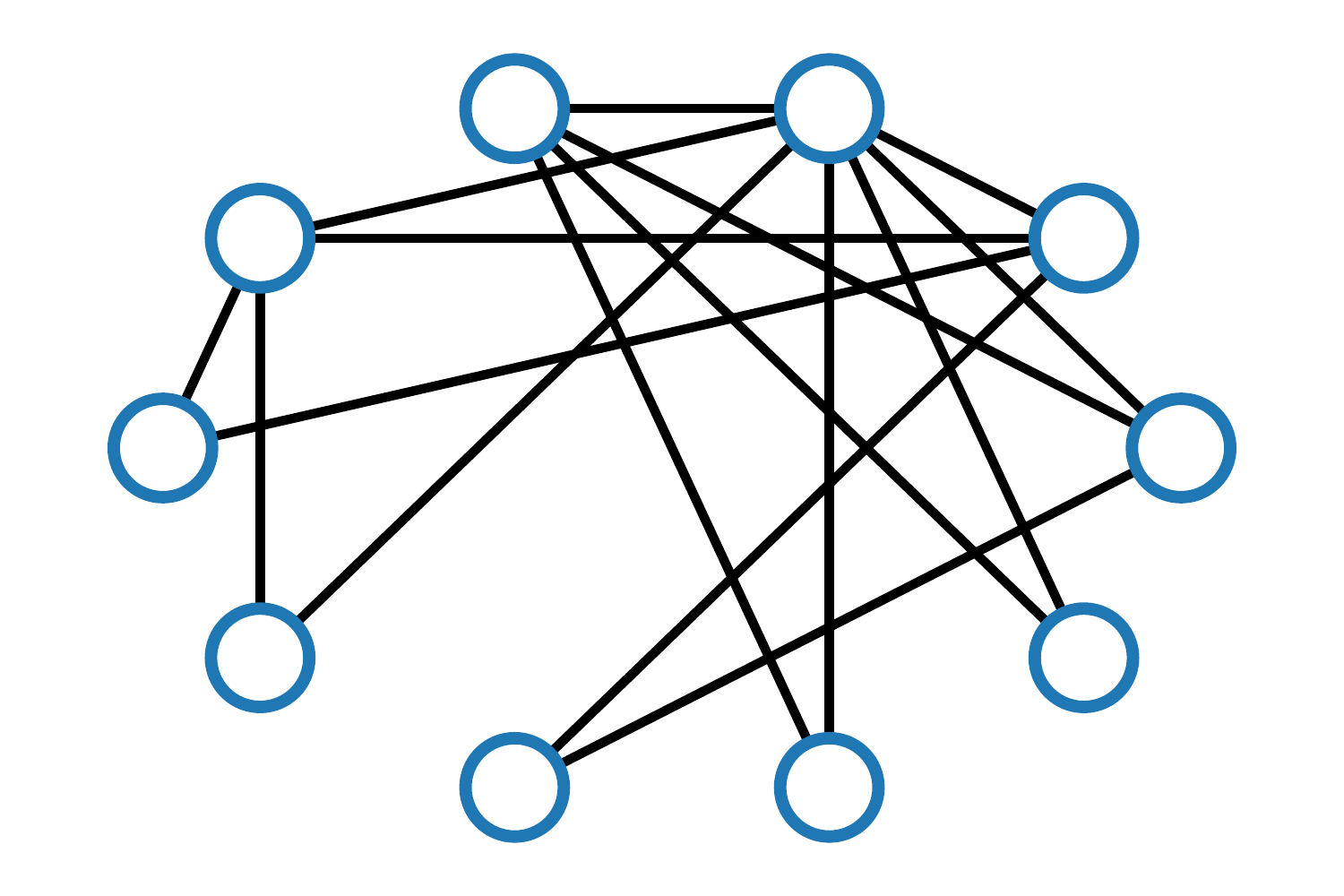}
\end{subfigure}%
\begin{subfigure}
  \centering
  \includegraphics[width=0.49\textwidth]{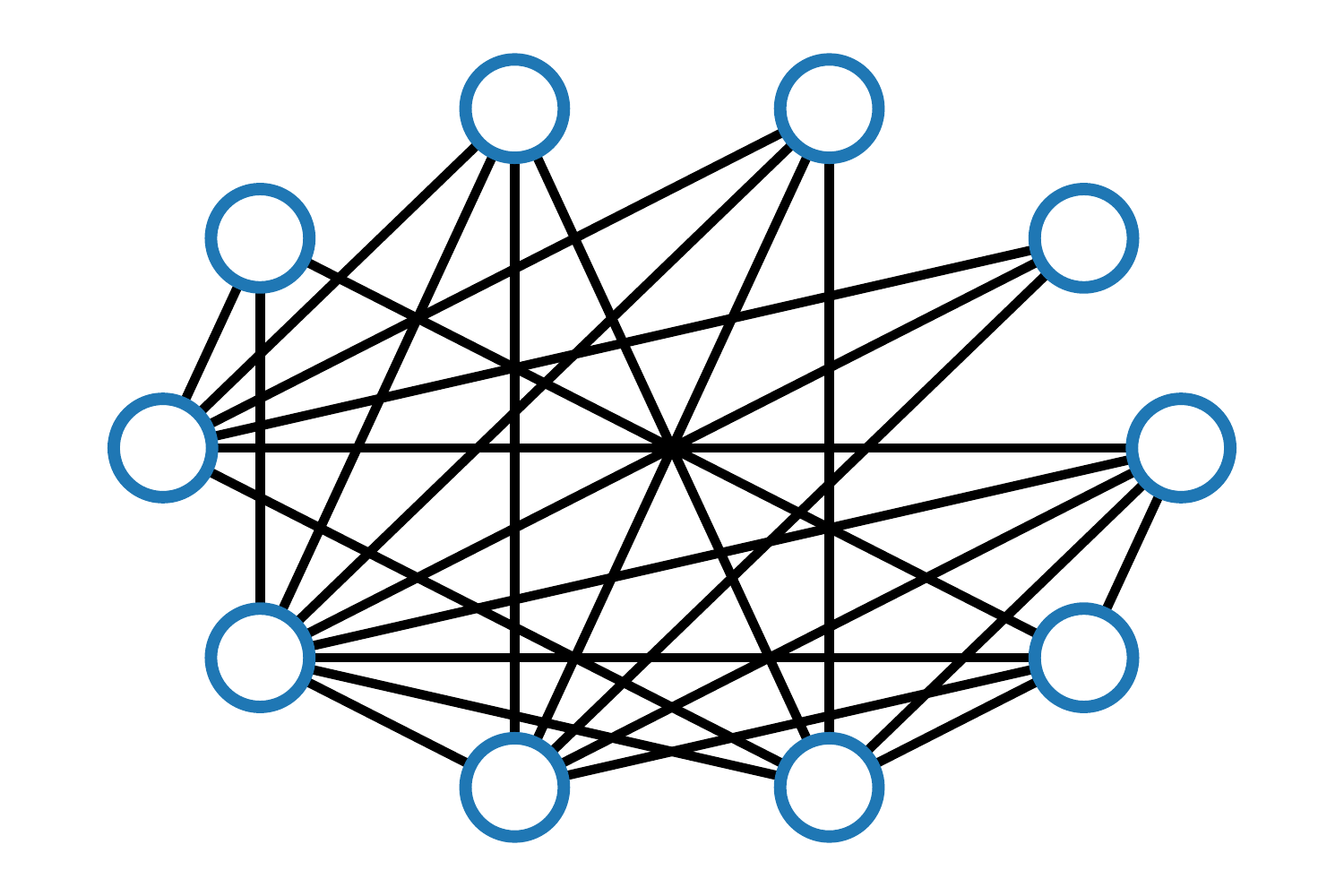}
\end{subfigure}
\caption{We generate two possible realizations of Barab\'asi-Albert graphs of size $10$ with $m=2$ (left) and $m=5$ (right).}
\label{fig:BA-graphs}
\end{figure}

In the real world, it is not difficult to find examples of graphs that have very different properties from the ER model. For example, social networks exhibit a \textbf{scale-free} property, \ie the degree distribution follows the so-called \textbf{power law}
\begin{align*}
    P(k\mid \gamma) = k^{-\gamma}
\end{align*}
parametrized by $\gamma$. In simple terms, this means that there are few vertices with high degree (the \quotes{hubs}) and many vertices with low degree. The construction of scale-free graphs follows the \textbf{preferential attachment process}, commonly known as \quotes{the rich get richer}\footnote{The \quotes{rich get richer} view can be found in Dirichlet Processes as well.}, and the Barab\'asi-Albert (BA) model \cite{barabasi_emergence_1999} is one way to define a distribution over scale-free graphs. The construction takes the size of the graph $N$ and a number of edges $m$ to add at each step. It starts by instantiating $n_0$ vertices, then it adds a new vertex $u$ and links it to $m$ existing vertices following a sort of preferential attachment criterion
\begin{align*}
    p_\uv = \frac{deg(v)}{\sum_{v'\leq v} deg(v')}.
\end{align*}
It can be shown that the degree distribution of the BA model follows the power-law distribution $P(k)=k^{-3}$. We conclude by showing, in Figure \ref{fig:BA-graphs}, two examples of BA graphs for varying values of the connectivity parameter $m$.

\section{What this Thesis is Not About}
\label{sec:other-topics}
This thesis focuses on deep learning for graphs with an emphasis on Bayesian techniques, but it would be inappropriate not to mention the different lines of research that address graph-related problems in alternative and original ways. Below, the reader can find a non-exhaustive list of some prominent approaches in the literature. Please bear in mind that the references below are just a few representative examples of a much vaster literature \cite{bronstein_geometric_2017,battaglia_relational_2018,bacciu_gentle_2020,wu_comprehensive_2020,ji_survey_2021}.

\subsection*{Kernels}
Kernel methods are one of the most long-standing and mature approaches that allow to compare graphs \cite{ralaivola_graph_2005, vishwanathan_graph_2010, shervashidze_weisfeiler-lehman_2011, frasconi_klog_2014, yanardag_deep_2015,kriege_survey_2020}. Briefly, a kernel is a function $k(\cdot,\cdot)$ that computes a similarity score for each pair of inputs in a given dataset $\dataset{}$. The resulting $|\dataset{}| \times |\dataset{}|$ matrix $K$ is generally required to be positive-definite, even though this constraint is sometimes overlooked without significant adverse effects. The \quotes{classical} drawback of most kernels is that the function $k$ has to be manually designed, and as such its choice may not be suitable for all tasks. Moreover, computing the kernel matrix can become unfeasible for very large datasets (with some exceptions \cite{shervashidze_weisfeiler-lehman_2011}). Some of the ways in which one can compute similarities between pairs of graphs is to take inspiration from the WL test of graph isomorphism \cite{shervashidze_weisfeiler-lehman_2011,morris_weisfeiler_2019} or to extract DAGs from each graph and then use tree-based kernels to compare them \cite{da_san_martino_tree-based_2012,da_san_martino_ordered_2016}. The matrix $K$, often called \textbf{Gram matrix}, can be incorporated into Support Vector Machines (SVMs) \cite{cortes_support-vector_1995} to address binary or multiclass graph classification.

Alternative methods may not be directly formulated as kernels but have an interpretation in terms of them. In particular, we mention a recent \textit{probabilistic} approach \cite{trentin_nonparametric_2018} that is somewhat related to graphlet (sub-graph) kernels \cite{shervashidze_efficient_2009} and was shown to be competitive against both kernels and deep learning models.

Clearly, when the properties of interest are known, kernels prove to be more than adequate to solve the task at hand. Whenever this is not the case, it might be preferable to \textit{learn} to extract features from a graph according to some unsupervised or supervised criterion, without posing handcrafted restrictions on the kind of patterns to look for.

\subsection*{Statistical Relational Learning}
Other classical approaches to deal with graph structures belong to the field of statistical relational learning \cite{koller_introduction_2007,de_statistical_2010}. This field combines probability theory, inductive logic, and learning to create data-driven models that possess a strong inductive bias rooted in logic rules. Examples to be ascribed to this field are Markov logic networks \cite{richardson_markov_2006}, which combine Markov random fields \cite{clifford_markov_1990} and first-order logic, and conditional random fields \cite{lafferty_conditional_2001} for classification of sequences. Differently from the methodologies described in this thesis, statistical relational learning does not typically rely on deep architecture to solve graph-related tasks, even though some hybrid approaches exist \cite{qu_gmnn_2019}.

\subsection*{Spectral Graph Theory}
\label{sec:spectral-graph-theory}
The objective of spectral graph theory is to mathematically characterize graphs through the analysis of the adjacency and Laplacian matrices. This is another well-studied topic that finds a diverse set of applications, from Laplacian smoothing \cite{sadhanala_graph_2016} to graph semi-supervised learning \cite{chapelle_semi-supervised_2006,calandriello_improved_2018} and spectral clustering \cite{von_luxburg_tutorial_2007}. We can also view the list of vertex features as a graph signal to be processed with ad-hoc signal-processing techniques: the Graph Fourier Transform \cite{hammond_wavelets_2011} has allowed to extend the formal definition of convolution over graph signals, and subsequent approaches \cite{bruna_spectral_2014} used approximations of spectral graph convolutions to learn graph filters.

A limitation of learned spectral techniques is the lack of generalization to new graph instances. In fact, everything depends on the eigen-decomposition of the specific Laplacian matrix, whose eigenvector matrix $Q$ is the orthonormal basis used to define the \textbf{Graph Fourier Transform} on the graph signal $\mathbf{f} \in \R{\Vset{g}}$:
\begin{align*}
    & \mathcal{F}(\mathbf{f}) = Q^T \mathbf{f} \\
    & \mathbf{f} = \mathcal{F}^{-1}(Q^T\mathbf{f}) = QQ^T \mathbf{f},
\end{align*}
where we used the ortogonality of $Q$ to obtain the inverse. Then, the convolution in the graph domain between a filter $\bm{\theta}$ and the graph signal can be written in a similar way to the usual Fourier analysis \cite{blackledge_chapter_2005,shuman_vertex_2016}:
\begin{align*}
(\mathbf{f} \otimes \boldsymbol{\theta}) =  QWQ^T\mathbf{f}
\end{align*}
where $W=diag(Q^T\bm{\theta})$ is the diagonal filter matrix (which can be learned \cite{bruna_spectral_2014}). The filter matrix, however, will only work with graphs that are identical, so it will not generalize to different graph instances. Also, computing the exact eigen-decomposition becomes unfeasible for large graphs. All those issues motivated the study of approximate techniques \cite{defferrard_convolutional_2016} using the truncated Chebyshev expansion \cite{hammond_wavelets_2011}. Later on, it was proposed to consider only the first term of said expansion \cite{kipf_semi-supervised_2017}, and the resulting approximation was used as a layer for a deep architecture for graphs.

\subsection*{Random Walks}

A random walk in a graph is a path starting from a vertex and exploring the surrounding neighborhood in a stochastic way. Random walks are often used to characterize a wider neighborhood of vertices: it is an attempt to acquire both local and non-local information to create meaningful vertex representations \cite{lovasz_random_1993,vishwanathan_graph_2010,ribeiro_struc2vec_2017,ivanov_anonymous_2018}. There are different ways in which random walks can be constructed and used: frameworks like Node2Vec \cite{grover_node2vec_2016} explore the surroundings of a vertex in a way that depends on the chosen hyper-parameters. Indeed, hyper-parameters determine whether to traverse the graph in depth or breadth search style. The objective to maximize is the likelihood of a vertex given the information extracted by the random walks. Similarly, methods like DeepWalk \cite{perozzi_deepwalk_2014}
model each random walk as a sentence, still maximizing a likelihood objective inspired by skip-grams. More recently, graph generation has been tackled with random walks \cite{bojchevski_netgan_2018}, whereas a connection between deep learning for graphs and random walks has been investigated in \cite{xu_representation_2018}.

\subsection*{Graph Generation}
 Graphs are discrete and combinatorial mathematical objects, and as such it is not obvious how to exactly define distributions over them. We have seen that random graphs define a process by which it is possible to generate skeletons of graphs, but now the question becomes \quotes{is it possible to \textbf{train} a model to generate graphs?}. Additionally, we would like to have our model creating graphs that are \textit{original} and variable in size. Many are the practical implications of such an approach, for instance developing new drugs conditioned on certain properties that must hold true, or similarly for the science of materials discovery.

Due to the nature of the input data, it is not easy to use gradient-based methods to approximate the underlying distribution $P(g)$. Instead, one usually conditions the generative process on a set of latent representations, either for the entire graph or its individual vertices. Most approaches can be therefore divided between \textbf{graph-level} and \textbf{vertex-level} decoding. The former generates an adjacency matrix starting from a latent graph representation \cite{jang_categorical_2017,simonovsky_graphvae_2018,de_cao_molgan_2018}, whereas the second connects vertices depending on the similarity of their latent representations \cite{kipf_variational_2016, grover_graphite_2019}. Note that graph-level decoders are generally sensitive to the ordering of the vertices because they assume a fixed ordering of the adjacency matrix, whereas node-level decoders do not suffer from this limitation. Among the generative graph models that are fully differentiable, we mention auto-encoder based generators \cite{kingma_auto-encoding_2014,tolstikhin_wasserstein_2018,simonovsky_graphvae_2018,liu_constrained_2018, samanta_nevae_2019,bradshaw_model_2019} and generative-adversarial networks \cite{goodfellow_generative_2014,fan_conditional_2019,wang_graphgan_2018,pan_adversarially_2018}. Finally, another family of models generates graphs as a result of a sequence of actions, which showed interesting generalization performances but is sensible to the vertex ordering \cite{li_learning_2018,you_graphrnn_2018,bacciu_graph_2019,bacciu_edge-based_2019,podda_deep_2020,podda_thesis_2021}.

\subsection*{Trustworthy AI for Complex Data}
Being able to determine if a model complies with the trustworthy AI principles, such as fairness, privacy, robustness, explainability, and transparency, remains an open and valuable research question. Indeed, robustness to adversarial attacks, \eg perturbations of the vertices or edges of topologically different graphs, guarantees that a model will behave as expected when deployed \cite{zugner_adversarial_2018,feng_graph_2019,zugner_certifiable_2019,bojchevski_certifiable_2019,yang_topology_2019,jin_latent_2019,zugner_certifiable_2020,zhang_detection_2021}. Similarly, fairness principles in graph applications have been analyzed \cite{bose_compositional_2019} to ensure that age/gender information about the entities is properly used and does not correlate too much with the target vale. While the probabilistic models presented in this thesis hold promise for what concerns interpretability and explainability (due to the explicit formulation of the causal effects), we leave such ideas for future research.

\subsection*{Theoretical Characterization of Models' \quotes{Expressiveness}}
There is a very active line of research devoted to the theoretical analysis of the discriminative power of deep learning methods for graphs. Among them, we mention studies on the effect of depth and width \cite{micheli_neural_2009,loukas_what_2020}, as well as enhancements to the main graph convolution mechanism that will be introduced later \cite{xu_how_2019,vignac_building_2020} and its theoretical characterization in terms of \textit{communication capacity} \cite{loukas_how_2020}. In addition, researchers have spent much effort in deriving equivalence relations between the $k$-dimensional WL test of graph isomorphism and specific deep learning architectures \cite{morris_weisfeiler_2019, maron_provably_2019,morris_weisfeiler_2020} that discriminate $k$-regular graphs. Finally, others have built lower and upper bounds on the expressiveness of specific classes of learning algorithms \cite{geerts_let_2021}.

\chapter{Principles of Deep Graph Networks}
\label{chapter:gentle-introduction}
\epigraph{\textit{Io non posso ritrar di tutti a pieno, \\ però che sì mi caccia il lungo tema, \\ che~molte~volte~al~fatto~il~dir~vien~meno.}}{\textit{Inferno - Canto IV}}
The goal of the chapter is to give a high-level overview of the field of \ML{} for graphs. Our main contribution is the standardization of the literature under a unified framework that let us look at similarities, differences, and novelties through the same lens. After some opening remarks on the main principles of contextual information processing, we build the discussion around the core building blocks of the field, such as neighborhood aggregation mechanisms, graph pooling, readout transductions and learning criteria. Equipped with this general understanding of deep learning for graphs, we shall then talk about scholarship issues in graph classification tasks. In fact, the rapid growth of interest in the field came at the price of poor reliability of empirical procedures. We shall discuss our effort in this direction, particularly how we carried out a fair, robust, and reproducible evaluation to help researchers avoid empirical malpractices in the future. To consolidate the theoretical notions, we will conclude the chapter with a real-world application from the field of molecular biosciences. By training a deep learning model on different protein realizations, we show that it is possible to efficiently predict the amount of \quotes{information loss} between the all-atom system we would like to study and one of its simpler but coarser representations. Overall, this \quotes{gentle} introduction creates fertile ground for what will come afterwards, that is, Bayesian deep learning models for graphs.

\clearpage
\section{Contextual Processing of Information}
\label{sec:contextual-processing}

To be able to put the recent burst of excitement about deep learning for graphs into the right perspective, we first go on a historical tour to show that the core ideas have already been there for more than twenty years. As a matter of fact, recent advancements of the field have not come without a certain forgetfulness of the pioneering methods that, however, shaped the field in its infant stages and are still relevant today.

\begin{figure}[ht]
    \hspace*{-1cm}
    \centering
    \resizebox{1\textwidth}{!}{\input{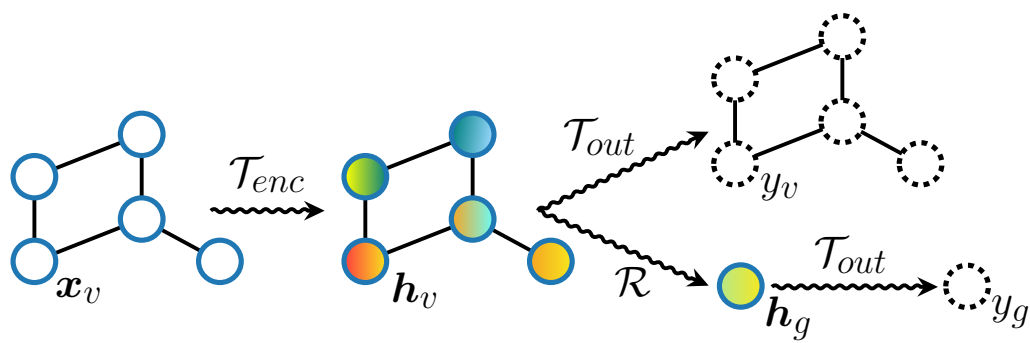}}
    \caption{High-level visualization of most SDL approaches. Each vertex of the input structure $g$ is mapped to a state $\boldh{v}$ after which, depending on the nature of the task, an IO-isomorphic output transduction $\T_{out}$ is applied \textit{or} all vertex states are first aggregated into a single state $\boldh{g}$ by a global aggregation function $\mathcal{R}$.}
    \label{fig:high-level}
\end{figure}

We pick up from the transduction $\T$ introduced in the previous chapter. Generally speaking, when solving a supervised task on structured data, we can decompose the transduction into two phases. First, there is an IO-isomorphic transduction $\T_{enc}$ that computes a vectorial encoding $\boldh{v}$ of each vertex $v$ in the structure; ideally, such an encoding (called \textbf{state} or \textbf{representation}) should capture information about the vertex's surroundings, in order to differentiate it from other vertex states. Then, if we care about individual entities' prediction, a \textbf{readout} transduction $\T_{out}$ has to map each vertex state to its corresponding output value $y_v$. Instead, when the task requires a single prediction $y_g$ for the whole structure, all vertex states have to be first aggregated into a single representation $\boldh{g}$  using a readout formed by the composition of a global aggregation $\mathcal{R}$ with an output transduction $\T_{out}$. This rather high-level scheme is summarized in Figure \ref{fig:high-level}, and it serves as our entry point into the world of \textbf{Structured-Data Learning} (SDL).

Hereinafter, we will be primarily concerned with ways to define the encoding transduction $\T_{enc}$, as this estabilishes \textit{how} vertex states are computed.\footnote{A reader interested in the theoretical aspects of the global aggregation and readout functions can refer to \cite{hammer_universal_2005,zaheer_deep_2017,xu_how_2019,wagstaff_limitations_2019} and the references therein.} Any SDL transduction relies on three main properties listed below
\begin{itemize}
    \item \textbf{Causality}: the encoding of an entity $v$ depends only on itself and its \textit{descendants} in the structure. The causality assumption intuitively acts as an inductive bias on the transduction by stating which dependencies between entities should be considered relevant. While strictly related to the class of input structures, causality assumptions and topology are actually different concepts. For instance, in an undirected sequence there are at least three straightforward causality assumptions one can make about the direction of the dependencies, with consequently different outcomes and \ML{} architectures.
    \item \textbf{Stationarity}: the encoding of an entity $v$ is the same regardless of $v$'s identity. Practically speaking, stationarity is deeply intertwined with the notion of weight sharing, as it defines how we can reuse the same encoding mechanism on every entity. We can also distinguish between \textit{full} stationarity, where we do not make any assumption on the class of structures under consideration, and \textit{positional} stationarity for DPAGs, in which we use different parameters for each position of the vertex's children to be considered \cite{bacciu_inputoutput_2013}.
    \item \textbf{Adaptivity}: the transduction is learned from the data. Whether the architecture is end-to-end differentiable or it admits closed-form update equations via the EM algorithm, the crux of the matter is that we want to avoid as much as possible any form of preprocessing of the structure. This is strongly related to the notion of \textbf{representation learning} \cite{bengio_representation_2013,goodfellow_deep_2016}.
\end{itemize}

In the abundant and pioneering literature for sequence, tree and DAG learning \cite{sperduti_supervised_1997,hochreiter_long_1997,frasconi_general_1998,bianucci_application_2000,diligenti_hidden_2003,micheli_contextual_2004,hammer_universal_2005,bacciu_compositional_2012,bacciu_inputoutput_2013,bacciu_modeling_2014}, the encoding transduction admits a recursive definition of the vertex state space, respectively.
This is possible because the causality assumptions on these structures, mostly inspired by the available topological ordering, do not incur in infinite loops that would generally make the definition of the vertex state unfeasible.

From now on, we shall use the term \textbf{contextual processing} when talking about computations whose output depends on the information encoded in the structure, \ie according to its topology and related causal assumptions. Also, the \textbf{context} of a vertex state $\boldh{v}$ shall be the set of states that directly or indirectly contribute to determining $\boldh{v}$.\footnote{Please refer to \cite{micheli_neural_2009} for a formal characterization of the context of a vertex state.}

\begin{figure}[ht]
    \centering
    \resizebox{0.8
    \textwidth}{!}{\input{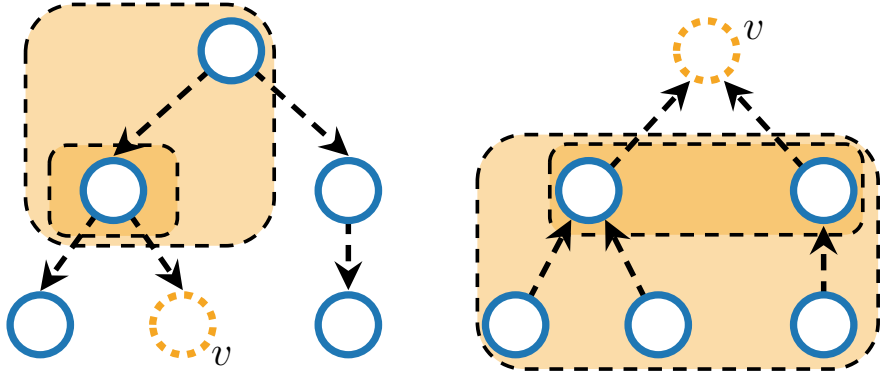}}
    \caption{Example of different causal assumptions on the same tree structure. Dashed arrows denote state dependencies rather than the actual topology, whereas encapsulated boxes show how the context of vertex $v$ increases as the recursive computation is unfolded.}
    \label{fig:context-dag}
\end{figure}

Let us make a concrete example to better understand the relation between causal assumptions and contextual processing of information. Regardless of whether the acyclic structure of Figure \ref{fig:context-dag} is directed or undirected, we can make different causal assumptions using its topology. Here, we assume top-down (left) and bottom-up (right) causal dependencies between vertex states. The state of vertex $v$, indicated by the orange dotted circle, is \textit{recursively} computed in terms of the states of the descendants. We use dashed edges to denote the state dependencies (the target depends on the source), which may differ from the original topological representation of the structure. Nested boxes aim to convey the amount of context involved at each recursive step in the definition of $\boldh{v}$. We observe how vertices can encode different kinds of contextual information, \ie the state of the parents up to the root or the entire subtree state rooted in $v$. The choice of a particular causal assumption is clearly task-dependent, since it encodes our bias about what kind of information vertex states should contain.

So far, we have considered \textbf{acyclic} structures, for which learning algorithms based on backpropagation through time \cite{baum_statistical_1966,werbos_backpropagation_1990} or structure \cite{goller_learning_1996,sperduti_supervised_1997,diligenti_hidden_2003,bacciu_generative_2012} exist and are practical to implement. Notwithstanding their importance, the above algorithms are generally unsuitable when it comes to learn vertex representations of a \textbf{cyclic} graph. In addition, some of these algorithms even assume that the size or number of parents/children are bounded by some value. On the other hand, we seek a practical methodology that can seamlessly treat cyclic graphs.

\section[Deep Learning for Graphs]{Deep Learning for Graphs \cite{bacciu_gentle_2020}}
\label{sec:dl4g}
Having defined the challenges ahead of us, it is time to describe the solutions proposed in the literature. Simply put, most works focus on the development of practical and effective models that \textit{automatically} extract features of interest from graphs of \textit{varying} topology while generalizing to \textit{unseen} instances. This is why the field is often referred to as \textbf{Graph Representation Learning} (GRL). In an attempt to disambiguate other terms \cite{scarselli_graph_2009,kipf_neural_2018}, in this thesis we will adopt the uniforming terminology of \textbf{Deep Graph Networks} (DGNs) \cite{bacciu_gentle_2020}. Below, we discuss how most DGNs address the presence of cycles, lack of a topological ordering, and variable topology in the dataset.

\subsection{Local Computation of Vertex States}
Two are the common solutions adopted in the literature to abstract from graphs of different topology, even though they are rarely stated explicitly. First of all, causality assumptions are relaxed, because taking into account the descendants of a vertex $v$ creates issues when that vertex belongs to a cycle. In other words, pairs of vertex states are assumed to be \textbf{independent} when conditioned on the states of their \textit{neighbors}. Secondly, due to the lack of a predefined total ordering of the neighboring vertices, in the most general case one has to assume \textit{full} stationarity of the encoding process. As a result, not only do we apply the same \textit{learned} function to all vertices, but we also need this function to be flexible enough to accept a variable number of neighboring states in any possible ordering. This is why \textbf{permutation invariant} functions are often employed in these models: their output does not change upon reordering of the input elements. Examples of such functions are the element-wise sum, mean, and product operators. Importantly, there exist conditions \cite{zaheer_deep_2017,wagstaff_limitations_2019} under which it is possible to express any continuous permutation invariant function $\Psi:\mathcal{X}^M \rightarrow \R{}$, with $\mathcal{X}$ being an uncountable set: if we consider a finite number $M$ of elements, then
\begin{align*}
    \Psi(x_1,\dots,x_M) = \phi(\sum_{i=1}^M \psi(x_i))
\end{align*}
with $\phi$ and $\psi$ being continuous functions that can be approximated by neural networks \cite{cybenko_approximation_1989}. For the rest of this work, we will use the Greek letter $\Psi$ to denote a permutation invariant function.


\subsection{Breaking Cycles via Iterations}
By now, the attentive reader may have noticed that the issue of mutual dependencies induced by cycles has not been solved yet. To see this, just imagine that a vertex is connected to itself via a self-loop. Clearly, we cannot compute its state using the aforementioned local approach because one of the neighboring vertex states is the state of the vertex itself. Therefore, other than being local, the processing of information in DGNs has to be \textbf{iterative}: the state of a vertex is conditioned on neighboring states computed \quotes{at some previous iteration}. If we already have some information on which to condition the vertex states, we can effectively \textbf{break cycles} in the structure with a simple iterative approximation.

\begin{figure}[ht]
    \hspace*{-1.5cm}
    \centering
    \resizebox{1\textwidth}{!}{\input{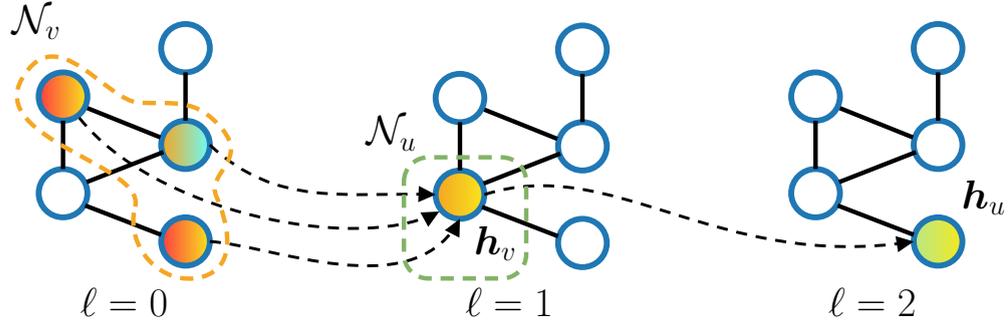}}
    \caption{We show how the context of a vertex spreads in an undirected graph as we iterate the local processing of information described above. Dashed arrows represent the context flow as we unfold the computational graph.}
    \label{fig:context-flow}
\end{figure}

Crucially, a local and iterative processing of information allows us to \textbf{propagate contextual information} across the graph. If we unfold the computation on the graph of Figure \ref{fig:context-flow}, we can see how at iteration $\lcurr=2$ the state $\boldh{u}$ depends on the state of $\boldh{v}$ at $\lcurr=1$. In turn, $\boldh{v}$ is obtained by applying a learned permutation invariant function to the neighboring states of iteration $\lcurr=0$. Therefore, by repeated iteration, each vertex increases its context according to the graph topology \cite{micheli_neural_2009}. In what follows, we shall refer to the state of vertex $u$ at iteration $\lcurr$ with the symbol $\boldhell{u}{\lcurr}$. To bootstrap the overall process, it is often the case that $\boldhell{u}{0} = \boldx{u}$.

A popular formalism to describe the above process is that of \quotes{\textbf{message passing}} \cite{angluin_local_1980,linial_locality_1992, naor_can_1995,peleg_distributed_2000,gilmer_neural_2017}. In particular, there are two operations associated with each vertex:
\begin{itemize}
    \item \textit{message dispatching}: compute a message for each vertex to be propagated to the neighbors. The message may depend on the vertex state as well as edge information.
    \item \textit{state update}: the state of each vertex is updated \textbf{in parallel} using the incoming messages.
\end{itemize}

Likewise, we can also imagine traversing the graph in no particular order to update the vertex states, and then iterate this traversal as many times as needed. This recalls the idea of Convolutional Neural Networks for images\footnote{Images can be represented as graphs where vertices are organized in a grid and for which a \quotes{natural} ordering exists.} \cite{lecun_convolutional_1995}, where at each layer a filter passes over all pixels to compute new values based on the pixels' surroundings and multiple layers increase the \textit{receptive field} (\ie the context). 

\subsection{Three Styles of Context Propagation}
In the above schemes, the notion of \quotes{iterative process} is sufficiently generic to encompass different styles of context diffusion. Therefore, we partition DGNs into three main families, namely \textbf{recurrent}, \textbf{feedforward}, and \textbf{constructive} architectures. We now separately describe their characteristics.

\paragraph*{Recurrent architectures.}
Recurrent \ML{} models for graphs treat the encoding process as a dynamical system, but at the same time they rely on contractive dynamics to ensure some convergence criterion can be met. The Graph Neural Network (GNN) \cite{scarselli_graph_2009} and the Graph Echo State Network (GraphESN) were the first models developed in this sense. On the one hand, GNN is a recurrent neural network that relies on constraints in the (supervised) objective function to ensure convergence. On the other hand, GraphESN brings the Reservoir Computing approach to the processing of graphs, inheriting convergence from the contractivity property of the \textit{untrained} pool of neurons. 
That said, it is also possible to fix in advance the number of iterations, regardless of whether convergence has been reached. This was the idea behind the Gated Graph Neural Network (GG-NN) \cite{li_gated_2016}.

Recurrent architectures treat the single iteration $\lcurr$ of the encoding process as a \quotes{time step} of the corresponding dynamical system, and they consist of a \textbf{single} layer of \textbf{recurrent} units to be repeatedly applied. Nevertheless, there exist multi-layered versions of these models such as the Fast and Deep Graph Neural Network (FDGNN) \cite{gallicchio_fast_2020}, which extends GraphESN to efficiently construct multi-resolution views of the graph.

\paragraph*{Feedforward architectures.}
From the perspective of feedforward models, iterations are \textbf{layers} of a possibly deep architecture, where each layer has its own parameters to be optimized. Stacking multiple layers is a way to \textit{compose} the context learned in a very flexible way, without being forced to impose contractive dynamics or use recurrent units. The Neural Network for Graphs (NN4G) \cite{micheli_neural_2009} was the first feedforward model for graphs to be developed, defining what was later re-discovered as the \quotes{spatial graph convolutional layer} \cite{kipf_semi-supervised_2017,velickovic_graph_2018}.

This family is the most known at both industrial and research levels, for its simplicity, ease of implementation, and competitive performances on many different tasks \cite{wu_comprehensive_2020}. At the same time, it inherits the same gradient-related issues of deep neural networks, in particular when trained in an end-to-end fashion \cite{bengio_learning_1994,li_deeper_2018}. Differently from deep networks for flat data, though, here depth serves two purposes, \ie automatically extracting features \textit{and} propagating contextual information across the graph. Nowadays, it is very straightforward to build these models, thanks to the collective effort of researchers that released easy-to-use libraries for quick development and experimentation \cite{fey_fast_2019,paszke_pytorch_2019,bacciu_gentle_2020}.

\paragraph*{Constructive architectures.}
A constructive model is a special case of a feedforward model where training occurs one layer at a time. As a consequence, constructive models do not necessarily suffer from oversmoothing of representations \cite{li_deeper_2018} or vanishing/exploding gradient effects. When used with a supervised criterion, this methodology allows to automatically determine the number of layers needed by the task, \ie the amount of context to propagate, for example using Cascade Correlation \cite{fahlman_cascade-correlation_1990}.

One of the major characteristics of constructive models is that they approach the task in a \textit{divide-et-impera} fashion, where each layer contributes to the solution of a sub-problem and subsequent layers try to better solve the task using the information extracted from previous layers. Notably, once a layer is trained its weights are frozen, meaning they do not change while training new layers. The NN4G belongs to the family of constructive approaches, as well as the \textbf{Contextual Graph Markov Model} (CGMM) \cite{bacciu_contextual_2018} of Chapter \ref{chapter:dbgn}.

\begin{figure}[ht]
    \centering
    \resizebox{1\textwidth}{!}{\tikzset{every picture/.style={line width=0.75pt}} 

\begin{tikzpicture}[x=0.75pt,y=0.75pt,yscale=-1,xscale=1]

\draw [line width=3.75]    (502.5,102) .. controls (502.5,140) and (187.42,95) .. (188.42,156) ;
\draw [line width=3.75]    (188.5,208) .. controls (188.5,243) and (110.5,193) .. (111.5,252) ;
\draw [line width=3.75]    (188.5,208) .. controls (188.5,243) and (263.5,194) .. (264.5,251) ;
\draw [line width=3.75]    (267.5,293) .. controls (267.5,315.76) and (299.22,303) .. (321.68,306.97) .. controls (333.76,309.1) and (343.15,316.07) .. (343.5,336) ;
\draw [line width=3.75]    (505.5,102) .. controls (505.5,140) and (823.42,98) .. (824.42,156) ;
\draw [line width=3.75]    (504.5,102) -- (504.5,161) ;
\draw [line width=3.75]    (504.5,210) -- (504.5,250.67) ;
\draw [line width=3.75]    (827.5,212) .. controls (827.5,250) and (752,199) .. (753,260) ;
\draw [line width=3.75]    (827.5,212) .. controls (827.5,250) and (910,197) .. (911,258) ;
\draw [line width=1.5]  [dash pattern={on 5.63pt off 4.5pt}]  (267.5,293) -- (182.33,368.56) ;
\draw [shift={(180.08,370.55)}, rotate = 318.41999999999996] [color={rgb, 255:red, 0; green, 0; blue, 0 }  ][line width=1.5]    (14.21,-4.28) .. controls (9.04,-1.82) and (4.3,-0.39) .. (0,0) .. controls (4.3,0.39) and (9.04,1.82) .. (14.21,4.28)   ;
\draw [line width=1.5]  [dash pattern={on 5.63pt off 4.5pt}]  (512.08,288.55) -- (567.02,346.37) ;
\draw [shift={(569.08,348.55)}, rotate = 226.47] [color={rgb, 255:red, 0; green, 0; blue, 0 }  ][line width=1.5]    (14.21,-4.28) .. controls (9.04,-1.82) and (4.3,-0.39) .. (0,0) .. controls (4.3,0.39) and (9.04,1.82) .. (14.21,4.28)   ;
\draw [line width=1.5]  [dash pattern={on 5.63pt off 4.5pt}]  (521.08,413.55) -- (470.16,466.39) ;
\draw [shift={(468.08,468.55)}, rotate = 313.94] [color={rgb, 255:red, 0; green, 0; blue, 0 }  ][line width=1.5]    (14.21,-4.28) .. controls (9.04,-1.82) and (4.3,-0.39) .. (0,0) .. controls (4.3,0.39) and (9.04,1.82) .. (14.21,4.28)   ;
\draw [line width=1.5]  [dash pattern={on 5.63pt off 4.5pt}]  (639.08,413.55) -- (696.87,466.52) ;
\draw [shift={(699.08,468.55)}, rotate = 222.51] [color={rgb, 255:red, 0; green, 0; blue, 0 }  ][line width=1.5]    (14.21,-4.28) .. controls (9.04,-1.82) and (4.3,-0.39) .. (0,0) .. controls (4.3,0.39) and (9.04,1.82) .. (14.21,4.28)   ;

\draw (499,84) node  [font=\Large] [align=left] {(Deep \textbf{Graph} Networks)};
\draw (190,194) node  [font=\Large] [align=left] {(Deep \textbf{Neural} Graph Networks)};
\draw (512,192) node  [font=\Large] [align=left] {(Deep \textbf{Bayesian} Graph Networks)};
\draw (114,276) node  [font=\Large] [align=left] {Recurrent};
\draw (264,274) node  [font=\Large] [align=left] {Feedforward};
\draw (345,349) node  [font=\Large] [align=left] {Constructive};
\draw (511,273) node  [font=\Large] [align=left] {Constructive};
\draw (190,173) node  [font=\Large] [align=left] {\textit{\textbf{DNGNs}}};
\draw (507,174) node  [font=\Large] [align=left] {\textit{\textbf{DBGNs}}};
\draw (829,192) node  [font=\Large] [align=left] {(Deep \textbf{Generative} Graph Networks)};
\draw (826,172) node  [font=\Large] [align=left] {\textit{\textbf{DGGNs}}};
\draw (504,62) node  [font=\Large] [align=left] {\textit{\textbf{DGNs}}};
\draw (742,274) node  [font=\Large] [align=left] {Node-level};
\draw (910.5,273) node  [font=\Large] [align=left] {Graph-level};
\draw (740,294) node  [font=\Large] [align=left] {decoder};
\draw (914,294) node  [font=\Large] [align=left] {decoder};
\draw (177.58,404.2) node  [font=\Large,color={rgb, 255:red, 31; green, 119; blue, 180 }  ,opacity=1 ] [align=left] {\begin{minipage}[lt]{214.17pt}\setlength\topsep{0pt}
\begin{center}
Graph Mixture Density Networks\\(Chapter 5)
\end{center}

\end{minipage}};
\draw (582.58,392.2) node  [font=\Large,color={rgb, 255:red, 31; green, 119; blue, 180 }  ,opacity=1 ] [align=left] {\begin{minipage}[lt]{170.11pt}\setlength\topsep{0pt}
\begin{center}
Contextual Graph Markov\\ Model (CGMM)\\(Chapter 4)
\end{center}

\end{minipage}};
\draw (469.58,495.55) node  [font=\Large,color={rgb, 255:red, 31; green, 119; blue, 180 }  ,opacity=1 ] [align=left] {\begin{minipage}[lt]{115.38pt}\setlength\topsep{0pt}
\begin{center}
Extended CGMM\\(Chapter 4)
\end{center}

\end{minipage}};
\draw (696.58,495.55) node  [font=\Large,color={rgb, 255:red, 31; green, 119; blue, 180 }  ,opacity=1 ] [align=left] {\begin{minipage}[lt]{96.59pt}\setlength\topsep{0pt}
\begin{center}
Infinite CGMM\\(Chapter 4)
\end{center}

\end{minipage}};

\end{tikzpicture}}
    \caption{A taxonomy of the various context propagation mechanisms, with the addition of the specific models developed in this thesis.}
    \label{fig:taxonomy}
\end{figure}
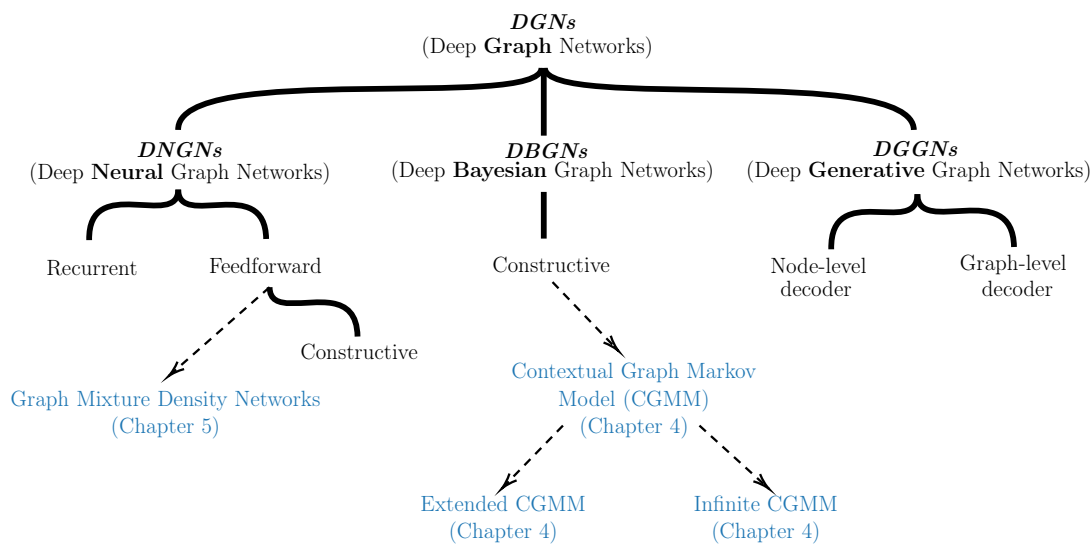

Before we dive deep into the building blocks of Deep Graph Networks, we provide an illustrative taxonomy summarizing what we said so far and highlighting where the models developed in this thesis stand. The taxonomy, provided in Figure \ref{fig:taxonomy}, is by no means exhaustive, but it encompasses how most works propagate contextual information. We also make the distinction between Deep \textbf{Neural} Graph Networks\footnote{Which are ambiguously referred to as Graph Neural Networks in the literature.} (DNGNs), Deep \textbf{Bayesian} Graph Networks (DBGNs), \ie probabilistic and deep models for graphs, and Deep \textbf{Generative} Graph Networks (DGGNs) \cite{podda_thesis_2021}, \ie models that are able to \textit{generate} new and original graphs. Notice that, in principle, one can also combine different context propagation mechanisms if the task is more complex than the usual, for instance by exploiting both feedforward and recurrent mechanisms to handle graphs that vary in time.

\subsection{Core Modules}

We are finally ready to describe the main mechanisms of Deep Graph Networks. Not only will these accompany us in the next chapters, making it easier to understand the rationale behind some of our technical choices, but they will also give a unifying view of the different approaches in the literature, which is one of the main contribution of this thesis.

\subsubsection*{Neighborhood Aggregation}
The definition of the permutation invariant function computing the local encoding of each vertex in the graph is arguably the key building block of any DGN. Indeed, this neighborhood aggregation function imposes an architectural bias that has important consequences on the representational power of the model under consideration. A very straightforward form of neighborhood aggregation is
\begin{align}
\boldhell{v}{\lnext} = \phi^{\lnext} \Big(\boldhell{v}{\lcurr},\ \Psi(\{\psi^{\lnext}(\boldhell{u}{\lcurr}) \mid u \in \N{v}{}\} ) \Big),
\label{eq:simple-aggregation}
\end{align}
where $\phi$ and $\psi$ are adaptive transformations of the input, \eg Multi Layer Perceptrons. It can be shown \cite{bacciu_gentle_2020,wu_comprehensive_2020} how Equation \ref{eq:simple-aggregation} is a generalization of some of the most known aggregation schemes, such as the Graph Convolutional Network \cite{kipf_semi-supervised_2017} and the Graph Isomorphism Network (GIN) \cite{xu_how_2019}. Besides, these architectures ignore any additional information on the nature of the relation between vertices, which is usually stored by \textbf{edge attributes}. In chemistry, for instance, it is common to have discrete edge features describing the type of bond between atoms as well as continuous values associated with their inter-atomic distance. In the former case of $\Aset{g}$ finite and discrete, one can extend the previous equation with additional parameters $w_{c_k}$ to be learned for each discrete edge label $c_k$:
\begin{align*}
\boldhell{v}{\lnext} = \phi^{\lnext} \Big(\boldhell{v}{\lcurr},\ \sum_{c_k \in \mathcal{A}}  \big( \Psi(\{\psi^{\lnext}(\boldhell{u}{\lcurr}) \mid u \in \N{v}{c_k}\} ) * w_{c_k} \big) \Big),
\end{align*}
where the symbol $*$ stands for multiplication between a scalar and a vector and we recall, from Section \ref{subsec:fundamentals}, that $\N{v}{c_k}$ is the subset of $v$'s neighbors whose connecting edges have label $c_k$. This is another form of stationarity over edge weights, since we are dealing with non-positional graphs. If graphs were positional, we could use a different weight for each position and each edge type. Also, please take note of how we grouped neighbors of $v$ according to their edge type; this is a practice that will be implemented in the next chapter. In the literature, NN4G \cite{micheli_neural_2009} and the Relational Graph Convolutional Network (R-GCN) \cite{schlichtkrull_modeling_2018} are just some of the models implementing this aggregation scheme.

On the contrary, if edge features $\boldsymbol{a}_\uv$ are continuous, one can easily combine vertex and edge states:
\begin{align*}
\boldhell{v}{\lnext} = \phi^{\lcurr} \Big(\boldhell{v}{\lcurr},\ \Psi(\{e^{\lnext}(\mathbf{a}_\uv)\cdot \psi^{\lnext}(\boldhell{u}{\lcurr}) \mid u \in \N{v}{}\} ) \Big),
\end{align*}
with $\cdot$ standing for the Hadamart product between vectors. The Message Passing Neural Network framework (MPNN) \cite{gilmer_neural_2017} and the Edge Conditioned Convolution (ECC) \cite{simonovsky_dynamic_2017} implement this kind of aggregation at the expense of extra computation for the edges, which can significantly slow down the algorithm.

A natural question that comes to mind at this point is \quotes{even though the graph is non-positional, should we treat all neighbors as equal?}. In fact, we may not, and that is the idea behind the \textbf{attention} mechanism \cite{vaswani_attention_2017} applied to the neighborhood aggregation of DNGNs. First of all, consider having a way to compute some sort of similarity score $\alpha_\uv$ between two vertex states $\boldh{u}$ and $\boldh{v}$. We can extend the aggregation mechanism of Equation \ref{eq:simple-aggregation} to re-weight neighbors according to such a score:
\begin{align*}
\boldhell{v}{\lnext} = \phi^{\lnext} \Big(\boldhell{v}{\lcurr},\ \Psi(\{\alpha^{\lnext}_\uv*\psi^{\lnext}(\boldhell{u}{\lcurr}) \mid u \in \N{v}{}\} ) \Big).
\end{align*}
We impose no restriction on how to compute $\alpha_\uv$ but for the procedure being differentiable or non-adaptive. Nothing prevents us from combining the above equations to obtain an attention score that depends, for example, on edge features. The Graph Attention Network (GAT) \cite{velickovic_graph_2018} was the first model to apply a multi-head attention mechanism to a Deep Graph Network, with potential advantages in terms of interpretability.

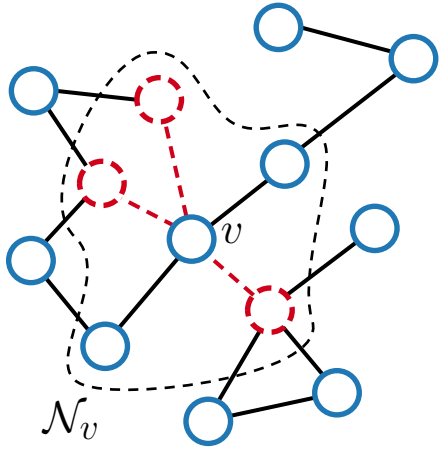
\begin{figure}[ht]
    \centering
    \resizebox{0.4\textwidth}{!}{\tikzset{every picture/.style={line width=0.75pt}} 

\begin{tikzpicture}[x=0.75pt,y=0.75pt,yscale=-1,xscale=1]

\draw [line width=2.25]    (504.83,176) -- (548.83,141) ;
\draw [line width=2.25]    (482.83,198.42) -- (438.03,251.7) ;
\draw [line width=2.25]    (542.33,254) -- (512.33,305.33) ;
\draw [line width=2.25]    (582.33,304) -- (519.33,317.67) ;
\draw [line width=2.25]    (559.67,251.33) -- (589,286.67) ;
\draw [line width=2.25]    (613.67,188) -- (563,226.67) ;
\draw [line width=2.25]    (643,62.67) -- (573,117.33) ;
\draw [line width=2.25]    (641.67,46) -- (572.67,27.5) ;
\draw [line width=2.25]    (413.67,130.67) -- (385.17,87.67) ;
\draw [line width=2.25]    (451.33,81.5) -- (391.33,74.83) ;
\draw [line width=2.25]    (383,188) -- (412.33,156) ;
\draw [line width=2.25]    (417,255.33) -- (382.33,212.67) ;
\draw  [dash pattern={on 5.63pt off 4.5pt}][line width=1.5]  (458.75,46.75) .. controls (492.25,39.75) and (499.25,85.25) .. (522.25,98.75) .. controls (545.25,112.25) and (573.75,88.25) .. (588.25,109.25) .. controls (602.75,130.25) and (569.25,206.25) .. (581.25,246.75) .. controls (593.25,287.25) and (442.75,307.25) .. (410.25,293.75) .. controls (377.75,280.25) and (436.25,213.75) .. (405.25,183.75) .. controls (387.39,166.46) and (396.75,125.94) .. (414.3,93.35) .. controls (427.21,69.39) and (444.55,49.72) .. (458.75,46.75) -- cycle ;
\draw  [color={rgb, 255:red, 31; green, 119; blue, 180 }  ,draw opacity=1 ][fill={rgb, 255:red, 255; green, 255; blue, 255 }  ,fill opacity=1 ][line width=3]  (544.33,129.5) .. controls (544.33,120.39) and (551.72,113) .. (560.83,113) .. controls (569.95,113) and (577.33,120.39) .. (577.33,129.5) .. controls (577.33,138.61) and (569.95,146) .. (560.83,146) .. controls (551.72,146) and (544.33,138.61) .. (544.33,129.5) -- cycle ;
\draw  [color={rgb, 255:red, 208; green, 2; blue, 27 }  ,draw opacity=1 ][dash pattern={on 7.88pt off 4.5pt}][line width=3]  (408.83,143.92) .. controls (408.83,134.8) and (416.22,127.42) .. (425.33,127.42) .. controls (434.45,127.42) and (441.83,134.8) .. (441.83,143.92) .. controls (441.83,153.03) and (434.45,160.42) .. (425.33,160.42) .. controls (416.22,160.42) and (408.83,153.03) .. (408.83,143.92) -- cycle ;
\draw  [color={rgb, 255:red, 31; green, 119; blue, 180 }  ,draw opacity=1 ][fill={rgb, 255:red, 255; green, 255; blue, 255 }  ,fill opacity=1 ][line width=3]  (411.33,265.5) .. controls (411.33,256.39) and (418.72,249) .. (427.83,249) .. controls (436.95,249) and (444.33,256.39) .. (444.33,265.5) .. controls (444.33,274.61) and (436.95,282) .. (427.83,282) .. controls (418.72,282) and (411.33,274.61) .. (411.33,265.5) -- cycle ;
\draw  [color={rgb, 255:red, 208; green, 2; blue, 27 }  ,draw opacity=1 ][dash pattern={on 7.88pt off 4.5pt}][line width=3]  (533.33,238.5) .. controls (533.33,229.39) and (540.72,222) .. (549.83,222) .. controls (558.95,222) and (566.33,229.39) .. (566.33,238.5) .. controls (566.33,247.61) and (558.95,255) .. (549.83,255) .. controls (540.72,255) and (533.33,247.61) .. (533.33,238.5) -- cycle ;
\draw [color={rgb, 255:red, 208; green, 2; blue, 27 }  ,draw opacity=1 ][line width=2.25]  [dash pattern={on 6.75pt off 4.5pt}]  (537.73,227.32) -- (503.63,197.3) ;
\draw [color={rgb, 255:red, 208; green, 2; blue, 27 }  ,draw opacity=1 ][line width=2.25]  [dash pattern={on 6.75pt off 4.5pt}]  (488.03,168.9) -- (472.03,97.7) ;
\draw [color={rgb, 255:red, 208; green, 2; blue, 27 }  ,draw opacity=1 ][line width=2.25]  [dash pattern={on 6.75pt off 4.5pt}]  (479.83,176) -- (439.83,155) ;
\draw  [color={rgb, 255:red, 208; green, 2; blue, 27 }  ,draw opacity=1 ][dash pattern={on 7.88pt off 4.5pt}][line width=3]  (451.33,81.5) .. controls (451.33,72.39) and (458.72,65) .. (467.83,65) .. controls (476.95,65) and (484.33,72.39) .. (484.33,81.5) .. controls (484.33,90.61) and (476.95,98) .. (467.83,98) .. controls (458.72,98) and (451.33,90.61) .. (451.33,81.5) -- cycle ;
\draw  [color={rgb, 255:red, 31; green, 119; blue, 180 }  ,draw opacity=1 ][fill={rgb, 255:red, 255; green, 255; blue, 255 }  ,fill opacity=1 ][line width=3]  (539.67,27.5) .. controls (539.67,18.39) and (547.05,11) .. (556.17,11) .. controls (565.28,11) and (572.67,18.39) .. (572.67,27.5) .. controls (572.67,36.61) and (565.28,44) .. (556.17,44) .. controls (547.05,44) and (539.67,36.61) .. (539.67,27.5) -- cycle ;
\draw  [color={rgb, 255:red, 31; green, 119; blue, 180 }  ,draw opacity=1 ][fill={rgb, 255:red, 255; green, 255; blue, 255 }  ,fill opacity=1 ][line width=3]  (639.67,50.83) .. controls (639.67,41.72) and (647.05,34.33) .. (656.17,34.33) .. controls (665.28,34.33) and (672.67,41.72) .. (672.67,50.83) .. controls (672.67,59.95) and (665.28,67.33) .. (656.17,67.33) .. controls (647.05,67.33) and (639.67,59.95) .. (639.67,50.83) -- cycle ;
\draw  [color={rgb, 255:red, 31; green, 119; blue, 180 }  ,draw opacity=1 ][fill={rgb, 255:red, 255; green, 255; blue, 255 }  ,fill opacity=1 ][line width=3]  (358.33,74.83) .. controls (358.33,65.72) and (365.72,58.33) .. (374.83,58.33) .. controls (383.95,58.33) and (391.33,65.72) .. (391.33,74.83) .. controls (391.33,83.95) and (383.95,91.33) .. (374.83,91.33) .. controls (365.72,91.33) and (358.33,83.95) .. (358.33,74.83) -- cycle ;
\draw  [color={rgb, 255:red, 31; green, 119; blue, 180 }  ,draw opacity=1 ][fill={rgb, 255:red, 255; green, 255; blue, 255 }  ,fill opacity=1 ][line width=3]  (356.33,201.5) .. controls (356.33,192.39) and (363.72,185) .. (372.83,185) .. controls (381.95,185) and (389.33,192.39) .. (389.33,201.5) .. controls (389.33,210.61) and (381.95,218) .. (372.83,218) .. controls (363.72,218) and (356.33,210.61) .. (356.33,201.5) -- cycle ;
\draw  [color={rgb, 255:red, 31; green, 119; blue, 180 }  ,draw opacity=1 ][fill={rgb, 255:red, 255; green, 255; blue, 255 }  ,fill opacity=1 ][line width=3]  (487.67,321.5) .. controls (487.67,312.39) and (495.05,305) .. (504.17,305) .. controls (513.28,305) and (520.67,312.39) .. (520.67,321.5) .. controls (520.67,330.61) and (513.28,338) .. (504.17,338) .. controls (495.05,338) and (487.67,330.61) .. (487.67,321.5) -- cycle ;
\draw  [color={rgb, 255:red, 31; green, 119; blue, 180 }  ,draw opacity=1 ][fill={rgb, 255:red, 255; green, 255; blue, 255 }  ,fill opacity=1 ][line width=3]  (583,300.17) .. controls (583,291.05) and (590.39,283.67) .. (599.5,283.67) .. controls (608.61,283.67) and (616,291.05) .. (616,300.17) .. controls (616,309.28) and (608.61,316.67) .. (599.5,316.67) .. controls (590.39,316.67) and (583,309.28) .. (583,300.17) -- cycle ;
\draw  [color={rgb, 255:red, 31; green, 119; blue, 180 }  ,draw opacity=1 ][fill={rgb, 255:red, 255; green, 255; blue, 255 }  ,fill opacity=1 ][line width=3]  (611,177.5) .. controls (611,168.39) and (618.39,161) .. (627.5,161) .. controls (636.61,161) and (644,168.39) .. (644,177.5) .. controls (644,186.61) and (636.61,194) .. (627.5,194) .. controls (618.39,194) and (611,186.61) .. (611,177.5) -- cycle ;
\draw  [color={rgb, 255:red, 31; green, 119; blue, 180 }  ,draw opacity=1 ][fill={rgb, 255:red, 255; green, 255; blue, 255 }  ,fill opacity=1 ][line width=3]  (475.33,185.5) .. controls (475.33,176.39) and (482.72,169) .. (491.83,169) .. controls (500.95,169) and (508.33,176.39) .. (508.33,185.5) .. controls (508.33,194.61) and (500.95,202) .. (491.83,202) .. controls (482.72,202) and (475.33,194.61) .. (475.33,185.5) -- cycle ;

\draw (521.83,181.5) node  [font=\Huge]  {$v$};
\draw (402.83,320.5) node  [font=\Huge]  {$\mathcal{N}_{v}$};

\end{tikzpicture}}
    \caption{Sampling neighbors allows to keep the computational complexity of the aggregation function fixed and, sometimes, provides better generalization performances.}
    \label{fig:sampling}
\end{figure}

From the above equations, it emerges how the time complexity of Deep Graph Networks is strictly related to the number of edges in the input graphs. Nevertheless, trying to scale DGNs to graphs with billions of edges poses two major challenges: first, the training time required is often unbearable for modest computing devices; secondly, the degree of some vertices is so high that aggregating all neighboring states can lead to numerical instability or oversmoothing, subject to the permutation invariant operator used. To mitigate these issues, \textbf{sampling} techniques have been proposed to reduce the set of neighbors to aggregate for each vertex. The idea is schematically represented in Figure \ref{fig:sampling}, and it has been adopted by architectures like FastGCN \cite{chen_fastgcn_2018} and GraphSAGE \cite{hamilton_inductive_2017} to improve the generalization performances on different tasks.
Moreover, sampling is not necessarily constrained to the immediate neighbors of a vertex: one can provide a more flexible notion of neighborhood, such as \quotes{all vertices at distance 2}, and sample from that set \cite{hamilton_inductive_2017}. This way, a wider and richer neighborhood can be explored, similarly to the random walks technique briefly mentioned at the end of last chapter.

To conclude this section, we provide a table with different neighborhood aggregation schemes under the same uniform mathematical notation introduced so far. Note that one can also combine different aggregations together to increase the expressiveness of the model; this approach is adopted by models like Principal Neighborhood Aggregation \cite{corso_principal_2020}.

\begin{table}[ht]
\centering
\small
\renewcommand{\arraystretch}{2.0}
\begin{tabular}{ll}
Model          & Neighborhood Aggregation  \\ \hline
NN4G \cite{micheli_neural_2009}      &   $\sigma \Big(\mathbf{w}^{\ell +1^T}\mathbf{x}_v + \sum_{i=0}^{\ell}\sum_{c_k \in \mathcal{C}} \sum_{u \in \mathcal{N}^{c_k}_v} w_{c_k}^i * \mathbf{h}_u^i \Big)$ \\
GNN \cite{scarselli_graph_2009}       &  $\sum_{u \in \mathcal{N}_v} MLP^{\ell+1}\Big(\mathbf{x}_u, \mathbf{x}_v, \mathbf{a}_{uv}, \mathbf{h}_u^\ell \Big) $  \\
GraphESN \cite{gallicchio_graph_2010}   &  $\sigma \Big(\mathbf{W}^{\ell+1}\mathbf{x}_u +\hat{\mathbf{W}}^{\ell+1}[\mathbf{h}^{\ell}_{u_1}, \dots, \mathbf{h}^{\ell}_{u_{\mathcal{N}_v}}]\Big) $  \\
GCN \cite{kipf_semi-supervised_2017}       &     $\sigma \Big(\mathbf{W}^{\ell +1} \sum_{u \in \mathcal{N}(v)} L_{vu}\mathbf{h}^\ell_u \Big)$ \\
GAT \cite{velickovic_graph_2018}      &   $\sigma\Big(\sum_{u \in \mathcal{N}_v} \alpha_{uv}^{\ell+1}*\mathbf{W}^{\ell+1}\mathbf{h}_u\Big)$     \\
ECC \cite{simonovsky_dynamic_2017} &    $\sigma\Big( \frac{1}{|\mathcal{N}_v|}\sum_{u \in \mathcal{N}_v} MLP^{\ell+1}(\mathbf{a}_{uv})^T \mathbf{h}_u^\ell \Big) $   \\
R-GCN \cite{schlichtkrull_modeling_2018}       &    $\sigma \Big(\sum_{c_k \in \mathcal{C}}\sum_{u \in \mathcal{N}_v^{c_k}} \frac{1}{|\mathcal{N}_v^{c_k}|}\mathbf{W}_{c_k}^{\ell +1} \mathbf{h}_u^\ell + \mathbf{W}^{\ell +1} \mathbf{h}_v^\ell\Big)$  \\
GraphSAGE \cite{hamilton_inductive_2017} &  $\sigma\Big(\mathbf{W}^{\ell+1}(\frac{1}{|\mathcal{N}_v|}[\mathbf{h}_v^\ell, \sum_{u \in \mathcal{N}_v} \mathbf{h}_u^\ell])\Big)  $  \\
CGMM \cite{bacciu_contextual_2018,bacciu_probabilistic_2020,atzeni_modeling_2021}      &  $ \sum_{i=0}^{\ell} w^i * \Big(\sum_{c_k \in \mathcal{C}} w_{c_k}^i * \big( \frac{1}{|\mathcal{N}_v^{c_k}|}\sum_{u \in \mathcal{N}^{c_k}_v} \mathbf{h}_u^i \big) \Big) $ \\
GIN \cite{xu_how_2019}       & $MLP^{\ell+1} \Big( \big(1 + \epsilon^{\ell+1} \big)\mathbf{h}_v^{\ell} + \sum_{u \in \mathcal{N}_v} \mathbf{h}_u^\ell \Big)$   \\
\end{tabular}
\caption{Here are some preeminent examples of neighborhood aggregation schemes  present in the literature. We use square brackets to denote concatenation, whereas $W, w$ and $\epsilon$ are weight to be learned. GraphESN's aggregation looks different because it assumes a maximum size of the neighborhood, but the core principles are the same. Also, we describe the \textit{mean} version of GraphSAGE, though variations are possible.}
\label{tab:summary-aggregation}
\end{table}

\subsubsection*{Graph Coarsening}
\begin{figure}[ht]
    \hspace{-2cm}
    \centering
    \resizebox{0.8\textwidth}{!}{\input{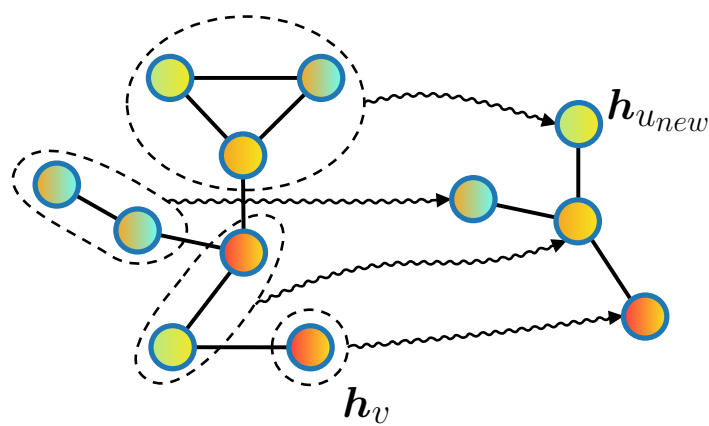}}
    \caption{A pooling layer coarsens vertex representations to obtained a reduced graph representation that should encode higher-level details.}
    \label{fig:pooling}
\end{figure}

Separate from neighborhood aggregation, graph \textbf{pooling} is an (optional) independent module of a deep feedforward architecture that coarsens the latent graph representations in order to reduce the number of vertices. The purpose of graph pooling is three-fold, \ie discover communities in the input graph, encode such knowledge in the vertex states, and finally to reduce the computational costs of the subsequent neighborhood aggregation modules. We can distinguish between \textit{adaptive} pooling, whose parameters are learned using gradient descent techniques, and \textit{topological} pooling, which leverages the topological properties of the graph with known non-adaptive algorithms. The idea of pooling is sketched in Figure \ref{fig:pooling}.

Adaptive pooling applies a differentiable transformation to vertex representations in order to produce soft cluster assignment scores, such as DiffPool \cite{ying_hierarchical_2018}. The drawback of these methods is that they produce a dense adjacency matrix in output, which can become even more costly than the original graph if the number of chosen clusters is sufficiently large. Other approaches like Top-k pooling \cite{gao_graph_2019} try to address this problem by retaining only the top-k vertices according to some ranking. Also, adaptive pooling techniques are not restricted to vertices, rather they can be applied to edges \cite{frederik_diehl_towards_2019} by collapsing vertices incident to the highest ranking edges.

Topological pooling, on the other hand, is non-adaptive  and inspired by classical community discovery algorithms. The striking computational advantage of this family of pooling methods is that coarsened graphs can often be \textit{precomputed}, thus significantly reducing the computational burden of the subsequent training phase. Among them, we mention spectral clustering approaches such as GRACLUS \cite{dhillon_weighted_2007} and ARMA filters \cite{bianchi_graph_2021}, as well as methods based on non-negative matrix factorization \cite{bacciu_non-negative_2019} and the k-plex cover algororithm \cite{bacciu_k_2020}.

Recently, there has been some criticism about the true benefits of pooling for graph classification on small datasets \cite{mesquita_2020_rethinking}. Local pooling was shown to become progressively invariant to cluster assignments of vertices, and simple baselines performed as well as methods employing pooling layers. Nevertheless, pooling can still be used to detect some form of community in the graph when we know there exists a latent hierarchy.

\subsubsection*{Global Aggregation}
Whenever the task requires it, it may be necessary to aggregate all vertex representations to produce a single graph state summarizing all the information extracted by the model. Because there exists no topological ordering among vertices in general, we almost always rely on another permutation invariant function to compute the graph state at each iteration $\lcurr$:
\begin{align}
    \boldhell{g}{\lcurr} = \Psi \Big( \{ f(\boldhell{v}{\lcurr}) \mid v \in \Vset{g} \} \Big).
\end{align}
Common choices for $f$ and $\Psi$ are the identity function and the element-wise sum, mean or max operators, even though nothing prevents us from using approximations of universal aggregators over multisets \cite{zaheer_deep_2017,wagstaff_limitations_2019}. In this manuscript, we will consider a graph representation that is the layer-wise concatenation of all graph states, in order to consider multiple \quotes{views} of the graph extracted by the model. Alternatives are possible though: \cite{li_gated_2016} applies a Long Short-Term Memory (LSTM) \cite{hochreiter_long_1997} to the sequence of graph states $\{\boldhell{g}{0},\dots,\boldhell{g}{\lcurr},\dots\}$, whereas Sort Pooling picks a subset of vertex states according to a lexicographic ordering of such states \cite{zhang_end--end_2018}.

\begin{figure}[ht]
    \hspace{-2cm}
    \centering
    \resizebox{1\textwidth}{!}{\input{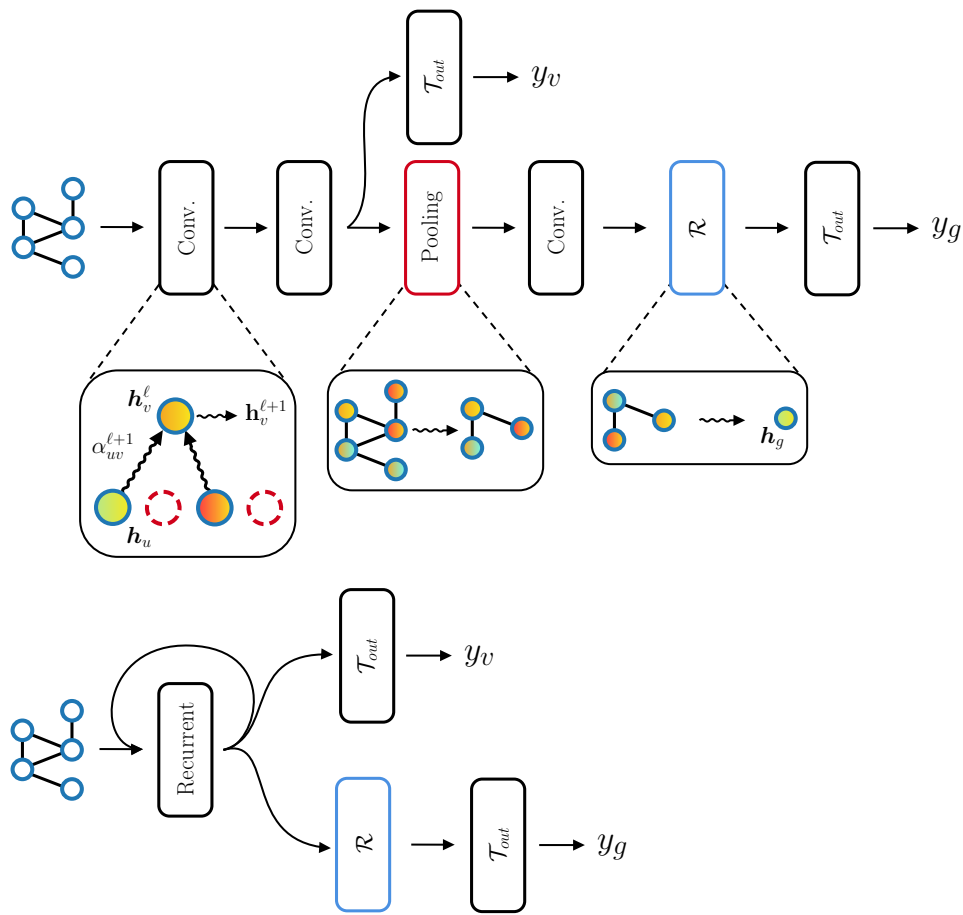}}
    \caption{Putting everything together: a comparative example between feedforward (top) and recurrent (bottom) DGNs. See the text for more details.}
    \label{fig:feedforward-vs-recurrent}
\end{figure}

To summarize the building blocks introduced so far, Figure \ref{fig:feedforward-vs-recurrent} sketches a comparison between a feedforward architecture and a recurrent model for graphs, where branches indicate that we are either solving vertex or graph related tasks. Note how the feedforward network uses differently parametrized layers, in contrast to the single layer of a recurrent network. The application of pooling and of a global transduction can only occur in graph related tasks, since the topology of the output is irremediably changed by these operations. On top of these architectures, which implement the transductions $\T_{end}$ and, optionally, $\mathcal{R}$ of Figure \ref{fig:high-level}, we can apply an output module implementing the $\T_{out}$ transduction.

\subsection{Learning Criteria}

Throughout the following chapters, we will deal with both unsupervised and supervised learning tasks such as maximum likelihood estimation on graphs, link prediction, vertex and graph classification. Therefore, it is useful to revise how researchers have approached these tasks in the past.

\subsubsection*{Unsupervised Learning}

The \textbf{maximum likelihood estimation} criterion is just one of the ways in which we can train a DGN in an unsupervised fashion. For instance, one can maximize the likelihood of all vertex features conditioned on the graph information, as described in detail in the following chapter. Whenever the graphical model does not allow a tractable computation of the likelihood, variational approximations can come in handy to train the model in a reasonable amount of time \cite{kipf_variational_2016,qu_gmnn_2019}. As an alternative, reasearchers have also proposed ways to directly capture the distribution of a graph $P(g|\theta)$, by combining a graph encoder with a radial basis function network, as well as providing a definition of \textbf{attributed} random graphs \cite{trentin_maximum_2009,bongini_recursive_2018}.

Predicting the existence of a link between entities can be very helpful in application domains such as drug repurposing and recommender systems \cite{ma_memory_2020,nguyen_graphdta_2020,xu_graphsail_2020}. \textbf{Link prediction} tasks use the structure of the graphs in the dataset to train a model to reconstruct the adjacency matrix and predict missing links, so in some sense they correspond to a form of \textit{self-supervision}. The most adopted loss is called \textbf{reconstruction loss}, adapted from auto-encoders:
\begin{align*}
    \mathcal{L}_{rec}(g) = \sum_{\puv}||\boldh{v} - \boldh{u}||^2,
\end{align*}
which also has a probabilistic formulation \cite{kipf_variational_2016}:
\begin{align*}
P( \puv \in \Eset{g} \mid \boldh{u}, \boldh{v}) = \sigma(\boldh{u}^T\boldh{v}).
\end{align*}
It is important to remark that the reconstruction loss does not take into account vertex or edge features, and that it is mainly based on the \textbf{homophily assumption}: adjacent vertices are more likely to share the same characteristics and/or target label \cite{macskassy_classification_2007}. In tasks where the homophily assumption is verified, it might be advantageous to use this loss as a \textit{regularizer}.

Another relevant unsupervised criterion is that of \textbf{graph clustering}, \ie partitioning a graph into groups of vertices that share some similarity. Self-Organizing Maps (SOMs) \cite{kohonen_self_1990} have been one of the most successful approaches to perform clustering on DAGs \cite{hagenbuchner_self_2003,hammer_general_2004,hammer_recursive_2004}, before being extended to cyclic graphs \cite{neuhaus_self_2005,hagenbuchner_graph_2009,wang_mgae_2017,bongini_recursive_2018}.

Borrowing ideas from information theory, \textbf{mutual information} approaches try to create representations that approximately maximize the mutual information between pairs of graphs. Deep Graph Infomax (DGI) \cite{velickovic_deep_2019} trains a discriminator to distinguish between a graph and its corrupted version, where the corruption algorithm must be defined a priori. The authors show that this kind of training indirectly maximizes the mutual information between graphs in the dataset. Similarly, the \textbf{entropy} of a categorical distribution can be used to regularize the soft-clustering matrix produced by adaptive pooling techniques such as DiffPool. Entropy can be used to encourage the pooling layer to produce one-hot assignments to the clusters for each vertex, even though it does not solve the \quotes{dense output adjacency matrix} problem.

\subsubsection*{Supervised Learning}

The most common supervised tasks in which DGNs are involved are vertex classification, graph classification, and graph regression. In \textbf{vertex classification}, the goal is to assign a target label to each vertex in the graph. We also distinguish between \textit{inductive} vertex classification, where the vertices to classify belong to unseen graphs, and \textit{tranductive} vertex classification, which consists of vertex predictions on a single (large) graph with some known vertex targets. Transductive vertex classification has suffered from serious reproducibility issues, due to arbitrary choices in the experimental setups regarding the same tasks \cite{shchur_pitfalls_2018}; in particular, it emerged that the data split was highly influential on these datasets. 
While there is a large stream of works around this \quotes{semi-supervised} setting, Bayesian approaches seem the more robust ones when few training labels are available \cite{zhang_bayesian_2019}. In particular, one can implement a Bayesian version of the GCN model and marginalize over the neural network parameters when it comes to predict a new vertex label. This method can be extended to a nonparametric scenario, giving rise to a Bayesian version of the Variational Graph Autoencoder \cite{kipf_variational_2016} that better models the adjacency matrix generation by taking into account noisy data \cite{pal_non_2020}.

As regards \textbf{graph classification and regression}, one usually applies a standard \ML{} predictor on top of the graph representation computed using the techniques described earlier. The objectives to be minimized are the usual Cross-Entropy (CE) loss or the Mean Squared Error (MSE) Notwithstanding the simplicity of this approach, we noticed the same troubling trends in scholarship on a series of graph classification benchmarks \cite{errica_fair_2020}. Therefore, a contribution of this thesis, which we shall introduce in a moment, will be about addressing such issues.

\subsubsection*{Self-supervised Learning}
Recently, many \textbf{self-supervision} objectives have been proposed to pretrain Deep Neural Graph Networks \cite{hu_strategies_2020}. These include predicting the surrounding neighborhood information of vertices and using attribute masking strategies similar to those employed in the Natural Language Processing field \cite{devlin_bert_2019} More generally, approaches to self-supervised learning for graphs can be divided into those that generate the feature and adjacency matrices, others than exploit contrastive learning based on information theory criteria, some that try to predict known properties of the graphs under consideration, and hybrid ones \cite{liu_graph_2021}.

\subsubsection*{Summary}
We have presented all the basic building blocks that allow us to reason about the main differences between deep architectures for graphs. To complement the discussion, we present a recap of the main properties of some DGNs in the literature. Specifically, Table \ref{tab:summary-properties} divides models by their style of context propagation, learning task, how layers are chosen, their intrinsic nature, and building blocks.

Broadly speaking, whether it is crafting a new DGN or choosing an existing one, one always has to keep in mind the characteristics of the task at hand, the available data, and the computational constraints. We can make a trivial example when considering the inductive bias of the neighborhood aggregation mechanism: if the task benefited from knowing the degree of each vertex, then a sum-based aggregation function would be the obvious choice. In contrast, when one wants to capture how neighboring representations are distributed around each vertex, a mean-based aggregation is an adequate, not to mention numerically stable, alternative. When the amount of supervised labels is limited, expressive models like GIN are prone to severely overfit the training data \cite{xu_how_2019}, so either one carefully applies a regularization strategy or a simpler, less parametrized aggregation like the ones of GraphSAGE or GCN is used. In cases where the amount of raw graphs is also much larger than the supervised samples or we need to quickly adapt to new tasks without retraining the entire model, unsupervised vertex/graph embedding mechanisms such as CGMM (Chapter \ref{chapter:dbgn}) may be a viable technique to consider. In terms of computational requirements, taking into account edges and their features usually leads to an increase in the parameters of each DGN. This is because models like GAT apply multiple adaptive, non-linear transformations to each pair of adjacent vertex representations to compute an importance score for each neighbor, whereas others such as ECC transform each edge feature vector by means of an MLP. When applied to large and dense graphs, these models quickly become computationally demanding (see, for instance, Section \ref{sec:scholarship-issues}). Here, exploiting any domain knowledge could be important to avoid such parametrizations and inject a favorable inductive bias into the model; as an example, if an atom is connected to others in space, the inverse of the inter-atomic distance could be exploited to diminish the importance of far-away atoms when aggregating neighbors in case the homophily assumption holds. More in general, whenever training is considered prohibitive because of hardware constraints and an extremely efficient solution is needed, GraphESN and FDGNN provide an advantageous trade-off between performances and a fully supervised training approach, and they can be considered good baselines against which to compare because of the randomized, untrained nature of the embedding construction.
To conclude these consideration, it has to be mentioned that very few works in the field have tried to automatically determine the \quotes{right} number of graph convolutional layers to use for the underlying task during training. In this sense, NN4G also stands out as a pioneering approach that applies the principle of Cascade Correlation to tackle the task in a divide-et-impera fashion, by training one layer at a time. We believe that this could be an understudied research direction, especially as regards unsupervised and self-supervised methods that could mitigate the need of cross-validating this crucial hyper-parameter.

\begin{table}[t]
\footnotesize
\begin{tabularx}{\textwidth}{XXXXX}
Model & Context & Embedding & Layers   &  Nature  \\ \hline
GNN \cite{scarselli_graph_2009}     & Recurrent & Supervised & Single & Neural  \\
NN4G \cite{micheli_neural_2009}     & Constructive   & Supervised  & Adaptive & Neural \\
GraphESN \cite{gallicchio_graph_2010} & Recurrent & Untrained & Single & Neural  \\
GCN \cite{kipf_semi-supervised_2017}     & Feedforward    &    Supervised          &    Fixed        &  Neural   \\
GG-NN \cite{li_gated_2016} & Recurrent    &    Supervised          &    Fixed        &  Neural   \\
ECC \cite{simonovsky_dynamic_2017}     &        Feedforward         &  Supervised    &  Fixed & Neural \\
GraphSAGE\cite{hamilton_inductive_2017}     &        Feedforward         &  Both    &      Fixed & Neural \\
CGMM \cite{bacciu_contextual_2018,bacciu_probabilistic_2020}     &       Constructive         &  Unsupervised    &      Fixed & Probabilistic \\
E-CGMM \cite{atzeni_modeling_2021}     &       Constructive         &  Unsupervised    &      Fixed & Probabilistic \\
iCGMM (\ref{sec:icgmm})           &      Constructive         &  Unsupervised    &      Fixed & Probabilistic \\
DGCNN \cite{zhang_end--end_2018}    &        Feedforward     &  Supervised    &      Fixed & Neural \\
DiffPool \cite{ying_hierarchical_2018}     &        Feedforward         &  Supervised    &  Fixed & Neural \\
GAT \cite{velickovic_graph_2018}    & Feedforward    &    Supervised          &       Fixed           & Neural          \\
R-GCN \cite{schlichtkrull_modeling_2018}    &       Feedforward         &  Supervised    &      Fixed & Neural \\
DGI \cite{velickovic_deep_2019}     &   Feedforward    &    Unsupervised          &       Fixed           & Neural \\
GMNN \cite{qu_gmnn_2019}     &   Feedforward    &    Both          &       Fixed           & Hybrid \\
GIN \cite{xu_how_2019}     &   Feedforward    &    Supervised          &       Fixed           & Neural \\
NMFPool \cite{bacciu_non-negative_2019}     &        Feedforward         &  Supervised    &  Fixed & Neural \\
SAGPool \cite{lee_self-attention_2019}     &        Feedforward         &  Supervised    &  Fixed & Neural \\
Top-k Pool \cite{gao_graph_2019}     &        Feedforward         &  Supervised    &  Fixed & Neural \\
FDGNN \cite{gallicchio_fast_2020}     &   Recurrent    &    Untrained      &       Fixed           & Neural \\
GMDN \cite{errica_graph_2021}     &   Feedforward    &    Supervised      &       Fixed           & Hybrid \\ \\
\end{tabularx}
\begin{tabularx}{\textwidth}{XXXXX}
Model & Edges        & Pooling  &                    Attention                  & Sampling                   \\  \hline
GNN \cite{scarselli_graph_2009}    & Continuous    & \ding{55} & \ding{55} & \ding{55} \\
NN4G \cite{micheli_neural_2009}     &Discrete    & \ding{55} & \ding{55} & \ding{55} \\
GraphESN \cite{gallicchio_graph_2010}    & \ding{55} & \ding{55} & \ding{55} & \ding{55} \\
GCN \cite{kipf_semi-supervised_2017}     &   \ding{55}             &    \ding{55}    &            \ding{55}    &   \ding{55} \\
GG-NN \cite{li_gated_2016} &  \ding{55}             &    \ding{55}    &            \ding{55}    &   \ding{55} \\
ECC \cite{simonovsky_dynamic_2017}     &        Continuous         &  Topological    &      \ding{55} & \ding{55} \\
GraphSAGE\cite{hamilton_inductive_2017}    &        \ding{55}         &  \ding{55}    &      \ding{55} & \ding{51} \\
CGMM \cite{bacciu_contextual_2018,bacciu_probabilistic_2020}     &        Discrete       &  \ding{55}    &      \ding{55} & \ding{55} \\
E-CGMM \cite{atzeni_modeling_2021}     &        Continuous       &  \ding{55}    &      \ding{55} & \ding{55} \\
iCGMM (\ref{sec:icgmm})     &        \ding{55}       &  \ding{55}    &      \ding{55} & \ding{55} \\
DiffPool \cite{ying_hierarchical_2018}    &        -         & Adaptive   &     -    &    - \\
DGCNN \cite{zhang_end--end_2018}     &        \ding{55}         & Topological   &      \ding{55} & \ding{55} \\
GAT \cite{velickovic_graph_2018}     &        \ding{55}         &  \ding{55}    &      \ding{51} & \ding{55} \\
R-GCN \cite{schlichtkrull_modeling_2018}     &        Discrete         &  \ding{55}    &      \ding{55} & \ding{55} \\
GMNN \cite{qu_gmnn_2019} &       -         &  -    &     - & - \\
DGI \cite{velickovic_deep_2019}     &        \ding{55}         &  \ding{55}    &      \ding{55} & \ding{51} \\
GIN \cite{xu_how_2019}     &        \ding{55}         &  \ding{55}    &      \ding{55} & \ding{55} \\
NMFPool \cite{bacciu_non-negative_2019}    &        -         & Topological   &     -    &    - \\
SAGPool \cite{lee_self-attention_2019}    &        -         & Adaptive   &     -    &    - \\
Top-k Pool \cite{gao_graph_2019}    &        -         & Adaptive   &     -    &    - \\
FDGNN \cite{gallicchio_fast_2020}  &     \ding{55}    &     \ding{55}          &         \ding{55}           &   \ding{51} \\
GMDN \cite{errica_graph_2021}     &   -    &    -      &       -           & - \\ \\
\end{tabularx}
\caption{Recap of DGNs' properties. When the symbol \quotes{-} is used, we mean \quotes{not applicable}, as the row refers to a framework rather than a single model.}
\label{tab:summary-properties}
\end{table}
\clearpage

\section[Scholarship Issues in Graph Classification]{Scholarship Issues in Graph Classification \cite{errica_fair_2020}}
\label{sec:scholarship-issues}
Experimental reproducibility and replicability are core aspects of empirical \ML{}. From time to time, researchers have warned their communities about flaws in scholarship regarding streams of scientific publications \cite{mcdermott_artificial_1976,lipton_troubling_2018,NAS_reproducibility_2019}. However, trying to correct bad practices does not take just one publication, rather a collective effort is required to acknowledge the current issues and take immediate action. Common examples of these troubling trends are the ambiguous or poorly detailed experimental settings, unfair comparison due to the use of different data features and/or data splits, cherry-picking of hyper-parameters on the basis of test set performances, and the impossibility of reproducing the results using the code provided by the authors. In turn, this means we are often unable to confidently assess which empirical methodology performs best on a given learning task.

The situation is not much different in the graph learning community. After the recent re-discovery of the core ideas and the subsequent exponential growth of scientific publications, it can be argued that little attention was devoted to ensure a fair and robust model assessment between models. A striking example can be found in some vertex classification benchmarks, where it was found that the use of different training/validation/test splits could completely alter the final performance ranking \cite{shchur_pitfalls_2018}. Indeed, in some papers, data splits were generated at random simply because there existed no common agreement on the evaluation criteria. Similarly, concerns have been raised about neural recommender systems, most of which cannot perform better than a very simple baseline \cite{dacrema_are_2019}.

This section describes our attempt to mitigate the lack of standardization of empirical comparisons in the \textbf{graph classification} scenario \cite{errica_fair_2020}. As a matter of fact, we observed that many practitioners did not provide thorough information about the two main steps of any \ML{} evaluation, namely \textbf{model selection} and \textbf{risk assessment}. Failure to keep these phases well separated often leads to over-optimistic estimates of the generalization performances, but it also generates confusion and doubts among other researchers while building on previous results; this can easily mislead them into repeating the same methodological errors. Before continuing, let us briefly recall the basics of risk assessment and model selection.

\paragraph*{Risk Assessment.} To provide an estimate of the generalization performance of each model, a risk assessment procedure has to be followed. Risk assessment relies on a test set that must be used only after the chosen model configuration has been trained. If a test set is not given in advance, one can adopt a simple holdout split or, like we did, a \emph{k-fold Cross Validation} (CV) \citep{stone_cross_1974,varma_bias_2006,cawley_overfitting_2010} scheme to generate $k$ different training/test partitions (called \textbf{folds}) of the dataset. For each partition, we should perform an internal model selection procedure (based on the training set \textbf{only}) that picks the best hyper-parameters for \emph{that specific partition}. This way, test data is \textbf{never} used to select the hyper-parameters, such as number of epochs, layers or hidden units. Note that, as model selection is performed independently for each fold, we obtain $k$ different \quotes{best} configurations. This is why one should talk about the performance of the \textit{class} of models rather than a single configuration. We would like to stress here that the best configuration overall does \textbf{not} exist because of the No Free Lunch Theorem \cite{wolpert_lack_1996}.

\paragraph*{Model Selection.} Inside each fold, the selection of the best hyper-parameters usually happens via another holdout strategy or an inner $k$-fold CV, where this time the \quotes{outer} training data is further partitioned into training and validation sets (unless a validation set is already available). Because the best hyper-parameters' configuration is selected on the basis of validation performances, the key thing to remember is that these results are \textit{biased} estimates of the true generalization capabilities of a model. Hence, it would be trivial to obtain state-of-the-art results by comparing models
on validation performances: just find the configuration that maximizes the performance metric on the validation set. This is a bad practice we clearly want to avoid.

\subsection{Chosen Criteria}
Similarly to what has been done in \cite{dacrema_are_2019}, we first listed some relevant requirements for reproducibility: \emph{i}) code for data preprocessing, model selection, and risk assessment is provided; \emph{ii}) data splits are available; \emph{iii}) data is split according to a stratification technique that preserves class proportions across all folds; \emph{iv}) results are reported using standard deviations, and they refer to model evaluation (test set) rather than model selection (validation set).

We then selected the DGNs to re-evaluate according to basic principles: \emph{i)} their graph classification performance obtained using a 10-fold cross validation; \emph{ii)} peer-reviewed status; \emph{iii)} architectural differences; \emph{iv)} popularity. We ended up choosing DGCNN \cite{zhang_end--end_2018}, DiffPool \cite{ying_hierarchical_2018}, ECC \cite{simonovsky_dynamic_2017}, GIN \cite{xu_how_2019}, and GraphSAGE \cite{hamilton_inductive_2017}, though the latter was not applied to graph classification tasks in the original paper. Table \ref{tab:reproducibility} summarizes our findings.
\begin{table}[t]
\begin{center}
\begin{tabular}{l c c c c}
    \toprule
    & \textbf{DGCNN} & \textbf{DiffPool} & \textbf{ECC} & \textbf{GIN}\\
    \midrule
    Data preprocessing code     & Y    & Y    & -     & Y \\
    Model selection code        & N    & N    & -     & N \\
    Model evaluation code       & Y    & Y    & -     & Y \\
    Data splits provided        & Y    & N    & N     & Y \\
    Label Stratification        & Y    & N    & -     & Y \\
    Report accuracy on test     & Y    & A    & A     & N \\
    Report standard deviations  & Y    & N    & N     & Y \\
    \bottomrule
\end{tabular}
\caption{Criteria for reproducibility considered in this work and their compliance among the considered models. (Y) indicates that the criterion is met, (N) indicates that the criterion is not satisfied, (A) indicates ambiguity (i.e. it is unclear whether the criteria is met or not), (-) indicates lack of information.}
\label{tab:reproducibility}
\end{center}
\end{table}

From the table, it seems that some of the most popular models from the literature did not meet all the listed criteria that would foster empirical reproducibility. Let us now expand the discussion about each model by highlighting the problems we found.

\paragraph*{DGCNN} In this paper, the architecture was fixed for all datasets. Although sub-optimal, learning rate and number of training epochs were tuned using only one of the 10 folds and then reused on all the other folds. We could not find the code to perform model selection despite the rest of it being publicly available. Moreover, the authors ran the 10-fold CV procedure 10 times\footnote{This was computationally feasible since model selection is performed only once per CV.} and reported the average of the 10 final scores, each of which had already been averaged over the 10 folds. As a result, the variance of the provided estimates was greatly reduced. This experimental setup, however, was different from the one used in other works, and thus we cannot reliably assess the variance of the models under the same setting.

\paragraph*{DiffPool} From both the paper and the provided code, it is unclear if reported results were obtained on the test set rather than the validation set. The authors stated that 10-fold CV was used, but standard deviations were not reported. There are some statements in the paper about applying early stopping on the validation set, but neither model selection code nor validation splits were made available. We also found that target stratification was not applied to the data splits and no random seed was set, hence we can assume the generated data splits were different each time the code was being executed.

\paragraph*{ECC} As in DiffPool, the paper lacks standard deviation values in the results. Likewise DGCNN, hyper-parameters were fixed in advance, hence it is not clear if and how model selection was performed. Importantly, there are no references in the code repository to data pre-processing, data stratification, data splitting, and model selection. This makes ECC the least reproducible model among those considered.

\paragraph*{GIN} Here we observed another kind of troubling trend. The authors did a good job in listing the ranges of hyper-parameters tried. However, as stated explicitly in the paper and in the public review discussion, they report the mean \textit{validation} accuracy of a 10-fold CV. In other words, the reported results refer to model selection and not to risk assessment. Furthermore, the code for model selection is not provided.

\paragraph*{GraphSAGE} This model is often used in other papers as a strong baseline \cite{ying_hierarchical_2018,xu_how_2019}. Nonetheless, the code to reproduce such experiments on graph classification has never been provided.

\paragraph*{Summary} It is this ample empirical inconsistency that has motivated a re-evaluation of these models within a rigorous, reproducible and fair environment. Our code has been publicly released alongside the data splits.\footnote{\url{https://github.com/diningphil/gnn-comparison}.}

\subsection{Experimental Setting}
\label{subsec:iclr-experimental-setting}
The assessment of the above models is carried out on 9 graph classification datasets, four of which are chemical and five social. We considered D\&D \cite{dd}, PROTEINS \cite{proteins}, NCI1 \cite{nci1} and ENZYMES \cite{enzymes} as binary classification chemical tasks, whereas IMDB-BINARY, IMDB-MULTI, REDDIT-BINARY, REDDIT-5K, and COLLAB \cite{yanardag_deep_2015} are social benchmarks. We report the statistics of these datasets in Table \ref{tab:datasets-table-iclr}.

\begin{table}[ht]
    \small
    \begin{center}
    \begin{tabular}{c l c c c c c }
    \toprule
    &  & \# Graphs & \# Classes & \# Vertices & \# Edges & \# Vertex feat.\\
    \toprule
    \multirow{4}{*}{\rotatebox[origin=c]{90}{\textsc{Chem.}}}
    &D\&D            & 1178 & 2 & 284.32 & 715.66 & 89 \\
    &ENZYMES       &  600 & 6 &  32.63 &  64.14 &  3+18 \\
    &NCI1          & 4110 & 2 &  29.87 &  32.30 & 37 \\
    &PROTEINS      & 1113 & 2 &  39.06 &  72.82 &  3 \\
    \midrule
    \multirow{5}{*}{\rotatebox[origin=c]{90}{\textsc{Social}}}
    &COLLAB     & 5000 & 3 & 74.49 & 2457.78 &  - \\
    &IMDB-BINARY   & 1000 & 2 &  19.77 &  96.53 &  - \\
    &IMDB-MULTI    & 1500 & 3 &  13.00 &  65.94 &  - \\
    &REDDIT-BINARY & 2000 & 2 & 429.63 & 497.75 &  - \\
    &REDDIT-5K     & 4999 & 5 & 508.82 & 594.87 &  - \\
    \bottomrule
    \end{tabular}
    \end{center}
    \caption{Dataset statistics. Following the literature, we use both the 18 continuous and 3 discrete vertex attributes in the case of ENZYMES. All other vertex features belong to a finite and discrete alphabet representing atom types.} \label{tab:datasets-table-iclr}
\end{table}

As the reader can observe, the social datasets lack any kind of feature information about vertices or edges. For this reason, we will double the re-evaluations on the social tasks to consider two scenarios, that is, one in which vertex features hold a constant value $1$ and another in which the vertex degree is treated as the sole continuous feature. This way, we can study the effect of the inductive bias imposed by different realizations of a DGN's layer.

It is worth mentioning that in the past different choices for the vertex features have been made \cite{ying_hierarchical_2018,xu_how_2019}, but the competing models were rarely compared under the same conditions.

\paragraph*{Structure-agnostic Baselines} The importance of proper baselines has been mostly underrated when it comes to DGNs. Yet, designing a good baseline is of paramount importance to discern between real and fallacious progress. In our specific case, testing whether structural information truly is meaningful for the task is more than just a double-check, as we will see. When DGNs performances closely match those of a structure-agnostic baseline, we can draw two conclusions: either the task does not need topological information to be solved or the models we have developed are not \quotes{powerful} enough. Whilst one may involve domain experts to check if the former conclusion is valid, the latter is more involved as multiple factors come into play, such as the amount of training data, the structural inductive bias we imposed through the architecture, and the hyper-parameters tried. On the contrary, a significant boost in performances can only indicate that the graph topology is relevant to solve the task.

Therefore, we adopted distinct baselines for the two families of datasets. On chemical datasets, with the exception of ENZYMES, we follow \cite{ralaivola_graph_2005} and implement the Molecular Fingerprint technique. A Molecular Fingerprint is obtained by first applying a global \textit{sum} aggregation $\mathcal{R}$, \ie counting the occurrences of all atom types in the graph, followed by a single-layer MLP with ReLU activations that implements $\T_{out}$. Instead, on social domains and ENZYMES (due to the presence of additional features), we follow \cite{zaheer_deep_2017} and learn permutation-invariant functions over sets of vertices. This is done by first transforming the vertex features with a single-layer MLP, which are then aggregated via \textit{sum} operator and passed to another single-layer MLP for the final classification. Hence, none of these baselines exploit the information contained in the adjacency matrix.

\begin{figure}[ht]
    \centering
    \resizebox{\textwidth}{!}{\input{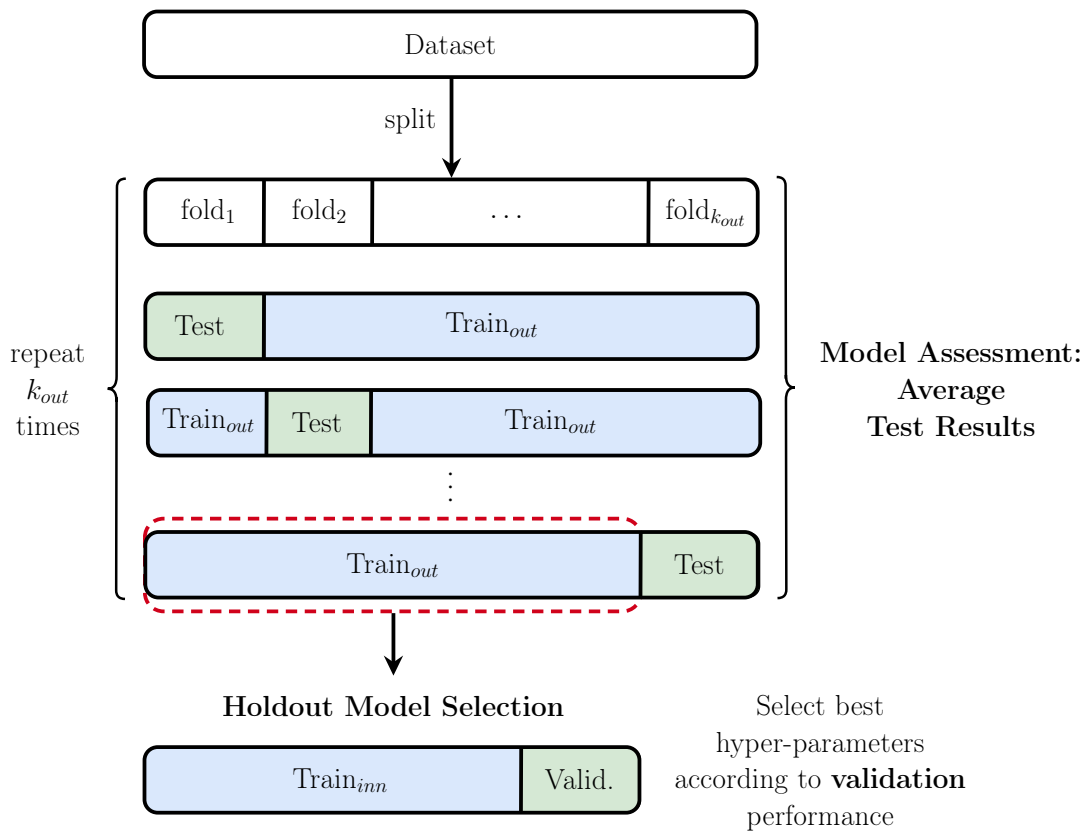}}
    \caption{The evaluation framework used for the re-evaluation of some DNGNs consists of a 10-fold cross validation for risk assessment, where we carry out an \textit{inner} holdout model selection \textbf{for each \textit{outer} fold}.}
    \label{fig:evaluation-scheme}
\end{figure}

\paragraph*{Setup and Hyper-parameters}
For the rest of the section, the evaluation setup shall consist of a 10-fold CV for risk assessment with a holdout model selection strategy inside each fold. This choice was made to keep the re-evaluation as close as possible to the procedure followed by most models. We also schematically represent it in Figure \ref{fig:evaluation-scheme}, where it becomes clearer how the computational requirements are proportional to $k_{out}$ and the number of configurations tried for each model selection.

The holdout strategy partitions the Train$_{out}$ set into Train$_{inn}$ and \textit{Validation} sets (90\% and 10\% respectively). After every model selection, the best configuration for the outer fold is re-trained 3 times on Train$_{out}$, holding out a random subset of the data (10\%) to perform early stopping. The separate training runs are needed to counteract the effect of an unfavorable random initialization on test performances. Finally, performances on the unseen test sets are averaged over the 3 final runs. We implemented early stopping \cite{prechelt_early_1998} with patience $n$, meaning we stop training when the validation metric of interest (the accuracy), has not improved for $n$ epochs. All data splits, with the exception of the random validation set in the final runs, have been precomputed, thus ensuring the models are cross-validated and assessed on the same data. Also, we applied a stratification strategy to ensure that class proportions are preserved. Table \ref{tab:exp-setup-iclr} summarizes what said so far.
\begin{table}[ht]
    \footnotesize
    \begin{minipage}{0.51\textwidth}
    \begin{algorithm}[H]
        \centering
        \caption{Model Assessment ($k$-fold CV)}
        \label{algo:model-assessment}
        \begin{algorithmic}[1]
            \State Input: Dataset $\mathcal{D}$, set of configurations $\Theta$
            \State Split $\mathcal{D}$ into $k$ folds $F_1,\dots, F_k$
            \For{$i\gets 1, \ldots, k$}
            \State train$_k$, test$_k$ $\gets$ $\big(\bigcup_{j \neq i}F_j\big)$, $F_i$
            \State best$_k$ $\gets$ Select(train$_k$, $\Theta$)
            \For{$r \gets 1, \ldots, R$}
                \State model$_r$ $\gets$ Train(train$_k$, best$_k$)
                \State p$_r$ $\gets$ Eval(model$_k$, test$_k$)
            \EndFor
            \State perf$_k$ $\gets$ $\sum_{r=1}^R \text{p}_r/R$
            \EndFor
            \State \textbf{return} $\sum_{i=1}^k \text{perf}_i/k$
        \end{algorithmic}
    \end{algorithm}\end{minipage}
    \hfill
    \begin{minipage}{0.46\textwidth}
    \begin{algorithm}[H]
        \centering
        \caption{Model Selection}
        \label{algo:model-selection}
        \begin{algorithmic}[1]
            \State Input: train$_k$, $\Theta$
            \State Split train$_k$ into \textit{train} and \textit{valid}
            \State p$_\theta$ = $\emptyset$
            \For{\textbf{each} $\theta \in \Theta $}
                \State model $\gets$ Train(train$_k$, $\theta$)
                \State p$_\theta$ $\gets$ p$_\theta\; \cup$ Eval(model, \textit{valid})
            \EndFor
            \State best$_\theta$ $\gets$ argmax$_\theta$ p$_\theta$
            \State \textbf{return} best$_\theta$
        \end{algorithmic}
    \end{algorithm}
    \end{minipage}
\caption{Pseudo-code for model assessment (left) and model selection (right). \quotes{Select} refers to the model selection procedure, whereas \quotes{Train} and \quotes{Eval} represent training and prediction phases, respectively.}
\label{tab:exp-setup-iclr}
\end{table}

Model selection relies on grid search, and the hyper-parameters' configurations for each model are shown in Table \ref{tab:hyper-parameters-iclr}. We always included the hyper-parameters mentioned in the original papers, but we also di our best to ensure a fair comparison in terms of number of parameters and configurations to try.

Computationally speaking, we had to run a very large number of experiments, which took months to complete. For each model, we tried a number of configurations ranging from 32 to 72, due to the varying number of hyper-parameters to select. The total effort amounted to more than 47000 training runs, which clearly required an extensive use of parallelism. We leveraged both multi-CPU and multi-GPU machines to complete these tasks in a reasonable amount of time. Nonetheless, training models such as ECC would have required more than 72 hours for a single training run on some social datasets. Allowing these models to complete their training would have dramatically slowed down the process; for these reason, due to the large amount of experiments to run and the limited amount of computational resources, we set a time limit of 72 hours to complete a single training run.

\clearpage
\begin{sidewaystable}[ht]
\scriptsize
\begin{tabular}{l|cccccccccccccc}
\toprule
                                                            & Layers                                                       & \begin{tabular}[c]{@{}c@{}}Convs\\ per \\ layer\end{tabular} & Batch size                                                   & \begin{tabular}[c]{@{}c@{}}Learning \\ rate\end{tabular}   & \begin{tabular}[c]{@{}c@{}}Hidden \\ units\end{tabular}                                               & Epochs & L2                                                         & Dropout                                             & Patience                                                       & Optimizer & Scheduler                                                                  & \begin{tabular}[c]{@{}c@{}}Dense\\ dim\end{tabular} & \begin{tabular}[c]{@{}c@{}}Embed.\\ dim\end{tabular} & \begin{tabular}[c]{@{}c@{}}Neighbors \\ Aggregation\end{tabular}                                              \\ \midrule
\begin{tabular}[c]{@{}l@{}}Baseline\\ chemical\end{tabular} & -                                                            & -                                                            & \begin{tabular}[c]{@{}c@{}}32\\ 128\end{tabular}             & \begin{tabular}[c]{@{}c@{}}1e-1\\ 1e-3\\ 1e-6\end{tabular} & \begin{tabular}[c]{@{}c@{}}32\\ 128\\ 256\end{tabular}                                                & 5000   & \begin{tabular}[c]{@{}c@{}}1e-2\\ 1e-3\\ 1e-4\end{tabular} & -                                                   & \begin{tabular}[c]{@{}c@{}}500, loss\\ 500, acc\end{tabular}   & Adam      & -                                                                          & -                                                   & -                                                    & sum                                                      \\ \midrule
\begin{tabular}[c]{@{}l@{}}Baseline IMDB\end{tabular}   & -                                                            & -                                                            & \begin{tabular}[c]{@{}c@{}}32\\ 128\end{tabular}             & \begin{tabular}[c]{@{}c@{}}1e-1\\ 1e-3\\ 1e-6\end{tabular} & \begin{tabular}[c]{@{}c@{}}32\\ 128\\ 256\end{tabular}                                                & 3000   & \begin{tabular}[c]{@{}c@{}}1e-2\\ 1e-3\\ 1e-4\end{tabular} & -                                                   &

\begin{tabular}[c]{@{}c@{}}500, loss\\ 500, acc\end{tabular}   & Adam      & -                                                                          & -                                                   & -                                                    & sum                                                      \\ \midrule
\begin{tabular}[c]{@{}l@{}}Base. COLLAB \\ and REDDIT\end{tabular}   & -                                                            & -                                                            & \begin{tabular}[c]{@{}c@{}}32\\ 128\end{tabular}             & \begin{tabular}[c]{@{}c@{}}1e-1\\ 1e-3\end{tabular} & \begin{tabular}[c]{@{}c@{}}32\\ 128\end{tabular}                                                & 3000   & \begin{tabular}[c]{@{}c@{}}1e-2\\ 1e-3\\ 1e-4\end{tabular} & -                                                   &

\begin{tabular}[c]{@{}c@{}}500, loss\\ 500, acc\end{tabular}   & Adam      & -                                                                          & -                                                   & -                                                    & sum                                                      \\ \midrule
\begin{tabular}[c]{@{}l@{}}Baseline\\ ENZYMES\end{tabular}  & -                                                            & -                                                            & 32                                                           & \begin{tabular}[c]{@{}c@{}}1e-1\\ 1e-3\\ 1e-6\end{tabular} & \begin{tabular}[c]{@{}c@{}}32\\ 64\\ 128\\ 256\end{tabular}                                           & 5000   & \begin{tabular}[c]{@{}c@{}}1e-2\\ 1e-3\\ 1e-4\end{tabular} & -                                                   & \begin{tabular}[c]{@{}c@{}}1000, loss\\ 1000, acc\end{tabular} & Adam      & -                                                                          & -                                                   & -                                                    & sum                                                      \\ \midrule
DGCNN                                                       & \begin{tabular}[c]{@{}c@{}}2\\ 3\\ 4\end{tabular}            & 1                                                            & \begin{tabular}[c]{@{}c@{}}50 (cpu)\\ 16 (gpu)\end{tabular}  & \begin{tabular}[c]{@{}c@{}}1e-4\\ 1e-5\end{tabular}        & \begin{tabular}[c]{@{}c@{}}32\\ 64\end{tabular}                                                       & 1000   & -                                                          & 0.5                                                 & \begin{tabular}[c]{@{}c@{}}500, loss\\ 500, acc\end{tabular}   & Adam      & -                                                                          & 128                                                 & -                                                    & mean                                                    \\ \midrule
DiffPool                                                    & \begin{tabular}[c]{@{}c@{}}1\\ 2\end{tabular}                & 3                                                            & \begin{tabular}[c]{@{}c@{}}20 (cpu)\\ 8 (gpu)\end{tabular}   & \begin{tabular}[c]{@{}c@{}}1e-3\\ 1e-4\\ 1e-5\end{tabular} & \begin{tabular}[c]{@{}c@{}}32\\ 64\end{tabular}                                                       & 3000   & -                                                          & -                                                   & \begin{tabular}[c]{@{}c@{}}500, loss\\ 500, acc\end{tabular}   & Adam      & -                                                                          & 50                                                  & \begin{tabular}[c]{@{}c@{}}64\\ 128\end{tabular}     & mean                                                      \\ \midrule
ECC                                                         & \begin{tabular}[c]{@{}c@{}}1\\ 2\end{tabular}                & 3                                                            & \begin{tabular}[c]{@{}c@{}}32 (cpu)\\ 8 (gpu)\end{tabular}   & \begin{tabular}[c]{@{}c@{}}1e-1\\ 1e-2\end{tabular}        & \begin{tabular}[c]{@{}c@{}}32\\ 64\end{tabular}                                                       & 1000   & -                                                          & \begin{tabular}[c]{@{}c@{}}0.05\\ 0.25\end{tabular} & \begin{tabular}[c]{@{}c@{}}500, loss\\ 500, acc\end{tabular}   & SGD       & ECC-LR                                                                     & -                                                   & -                                                    & sum                                                      \\ \midrule
GIN                                                         & \begin{tabular}[c]{@{}c@{}}see\\ hidden\\ units\end{tabular} & 1                                                            & \begin{tabular}[c]{@{}c@{}}32\\ 128\end{tabular}             & 1e-2                                                       & \begin{tabular}[c]{@{}c@{}}32 (5 layers)\\ 64 (5 layers)\\ 64 (2 layers)\\ 32 (3 layers)\end{tabular} & 1000   & -                                                          & \begin{tabular}[c]{@{}c@{}}0\\ 0.5\end{tabular}     & \begin{tabular}[c]{@{}c@{}}500, loss\\ 500, acc\end{tabular}   & Adam      & \begin{tabular}[c]{@{}c@{}}Step-LR\\ (step: 50,\\ gamma: 0.5)\end{tabular} & -                                                   & -                                                    & sum                                                      \\ \midrule
GraphSAGE                                                   & \begin{tabular}[c]{@{}c@{}}3\\ 5\end{tabular}                & 1                                                            & \begin{tabular}[c]{@{}c@{}}32 (cpu)\\ 16 (gpu)\end{tabular} & \begin{tabular}[c]{@{}c@{}}1e-2\\ 1e-3\\ 1e-4\end{tabular} & \begin{tabular}[c]{@{}c@{}}32\\ 64\end{tabular}                                                       & 1000   & -                                                          & -                                                   & \begin{tabular}[c]{@{}c@{}}500, loss\\ 500, acc\end{tabular}   & Adam      & -                                                                          & -                                                   & -                                                    & \begin{tabular}[c]{@{}c@{}}mean\\ max\\ sum\end{tabular} \\
\bottomrule
\end{tabular}
\caption{Hyper-parameters tried during each grid-search model selection.}
\label{tab:hyper-parameters-iclr}
\end{sidewaystable}
\clearpage

\subsection{Results}
\label{subsec:iclr-results}
We present the results on chemical and social benchmarks in Tables \ref{tab:chemical-iclr} and \ref{tab:social-iclr}, respectively. Some observations can be made: first of all, none of the DGNs seem to improve over the performance of the structure-agnostic baseline on three out of four chemical datasets. On the other hand, the baseline cannot reach the same performance of DGNs on the NCI1 dataset. To confirm that this is not due to the under-parametrization of the baseline, we trained a configuration with 10000 hidden units and no regularization. The training accuracy reached a modest 67\%, whereas a DGN like GIN can easily overfit the training set. This provides unambiguous evidence that structural information is actually relevant for the task. In social datasets, the addition of vertex degrees makes the baseline very competitive w.r.t. most models, but the GIN model has the best accuracy scores in almost all social tasks.
\begin{table}[ht]
\centering
\small
\begin{tabular}{l c c c c c}
\toprule
     & \textbf{D\&D} & \textbf{NCI1} & \textbf{PROTEINS} & \textbf{ENZYMES}\\
\midrule
 Baseline & $\mathbf{78.4}\pm 4.5 $ &  $69.8 \pm 2.2 $ &  $\mathbf{75.8} \pm 3.7 $ &  $\mathbf{65.2}\pm 6.4 $ \\
 DGCNN & $76.6 \pm 4.3 $ &  $76.4 \pm 1.7 $ &  $72.9 \pm 3.5 $ &  $38.9 \pm 5.7 $   \\
 DiffPool & $75.0 \pm 3.5 $ &  $76.9 \pm 1.9 $ &  $73.7 \pm 3.5 $ &  $59.5 \pm 5.6 $   \\
 ECC & $72.6 \pm 4.1 $ &  $76.2 \pm 1.4 $ &  $72.3 \pm 3.4 $ &  $29.5 \pm 8.2 $   \\
 GIN & $75.3 \pm 2.9 $ &  $\mathbf{80.0} \pm 1.4 $ &  $73.3 \pm 4.0 $ &  $59.6 \pm 4.5 $   \\
 GraphSAGE & $72.9 \pm 2.0 $ &  $76.0 \pm 1.8 $ &  $73.0 \pm 4.5 $ &  $58.2 \pm 6.0 $   \\
\bottomrule
\end{tabular}
\caption{Results on chemical datasets with mean accuracy and standard deviation are reported. Best average performances are highlighted in bold.}
\label{tab:chemical-iclr}
\end{table}
\begin{table}[ht]
\renewcommand{\arraystretch}{1.1}
\centering
\small
\begin{tabular}{llccccc}
\toprule
     & & \textbf{IMDB-B} & \textbf{IMDB-M} & \textbf{REDDIT-B} & \textbf{REDDIT-5K} & \textbf{COLLAB}\\
    \midrule
\multirow{6}{*}{\rotatebox[origin=c]{90}{\textsc{No Features}}}
& Baseline & $50.7 \pm 2.4 $ &  $36.1 \pm 3.0 $ &  $72.1 \pm 7.8 $ &  $35.1 \pm 1.4 $ &  $55.0 \pm 1.9 $   \\
& DGCNN & $53.3 \pm 5.0 $ &  $38.6 \pm 2.2 $ &  $77.1 \pm 2.9 $ &  $35.7 \pm 1.8 $ &  $57.4 \pm 1.9 $   \\
& DiffPool & $68.3 \pm 6.1 $ &  $45.1 \pm 3.2 $ &  $76.6 \pm 2.4 $ &  $34.6 \pm 2.0 $ &  $67.7 \pm 1.9 $   \\
& ECC & $67.8 \pm 4.8 $ &  $44.8 \pm 3.1 $ &   OOR &   OOR &   OOR   \\
& GIN & $66.8 \pm 3.9 $ &  $42.2 \pm 4.6 $ &  $\mathbf{87.0} \pm 4.4 $ &  $\mathbf{53.8} \pm 5.9 $ &  $\mathbf{75.9} \pm 1.9 $   \\
& GraphSAGE & $\mathbf{69.9}\pm 4.6 $ &  $\mathbf{47.2}\pm 3.6 $ &  $86.1 \pm 2.0 $ &  $49.9 \pm 1.7 $ &  $71.6 \pm 1.5 $   \\

\midrule
\multirow{6}{*}{\rotatebox[origin=c]{90}{\textsc{With Degree}}}
& Baseline & $70.8 \pm 5.0 $ &  $\mathbf{49.1} \pm 3.5 $ &  $82.2 \pm 3.0 $ &  $52.2 \pm 1.5 $ &  $70.2 \pm 1.5 $   \\
& DGCNN & $69.2 \pm 3.0 $ &  $45.6 \pm 3.4 $ &  $87.8 \pm 2.5 $ &  $49.2 \pm 1.2 $ &  $71.2 \pm 1.9 $   \\
& DiffPool & $68.4 \pm 3.3 $ &  $45.6 \pm 3.4 $ &  $89.1 \pm 1.6 $ &  $53.8 \pm 1.4 $ &  $68.9 \pm 2.0 $   \\
& ECC & $67.7 \pm 2.8 $ &  $43.5 \pm 3.1 $ &   OOR &   OOR &   OOR   \\
& GIN & $\mathbf{71.2} \pm 3.9 $ &  $48.5 \pm 3.3 $ &  $\mathbf{89.9} \pm 1.9 $ &  $\mathbf{56.1} \pm 1.7 $ &  $\mathbf{75.6} \pm 2.3 $   \\
& GraphSAGE & $68.8 \pm 4.5 $ &  $47.6 \pm 3.5 $ &  $84.3 \pm 1.9 $ &  $50.0 \pm 1.3 $ &  $73.9 \pm 1.7 $   \\
\bottomrule
\end{tabular}
\caption{Results on social datasets with mean accuracy and standard deviation are reported. Best average performances are highlighted in bold. OOR means Out of Resources, either time ($>$ 72 hours for a single training) or GPU memory. }
\label{tab:social-iclr}
\end{table}

From these results, it is clear how structure-agnostic baselines represent an essential tool to understand the impact of using DGNs. But we can extract further insights too: since structural features are known to correlate with molecular properties \cite{vanrossum_relation_1963}, it is possible that the actual DGNs are still not able to extract what is needed to solve the above chemical tasks. Also, the relatively high standard deviations should suggest caution when arguing that a model performs better than another because of small (averaged) performance gains. It is highly likely, in fact, that such performance fluctuations are caused by random initializations, rather than being actual empirical progress.

It is also interesting to see how the addition of a simple feature like the vertex degree is able to provide significant performance gains on the social datasets. Indeed, the baseline provides from 10\% to 20\% better accuracy with this kind of information, and it even achieves state of the art results on IMDB-BINARY. As regards DGNs, instead, the effect of the degree seems to be less relevant, which is reasonable since the first layer can (in principle) compute the degree by summing neighboring features. One notable exception is DGCNN, which explicitly needs the degree as a vertex feature to improve the performances. Last but not least, the addition of this feature produces completely different rankings, much alike what happened in \cite{shchur_pitfalls_2018}. This demonstrates how important it is to compare different methods while using the same set of features.

Since the degree of a vertex can be computed with a simple sum-based neighborhood aggregation, we compare the median \quotes{best} number of layers chosen across the 10 different folds in the two social settings. Results are reported in Table \ref{tab:num-layers-iclr}. There exists a general trend, with the exception of GraphSAGE, in which the best number of layers is reduced by approximately 1 when we add the degree feature. Therefore, our intuitive reasoning about the inductive bias of DGNs architectures seems supported by evidence.

\begin{table}[ht]
\renewcommand\arraystretch{1.2}
\small
\begin{center}
\begin{tabular}{lcccccccccc}
\cmidrule{2-11}
          & \multicolumn{2}{c}{\textbf{IMDB-B}} & \multicolumn{2}{c}{\textbf{IMDB-M}} & \multicolumn{2}{c}{\textbf{REDDIT-B}} & \multicolumn{2}{c}{\textbf{REDDIT-M}} & \multicolumn{2}{c}{\textbf{COLLAB}} \\
          \cmidrule(lr){2-3}\cmidrule(lr){4-5}\cmidrule(lr){6-7}\cmidrule(lr){8-9}\cmidrule(lr){10-11}
          & \textbf{1}           & \textbf{DEG} & \textbf{1}           & \textbf{DEG} & \textbf{1}           & \textbf{DEG} & \textbf{1}           & \textbf{DEG} & \textbf{1}           & \textbf{DEG}          \\
          \midrule
\textbf{DGCNN}     & 3           & 3            & 3.5          & 3           & 4          & 3            & 3          & 2            & 4           & 2            \\
\textbf{DiffPool}  & 1           & 2            & 2            & 1           & 2          & 2            & 2          & 1            & 2           & 1.5          \\
\textbf{ECC}       & 1           & 2            & 1            & 1           & -          & -            & -          & -            & -           & -            \\
\textbf{GIN}       & 3           & 2            & 4            & 2           & 4          & 4            & 4          & 3            & 4           & 4            \\
\textbf{GraphSAGE} & 4           & 3            & 5            & 4           & 3          & 4            & 3          & 5            & 3           & 5\\
\bottomrule
\end{tabular}
\end{center}
\caption{We report the median number of selected layers per model, depending on whether vertex degrees are used as input features or not. A \quotes{1} indicates that an uninformative feature is used as the vertex label.}
\label{tab:num-layers-iclr}
\end{table}

\begin{figure}
    \centering
    \includegraphics[width=\textwidth]{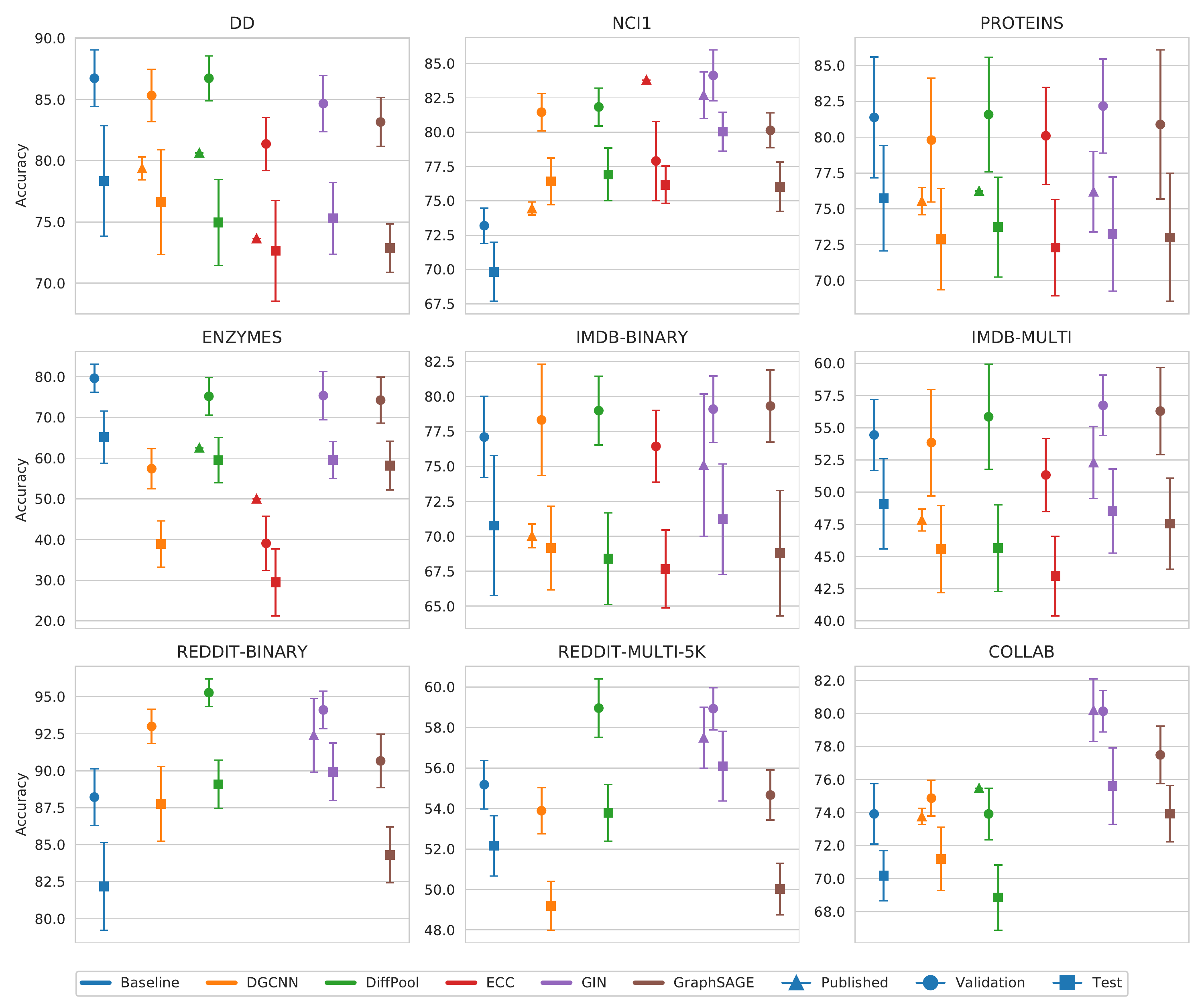}
   \caption{Chemical and social (with degree) benchmark results are shown together with published results (when available). For each of them, we report validation and test accuracy of the evaluated models, together with published results (when available).}
   \label{fig:vl-vs-test-iclr}
\end{figure}

Finally, to show that our estimates are actually much lower than what has been reported in the literature, we visually compare our averaged values with those of the original papers. In addition, we plot the best validation scores averaged across the 10 different model selections, so that we can clearly see how far from the (empirical) truth we can get when reporting validation scores. Figure \ref{fig:vl-vs-test-iclr} confirms that the gap between the validation and test estimates is usually consistent, with validation scores overestimating the true generalization performances of the model.

To conclude this section, we briefly mention that subsequent effort has been made to put together larger datasets \cite{hu_open_2020,dwivedi_benchmarking_2020} and standardize the evaluation process\footnote{\url{https://github.com/diningphil/PyDGN}.}. Open Graph Benchmark \cite{hu_open_2020} is a collection of graph-related tasks, each with its own evaluation process with fixed performance metrics. In another work \cite{dwivedi_benchmarking_2020}, larger benchmarks related to chemistry, the travel-salesman problem, and image classification are proposed, together with an assessment of some models taken from the literature. It should be noted that in \cite{dwivedi_benchmarking_2020} the numbers are approximated estimates of the generalization performances of each model, as a proper model selection has not been carried out due to time constraints. Therefore, it could be unclear whether subsequent improvements w.r.t. those numbers will be caused by the model selection itself or by the actual improvement of a particular architecture over others.
\clearpage

\section[Application to Molecular Biosciences]{Application to Molecular Biosciences \cite{errica_deep_2021}}
\label{sec:application-molecular}
Now that we have discussed the building blocks of deep learning for graphs as well as our attempt to tackle some of its scholarship issues, we shall provide an example of a practical application from the field of molecular biosciences \cite{errica_deep_2021}, more specifically related to molecular dynamics.

Molecular dynamics simulations \cite{md_general_method, md_sim_biomol} are a very useful tool when it comes to investigate properties of matter. Classical all-atom simulations have allowed researchers to ultimately understand a large variety of physical systems, from metals and fluids to biological entities like proteins. As these systems grow larger, the computational costs and the intuitive understanding of the systems' behavior become increasingly challenging. In the soft and biological matter field, \textbf{coarse-graining} methods provide ways to extract relevant properties of a macro-molecular system \cite{marrink2007martini, Takada2012, Potestio2014, Saunders2013}. To do so, the system is first \quotes{simplified} into a higher-level representation where the constituent units are called \CG{} \textbf{sites}.

\begin{figure}[b]
    \centering
    \includegraphics[width=\textwidth]{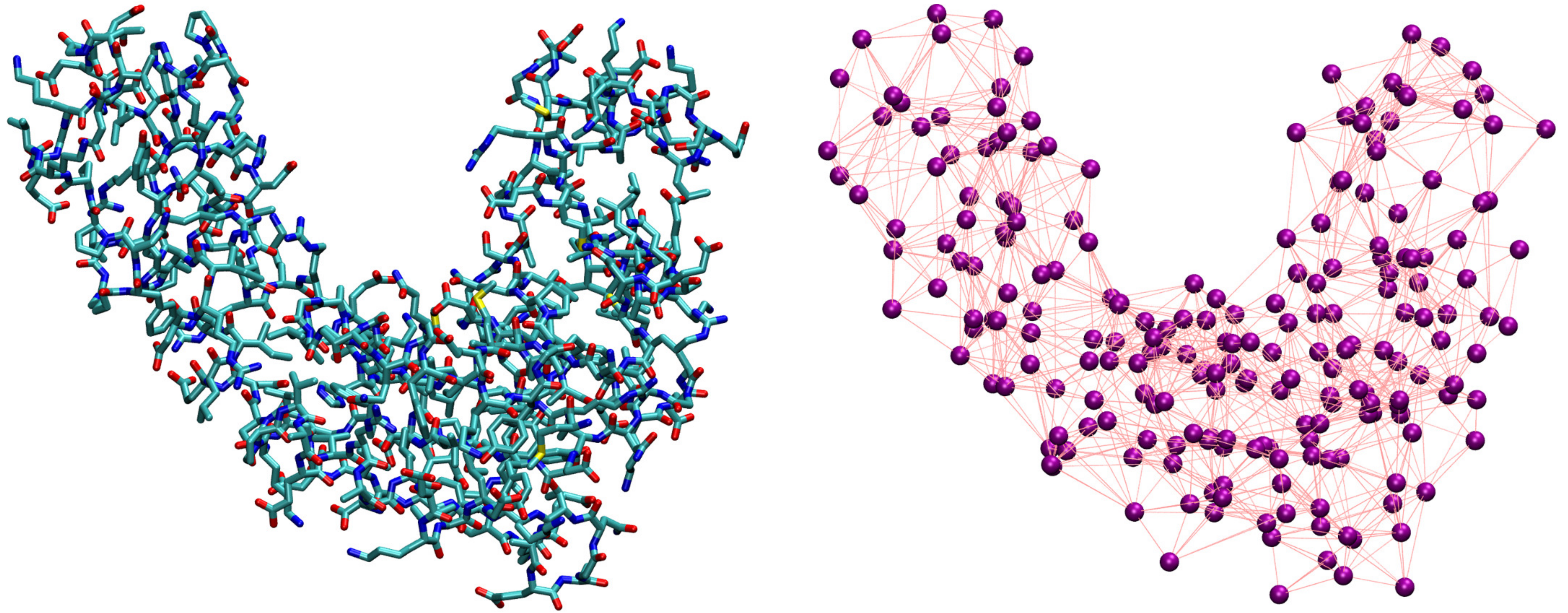}
   \caption{Comparing an all-atom system (a protein, left) with one of its possible \CG{} representations (right). Purple vertices are \CG{} sites.}
   \label{fig:cg-sites}
\end{figure}

Defining a \CG{} representation requires two things: first, a \textbf{mapping} from the units of the original system to the \CG{} sites; second, the set of effective \textbf{interactions} between the sites, so that we can reproduce a posteriori the emergent properties of the original system through this simplified representation. Figure \ref{fig:cg-sites} depicts one such example. While there has been a substantial research effort in defining \CG{} potentials \cite{noid2008multiscale,noid_mapping, Shell2008}, the study of the mapping itself has been less investigated. Sites are often selected on the basis of chemical or physical criteria that do not take into account the local and global environment of each constituent in the original system \cite{kmiecik2016coarse}.

Nevertheless, this approach has an evident limitation: any \CG{} process implies some degree of information loss, so it would be appropriate to \textbf{automatically} find the mapping that minimizes the loss of information about the system's overall behavior. There have been attempts to solve via graph-theoretical analyses \cite{depabloJCTC2019}, geometric criteria \cite{bereau2015automated} and machine learning \cite{murtola2007conformational, wang2019coarse, li2020graph}. The underlying idea of all these work is that the optimal \CG{} representation can be found in a subset of the original features. In addition, there are statistical mechanics-based strategies that address the issue by means of minimization of the so-called \textbf{mapping entropy} $S_{map}$ \cite{giulini2020information}, a measure of the dissimilarity between the probability density of the original configuration and that of the lower-resolution description. \cite{foley2015impact,  rudzinski_2011, Shell2008, Shell2012}.

The main shortcomings of mapping entropy minimization are the costs of computing the $S_{map}$ of a single \CG{} representation as well the combinatorial size of the search space. Therefore, in this section, we propose to train a Deep Graph Network that predicts the mapping entropy associated with a specific \CG{} representation of a given protein. If we managed to achieve good performances, we could incorporate the much more efficient DGN into the Wang-Landau enhanced sampling algorithm \cite{wang2001determining,wang2001efficient,shell2002generalization,barash2017control} so as to carry out a quasi-exhaustive exploration of a biomolecule's mapping space.

\subsection{Datasets}
\begin{figure}[b]
    \centering
    \hspace{-2cm}
    \includegraphics[width=\textwidth]{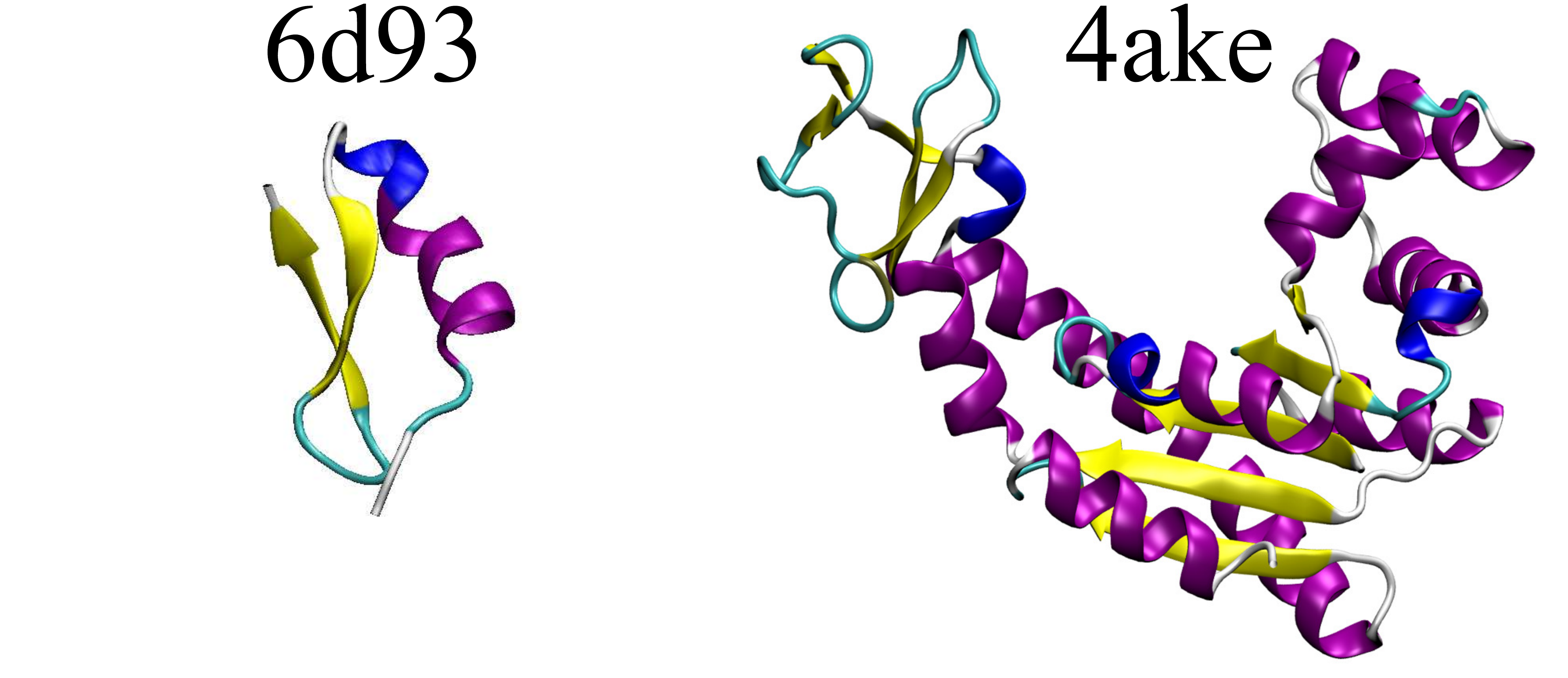}
    \caption{The \textit{tamapin}, \aka \textit{6d93},  and the open conformation of \textit{adenylate kinase}, \aka \textit{4ake}. Though smaller, \textit{6d93} possesses all the elements of proteins' secondary structures. On the other hand, \textit{4ake} is larger and has has a much wider structural variability.}
    \label{fig:proteins}
\end{figure}

\begin{figure}[ht]
    \centering
    \hspace{-2cm}
    \includegraphics[width=0.8\textwidth]{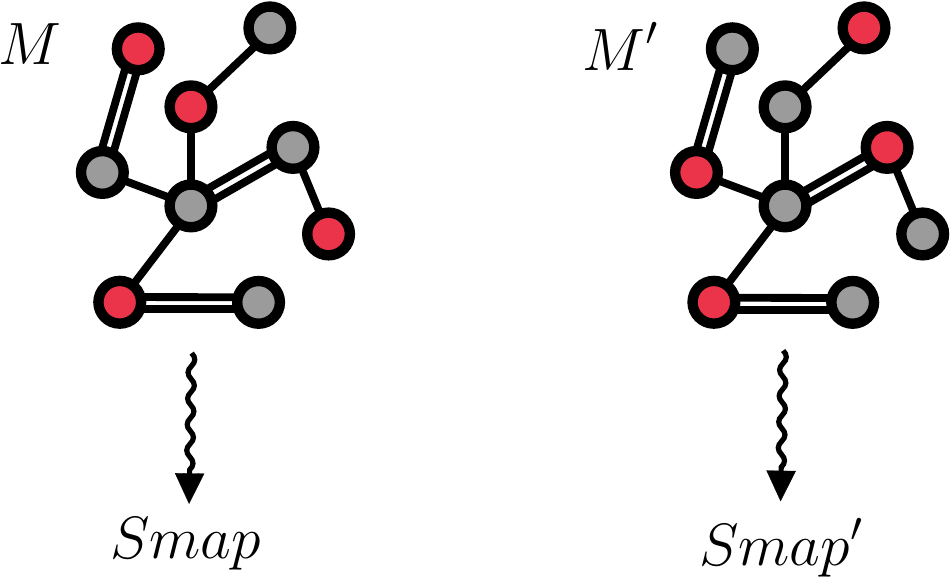}
    \caption{Each protein is converted to a graph, where vertices hold atomic features and the \quotes{site} information (in red). For the same protein and different mappings, we seek to predict the corresponding mapping entropies.}
    \label{fig:proteintograph}
\end{figure}

The evaluation focuses on two proteins called \textit{6d93} and \textit{4ake}, extracted from \cite{giulini2020information}. The former is a mutant of \textit{tamapin}, a toxin of the Indian red scorpion, whereas the latter is the open conformation of the \textit{adenylate kinase}, an enzyme inside the cell. A schematic representation of both proteins is shown in Figure \ref{fig:proteins}. The task is a \textit{regression} problem in which, given a protein and a specific choice for the mapping, we need to predict the associated mapping entropy.
To build the dataset, we first represent each protein as a graph, where vertices encode heavy atoms and edges connect pairs of atoms whose atomic distance is closer than 1\textit{nm} in the reference structure. We incorporate a number of binary properties into each vertex's features, which are described in Table \ref{tab:features-frontiers}, whereas edge features consist of a single continuous value encoding the inverse atomic distance. A schematic representation of a protein as a graph, with different mappings and therefore $S_{map}$ values, is shown in Figure \ref{fig:proteintograph}

\begin{table}[ht]
\centering
\begin{tabular}{@{}ll@{}}
\toprule
Feature name \ \ \ \ \ & Description \\ \hline
\addlinespace[0.1cm] C & Carbon atom \\
N & Nitrogen atom \\
O & Oxygen atom \\
S & Sulphur atom \\
HPhob & Part of a hydrophobic residue\\
Amph & Part of a amphipathic residue\\
Pol & Part of a polar residue \\
Ch & Part of a charged residue \\
Bkb & Part of the protein backbone \\
Site & \textbf{Atom selected as a CG site}\\
\hline
\end{tabular}
\caption{A list of the binary features used to describe the properties of each atom in the protein representation.}
\label{tab:features-frontiers}
\end{table}

\begin{table}[ht]
\centering
\begin{tabular}{@{}llll@{}}
\toprule
Protein &  CPU time & Walltime & Single measure \\ \hline
\addlinespace[0.1cm] \textit{6d93} & $40.7$ days  & $2.55$ days & $\simeq 2.1 $ mins  \\
\textit{4ake} & $153.9$ days & $3.20$ days   & $\simeq 8.0 $ mins \\  \hline
\end{tabular}
\caption{Computational costs of all-atom simulations and mapping entropy calculations for the two investigated proteins. \textit{CPU time} (respectively \textit{Walltime}) represents the time (user time) necessary to simulate 200ns. \textit{Single measure} is the amount of time that is required to compute, on a single core, the $S_{map}$ of a given mapping.}
\label{tab:all-atom-simulation-costs}
\end{table}

The samples in the dataset are constructed by taking the \textbf{same} graph representation of the protein and changing the binary attribute \quotes{Site} depending on the \CG{} configuration we want to represent. If an atom is selected as a site, then the attribute is set to 1 and 0 otherwise. Note that we retain the atoms that are not selected by the \CG{} configuration: the underlying idea is to make the DGN learn the relation between the sites' position in the protein and the mapping entropy value. To find a good estimate for the target value, we carried out expensive all-atom simulations on these proteins. We ended up with 4968 and 1968 labeled samples, and we summarize other dataset statistics in Table \ref{tab:protein-statistics}. The distribution of the target values for both datasets is such that there is negligible overlap between the random and optimized\footnote{Using a simulated annealing approach \cite{giulini2020information}.} mappings, meaning that the best $S_{map}$ values cannot be reached by a mere random exploration of the mapping space. Moreover, the mapping entropy is proportional to the system's size, as the lower bound in mapping entropy of \textit{4ake} $(\approx 90)$ is almost one order of magnitude higher than that of \textit{6d93} $(\approx 10)$. Computationally speaking, Table \ref{tab:all-atom-simulation-costs} reports that the mapping entropy calculations of a single \CG{} configuration can take up to 8 minutes for the larger protein \textit{4ake}.

\begin{table}[ht]
    \centering
    \begin{tabular}{lcccc}
    \toprule
         Protein & Vertices & Edges & Avg. Degree & Dataset Size \\ \hline
         \addlinespace[0.1cm] \textit{6d93}   & 230 & 21474 & 93 & 4968\\
         \textit{4ake}  & 1656  & 207618 & 125 & 1968\\ \hline
    \end{tabular}
    \caption{Dataset statistics.}
    \label{tab:protein-statistics}
\end{table}

\subsection{Experimental Setting}
We experiment with a structure-agnostic baseline like the one introduced in Section \ref{sec:scholarship-issues} for social tasks and an edge-aware DGN. The neighborhood aggregation extends that of \cite{xu_how_2019} as follows:
\begin{align*}
     \boldhell{v}{\lnext} &= MLP^{\lcurr} \Big( \big(1 + \epsilon^{\lcurr} \big)*\boldhell{v}{\lcurr}  + \sum_{u \in \N{v}{}} \boldhell{v}{\lcurr} * a_{uv} \Big),
\end{align*}
where $*$ denotes element-wise scalar multiplication, $\epsilon^{\lcurr} \in \R{}$ is an adaptive weight of the model, and $a_\uv$ is the inverse atomic distance between atoms. Practically speaking, we want to penalize the contribution of neighbors that are farther away according to the protein topology. Then, we apply a \textbf{site-aware} readout function that learns to weight the contribution of site ($w_s$) and non-site ($w_n$) atoms belonging to the disjoint sets $\mathcal{V}^s_g \subset \Vset{g}$ and $\mathcal{V}^s_g \subset \Vset{g}$:
\begin{align*}
     \hat{S}_{map} &= \mathbf{w}_{out}^T \Big(\sum_{u \in \mathcal{V}_g^s} \big([\boldhell{u}{1},\dots,\boldhell{u}{L}]*w_s\big) + \sum_{u \in \mathcal{V}_g^n} \big([\boldhell{u}{1},\dots,\boldhell{u}{L}]*w_n\big) \Big),
\end{align*}
where $L$ is the chosen number of layers, $\mathbf{w}_{out} \in \mathbb{R}^{K*L}$ is a vector of parameters to be learned, and square brackets denote concatenation of the different vertex states computed at different layers.

To assess the performance of each model, we first split the dataset into training, validation and test realisations, following an 80\%/10\%/10\% hold-out strategy. During model selection. we applied early stopping to select the training epoch with the best validation score, and the chosen model was evaluated on the unseen test set. The evaluation metric for our regression problem is the coefficient of determination (or R$^2$-score); this score ranges from $-\infty$ (worst predictor) to $1$ (best predictor).

For the purpose of this application, and due to the computational costs necessary to train a DGN on these datasets, we opted for selecting the hyper-parameters via a manual experimental screening on the validation set performances. Eventually, we chose a DGN depth of $L=5$,  and we implemented each $MLP$ as a one-layer feed-forward network with $K=64$ hidden units followed by an element-wise rectifier linear unit (ReLU) activation function \cite{glorot_deep_2011}. The loss function was the Mean Absolute Error (MAE). The optimization algorithm was Adam \cite{kingma_adam_2015} with a learning rate of $0.001$ and no regularization. We trained for a maximum of $10000$ epochs with early stopping patience of $1000$ epochs and mini-batch size $8$, accelerating the training using a GPU with 16GB of memory. Instead, we chose $K = 1024$ hidden units for the baseline while keeping the rest unchanged.

The subsequent exploration of the mapping space is carried out with the Wang-Landau sampling scheme. The parameters governing the sampler are the result of previous studies and expert knowledge \cite{giulini2020information}, and they do not influence the training of the DGN. Therefore, in the interest of readability, we refer the reader to \cite{errica_deep_2021} for a thorough description of the whole sampling process as well as the dataset-specific parameters used to explore the mapping space.\footnote{\url{https://github.com/CIML-VARIAMOLS/GRAWL}.}

\subsection{Results}
We start by looking at the prediction performances of the aforementioned models. Table \ref{tab:frontiers-results} reports the R$^2$ score and MAE in training, validation and test. While the baseline provides a surprisingly high score on \textit{6d93}, we also observe that the DGN has much better performances on both datasets. Indeed, it achieves extremely low values of MAE for \textit{6d93}, with an R$^2$ score higher than $0.95$ in all cases. The model performs slightly worse in the case of \textit{4ake}: the result of R$^2 = 0.84$ on the test set is still acceptable, although the gap with the training set (R$^2 = 0.92$) is non-negligible.

\begin{table}[ht]
    \centering
    \small
    \begin{tabular}{lcccccc}
    \toprule
         Model / Protein    & TR MAE & TR R$^2$ & VL MAE & VL R$^2$ & TE MAE & TE R$^2$ \\ \hline
         \addlinespace[0.1cm] Baseline / \textit{6d93}  & 0.55 & 0.86 & 0.63 & 0.83 & 0.65 & 0.82 \\
         DGN / \textit{6d93}  & 0.13 & 0.99 & 0.33 & 0.95 & 0.33 & \textbf{0.96} \\  \hline
         Baseline / \textit{4ake} & 1.78 & 0.70  & 1.75 & 0.65 & 1.86 & 0.69 \\
         DGN / \textit{4ake} & 0.91 & 0.92  & 1.2 & 0.85 & 1.35 & \textbf{0.84} \\ \hline
    \end{tabular}
    \caption{Mapping entropy prediction results on the training (TR), validation (VL) and test (TE) sets for the two analysed proteins. We display both the R$^2$ score and the mean average error (MAE, $kJ/\text{mol}/K$).}
    \label{tab:frontiers-results}
\end{table}

\begin{figure}[ht]
\begin{subfigure}
  \centering
  \includegraphics[width=0.49\textwidth]{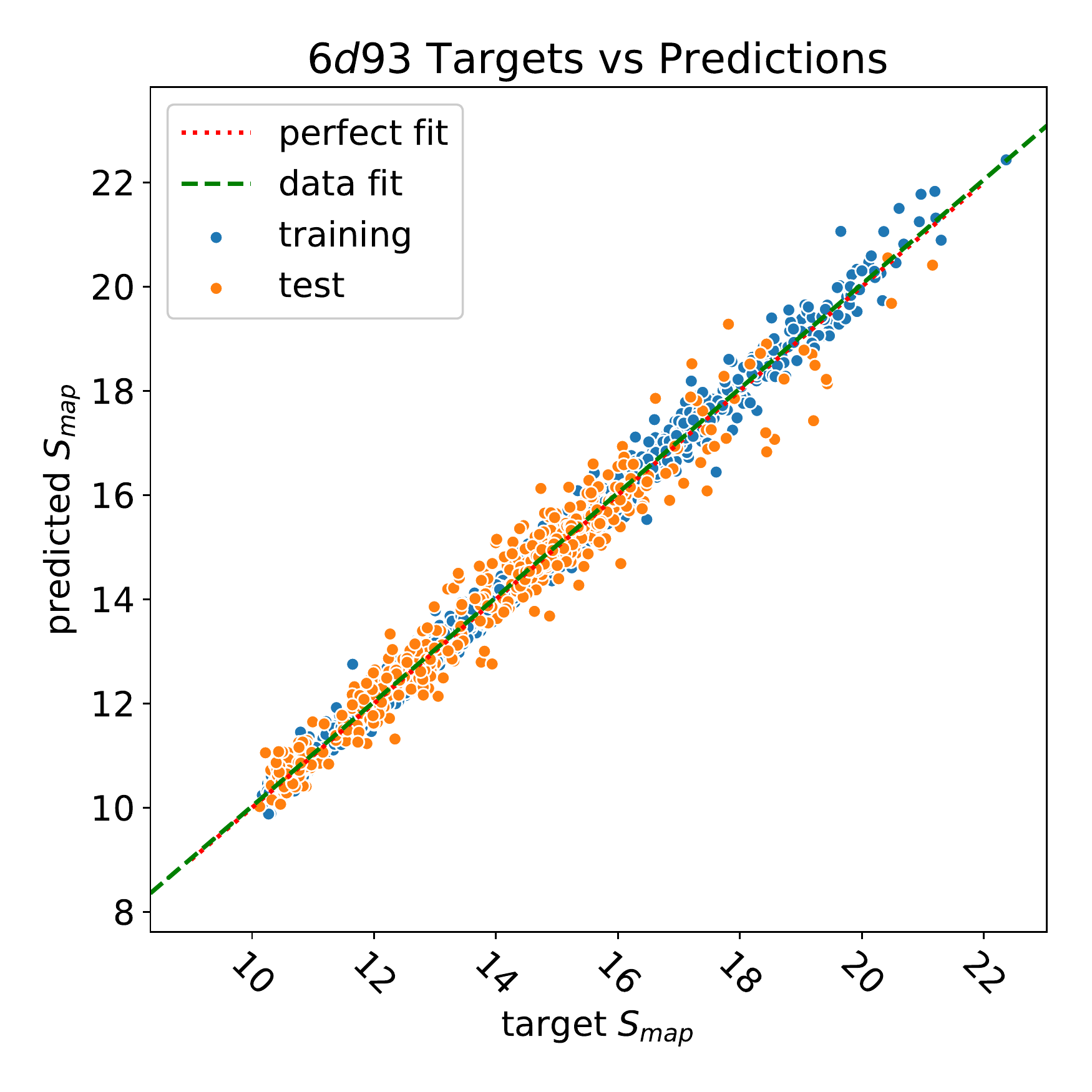}
\end{subfigure}%
\begin{subfigure}
  \centering
  \includegraphics[width=0.49\textwidth]{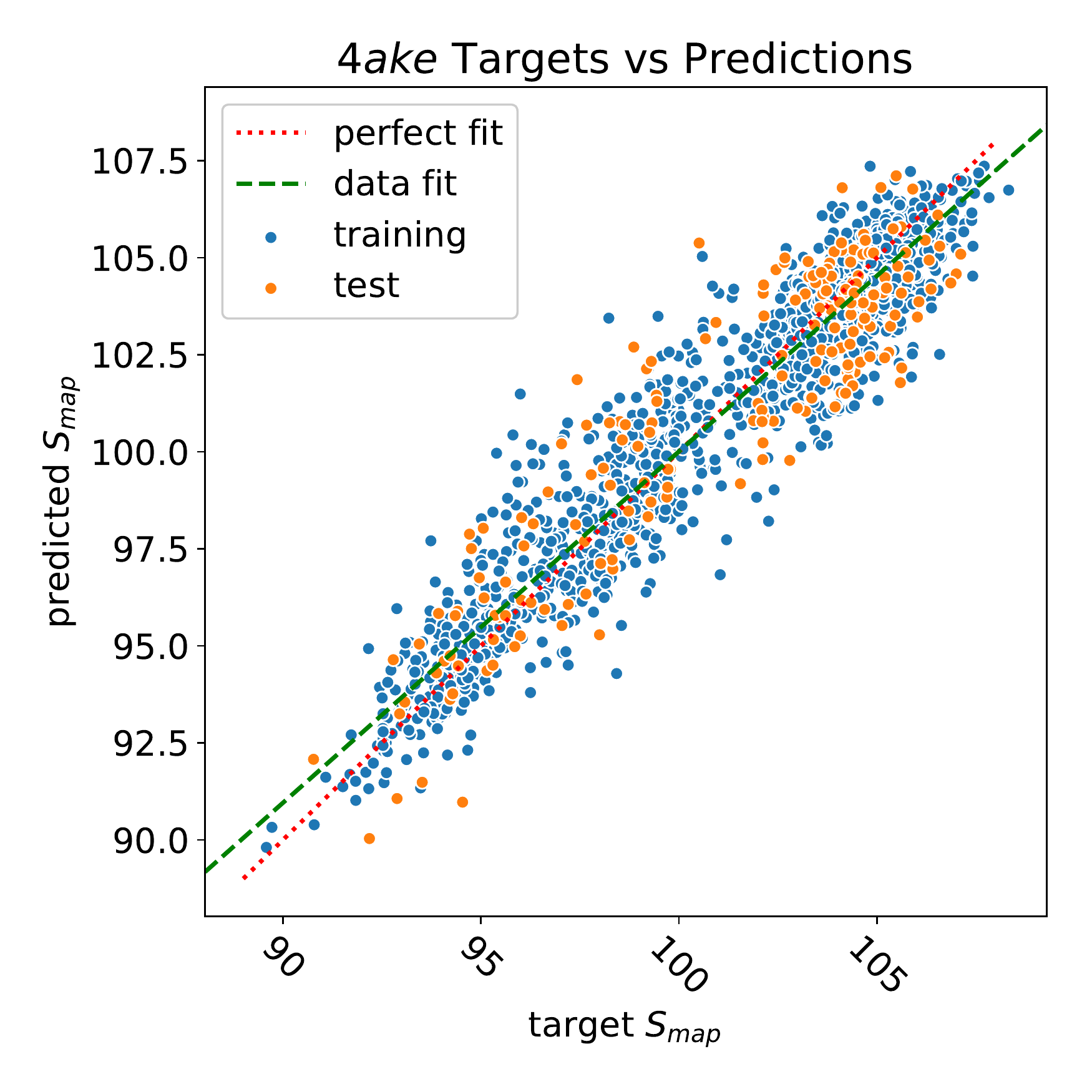}
\end{subfigure}%
\caption{Scatter plot of predictions against the ground truth for both datasets.}
\label{fig:predvstest}
\end{figure}

In Figure \ref{fig:predvstest}, we plot predicted values for training and test samples against their ground truth. Ideally, a perfect result would correspond to the points lying on the diagonal dotted line. As regards \textit{6d93}, we can get pretty close to the true training and test targets. The deviation from the perfect fit becomes wider for \textit{4ake}, but there are no relevant outliers to report, a good sign of the DGN's generalization performances. By closely inspecting the \textit{4ake} scatter plot, we observe that the DGN slightly overestimates the mapping entropy of optimized \CG{} samples, \ie $S_{map}\lesssim 100~kJ/\text{mol}/K$. Likewise, the opposite is true for $S_{map}\gtrsim 100~kJ/\text{mol}/K$, with random \CG{} mappings being slightly underestimated.

The dissimilarity in performance between the two proteins is not surprising, given expert knowledge about their nature. In fact, we already mentioned how \textit{adenylate kinase} is larger and more complex than the \textit{tamapin} mutant. The datasets sizes are necessarily very different due to the heavy computational requirements to label \textit{4ake} samples. It is then natural to expect that training a model with excellent generalization performance on \textit{4ake} would be harder than the other task. We would like to emphasize, however, that a completely adaptive DGN was able to approximate, in both structures, the long and computationally intensive algorithm for estimating the mapping entropy \cite{giulini2020information}. Even more significant is the fact that the model relies on a combination of static structural information and a few simple vertex features. In other words, the DGN operates in the absence of direct knowledge about the complex dynamical behavior of the two systems, in contrast to the onerous molecular dynamics simulations.

To confirm that the DGN provides a computational advantage with respect to the simulations, we report in Table~\ref{tab:frontiers-time-inference} the time required to predict a single $S_{map}$ output and compare it to the ground truth algorithm. To provide a fair comparison between the algorithm of \cite{giulini2020information}, which relies on a CPU machine, we compute prediction times on both CPU and GPU. Overall, we can see that the DGN inference phase is $2-5$ orders of magnitude faster than the original algorithm, depending on the hardware used. Notably, the speedup is associated with a fairly good predictive accuracy of the \ML{} model. To sum up, such drastic speedup of the trained model allows us to carry out a much wider exploration of the $S_{map}$ landscape of both protein systems.
\begin{table}[ht]
\centering
\begin{tabular}{@{}llll@{}}
\toprule
Protein &  Single measure & Inference GPU (CPU) & Improvement GPU (CPU)  \\ \hline
\addlinespace[0.1cm] \textit{6d93} & $\simeq 2.1 $ mins  & $\simeq 0.9 (98.7)$ ms & $\simeq 140000\times (1276 \times)$\\
\textit{4ake} & $\simeq 8.0 $ mins & $\simeq 4.8 (1103.2) $ ms & $\simeq 100000\times (435 \times)$\\  \hline
\end{tabular}
\caption{Time comparison between the original mapping entropy algorithm and the inference phase of the DGN.}
\label{tab:frontiers-time-inference}
\end{table}

If we embed the trained DGN in the Wang-Landau sampler, we can better approximate the distribution of the mapping entropy values for \textit{6d93} and \textit{4ake}. Put differently, we can better estimate how many \CG{} representations (sampled from the mapping space of each protein) exhibit a specific amount of information loss with respect to the all-atom system. To reach convergence of the sampling protocol, we had to probe approximately $4.8\times 10^6$ and $3\times 10^7$ different mappings for \textit{6d93} and \textit{4ake}, respectively. Clearly, such an extensive sampling was made feasible by the speedup attained by the proposed DGN.

\begin{figure}[ht]
\centering
\includegraphics[width=\textwidth]{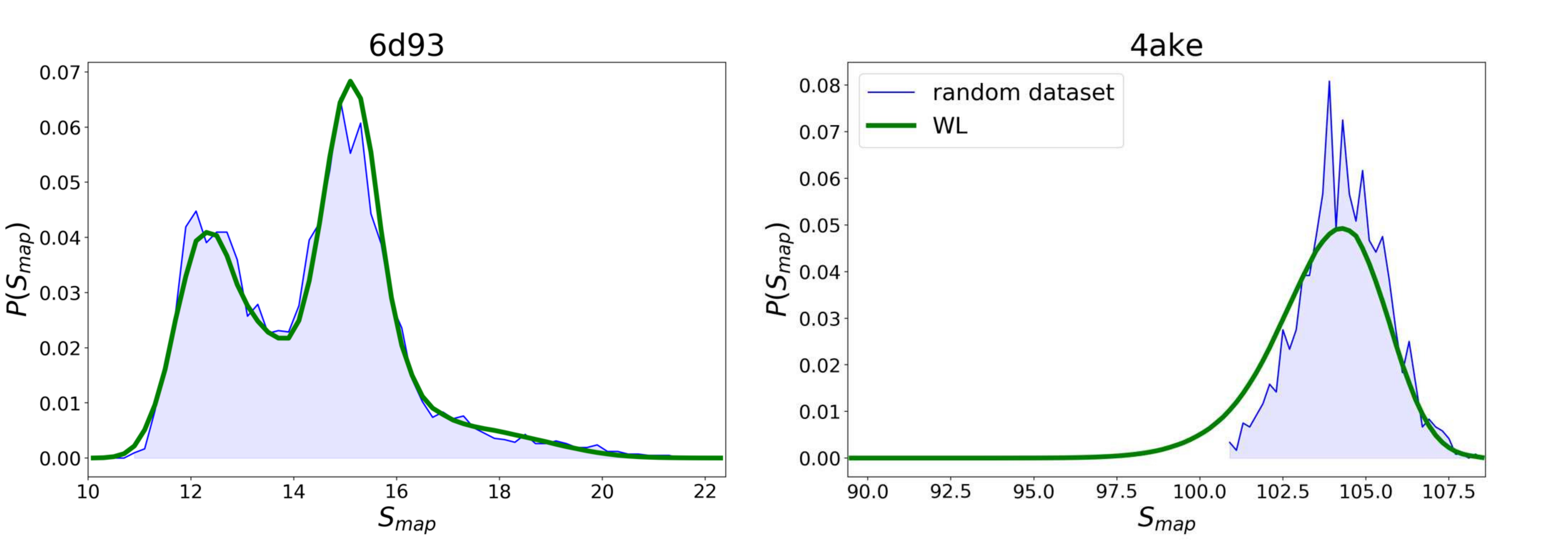}
\caption{Comparing probability densities $P(S_{map})$ for the two proteins. The smooth distribution produced by the DGN (green lines) is similar to that generated by a random sampling of mappings (blue areas). $S_{map}$ values are in $kJ/\text{mol}/K$. WL here stands for Wang-Landau.}
\label{fig:wl_dofs}
\end{figure}

The distributions $P(S_{map})$ of both \textit{6d93} and \textit{4ake} are shown in Figure \ref{fig:wl_dofs}. In the former case, the sampling scheme produces a probability density that is fully compatible with the (normalized) histograms of the target values. Also, notice that the statistical weight of the optimized mappings here is negligible, but nonetheless this result is definitely non-trivial, as it proves that the trained DGN of \textit{6d93} is not in an overfitting regime and can predict the correct population of the true mapping entropy landscape.

As regards \textit{4ake}, the agreement between the two curves presented is still remarkable but not as precise as before. The slight mismatch is understandable if we consider the above regression scores: the DGN tends to underestimate (respecrively overestimate) the mapping entropy associated with random (optimized) \CG{} representations.

\clearpage

\section{Summary}
In this chapter, we have described the basic building blocks of deep learning methodologies able to learn from graph-structured data. In doing so, we managed to define different methods under the same uniform mathematical notation, so that we could easily understand peculiarities among the most popular neighborhood aggregation mechanisms in the literature as well as their nature \cite{bacciu_gentle_2020,wu_comprehensive_2020}. In turn, this also highlighted how the main underlying mechanisms had been already there for more than 10 years, in spite of the recent wave of re-discovery caused by the growing interest in the field. However, such an incredibly rapid stream of publications lacked a standardized, fair and robust experimental procedure in both vertex \cite{shchur_pitfalls_2018} and graph classification tasks, so we provided an \textit{empirical} re-evaluation of some of the most known works across a substantial number of benchmarks \cite{errica_fair_2020}. To convince the reader that these approaches are truly useful when applied to real-world problems, an application to the field of molecular biosciences was presented, in which we demonstrated how we can study a molecular system by quickly finding solutions with minimum mapping entropy \cite{errica_deep_2021}.

It is now time to move to the core topics of this thesis. We shall be dealing with a deep, fully probabilistic framework to learn from graphs with varying topology. We will use the knowledge acquired in the previous chapters to simplify the exposition and focus on the technical details that position such framework into the family of Deep Bayesian Graph Networks.

\chapter{Deep Bayesian Graph Networks}
\label{chapter:dbgn}
\epigraph{\textit{Lo mio maestro allora in su la gota \\ \mbox{destra si volse indietro, e riguardommi;}\\ poi disse: «Bene ascolta chi la nota».}}{\textit{Inferno - Canto XV}}

Deep Bayesian Graph Networks are fully probabilistic models for graphs whose architecture implements the principles of local and iterative computation described in the previous chapter. We will commence the discussion with the Contextual Graph Markov Model (\CGMM{}), a model bridging the gap between the NN4G \cite{micheli_neural_2009} and the recursive Bottom-up Hidden Tree Markov Model \cite{bacciu_bottom_2010}. We shall show how the neighborhood aggregation can be formalized and handled in a deep Bayesian framework, and we will evaluate its effectiveness on vertex and graph classification tasks. Then, we will extend \CGMM{} to the processing of arbitrary edge features. The resulting model, called Extended \CGMM{} (\ECGMM{}), uses an additional Bayesian network to model the generation of edge features, and its functioning is deeply intertwined with the original \CGMM{}'s graphical model. \ECGMM{} exhibits a form of dynamic neighborhood aggregation that contributes to the better performances of the model on graph classification, graph regression, and link prediction tasks. The third and last methodological contribution is the Infinite Contextual Graph Markov Model (\iCGMM{}), which extends \CGMM{} to the Bayesian nonparametric setting using an HDP. \iCGMM{} is capable of automatically selecting, on the basis of the available data, almost all \CGMM{}'s hyper-parameters, including the number of latent states at each layer. Empirically, we will bring evidence that \iCGMM{} has comparable or better performances than the original model while drastically reducing the size of the graph embeddings. We conclude the chapter with a real-world malware detection application that exploits the above models.

\clearpage
\section[The Contextual Graph Markov Model]{The Contextual Graph Markov Model \cite{bacciu_contextual_2018,bacciu_probabilistic_2020}}
\label{sec:cgmm}
The core contributions of this thesis, which we are about to present, are inspired by both the probabilistic topics of Chapter \ref{chapter:background} and the underlying principles of Deep Graph Networks developed in Chapter \ref{chapter:gentle-introduction}. For the rest of the chapter, we shall therefore depart from purely neural architectures and focus on a novel probabilistic framework to learn representations of graphs or vertices.

This section is devoted to the introduction of the Contextual Graph Markov Model (\CGMM{}) \cite{bacciu_contextual_2018,bacciu_probabilistic_2020}. As in the previous chapter, we shall adopt a top-down approach and first list the four main characteristics of the $\mathcal{T}_{enc}$ mapping:
\begin{itemize}
    \item \textbf{Unsupervised}. The model relies on the maximization of an unsupervised learning criterion, that is, the likelihood of the graph's entities, to adjust its parameters and construct vertex/graph embeddings. In principle, this means that the model can exploit large amounts of unlabelled data to produce richer vertex/graph embeddings on a given domain.
    \item \textbf{Fully Probabilistic}. Contrarily to other methods, which formalize the learning objective in probabilistic terms but approximate probability distributions with neural networks \cite{qu_gmnn_2019}, \CGMM{} relies on Bayesian networks to capture the 
    latent factors of vertex features. This makes \CGMM{} a fully probabilistic model and requires, as we will see, a completely probabilistic formulation of the neighborhood aggregation previously discussed.
    \item \textbf{Deep (Constructive)}. Following the principles of Deep Graph Networks, \CGMM{} is a deep feedforward architecture, where each layer is a distinct Bayesian network. This is enough to distinguish \CGMM{} from SRL approaches or recursive Bayesian networks for trees \cite{diligenti_hidden_2003,bacciu_generative_2012,bacciu_inputoutput_2013}, where the structure is taken into account in the formalization of the probabilistic model rather than by the message passing scheme of DGNs. In addition, and similarly to NN4G \cite{micheli_neural_2009}, the model is built in a constructive fashion by training one layer at a time. Once a \CGMM{} layer has been trained, it is \textit{frozen} and never modified again.
    \item \textbf{Efficient}. Last but not least, the model has the same asymptotic complexity as most DNGNs, being linear in the number of edge. Therefore, the model is amenable to large scale graph processing.
\end{itemize}
These characteristics make \CGMM{} a rather peculiar approach in the landscape of Deep Graph Networks. To show the richness of the extracted graph embeddings, we will use them in combination with a neural readout to tackle vertex and graph classification tasks.

\subsection{Layer Definition}
We have already seen how each layer $\lcurr$ of a DGN is responsible for the creation of intermediate vertex representations $\boldhell{u}{\lnext}$. Likewise, each \CGMM{}'s probabilistic layer assumes that the generation of vertex' features $\boldx{u}$ depends on some \textbf{latent factor} that we would like to capture. Hereinafter, for the purposes of this thesis, we will assume to deal with a single discrete or continuous feature $x_u$.

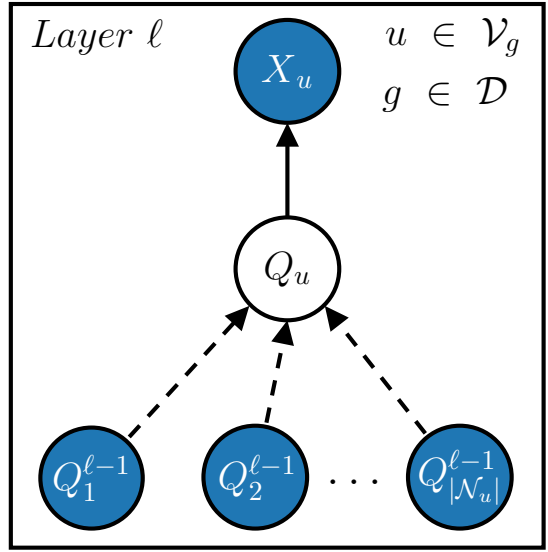
\begin{figure}[ht]
\begin{center}
\centerline{\resizebox{0.5\textwidth}{!}{\tikzset{every picture/.style={line width=0.75pt}} 

\begin{tikzpicture}[x=0.75pt,y=0.75pt,yscale=-1,xscale=1]

\draw [line width=1.5]  [dash pattern={on 5.63pt off 4.5pt}]  (194.52,255.42) -- (267.96,174.69) ;
\draw [shift={(270.65,171.73)}, rotate = 492.29] [fill={rgb, 255:red, 0; green, 0; blue, 0 }  ][line width=0.08]  [draw opacity=0] (11.61,-5.58) -- (0,0) -- (11.61,5.58) -- cycle    ;
\draw [line width=1.5]  [dash pattern={on 5.63pt off 4.5pt}]  (273.5,254.97) -- (288.2,182.65) ;
\draw [shift={(289,178.73)}, rotate = 461.49] [fill={rgb, 255:red, 0; green, 0; blue, 0 }  ][line width=0.08]  [draw opacity=0] (11.61,-5.58) -- (0,0) -- (11.61,5.58) -- cycle    ;
\draw  [fill={rgb, 255:red, 31; green, 119; blue, 180 }  ,fill opacity=1 ][line width=1.5]  (264,57.73) .. controls (264,43.93) and (275.19,32.73) .. (289,32.73) .. controls (302.81,32.73) and (314,43.93) .. (314,57.73) .. controls (314,71.54) and (302.81,82.73) .. (289,82.73) .. controls (275.19,82.73) and (264,71.54) .. (264,57.73) -- cycle ;
\draw [line width=1.5]  [dash pattern={on 5.63pt off 4.5pt}]  (372.17,254.97) -- (309.51,174.88) ;
\draw [shift={(307.05,171.73)}, rotate = 411.96000000000004] [fill={rgb, 255:red, 0; green, 0; blue, 0 }  ][line width=0.08]  [draw opacity=0] (11.61,-5.58) -- (0,0) -- (11.61,5.58) -- cycle    ;
\draw  [line width=1.5]  (264,153.73) .. controls (264,139.93) and (275.19,128.73) .. (289,128.73) .. controls (302.81,128.73) and (314,139.93) .. (314,153.73) .. controls (314,167.54) and (302.81,178.73) .. (289,178.73) .. controls (275.19,178.73) and (264,167.54) .. (264,153.73) -- cycle ;
\draw [line width=1.5]    (289,128.73) -- (289,97.07) -- (289,86.73) ;
\draw [shift={(289,82.73)}, rotate = 450] [fill={rgb, 255:red, 0; green, 0; blue, 0 }  ][line width=0.08]  [draw opacity=0] (11.61,-5.58) -- (0,0) -- (11.61,5.58) -- cycle    ;
\draw  [fill={rgb, 255:red, 31; green, 119; blue, 180 }  ,fill opacity=1 ][line width=1.5]  (169.83,255.42) .. controls (169.83,241.86) and (180.89,230.87) .. (194.52,230.87) .. controls (208.15,230.87) and (219.21,241.86) .. (219.21,255.42) .. controls (219.21,268.98) and (208.15,279.97) .. (194.52,279.97) .. controls (180.89,279.97) and (169.83,268.98) .. (169.83,255.42) -- cycle ;
\draw  [fill={rgb, 255:red, 31; green, 119; blue, 180 }  ,fill opacity=1 ][line width=1.5]  (248.5,254.97) .. controls (248.5,241.16) and (259.69,229.97) .. (273.5,229.97) .. controls (287.31,229.97) and (298.5,241.16) .. (298.5,254.97) .. controls (298.5,268.77) and (287.31,279.97) .. (273.5,279.97) .. controls (259.69,279.97) and (248.5,268.77) .. (248.5,254.97) -- cycle ;
\draw  [fill={rgb, 255:red, 31; green, 119; blue, 180 }  ,fill opacity=1 ][line width=1.5]  (347.17,254.97) .. controls (347.17,241.16) and (358.36,229.97) .. (372.17,229.97) .. controls (385.97,229.97) and (397.17,241.16) .. (397.17,254.97) .. controls (397.17,268.77) and (385.97,279.97) .. (372.17,279.97) .. controls (358.36,279.97) and (347.17,268.77) .. (347.17,254.97) -- cycle ;
\draw  [line width=1.5]  (155.58,25) -- (412.58,25) -- (412.58,289.25) -- (155.58,289.25) -- cycle ;

\draw (197,41.33) node  [font=\Large] [align=left] {$\displaystyle Layer$ $\displaystyle \ell $};
\draw (372.17,254.97) node  [font=\Large]  {$\textcolor[rgb]{1,1,1}{Q}\textcolor[rgb]{1,1,1}{_{|\mathcal{N}_{u} |}^{\ell -1}}$};
\draw (274.5,254.97) node  [font=\Large]  {$\textcolor[rgb]{1,1,1}{Q}\textcolor[rgb]{1,1,1}{_{2}^{\ell -1}}$};
\draw (194.52,255.42) node  [font=\Large]  {$\textcolor[rgb]{1,1,1}{Q}\textcolor[rgb]{1,1,1}{_{1}^{\ell -1}}$};
\draw (289,57.73) node  [font=\Large]  {$\textcolor[rgb]{1,1,1}{X}\textcolor[rgb]{1,1,1}{_{u}}$};
\draw (289,153.73) node  [font=\Large]  {$Q_{u}$};
\draw (323,257) node  [font=\LARGE,color={rgb, 255:red, 0; green, 0; blue, 0 }  ,opacity=1 ] [align=left] {$\displaystyle \dotsc $};
\draw (368.95,41.12) node  [font=\Large]  {$u\ \in \ \mathcal{V}_{g}$};
\draw (365.05,68.69) node  [font=\Large]  {$g\ \in \ \mathcal{D}$};

\end{tikzpicture}}}
\end{center}
\caption{Graphical model of a generic layer $\lcurr$ of \CGMM{}. Dashed arrows denote the flow of contextual information coming from previous layers.}
\label{fig:cgmm-basic-layer}
\end{figure}
We present the Bayesian network of a generic \CGMM{} layer $\lcurr$ in Figure \ref{fig:cgmm-basic-layer}. This is a \textbf{conditional mixture model} where we formally associate each vertex feature with an observable variable $X_u$, whose \textbf{adaptive} \textbf{emission} distribution $P(X_u \mid \Q{u}{})$ is conditioned on the latent \textbf{categorical} variable $\Q{u}{}$ with $C$ attainable values.

Intuitively, the latent variable $\Q{u}{}$ plays the role of the hidden state $\boldhell{u}{\lnext}$ in the general formulation devised in the previous chapter; that said, we follow the notational convention of Chapter \ref{chapter:background} when referring to random variables.

At first, it may seem that a strong assumption has been imposed here, \ie all vertices are \iid{}. 
In other words, we are completely disregarding the structural dependency between the set $\{X_u \mid u \in \Vset{g}\}$ of variables. However, having already outlined the basic principles of Deep Graph Networks, it should be clearer why this assumption works well: structural information has been encoded into the neighboring observable variables $\boldQ{\N{u}{}}{\lprev}=\{\Q{1}{\lprev},\dots,\Q{|\N{u}{}|}{\lprev}\}$ computed at the previous layer. This is why, with slight abuse of notation, dashed arrows in the figure indicate that there is contextual information flowing from the previous (frozen) layer $\lprev$. We will also use the symbol $\boldqell{v}{\lprev}$ to refer to the categorical distribution over the $C$ states (\ie a vector) inferred for the \textit{latent} variable $\Q{v}{\lprev}$ when training the previous layer. In addition, the $j$-th component of such distribution shall be $q^{\lprev}_v(j)$.

Formally, we can define the likelihood of a graph $g$ at layer $\lcurr$ as
\begin{align}
\mathcal{L}(\boldsymbol{\theta} \mid g) = P(g \mid \boldsymbol{\theta}) = \prod_{u \in \Vset{g}} \sum_{i=1}^{C}  \underbrace{P_{\boldsymbol{\theta}}(X_u=x_u | \Q{u}{} = i)}_{emission} P(\Q{u}{} = i \mid \boldQ{\N{u}{}}{\lprev}).
\label{eq:basic-cgmm-graph-likelihood}
\end{align}
As in standard mixture models, we have introduced the latent variable $\Q{u}{}$ in the equation via marginalization. However, not knowing the size of $\N{u}{}$ for each vertex $u$ makes the definition of the posterior distribution $P(\Q{u}{} = i \mid \boldQ{\N{u}{}}{\lprev})$ quite hard to formalize, and conditioning on all neighboring states becomes rapidly intractable because of the exponential growth in the number of possible combinations (\ie $\mathcal{O}(C^{|\N{u}{}|})$). What is worse, the cardinality of the neighbors may vary, so either we assume a maximum size of all neighbors' sets or we rely on permutation invariant functions like DGNs do. We opt for the latter option and weigh the contributions of the neighboring states equally using the \textbf{mean} operator:
\begin{align}
P(\Q{u}{} = i \mid \boldQ{\N{u}{}}{\lprev}) \approx \frac{1}{|\N{u}{}|}\sum_j^C \underbrace{P_{\boldsymbol{\theta}}(\Q{u}{} = i \mid \Q{*}{\lprev}=j)}_{transition}\sum_{v \in \N{u}{}}q^{\lprev}_v(j).
\label{eq:basic-cgmm-avg-neighbourhood}
\end{align}
We can intuitively understand the last equation by imagining that all neighboring variables have been collapsed into a \quotes{macro-variable} $\Q{*}{\lprev}$, whose categorical distribution is given by the element-wise mean of the individual distributions (viewed as $C$-sized vectors), \ie the probability of $\Q{*}{\lprev}$ being in state $j$ is $\frac{1}{|\N{u}{}|}\sum_{v \in \N{u}{}}q^{\lprev}_v(j)$.

While it is true that each neighbor is weighted equally (\ie the $\frac{1}{|\N{u}{}|}$ term), the \textbf{adaptive} \textbf{transition} distribution weights the contribution of any neighboring state differently according to the arrival state $i$. Crucially, since we assume \textbf{full stationarity} of \textbf{all} adaptive distributions, the identities of a neighbor or the vertex itself are irrelevant to the parametrization of such distributions.

Another peculiar characteristic of this aggregation is that we do not necessarily weight the most likely state of each neighboring variable $\Q{*}{\lprev}$, but rather we consider the entire probability mass specified in the distribution $\boldqell{v}{\lprev}$: Section \ref{subsec:inference} will provide more details about this point.

\subsection{Enhancing the Neighborhood Aggregation}
The neighborhood aggregation scheme presented above ensures that the rightmost term of Equation \ref{eq:basic-cgmm-graph-likelihood} is still a valid probability, thus allowing us to find closed-form solutions when training the layer with the exact EM algorithm (details are provided later). However, this aggregation is limited in two respects: first, it does not take into account more than one previous layer, similarly to what skip connections do in deep neural networks \cite{he_deep_2016,micheli_neural_2009,kipf_semi-supervised_2017}; secondly, it ignores the presence of edge features. Inspired by bottom-up generative models for tree-structure data \cite{bacciu_generative_2012,bacciu_inputoutput_2013}, we address these limitations by means of the so-called Switching Parent (SP) approximation \cite{saul_mixed_1999,bacciu_inputoutput_2013}.

The goal is to modify the above equations to consider contributions from an arbitrary subset $\mathbb{L}(\lcurr)$ of previous layers as well as a finite number of \textbf{discrete} edge labels; to this aim, we introduce the random categorical variables $L_u$ and $S_u$, respectively. Mathematically, the role of a Switching Parent variable $\Xi$ is to decompose a complex conditional distribution over variables (let us call them $I$) into a convex combination of simpler ones
\begin{equation*}
P(I_0=i_0| I_1=i_1,\dots,I_k=i_k) \approx \sum_{\xi=1}^k P(\Xi = \xi)P^{\xi}(I_0=i_0| I_{\xi}=i_{\xi}),
\end{equation*}
where the rightmost \textbf{transition} probability depends on the value $\xi$ of the SP variable.

\begin{figure}[ht]
\begin{center}
\centerline{\resizebox{0.9\textwidth}{!}{\tikzset{every picture/.style={line width=0.75pt}} 

\begin{tikzpicture}[x=0.75pt,y=0.75pt,yscale=-1,xscale=1]

\draw  [fill={rgb, 255:red, 31; green, 119; blue, 180 }  ,fill opacity=1 ][line width=1.5]  (244,78) .. controls (244,64.19) and (255.19,53) .. (269,53) .. controls (282.81,53) and (294,64.19) .. (294,78) .. controls (294,91.81) and (282.81,103) .. (269,103) .. controls (255.19,103) and (244,91.81) .. (244,78) -- cycle ;
\draw [line width=1.5]  [dash pattern={on 5.63pt off 4.5pt}]  (444.25,290.92) -- (290.63,194.05) ;
\draw [shift={(287.25,191.92)}, rotate = 392.23] [fill={rgb, 255:red, 0; green, 0; blue, 0 }  ][line width=0.08]  [draw opacity=0] (11.61,-5.58) -- (0,0) -- (11.61,5.58) -- cycle    ;
\draw  [line width=1.5]  (244,174) .. controls (244,160.19) and (255.19,149) .. (269,149) .. controls (282.81,149) and (294,160.19) .. (294,174) .. controls (294,187.81) and (282.81,199) .. (269,199) .. controls (255.19,199) and (244,187.81) .. (244,174) -- cycle ;
\draw  [line width=1.5]  (338,119) .. controls (338,105.19) and (349.19,94) .. (363,94) .. controls (376.81,94) and (388,105.19) .. (388,119) .. controls (388,132.81) and (376.81,144) .. (363,144) .. controls (349.19,144) and (338,132.81) .. (338,119) -- cycle ;
\draw [line width=1.5]    (340.25,131.58) -- (293.75,157.16) ;
\draw [shift={(290.25,159.08)}, rotate = 331.19] [fill={rgb, 255:red, 0; green, 0; blue, 0 }  ][line width=0.08]  [draw opacity=0] (11.61,-5.58) -- (0,0) -- (11.61,5.58) -- cycle    ;
\draw [line width=1.5]    (269,149) -- (269,117.33) -- (269,107) ;
\draw [shift={(269,103)}, rotate = 450] [fill={rgb, 255:red, 0; green, 0; blue, 0 }  ][line width=0.08]  [draw opacity=0] (11.61,-5.58) -- (0,0) -- (11.61,5.58) -- cycle    ;
\draw  [fill={rgb, 255:red, 31; green, 119; blue, 180 }  ,fill opacity=1 ][line width=1.5]  (60.83,302.23) .. controls (60.83,288.43) and (72.03,277.23) .. (85.83,277.23) .. controls (99.64,277.23) and (110.83,288.43) .. (110.83,302.23) .. controls (110.83,316.04) and (99.64,327.23) .. (85.83,327.23) .. controls (72.03,327.23) and (60.83,316.04) .. (60.83,302.23) -- cycle ;
\draw  [fill={rgb, 255:red, 31; green, 119; blue, 180 }  ,fill opacity=1 ][line width=1.5]  (119.5,302.23) .. controls (119.5,288.43) and (130.69,277.23) .. (144.5,277.23) .. controls (158.31,277.23) and (169.5,288.43) .. (169.5,302.23) .. controls (169.5,316.04) and (158.31,327.23) .. (144.5,327.23) .. controls (130.69,327.23) and (119.5,316.04) .. (119.5,302.23) -- cycle ;
\draw  [fill={rgb, 255:red, 31; green, 119; blue, 180 }  ,fill opacity=1 ][line width=1.5]  (288.83,302.23) .. controls (288.83,288.43) and (300.03,277.23) .. (313.83,277.23) .. controls (327.64,277.23) and (338.83,288.43) .. (338.83,302.23) .. controls (338.83,316.04) and (327.64,327.23) .. (313.83,327.23) .. controls (300.03,327.23) and (288.83,316.04) .. (288.83,302.23) -- cycle ;
\draw  [fill={rgb, 255:red, 31; green, 119; blue, 180 }  ,fill opacity=1 ][line width=1.5]  (348.5,302.23) .. controls (348.5,288.43) and (359.69,277.23) .. (373.5,277.23) .. controls (387.31,277.23) and (398.5,288.43) .. (398.5,302.23) .. controls (398.5,316.04) and (387.31,327.23) .. (373.5,327.23) .. controls (359.69,327.23) and (348.5,316.04) .. (348.5,302.23) -- cycle ;
\draw  [fill={rgb, 255:red, 31; green, 119; blue, 180 }  ,fill opacity=1 ][line width=1.5]  (442.17,302.23) .. controls (442.17,288.43) and (453.36,277.23) .. (467.17,277.23) .. controls (480.97,277.23) and (492.17,288.43) .. (492.17,302.23) .. controls (492.17,316.04) and (480.97,327.23) .. (467.17,327.23) .. controls (453.36,327.23) and (442.17,316.04) .. (442.17,302.23) -- cycle ;
\draw  [fill={rgb, 255:red, 31; green, 119; blue, 180 }  ,fill opacity=1 ][line width=1.5]  (205.5,302.23) .. controls (205.5,288.43) and (216.69,277.23) .. (230.5,277.23) .. controls (244.31,277.23) and (255.5,288.43) .. (255.5,302.23) .. controls (255.5,316.04) and (244.31,327.23) .. (230.5,327.23) .. controls (216.69,327.23) and (205.5,316.04) .. (205.5,302.23) -- cycle ;
\draw [line width=1.5]  [dash pattern={on 5.63pt off 4.5pt}]  (356.83,282.06) -- (284.79,199.08) ;
\draw [shift={(282.17,196.06)}, rotate = 409.03] [fill={rgb, 255:red, 0; green, 0; blue, 0 }  ][line width=0.08]  [draw opacity=0] (11.61,-5.58) -- (0,0) -- (11.61,5.58) -- cycle    ;
\draw [line width=1.5]  [dash pattern={on 5.63pt off 4.5pt}]  (304.83,279.39) -- (277.49,201.16) ;
\draw [shift={(276.17,197.39)}, rotate = 430.73] [fill={rgb, 255:red, 0; green, 0; blue, 0 }  ][line width=0.08]  [draw opacity=0] (11.61,-5.58) -- (0,0) -- (11.61,5.58) -- cycle    ;
\draw [line width=1.5]  [dash pattern={on 5.63pt off 4.5pt}]  (236.83,276.06) -- (259.01,202.55) ;
\draw [shift={(260.17,198.72)}, rotate = 466.79] [fill={rgb, 255:red, 0; green, 0; blue, 0 }  ][line width=0.08]  [draw opacity=0] (11.61,-5.58) -- (0,0) -- (11.61,5.58) -- cycle    ;
\draw [line width=1.5]  [dash pattern={on 5.63pt off 4.5pt}]  (162.17,281.39) -- (248.68,194.23) ;
\draw [shift={(251.5,191.39)}, rotate = 494.79] [fill={rgb, 255:red, 0; green, 0; blue, 0 }  ][line width=0.08]  [draw opacity=0] (11.61,-5.58) -- (0,0) -- (11.61,5.58) -- cycle    ;
\draw [line width=1.5]  [dash pattern={on 5.63pt off 4.5pt}]  (107.5,285.39) -- (243.57,189.03) ;
\draw [shift={(246.83,186.72)}, rotate = 504.7] [fill={rgb, 255:red, 0; green, 0; blue, 0 }  ][line width=0.08]  [draw opacity=0] (11.61,-5.58) -- (0,0) -- (11.61,5.58) -- cycle    ;
\draw  [line width=1.5]  (338,195) .. controls (338,181.19) and (349.19,170) .. (363,170) .. controls (376.81,170) and (388,181.19) .. (388,195) .. controls (388,208.81) and (376.81,220) .. (363,220) .. controls (349.19,220) and (338,208.81) .. (338,195) -- cycle ;
\draw [line width=1.5]    (363,170) -- (363,148) ;
\draw [shift={(363,144)}, rotate = 450] [fill={rgb, 255:red, 0; green, 0; blue, 0 }  ][line width=0.08]  [draw opacity=0] (11.61,-5.58) -- (0,0) -- (11.61,5.58) -- cycle    ;
\draw [line width=1.5]    (337.75,190.25) -- (297.14,180.67) ;
\draw [shift={(293.25,179.75)}, rotate = 373.28] [fill={rgb, 255:red, 0; green, 0; blue, 0 }  ][line width=0.08]  [draw opacity=0] (11.61,-5.58) -- (0,0) -- (11.61,5.58) -- cycle    ;
\draw  [line width=1.5]  (52.55,44) -- (500.55,44) -- (500.55,339.42) -- (52.55,339.42) -- cycle ;

\draw (269,174) node  [font=\Large]  {$Q_{u}$};
\draw (363,119) node  [font=\Large]  {$S_{u}$};
\draw (269,78) node  [font=\Large]  {$\textcolor[rgb]{1,1,1}{X_{u}}$};
\draw (85.83,302.23) node  [font=\Large]  {$\textcolor[rgb]{1,1,1}{Q}\textcolor[rgb]{1,1,1}{_{1}^{1}}$};
\draw (144.5,302.23) node  [font=\Large]  {$\textcolor[rgb]{1,1,1}{Q_{2}^{1}}$};
\draw (467.17,302.23) node  [font=\Large]  {$\textcolor[rgb]{1,1,1}{Q_{|\mathcal{N}_{u} |}^{\ell -1}}$};
\draw (373.5,302.23) node  [font=\Large]  {$\textcolor[rgb]{1,1,1}{Q_{2}^{\ell -1}}$};
\draw (313.83,302.23) node  [font=\Large]  {$\textcolor[rgb]{1,1,1}{Q_{1}^{\ell -1}}$};
\draw (230.5,302.23) node  [font=\Large]  {$\textcolor[rgb]{1,1,1}{Q_{|\mathcal{N}_{u} |}^{1}}$};
\draw (187.9,306.27) node  [font=\large,color={rgb, 255:red, 0; green, 0; blue, 0 }  ,opacity=1 ]  {$\mathbf{\dotsc }$};
\draw (271.5,306.27) node  [font=\large,color={rgb, 255:red, 0; green, 0; blue, 0 }  ,opacity=1 ]  {$\mathbf{\dotsc }$};
\draw (416.5,305.77) node  [font=\large,color={rgb, 255:red, 0; green, 0; blue, 0 }  ,opacity=1 ]  {$\mathbf{\dotsc }$};
\draw (94,69) node  [font=\Large] [align=left] {$\displaystyle Layer$\textit{ }$\displaystyle \ell $};
\draw (363,195) node  [font=\Large]  {$L_{u}$};
\draw (453.2,67.12) node  [font=\Large]  {$u\ \in \ \mathcal{V}_{g}$};
\draw (452,101.69) node  [font=\Large]  {$g\ \in \ \mathcal{D}$};

\end{tikzpicture}}}
\end{center}
\caption{Graphical model of the \quotes{full} \CGMM{} at layer $\lcurr$. The SP variables weight the frozen neighboring states in relation to their layer and edge type. Dashed arrows denote the flow of contextual information coming from previous layers.}
\label{fig:cgmm-extended-layer}
\end{figure}
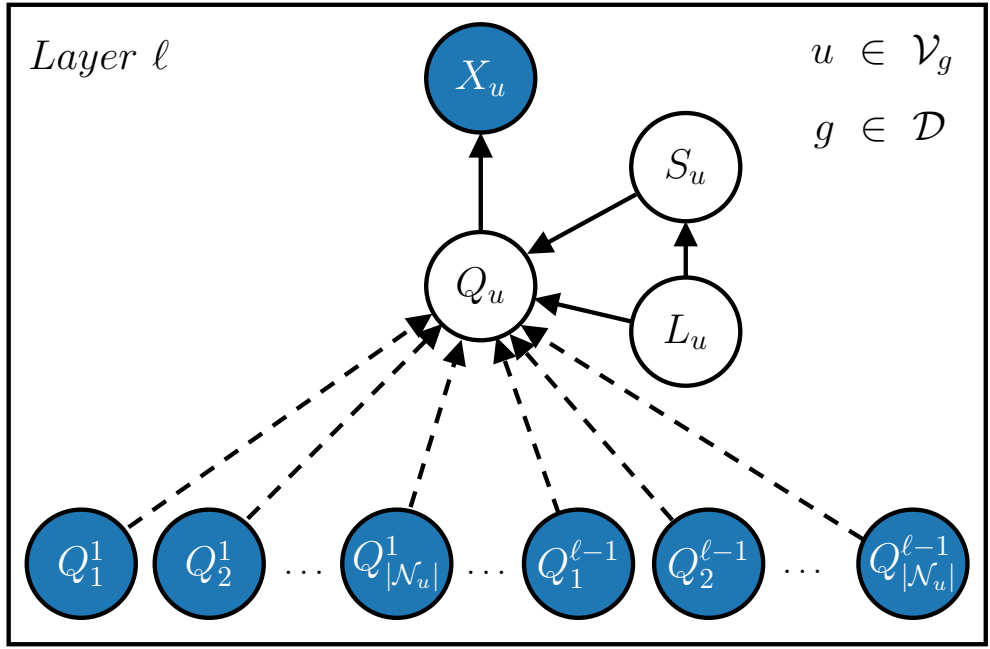

The finite cardinality of the sets $\mathbb{L}(\lcurr)$ and $\Aset{g}$ makes it possible to apply the SP approximation to our \CGMM{} layer. The SP technique will first assign a specific weight to frozen neighboring states computed at different layers. In addition, for each layer, neighbors of vertex $u$ connected with diverse edge types will be weighted differently as well. If we go back to the \quotes{macro-state} idealization, this corresponds to grouping neighboring variables into many macro-states, according to their relation with the previous layers and edge types. We give a graphical overview of the extended probabilistic layer in Figure \ref{fig:cgmm-extended-layer}

Hence, we can see Equation \ref{eq:basic-cgmm-avg-neighbourhood} as a special case of the following neighborhood aggregation (considering the extended set of neighboring observables $\boldQ{\N{u}{}}{\mathbb{L}(\lcurr)}$):
\begin{align}
P(\Q{u}{} = i \mid \boldQ{\N{u}{}}{\mathbb{L}(\lcurr)}) \approx & \sum_{\lcurr' \in \mathbb{L}(\lcurr)} \underbrace{P_{\boldsymbol{\theta}}(L_u = \lcurr')}_{\text{\textit{SP layer}}} \sum_{a=1}^{|\Aset{g}|} \underbrace{P^{\lcurr'}_{\boldsymbol{\theta}}(S_u = a)}_{\text{\textit{SP edge}}} \times \nonumber \\
& \times \frac{1}{|\N{u}{\lcurr',a}|}\sum_j^C \underbrace{P^{\lcurr',a}_{\boldsymbol{\theta}}(\Q{u}{} = i \mid \Q{*}{\lcurr',a}=j)}_{\text{\textit{SP-aware transition}}}\sum_{v \in \N{u}{\lcurr',a}}q^{\lcurr'}_v(j),
\label{eq:cgmm-sp-decomposition}
\end{align}
where $\N{u}{\lcurr',a}=\N{u}{a}$ defines the subset of neighboring observables computed at layer $\lcurr'$, whose associated vertices are connected to $u$ with edge label $a$. Notice how we have adopted \textbf{positional stationarity} for the transition distribution and the switching parent $S_u$: the distributions are dependent on the layer and edge type we are considering. Similarly, the variable $\Q{*}{\lcurr',a}$ identifies the macro-state obtained by averaging the neighboring contributions in $\N{u}{a}$.

Computationally speaking, each term $\frac{1}{|\N{u}{a}|}q^{\lcurr'}_v(j)$ is constant and can be pre-computed before training the current \CGMM{} layer. Thanks to the incremental construction, this can substantially speed up the training process. From now on, when needed, we will talk about \quotes{pre-computed statistics} or simply \textbf{statistics}.

The Switching Parent approximation fits well inside the \CGMM{} layer because it does not require reasoning about all layers \textit{simultaneously}. Different approaches, such as Recurrent Neural Networks or Hidden Markov Models, assume that the \quotes{history} of states is not frozen and can change altogether through a \textit{shared} transition function across layers. This design choice, however, would reintroduce the mutual dependencies between unobserved variables that we are trying to break with the incremental construction or, more generally, with the local and iterative processing of information described in Chapter \ref{chapter:gentle-introduction}. Thus, the SP variables provide a formal way to consider \quotes{skip connections} and discrete edge features in a \textbf{probabilistic} framework.

To summarize, the likelihood of a graph under the extended formulation of \CGMM{} can be therefore written as
\begin{align}
\likelihood(\boldsymbol{\theta} \mid g) & = \prod_{u \in \Vset{g}} \sum_{i=1}^{C}  P_{\boldsymbol{\theta}}(X_u=x_u | \Q{u}{} = i) P(\Q{u}{} = i \mid \boldQ{\N{u}{}}{\mathbb{L}(\lcurr)}) \approx \nonumber \\
& \approx \prod_{u \in \Vset{g}} \sum_{i=1}^{C}  \underbrace{P_{\boldsymbol{\theta}}(X_u=x_u | \Q{u}{} = i)}_{emission}\sum_{\lcurr' \in \mathbb{L}(\lcurr)} \underbrace{P_{\boldsymbol{\theta}}(L_u = \lcurr')}_{\text{\textit{SP layer}}} \sum_{a=1}^{|\Aset{g}|} \underbrace{P^{\lcurr'}_{\boldsymbol{\theta}}(S_u = a)}_{\text{\textit{SP edge}}} \times \nonumber \\
& \times \frac{1}{|\N{u}{a}|}\sum_j^C \underbrace{P^{\lcurr',a}_{\boldsymbol{\theta}}(\Q{u}{} = i \mid \Q{*}{\lcurr',a}=j)}_{\text{\textit{SP-aware transition}}}\sum_{v \in \N{u}{a}}q^{\lcurr'}_v(j).
\label{eq:cgmm-extended-likelihood}
\end{align}

\subsection{Training}
When training the $\lcurr$-th layer of \CGMM{}, we maximize the likelihood of the data using the EM algorithm, by assuming that the graphs in the dataset $\dataset{}$ are \iid{}. Similarly to the standard mixture model training, to compute the E-step we introduce the set of indicator random variables $\boldsymbol{Z}$. In particular, $Z_{ui\lcurr' aj}=1$ if the latent variable $\Q{u}{}$ of vertex $u$ has value $i$ while its observable neighboring variables coming from layer $\lcurr'$ are in state $j$ and are connected to $u$ with edge type $a$, and 0 otherwise. Note that there can also be other indicator variables, \eg $Z_{ui\lcurr'a}$, where we express no interest in knowing the value of one or more subscripts of $Z_{ui\lcurr' aj}$. Using knowledge coming from $\boldsymbol{Z}$, we can write the complete log-likelihood formula to be maximized (omitting $\boldsymbol{\theta}$ to simplify the notation):
\begin{align}
\log \likelihood_c(\boldsymbol{\theta}\mid \boldsymbol{Z},\dataset{}) & = \log \prod_{\substack{g \in \dataset{} \\ u \in \Vset{g}}}\prod_{i}^{C} \Bigg\{ P(x_u | \Q{u}{} = i) \prod_{\lcurr' \in \mathbb{L}(\lcurr)} \bigg\{P(L_u = \lcurr')\prod_{a=1}^{|\Aset{g}|} \Big\{P^{\lcurr'}(S_u = a) \nonumber \times \\ & \times \prod_j^C\big\{\frac{P^{\lcurr',a}(\Q{u}{}=i | \Q{*}{\lcurr',a}=j) \sum_{v \in \N{u}{a}}q^{\lcurr'}_v(j)}{|\N{u}{a}|} \big\}^{Z_{ui\lcurr' aj}} \Big\}^{Z_{ui\lcurr' a}} \bigg\}^{Z_{ui\lcurr'}} \Bigg\}^{Z_{ui}} \nonumber \\
& = \sum_{\substack{g \in \dataset{} \\ u \in \Vset{g}}} \sum_{i}^{C} Z_{ui} \log P(x_u | \Q{u}{} = i) + \sum_{\substack{g \in \dataset{} \\ u \in \Vset{g}}} \sum_{i}^{C} \sum_{\lcurr' \in \mathbb{L}(\lcurr)} Z_{ui\lcurr'} \log P(L_u = \lcurr') \nonumber\\
& + \sum_{\substack{g \in \dataset{} \\ u \in \Vset{g}}} \sum_{i}^{C} \sum_{\lcurr' \in \mathbb{L}(\lcurr)} \sum_{a=1}^{|\Aset{g}|} Z_{ui\lcurr' a} \log P^{\lcurr'}(S_u = a) \nonumber \\
& + \sum_{\substack{g \in \dataset{} \\ u \in \Vset{g}}} \sum_{i}^{C} \sum_{\lcurr' \in \mathbb{L}(\lcurr)} \sum_{a=1}^{|\Aset{g}|} \sum_j^C Z_{ui\lcurr' aj} \log \frac{P^{\lcurr',a}(\Q{u}{}=i | Q_{*}^{\lcurr',a}=j)\sum_{v \in \N{u}{a}}q^{\lcurr'}_v(j)}{|\N{u}{a}|}.
\label{eq:cgmm-complete-log-likelihood}
\end{align}
At layer $\lcurr=0$, since there are no neighboring states to consider, the problem reduces to maximizing the complete log-likelihood of a standard mixture model (Section \ref{subsec:mixture-model-basic}).

\subsubsection*{E-step}

The E-step of the EM algorithm requires to compute the \textbf{expectation} of the complete log-likelihood \wrt{} $\boldsymbol{Z}$. Thanks to the properties of expectation, this is equivalent to replace each indicator variable in Equation \ref{eq:cgmm-complete-log-likelihood} with its conditional expectation. Therefore, we define the following terms:
\begin{align*}
& \mathbb{E}[Z_{ui} | \dataset{},\boldQ{\N{u}{}}{\mathbb{L}(\lcurr)}] = P(\Q{u}{} = i | \dataset{},\boldQ{\N{u}{}}{\mathbb{L}(\lcurr)}) \\
& \mathbb{E}[Z_{ui\lcurr'} | \dataset{},\boldQ{\N{u}{}}{\mathbb{L}(\lcurr)}] = P(\Q{u}{} = i, L_u = \lcurr' | \dataset{},\boldQ{\N{u}{}}{\mathbb{L}(\lcurr)}) \\
& \mathbb{E}[Z_{ui\lcurr' a} | \dataset{},\boldQ{\N{u}{}}{\mathbb{L}(\lcurr)}] = P(\Q{u}{} = i, L_u = \lcurr', S_u = a | \dataset{},\boldQ{\N{u}{}}{\mathbb{L}(\lcurr)}) \\
& \mathbb{E}[Z_{ui\lcurr' aj} | \dataset{},\boldQ{\N{u}{}}{\mathbb{L}(\lcurr)}] = P(\Q{u}{} = i, L_u = \lcurr', S_u = a, K_u = j | \dataset{},\boldQ{\N{u}{}}{\mathbb{L}(\lcurr)}),
\end{align*}
noting that the first three terms can be straightforwardly obtained from the last one via marginalization. To formally model the aggregation process, we had to introduce the \quotes{macro-state} categorical variable $K_u$ with $C$ possible values, such that $P^{\lcurr',a}(K_u = j) = \sum_{v \in \N{u}{a}}q^{\lcurr'}_v(j)/|\N{u}{a}|$.
Consequently, we can apply the Bayes Theorem on $\mathbb{E}[Z_{ui\lcurr' aj} | \dataset{},\boldQ{\N{u}{}}{\mathbb{L}(\lcurr)}]$, yielding
\begin{align*}
& \mathbb{E}[Z_{ui\lcurr' aj} | \dataset{},\boldQ{\N{u}{}}{\mathbb{L}(\lcurr)}] = P(\Q{u}{} = i, L_u = \lcurr', S_u = a, K_u = j | \dataset{},\boldQ{\N{u}{}}{\mathbb{L}(\lcurr)}) \nonumber \\
& = \frac{P(x_u | \Q{u}{} = i)P(\Q{u}{} = i, L_u = \lcurr', S_u = a, K_u = j | \boldQ{\N{u}{}}{\mathbb{L}(\lcurr)})}{P(x_u|\boldQ{\N{u}{}}{\mathbb{L}(\lcurr)})} \nonumber \\
& = \frac{P(x_u | i)P(i | L_u = \lcurr', S_u = a, K_u = j, \boldQ{\N{u}{}}{\mathbb{L}(\lcurr)})P(L_u = \lcurr')P^{\lcurr'}(S_u = a)P^{\lcurr',a}(K_u = j)}{Z_{norm}} \nonumber \\
& = \frac{P(x_u | \Q{u}{} = i)P^{\lcurr',a}(\Q{u}{} = i | \Q{*}{\lcurr',a}=j)P(L_u = \lcurr')P^{\lcurr'}(S_u = a)P^{\lcurr',a}(K_u = j)}{Z_{norm}} \nonumber \\
& = \frac{P(x_u | \Q{u}{} = i)P(L_u = \lcurr')P^{\lcurr'}(S_u = a)P^{\lcurr',a}(Q=i | \Q{*}{\lcurr',a}=j)P^{\lcurr',a}(K_u = j)}{Z_{norm}},
\end{align*}
where $Z_{norm}$ is the normalization term, obtained by $P(x_u|\boldQ{\N{u}{}}{\mathbb{L}(\lcurr)})$ via marginalization over all the latent variables (including $K_u$).

As we can see, the E-step can be computed without relying on variational approximations. Also, these operations can be easily parallelized due to the \iid assumption between vertices, making the training of each layer scalable to larger graphs and amenable to GPU processing.
\subsubsection*{M-step}
We use the posterior probabilities obtained in the E-step to update the parameters of the \CGMM{} layer. After the addition of a suitable Lagrange multiplier to enforce probability requirements, we can obtain closed-form solutions for each  adaptive distribution $P_{\boldsymbol{\theta}}$ in the following way \cite{bacciu_probabilistic_2020}:
\begin{enumerate}
    \item Compute the gradient of the \textbf{expected} complete log-likelihood \wrt{} $P_{\boldsymbol{\theta}}$. The resulting equation will depend on the Lagrange multiplier;
    \item Compute the gradient of the \textbf{expected} complete log-likelihood \wrt{} the Lagrange multiplier, and plug the result into the resulting equation of the previous point.
\end{enumerate}

We end up with the following update equations (omitting the subscript $u$ to express stationarity of the learned distributions):
\paragraph*{Transition Distribution}
\begin{align*}
& P^{\lcurr',a}(Q = i | \Q{*}{\lcurr',a} = j) = \frac{\sum_{\substack{g \in \dataset{} \\ u \in \Vset{g}}}\mathbb{E}[z_{ui\lcurr' aj}|\dataset{},\boldQ{\N{u}{}}{\mathbb{L}(\lcurr)}]}{ \sum_{\substack{g \in \dataset{} \\ u \in \Vset{g}}} \sum_{i'=1}^C \mathbb{E}[z_{ui'\lcurr' aj}|\dataset{},\boldQ{\N{u}{}}{\mathbb{L}(\lcurr)}]}.
\end{align*}
\paragraph*{Switching Parents Distributions}
\begin{align*}
& P(L = \lcurr') = \frac{\sum_{\substack{g \in \dataset{} \\ u \in \Vset{g}}} \sum_{i=1}^C \mathbb{E}[z_{ui\lcurr'} | \dataset{},\boldQ{\N{u}{}}{\mathbb{L}(\lcurr)}]}{\sum_{\substack{g \in \dataset{} \\ u \in \Vset{g}}} \sum_{i=1}^C \sum_{\lcurr'' \in \mathbb{L}(\lcurr)} \mathbb{E}[z_{ui\lcurr''} | \dataset{},\boldQ{\N{u}{}}{\mathbb{L}(\lcurr)}]}, \\
& P^{\lcurr'}(S = a) = \frac{\sum_{\substack{g \in \dataset{} \\ u \in \Vset{g}}} \sum_{i}^{C} \mathbb{E}[z_{ui\lcurr' a} | \dataset{},\boldQ{\N{u}{}}{\mathbb{L}(\lcurr)}]}{\sum_{\substack{g \in \dataset{} \\ u \in \Vset{g}}} \sum_{i}^{C} \sum_{a'=1}^{|\Aset{g}|} \mathbb{E}[z_{ui\lcurr' a'} | \dataset{},\boldQ{\N{u}{}}{\mathbb{L}(\lcurr)}]}.
\end{align*}
\paragraph*{Categorical Emission Distribution}
\begin{align*}
    & P(X=k|Q=i) = \frac{\sum_{\substack{g \in \dataset{} \\ u \in \Vset{g}}} \delta(x_u, k) \mathbb{E}[z_{ui} | \dataset{},\boldQ{\N{u}{}}{\mathbb{L}(\lcurr)}]}{\sum_{\substack{g \in \dataset{} \\ u \in \Vset{g}}}\sum_{k'}^K \delta(x_u, k') \mathbb{E}[z_{ui} | \dataset{},\boldQ{\N{u}{}}{\mathbb{L}(\lcurr)}]}. \\
\end{align*}
\paragraph*{Gaussian Emission Distribution}
\begin{align}
& \mu_i = \frac{\sum_{\substack{g \in \dataset{} \\ u \in \Vset{g}}} x_u \mathbb{E}[z_{ui} | \dataset{}, \boldQ{\N{u}{}}{\mathbb{L}(\lcurr)}]}{\sum_{\substack{g \in \dataset{} \\ u \in \Vset{g}}} \mathbb{E}[z_{ui} | \dataset{}, \boldQ{\N{u}{}}{\mathbb{L}(\lcurr)}]}, \nonumber \\
& \sigma_i = \sqrt{\frac{\sum_{\substack{g \in \dataset{} \\ u \in \Vset{g}}} \mathbb{E}[z_{ui} | \dataset{}, \boldQ{\N{u}{}}{\mathbb{L}(\lcurr)}](x_u - \mu_i)^2}{\sum_{\substack{g \in \dataset{} \\ u \in \Vset{g}}} \mathbb{E}[z_{ui} | \dataset{}, \boldQ{\N{u}{}}{\mathbb{L}(\lcurr)}]}}. \nonumber
\end{align}
As already mentioned, for the purposes of this thesis we will mainly focus on categorical and univariate Gaussian emission distributions. For multidimensional features, one could simply assume their conditional independence or rely on the vast amount of literature available \cite{bishop_pattern_2006,barber_bayesian_2012}.

\subsection{Inference}
\label{subsec:inference}
During inference, we compute the most likely \textit{index} associated with the posterior of $Q_u$ as representative for vertex $u$. In other words, it assigns $u$ to one of the $C$ potential clusters. Formally, this can be expressed as
\begin{equation}
\label{eq:inference}
\max_i P(\Q{u}{}=i|g,\boldQ{\N{u}{}}{\mathbb{L}(\lcurr)}) = \max_i \frac{P(x_u| \Q{u}{} = i)P(\Q{u}{} = i | \boldQ{\N{u}{}}{\mathbb{L}(\lcurr)})}{\cancel{P(x_u|\boldQ{\N{u}{}}{\mathbb{L}(\lcurr)})}}.
\end{equation}
The equivalence is obtained by straightforward application of the Bayes Theorem; moreover, the denominator does not contribute to the maximization because it is independent of the state $i$, so it can be ignored.

Mathematically speaking, when training a \CGMM{} layer, we formalized the neighborhood aggregation using the entire \textit{frozen} posterior distribution of each vertex rather than its most likely state. Note, however, that our formalization is general enough to allow the use of either the former (continuous) representation or its one-hot variant, \ie collapsing all probability mass into the most likely state.

To understand why this matters, let us assume $C=3$ and consider the frozen posterior distribution of a vertex inferred at the previous layer being $(0.4, 0, 0.6)$ or $(0, 0.4, 0.6)$. Clearly, collapsing all the posterior mass onto the most likely state discards important information about the probability of being in the others. Therefore, there is a trade-off between a less noisy but approximate one-hot representation and a possibly noisy but exact one. We treat this choice as a hyper-parameter to be selected.

\subsection{Building Graph Representations}
\begin{figure}[h]
\begin{center}
\centerline{\resizebox{1\textwidth}{!}{\tikzset{every picture/.style={line width=0.75pt}} 

\begin{tikzpicture}[x=0.75pt,y=0.75pt,yscale=-1,xscale=1]

\draw  [fill={rgb, 255:red, 255; green, 255; blue, 255 }  ,fill opacity=1 ][line width=1.5]  (130,36) -- (227,36) -- (227,133) -- (130,133) -- cycle ;
\draw  [fill={rgb, 255:red, 255; green, 255; blue, 255 }  ,fill opacity=1 ][line width=1.5]  (150,67) -- (247,67) -- (247,164) -- (150,164) -- cycle ;
\draw [line width=1.5]    (281,93) -- (390.5,93) ;
\draw [shift={(394.5,93)}, rotate = 180] [fill={rgb, 255:red, 0; green, 0; blue, 0 }  ][line width=0.08]  [draw opacity=0] (11.61,-5.58) -- (0,0) -- (11.61,5.58) -- cycle    ;
\draw  [fill={rgb, 255:red, 255; green, 255; blue, 255 }  ,fill opacity=1 ][line width=1.5]  (427,44.5) -- (621,44.5) -- (621,141.5) -- (427,141.5) -- cycle ;
\draw [line width=1.5]    (668,93) -- (777.5,93) ;
\draw [shift={(781.5,93)}, rotate = 180] [fill={rgb, 255:red, 0; green, 0; blue, 0 }  ][line width=0.08]  [draw opacity=0] (11.61,-5.58) -- (0,0) -- (11.61,5.58) -- cycle    ;
\draw  [fill={rgb, 255:red, 255; green, 255; blue, 255 }  ,fill opacity=1 ][line width=1.5]  (806,73) -- (1000,73) -- (1000,113) -- (806,113) -- cycle ;
\draw   (805.75,122.5) .. controls (805.75,127.17) and (808.08,129.5) .. (812.75,129.5) -- (893.25,129.5) .. controls (899.92,129.5) and (903.25,131.83) .. (903.25,136.5) .. controls (903.25,131.83) and (906.58,129.5) .. (913.25,129.5)(910.25,129.5) -- (993.75,129.5) .. controls (998.42,129.5) and (1000.75,127.17) .. (1000.75,122.5) ;

\draw (198.5,115.5) node  [font=\Large] [align=left] {$\displaystyle |\mathcal{V}_{g} |\times C$};
\draw (87,60) node  [font=\Large] [align=left] {$\displaystyle Layer\ 1$};
\draw (104,157) node  [font=\Large] [align=left] {$\displaystyle Layer\ 2$};
\draw (336,112) node  [font=\Large] [align=left] {$\displaystyle concatenate$};
\draw (524,93) node  [font=\Large] [align=left] {$\displaystyle |\mathcal{V}_{g} |\times 2C$};
\draw (723,112) node  [font=\Large] [align=left] {$\displaystyle aggregate$};
\draw (903,93) node  [font=\Large] [align=left] {$\displaystyle 1\times 2C$};
\draw (903.5,159) node  [font=\LARGE] [align=left] {$\displaystyle \boldsymbol{h}_{g}$};

\end{tikzpicture}}}
\caption{Example of a graph embedding construction for a 2-layer \CGMM{}. Each layer outputs a representation of size $C$ for each vertex $u \in \Vset{g}$. After a concatenation step, vertex representations (of size $2C$) are aggregated into a graph embedding. The choice of the aggregation function, \eg sum or mean, influences the final result.}
\label{fig:embedding-construction}
\end{center}
\end{figure}
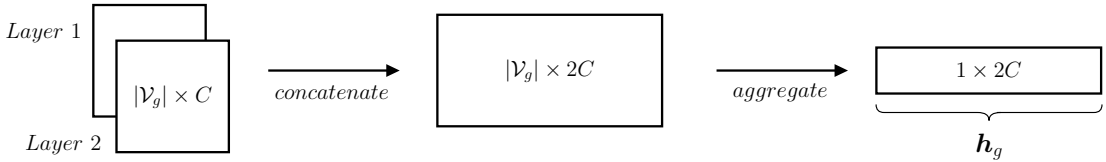

After training \CGMM{} for $L$ layers, we can finally build a graph representation. Figure \ref{fig:embedding-construction} schematizes the two-step process for $L=2$. First of all, inferred vertex representations from all layers are concatenated into $|\mathcal{V}_g|$ vectors of size $C\times L$; concatenation is the most conservative choice when one wants to prevent loss of information. After that, these vectors are aggregated to obtain the final graph fingerprint, and we treat the choice of the permutation invariant function as a hyper-parameter. Indeed, it is reasonable to assume that the best choice is task-dependent; for instance, the \textit{mean} aggregation abstracts from the size of a graph and focuses on variations in vertex distributions, whereas the \textit{sum} encodes the most prevalent features in the vertex embedding space. To simplify our analysis, save computational resources, and generate task-agnostic graph representations, we do not incorporate \textit{adaptive} aggregation functions in the process.
\subsection{Trade-offs of Vertex Representations}
\label{subsec:trade-offs-vertex-representations}
The $C-sized$ representation of a vertex at layer $\lcurr$ is also called a \textbf{unigram}. While this is the most straightforward way to obtain a vertex embedding, we can still build \textit{structure-aware} representations called \textbf{bigrams}. A \textit{bigram} is a $C^2$-sized vector which reflects how neighbors of a vertex are distributed. Formally, the bigram $\Phi(u)$ \cite{bacciu_generative_2018} of a vertex $u$ is defined as
\begin{align*}
\Phi_{i_j}(u) = \sum_{\substack{v \in \mathcal{N}(u)}} q_u(i) q_v(j), \ \ \ i,j \in {1,\dots,C}. \nonumber
\end{align*}
To increase the richness of vertex and graph representations, whenever a bigram is used we concatenate it with its corresponding unigram, thus obtaining a \textbf{unibigram}. Here we have another trade-off to consider: unigrams are clearly less expensive to compute and store, but unibigrams carry more information.

\subsection{Complexity and Scalability}
Thanks to \CGMM{}'s architectural flexibility, the cost of training each layer ranges from constant (\eg $\mathbb{L}(\lcurr)=\{\lprev\}$) to depth-specific (\eg $\mathbb{L}(\lcurr)=\{1,\dots,\lprev\}$). Time and space complexity of a training epoch on a single graph are bounded by the cost of computing the E-step, which is $\mathcal{O}(|\Vset{g}|(|\mathbb{L}(\lcurr)|C^2 + KC) )$, where $K$ stands for the number of vertex features.
Instead, computing the statistics after training one layer has time complexity $\mathcal{O}(|\Eset{g}|)$ because we just need to access the structure. The overall computation is therefore bounded by the sum of these two asymptotic terms that can be written as $\mathcal{O}(|\mathcal{V}_{\dataset{}}| + |\mathcal{E}_{\dataset{}}|)$.

Similarly to DNGNs, the \textit{i.i.d} assumption on vertices allows to easily implement mini-batch training, with which we can arbitrarily reduce the memory fingerprint at run-time; this is especially important in hardware-constrained scenarios. Also, data parallelism can be trivially achieved by distributing the epoch's mini-batches on different CPUs or clusters of machines. For these reasons, \CGMM{} is a suitable candidate to handle large-scale graph learning.\footnote{\url{https://github.com/diningphil/CGMM}.}

\subsection{Limitations}
\label{subsec:cgmm-limitations}
Due to the parametrized mean aggregation of neighboring observables, care must be taken when discriminating between structures with different connectivity but \textbf{same local distributions}. Indeed, when the distributions of the neighbourhood's states of two vertices are identical, \CGMM{} cannot differentiate between them regardless of their connectivity. 
We can mitigate this issue by embedding the notion of vertex degree into the neighborhood aggregation mechanism. To do so, we consider $deg_{max}(g)$, \ie the maximum degree of a graph $g$, and we intuitively connect each vertex $u$ to $deg_{max}(g)-deg(u)$ dummy neighbors in a special hidden state $\perp$ called \textit{bottom}. To maximize flexibility via the SP distributions, such dummy neighbors are connected to $u$ with a \emph{dedicated edge type}. Practically speaking, this means the statistics $\boldQ{\N{u}{}}{\mathbb{L}(\lcurr)}$ will contain information about vertex $u$'s degree, and such information is well-separated from the contextual information thanks to the use of the $\perp$ hidden state. Finally, note that we can also encode each vertex $u$'s degree into $x_u$ and use a Gaussian emission distribution to model its generation.

There are classes of graphs that cannot be distinguished so easily by \CGMM{}, such as \textit{k-regular graphs}, that is those such that $deg(u)=k $ $\forall u \in g$. In particular, this is the family of structures that can be discriminated by the $k$-dim WL isomorphism test. Recently, \citet{xu_how_2019} showed that almost all DNGNs are at most as powerful as the 1-dim WL test, but a similar proof for CGMM will be the subject of future works.

\subsection*{Summary}
\begin{figure}[h]
\begin{center}
\centerline{\resizebox{1\textwidth}{!}{\input{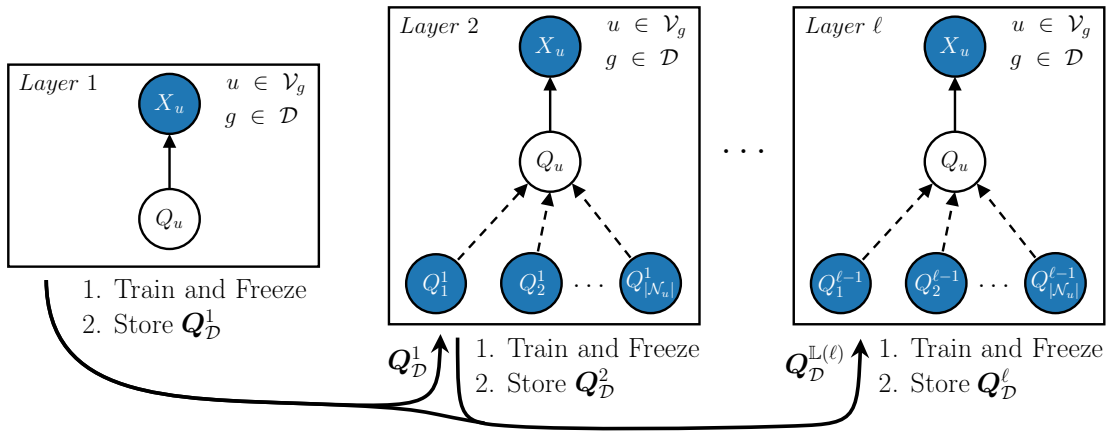}}}
\caption{High-level description of \CGMM{}'s incremental construction, in light of the description given by Algorithm \ref{tab:cgmm-pseudocode}. Each layer is trained individually before being frozen; after that, statistics are computed and passed to the subsequent layers. The process can be repeated as many times as desired.}
\label{fig:cgmm-incremental-construction}
\end{center}
\end{figure}

To summarize what we said so far, we detail the pseudo-code of the incremental training procedure in Algorithm \ref{tab:cgmm-pseudocode}, up to the construction of graph representations. In addition, we visually sketch \CGMM{}'s incremental construction in Figure \ref{fig:cgmm-incremental-construction}.
\begin{table}[ht]
    \footnotesize
    \begin{minipage}{\textwidth}
    \begin{algorithm}[H]
        \centering
        \caption{Probabilistic Incremental Training}
        \label{tab:cgmm-pseudocode}
        \begin{algorithmic}[1]
              \State Input: dataset $\dataset{}$, maximum number of layers $\lcurr_{max}$ and epochs epoch$_{max}$.
              \State Output: dataset of vertex and graph representations
              \For{$\lcurr \gets 1$,$\ldots$, $\lcurr_{max}$}
                \State Initialize layer $\lcurr$ according to $\mathbb{L}(\lcurr),|\Aset{}|$ and $C$
                \State Load $\boldQ{\dataset{}}{\mathbb{L}(\lcurr)} = \{\boldQ{\N{u}{a}}{\lcurr',a} \mid \lcurr' \in \mathbb{L}(\lcurr), a \in |\Aset{g}|, u \in g, g \in \dataset{}\}$
                \For{epoch $\gets 1$,$\ldots$, epoch$_{max}$}
                    \State $\Delta_{likelihood}$, \emph{posteriors} $\leftarrow$ E-Step($\dataset{}, \boldQ{\dataset{}}{\mathbb{L}(\lcurr)}$)
                    \State M-Step(\emph{posteriors})
                    \If{$\Delta_{likelihood} <$ threshold}
                        \State \textbf{break;}
                    \EndIf
                \EndFor
                \State $\boldQ{\dataset{}}{\lcurr} \leftarrow$ Inference($\dataset{}$)
                \State Store $\boldQ{\dataset{}}{\lcurr}$
              \EndFor
              \State $R_{\mathcal{V}_\dataset{}} \leftarrow$ concatenate($\{\boldQ{\dataset{}}{\lcurr'},\dots,\boldQ{\dataset{}}{\lcurr_{max}}\}$)
              \State  $R_{\dataset{}} \leftarrow$ aggregate($R_{\mathcal{V}_\dataset{}}$)
              \State  \textbf{return}  $R_{\mathcal{V}_\dataset{}}, R_{\dataset{}}$
        \end{algorithmic}
    \end{algorithm}
    \end{minipage}
\end{table}

\subsection{Experimental Setting}
\label{subsec:cgmm-exp-setting}
This section reports our experimental findings on some of the common graph classification benchmarks listed in Section \ref{subsec:iclr-experimental-setting} as well as a vertex classification task. To go more in depth, we study the effect of layering on performances, carry out ablation studies, and graphically show an example of context propagation across layers.

\subsubsection*{Datasets}
As regards graph classification, we compare \CGMM{} against PROTEINS, DD, NCI1, IMDB-BINARY, IMDB-MULTI, REDDIT-BINARY, REDDIT-5K, and COLLAB from Table \ref{tab:datasets-table-iclr}: we leave ENZYMES out of the evaluation due its very small size and nature of the features. The absence of vertex features in the social datasets constitutes a degenerate case that prevents \CGMM{} from learning, \ie the neighboring states will always have the same distributions. As done in \cite{niepert_learning_2016,errica_fair_2020}, when a vertex has no features, we add its degree as a continuous value as well as the \textit{bottom} states discussed in Section \ref{subsec:cgmm-limitations}. \\
In addition, to test \CGMM{} performance on vertex classification tasks, we use the protein-protein interaction data set (PPI) introduced in \cite{hamilton_inductive_2017}. We have already highlighted the troubling trends regarding some of the most common vertex classification benchmarks \cite{shchur_pitfalls_2018}, so we prefer to stick to a dataset which is large and has a well-defined evaluation protocol. In this task, we are given a set of distinct large graphs, and our goal is to classify their vertices. Table \ref{tab:ppi-details} provides some details about the PPI benchmark.

\begin{table}[ht]
\centering
\begin{tabular}{l c c c c c} \hline
 & \# Graphs & Classes & \# Vertices   & \# Edges    & \# Vertex feat. \\ \hline
\textbf{PPI}        & 24     & 121     & 2372.67  & 34113.17      & 50       \\ \hline
\end{tabular}
\caption{Dataset statistics for the PPI vertex classification benchmark. In this task, we simultaneously have to predict 121 binary labels.}
\label{tab:ppi-details}
\vskip -0.1in
\end{table}

\subsubsection*{Hyper-parameters and Evaluation Protocol}
This section describes the hyper-parameters tested as well as the chosen model assessment and selection procedures. However, for some of the baselines considered, there is no standardized evaluation protocol to follow such as the one of Section \ref{sec:scholarship-issues}. For this reason, in order to compare with kernel methods and other DNGNs, we will report two experimental setups. The first uses random data splits, whereas the second follows \cite{errica_fair_2020} and provides a more reliable performance comparison. Both setups rely on grid searches. \\
In Table \ref{tab:cgmm-hyperparams}, we list the hyper-parameters needed by \CGMM{} and by the subsequent supervised classifier working on the unsupervised vertex/graph representations. We achieved constant per-layer complexity by setting $|\mathbb{L}(\lcurr)|=\{\lprev\}$, but we also evaluated the impact of considering all previous layers using the SP technique. The number of EM epochs is fixed to $10$, because the likelihood always stabilized around that value in our preliminary experiments. We also tried both \textit{continuous} and \textit{one-hot} vertex representations, and the global aggregation was either \textit{sum} or \textit{mean}. For some datasets, the preliminary experiments revealed that one global aggregation was dramatically better than the other, so we exploited this fact to reduce the number of combinations to test.  \\
The different configurations of a \CGMM{}'s layer mainly depend on \textit{two} hyper-parameters only: the number of hidden states $C$ and the number of layers. We can thus exploit the incremental nature of our model to further reduce the dimension of the grid search space: for each value $C$, we trained a single network for a maximum of $L$ layers and then \quotes{cut} it to obtain the necessary configurations of depth $L'<L$. In addition, once the embeddings of all possible \CGMM{}'s configurations have been computed and stored, we can explore many combinations for the classifier without having to re-train \CGMM{} every time. \\
Speaking of classifiers, we considered a logistic regressor and a more flexible alternative in the form of a one-layer MLP with ReLU activations. Both of them are trained with Adam \cite{kingma_adam_2015} and Cross-Entropy loss (Mean Squared Error for PPI). To contrast overfitting, we introduced L2 regularization and an early stopping technique.

\begin{sidewaystable}[ht]
\centering
\begin{tabular}{l|cccccc} \hline
     & \textbf{D\&D} & \textbf{NCI1} & \textbf{PROTEINS} & \textbf{IMDB-*} & \textbf{COLLAB} & \textbf{PPI} \\\hline
C    & \{5,10,20\} &  \{5,10,20\} &  \{5,10,20\} &  \{5,10,20\} &  \{5,10,20\} & \{5,10,20\}\\
$\mathbb{L}(\lcurr)$  &  $\{\lprev\}$ & $\{\lprev\}$ & $\{\lprev\}$  & $\{\lprev\}$  & $\{\lprev\}$ & $\{\lprev\}$\\
\# layers  & \{5,10,15,20\} & \{10,20\} & \{5,10,15,20\} & \{5,10,15,20\} & \{5,10,15,20\} & \{5,10,20\} \\
EM epochs  & 10  & 10  &  10   &  10  &  10 & 10\\
One-Hot/Continuous Vertex Repr.  &  both  &  both   &           both  &  both   &  both & both \\
Unigram/Unibigram  &  both  &  both   &      both   &  both   &  both & unigram \\
Sum/Mean Global Aggr.  &  sum  &  both   &      sum   &  both   &  mean & - \\
Classifier  &  \{mlp\}  &  \{mlp\}   &      \{mlp\}     &  \{logistic, mlp\}   &  \{logistic, mlp\} & \{mlp\} \\
\# Hidden Units &  \{8,16,32,128\}  &  \{32,128\}   &      \{8,16,32,128\}   &  \{8,32,128\}   &  \{32,64,128\} & \{128,256,512\} \\
Learning Rate &  \{1e-3,1e-4\}  &  \{1e-3\}   &      \{1e-3,1e-4\}  &  \{1e-3,1e-4\}   &  \{1e-3,1e-4\} & \{1e-2,1e-3\} \\
L2 Weight Decay &  \{1e-2,5e-2,5e-3\}  &  \{1e-3,5e-4\}   &      \{1e-2,5e-2,5e-3\}   &  \{1e-2,1e-3\}   &  \{1e-3,5e-3,5e-4\} & \{0.,1e-5\} \\
Classifier Epochs  &  5000  &  2000   &      5000   &     5000  &  5000 & 5000 \\
Early-Stopping &  1000  &  100   &      1000     & 1000  &  1000 &  500 \\
Batch Size &  100  &  200   &      100     & 100  &  100 & 20 \\ \hline
\end{tabular}
\caption{Hyper-parameters tried during model selection. IMDB-* refers to both binary and multi-class tasks. The number of hidden units per layer are ignored when using logistic regression.}
\label{tab:cgmm-hyperparams}
\end{sidewaystable}

\paragraph*{Different setups across models.}
As anticipated, in the first evaluation setup we created random (stratified) data splits for the graph classification tasks. We followed a \textbf{Double Cross-Validation} strategy, with 10 external folds for risk assessment and 5 internal folds for model selection. All the other methodologies are evaluated according to a 10-fold CV strategy for risk assessment, so overall we expect the results to be roughly comparable. For an in-depth analysis of these baselines, the reader is referred to \cite{bacciu_probabilistic_2020}.
Regarding early stopping, we used the Generalization Loss \cite{prechelt_early_1998} with $\alpha=5$, which is considered to be a good compromise between training time and performances. In this respect, Table \ref{tab:cgmm-hyperparams} reports the number of epochs after which early stopping starts; at the beginning of training, the validation loss smoothly oscillated and accuracy did not steadily increase, we believe stopping too early would have not been beneficial to get reliable performance estimates. On the contrary, the training/validation/test data splits of PPI were given, so we chose a simpler holdout approach.

\paragraph*{Same setup across models.}
Here, we took advantage of the evaluation protocol of Section \ref{sec:scholarship-issues} to robustly re-evaluate \CGMM{} against the most popular DNGNs. The only difference in the hyper-parameters regards the early stopping technique: for simplicity, we chose to use a simple patience-based stopping criterion, whose values correspond to those in Table \ref{tab:cgmm-hyperparams}. Instead, we experimented with the same hyper-parameters of COLLAB on both REDDIT datasets.

\clearpage
\subsection{Results}
We now present \CGMM{}'s results on graph and vertex classification, along with empirical studies on the beneficial effects of depth. These will provide us with additional hints on \CGMM{}'s ability to extract useful information in an unsupervised fashion. We also study the impact of the layer-wise SP variable as part of our ablation studies. Finally, we visualize how the model propagates contextual information across the graph.

\subsubsection*{Graph Classification}
\label{sub:graph-classification}
We evaluate the performance of our method against different kernels and deep learning techniques for graph classification. Table \ref{tab:kernel-costs} provides a comparison between the kernels considered and \CGMM{} in terms of computational costs. As we can see, some kernels can be inadequate when it comes to large scale training and inference because of their (at least) quadratic time complexity in the number of graphs. Moreover, the considered kernels for graphs are not applicable to continuous vertex features, which limits their applicability to different domains.
\begin{table}[ht]
\centering
\begin{tabular}{l|cc} \hline
\textbf{Kernel}  & \textbf{Cost} & \textbf{Reference} \\\hline
\textsc{GK}   & $\mathcal{O}(|\dataset{}|^2 n d^{k-1})$   &  \cite{shervashidze_efficient_2009}  \\
\textsc{RW}   & $\mathcal{O}(|\dataset{}|^2 n^3)$   &  \cite{vishwanathan_graph_2010}   \\
\textsc{PK}   & $\mathcal{O}(m(h-1) + h|\dataset{}|^2 n)$   &  \cite{neumann_efficient_2012}  \\
\textsc{WL}   & $\mathcal{O}(|\dataset{}|hm+|\dataset{}|^2hn)$   &  \cite{shervashidze_weisfeiler-lehman_2011} \\ \hline
\CGMM{} & $\mathcal{O}(L(|\dataset{}|n + |\dataset{}|m))$ & \\ \hline
\end{tabular}
\caption{Computational costs of graph kernels compared to \CGMM{}. We assume that all graphs have size $n=|\mathcal{V}_g|$, $m=|\mathcal{E}_g|$ edges and maximum degree $d$. Moreover, $k$ is the size of the graphlets (\ie subgraphs) counted by GK, and $h$ is the number of iterations needed by different procedures to compute the final similarity scores.}
\label{tab:kernel-costs}
\end{table}

\paragraph*{Different setups across models.}
Results for graph classification, under the first of the two evaluation protocols considered, are shown in Table \ref{tab:results-table}. \CGMM{} performs well in all data sets (scoring top-3 on five of them), even though the probabilistic architecture was not trained to solve a classification task. In particular, we achieve competing results on all three collaborative data sets, and we improve the best result on NCI1. This suggests that learning the distribution of a vertex's neighbourhood at different abstraction's levels produces a rich unsupervised graph representation. As a matter of fact, in 9 out of 10 external folds on NCI1, the model selection procedure chose a configuration with 20 layers; in contrast, the DNGNs of Table \ref{tab:results-table} exploit a maximum of 4 graph convolutions. The results also highlight that \CGMM{} can perform well even when the only source of information is structural, \ie the degree of a vertex.

\begin{table}[ht]
\scriptsize
\centering
\setlength{\tabcolsep}{4.5pt}
\begin{tabular}{lcccccc}  \hline
     & \textbf{D\&D} & \textbf{NCI1} & \textbf{PROTEINS} & \textbf{IMDB-B} & \textbf{IMDB-M} & \textbf{COLLAB} \\\hline
\textsc{GK} \cite{shervashidze_efficient_2009}   & $74.38\pm0.7$ & $62.49\pm0.3$ & $71.39\pm0.3$ & -    & -    & -    \\
\textsc{RW} \cite{vishwanathan_graph_2010}   & $>3$ days  & $>3$ days  & $59.57\pm0.2$ & -    & -    & -    \\
\textsc{PK} \cite{neumann_efficient_2012}   & $78.25\pm0.5$ & $82.54\pm0.5$ & $73.68\pm0.7$ & -    & -    & -    \\
\textsc{WL} \cite{shervashidze_weisfeiler-lehman_2011}   & $78.34\pm0.6$ & $\mathbf{84.46}\pm0.5$ & $74.68\pm0.5$ & -    & -    & -    \\ \hline
\textsc{ARMA} \cite{bianchi_graph_2021} & $74.86$ & - & $75.12$ & -    & -    & -    \\
\textsc{PSCN} \cite{niepert_learning_2016}  & $76.27\pm2.6$ & $76.34\pm1.7$ & $75.00\pm2.5$ & $71.00\pm2.3$ & $45.23\pm2.8$ & $72.60\pm2.15$ \\
\textsc{DCNN} \cite{atwood_diffusion-convolutional_2016}  & $58.09\pm0.5$ & $56.61\pm1.0$ & $61.29\pm1.6$ & $49.06\pm1.4$ & $33.49\pm1.4$ & $52.11\pm0.7$ \\
\textsc{ECC} \cite{simonovsky_dynamic_2017}  &  $72.54$  & $76.82$  &  -    & - & - & - \\
\textsc{DGK} \cite{yanardag_deep_2015}  &  -  & $62.48\pm0.3$ & $71.68\pm0.5$ & $66.96\pm0.6$ & $44.55\pm0.5$ & $73.09\pm0.3$ \\
\textsc{DGCNN} \cite{zhang_end--end_2018}  & $\mathbf{79.37}\pm0.9$ & $74.44\pm0.5$ & $75.54\pm0.94$ & $70.03\pm0.86$ & $47.83\pm0.9$ & $73.76\pm0.5$ \\
\textsc{PGC-DGCNN} \cite{tran_filter_2018} & $78.93\pm0.9$ & $76.13\pm0.7$ & $\mathbf{76.45}\pm1.02$ & $71.62\pm1.2$ & $47.25\pm1.4$ & $75.00\pm0.58$ \\ \hline
\CGMM{}-nb   & $77.35\pm1.6$ & $77.02\pm1.8$ & $75.11\pm2.8$ & $71.07\pm3.5$ & $47.36\pm3.4$ & $73.3\pm2.9$ \\
\CGMM{}-full   & $77.20\pm3.1$ & $76.94\pm1.6$ & $75.45\pm4.4$ & $\mathbf{72.30}\pm3.5$ & $49.42\pm3.6$ & $\mathbf{76.06}\pm2.4$ \\
\CGMM{}   & $77.15\pm3.5$ & $\mathbf{77.80}\pm1.9$ & $75.56\pm3.0$ & $72.1\pm2.3$ & $\mathbf{49.73}\pm1.6$ & $75.50\pm2.74$ \\ \hline
\end{tabular}
\caption{\CGMM{}'s results of a 10-Fold Double Cross Validation for graph classification. Best results are reported in bold. We report \CGMM{}'s accuracy on NCI1 in bold because it performs better than the other neural models. \CGMM{}-nb indicates that the model is not using bigram features, whereas \CGMM{}-full represents the extended \CGMM{} where each layer exploits all previous layers.}
\label{tab:results-table}
\end{table}

Note that kernels can process and compare graphs more explicitly than DGNs: one of the reasons why the WL kernel has higher accuracy on NCI1 may be due to the kind of structural patterns used to compute the similarity score. Still, when the number and size of the graphs to consider increases, using these kernels becomes challenging.

\paragraph*{Same setup across models.}
If re-evaluated under the rigorous setup of Section \ref{sec:scholarship-issues}, we can appreciate how \CGMM{} is still competitive against the new pool of models, with special mention for the social tasks. The structure agnostic baseline, instead, still leads on D\&D, PROTEINS, and IMDB-MULTI.

In addition, please note how large the performance gap can be on some datasets \wrt{} the former evaluation protocol, with approximately 3 points less on D\&D and PROTEINS for \DGCNN{} and even \CGMM{} (though standard deviation are relatively high). This is further confirmation that to clearly evaluate progress one should at least keep the experimental setting identical for all methods.
\begin{table}[ht]
\centering
\begin{tabular}{l c c c c}
\toprule
     & \textbf{D\&D} & \textbf{NCI1} & \textbf{PROTEINS}\\
\midrule
 \Baseline & $\mathbf{78.4}\pm 4.5 $ &  $69.8 \pm 2.2 $ &  $\mathbf{75.8} \pm 3.7 $ \\
 \DGCNN & $76.6 \pm 4.3 $ &  $76.4 \pm 1.7 $ &  $72.9 \pm 3.5 $   \\
 \DiffPool & $75.0 \pm 3.5 $ &  $76.9 \pm 1.9 $ &  $73.7 \pm 3.5 $  \\
 \ECC & $72.6 \pm 4.1 $ &  $76.2 \pm 1.4 $ &  $72.3 \pm 3.4 $  \\
 \GIN & $75.3 \pm 2.9 $ &  $\mathbf{80.0} \pm 1.4 $ &  $73.3 \pm 4.0 $    \\
 \GraphSAGE & $72.9 \pm 2.0 $ &  $76.0 \pm 1.8 $ &  $73.0 \pm 4.5 $  \\ \hline
 \CGMM{}  & $74.9 \pm 3.4 $ & $ 76.2 \pm2.0$ & $74.0 \pm 3.9$ \\
\bottomrule
\end{tabular}
\caption{Mean and standard deviation results on chemical datasets of a 10-fold Cross Validation (setup of Section \ref{sec:scholarship-issues}). Best results are reported in bold.}
\label{tab:cgmm-iclr-chemical-results}
\end{table}
\begin{table}[ht]
\small
\centering
\begin{tabular}{lccccc}
\toprule
     & \textbf{IMDB-B} & \textbf{IMDB-M} & \textbf{REDDIT-B} & \textbf{REDDIT-5K} & \textbf{COLLAB}\\
    \midrule
 \Baseline & $70.8 \pm 5.0 $ &  $\mathbf{49.1} \pm 3.5 $ &  $82.2 \pm 3.0 $ &  $52.2 \pm 1.5 $ &  $70.2 \pm 1.5 $   \\
 \DGCNN & $69.2 \pm 3.0 $ &  $45.6 \pm 3.4 $ &  $87.8 \pm 2.5 $ &  $49.2 \pm 1.2 $ &  $71.2 \pm 1.9 $   \\
 \DiffPool & $68.4 \pm 3.3 $ &  $45.6 \pm 3.4 $ &  $89.1 \pm 1.6 $ &  $53.8 \pm 1.4 $ &  $68.9 \pm 2.0 $   \\
 \ECC & $67.7 \pm 2.8 $ &  $43.5 \pm 3.1 $ &   - &   - &   -   \\
 \GIN & $71.2 \pm 3.9 $ &  $48.5 \pm 3.3 $ &  $\mathbf{89.9} \pm 1.9 $ &  $\mathbf{56.1} \pm 1.7 $ &  $75.6 \pm 2.3 $   \\
 \GraphSAGE & $68.8 \pm 4.5 $ &  $47.6 \pm 3.5 $ &  $84.3 \pm 1.9 $ &  $50.0 \pm 1.3 $ &  $73.9 \pm 1.7 $   \\
 \midrule
 \CGMM{}  & $\mathbf{72.7} \pm 3.6$ & $47.5 \pm 3.9$ & $88.1 \pm 1.9$ & $52.4 \pm 2.2$ & $\mathbf{77.32} \pm 2.2$    \\ \bottomrule
\end{tabular}
\caption{Mean and standard deviation results on social datasets of a 10-fold Cross Validation (setup of Section \ref{sec:scholarship-issues}). Best results are reported in bold. Note that the degree is the sole vertex feature used by all models.}
\label{tab:cgmm-iclr-social-results}
\end{table}

Overall, \CGMM{} has proved to be a satisfactory unsupervised model, given that the richness of its graph embeddings allowed us to get very close to the state of the art.

\subsubsection*{Vertex Classification}
\label{sub:vertex-classification}
We now turn our attention to vertex classification, specifically on the PPI benchmark. Following the literature \cite{velickovic_deep_2019}, we compare \CGMM{} against \GraphSAGE{} and DGI, as well as a structure-agnostic baseline that applies logistic regression to the vertex features. GraphSAGE, DGI, and \CGMM{} share a first pre-training step, in which vertex embeddings are learned in an unsupervised fashion before feeding the learned vertex representations to a supervised classifier. Results are shown in Table \ref{tab:results-ppi-table}: we observe that \CGMM{} has very good performances, improving against all GraphSAGE variants but for DGI. Considering that DGI uses GraphSAGE as part of its framework, it seems that the learning procedure is what generates the gap between the two methods. In fact, while GraphSAGE relies on a link prediction loss and an entropy penalization term to learn vertex representations, DGI learns to discriminate vertices according to a contrastive noise procedure.

 \begin{table}[th]
 \centering
\begin{tabular}{l|cc} \hline
                & \textbf{Data Used} & \textbf{Micro F1} \\ \hline
\textsc{Baseline}        & $\Vset{g}$         & 42.2     \\
\GraphSAGE-GCN   & $\Vset{g}, \Eset{g}$      & 46.5     \\
\GraphSAGE-mean  & $\Vset{g}, \Eset{g}$      & 48.6     \\
\GraphSAGE-LSTM  & $\Vset{g}, \Eset{g}$      & 48.2     \\
\GraphSAGE-pool  & $\Vset{g}, \Eset{g}$      & 50.2     \\
\DGI             & $\Vset{g}, \Eset{g}$      & $\mathbf{63.8}$ \\ \hline \CGMM{}-full            & $\Vset{g}, \Eset{g}$      & 58.4    \\
\CGMM{}            & $\Vset{g}, \Eset{g}$      & $60.2$    \\ \hline
\end{tabular}
\caption{\CGMM{}'s results of inductive vertex classification on PPI. We report the Micro Average F1 score across the 121 target labels.}
\label{tab:results-ppi-table}
\end{table}

\subsubsection*{Hyper-parameters' Analysis}
\label{sub:ablation studies}
We enrich our empirical analysis with three further studies: the first concerns the impact of the unibigram technique on performances; the second inspects the potential performance advantages of using all previous layers when training a new layer; the last one investigates whether a wider model with fewer layers can perform as well as a deep model with fewer hidden states. With the exception of the second study, the other analyses will use the information coming from the previous layer, \ie $\mathbb{L}(\lcurr)=\{\lprev\}$.

\paragraph*{Unibigram Ablation}
In Table \ref{tab:results-table}, we re-evaluated the model on all data sets by constraining \CGMM{} to only use unigrams (\CGMM{}-nb). Results indicate a slight performance drop on chemical data sets and a larger decrease in social data sets. This suggests that, especially when we only have access to structural information (\ie the degree distribution), computing a graph representation that takes the structure into account can be helpful. Nevertheless, \CGMM{}-nb performances still remain good with respect to the state of the art.

\paragraph*{On the Impact of Previous Layers}
We repeated all experiments by conditioning each layer of the architecture on the entire subset of previous layers, \ie $\mathbb{L}(\lcurr)=\{1,\dots,\lprev\}$. This way, each layer is free to weigh the previous layers (thanks to the SP variable) to maximize the likelihood of each graph. Results (Table \ref{tab:results-table} and \ref{tab:results-ppi-table}, \CGMM{}-full) indicate that the model performs almost always on par \wrt{} the significantly more efficient version that does not use the SP variable $L_u$. Nonetheless, despite the negligible performance advantage obtained on these tasks, we recommend treating the use of $L_u$ as a hyper-parameter of the model when dealing with other vertex or graph classification tasks.

\begin{figure}[ht]
\centering
\includegraphics[width=0.9\columnwidth]{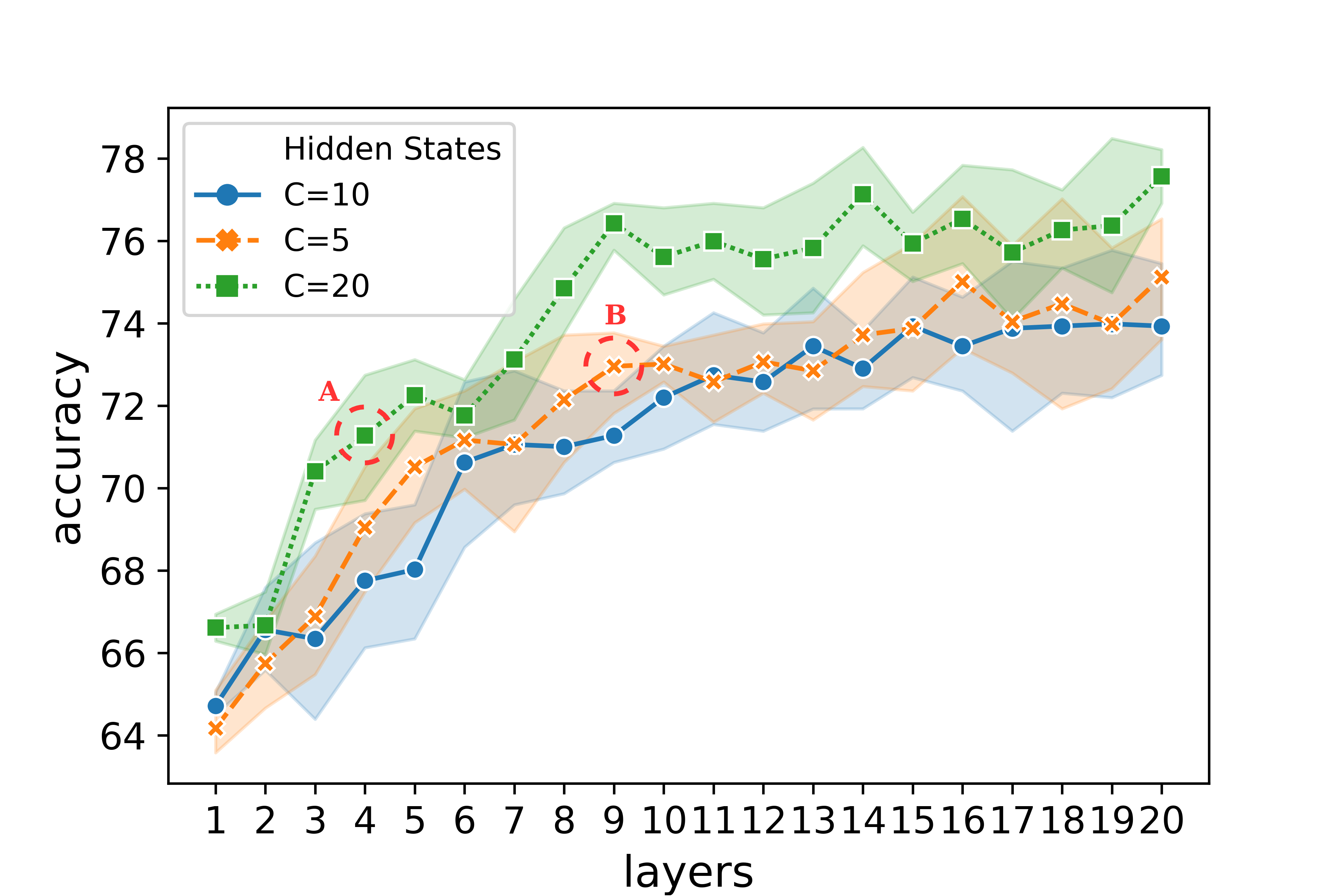}
\caption{This picture shows that using a large $C$ but few layers is not always enough to reach the same performance of a deeper network with a smaller $C$. Points $A$ and $B$ are associated with representations of dimensions $C=60$ and $C=45$, respectively. However, the latter has better validation performances if we consider a logistic regressor on NCI1, which means that the flow of context is indeed more beneficial than increasing the number of hidden states $C$. Results are averaged over five independent runs and standard deviations are shown as colored bands.}
\label{fig:states_vs_depth}
\end{figure}

\paragraph*{Sensitivity Analysis}
When designing any deep network (let it be neural or probabilistic), it is useful to analyze the relation between the dimension of each layer's hidden representation and the number of layers in terms of performance variations. For DGNs the depth of the architecture is functional to context spreading, so we would expect that having a larger hidden representation for each layer is not enough to compensate for the flow of information between vertices. We provide an example in Figure \ref{fig:states_vs_depth} to show how the validation accuracy of a logistic regressor on NCI1 varies while changing the number of hidden states $C$ and \CGMM{}'s depth. It can be seen that the graph representation associated with point $A$ (of size $C=60$) is not sufficient to achieve the same performance of the graph representation associated with point B ($C=45$). This means that depth is crucial to encode more information.

\begin{figure}[ht]
 \begin{subfigure}
     \centering
     \includegraphics[width=0.49\columnwidth]{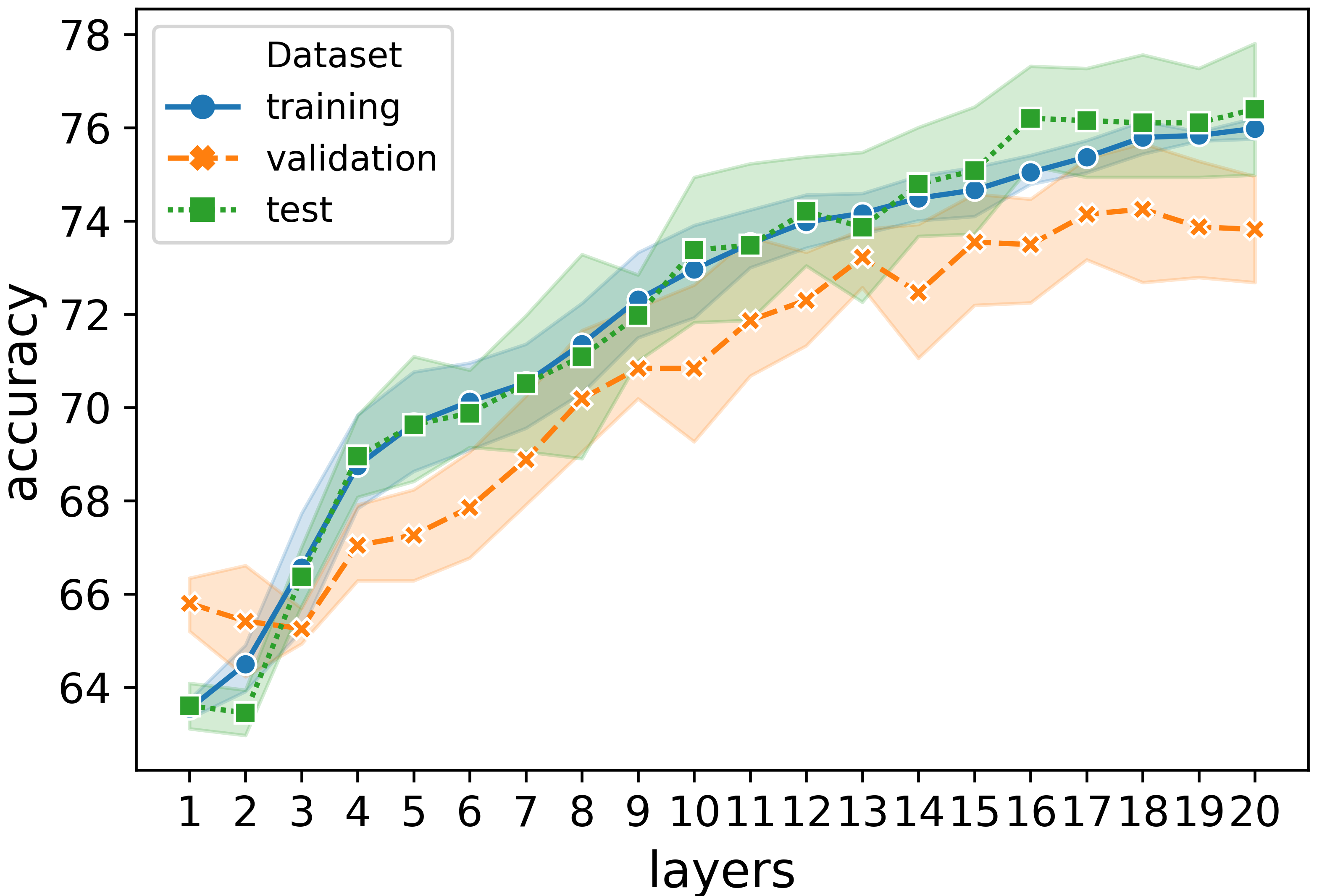}
 \end{subfigure}
 \begin{subfigure}
     \centering
     \includegraphics[width=0.49\columnwidth]{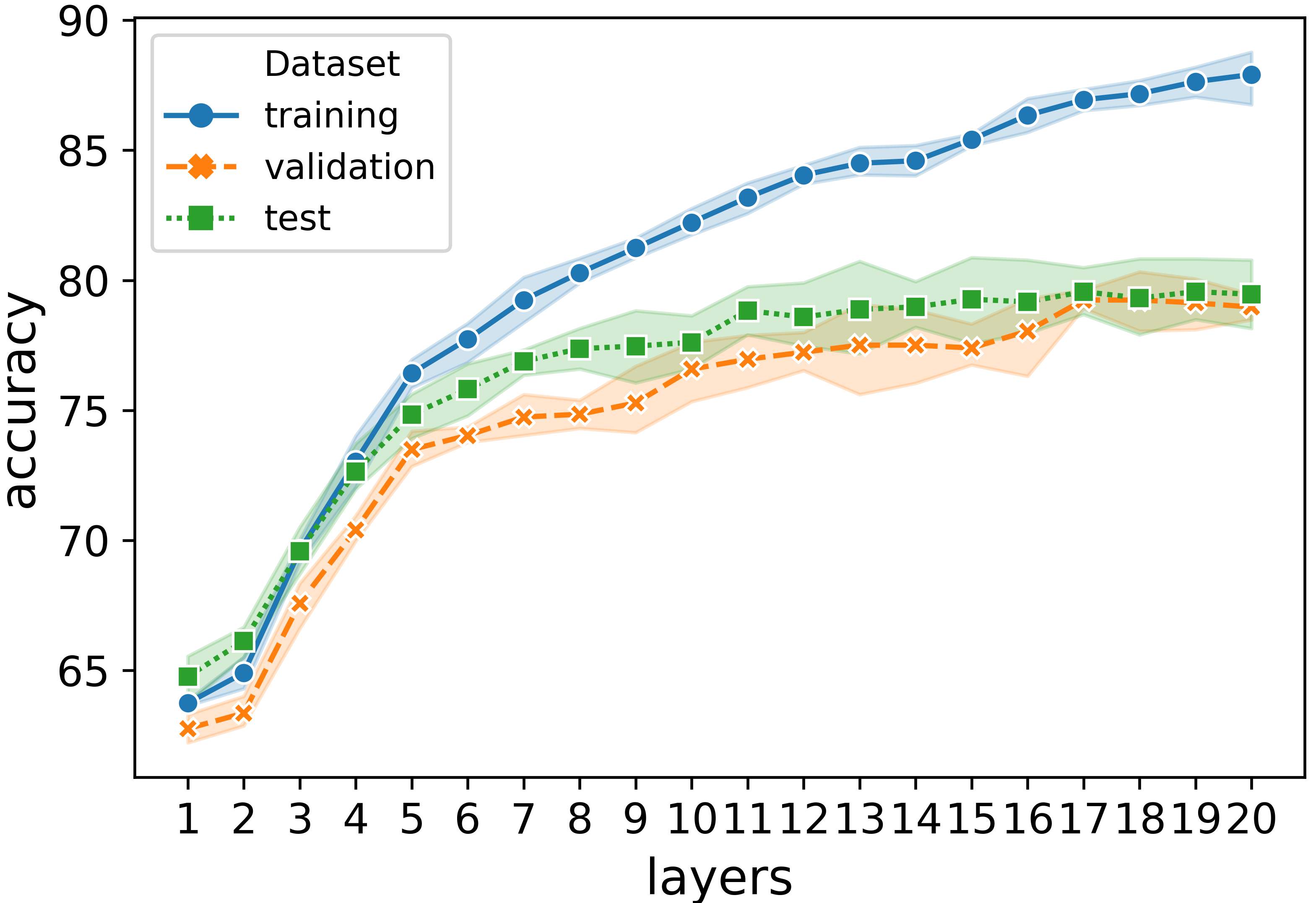}
 \end{subfigure}
 \begin{subfigure}
     \centering
     \includegraphics[width=0.49\columnwidth]{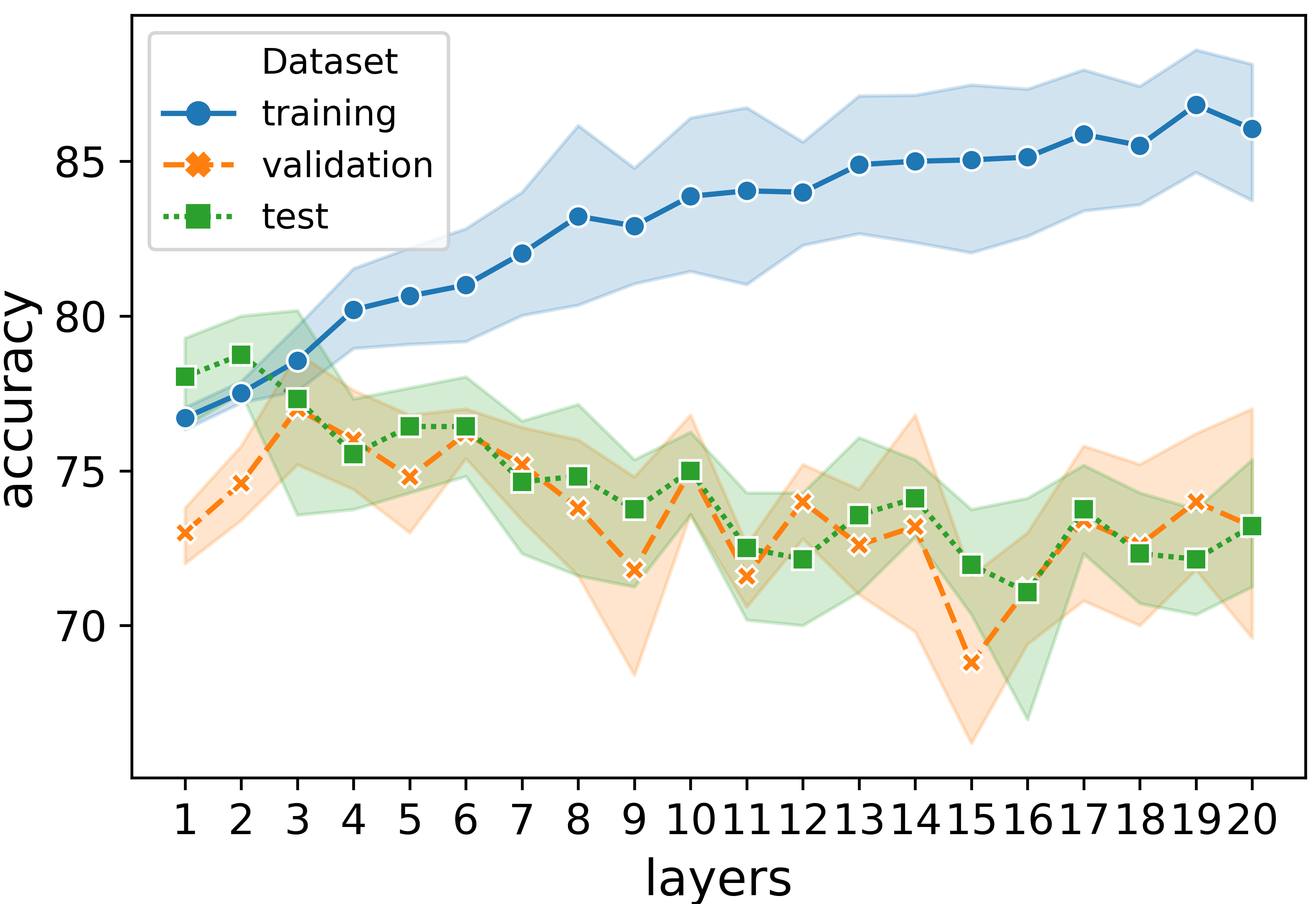}
 \end{subfigure}
 \begin{subfigure}
     \centering
     \includegraphics[width=0.49\columnwidth]{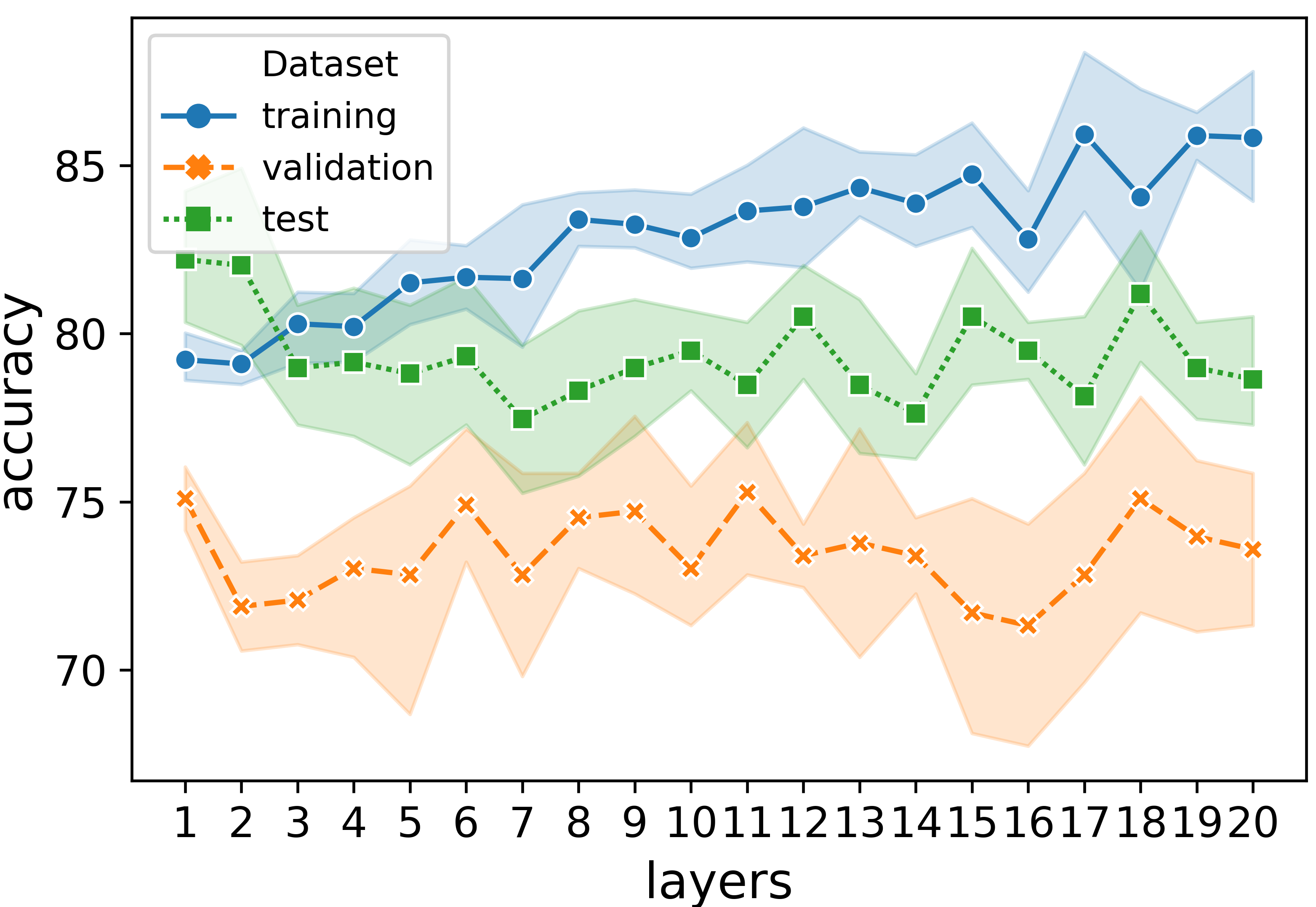}
 \end{subfigure}
 \caption{Stability experiments on NCI1 with logistic regressor (top left) and MLP (top right), PROTEINS with MLP (bottom left) and D\&D with MLP (bottom right), which show how accuracy varies from 1 to 20 layers. Results are averaged over five independent runs. Colored bands denote standard deviation. While the effect of depth is beneficial for NCI1, a different behavior emerge on PROTEINS and D\&D, which is not to be attributed to random splits.}%
 \label{fig:stability}%
\end{figure}

\paragraph*{On the Effects of Depth}
This Section is meant to answer to two further research questions. First, we want to quantify the effect of depth on the architecture when coupled with a classifier. The second point is about understanding how much the model's performance is affected by a random initialization of the layers.
To address both questions, we took a random train-validation-test split of NCI1 to conduct a new experiment. With its 4110 graphs, NCI1 was chosen to minimize the effect of the data split on results and consequently on the random initialization of the classifier. In contrast, the other chemical data sets seem more data split-dependent. We trained a $20$-layer \CGMM{} for some of the configurations defined in Table \ref{tab:cgmm-hyperparams}, and we repeated each process five times averaging the results. Figure \ref{fig:stability} reports the accuracy versus the number of layers for such configurations, with logistic regression or MLP classifiers. We see that, in both cases (top part of the figure), depth has a beneficial effect on test accuracy, with slightly worse results on test accuracy when using logistic regression due to its strong bias. Notice how validation and test curves tend to an asymptote after ten layers; this information may be used as stopping criterion when constructing the architecture for supervised tasks, as proposed in \cite{marquez_deep_2018} for convolutional networks on images.

One interesting thing to notice is that we do not necessarily incur in the curse of dimensionality as the size of the fingerprint grows larger and larger with the layers. This cold be ascribed to the fact that the fingerprint construction at layer $\lcurr$ is guided by layer $\lprev$, and this dependency generates a completely different learning problem at each layer. This may explain why we do not quickly overfit the training data after the $20$ layers.

In addition, since the training accuracy on NCI1 does not significantly vary between different runs (due to different weights' initialization), we get an indication that the pooling strategy of \cite{fahlman_cascade-correlation_1990}, used in our preliminary contribution \cite{bacciu_contextual_2018} brings a negligible advantage to \CGMM{}. This also holds for PROTEINS and D\&D when the architecture is very shallow (up to three layers), though results need to be taken with a pinch of salt because of the discussion of Section \ref{subsec:iclr-results}. As a matter of fact, it is still an open question whether a dataset is \quotes{too simple} or the graph convolutions devised so far are unable to extract the relevant features from the graphs.

\subsubsection*{Visualization: a Case Study}
\label{sub:qualitative}
\begin{figure}[ht]
\centering
\includegraphics[width=1\columnwidth]{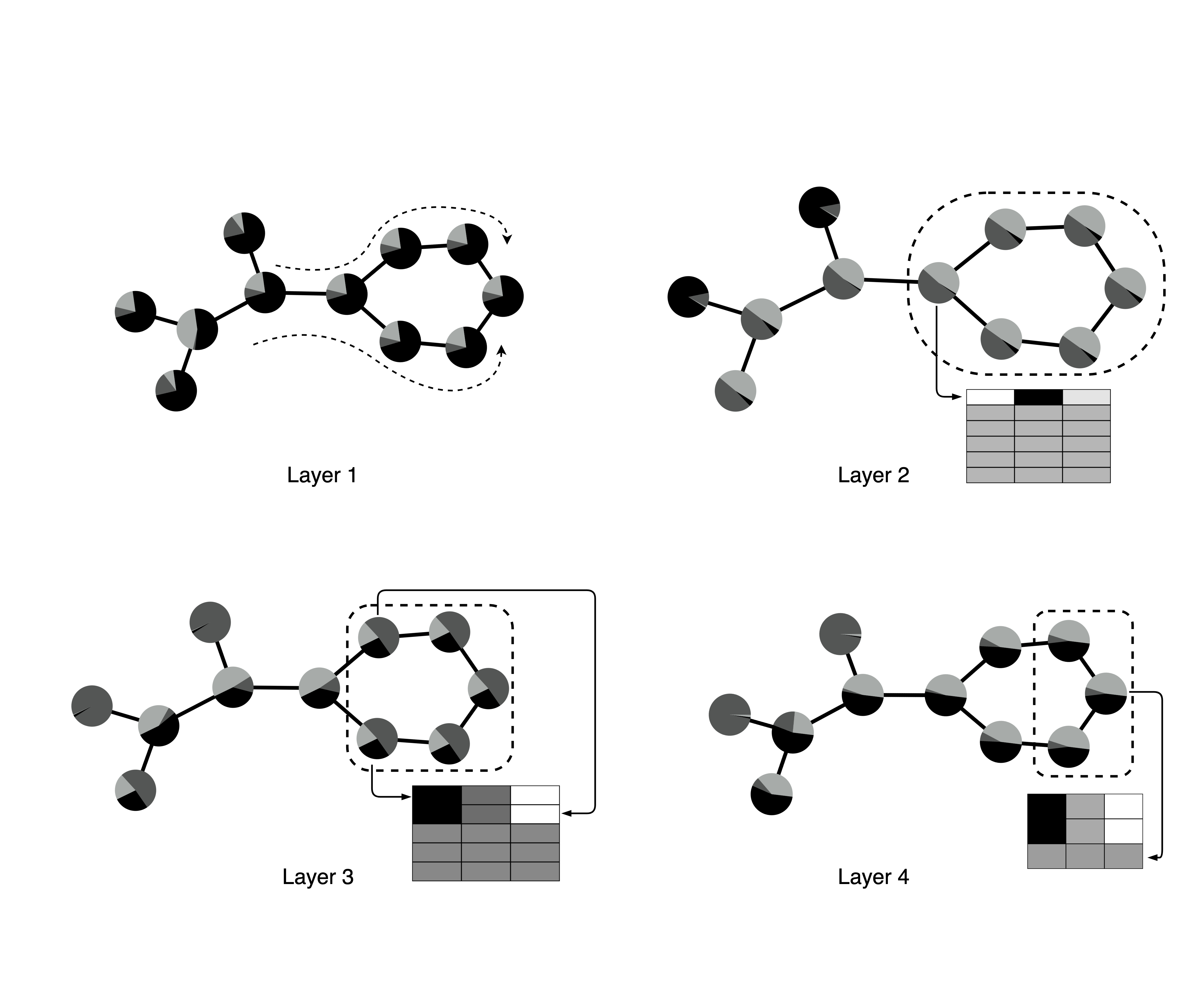}
\caption{Context flow on a real NCI1 graph. We focus on the highly regular rightmost subgraph, which is influenced by the left part of the structure. Posterior's heatmaps show that, although relatively small, posteriors of vertices change as expected.}
\label{fig:real-context-flow}
\end{figure}

This part provides a visual exploration and interpretation of \CGMM{}'s internal dynamics. This kind of analysis is meant to demonstrate how the model extracts different patterns at each layer of the architecture in a way that is consistent with what stated Chapter \ref{chapter:gentle-introduction}. Therefore, Figure \ref{fig:real-context-flow} sketches how information spreads in a real NCI1 molecule. We represent each vertex's posterior as a pie chart with $C$ different colours; in this experiment, we use a 4-layer \CGMM{} with $C=3$. Please keep in mind that colours assignment between different layers is irrelevant. The rightmost six atoms of the molecule have the same atomic symbol, so they will be assigned an identical state at layer 1. What is more, these six atoms alone form a 2-regular subgraph, which means that their state can only change if context flows from left to right, as shown by the dashed arrows on the top-left side of the figure. If it were not for the five leftmost vertices context could not flow, because all neighborhoods would look identical to the model. At each layer, we highlight the vertices of interest inside a dashed regions, and the associated heatmap of states (one vertex's posterior per row) proves that information flows as expected.

\subsection{Summary}
\label{subsec:cgmm-summary}
In this section we have introduced the core contribution of the thesis. The Contextual Graph Markov Model is a deep, unsupervised, fully probabilistic, and efficient way to compute vertex/graph embeddings. It brings together the principles of Deep Graph Networks and the probabilistic tools of Bayesian networks. In the experiments, we have shown that the model allows a common classifier to reach the state of the art on different classification tasks, even though the input embeddings were not specifically calibrated for supervised learning.

Despite our enthusiasm, there are still many ways in which we can improve \CGMM{}. Throughout the thesis, we saw that regular graphs cannot be discriminated by the model, or that the mean aggregator is not as expressive as the sum. Solving this issues in \CGMM{} while maintaining the fully probabilistic nature of the method is not an easy task. The same can be said for graphical extensions of the model to make it dependent on a target label: implementing a permutation invariant readout transduction that is also tractable remains an unsolved problem, to the best of our knowledge. Notably, the use of a SP variable to model the global aggregation would not work, because the SP does not consider mutual dependencies between the elements to aggregate. For these reasons, a potential future direction would be to devise an efficient sum-based neighbor aggregator and/or readout such that we can still compute closed form solutions when training each layer of our architecture. These two contributions would greatly enhance the expressivity of the probabilistic framework we have introduced.

Another important limitation of \CGMM{} is that its applicability is restricted to the use of discrete edge labels. This is problematic when we want to take into account the distance between entities in the graph or a more complex edge feature that lives in a multi-dimensional space. We know, however, that this problem can be addressed, and the next section will be just devoted to that.

\clearpage
\section[Beyond Discrete Edge Features]{Beyond Discrete Edge Features \cite{atzeni_modeling_2021}}
\label{sec:ecgmm}
We turn our attention to the problem of modeling (possibly multidimensional) continuous edge features in our deep and fully probabilistic framework. The benefits of the methodology we are about to propose are multi-faceted. First and foremost, \textbf{we extend CGMM} to enlarge the classes of graphs it can handle. Secondly, we show that the unsupervised model can build richer graph representations \textit{even in the absence of edge features}.

It is easy to see why we cannot keep using the Switching Parent technique in the presence of continuous edge features. If the support of the \pdf{} associated with the SP variable was infinite, we would have to replace the summation over the discrete states with an integral. By doing that, we would lose the closed-form solutions to the MLE estimation problem as well as the convergence guarantees of the EM algorithm. Because we want to preserve these nice characteristics of \CGMM{}, we have to resort to a different solution.

The key aspect of our contribution is the following: we can \textbf{learn} to \quotes{discretize} edge features so as to use them in the original \CGMM{} model. To do that, we adopt an architectural approach in which we train a secondary Bayesian network to model the generation of edge features. The (discrete) edge state will then be used by the SP variable of the original \CGMM{} model. Empirically, we will show that this is better than using a hand-made heuristic to obtain discrete edge features, and the advantage persists even when edge features are \textbf{absent} from the graph.

In the following sections, we will briefly introduce the mathematical variations to the model, which we will call Extended Contextual Graph Markov Model (\ECGMM{}), noting how the derivations do not change in any way. The asymptotic complexity will remain linear in the number of edges, so the model will still be fairly efficient. Then, we will highlight the performance improvement against \CGMM{} on graph classification benchmarks, which can be attributed to the richness of the learned graph representations. Moreover, we will study the impact of \ECGMM{} on a graph regression and three link prediction tasks, to show the advantage of explicitly modeling the generation of edge features.

\clearpage
\subsection{Layer Definition}

\begin{figure}[ht]
\begin{center}
\centerline{\resizebox{1\textwidth}{!}{\tikzset{every picture/.style={line width=0.75pt}} 

\begin{tikzpicture}[x=0.75pt,y=0.75pt,yscale=-1,xscale=1]

\draw [line width=1.5]  [dash pattern={on 5.63pt off 4.5pt}]  (103.5,284.97) -- (128.04,185.62) ;
\draw [shift={(129,181.73)}, rotate = 463.88] [fill={rgb, 255:red, 0; green, 0; blue, 0 }  ][line width=0.08]  [draw opacity=0] (11.61,-5.58) -- (0,0) -- (11.61,5.58) -- cycle    ;
\draw [line width=1.5]  [dash pattern={on 5.63pt off 4.5pt}]  (43.83,284.97) -- (109.39,179.4) ;
\draw [shift={(111.5,176)}, rotate = 481.84] [fill={rgb, 255:red, 0; green, 0; blue, 0 }  ][line width=0.08]  [draw opacity=0] (11.61,-5.58) -- (0,0) -- (11.61,5.58) -- cycle    ;
\draw  [fill={rgb, 255:red, 31; green, 119; blue, 180 }  ,fill opacity=1 ][line width=1.5]  (104,60.73) .. controls (104,46.93) and (115.19,35.73) .. (129,35.73) .. controls (142.81,35.73) and (154,46.93) .. (154,60.73) .. controls (154,74.54) and (142.81,85.73) .. (129,85.73) .. controls (115.19,85.73) and (104,74.54) .. (104,60.73) -- cycle ;
\draw [line width=1.5]  [dash pattern={on 5.63pt off 4.5pt}]  (197.17,284.97) -- (146.28,182.58) ;
\draw [shift={(144.5,179)}, rotate = 423.57] [fill={rgb, 255:red, 0; green, 0; blue, 0 }  ][line width=0.08]  [draw opacity=0] (11.61,-5.58) -- (0,0) -- (11.61,5.58) -- cycle    ;
\draw  [line width=1.5]  (104,156.73) .. controls (104,142.93) and (115.19,131.73) .. (129,131.73) .. controls (142.81,131.73) and (154,142.93) .. (154,156.73) .. controls (154,170.54) and (142.81,181.73) .. (129,181.73) .. controls (115.19,181.73) and (104,170.54) .. (104,156.73) -- cycle ;
\draw [line width=1.5]    (129,131.73) -- (129,100.07) -- (129,89.73) ;
\draw [shift={(129,85.73)}, rotate = 450] [fill={rgb, 255:red, 0; green, 0; blue, 0 }  ][line width=0.08]  [draw opacity=0] (11.61,-5.58) -- (0,0) -- (11.61,5.58) -- cycle    ;
\draw  [fill={rgb, 255:red, 31; green, 119; blue, 180 }  ,fill opacity=1 ][line width=1.5]  (18.83,284.97) .. controls (18.83,271.16) and (30.03,259.97) .. (43.83,259.97) .. controls (57.64,259.97) and (68.83,271.16) .. (68.83,284.97) .. controls (68.83,298.77) and (57.64,309.97) .. (43.83,309.97) .. controls (30.03,309.97) and (18.83,298.77) .. (18.83,284.97) -- cycle ;
\draw  [fill={rgb, 255:red, 31; green, 119; blue, 180 }  ,fill opacity=1 ][line width=1.5]  (78.5,284.97) .. controls (78.5,271.16) and (89.69,259.97) .. (103.5,259.97) .. controls (117.31,259.97) and (128.5,271.16) .. (128.5,284.97) .. controls (128.5,298.77) and (117.31,309.97) .. (103.5,309.97) .. controls (89.69,309.97) and (78.5,298.77) .. (78.5,284.97) -- cycle ;
\draw  [fill={rgb, 255:red, 31; green, 119; blue, 180 }  ,fill opacity=1 ][line width=1.5]  (172.17,284.97) .. controls (172.17,271.16) and (183.36,259.97) .. (197.17,259.97) .. controls (210.97,259.97) and (222.17,271.16) .. (222.17,284.97) .. controls (222.17,298.77) and (210.97,309.97) .. (197.17,309.97) .. controls (183.36,309.97) and (172.17,298.77) .. (172.17,284.97) -- cycle ;
\draw [line width=1.5]    (209,76.73) -- (150.07,135.98) ;
\draw [shift={(147.25,138.82)}, rotate = 314.85] [fill={rgb, 255:red, 0; green, 0; blue, 0 }  ][line width=0.08]  [draw opacity=0] (11.61,-5.58) -- (0,0) -- (11.61,5.58) -- cycle    ;
\draw  [fill={rgb, 255:red, 255; green, 255; blue, 255 }  ,fill opacity=1 ][line width=1.5]  (184,76.73) .. controls (184,62.93) and (195.19,51.73) .. (209,51.73) .. controls (222.81,51.73) and (234,62.93) .. (234,76.73) .. controls (234,90.54) and (222.81,101.73) .. (209,101.73) .. controls (195.19,101.73) and (184,90.54) .. (184,76.73) -- cycle ;
\draw [line width=1.5]    (526,78.73) -- (467.07,137.98) ;
\draw [shift={(464.25,140.82)}, rotate = 314.85] [fill={rgb, 255:red, 0; green, 0; blue, 0 }  ][line width=0.08]  [draw opacity=0] (11.61,-5.58) -- (0,0) -- (11.61,5.58) -- cycle    ;
\draw  [fill={rgb, 255:red, 31; green, 119; blue, 180 }  ,fill opacity=1 ][line width=1.5]  (421,62.73) .. controls (421,48.93) and (432.19,37.73) .. (446,37.73) .. controls (459.81,37.73) and (471,48.93) .. (471,62.73) .. controls (471,76.54) and (459.81,87.73) .. (446,87.73) .. controls (432.19,87.73) and (421,76.54) .. (421,62.73) -- cycle ;
\draw  [line width=1.5]  (421,158.73) .. controls (421,144.93) and (432.19,133.73) .. (446,133.73) .. controls (459.81,133.73) and (471,144.93) .. (471,158.73) .. controls (471,172.54) and (459.81,183.73) .. (446,183.73) .. controls (432.19,183.73) and (421,172.54) .. (421,158.73) -- cycle ;
\draw  [fill={rgb, 255:red, 255; green, 255; blue, 255 }  ,fill opacity=1 ][line width=1.5]  (501,78.73) .. controls (501,64.93) and (512.19,53.73) .. (526,53.73) .. controls (539.81,53.73) and (551,64.93) .. (551,78.73) .. controls (551,92.54) and (539.81,103.73) .. (526,103.73) .. controls (512.19,103.73) and (501,92.54) .. (501,78.73) -- cycle ;
\draw [line width=1.5]    (446,133.73) -- (446,102.07) -- (446,91.73) ;
\draw [shift={(446,87.73)}, rotate = 450] [fill={rgb, 255:red, 0; green, 0; blue, 0 }  ][line width=0.08]  [draw opacity=0] (11.61,-5.58) -- (0,0) -- (11.61,5.58) -- cycle    ;
\draw  [fill={rgb, 255:red, 31; green, 119; blue, 180 }  ,fill opacity=1 ][line width=1.5]  (465.83,286.97) .. controls (465.83,273.16) and (477.03,261.97) .. (490.83,261.97) .. controls (504.64,261.97) and (515.83,273.16) .. (515.83,286.97) .. controls (515.83,300.77) and (504.64,311.97) .. (490.83,311.97) .. controls (477.03,311.97) and (465.83,300.77) .. (465.83,286.97) -- cycle ;
\draw  [fill={rgb, 255:red, 31; green, 119; blue, 180 }  ,fill opacity=1 ][line width=1.5]  (382.5,286.97) .. controls (382.5,273.16) and (393.69,261.97) .. (407.5,261.97) .. controls (421.31,261.97) and (432.5,273.16) .. (432.5,286.97) .. controls (432.5,300.77) and (421.31,311.97) .. (407.5,311.97) .. controls (393.69,311.97) and (382.5,300.77) .. (382.5,286.97) -- cycle ;
\draw [line width=1.5]  [dash pattern={on 5.63pt off 4.5pt}]  (481.83,264.12) -- (454.49,185.9) ;
\draw [shift={(453.17,182.12)}, rotate = 430.73] [fill={rgb, 255:red, 0; green, 0; blue, 0 }  ][line width=0.08]  [draw opacity=0] (11.61,-5.58) -- (0,0) -- (11.61,5.58) -- cycle    ;
\draw [line width=1.5]  [dash pattern={on 5.63pt off 4.5pt}]  (413.83,260.79) -- (436.01,187.29) ;
\draw [shift={(437.17,183.46)}, rotate = 466.79] [fill={rgb, 255:red, 0; green, 0; blue, 0 }  ][line width=0.08]  [draw opacity=0] (11.61,-5.58) -- (0,0) -- (11.61,5.58) -- cycle    ;
\draw  [line width=1.5]  (8.55,23.58) -- (314.58,23.58) -- (314.58,319) -- (8.55,319) -- cycle ;
\draw  [line width=1.5]  (324.55,23.58) -- (640.58,23.58) -- (640.58,319) -- (324.55,319) -- cycle ;

\draw (129,156.73) node  [font=\Large]  {$Q_{u}$};
\draw (129,60.73) node  [font=\Large]  {$\textcolor[rgb]{1,1,1}{X}\textcolor[rgb]{1,1,1}{_{u}}$};
\draw (197.17,284.97) node  [font=\large]  {$\textcolor[rgb]{1,1,1}{Q}\textcolor[rgb]{1,1,1}{_{|\mathcal{N}_{u} |}^{\ell -1}}$};
\draw (103.5,284.97) node  [font=\Large]  {$\textcolor[rgb]{1,1,1}{Q}\textcolor[rgb]{1,1,1}{_{2}^{\ell -1}}$};
\draw (43.83,284.97) node  [font=\Large]  {$\textcolor[rgb]{1,1,1}{Q}\textcolor[rgb]{1,1,1}{_{1}^{\ell -1}}$};
\draw (209,76.73) node  [font=\Large]  {$S_{u}$};
\draw (148.4,282.43) node  [font=\large,color={rgb, 255:red, 0; green, 0; blue, 0 }  ,opacity=1 ]  {$\mathbf{\dotsc }$};
\draw (446,158.73) node  [font=\Large]  {$Q_{uv}$};
\draw (526,78.73) node  [font=\Large]  {$S_{uv}$};
\draw (446,62.73) node  [font=\Large]  {$\textcolor[rgb]{1,1,1}{A}\textcolor[rgb]{1,1,1}{_{uv}}$};
\draw (490.83,286.97) node  [font=\Large]  {$\textcolor[rgb]{1,1,1}{Q}\textcolor[rgb]{1,1,1}{_{v}^{\ell -1}}$};
\draw (407.5,286.97) node  [font=\Large]  {$\textcolor[rgb]{1,1,1}{Q}\textcolor[rgb]{1,1,1}{_{u}^{\ell -1}}$};
\draw (399,217.73) node  [font=\Large]  {$A_{s}$};
\draw (489,217.73) node  [font=\Large]  {$A_{d}$};
\draw (52.86,46.58) node  [font=\Large] [align=left] {$\displaystyle Layer$\textit{ }$\displaystyle \ell $};
\draw (270.24,264.7) node  [font=\Large]  {$u\ \in \ \mathcal{V}_{g}$};
\draw (271.42,299.27) node  [font=\Large]  {$g\ \in \ \mathcal{D}$};
\draw (368.86,46.58) node  [font=\Large] [align=left] {$\displaystyle Layer$\textit{ }$\displaystyle \ell $};
\draw (580.24,264.7) node  [font=\Large]  {$( u,v) \ \in \ \mathcal{E}_{g}$};
\draw (597.42,299.27) node  [font=\Large]  {$g\ \in \ \mathcal{D}$};

\end{tikzpicture}}}
\end{center}
\caption{Graphical model of a generic layer $\lcurr$ of \ECGMM{}. Dashed arrows denote the flow of contextual information coming from previous layers. We have omitted the SP variables $L_u$ and $L_\uv$ for simplicity of exposition.}
\label{fig:ecgmm-basic-layer}
\end{figure}
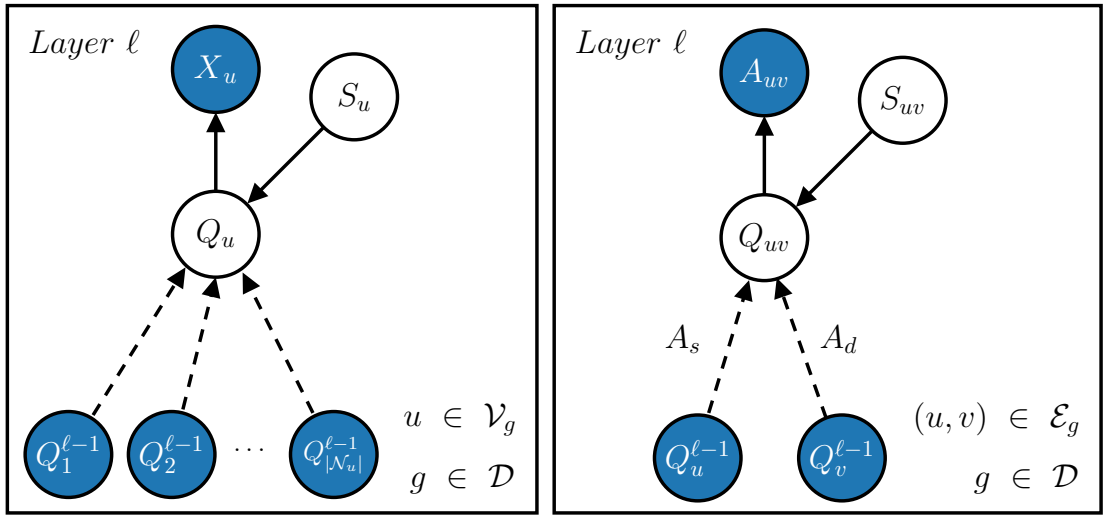

The graphical model of each \ECGMM{}'s layer is represented in Figure \ref{fig:ecgmm-basic-layer}: an additional Bayesian network, very similar to that of \CGMM{}, is responsible for modeling the generation each edge feature $a_\uv$ through a latent categorical variable $Q_{uv}$. This variable can take $C_E$ different values, in contrast to the number of latent states of the original models that we will call $C_V$ from now on. In principle, edges act as fictitious vertices whose neighbors are the source and destination vertex states inferred and frozen at the previous layers. In the interest of clarity, we will omit the layer-wise SP variable and consider contributions coming from the previous layer only. In addition, the \quotes{edge component} on the right uses two special discrete edge labels, one for the source ($A_s$) and one for the destination $A_d$ states, to model the direction of the edge under consideration. If the graph is undirected, we can assume a uniform distribution for the SP variable $S_\uv$.

Formally, we model the generation of vertex and edge features as follows:
\begin{align*}
    & P(x_u \mid \boldQ{\N{u}{}}{\lprev}, \boldQ{\Eset{u}{}}{\lprev}) = \sum_{i=1}^{C_V} \underbrace{P(x_u \mid \Q{u}{}=i)}_\text{vertex emission}P(\Q{u}{} = i \mid \boldQ{\N{u}{}}{\lprev}, \boldQ{\Eset{u}{}}{\lprev}) \\
    & P(a_\uv \mid  \Q{u}{\lprev}, \Q{v}{\lprev}) = \sum_{i=1}^{C_E} \underbrace{P(a_\uv \mid \Q{\uv}{}=i)}_\text{edge emission}P(\Q{\uv}{}=i \mid \Q{u}{\lprev}, \Q{v}{\lprev}),
\end{align*}
where $\boldQ{\Eset{u}{}}{\lprev}$ denotes the set of states inferred by the edge component at the previous layer. Likewise \CGMM{}, when $\lcurr=0$, the equations simplify and the layer implements two standard mixture models that do not consider contextual information.

Thanks to the fact that frozen edge states are inferred from a categorical variable, we approximate the conditional distribution of the vertex component as
\begin{align*}
    & P(\Q{u}{} = i \mid \boldQ{\N{u}{}}{\lprev}, \boldQ{\Eset{u}{}}{\lprev}) = \sum_{a=1}^{C_E}\underbrace{P(S_u=a)}_{SP}\underbrace{P^{a}(Q_u=i\mid \boldQ{\N{u}{a}}{\lprev},\boldQ{\Eset{u}{}}{\lprev})}_\text{vertex transition}.
\end{align*}
This last equation, while apparently similar to the one of \CGMM{}, shows the interplay between the vertex-centric and edge-centric components of \ECGMM{}, which makes it possible to incorporate arbitrary edge information in the fully probabilistic neighborhood aggregation scheme.

Similarly, the rightmost term of $P(a_\uv \mid  \Q{u}{\lprev}, \Q{v}{\lprev})$ can be decomposed as
\begin{align*}
    & P(\Q{\uv}{}=i\mid \Q{u}{\lprev}, \Q{v}{\lprev}) =\sum_a^{A_s, A_d} \underbrace{P(S_{uv} = a)}_{SP}\underbrace{P^{a}(\Q{\uv}{}=i\mid \Q{a}{\lprev})}_\text{edge transition},
\end{align*}
where we remind that $A_s$ and $A_d$ are the discrete labels assigned to source and destination vertices, respectively, and $\Q{a}{\lprev}$ is $\Q{u}{\lprev}$ if $a=A_s$, and $\Q{v}{\lprev}$ otherwise.

The last brick in the formalization of the model is the definition of the transition distribution $P^{a}(Q_u=i\mid \boldQ{\N{u}{a}}{\lprev}, \boldQ{\Eset{u}{}}{\lprev})$. Using the additional edge information we have, we can write
\begin{align*}
    & P^{a}(Q_u=i\mid \boldQ{\N{u}{a}}{\lprev}, \boldQ{\Eset{u}{}}{\lprev}) =  \sum_{j=1}^{C_V}P^a(\Q{u}{}=i \mid \Q{*}{\lprev}=j)\sum_{v\in\N{u}{a}}q_v^{\lprev}(j)\frac{q_{uv}^{\lprev}(a)}{\sum_{v\in\N{u}{a}}q^{\lprev}_\uv(a)}
\label{eq:vertex-aggr}
\end{align*}
where we recall that $q_u(j)$ and $q_{uv}(j)$ are the $j$-th components of the inferred states (represented as a vector) inferred at the previous layer. The transition distribution we just presented is a generalization of Equation \ref{eq:cgmm-sp-decomposition}, where we have exploited the posterior of each edge to weight the contribution of the individual neighbors. Moreover, as in \CGMM{}, we assume full stationarity on vertices and on edges, meaning that we share the parameters of the emission, transition, and SP distributions across all vertices or edges depending on the component of \ECGMM{}.

To train an \ECGMM{} layer, and similarly for inference, it is sufficient to apply EM to the two \textbf{independent} Bayesian networks, and use their inferred states as statistics for the subsequent layer of the architecture. We remark that. mathematically speaking, we are still dealing with conditional mixture models. Therefore, at the cost of training an additional network for edges, which shares a similar time complexity as the original \CGMM{}, we obtain a deep architecture capable of building \textbf{both vertex and edge embeddings} from raw graphs, something that not many other DGNs can do to the best of our knowledge.

\subsection{Dynamic Neighborhood Aggregation}
\begin{figure}[ht]
    \begin{center}
    \hspace{-3cm}
    \centerline{\resizebox{1.1\textwidth}{!}{\input{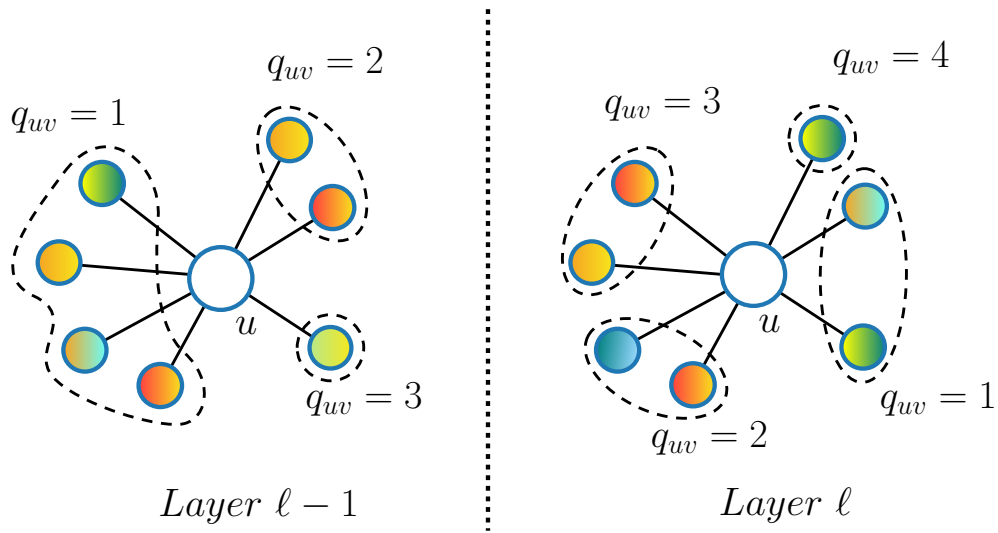}}}
    \end{center}
    \caption{We show an example of dynamic neighborhood aggregation with $C_E=4$. At layer $\lprev$, the neighbors of vertex $u$ are split into 3 groups according to the \textbf{edge} states computed at layer $\lcurr-2$. Because edge states vary, at layer $\lcurr$ a different grouping of the same neighbors can be induced.} 
    \label{fig:dynamic-aggr}
\end{figure}

There is one subtle but very important difference between \CGMM{} and \ECGMM{} that could potentially contribute to the richness of vertex embeddings produced by the latter. Whenever edge features are missing but a dummy feature is used in their place, the edge latent states can still vary across the graph because they depend on the \textit{source} and \textit{destination} frozen states. Therefore, at different layers, the posterior distributions of the same edge may be different regardless of the absence of real edge features. From a methodological point of view, this allows for \textbf{different} groupings of the \textbf{same} vertex neighbors at different layers. On the other hand, CGMM always groups neighbors in the same way, since it relies on static and discrete edge features. We denote this peculiar characteristic of \ECGMM{} with the term \quotes{dynamic neighborhood aggregation}, which is sketched in Figure \ref{fig:dynamic-aggr}. We believe this is the main reason why \ECGMM{} shows significant performance improvements with respect to \CGMM{} on the graph classification benchmarks, as we will see in the following.

\subsection{Complexity and Scalability}
\ECGMM{} shares an asymptotic efficiency comparable, in both time and space, to \CGMM{}. At a given layer $\lcurr$, the complexity of \ECGMM{}'s vertex component is bounded by $\mathcal{O}(|\Vset{g}|  (C_E  C_V^2 + KC_V)$, where $K$ is the number of vertex features. To compute the posterior and bigram of each vertex, time complexity is again $\mathcal{O}(|\Eset{g}|)$. Hence, the overall time complexity becomes $\mathcal{O}(|\Vset{g}| + |\Eset{g}|)$. Instead, the edge component of \ECGMM{} is bounded by $\mathcal{O}(|\Eset{g}|(2C_EC_V + KC)$. Clearly, the time complexity of our extension will be always strictly greater than \CGMM{}; however, asymptotically speaking, it is still controlled by $\mathcal{O}(|\Vset{g}| + |\Eset{g}|)$ whenever $C_V \ll |\Vset{g}|, C_V \ll |\Eset{g}|, C_E \ll |\Vset{g}|$, and $C_E \ll |\Eset{g}|$.

\subsection{Embeddings Construction}
Differently to \CGMM{}, after the inference phase we will obtain both vertex and edge representations At model selection time, we can still choose to concatenate a vertex unigram and bigram (obtaining a \textit{unibigram}) or not. However, to obtain graph representations at each layer $\lcurr$, we shall independently aggregate all vertex and edge representations in the graph via permutation-invariant operators, such as the mean or the sum, followed by concatenation of the two resulting vectors. The final unsupervised graph embedding is then the concatenation of these embeddings across all layers of the deep architecture.

\subsection{Experimental Setting}
To assess the performances of \ECGMM{}, we will carry out three different evaluations. First, we will compare against the same set of graph classification benchmarks on which we evaluated \CGMM{}. Secondly, we will show the importance of adaptively discretizing edge features on a graph regression benchmark. Finally, we will compare \CGMM{} and \ECGMM{} on the three link prediction tasks following a rigorous evaluation scheme.\footnote{\url{https://github.com/diningphil/E-CGMM}}.

\subsubsection*{Graph Classification}
Please recall that the datasets under consideration do not provide edge attributes, so the scope of our analysis is to evaluate the richness of graph embeddings given by the \textbf{dynamic aggregation mechanism} as well as the use of posterior edge probabilities. We follow the empirical evaluation of Section \ref{sec:scholarship-issues}.

In terms of hyper-parameters, we set $C_V$ to $20$\footnote{This value was shown to guarantee better or on-par performances than smaller values during our \CGMM{} experiments, therefore we fixed it to reduce the already large number of configurations to try.} whereas $C_E$ was chosen from $\{5, 10\}$. The other hyper-parameters are selected according to Table \ref{tab:cgmm-hyperparams}, to keep the model selection as similar as possible between the two methods.

\subsubsection*{Graph Regression}
To understand the importance of handling continuous edge information, we consider the QM7b graph regression task \cite{qm7b, qm7b2}, a chemical dataset composed of more than 7k organic molecules. The task is to predict $14$ continuous properties of each molecule, where a molecule is associated with a Coulomb interaction matrix that is used to extract vertex and edge features. The entry $(i, j)$ of the Coulomb matrix is proportional to the product of nuclear charges of atoms $i$ and $j$ and inversely proportional to their distance (for non-diagonal elements).
In the dataset, there are 6 different continuous diagonal elements of the matrix, which are used as discrete labels for the vertices. On the other hand, we consider edges associated with matrix entries greater than $0.52$ to induce sparsity on the graph. To quantitatively evaluate the goodness of \ECGMM{}, we rely on the Mean Absolute Error (MAE) objective function.

To show that a naive edge discretization technique may not work well, we consider an alternative representation of the molecules where we discretize the continuous edge features using $10$ bins of equal widths, so that we can train \CGMM{} using the SP technique.

We used the same model selection and assessment setup of graph classification to get reliable performance estimates. We tried different configurations with depth in $\{10, 20\}$, $C_E$ in $\{5, 10\}$, $C_V$ in $\{10, 20\}$ for CGMM and $20$ for E-CGMM, continuous or discrete frozen states, unigrams or unibigrams, and sum or mean aggregation. The MLP configurations, after a preliminary screening on the validation set to reduce the number of configurations, were: $2000$ maximum epochs, hidden layer dimension in $\{32, 128\}$, and learning rate equal to $5\cdot 10^{-4}$. Early stopping's patience was set to $100$, and the $L2$ regularization had weight decay $10^{-4}$.

\subsubsection*{Link Prediction}
Since \ECGMM{} can model the generation of edges at each layer, we also tested some link prediction benchmarks, namely Cora, Citeseer and Pubmed \cite{shchur_pitfalls_2018}. These are three citation networks (\ie undirected graphs) in which vertices represent documents, and edges represent citations. Because there is no standardized evaluation on these datasets when it comes to link prediction, the goal of these experiments will be to just show how \ECGMM{} can tackle link prediction by design, with better performances than \CGMM{}.

To adopt a robust experimental protocol even in this third task, we perform 10-fold cross validation for model assessment, with an hold-out technique for model selection. Yet, the difference with the previous tasks lies in how we build each outer fold. In this case, portions of positive edges in each graph are used as validation and test sets, together with a randomly selected subset of negative edges. The remaining edges are used to train both our models. Since \CGMM{} does not produce edge embeddings, we used an MLP to predict whether a link $(u,v)$ exists starting from the mean embedding $\boldhell{\uv}{}=(\boldhell{u}{}+\boldhell{v}{})/2$. In particular, we chose the mean operator over concatenation because the latter would introduce asymmetries, \ie learning difficulties, when modeling the existing undirected edges.
To infer the existence of an undirected link using \ECGMM{}, we construct the mean embedding $\boldhell{\uv}{}$ at each layer using the posterior distributions of $Q_\uv$ and $Q_\vu$.

For every dataset and for both models, we choose the following hyper-parameters: $C_V \in \{10, 20\}$ and number of layers in $\{2, 4, 6, \dots , 20\}$, $C_E \in \{5, 10\}$ (\ECGMM{} only), discrete or continuous vertex representations, and unigrams or unibigrams. For the MLP, we chose the learning rate in $\{10^{-3}, 10^{-4}, 10^{-5}\}$, the hidden dimension in $\{128, 256\}$, and weight decay in  $\{10^{-3}, 10^{-5}\}$.

\subsection{Results}
We now present our empirical findings starting from graph classification to graph regression and link prediction. The goal of this section is simply to show that \ECGMM{} almost consistently improves the metrics of interest.

\subsubsection*{Graph Classification}
Chemical and social graph classification results are detailed in Tables \ref{tab:ecgmm-iclr-chemical-results} and \ref{tab:ecgmm-iclr-social-results}. They show how \ECGMM{} is basically on par with \CGMM{} on those tasks where the baseline is able to get near to or better than the state of the art, but it improves on the others. Notably, there is a substantial gap on NCI1 and both REDDIT tasks. We believe such improvements are attributable to the new capabilities introduced with the edge component of each layer, \ie the ability to dynamically aggregate neighbors, in a way that explicitly and adaptively depends on the local connections, and the \quotes{new} graph embedding enriched with global edge information.
\begin{table}[ht]
\centering
\caption{Mean and standard deviation results on chemical datasets of a 10-fold Cross Validation (setup of Section \ref{sec:scholarship-issues}). Best results are reported in bold.}
\label{tab:ecgmm-iclr-chemical-results}
\begin{tabular}{l c c c c}
\toprule
     & \textbf{D\&D} & \textbf{NCI1} & \textbf{PROTEINS}\\
\midrule
 \Baseline & $\mathbf{78.4}\pm 4.5 $ &  $69.8 \pm 2.2 $ &  $\mathbf{75.8} \pm 3.7 $ \\
 \DGCNN & $76.6 \pm 4.3 $ &  $76.4 \pm 1.7 $ &  $72.9 \pm 3.5 $   \\
 \DiffPool & $75.0 \pm 3.5 $ &  $76.9 \pm 1.9 $ &  $73.7 \pm 3.5 $  \\
 \ECC & $72.6 \pm 4.1 $ &  $76.2 \pm 1.4 $ &  $72.3 \pm 3.4 $  \\
 \GIN & $75.3 \pm 2.9 $ &  $\mathbf{80.0} \pm 1.4 $ &  $73.3 \pm 4.0 $    \\
 \GraphSAGE & $72.9 \pm 2.0 $ &  $76.0 \pm 1.8 $ &  $73.0 \pm 4.5 $  \\
 \CGMM{}  & $74.9 \pm 3.4 $ & $ 76.2 \pm2.0$ & $74.0 \pm 3.9$ \\  \hline
 \ECGMM{} & $73.9 \pm4.1$ & $ 78.5 \pm 1.7$ & $73.3 \pm 4.1$ \\
\bottomrule
\end{tabular}
\end{table}
\begin{table}[ht]
\small
\centering
\caption{Mean and standard deviation results on social datasets of a 10-fold Cross Validation (setup of Section \ref{sec:scholarship-issues}). Best results are reported in bold. Note that the degree is the sole vertex feature used by all models.}
\label{tab:ecgmm-iclr-social-results}
\begin{tabular}{lccccc}
\toprule
     & \textbf{IMDB-B} & \textbf{IMDB-M} & \textbf{REDDIT-B} & \textbf{REDDIT-5K} & \textbf{COLLAB}\\
    \midrule
 \Baseline & $70.8 \pm 5.0 $ &  $\mathbf{49.1} \pm 3.5 $ &  $82.2 \pm 3.0 $ &  $52.2 \pm 1.5 $ &  $70.2 \pm 1.5 $   \\
 \DGCNN & $69.2 \pm 3.0 $ &  $45.6 \pm 3.4 $ &  $87.8 \pm 2.5 $ &  $49.2 \pm 1.2 $ &  $71.2 \pm 1.9 $   \\
 \DiffPool & $68.4 \pm 3.3 $ &  $45.6 \pm 3.4 $ &  $89.1 \pm 1.6 $ &  $53.8 \pm 1.4 $ &  $68.9 \pm 2.0 $   \\
 \ECC & $67.7 \pm 2.8 $ &  $43.5 \pm 3.1 $ &   - &   - &   -   \\
 \GIN & $71.2 \pm 3.9 $ &  $48.5 \pm 3.3 $ &  $\mathbf{89.9} \pm 1.9 $ &  $\mathbf{56.1} \pm 1.7 $ &  $75.6 \pm 2.3 $   \\
 \GraphSAGE & $68.8 \pm 4.5 $ &  $47.6 \pm 3.5 $ &  $84.3 \pm 1.9 $ &  $50.0 \pm 1.3 $ &  $73.9 \pm 1.7 $   \\

 \CGMM{}  & $\mathbf{72.7} \pm 3.6$ & $47.5 \pm 3.9$ & $88.1 \pm 1.9$ & $52.4 \pm 2.2$ & $77.32 \pm 2.2$    \\ \midrule
 \ECGMM{} & $70.7 \pm 3.8$ & $48.3 \pm 4.1 $& $89.5 \pm 1.3$ & $53.7 \pm 1.0 $ & $\mathbf{77.45} \pm 2.3$  \\ \bottomrule
\end{tabular}
\end{table}

\subsubsection*{Graph Regression}
Table \ref{tab:graph_regr} reports our graph regression analysis. We can see that E-CGMM performs better than both versions of \CGMM{}, \ie one that ignores edge attributes and the other working on discretized edge labels. In particular, there is a relative improvment in MAE of 17\%-19\% that was to be expected, given the importance of the information contained in the Coulomb interaction matrix. Such an improvement also proves the inadequacy of non-adaptive edge discretization techniques, which not only may force the user to take decisions a-priori but might also cause loss of relevant information.

\begin{table}[ht]
    \centering
    \begin{tabular}{l c  c}
     \hline
     & MAE & Relative Improvement\\
     \hline
     \CGMM{}-no edge attributes & $1.52 \pm 0.05$ & $19\%$ \\
     \CGMM{}-discretized edges   & $1.49 \pm 0.07$ & $17\%$\\ \hline
     \ECGMM{} & $\mathbf{1.23} \pm 0.06$ & - \\
     \hline
    \end{tabular}
    \vskip 0.1in
    \caption{Graph regression results and relative improvement of E-CGMM compared to CGMM. Best results are in bold. CGMM results are reported for both a version of the dataset with no edge attributes as well as for discretized edge labels.}
    \label{tab:graph_regr}
\end{table}

\subsubsection*{Link Prediction}
We conclude our analysis with link prediction experiments, summarized in Table \ref{tab:link_pred}. The numbers indicate a substantial improvements with respect to \CGMM{} on every dataset tested, with an average accuracy increase of 3\%-4\%. By modeling the generation of positive and negative edges, \ECGMM{} captures the conditional distribution of the edges given the frozen vertex states, thus building more informative edge posteriors. On the other hand, the way in which we have built edge representations with \CGMM{} is yet another a-priori choice that we have managed to avoid with this new method.
\begin{table}[h]
    \centering
    \begin{tabular}{l c  c  c}
     \hline
      & Cora & Citeseer & Pubmed \\
     \hline
     \CGMM{}& $82.62 \pm 1.8$ & $74.47 \pm 2.2$ & $77.09 \pm 1.9$ \\
     \ECGMM{} & $\mathbf{86.76} \pm 2.3$ & $\mathbf{77.69} \pm 1.7$ & $\mathbf{81.58} \pm 1.6$ \\
     \hline
    \end{tabular}
    \vskip 0.1in
    \caption{Comparison between \ECGMM{} and \CGMM{} on link prediction tasks. Best results are reported in bold.}
    \label{tab:link_pred}
\end{table}

\subsection{Summary}
We have extended our fully probabilistic framework with an \quotes{edge-aware} version of the Contextual Graph Markov Model. \ECGMM{} allows us the process a broader class of graphs with potentially arbitrary edge features. To achieve this goal, we took an architectural approach by introducing an additional Bayesian network responsible for the generative modeling of edge features. Thanks to the richer graph embeddings produced, we have observed empirical improvements with respect to \CGMM{} on three different tasks, while keeping the asymptotic complexity linear in the number of edges.

It is worth noticing that many of the future directions lied down in Section \ref{subsec:cgmm-summary} would implicitly apply here, as they involve modifying the Bayesian network rather than architectural aspects. Combined with the benefits of explicitly modeling edge features, we believe there is still room for improvement on both technical and empirical sides.

\clearpage
\section{The Infinite Contextual Graph Markov Model}
\label{sec:icgmm}

As with most Deep Graph Networks, one inherent limitation of \CGMM{} is the absence of a mechanism to learn the size of each layer's latent representation. This is also related to one of the most challenging problems of \ML{}, that is, the selection of appropriate hyper-parameters for the task at hand. In fact, due to the data-dependent nature of the learning problem, we have seen throughout this thesis that there exists no single model configuration that works well in every situation. One usually relies on model selection techniques such as grid or random searches \cite{bergstra_random_2012}, where the hyper-parameters configurations to try are chosen a-priori by the \ML{} practitioner.

In Chapter \ref{chapter:background}, however, we have briefly introduced Bayesian nonparametric methods, in particular HDP mixture models, that automatically choose the \quotes{right} amount of clusters to use \cite{gershman_tutorial_2012}. We recall that, in the BNP literature, the complexity of the models, \eg the number of states, automatically grows \textit{with the data} \cite{teh_hdp_2006}. Since each \CGMM{} is essentially a conditional mixture model, it would make sense to apply a BNP treatment to \CGMM{} in order to automatize the choice of its hyper-parameters.

In this section, we present our last methodological contribution to the family of Deep Bayesian Graph Networks. The principal difficulty of extending \CGMM{} lies in how to handle the variable-size number of neighbors of each vertex inside the BNP framework, which in \CGMM{} is solved by (possibly weighted) convex combinations of the frozen neighbors' posteriors. We shall see how the notion of a neighboring \textbf{macro-state} will be particularly useful in this context. It is thanks to this realization that we will be able to replace the \CGMM{} layer with an HDP mixture model, without incurring in major technical challenges.

The resulting model, called Infinite Contextual Graph Markov Model (\iCGMM{}), can generate as many latent states as needed to solve the unsupervised density estimation task at each layer. To the extent of our knowledge, this is the first \textbf{deep}, \textbf{Bayesian nonparametric} model for \textbf{graph processing}. To increase its efficiency, and despite the existence of variational inference alternatives \cite{wang_truncation_2012,bryant_truly_2012,hoffman_stochastic_2013, hughes_reliable_2015}, we opted for a straightforward and faster heuristic that scales to the social datasets considered in this thesis, requires little code modification, and works as well as the original implementation.

We compare \iCGMM{} against \CGMM{}, \ECGMM{} and end-to-end supervised methods on the graph classification tasks of Section \ref{subsec:cgmm-exp-setting}. Results show that \iCGMM{} performs on par or better than \CGMM{}. We complement the analysis with studies on the effects of depth and generation of our model's latent states. All in all, we believe that \iCGMM{} is an important (if not the first) step towards a theoretically grounded and automatic construction of Deep Bayesian Graph Networks.

\subsection{Layer Definition}
There are different ways to define an HDP mixture model, but we will mainly use the stick-breaking construction of Section \ref{subsubsec:dpmm}. Nevertheless, another representation exists, called Chinese Restaurant Franchise (CRF), which will be needed in the following. The CRF extends the CRP to hierarchical models, by assuming that there are $\bar{C}$ \quotes{restaurants}, \ie the known groups assigned to observations in an HDP. Each observation, in our case the variable $X_u$, is called a \quotes{customer}. Following the HDP mixture model literature, each vertex $u$ must be already assigned to one of the $\bar{C}$ different groups. Hence, we will use the term $n_j$ to indicate the number of customers eating at restaurant $j$, \ie given a graph $g$ it must hold $\sum_{j=1}^{\bar{C}} n_j=|\Vset{g}|$. In addition, a latent state $c$, modeled by the mixture variable $\Q{u}{}$, will be assigned to each vertex.

The value that $\Q{u}{}$ takes, namely $q_u$, specifies one of the emission components of the \textbf{possibly infinite} mixture model, which is parametrized by $\bm{\theta}_c, c \in \mathbb{N}$. Continuing with the CRF metaphor, we can say that a customer $u$ goes to restaurant $j_u$ and sits at one of the \textbf{tables} $t_u$ where \textbf{dish} $q_u=c$ is served. Importantly, we assume stationarity of the emission distributions with respect to the groups, meaning there is a form of parameter sharing of the emission distributions \textbf{across different groups}. It is appropriate to remark that the assignment of a customer $u$ to a specific table $t_u$ is unnecessary in the Stick-breaking formulation we will use, but the CRF notion of \quotes{table assignment} will provide an exact and efficient way to solve some technical challenges in the implementation of the model. Finally, please recall from Section \ref{subsubsec:stick-breaking} that even if the number of the possible latent states is infinite, \textbf{only a finite} number of them will be used in the model's implementation. Hence, we refer to this finite value of clusters with the usual symbol $C$.

\begin{figure}[ht]
    \centering
    \resizebox{0.8\textwidth}{!}{\input{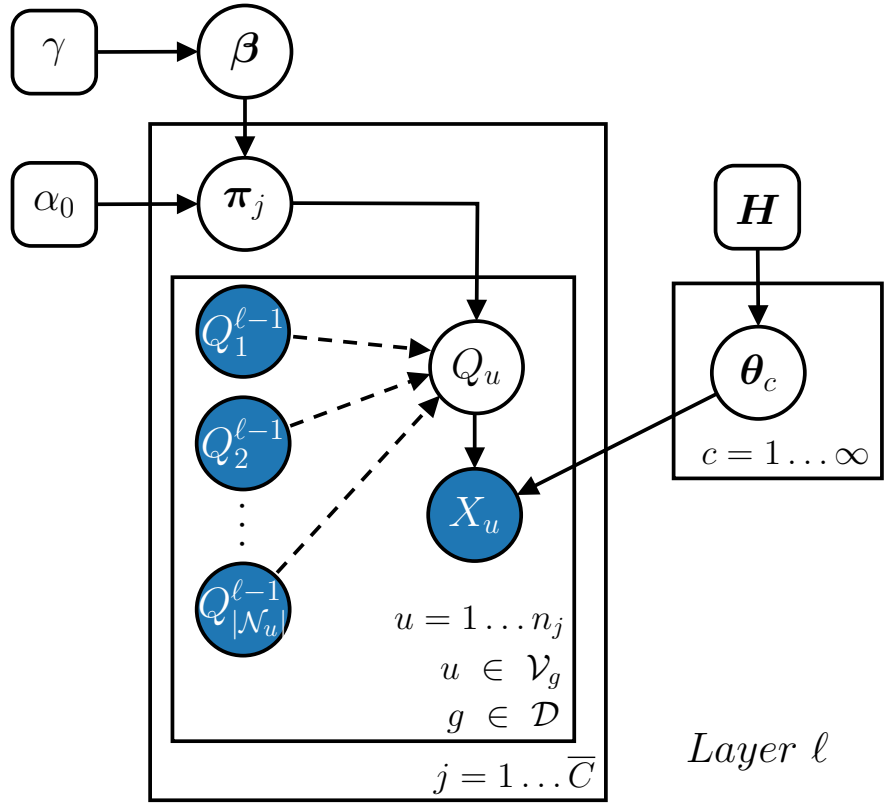}}
    \caption{Graphical model of a generic \iCGMM{} layer $\lcurr$, where observable variables are blue circles, latent ones are white ones, and white boxes denote prior knowledge. This is an HDP mixture model where the group $j_u$ for each observation $X_u$ is determined by the set of neighboring states. The symbol $\bar{C}$ corresponds to the selected number of states determined at the previous layer of the deep incremental architecture. Dashed arrows denote the flow of contextual information.}
    \label{fig:layer}
\end{figure}
The graphical model of a generic \iCGMM{} layer $\lcurr$ is shown in Figure \ref{fig:layer}, where most of the notation has already been described in Section \ref{fig:hdpmm}. We model the generative process of the observable vertex feature $X_u$ conditioned on a set of observable variables of neighboring vertices $\boldQ{\N{u}{}}{\lprev} = \{\boldQ{v}{\lprev} \in [0,1]^{\bar{C}} \mid v \in \mathcal{N}_u\}$, \ie the usual vectors of probabilities inferred and frozen at the previous layer. It follows that in \iCGMM{} each layer has a different number of groups $\bar{C}_{\lcurr}$; when clear from the context, we will omit the symbol $\lcurr$ to ease the notation.

Overall, the generative process of a single \iCGMM{} layer can be formalized as follows:
\begin{equation}
	\begin{alignedat}{2}
		\bm{\beta} \mid \gamma &\sim \text{Stick}(\gamma) &\qquad\qquad j_u \mid \boldQ{\N{u}{}}{\lprev} &= \psi(\boldQ{\N{u}{}}{\lprev}) \\
		\bm{\pi}_j \mid \bm{\beta}, \alpha_0 &\sim \text{DP}(\alpha_0, \bm{\beta}) & \qquad\qquad q_u \mid j_u, (\bm{\pi}_j)_{j=1}^{\bar{C}} &\sim \bm{\pi}_{j_u}\\
		\bm{\theta} \mid \bm{H} &\sim \bm{H} & x_u \mid q_u, (\bm{\theta})_{c=1}^{\infty} &\sim F(\bm{\theta}_{q_u}),
	\end{alignedat}
\end{equation}
where $F(\bm{\theta}_{q_u})$ denotes the emission distribution parametrized by $\bm{\theta}_{q_u}$, and $j_u$ is the group of vertex $u$ chosen according to a permutation invariant function $\psi(\boldQ{\N{u}{}}{})$. To generate a possibly infinite number of emission distributions, we sample from a prior distribution $\bm{H}$. Instead, we sample the distribution $\bm{\beta}$ via the Stick-breaking process $\text{Stick}(\gamma)$ (Section \ref{subsubsec:stick-breaking}). In turn, $\bm{\beta}$ is used by a DP to generate the distribution $\bm{\pi}_j$, responsible for sampling the dish $q_u$ that customer $u$ eats at restaurant $j_u$.

\subsubsection*{Deterministic Choice of the Group}
The crux of the matter is that we need a sensible way to assign each observation to one of the $\bar{C}$ groups. Contrarily to a standard HDP, in which the groups are somehow known, here we cannot rely on a-priori information to make the assignments at each layer. The reason is two-fold: i) we would need an oracle for the layer-wise assignments, since these do not have a straightforward interpretation in our context; ii) if we fixed in advance the group for each observation for all layers, there would be no contextual information to spread across the graph, and therefore all HDPs would be truly independent between each other. It follows that, to induce a dependency between subsequent layers, we must again rely on the frozen states of each vertex as \CGMM{}. These states are indeed the only actionable information we have in order to propagate information across the graph.

The question now becomes how to obtain a group from the set of frozen states of each vertex. In an attempt to be as similar as possible to \CGMM{}, we can reuse the concept of \textbf{macro-state} introduced in the previous sections. By considering the simplest version of a \CGMM{} layer, \ie the one without the SP variables, we can take the mean of the neighboring states and assign vertex $u$ to the most likely position $j_u$ in the resulting macro-state vector. Two important observations follow:
\begin{itemize}
    \item The $j_u$ is chosen \textbf{deterministically}, as in standard HDPs, because the states have been frozen in advance;
    \item The value of $j_u$ changes at each layer according to the distributions of neighboring states. In other words, $j_u$ is the \textbf{sole} responsible for information propagation at each layer.
\end{itemize}

Formally, to exploit the structural information of the graph and to stick as much as possible to the original \CGMM{} formalism, we chose to select $j_u$ according to this straightforward rule:
\begin{align}
    & j_u = \psi( \boldQ{\N{u}{}}{\lprev}) = \argmax_{j \in \{1,\dots,\bar{C}_{\lcurr}\}} \Big(\frac{1}{|\N{u}{}|}\sum_{v \in \N{u}{}} \boldQ{v}{\lprev}\Big)_j.
    \label{eq:ju}
\end{align}
Thanks to Equation \ref{eq:ju}, vertices with the same features may have a different latent state $c$, due to the fact that they are assigned to different groups, \ie different $\bm{\pi}_j$, on the basis of their neighborhood. Again, this mimics the role of the \CGMM{} transition distribution but in an HDP.

If we wanted to reason using the CRF jargon, Eq. \ref{eq:ju} could be equivalent to have a customer go at the restaurant that was recommended the most by the customer's friends. As in \cite{bacciu_probabilistic_2020}, we chose to average the parameters of the richer distributions $\boldQ{v}{\lprev}, \forall v \in \N{u}{}$ rather than perform majority voting amongst the most likely state of every neighbor. Still, notice that the former choice reduces to the latter when the discrete distributions collapse all their probability mass into a single state (one-hot representation).

Finally, and similarly to \CGMM{}, the very first layer of \iCGMM{} is just an HDP mixture model with one group, as no neighboring states have been inferred yet. The reason why we did not choose a simpler DP mixture model is that an HPD tends to generate a fewer number of latent states in our experiments.

\subsubsection*{On Exchangeability}
Every \iCGMM{} relies on the \textit{exchangeability} assumption of DPs to be trained. Recall that exchangeability (informally) states that the observations $x_u$ of our dataset are not independent but the order in which we look at them does not matter \cite{orbanz_bayesian_2010}. Exchangeability is trivially satisfied in \iCGMM{}, because the observations $X_u$ are assumed to be \iid{} when conditioned on the neighboring states.

Summing up, we depart from the basic \CGMM{} layer in more than one way. First and foremost, we do not parametrize nor learn the \CGMM{} transition distribution, which was responsible for the convex combination of neighboring states when computing the E-step of the EM algorithm. Instead, we rely on the most probable choice of the group $j_u$ that is encoded by the neighbors' macro-state. Secondly, due to the sheer complexity of the Bayesian nonparametric treatment, we do not train the model via EM as done with \CGMM{} and \ECGMM{}; instead, we will exploit Gibbs sampling to compute the quantities of interest. Nonetheless, apart from the conceptual similarities, \iCGMM{} retains one important characteristic of \CGMM{}, \ie it prevents vanishing gradient effects and over-smoothing by default \cite{bacciu_probabilistic_2020}, thus allowing us to construct deeper architectures that freely propagate contextual information.

\subsection{Inference}
\label{subsec:icgmm-inference}
The inference phase determines the latent state of $u$ and updates the \iCGMM{}'s parameters. This happens at each iteration of the HDP Gibbs sampling algorithm \cite{neal_markov_2000,teh_hdp_2006,fox_sticky_2007}. Note that it is also possible to iteratively estimate the hyper-parameters $\alpha_0$ and $\gamma$: whenever that is the case, we shall append a subscript \textit{\quotes{auto}} to our model's name. We start with the latent indicator variable $q_u$, which is sampled from the following conditional probability
\begin{equation}
	\boldqell{u}{\lcurr}(c) = P(q_u=c \mid j_u=j, \bm{Q}^{-u}, \bm{\beta}, \bm{\theta}, \bm{x}) \propto
	(\alpha_0\beta_c+n_{jc}^{-u})f(x_u \mid \theta_c), \quad c \in \{1,\dots,C+1\},\\
	\label{eq:z_sampling}
\end{equation}
where $C$ denotes the number of current states in the mixture model, $f$ is the \textit{p.d.f.} associated with $F$, $\bm{Q}^{-u}$ is the set of latent states assigned so far to each vertex, and the distribution $\bm{\pi}_j$ has been integrated out \cite{teh_hdp_2006}. Here, $n_{jc}^{-u}$ indicates the number of observables associated with latent state $c$ and belonging to group $j$. Whenever we have that $q_u = C+1$, we create a new state and sample a new emission distribution $\bm{\theta}_{C+1}$ from $\bm{H}$. On the contrary, if at the end of an iteration there are no observables associated with state $c$, we can remove that state and decrease $C$ by 1. This is how the HDP, and hence \iCGMM{}, varies in complexity to fit the data distribution.

In the HDP stick-breaking representation, we require an auxiliary variable method to sample the base distribution $\bm{\beta}$ \cite{teh_hdp_2006}. We therefore introduce the auxiliary variables $\bm{m} = \{m_{jc} \mid \forall j \in \{1,\dots,\bar{C}\}, \forall c \in \{1,\dots,C\}\}$ that need to be sampled in order to compute $\bm{\beta}$. However, since $m_{jc}$ is dependent on $n_{jc}$, the sampling step of these variables is very inefficient for large values of $n_{jc}$, as the probability values are proportional the Stirling number of the first-kind $s(n_{jc}, \cdot)$
\cite{fox_stickyhdp_2008}. Luckily, we can avoid this step thanks to the CRF formulation, as anticipated. The key point is that the value $m_{jc}$ corresponds to the number of tables where dish $q_u=c$ is served at group $j$ in the CRF representation \cite{teh_hdp_2006,fox_sticky_2007}; thus, we can compute each $m_{jc}$ by simply simulating the table assignments process in addition to the Stick-breaking machinery.

Knowing that customer $u$ is eating dish $q_u=c$, its table assignment $t_u$ can be sampled according to:
\begin{equation}
	P(t_u=t \mid q_u=c, j_u=j, \bm{c}, \bm{t}^{-u}, \bm{\beta},\alpha_0) \propto
	\begin{cases}
		\tilde{n}^{-u}_{jt}, &\quad \forall t \ \text{s.t.} \ c_{jt}=c;\\
		\alpha_0\beta_c, &\quad t = t_{new},\\
	\end{cases}
	\label{eq:t_sampling}
\end{equation}
where $\bm{t}^{-u}$ represents the tables assigned to each vertex up to now, $c_{jt} \in \bm{c}$ specifies the dish assigned to table $t$ at restaurant $j$ and $\tilde{n}_{jt}^{-u}$ denotes the number of customers (except $u$) sitting at table $t$ of restaurant $j$. Since we know the dish $q_u$ selected by the customer $u$, there is zero probability that the customer sits to a table where that dish is not served. The creation and deletion of tables is very similar to that of Equation \ref{eq:z_sampling}, so we skip it in the interest of the exposition and refer to the pseudocode at the end of the section for a complete treatment. Practically speaking, these auxiliary variables are counters that can be updated in parallel to the Stick-breaking implementation.

After computing $m_{jc}$, \ie $m_{jc}=\sum_{t'} \mathbb{I}[c_{jt'}=c]$, the base distribution $\bm{\beta}$ is updated using:
\begin{equation}
	\bm{\beta} \mid \bm{Q},\bm{m} \sim \text{Dir}(\sum_{j=1}^{\bar{C}}{m_{j1}},\dots,\sum_{j=1}^{\bar{C}}{m_{jC}, \gamma}),
	\label{eq:b_sampling}
\end{equation}
where Dir stands for the Dirichlet distribution and $\bm{Q}$ is the set of latent states assigned to the vertices. The last step of the Gibbs sampling aims to update the emission parameters $\bm{\theta}$ using its posterior given the observable variables:
\begin{equation}
	P(\bm{\theta}_c \mid \bm{Q},\bm{x}) \propto h(\bm{\theta}_c) \prod_{\forall u \mid q_u=c}f(x_u \mid \bm{\theta}_c).
	\label{eq:emission_sampling}
\end{equation}
By choosing the family of the base distribution $\bm{H}$ to be a conjugate prior for $F$, \eg a Dirichlet distribution for Categorical emissions or a Normal-Gamma distribution for Normal emissions, we can compute the posterior in closed form using the usual
data statistics, summarized below.

\subsubsection*{Posterior of the Emission Distribution}
\label{appendix:emission-equations}
We consider the two cases of a discrete and continuous vertex feature.

\paragraph*{Categorical Emission}
Let the emission distribution be a categorical distribution with $K$ possible states. When creating a new state, we can sample the emission parameter according to a Dirichlet distribution, which is a conjugate prior for the categorical distribution:
\begin{equation}
    \theta_c \sim \text{Dir}(\eta,\dots,\eta),
\end{equation}
where the subscript $c$ indicates the new mixture component. Thanks to the conjugate prior, the emission parameters can be updated by sampling its Dirichlet posterior distribution:
\begin{equation}
    \theta'_c \sim \text{Dir}(\eta + N_c^1,\dots,\eta+N_c^K),
\end{equation}
where $N_c^k$ indicates the number of times the observed label $k$ has been associated with the latent state $c$, \ie $N^c_k = \sum_u \mathbb{I}[q_u=c \land x_u=k]$.

\paragraph*{Gaussian Emission}
Similarly to the categorical case, let the emission distribution be an univariate Gaussian. In this case, for each new state, we can sample the emission parameter according to a Normal-Gamma distribution:
\begin{align}
    \mu_c &\sim \mathcal{N}(\mu_0, 1/(\lambda_0\tau_c))\\
    \tau_c &\sim \text{Gamma}(a_0, b_0),
\end{align}
where the subscript $c$ indicates a mixture component ant $\tau_c$ is the inverse of the variance. Then, the emission parameters of the Gaussian can be updated as follows:
\begin{align}
    \mu'_c &\sim \mathcal{N}\left(\frac{\lambda_0\mu_0 + N_c \bar{x}_c}{\lambda_0+N_c}, \frac{1}{(\lambda_0+N_c)\tau'_c}\right)\\
    \tau'_c &\sim \text{Gamma}\left(a_0 + \frac{N_c}{2}, b_0 + \frac{1}{2}\left(N_cs_c+\frac{\lambda_0N_c(\bar{x}_c-\mu_0)^2}{\lambda_0 + N_c}\right)\right),
\end{align}
where $N_c$ indicates the number of observed labels associated with the latent state $c$ (\ie $N_c = \sum_u \mathbb{I}[q_u=c]$), $\bar{x}_c$ is the mean of the data associated with the class $c$ (\ie $\bar{x}_c = \frac{1}{N_c} \sum_{\forall u \mid q_u=c} x_u$), and $s_c$ is the variance of the data associated with the class $c$ (\ie $s_c = \frac{1}{N_c} \sum_{\forall u \mid q_u=c} (x_u - \bar{x}_u)^2$).

\subsubsection*{Sampling $\alpha_0$ and $\gamma$}
Following \citep{teh_hdp_2006}, the concentration parameter $\alpha_0$ and $\gamma$ can be updated between Gibbs sampling iterations by exploiting an auxiliary variable schema. Let us start with the former, by assuming that $\alpha_0$ has a Gamma prior distribution $\text{Gamma}(a, rate=b)$ (\ie $\alpha_0 \sim \text{Gamma}(a, b)$). Then, we define the auxiliary variables $w_1,\dots,w_{\bar{C}}$ and $s_1,\dots,s_{\bar{C}}$, where each $w_j$ variable takes a value between $0$ and $1$, and each $s_j$ is a binary variable. Then, the value of $\alpha_0$ can be sampled according to the following schema:
\begin{align}
    w_j &\sim \text{Beta}(\alpha_0+1, n_{j.}),\label{eq:alpha_sampling_1}\\
    s_j &\sim \text{Bernoulli}\left(\frac{n_{j.}}{n_{j.} + \alpha_0}\right),\label{eq:alpha_sampling_2}\\
    \alpha_0 &\sim \text{Gamma}\left(a+m_{..}-\sum_{j=1}^{\bar{C}}s_j, b-\sum_{j=1}^{\bar{C}}\log w_j\right),\label{eq:alpha_sampling_3}
\end{align}
where $n_{j.}$ is the number of costumer eating in the $j$-th restaurant, and $m_{..}$ is the total number of tables in all the restaurants.

Similarly, assuming that the hyper-parameter $\gamma$ has a gamma prior distribution, \ie $\gamma \sim \text{Gamma}(a',b')$, then its value can be updated by following the auxiliary variable schema below \citep{teh_hdp_2006, fox_stickyhdp_2008}:
\begin{align}
    r &\sim \text{Beta}(\gamma+1,m_{..}),\label{eq:gamma_sampling_1}\\
    p &\sim \text{Bernoulli}\left(\frac{m_{..}}{m_{..}+\gamma}\right),\label{eq:gamma_sampling_2}\\
    \gamma &\sim \text{Gamma}(a'+C-p, b'-\log r). \label{eq:gamma_sampling_3}
\end{align}

After training for a number of Gibbs sampling iterations, we predict the latent states for unseen data points by simply applying Equation \ref{eq:z_sampling}, where all the statistics, \eg $n_{jc} \ \forall j,c$, have been stored after training and never updated again.

To facilitate the practical understanding of our model, Algorithm \ref{alg:gibbs_icgmm} provides the pseudocode of the Gibbs sampling method employed by \iCGMM{}.
\begin{table}[p]
    \begin{minipage}{\textwidth}
        \scriptsize
        \begin{algorithm}[H]
            \caption{Gibbs sampling method for exact \iCGMM{}}\label{alg:gibbs_icgmm}
            \begin{algorithmic}[H]
            \small
            \Require{A dataset of graphs $\mathcal{D} = \{ g_1, \dots, g_N \}$. Initialize $C=1$, $\bm{\theta} = \{\theta_{1}\}$ (where $\theta_{1} \sim H$), $\mathcal{T}_j=\emptyset$ (for all groups $j$), $\bm{q}=\bm{t}=\bm{c}=\bm{\bot}$,  and $\bm{n} = \bm{\tilde{n}} = \bm{0}$.}
            \Repeat
                \For{$g \in \mathcal{D}$}\Comment{For each graph}
                    \For{$u \in \mathcal{V}_{g}$}\Comment{For each vertex}

                        \State \emph{// assign the group}
                        \State $j_u \leftarrow \psi(\boldqell{\N{u}{}}{}) $ \Comment{Can be done once $\forall u$}

                        \vspace{5pt}
                        \State \emph{// assign the dish}
                        \State $n_{j_uq_u}\leftarrow n_{j_uq_u}-1$\Comment{If $q_u \neq \bot$, remove $q_u$ from the counting}
                        \State $q_u \leftarrow \text{SAMPLING}(j_u, \bm{n},\bm{\theta},\bm{x},\bm{\beta},\alpha_0)$\Comment{Sample the dish according to Eq. \ref{eq:z_sampling}}
                        \If{$q_u$ is new}\Comment{Create a new state}
                            \State $\theta_{\text{new}} \sim H$
                            \State $\boldsymbol{\theta} \leftarrow \boldsymbol{\theta} \cup \{\theta_{\text{new}}\}$
                            \State $C \leftarrow C+1$
                            \State $n_{jq_u}\leftarrow0\quad \forall j \in \{1,\dots, \bar{C}\} \,$ \Comment{Initialize the counters}
                        \EndIf
                        \State $n_{j_uq_u}\leftarrow n_{j_uq_u}+1 $\Comment{Update the counter}

                        \vspace{5pt}
                        \State \emph{// assign the table}
                        \State{$\tilde{n}_{j_ut_u} \leftarrow \tilde{n}_{j_ut_u}-1$}\Comment{If $t_u \neq \bot$, remove $t_u$ from the counting}
                        \State $t_u \leftarrow \text{SAMPLING}(j_u, q_u,\bm{c},\bm{\tilde{n}}, \bm{\beta},\alpha_0)$\Comment{Sample the table according to Eq. \ref{eq:t_sampling}}
                        \If{$t_u$ is new}\Comment{Create a new table}
                            \State $\mathcal{T}_j \leftarrow \mathcal{T}_j \cup \{t_u\}$
                            \State$c_{j_ut_u} \leftarrow q_u$\Comment{Save the dish-table assignment}
                            \State $m_{j_uq_u} \leftarrow m_{j_uq_u}+1$ \Comment{Update the table count}
                            \State $\tilde{n}_{j_ut_u} \leftarrow 0$ \Comment{Initialize customer counter}
                        \EndIf
                        \State{$\tilde{n}_{j_ut_u} \leftarrow \tilde{n}_{j_ut_u}+1$}
                    \EndFor
                \EndFor

                \vspace{5pt}
                \State \emph{// remove unused dishes}
                \For{$c \in \{1,\dots,C\}$}
                    \If{$\sum_{j=1}^{\bar{C}}n_{jc} = 0$}\Comment{No customers eat the dish $c$}
                        \State $\bm{\theta} \leftarrow \bm{\theta} \setminus \{\theta_c\}$
                        \State $C \leftarrow C-1$
                    \EndIf
                \EndFor

                \vspace{5pt}
                \State \emph{// remove empty tables}
                \For{$j \in \{1,\dots,\bar{C}\}$}
                    \For{$t \in \mathcal{T}_j$}
                        \If{$\tilde{n}_{jt} =0$}\Comment{No customers eat at the table $t$ in the restaurant $j$}
                            \State $\mathcal{T}_j \leftarrow \mathcal{T}_j \setminus \{t\}$
                            \State $m_{jc_{jt}} \leftarrow m_{jc_{jt}}-1$
                        \EndIf
                    \EndFor
                \EndFor

                \vspace{5pt}
                \State \emph{// update model parameters}
                \State $\bm{\beta} \leftarrow \text{SAMPLING}(\bm{q},\bm{m})$ \Comment{Sample according to Eq. \ref{eq:b_sampling}}
                \State $\bm{\theta} \leftarrow \text{SAMPLING}(\bm{q},\bm{x})$ \Comment{Sample according to Eq. \ref{eq:emission_sampling}}

                \vspace{5pt}
                \If{\iCGMM$_{auto}$}
                    \State $\alpha_0 \leftarrow \text{SAMPLING}(a,b,\bm{n})$
                    \Comment{Sample according to Eq. \eqref{eq:alpha_sampling_1}, \eqref{eq:alpha_sampling_2}, \eqref{eq:alpha_sampling_3}}
                    \State $\gamma \leftarrow \text{SAMPLING}(a',b', \bm{m})$
                    \Comment{Sample according to Eq. \eqref{eq:gamma_sampling_1}, \eqref{eq:gamma_sampling_2}, \eqref{eq:gamma_sampling_3}}
                \EndIf
            \Until{stopping criteria}
            \end{algorithmic}
        \end{algorithm}
    \end{minipage}
\end{table}

\subsection{Faster Inference with Vertex Batches}
\label{subsec:faster-inference}
Due to the sequential nature of the above inference process, a naive implementation is slow when applied to the larger social graphs considered in this thesis.  In the literature, there exist several exact distributed inference methods for the HDP \cite{Lovell12parallelmarkov, Williamson2014, Chang2014, Ge2015}), but their effectiveness might be limited due to the unbalanced workload among workers or the elevated rejection rate \cite{Gal2014pitfalls}. Similarly, there are variational inference approximations \cite{wang_truncation_2012,bryant_truly_2012,hoffman_stochastic_2013, hughes_reliable_2015} that substantially differ from the approach taken here, but their investigation will be subject of future works.

We prefer to speed-up the inference procedure by introducing a straightforward heuristic rather than relying on an exact distributed computation. As suggested in \cite{Gal2014pitfalls}, an approximated inference procedure may indeed suffice for many problems, so what we propose is to perform sampling for a batch of vertex observations altogether. This way, the necessary statistics are updated in batch rather than individually, and matrix operations can be used to gain efficiency.

To keep the quality of the approximation as close as possible to the original Gibbs Sampling algorithm, we choose $1$ graph as the size of the batch. Such a trade-off provides a  CPU speedup of up to $60\times$ at training time, and we empirically observed that performances remain unchanged \wrt{} the original version on the smaller chemical tasks considered so far. While this faster version of \iCGMM{}, which we call \iCGMM{}$_f$, does not strictly adhere to the technical specifications of the previous Section, we believe that the pros largely outperform the cons. Table \ref{tab:speedup-icgmm} reports the speedup gains on some tasks by comparing the same configurations.

\begin{table}[ht]
    \begin{center}
    \begin{tabular}{c l c c }
    \toprule
    & & \iCGMM{} & \iCGMM{}$_f$ \\
    \toprule
    & & \textbf{ref.} & \textbf{min}/\textbf{max} \\
    \cmidrule{3-4}
    \multirow{3}{*}{\rotatebox[origin=c]{90}{\textsc{Chem.}}}
    & D\&D            & 1$\times$ & $17.8\times$/$30.8\times$ \\
    & NCI1          & 1$\times$ & $3.1\times$/$5.1\times$ \\
    & PROTEINS      & 1$\times$ &  $4.2\times$/$5.7\times$ \\
    \midrule
    \multirow{5}{*}{\rotatebox[origin=c]{90}{\textsc{Social}}}
    & IMDB-B    & 1$\times$ & $2.4\times$/$5.1\times$ \\
    & IMDB-M    & 1$\times$ & $1.6\times$/$3.6\times$ \\
    & REDDIT-B  & 1$\times$ & $11.1\times$/$45.6\times$ \\
    & REDDIT-5K & 1$\times$ & $36.7\times$/$60.6\times$ \\
    & COLLAB    & 1$\times$ & $3.1\times$/$8.6\times$ \\
    \bottomrule
    \end{tabular}
    \caption{Approximate minimum and maximum speedup across different configurations between the exact \iCGMM{} and the faster version of \iCGMM{}.}
    \label{tab:speedup-icgmm}
    \end{center}
\end{table}

\subsection{Limitations}
\label{sec:limitations}
It should be clear by now that, due to the complexity of the BNP treatment, one limitation is that naive Gibbs sampling does not scale easily to very large datasets. Yet, the vertex independence assumption made by \CGMM{} enables a faster batch computation, which can also be run on a GPU.

The second limitation of \iCGMM{} is that edge features are not taken into account, differently from \CGMM{} and \ECGMM{}. One of our research directions for the future will be to investigate a potential extension of \iCGMM{} to discrete edge features, perhaps by still using the SP variables in a fully Bayesian fashion.

\subsection{Experimental Setting}
\label{subsec:icgmm-exp}
Similarly to the previous models, we evaluated the performances of \iCGMM{} using the fair, robust, and reproducible evaluation setup for graph classification defined in Section \ref{sec:scholarship-issues}. We first tested the \quotes{exact} and faster Gibbs sampling versions of \iCGMM{} on the three chemical datasets D\&D, NCI1, and PROTEINS. When considering social datasets, instead, we only evaluated \iCGMM{}$_f$ due to its speedup on larger graphs.\footnote{\url{https://github.com/diningphil/iCGMM}.}

We have discussed how \iCGMM{} can automatize the choice its hyper-parameters, \eg the size of the latent representation. In general, the choice of the Bayesian hyper-parameters is much less important than that of the number of states $C$, as in principle one can recursively introduce hyper-priors over these hyper-parameters \citep{bernardo_bayesian_2009,goel_information_1981}. That said, since this is the first work to study HDP methods in the context of graph classification, we both i) explored the hyper-parameter space to best assess and characterize the behaviour of the model and ii) introduced hyper-priors to estimate $\alpha_0$ and $\gamma$ at each layer, thus further reducing the need for an extensive model selection.
For the chemical tasks, the prior $\bm{H}$ over the emission parameters $\bm{\theta}_c$ was the uniform Dirichlet distribution. The range of \iCGMM{} hyper-parameters tried in this case were: number of layers $\in \{5,10,15,20\}$, $\alpha_0 \in \{1,5\}$, $\gamma \in \{1,2,3\}$, unibigram aggregation $\in \{\text{sum}, \text{mean}\}$, and Gibbs sampling iterations $\in \{10, 20, 50\}$. Instead, for the social tasks we implemented a Normal-Gamma prior $\bm{H}$ over a Gaussian distribution. Here the prior is parametrized by the following hyper-priors: $\mu_0$, the mean vertex degree extracted from the data; $\lambda_0$, which is inversely proportional to the prior variance of the mean; and $(a_0, b_0)$, whose ratio $t=\frac{b_0}{a_0}$ represents the expected variance of the data. The \iCGMM{} hyper-parameters here were: number of layers $\in \{5,10,15,20\}$, $\lambda_0 \in \{\text{1e-6}\}$, $a_0\in \{1.\}$, $b_0 \in \{0.09, 1.\}$, $\alpha_0 \in \{1,5,10\}$, $\gamma \in \{2,5,10\}$, unibigram aggregation $\{\text{sum}, \text{mean}\}$, and Gibbs Sampling iterations $\in \{100\}$.
To further automate learning of \iCGMM{}'s unsupervised layers, we place uninformative $Gamma(1, rate=0.01)$ hyper-priors on both $\alpha_0^{\ell},\gamma^{\ell}$ hyper-parameters. To prevent the model from getting stuck in a local minimum on COLLAB (due to bimodal degree distribution and large variances), we tried $\lambda_0 \in \{\text{1e-4},\text{1e-5}\}$.

To conclude, we list the hyper-parameters tried for the one-layer MLP classifier trained on the unsupervised graph embeddings: optimizer $\in \{\text{Adam}\}$, batch size $\in \{32\}$, hidden units $\in \{32, 128\}$, learning rate $\in \{\text{1e-3}\}$, L2 regularization $\in \{0., \text{5e-4}\}$, epochs $\in \{2000\}$, ReLU activation, and early stopping on validation accuracy with patience 300 on chemical tasks and 100 on social ones.

\subsection{Results}
The empirical results on chemical and social benchmarks are reported in Tables \ref{tab:icgmm-chemical-results} and \ref{tab:icgmm-social-results}, respectively. There are several observations to be made, starting with the chemical tasks. First of all, \iCGMM{} performs similarly to \CGMM{}, \ECGMM{}, and most of the \textit{supervised} neural models; this suggests that the selection of $j_u$ based on the neighboring recommendations is a subtle but effective form of information propagation between the vertices of the graph. In addition, results indicate that we have succeeded in effectively automatizing the choice of the number of latent states without compromising the accuracy, which was the main goal of this work. Finally, \iCGMM{}$_f$ performs as well as the exact version, and for this reason we safely applied the faster variant to the larger social datasets (including IMDB-B and IMDB-M to ease the exposition).
\begin{table}[t]
\footnotesize
\centering
\begin{tabular}{l c c c c}
\toprule
     & \textbf{D\&D} & \textbf{NCI1} & \textbf{PROTEINS}\\
\midrule
 \Baseline & $\mathbf{78.4}\pm 4.5 $ &  $69.8 \pm 2.2 $ &  $\mathbf{75.8} \pm 3.7 $ \\
 \DGCNN & $76.6 \pm 4.3 $ &  $76.4 \pm 1.7 $ &  $72.9 \pm 3.5 $   \\
 \DiffPool & $75.0 \pm 3.5 $ &  $76.9 \pm 1.9 $ &  $73.7 \pm 3.5 $  \\
 \ECC & $72.6 \pm 4.1 $ &  $76.2 \pm 1.4 $ &  $72.3 \pm 3.4 $  \\
 \GIN & $75.3 \pm 2.9 $ &  $\mathbf{80.0} \pm 1.4 $ &  $73.3 \pm 4.0 $    \\
 \GraphSAGE & $72.9 \pm 2.0 $ &  $76.0 \pm 1.8 $ &  $73.0 \pm 4.5 $  \\
 \CGMM  & $74.9 \pm 3.4 $ & $ 76.2 \pm2.0$ & $74.0 \pm 3.9$ \\
 \ECGMM & $73.9 \pm4.1$ & $ 78.5 \pm 1.7$ & $73.3 \pm 4.1$   \\
 \midrule
 \iCGMM  & $75.6 \pm 4.3$ & $76.5 \pm1.8$ & $ 72.7 \pm3.4$  \\
 \iCGMM$_f$  & $75.0 \pm 5.6$ & $76.7 \pm1.7$ & $73.3 \pm 2.9$  \\
 \iCGMM$_{auto}$  & $76.3 \pm 5.6$ & $77.6 \pm1.5$ & $ 73.1 \pm3.9$  \\
 \iCGMM$_{f_{auto}}$  & $75.1 \pm 3.8$ & $76.4 \pm1.4$ & $73.2 \pm 3.9$  \\
\bottomrule
\end{tabular}
\caption{Results on chemical datasets (mean accuracy and standard deviation) are shown. The best performances are highlighted in bold.}
\label{tab:icgmm-chemical-results}
\end{table}
\begin{table}[t]
\footnotesize
\centering
\begin{tabular}{lccccc}
\toprule
     & \textbf{IMDB-B} & \textbf{IMDB-M} & \textbf{REDDIT-B} & \textbf{REDDIT-5K} & \textbf{COLLAB}\\
    \midrule
 \Baseline & $70.8 \pm 5.0 $ &  $\mathbf{49.1} \pm 3.5 $ &  $82.2 \pm 3.0 $ &  $52.2 \pm 1.5 $ &  $70.2 \pm 1.5 $   \\
 \DGCNN & $69.2 \pm 3.0 $ &  $45.6 \pm 3.4 $ &  $87.8 \pm 2.5 $ &  $49.2 \pm 1.2 $ &  $71.2 \pm 1.9 $   \\
 \DiffPool & $68.4 \pm 3.3 $ &  $45.6 \pm 3.4 $ &  $89.1 \pm 1.6 $ &  $53.8 \pm 1.4 $ &  $68.9 \pm 2.0 $   \\
 \ECC & $67.7 \pm 2.8 $ &  $43.5 \pm 3.1 $ &  - &  - &  -   \\
 \GIN & $71.2 \pm 3.9 $ &  $48.5 \pm 3.3 $ &  $89.9 \pm 1.9 $ &  $\mathbf{56.1} \pm 1.7 $ &  $75.6 \pm 2.3 $   \\
 \GraphSAGE & $68.8 \pm 4.5 $ &  $47.6 \pm 3.5 $ &  $84.3 \pm 1.9 $ &  $50.0 \pm 1.3 $ &  $73.9 \pm 1.7 $   \\
 \CGMM  & $72.7 \pm 3.6$ & $47.5 \pm 3.9$ & $88.1 \pm 1.9$ & $52.4 \pm 2.2$ & $77.32 \pm 2.2$    \\
 \ECGMM & $70.7 \pm 3.8$ & $48.3 \pm 4.1 $& $89.5 \pm 1.3$ & $53.7 \pm 1.0 $ & $77.45 \pm 2.3$  \\
\midrule
 \iCGMM$_f$  & $\textbf{73.0} \pm 4.3$ & $48.6 \pm 3.4 $ & $91.3 \pm 1.8$ &  $55.5 \pm 1.9$  & $78.6 \pm 2.8$  \\
 \iCGMM$_{f_{auto}}$  & $71.8 \pm 4.4$ & $49.0 \pm 3.8 $ & $\mathbf{91.6} \pm 2.1$ &  $55.6 \pm 1.7$  & $\mathbf{78.9} \pm 1.7$  \\\bottomrule
\end{tabular}
\caption{Results on social datasets (mean accuracy and standard deviation) are shown, where the vertex degree is used as the only vertex feature. The best performances are highlighted in bold.}
\label{tab:icgmm-social-results}
\end{table}

Moving to the social datasets, we observe that \iCGMM{} achieves better average performances than other methods on IMDB-B, REDDIT-B and COLLAB. One possible reason for such an improvement with respect to \CGMM{} variants may be how the emission distributions are initialized. On the one hand, and differently from the chemical tasks, \CGMM{} and \ECGMM{} use the $k$-means algorithm (with fixed $k$=$C$), to initialize the mean values of the $C$ Gaussian distributions, which can be stuck in a local minimum around the most frequent degree values. One the other hand, \iCGMM{} adopts a fully Bayesian treatment, which combined with the automatic selection of the latent states allows to better model outliers by adding a new state when the posterior probability of a data point is too low.

Similarly to the analysis done earlier for \CGMM{}, we will try to shed more light into the improved generalization performances of \iCGMM{}, by analyzing the exact model from a layer-wise perspective.

\begin{figure}[th]
 \centering
 \subfigure[Effect of depth on training/validation accuracy\label{fig:depth-effect-nci1}]{{\includegraphics[width=0.46\columnwidth]{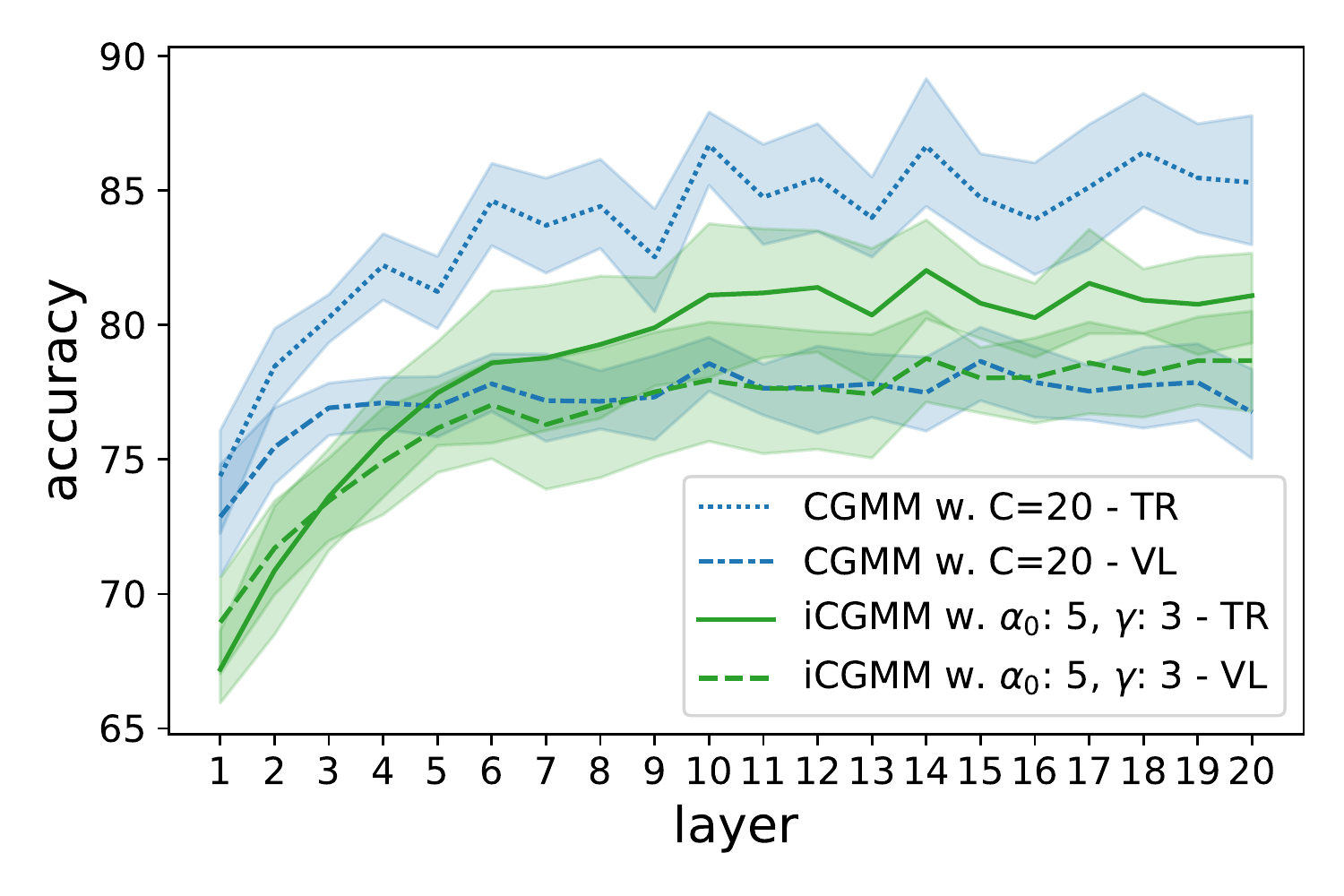} }}%
 \qquad
 \subfigure[Average VL accuracy (solid line) and number of chosen states (dashed line)  w.r.t $\alpha_0$ and $\gamma$ values\label{fig:avg-acc-and-states}]{{\includegraphics[width=0.46\columnwidth]{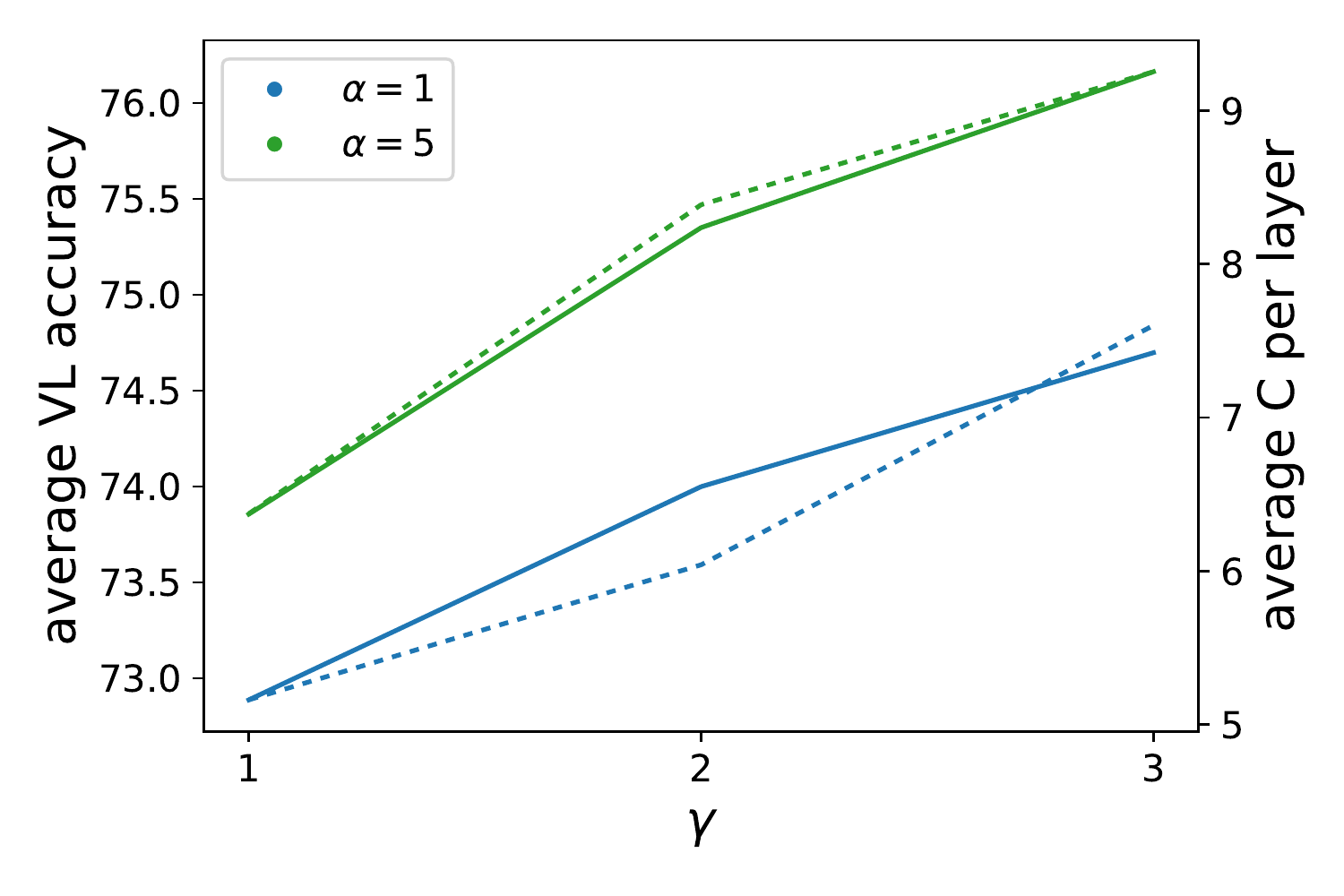}}}\\%
\caption{Figures \ref{fig:depth-effect-nci1} and \ref{fig:avg-acc-and-states} analyze the relation between depth, performances, and the number of chosen states on NCI1.}
 \label{fig:depth-hyper-parameter-study}%
\end{figure}

\subsubsection*{On the effectiveness of depth and hyper-parameters}
To confirm our intuition about the benefits of the proposed information propagation mechanism, Figure \ref{fig:depth-effect-nci1} shows the NCI1 training and validation performances of both \CGMM{} and \iCGMM{} as we add more layers. For simplicity, we picked the best \iCGMM{} configuration on the first external fold, and we compared it against the \CGMM{} configuration with the most similar performances. Note that $C=20$ was the most frequent choice of \CGMM{} states by the best model configurations across the 10 outer folds: this is because having more emission distributions to choose from allows the \CGMM{} model to find better local minima, whereas \iCGMM{} can automatically add states whenever the data point's sampling probabilities are too low.
We trained the same classifier at different depths, and we averaged scores across the 10 outer folds. The validation performance of both models are similar, with an asymptotic behavior as we reach 20 layers; hence, depth remains fundamental to improve the generalization performances \cite{bacciu_probabilistic_2020}. Importantly, we see that gap between \iCGMM{} training and validation scores is thinner than its non-BNP counterpart, suggesting that there is less overfitting of the data.

We now study how \iCGMM{} behaves as we vary the main hyper-parameters $\alpha_0$ and $\gamma$. We continue our experimentation on NCI1; Figure \ref{fig:avg-acc-and-states} depicts the average validation performance and number of states $C$ over all configurations and folds, subject to changes of $\alpha_0$ and $\gamma$ values. The trend indicates how greater values for both hyper-parameters achieve, on average, better validation performance. Also, smaller values of the two hyper-parameters tend to strongly regularize the model by creating fewer states, with consequent reduction in validation accuracy.

\begin{figure}[t!]
 \centering
\subfigure[\# states chosen at each layer\label{fig:chosen-states-nci1}]{{\includegraphics[width=0.46\columnwidth]{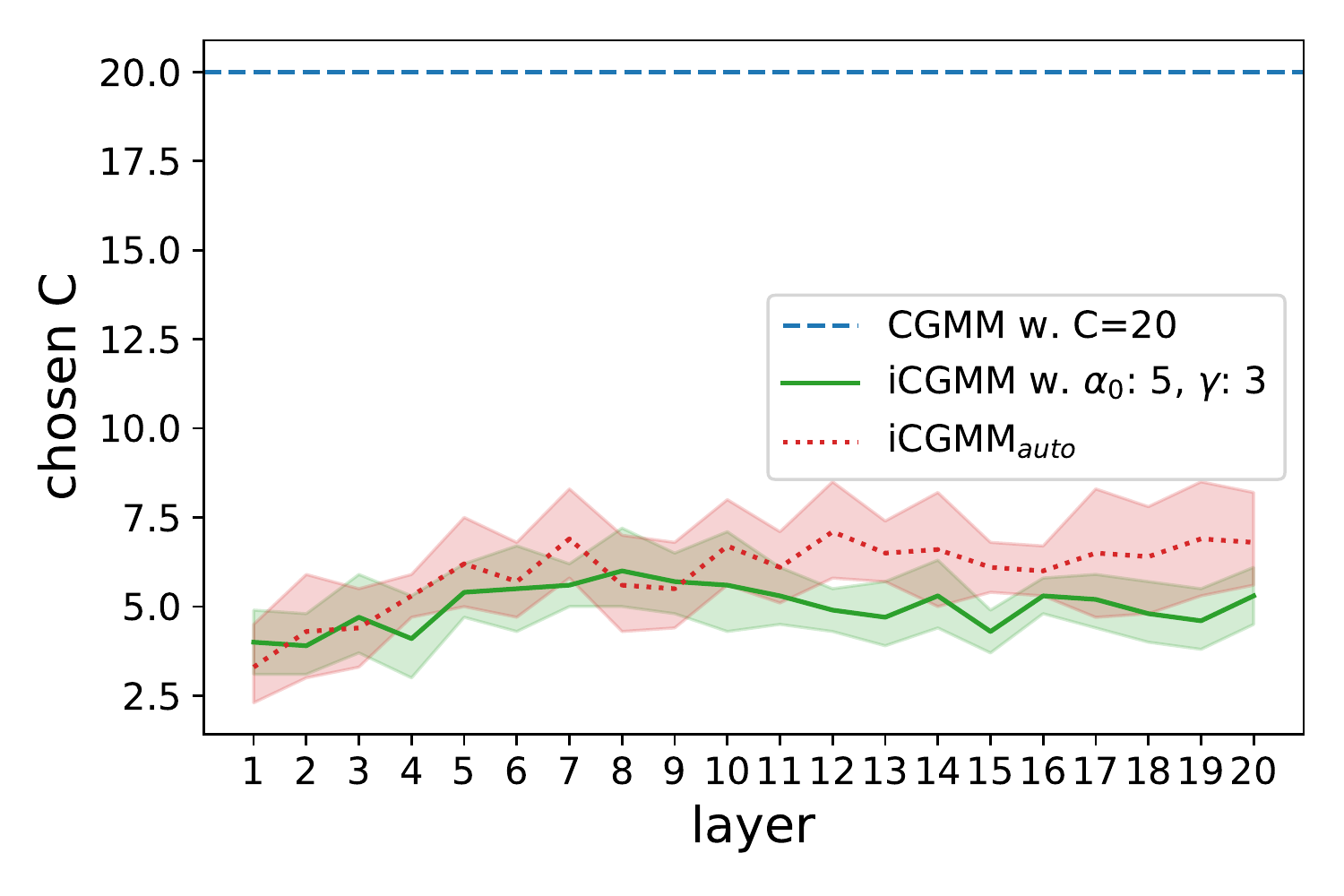}}}%
 \qquad
 \subfigure[Cumulative graph embedding size on NCI1\label{fig:cumulative-embedding-size-nci1}]{{\includegraphics[width=0.46\columnwidth]{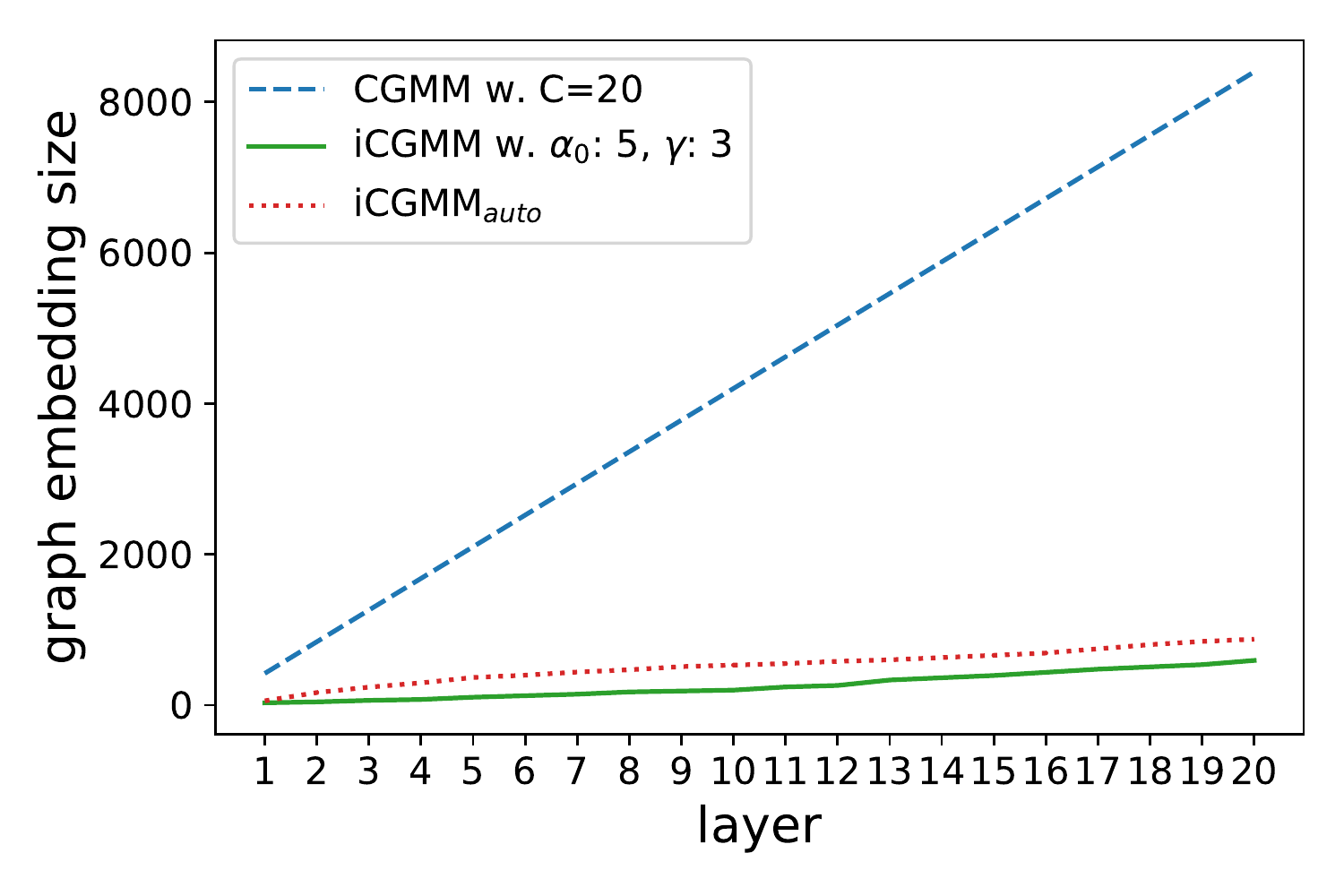}}}%
  \caption{We show comparative results on the size and quality of graph embeddings between \CGMM{} and \iCGMM{}. Overall, \iCGMM{} generates $\approx$ 0 unused latent states, with consequent savings in terms of memory and compute time of the classifier with respect to \CGMM{}. See the text for more details.}%
 \label{fig:icgmm-stability}%
\end{figure}

\subsubsection*{On the quality of graph embeddings}
So far, we have argued that \iCGMM{} selects the appropriate number of states for its unsupervised task at each layer. As a matter of fact, Figure \ref{fig:chosen-states-nci1} reports such a statistic on the same NCI1 configuration as before: \iCGMM{} preferred a lower number of latent states than \CGMM{}, \ie around $5$ per layer.In turn, the resulting graph embeddings become much smaller, with important savings in terms of memory footprint and computational costs to train the subsequent classifier. Figure \ref{fig:cumulative-embedding-size-nci1} displays the cumulative graph embedding size across layers, using the unibigram representation without loss of generality. We see that, when compared with \CGMM{} ($C$=20), the size of graph embeddings produced by \iCGMM{} is approximately 7\% of those of the original model, while still preserving the same performance as \CGMM{}.

\subsubsection*{On the automatic estimation of $\alpha^{\ell}$ and $\gamma^{\ell}$}
We conclude this work with a performance analysis of the fully automated versions of \iCGMM{} and \iCGMM{}$_f$, namely those with an \textit{\quotes{auto}} subscript in Tables \ref{tab:icgmm-chemical-results} and \ref{tab:icgmm-social-results}; in particular, we observe no statistically significant performance differences with respect to the original models. By estimating all hyper-parameters of our models using uninformative priors, we almost always (but for COLLAB) managed to \textit{avoid the model selection} for the unsupervised graph embeddings creation. In turn, this amounted to a $6\times$ reduction in the overall number of configurations to be tried, but most importantly it frees the user from making hard choices about which configurations of hyper-parameters to try. Also, we observe that the number of chosen states and the consequent graph embedding size is very similar to that of \iCGMM{} with $\alpha_0=5,\gamma=3$, but this time the two hyper-parameters have been estimated by the model on the basis of the data. \\

\subsection{Summary}
The Infinite Contextual Graph Markov model is the last methodological contribution of the chapter. We have shown how to bridge the two distant fields of Bayesian nonparametrics and deep learning for graphs in order to build a DBGN whose complexity grows with the data. \iCGMM{} has demonstrated very competitive performances with respect to the (supervised) state of the art, thanks to an information propagation mechanism that is inspired from the concepts of Chapter \ref{chapter:gentle-introduction} but adapted to work with HDPs. Not only does this model automatically select the number of hidden states for each layer, but we can also estimate almost all hyper-parameters at each layer using uninformative hyper-priors. In turn, we can get lower memory and computational footprints without sacrificing the overall predictive performances, at least in the tasks studied so far.

It still remains to be seen whether or not more complex aggregation mechanisms could be applicable to \iCGMM{}. Our attempts at choosing the group $j$ for each observation in a stochastic way, \ie by sampling from the macro-state distribution at each Gibbs sampling iteration, failed to converge or performed poorly. Moreover, there is the necessity to scale up to larger graphs, which may be achieved by distributed Gibbs sampling or variational inference procedures. We leave these interesting directions to future works, confident that the cross-fertilization of ideas between different fields will further enhance the representational power of Deep Bayesian Graph Networks.

\clearpage
\section[Application to Malware Classification]{Application to Malware Classification \cite{errica_robust_2021}}
\label{sec:application-security}

To conclude the chapter, we tackle a real-world malware classification problem using the DBGNs introduced so far \cite{errica_robust_2021}. The task of detecting malicious behavior using static analysis is indeed one fundamental process to protect devices, networks and users' personal data. By looking at how the program is written, we want to automatically find patterns that allow us to distinguish whether a program is to be \textit{trusted} or belongs to a specific \textit{malware} family.

As anti-malware companies become better at finding known patterns, so do malware writers that rely on \textit{obfuscation} techniques to elude common pattern checks. There are two main categories of obfuscation: \textbf{intra-procedural}, \ie it modifies procedural code without changing the interaction with the rest of the program, or \textbf{inter-procedural}, \ie it alters the structure of the program by also adding \textit{call} or \textit{invoke} statements. While the latter is certainly more difficult to detect, it is also much more delicate to use as some mechanisms (\eg call-return and parameter passing) may rely on information known only at run-time and can introduce concurrency problems.

Intra-procedural techniques are widely used and suffice to fool a number of static code analysis tools \cite{bacci2018detection}. Recent works test their approaches on the most common obfuscation techniques \cite{suarez2017droidsieve} or group them by their magnitude of edits on the code \cite{maiorca2015stealth}. In this context, we investigate the problem of malware classification using DBGNs, where the program is represented as a \textbf{Call Graph} (CG), \ie a graph where vertices are procedures and edges denote calls to other procedures. Differently from the literature, we consider obfuscation techniques based on their influence on the CG topology.

Many non-adaptive malware detection solutions based on CGs exploit graph-signatures, similarity algorithms and graph-kernels \cite{shang2010detecting, elhadi2014enhancing,gascon_structural_2013}. In conjunction with formal methods, these approaches achieve excellent accuracy, but the analysis is time-consuming and requires domain-level expertise for the temporal logic formulas generation \cite{iadarola2020call}. Instead, most \ML{} approaches are generally more efficient and rely on static analysis features included in the graphs, such as opcodes frequencies \cite{canfora2015effectiveness} and control/data dependencies \cite{kapoor2016control}, some of which are easily vulnerable to intra-procedural obfuscation.

Our contribution, apart from showing a practical application of the methodologies introduced so far, is to propose a malware classification method based solely on the \textbf{CG topology}. This way, it is possible to show that the approach is intrinsically robust to intra-procedural obfuscation techniques. We exploit CGMM, E-CGMM, and iCGMM to construct CG embeddings that are then fed to a \ML{} classifier. Note that, while methods exist to certify robustness of DGNs to vertex perturbations \cite{zugner_certifiable_2019}, our approach does not need such certificates as it only focuses on the structure.

\subsection{Methodology}
\begin{figure}[ht]
\centering
\includegraphics[width=\textwidth]{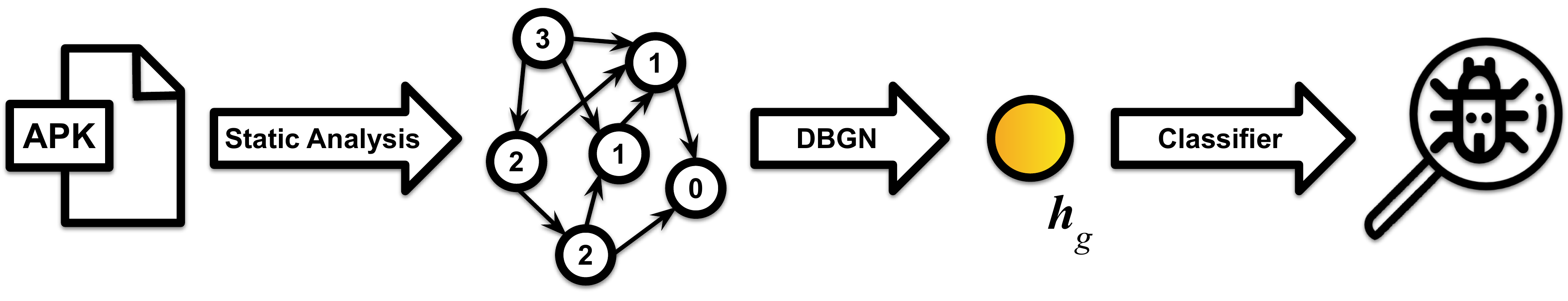}
\caption{Given an Android Application Package (APK) \wlog{}, we apply static analysis to construct a CG, where vertices represent methods and arrows denote how methods are intertwined. For our purposes, the sole vertex feature we use is the out-degree of each vertex. Then, the CGMM model transforms the input graph into an embedding that is used for the final classification.}
\label{fig:cnr-methodology}
\end{figure}

We sketch the overall methodology in Figure \ref{fig:cnr-methodology}. Assuming we have a CG dataset, we employ an unsupervised DBGN to generate graph embeddings encoding CG structural information.
We already know from the previous sections that, to enable learning with the current DBGNs, vertex features are essential. To avoid relying on features vulnerable to intra-obfuscation, we choose the out-degree of each vertex, which encodes how many calls are made by the caller. Therefore, the emission distribution $P(x_u | Q_u = i)$ will be a univariate Gaussian. Clearly, using the degree feature is just one the possible choices, but it proved to be quite effective so far.

It is worth that the unsupervised training can significantly accelerate the model selection phase, since graph embeddings need be computed only once and the downstream classifier works on simple vectors. Also, the final graph embedding is (again) the concatenation of the aggregated vertex/edge posteriors produced at each layer of the DBGNs developed in this thesis.

\subsection{Experimental Setting}

We now describe how we converted a set of Android applications, \ie \path{.apk} files, into a CG dataset; nevertheless, provided a static analysis tool is available, it is straightforward to apply this methodology to other environments as well. First, of all, each \path{.apk} file is decompressed and the Java bytecode is decompiled into Jimple, an intermediate representation language, using the Soot Framework \cite{vallee2010soot}. During decompilation, the code is analyzed to generate a CG\footnote{Soot transformation: \url{https://github.com/Djack1010/graph4apk}.}, where  vertices represent methods, \ie a procedure or function construct, and \textbf{directed} edges denote calls from caller to called  vertices, \ie when an \textit{invoke} or \textit{call} statement is present in the method. Our analysis only considers methods in the application packages, thus discarding calls to library functions or external packages. Notably, the generated CGs do not contain information about the methods statements, \eg variables, declaration, and dependencies, on the  vertices; instead, as already mentioned, we add the out-degree as the sole vertex feature to be able to train the probabilistic models. Hence, our methodology is intrinsically robust to intra-procedural obfuscations techniques, such as code reordering/removal, junk code insertion, instruction substitution, control flow modifications, identifiers and variables renaming/encryption, and repacking  \cite{bacci2018detection}. Indeed, these obfuscation techniques modify each method's statements but they do not alter the number of \textit{invoke} or \textit{call} statements, \ie the initial CG is exactly the same as any intra-procedurally obfuscated CG.

Malware samples were collected from the AMD and previous work datasets \cite{iadarola2021towards}, and the benign samples were downloaded from Google Play. Both the malware and the trusted applications were verified with VirusTotal, to ensure either their maliciousness or trustiness. The resulting dataset consists of 5669 samples of real-world malware, split into 8 classes, where one represents the trusted software (1762 samples) and the others stand for different malware families, namely \textit{Airpush} (736 samples), \textit{Dowgin} (1040 samples), \textit{FakeInst} (190 samples), \textit{Kuguo} (879 samples), \textit{Youmi} (959 samples), \textit{Fusob} (73 samples), and \textit{Mecor} (30 samples). Dataset statistics are described in Table \ref{tab:cnr-dataset}.
\begin{table}[ht]
\centering
\small
\begin{tabular}{ccccccc}
\hline
\# Graphs & \# Classes & Avg $|\Vset{g}|$ & Avg $|\Eset{g}|$ & Min Degree & Max Degree & Avg Degree \\ \hline
5669               & 8                   & 5069             & 3267             & 0                & 618              & 0.58             \\ \hline
\end{tabular}
\caption{Dataset statistics. Graphs are large but sparse, and the average out-degree is low because all calls to external libraries have been removed from the CG.}
\label{tab:cnr-dataset}
\end{table}

To assess the performance of the DBGNs on our CG dataset, we split the data according to a stratified hold-out strategy, with 80\% of the data for training, 10\% for validation and 10\% for test.\footnote{\url{https://github.com/diningphil/robust-call-graph-malware-detection}.} To empirically evaluate the impact of the structure in the dataset, we follow Section \ref{sec:scholarship-issues} and introduce a structure-agnostic baseline. The baseline applies an MLP to the vertex features, performs global aggregation and then applies a linear output layer. We performed grid-search model selection for all models, with early stopping monitoring the classification accuracy.

The hyper-parameters tried for the baseline were: hidden units $\in$ \{32,64,128\}, 2000 epochs, batch size 128, global aggregation $\in$ \{sum, mean\}, Adam Optimizer with learning rate $\in$ \{0.01, 0.001\}, patience $\in$ \{50\}. As regards CGMM, instead, we selected the best model across the following configurations: 20 states, layers $\in$ $\{10, 20\}$, 10 EM epochs, posterior version $\in$ \{discrete, continuous\}, embedding version $\in$ \{unigram, unibigram\}, global aggregation $\in$ \{sum, mean\}, batch size 64, 2000 epochs, hidden units $\in$ \{32,64,256, 512\}, Adam Optimizer with learning rate $\in$ \{0.0001\} and weight decay $\in$ \{0., 0.0005\}, and patience 100. \ECGMM{} shares the same hyper-parameters' set but for $C_E$ in $\{5, 10\}$. Finally, we tried the same \iCGMM{}$_{f}$ and \iCGMM{}$_{f_{auto}}$ embedding configurations of Section \ref{subsec:icgmm-exp} (but for $\lambda_0 \in \{\text{1e-4,1e-5}\}$), with the $\mu_0$ being set to $0.58$, \ie the empirical mean out-degree of the dataset. However, we used the same classifier's configurations as the two models above.

\subsection{Results}

\begin{table}[t]
    \centering
    \small
    \begin{tabular}{lcccccc}
    \hline
          & \textsc{TR Loss} & \textsc{TR Acc.} & \textsc{VL Loss} & \textsc{VL Acc.} & \textsc{TE Loss} & \textsc{TE Acc.} \\ \hline
         \textsc{Baseline} & $1.2_{\pm 0.05}$ & $55.6_{\pm 0.5}$ & $1.1_{\pm 0.01}$ & $60.6_{\pm 0.9}$ & $1.1_{\pm 0.03}$ & $56.7_{\pm 0.5}$ \\
         \CGMM{} & $0.01_{\pm 0.01}$ & $99.8_{\pm 0.4}$  & $0.16_{\pm 0.01}$ & $97.9_{\pm 0.2}$ & $0.13_{\pm 0.01}$ & $96.4_{\pm 0.6}$ \\
         \ECGMM{} & $0.03_{\pm0.02}$ & $99.4_{\pm0.6}$ & $0.59_{\pm0.003}$ & $98.4_{\pm0.3}$ & $0.19_{\pm0.02}$ & $\mathbf{97.3}_{\pm 0.4}$ \\
          \iCGMM{}$_f$ & $0.05_{\pm0.01}$ & $98.7_{\pm0.5}$ & $0.27_{\pm0.04}$ & $94.8_{\pm0.5}$ & $0.35_{\pm0.03}$ & $93.6_{\pm0.6}$ \\
          \iCGMM{}$_{f_{auto}}$ & $0.07_{\pm0.03}$ & $97.93_{\pm0.9}$ & $0.25_{\pm0.02}$ & $95,8_{\pm0.5}$ & $0.42_{\pm0.1}$ & $92.7_{\pm0.5}$ \\ \hline
    \end{tabular}
    \caption{Malware classification results (mean and standard deviation) on training (TR), validation (VL) and test (TE) sets. We display both the Cross-Entropy loss as well as the multi-class accuracy. Results are averaged over 3 final runs.}
    \label{tab:results}
\end{table}
\begin{figure}[ht]
\begin{center}
\centerline{\resizebox{0.8\textwidth}{!}{\includegraphics{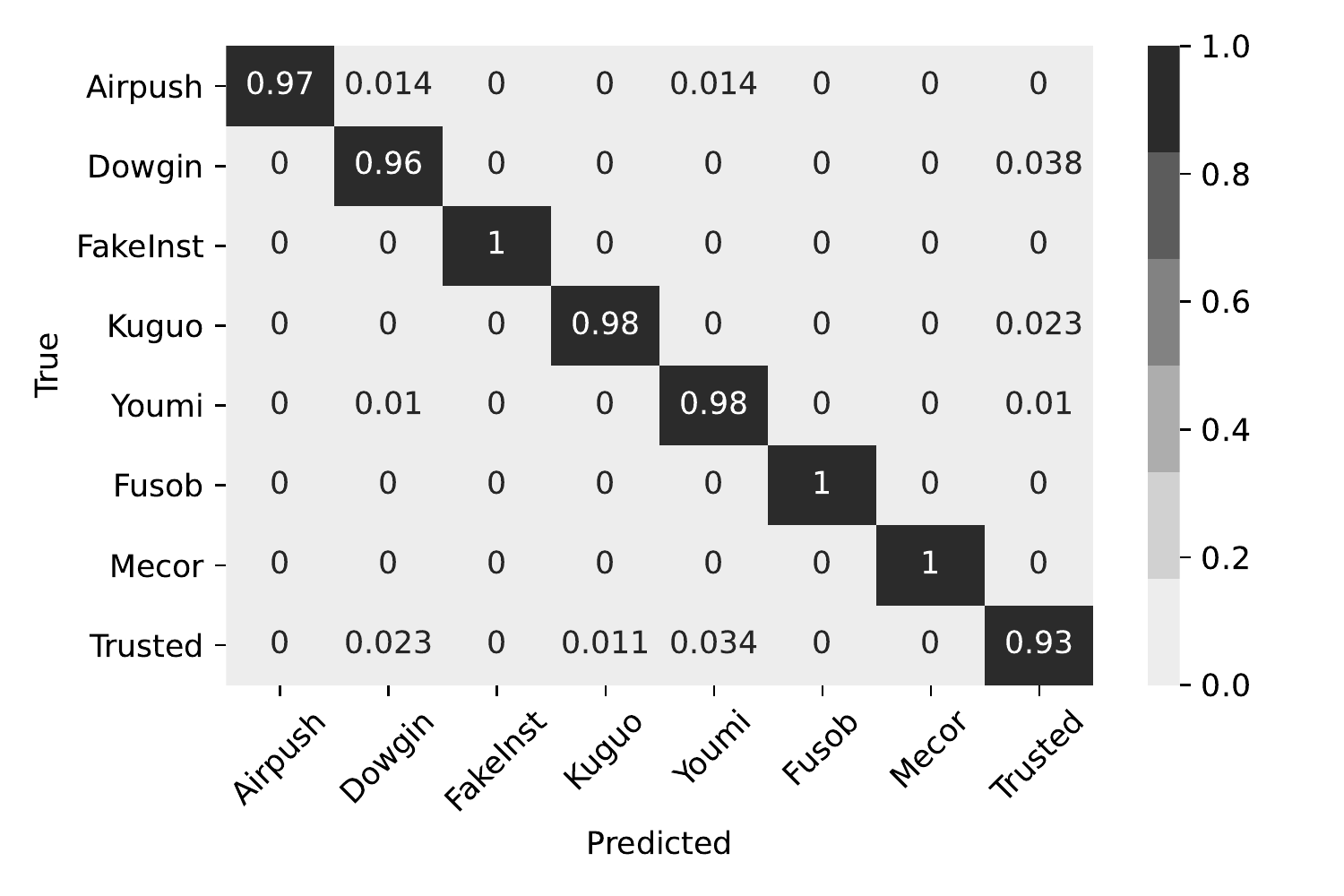}}}
\vspace{-0.5cm}
\caption{Row-normalized confusion matrix of CGMM computed on the test set.}
\label{fig:confusion-matrix}
\end{center}
\vspace{-1cm}
\end{figure}
Results are shown in Table \ref{tab:results}. As we can see, the structural variability in the dataset is such that a structure-agnostic baseline cannot accurately classify instances by merely looking at the out-degree statistics of the graph. Instead, the three DBGNs are able to extract structural patterns that allow the subsequent classifier to achieve a very good accuracy on the test set (and Macro F1 score of approximately 97\% for the best performing solution). Here, it seems that the configurations tried for \iCGMM{}$_{f}$ and \iCGMM{}$_{f_{auto}}$ led to a slight underfitting \wrt{} the other two models, which is probably due to the chosen ranges of hyper-parameters and hyper-priors for the peculiar out-degree distribution of the dataset.

These results support our knowledge that different malware families share detectable topological similarities and hence our hypothesis on the robustness of the approach to intra-procedural obfuscations. In fact, we were able to accurately detect such similarities without relying on non-adaptive procedures, domain expertise, and static analysis' vertex features that are susceptible to obfuscation techniques. In addition, the confusion matrix of Figure \ref{fig:confusion-matrix} shows how accuracy does not decrease for the most imbalanced classes, \eg Fusob and Mecor. Rather, the classifier achieves perfect classification on those test samples. Finally, to empirically confirm that the proposed approach is robust to intra-procedural obfuscation methods, we also performed inference with \CGMM{} on an obfuscated subset of test malwares (261 out of 391, due to intrinsic difficulties in the process, \eg sometimes the obfuscated code did not compile) using the Code Reordering and Junk Code techniques in \cite{bacci2018detection}. \CGMM{} achieved a 99.6\% accuracy.

\section{Summary}
We have introduced Deep Bayesian Graph Networks, a probabilistic alternative to Deep Neural Graph Networks to extract information from graphs of varying size and topology. Throughout the chapter, we have shown how it is possible to formalize the main building blocks of Deep Graph Networks using the well known tools of Bayesian inference. Our contributions rely on an incremental construction to break the mutual (and possibly cyclic) dependencies between latent variables associated with vertices, and the quality of the unsupervised embeddings is such that we managed to compete with state of the art supervised DNGNs on classification and regression tasks.

It may be worth mentioning that the probabilistic framework presented in this thesis is general enough to be extended in many ways, from the development of a \quotes{supervised \CGMM{}} to the introduction of attention and more general aggregation mechanisms. In a sense, \ECGMM{} and \iCGMM{} are examples of architectural and Bayesian nonparametric extensions of the basic \CGMM{}, respectively, but it is not difficult to imagine variations of the graphical model that take advantage of variational bounds or automatic estimation of other hyper-parameters.

What is more, we still have to investigate the scenario in which DBGNs should excel at, namely pre-training of vertex/graph embeddings on a huge amount of \textbf{unlabelled} raw data. This was mainly due to the absence of large datasets which, however, are becoming more and more available these days \cite{hu_open_2020}.

\chapter{Graph Mixture Density Networks}
\label{chapter:hybrid}

\epigraph{\textit{La faccia sua era faccia d’uom giusto, \\ tanto benigna avea di fuor la pelle, \\ e d’un serpente tutto l’altro fusto;}}{\textit{Inferno - Canto XVII}}

In this final methodological chapter, we aim at building a \textbf{hybrid} model that gets the best of the two worlds presented so far, namely neural and Bayesian networks, in the context of deep learning for graphs. More specifically, our contribution is motivated by the need of modeling multimodal output distributions conditioned on topologically varying graphs: in this respect, we are extending the Mixture Density Network model to the processing of structured-data. What we call Graph Mixture Density Network (\GMDN{}) \cite{errica_graph_2021} is basically the combination of a graph encoder, \eg a DGN, and yet another conditional mixture model. This time, however, the overall architecture is a feedforward (but not constructive) DNGN. We shall present practical reasons as to why such a model is necessary, and we will formalize learning withing the framework of Generalized Expectation Maximization. A \GMDN{} is particularly suited to express uncertainty about the possible continuous output values associated with an input graph. \GMDN{} can tackle predictions of stochastic events, like the final outcome of an epidemic given the initial network, but it also can be applied to standard regression problems to better understand the data at hand. We complement the discussion with an alternative way to solve link prediction problems, using a measure of distance between two vertices' multimodal distributions. All in all, we will see how \GMDN{} can be a useful tool to \textit{i)} better analyze the data, as uncertainty usually arises from stochasticity, noise, or under-specification of the system of interest, and \textit{ii)} train Deep Graph Networks which can provide further insights into their predictions and their trustworthiness.

\section{Motivations}
In Section \ref{subsec:mdn}, we discussed how the classical assumptions we make in regression problems do not hold anymore when the output distribution is multimodal. The Mixture Density Network \cite{bishop_mixture_1994} was designed to produce multimodal target distributions, but the input data has to be of vectorial nature.

In terms of applications, MDNs have been recently applied to epidemic simulation prediction \cite{davis_use_2020}. The goal is to predict the multimodal distribution of the total number of infected cases under a compartmental model such as the \textbf{stochastic} Susceptible-Infectious-Recovered (SIR) model \cite{kermack_contribution_1927}. With SIR, each individual in the network can be in one of the three states (S,I or R), and there are very simple update rules to transition from one state to another, depending on the connectivity of the network and two parameters: i) infectivity $\beta$ and ii) recovery $\gamma$. In the paper \cite{davis_use_2020}, the authors show that, given samples of SIR simulations with different infectivity and recovery parameters, the MDN could approximate the output distribution using a mixture of binomials. This result is a remarkable step in approximating way more complex compartmental models in a fraction of the time originally required, similarly to what has been done, for example, in material sciences \cite{pilania_accelerating_2013} and molecular biosciences \cite{errica_deep_2021}. However, the work of \cite{davis_use_2020} makes the strong assumption that the infected network is a complete graph. In fact, as stated in \cite{opuszko_impact_2013}, arbitrary social interactions in the network play a fundamental role in the spreading of a disease, so predictive models should be able to take the topology into account \cite{valenchon_multiple_2019}.

Throughout this thesis, we had the chance to see that many real-world problems are best solved with relational data, where the structure substantially impacts the possible outcomes. For these reasons, we shall propose a hybrid approach to handle multimodal target distributions conditioned on graphs, namely the Graph Mixture Density Network (\GMDN{}). This model can output distributions for either the whole structure or its individual entities. We shall use the likelihood as a metric for this kind of conditional density estimation tasks \cite{nowicki_estimation_2001}, since it tells us how well the model is fitting the empirical data distribution. Overall, \GMDN{} extends the capabilities of all DNGNs whose output is restricted to unimodal distributions.

\clearpage
\section[Model Definition]{Model Definition \cite{errica_graph_2021}}

We aim to learn the conditional distribution $P(y_g|g)$, with $y_g$ being the continuous target label(s) associated with an input graph $g$ in the dataset $\dataset{}$\footnote{Note that the process to output vertex-specific distributions is almost identical, with the exception that global aggregation is not performed.}. We assume the target distribution to be multimodal, and as such it cannot be well modeled by current DGNs.

\begin{figure*}[ht]
    \hspace{-1.5cm}
    \centering
    \resizebox{\textwidth}{!}{\input{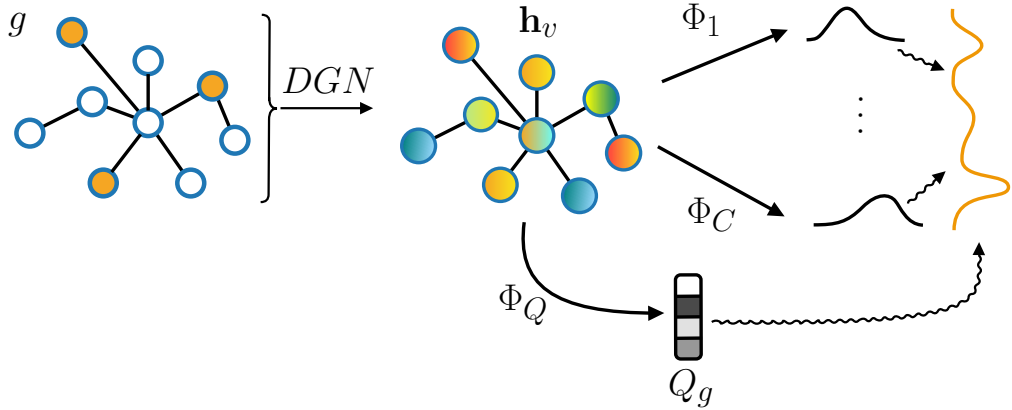}}
    \caption{From a high-level perspective, we first obtain a state $\boldhell{v}{}$ for each vertex by applying an isomorphic transduction with a DGN encoder. Then, for each graph $g$, a subsequent transformation $\Phi_Q$ generates the mixing probability vector $Q_g \in [0,1]^C$ that combines the $C$ different distributions produced by the sub-networks $\Phi_1,\dots,\Phi_C$.}
    \label{fig:gmdn-high-level}
\end{figure*}

As sketched in Figure \ref{fig:gmdn-high-level}, we seek a DGN that performs an isomorphic transduction of the graph to obtain vertex representations $\boldhell{\Vset{g}}{}$ as well as a set of mixing weights $Q_g \in [0,1]^C$ that sum to 1, where $C$ is the number of unimodal output distributions we want to mix. Given $\boldhell{\Vset{g}}{}$, we then apply $C$ different sub-networks $\Phi_1,\dots,\Phi_C$ that produce the parameters $\theta_1,\dots,\theta_C$ of $C$ output distributions, respectively.

In principle, we could mix distributions from different families, but this poses several issues, such as finding a rationale for their choice and choosing how many of them to use for each family. In light of these considerations, we stick to a single family for simplicity of exposition. Finally, combining the $C$ unimodal output distributions with the mixing weights $Q_g$ produces a multimodal output distribution.

\begin{figure}[ht]
    \centering
    \resizebox{0.4\textwidth}{!}{\tikzset{every picture/.style={line width=0.75pt}} 

\begin{tikzpicture}[x=0.75pt,y=0.75pt,yscale=-1,xscale=1]

\draw [line width=1.5]    (211.8,181) -- (149.69,108.78) ;
\draw [shift={(147.08,105.75)}, rotate = 409.3] [fill={rgb, 255:red, 0; green, 0; blue, 0 }  ][line width=0.08]  [draw opacity=0] (11.61,-5.58) -- (0,0) -- (11.61,5.58) -- cycle    ;
\draw [line width=1.5]    (211.8,181) -- (157,181.22) ;
\draw [shift={(153,181.23)}, rotate = 359.77] [fill={rgb, 255:red, 0; green, 0; blue, 0 }  ][line width=0.08]  [draw opacity=0] (11.61,-5.58) -- (0,0) -- (11.61,5.58) -- cycle    ;
\draw [line width=1.5]    (128,181.23) -- (128,118.23) ;
\draw [shift={(128,114.23)}, rotate = 450] [fill={rgb, 255:red, 0; green, 0; blue, 0 }  ][line width=0.08]  [draw opacity=0] (11.61,-5.58) -- (0,0) -- (11.61,5.58) -- cycle    ;
\draw  [fill={rgb, 255:red, 255; green, 255; blue, 255 }  ,fill opacity=1 ][line width=1.5]  (103,181.23) .. controls (103,167.43) and (114.19,156.23) .. (128,156.23) .. controls (141.81,156.23) and (153,167.43) .. (153,181.23) .. controls (153,195.04) and (141.81,206.23) .. (128,206.23) .. controls (114.19,206.23) and (103,195.04) .. (103,181.23) -- cycle ;
\draw  [fill={rgb, 255:red, 31; green, 119; blue, 180 }  ,fill opacity=1 ][line width=1.5]  (103,89.23) .. controls (103,75.43) and (114.19,64.23) .. (128,64.23) .. controls (141.81,64.23) and (153,75.43) .. (153,89.23) .. controls (153,103.04) and (141.81,114.23) .. (128,114.23) .. controls (114.19,114.23) and (103,103.04) .. (103,89.23) -- cycle ;
\draw  [line width=1.5]  (89.67,60.33) -- (277.08,60.33) -- (277.08,322.42) -- (89.67,322.42) -- cycle ;
\draw [line width=1.5]    (128,265.23) -- (212.88,265.19) -- (212.52,202.73) ;
\draw [shift={(212.5,198.73)}, rotate = 449.68] [fill={rgb, 255:red, 0; green, 0; blue, 0 }  ][line width=0.08]  [draw opacity=0] (11.61,-5.58) -- (0,0) -- (11.61,5.58) -- cycle    ;
\draw  [fill={rgb, 255:red, 31; green, 119; blue, 180 }  ,fill opacity=1 ][line width=1.5]  (103,265.23) .. controls (103,251.43) and (114.19,240.23) .. (128,240.23) .. controls (141.81,240.23) and (153,251.43) .. (153,265.23) .. controls (153,279.04) and (141.81,290.23) .. (128,290.23) .. controls (114.19,290.23) and (103,279.04) .. (103,265.23) -- cycle ;
\draw  [fill={rgb, 255:red, 31; green, 119; blue, 180 }  ,fill opacity=1 ][line width=1.5]  (194.22,173.42) .. controls (194.22,167.89) and (198.69,163.42) .. (204.22,163.42) -- (219.38,163.42) .. controls (224.91,163.42) and (229.38,167.89) .. (229.38,173.42) -- (229.38,188.58) .. controls (229.38,194.11) and (224.91,198.58) .. (219.38,198.58) -- (204.22,198.58) .. controls (198.69,198.58) and (194.22,194.11) .. (194.22,188.58) -- cycle ;

\draw (128,181.23) node  [font=\LARGE]  {$Q_{g}$};
\draw (128,89.23) node  [font=\LARGE,color={rgb, 255:red, 255; green, 255; blue, 255 }  ,opacity=1 ]  {$Y_{g}$};
\draw (128,265.23) node  [font=\LARGE,color={rgb, 255:red, 255; green, 255; blue, 255 }  ,opacity=1 ]  {$G$};
\draw (212.2,182.8) node  [font=\LARGE,color={rgb, 255:red, 255; green, 255; blue, 255 }  ,opacity=1 ]  {$\boldsymbol{h}_{\mathcal{V}_{g}}$};
\draw (236.63,306) node  [font=\LARGE]  {$g\ \in \ \mathcal{D}$};

\end{tikzpicture}}
    \caption{\GMDN{}'s Bayesian network is almost identical to that of a Mixture Density Network (see Figure \ref{fig:mdn-graphical-model}), with the exception that the input observable has distribution with support over graphs rather than flat data.}
    \label{fig:gmdn-graphical-model}
\end{figure}
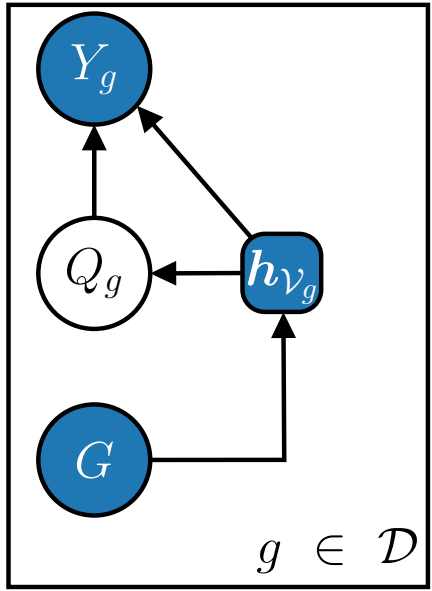

More formally, we learn the conditional distribution $P(y_g|g)$ using the Bayesian network of Figure \ref{fig:gmdn-graphical-model}. We solve the CDE problem via maximum likelihood estimation, which reflects the probability that an output $y$ is generated from a graph $g$. Given an hypotheses space $\mathcal{H}$, we therefore seek the following hypothesis:
\begin{align*}
& h_{MLE} =  \argmax_{h\in\mathcal{H}}P(\dataset{}|h) = arg\max_{h\in\mathcal{H}} \prod_{g\in\dataset{}}\sum_{i=1}^C P(y_g|Q_g=i, g)P(Q_g=i|g),
\end{align*}
where we introduced the latent variable $Q_g$ via marginalization. We model the distributions in the equation by means of DGNs, given that a graph $g$ may have a variable number of vertices and edges. In this respect, we choose the convolution of the Graph Isomorphism Network (GIN) \cite{xu_how_2019} in our experiments. Also, the final vertex representation $\boldhell{v}{}$ is given by the concatenation of all $L$ intermediate states, where $L$ is the chosen number of layers.

Since we care about producing a single graph-related distribution, representations $\boldhell{\Vset{g}}{}$ have to be globally aggregated with a function $\Psi_g$
\begin{align*}
    \boldhell{g}{} = r_g(\boldhell{\Vset{g}}{})=\Psi_g \Big( \{ f_r(\boldhell{v}{}) \mid v \in \Vset{g} \} \Big),
\end{align*}
where $f_r$ could be a linear model or an MLP. Likewise, the mixing weights can be computed using a function $r^Q_g$ as follows:
\begin{align*}
 P(Q_g|g) = \sigma(r^Q_g(\boldhell{\Vset{g}}{})),
\end{align*}
where $\sigma$ is the softmax over the components of the aggregated vector.

To learn the emission $P(y_g|Q^i_g,g), \ i=1,\dots,C$, we have to implement a sub-network $\Phi_i$ that outputs the parameters of the chosen distribution. For instance, if the distribution was a multivariate Gaussian, we would have
\begin{align*}
\boldsymbol{\mu_i},\boldsymbol{\Sigma_i} = \Phi_i(\boldhell{g}{}) = f_i(r^i_g(\boldhell{\Vset{g}}{})),
\end{align*}
with $f_i$ being defined as $f_r$ above. Note that vertex-prediction tasks do not need a global aggregation phase, so the mixing weights and emission transformations would be individually applied to $\boldhell{v}{} \ \forall v \in \Vset{g}$.

It is important to remark that we share $\boldhell{\Vset{g}}{}$ between all sub-networks; this is different from the so-called Mixture of Experts approach \cite{jacobs_adaptive_1991,jordan_hierarchical_1994}, in which a different set of vertex representations would be created for each sub-network. This form of weight sharing reduces the number of parameters and pushes the model to extract all the relevant structural information into a single representation for each vertex. Last but not least, using multiple DGN encoders can easily become computationally intractable for large datasets.

\section{Training} We train the \GMDN{} model using the Expectation-Maximization (EM) framework \cite{dempster_maximum_1977} for MLE estimation. We continue choosing EM for the local convergence guaranteees that it offers with respect to other solutions, and since its effectiveness has already been proved on the probabilistic graph models introduced so far. By introducing the usual indicator variable $z^g_i \in \boldsymbol{Z}$, which is one when $Q_g$ has latent state $i$, we can compute the lower bound of the log-likelihood as in standard mixture models \cite{jordan_hierarchical_1994,corduneanu_variational_2001}:
\begin{align}
    & \mathbb{E}_{\boldsymbol{Z}|\dataset{}}[\log \mathcal{L}_c(h|\dataset{})] = \sum_{g \in \dataset{}}\sum_{i=1}^C E[z_i^g\vert\dataset{}] \log \Big(P(y_g|Q^i_g,g)P(Q_g|g) \Big)
\label{eq:gmdn-e-step}
\end{align}
where $\log \mathcal{L}_c(h|\dataset{})$ is the complete log likelihood.

The E-step of the EM algorithm can be performed analytically by computing the posterior probability of the indicator variables:
\begin{align*}
    E[z_i^g\vert\dataset{}]=P(z^g_i=1|g)=\frac{1}{Z_{norm}}P(y_g|Q^i_g,g)P(Q_g|g)
\end{align*}
where $Z_{norm}$ is a normalization term obtained via marginalization. On the other hand, we do not have closed-form solutions for the M-step because of the non-linear transformations $\Phi$ used. Hence, we perform a gradient ascent step to maximize Equation \ref{eq:gmdn-e-step}. This is an instance of the GEM algorithm (Section \ref{subsec:em}), which still guarantees convergence to a local minimum if each optimization step improves Equation \ref{eq:gmdn-e-step}. Finally, we introduce an optional Dirichlet regularizer $\pi$ with hyper-parameter $\boldsymbol{\alpha}=(\alpha_1,\dots,\alpha_C)$ on the distribution $P(Q_g|g)$. The prior distribution serves to prevent the posterior probability mass of the from collapsing onto a single state; this is a well-known problem that has been addressed in the literature through specific constraints \cite{eigen_learning_2013} or entropic regularization terms \cite{pereyra_regularizing_2017}. Eventually, the objective to be maximized becomes
\begin{align}
\underbrace{\mathbb{E}_{\boldsymbol{Z}|\dataset{}}[\log \mathcal{L}_c(h|\dataset{})]}_\text{original objective} +  \underbrace{\sum_{g \in \dataset{}}\log \pi(Q_g|\boldsymbol{\alpha})}_\text{Dirichlet regularizer},
\label{eq:objective}
\end{align}
where we note that $\boldsymbol{\alpha}=\boldsymbol{1}^C$ corresponds to a uniform prior, \ie no regularization. Maximizing Equation \ref{eq:objective} still preserves the convergence guarantees of GEM if the original objective increases at each step.

\section{Encoding the structure via distribution distances}
In the DGN literature, a common regularization technique encourages adjacent vertex representations to be similar in the Euclidean space and dissimilar otherwise \cite{kipf_semi-supervised_2017}. This can be achieved by computing the dot product of pairs of vertex representations followed by sigmoidal activation (to obtain a \quotes{probability} of being adjacent). Ideally, this regularization term should help the model focus on structural patterns rather than overfitting vertex features.

This way, however, the \textbf{space} of vertex representations is \textit{explicitly} constrained, and we argue that this may limit the amount of information that can be encoded into $\boldhell{\Vset{g}}{}$ about the main classification/regression task. For this reason, we propose the first insights into a \GMDN{}-based technique that \textit{implicitly} embeds structural information into vertex representations.  The idea to use \GMDN{} to produce separate vertex distributions other than those required for the main task. Then, we can encourage the distance between pairs of such vertex distributions to be close if the vertices are indeed adjacent. For the Data Processing Inequality \cite{cover_elements_1999},  it follows that vertex representations obtained in this way will encode structural information, but there will be \textbf{no explicit} constraint on the space they live in.

\begin{figure}[ht]
    \hspace{0cm}
    \centering
    \resizebox{0.8\textwidth}{!}{\input{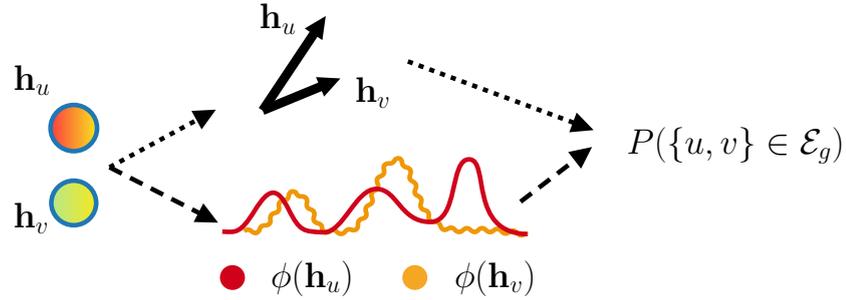}}
    \caption{In explicit regularization (top), adjacent vertex representations must be aligned in the Euclidean space. Instead, we propose to implicitly encode adjacency information (bottom) into the representations $\boldhell{\Vset{g}}{}$ by minimizing the distance between adjacent vertex \textbf{distributions}.}
    \label{fig:distance-intuition}
\end{figure}

While the application of this strategy to regularization seems promising, we should first investigate whether it is actually possible to learn appropriate distribution distances that encode the adjacency information. This thesis will take a step in this direction, rather than focusing on regularization benefits, by analyzing the ability of different distance functions to solve link prediction tasks. We graphically sketch the idea behind this experiment in Figure \ref{fig:distance-intuition}.

In this context, mixtures of Gaussians prove useful, as there are many possible choices for the distance function. An example is the closed-form L2 distance between two Gaussian mixture distributions P and Q described in \cite{helen_query_2007}. We define the L2 distance as

\begin{align}
L_2^2(P,Q) = \int_{\R{}} (p(\boldx{})-q(\boldx{}))^2 d\boldx{}.
\end{align}

This function sums the point-wise squared distances between the \textit{pdfs} of the two distributions, and it is not difficult to implement in matrix form for univariate Gaussian mixtures.

Lastly, mapping each vertex representation into a one-dimensional distribution could also be used as a structure-aware dimensionality reduction technique, in contrast to task-agnostic alternatives commonly used in the literature \cite{hinton_stochastic_2003,maaten_visualizing_2008}.

We now describe how we implement the different distances between pairs of Gaussian mixture distributions.
\subsection{L2 Distance}
This is the usual squared Euclidean distance
\begin{align}
L_2^2(P,Q) = \int_{\mathbb{R}} (p(\boldx{})-q(\boldx{}))^2 d\boldx{} \nonumber
\end{align}
which, for mixture of distributions, can be written as\footnote{We follow the straightforward derivation of \cite{you_l2_2019}.}
\begin{align}
& L_2^2(P,Q) = \int_{\mathbb{R}} \big(\sum_i^C \alpha_i p_i(\boldx{}) - \sum_j^C \beta_j q_j(\boldx{}) \big)^2 d\boldx{} \nonumber \\
& = \sum_{i,j} \alpha_i \alpha_j \int_{\mathbb{R}} p_i(\boldx{}) p_j(\boldx{}) d\boldx{} + \beta_i \beta_j \int_{\mathbb{R}} q_i(\boldx{}) q_j(\boldx{}) d\boldx{} \nonumber \\
& -2\sum_{i,j}\alpha_i \beta_j \int_{\mathbb{R}} p_i(\boldx{}) q_j(\boldx{}) d\boldx{}, \nonumber
\end{align}
where $\bm{\alpha}$ and $\bm{\beta}$ are the mixture weight vectors of the two distributions. In general, we can compute the integral of the product of two Gaussians as
\begin{align}
\begin{small}
\int_{\mathbb{R}} \mathcal{N}(\boldx{} \mid \boldsymbol{\mu_1}, \boldsymbol{\Sigma_1})\mathcal{N}(\boldx{} \mid \boldsymbol{\mu_2}, \boldsymbol{\Sigma_2})d\boldx{} = \mathcal{N}(\boldsymbol{\mu_1} \mid \boldsymbol{\mu_2}, (\boldsymbol{\Sigma_1} + \boldsymbol{\Sigma_2})) \nonumber
\end{small}
\end{align}
where we have used a known property of the product of Gaussians (see Section 8.1.8 of \cite{petersen_matrix_2012}) and the fact that the integral of a density function sums to 1. Therefore, if we define
\begin{align}
A_{i,j} = \int_{\mathbb{R}} p_i(\boldx{})p_j(\boldx{})d\boldx{} = \mathcal{N}(\mu^P_i \mid \mu^P_j, (\sigma^P_i + \sigma^P_j)) \nonumber \\
B_{i,j} = \int_{\mathbb{R}} q_i(\boldx{})q_j(\boldx{})d\boldx{} = \mathcal{N}(\mu^Q_i \mid \mu^Q_j, (\sigma^Q_i + \sigma^Q_j)) \nonumber \\
C_{i,j} = \int_{\mathbb{R}} p_i(\boldx{})q_j(\boldx{})d\boldx{} = \mathcal{N}(\mu^P_i \mid \mu^Q_j, (\sigma^P_i + \sigma^Q_j)) \nonumber
\end{align}
then the Euclidean distance can be computed as
\begin{align}
& L_2^2(P,Q) = \sum_{i,j} \alpha_i \alpha_j A_{i,j} + \beta_i \beta_j B_{i,j} -2\sum_{i,j}\alpha_i \beta_j C_{i,j}. \nonumber
\end{align}

\subsection{Jeffrey's Distance}
The Jeffrey's distance can be thought as the symmetric version of the KL Divergence or, equivalently, as double the $\alpha$-JS divergence with $\alpha=1$. We use the $\alpha$-JS divergence implementation, though the difference \wrt{} Jeffrey's lies only in a constant value. We consider a weighted sum of $C$ distances (univariate case) that relies on the corresponding mixing weights:
\begin{align}
& J_w(P, Q) = \frac{1}{2}\sum_{i=1}^C w_i^P KL(P_i \mid\mid Q_i) + w_i^QKL(Q_i \mid\mid P_i) \nonumber \\
& \text{where} \ KL(P \mid\mid Q) = \log\frac{\sigma_Q}{\sigma_P} + \frac{\sigma_P^Q + (\mu_P - \mu_Q)^2}{2\sigma_Q^2} - \frac{1}{2}. \nonumber
\end{align}

\subsection{Bhattacharyya's Distance}
Similarly, we define the weighted sum of Bhattacharyya's distances (univariate case) as
\begin{align}
& B_w(P, Q) = \sum_{i=1}^C \int_{\mathbb{R}}\sqrt{(w_i^Pp_i(x)w_i^Qq_i(x))}dx \nonumber \\
& = \sum_{i=1}^C \sqrt{w_i^Pw_i^Q}\int_{\mathbb{R}}\sqrt{p_i(x)q_i(x)}dx \nonumber \\
& = \sum_{i=1}^C \sqrt{w_i^Pw_i^Q} \Big( \frac{1}{8}\frac{2(\mu_P - \mu_Q)^2}{\sigma_1 + \sigma_2} + \frac{1}{2}\log(\frac{\sigma_P + \sigma_Q}{2\sqrt{\sigma_P\sigma_Q}}) \Big). \nonumber
\end{align}

\section{Experiments}
\label{sec:gmdn-experiments}

We now describe the datasets, experiments, evaluation process and hyper-parameters used to empirically study \GMDN{}. Our goal is to show how \GMDN{} can fit multimodal distributions conditioned on a graph better than using MDNs or DGNs individually. To do so, we publicly release large datasets of stochastic SIR simulations whose results depend on the underlying network, rather than assuming uniformly distributed connections as in \cite{davis_use_2020}. The datasets have been generated using random graphs from the Barabasi-Albert (BA) and Erdos-Renyi (ER) families. While ER graphs do not preserve social networks' properties, here we are just interested in the emergence of multimodal outcome distributions rather than biological plausibility. That said, future investigation will cover more realistic cases, for instance using the Block Two-Level Erdos–Renyi model \cite{seshadhri_community_2012}. We also apply the model on two molecular graph regression benchmarks, to analyze the performances of \GMDN{} on real-world data.

Two additional experiments complement the exposition: first, we analyze whether training on a particular family of graphs exhibits transfer properties; if that is the case, then the model has learned how to make informed predictions about different (let alone completely new) structures. Secondly, we study whether we can use vertex-specific multimodal distribution to perform link prediction. We recall that the main goal is to decouple the role of vertex representations from the objects used to compute the pair-wise link prediction scores.\footnote{\url{https://github.com/diningphil/graph-mixture-density-networks}.}

\begin{figure}[t]
\centering
\includegraphics[width=0.7\columnwidth]{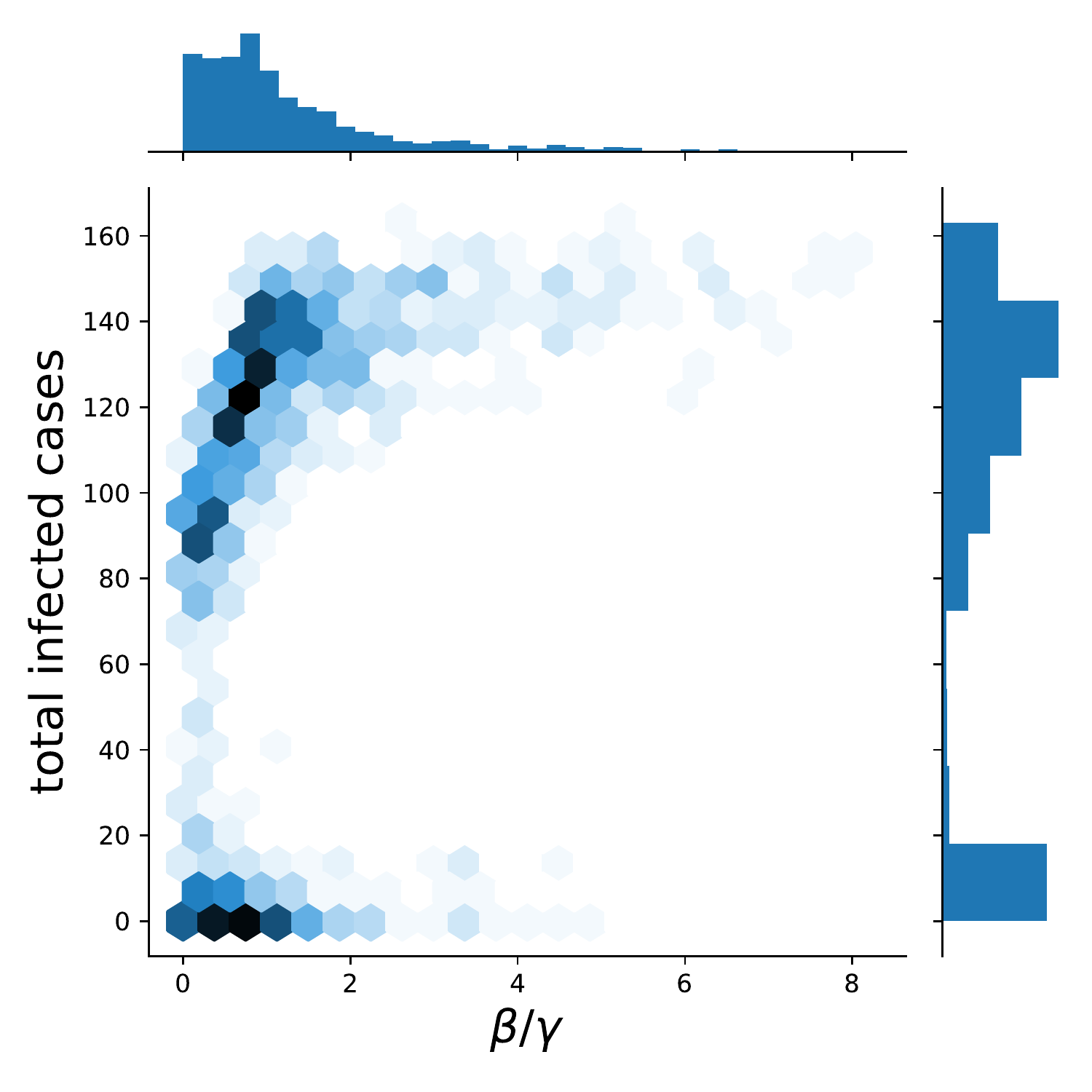}
\caption{Given a single network and specific choices for $R0=\beta/\gamma$, the repeated simulation of the stochastic SIR model is known to produce different outcomes. Here we plot the outcome distributions of 1000 SIR simulations on an Erdos-Renyi network of size 200. We follow \cite{davis_use_2020} and sample $\beta$ and $\gamma$ uniformly, rather than their ratio, because higher ratios correspond to less interesting behaviors, \ie the distribution becomes unimodal. Depending on the input structure, the distribution of the total infected cases may be multimodal or not, and \GMDN{} should recognize this phenomenon. In our simulations, larger networks exhibited less multimodality; hence, without loss of generality, we focus on larger datasets of smaller graphs.}
\label{fig:distribution-infected}
\end{figure}

\subsection{Datasets} We simulated the well-known stochastic SIR epidemiological model on Barabasi-Albert graphs of size 100 (BA-100), generating 100 random graphs for different connectivity values (2, 5, 10 and 20). Borrowing ideas from \cite{davis_use_2020}, for each configuration, we run 100 simulations for each different initial infection probability (1\%, 5\%, 10\%) sampling the infectivity parameter $\beta$ from $[0, 1]$ and the recovery parameter $\gamma$ from $[0.1, 1]$. We also carry out simulations for Erdos-Renyi graphs (ER-100), this time with connectivity parameters 0.01, 0.05, 0.1, and 0.2. The resulting total number of simulations (\ie samples) in each dataset is 120.000, and the goal is to predict the distribution of the total infected cases at the end of a simulation. Vertex features consist of $\beta$, $\gamma$, their ratio $R0=\beta/\gamma$, a constant value $1$, and a binary value that indicates whether that vertex is infected or not at the beginning of the simulation. Moreover, to test the transfer learning capabilities of GMDN on graphs with different structural properties (according to the chosen random graph model), we constructed six additional simulation datasets where graphs have different sizes, \ie 50, 200 and 500. An instance of simulation results is summarized in Figure \ref{fig:distribution-infected}; we observe that the outcome distribution of repeated simulations on a single graph leads to a multimodal distribution, in accord with \cite{opuszko_impact_2013}. Therefore, in principle, being able to accurately and efficiently predict the outcome distribution of a (possibly complex) epidemiological model can significantly impact the preparations for an incumbent sanitary emergency.

When dealing with real-world graph regression tasks, especially in the chemical domain, we usually do not expect such a conspicuous emergence of multimodality in the output distribution. Indeed, the properties of each molecule are assumed to be regulated by natural laws, but the information we possess about the input representation may be incomplete and/or noisy. Similarly, the way the model processes the input has an impact on the overall uncertainty; for instance, disregarding bond information makes graphs appear isomorphic to the model while they are indeed not so. As such, knowing the confidence of a trained regressor for a specific outcome becomes invaluable to better understand the data, the model behavior, and, ultimately, to determine the trust we place in each prediction. Therefore, we will evaluate our model on the large chemical benchmarks alchemy\_full \cite{chen_alchemy_2019} and ZINC\_full \cite{irwin_zinc_2012,bresson_two_2019} made of 202579 and 249456 molecules, respectively. The task of both datasets is the prediction of continuous chemical properties (12 for the former and 1 for the latter) associated with each molecule representation (9 and 28 vertex features, respectively). As in \cite{chen_alchemy_2019}, the GIN convolution used only exploits the existence of a bond between atoms. In the considered datasets, this gives rise to isomorphic representations of different molecules when bond types or 3D coordinates are not considered (or simply ignored by the trained model). The same phenomena, in different contexts and forms, can occur whenever the original data or its choice of representation lack part of the information to solve a task.

Finally, to start studying the feasibility of \GMDN{} as a link predictor, we will make use of the same Cora and Pubmed datasets introduced in the context of \ECGMM{}.

\subsection{Evaluation Setup.} We assess the performance of different models using a holdout strategy for all datasets ($80\%/10\%/10\%$ split). Given the size of the datasets, we believe that a simple holdout is sufficient to assess the performances of the different models considered. To make the evaluation even more robust for the epidemic datasets, different simulations about the same graph cannot appear in both training and test splits. The metric of interest is the log-likelihood of the data ($\log \mathcal{L}$), which captures how well we can fit the target distribution and the model's uncertainty with respect to a particular output value. We also report the Mean Average Error (MAE) on the real-world benchmarks for completeness. However, the MAE does not reflect the model's uncertainty about the output, as we will show.

Instead, we split the links of Cora and Pubmed according to a bootstrap sampling technique. We created ten different 85\%/5\%/10\% link splits with an equal number of true (class 1) and false (class 0) edges. We recall that this setup is more robust than using a single split \cite{kipf_semi-supervised_2017,shchur_pitfalls_2018}, but it is ten times more expensive because we must perform a model selection for each split. We treat sampled false edges as directed for a better exploration of the space of unconnected pairs of vertices. We use an L1 loss with target distance 0 for the positive class and 2 for the negative class. When it comes to computing classification scores, we convert each distance $d$ into a probability using the continuous function $1/(1+d)$ (though hard thresholds are also possible). Following the literature, we evaluate the classification performance using the area under the curve (AUC) and the average precision (AP).

We perform model selection via grid search for all the models presented. For each of them, we select the best configuration on the validation set using early stopping with patience \cite{prechelt_early_1998}. As regards holdout, to avoid an unlucky random initialization of the chosen configuration, we average the model's performance on the unseen test set over ten final training runs. Instead, 3 final training runs are used for the link prediction experiments. Similarly to the model selection phase, in all these final training runs we use early stopping on a validation set extracted from the training set (10\% of the training data).

\paragraph*{Baselines and hyper-parameters}
On the synthetic and chemical tasks, we compare \GMDN{} against four different baselines. First, RAND predicts the uniform probability over the finite set of possible outcomes, thus providing the threshold log-likelihood score above which predictions are useful. Instead, HIST computes the normalized frequency histogram of the target values given the training data, which is then converted into a discrete probability. While on epidemic simulations we can use the graph's size as the number of histogram bins to use, on the chemical benchmarks this number must be treated as a hyper-parameter and manually cross-validated against the validation set. HIST is used to test whether multimodality is useful when a model does not take the structure into account. Finally, we have MDN and DGN, which are, in a sense, ablated versions of \GMDN{}. Indeed, MDN ignores the input structure, whereas DGN cannot model multimodality. Neural models are trained to output unimodal (DGN) or multimodal (MDN, \GMDN{}) binomial distributions for the epidemic simulation datasets and isotropic Gaussians for the chemical ones. The sub-networks $\Phi_i$ are linear models, and the graph convolutional layer is adapted from \cite{xu_how_2019}.

For link prediction, we test a Graph Auto-Encoder (GAE) \cite{kipf_variational_2016} as a strong baseline that computes the dot product between pairs of vertex representations. Then, we test different versions of \GMDN{} according to the distributional distance used: L2 distance (\GMDN{}-L2), weighted Jeffrey distance (\GMDN{}-J), and weighted Bhattacharyya distance (\GMDN{}-B). We now list the hyper-parameters tried for each model:
\begin{itemize}
\setlength\itemsep{-0.4em}
\item MDN: $C$ $\in$ \{2,3,5\}, hidden units per convolution $\in$ \{64\}, neighborhood aggregation $\in$ \{sum\}, global aggregation $\in$ \{sum, mean\}, $\boldsymbol{\alpha}$ $\in$ \{$\boldsymbol{1}^C$, $\boldsymbol{1.05}^C$\}, epochs $\in$ \{2500\}, $\Phi_i$ $\in$ \{\text{Linear model}\},  Adam Optimizer with learning rate $\in$ \{0.0001\}, full batch, patience $\in$ \{30\}.
\item \GMDN{}: $C$ $\in$ \{3,5\}, number of layers $\in$ \{2,5,7\}, hidden units per convolution $\in$ \{64\}, neighborhood aggregation $\in$ \{sum\}, global aggregation  $\in$ \{sum, mean\}, $\boldsymbol{\alpha}$ $\in$ \{$\boldsymbol{1}^C$, $\boldsymbol{1.05}^C$\}, epochs $\in$ \{2500\}, $\Phi_i$ $\in$ \{\text{Linear model}\},  Adam Optimizer with learning rate $\in$ \{0.0001\}, full batch, patience $\in$ \{30\}.
\item DGN:  same as \GMDN{} but $C$ $\in$ \{$1$\} (that is, it outputs a unimodal distribution).
\item \textsc{GAE}:  number of layers $\in$ \{1,2,3\}, hidden units per convolution $\in$ $\{32,64$,$128,256,512\}$, neighborhood aggregation $\in$ \{sum\}, epochs $\in$ \{5000\},  Adam Optimizer with learning rate $\in$ \{0.01, 0.001\}, full batch,  patience $\in$ \{1000\}.
\item \textsc{GMDN-L2/J/B}: $C$ $\in$ \{2,5,10,20,30,50\}, number of layers $\in$ \{1\}\footnote{After evaluating GAE, we observed that 1 layer was sufficient to achieve the best results.}, hidden units per convolution $\in$ \{128,256,512,1024\}, neighborhood aggregation $\in$ \{sum, mean\},$\boldsymbol{\alpha}$ $\in$ \{$\boldsymbol{1}^C$\}, epochs $\in$ \{2500\}, $\Phi_i$ $\in$ \{\text{Linear model} \},  Adam Optimizer with learning rate $\in$ \{0.01, 0.001\}, full batch,  patience $\in$ \{200 (\textsc{GMDN-J}), 500\}.
\end{itemize}
Note that we kept the maximum number of epochs intentionally high as we use early stopping to halt training. Also, the results of the experiments hold regardless of the DGN variant used, given the fact that DGNs output a single value rather than a complex distribution. In other words, we are comparing \textit{families} of models rather than specific architectures.

\section{Results}
\label{sec:results}
We discuss our findings starting from the main empirical study on epidemic simulations, which includes CDE results and transferability of the learned knowledge. Then, we report results obtained on the real-world chemical tasks, highlighting the importance of capturing a model's uncertainty about the output predictions. Finally, we show the first insights into using distributional distances to tackle link prediction tasks.

\subsection{Epidemic Simulation Results}
We begin by analyzing the results obtained on BA-100 and ER-100 in Table \ref{tab:gmdn-results}. We notice that \GMDN{} has better test log-likelihoods than the other baselines, with larger performance gains on ER-100.  Being \GMDN{} the only model that considers both structure and multimodality, such an improvement was to be expected. However, it is particularly interesting that HIST has a better log-likelihood than MDN on both tasks. By combining this fact with the results of DGN, we come to two conclusions. First, the structural information seems to be the primary factor of performance improvement; this should not come as a surprise since the way an epidemic develops depends on how the network is organized (despite we are not aiming for biological plausibility). Secondly, none of the baselines can get close enough to \GMDN{} on ER-100, indicating that this task is harder to solve by looking individually at structure or multimodality. In this sense, BA-100 might be considered an easier task than ER-100, and this is plausible because emergence of multimodality on the former task seems slightly less pronounced in the SIR simulations. For completeness, we also tested an intermediate baseline where DGN is trained with L1 loss followed by MDN on the graph embeddings. Results displayed a $\log\mathcal{L} \approx -16$ on both datasets, probably because the DGN creates similar graph embeddings for different distributions with the same mean, with consequent severe loss of information.

\begin{table}[ht]
\centering
\begin{tabular}{lcccc}
\toprule
     & \textbf{\textsc{BA-100}} & \textbf{\textsc{ER-100}} & \textbf{Structure}       & \textbf{Multimodal}        \\ \midrule
RAND & -4.60       & -4.60       &    \absent      &    \absent        \\
HIST & -1.16       & -2.32       &    \absent      &    \present        \\
MDN  & -1.17(.05)       & -2.54(.07)       &    \absent     &    \present       \\
DGN  & -0.90(.35)       & -1.96(.16)            &    \present      &    \absent        \\ \midrule
\GMDN{} & \textbf{-0.67}(.02) & \textbf{-1.56}(.04) & \present   &    \present      \\ \bottomrule
\end{tabular}
\caption{Results on BA-100 e ER-100 (12.000 test samples each). A higher log-likelihood corresponds to better performances. \GMDN{} improves the performance on both tasks, showing the advantages of that taking into account both multimodality and structure. Neural models' results are averaged over 10 runs, and standard deviation is reported in brackets.}
\label{tab:gmdn-results}
\end{table}

\begin{figure}[ht]
\centering
\begin{subfigure}
  \centering
  \includegraphics[width=0.49\textwidth]{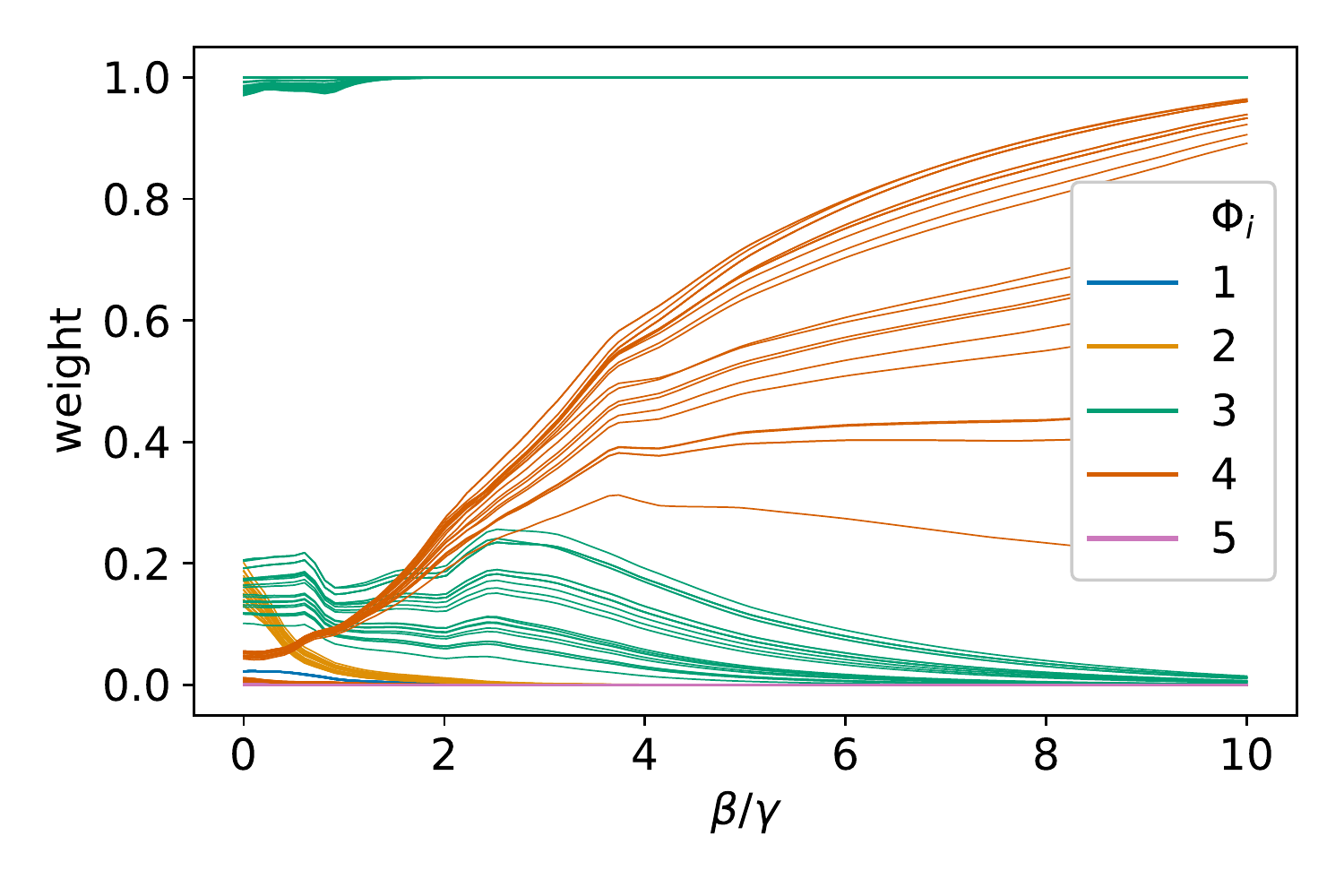}
\end{subfigure}%
\begin{subfigure}
  \centering
  \includegraphics[width=0.49\textwidth]{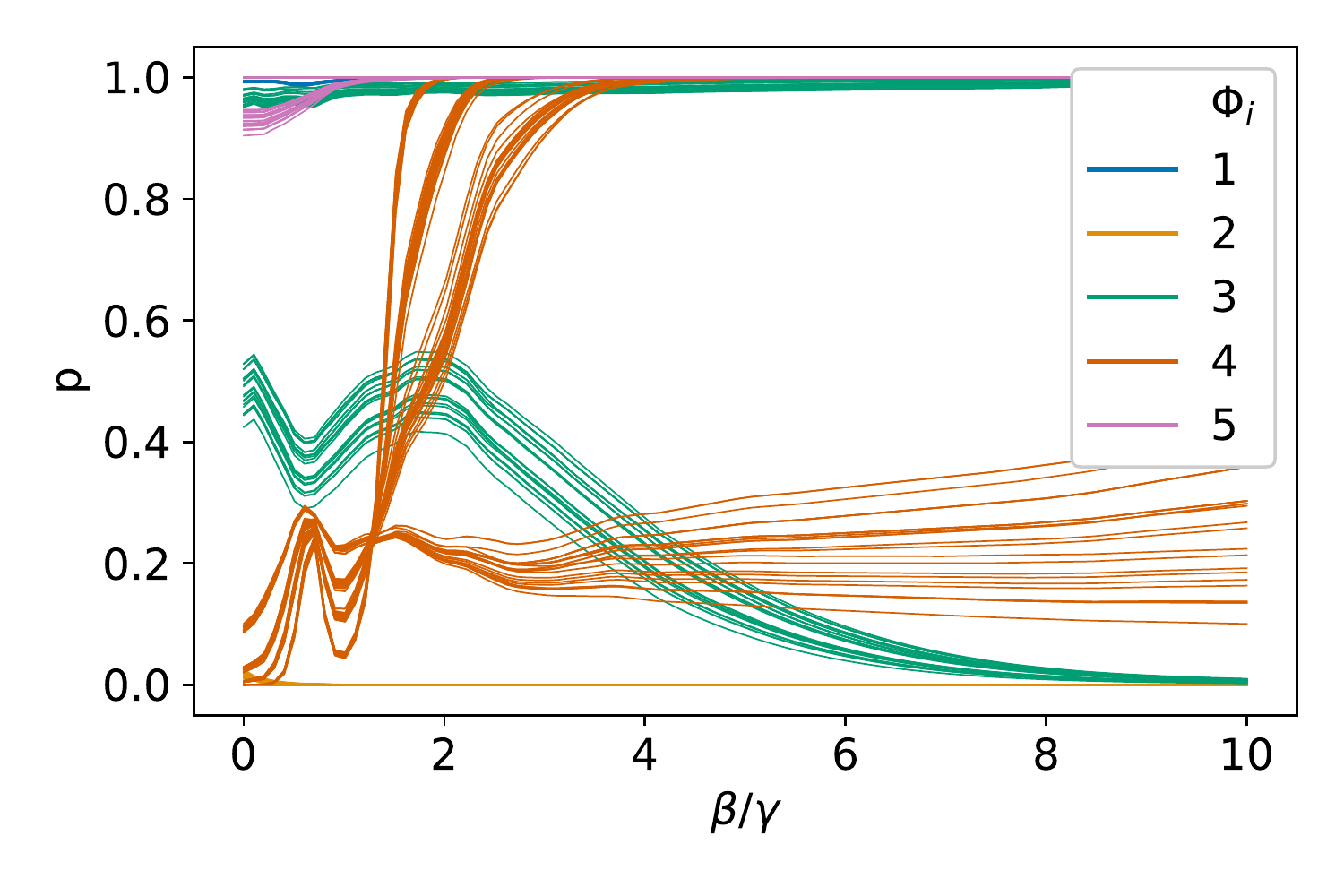}
\end{subfigure}
\caption{The trend of the mixing weights (left) and binomial coefficient (right) for each one of five sub-networks is shown on 100 ER-100 graphs. We vary the ratio between infection and recovery rate to inspect the behavior of the \GMDN{}. Here, we see that sub-network 4 can greatly change the binomial output distribution in a way that depends on the input graph.}
\label{fig:parameters-distributions}
\end{figure}

Similarly to what has been done in \cite{bishop_mixture_1994} and \cite{davis_use_2020}, we analyze how the mixing weights and the distribution parameters vary on a particular \GMDN{} instance. We use $C$=5 and track the behavior of each sub-network for 100 different ER-100 graphs. Figure \ref{fig:parameters-distributions} shows the trend of the mixing weights (left) and of the binomial parameters $p$ (right) for different values of the ratio $R0=\beta/\gamma$. We immediately see that many of the sub-networks are \quotes{shut down} as the ratio grows. In particular, sub-networks 3 and 4 are the ones that control \GMDN{}'s output distribution the most, though for high values of R0 only one sub-network suffices. These observations are concordant with the behavior of Figure \ref{fig:distribution-infected}: when the infectivity rate is much higher than the recovery rate, the target distribution becomes unimodal. The analysis of the binomial parameter for sub-network 4 provides another interesting insight. We notice that, depending on the input graph, the sub-network leads to two possible outcomes: the outbreak of the disease or a partial infection of the network. Note that this is a behavior that \GMDN{} can model whereas the classical MDN cannot.

\begin{figure}[ht]
    \centering
    \includegraphics[width=0.8\columnwidth]{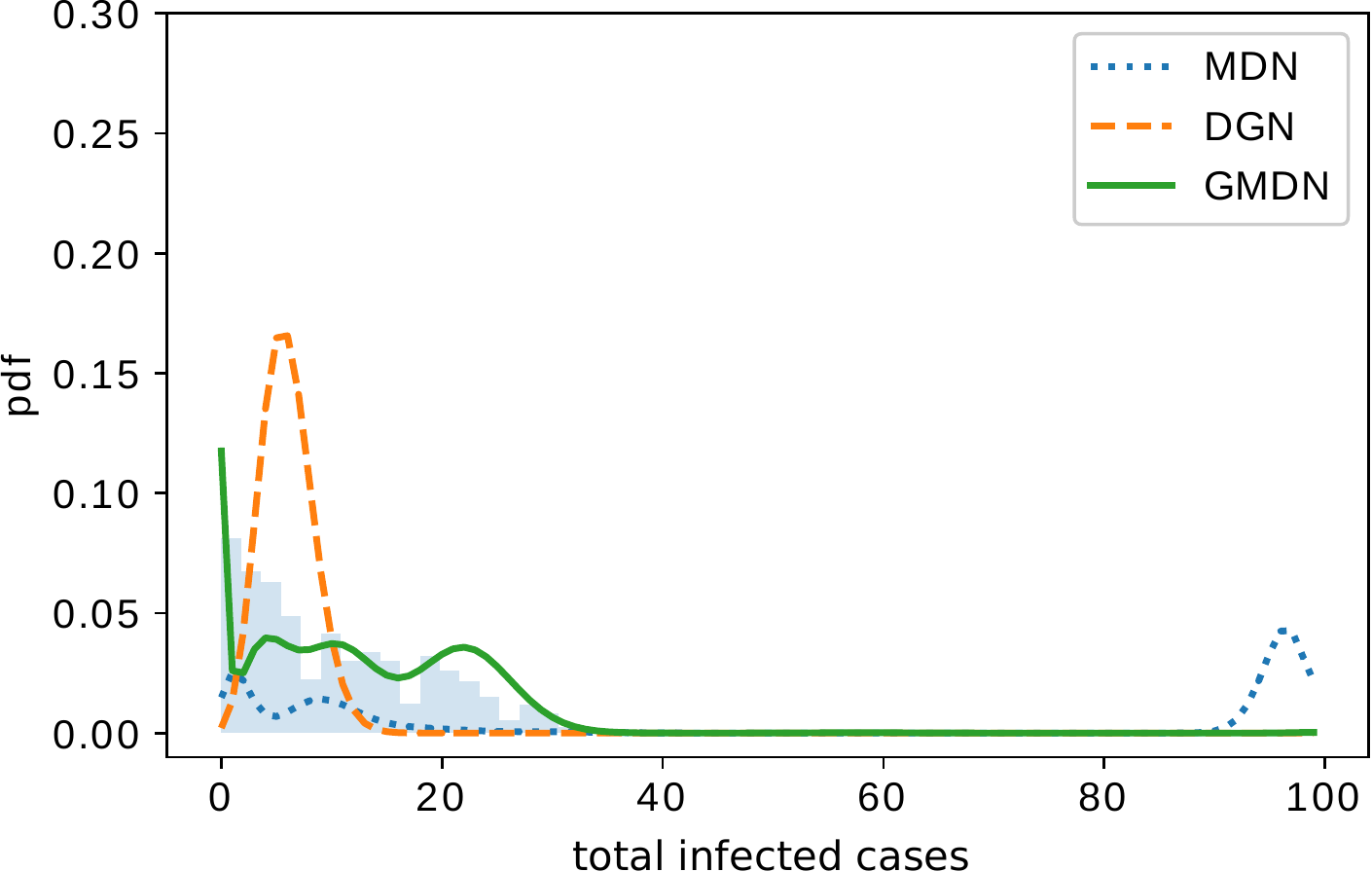}
    \caption{Output distributions of MDN, DGN, and \GMDN{} on an ER graph of size 100. As we can see, the \GMDN{} can provide a rich multimodal distribution conditioned on the structure close to that generated by SIR simulations (blue histogram).}
    \label{fig:erdos-renyi}
\end{figure}

To provide further evidence about the benefits of the proposed model, Figure \ref{fig:erdos-renyi} shows the output distributions of MDN, DGN and \GMDN{} for a given sample of the ER-100 dataset. We also plot the result of SIR simulations on that sample as a blue histogram (ground truth). Some observations can be made. First, the MDN places the output probability mass at both sides of the plot. This choice is understandable considering the lack of knowledge about the underlying structure (see also Table \ref{tab:gmdn-results}) and the fact that likely output values tend to be polarized at the extremes (see \eg Figure \ref{fig:distribution-infected}). Secondly, the DGN can process the structure but cannot model more than one outcome. Therefore, and coherently with \cite{bishop_mixture_1994} for vectorial data, the DGN unique mode lies in between those of \GMDN{} that account for the majority of \GMDN{} probability mass.  In contrast, \GMDN{} produces a multimodal and structure-aware distribution that closely follows the ground truth.

\subsection{Transfer Results}
\begin{figure}[ht]
    \centering
    \includegraphics[width=0.8\columnwidth]{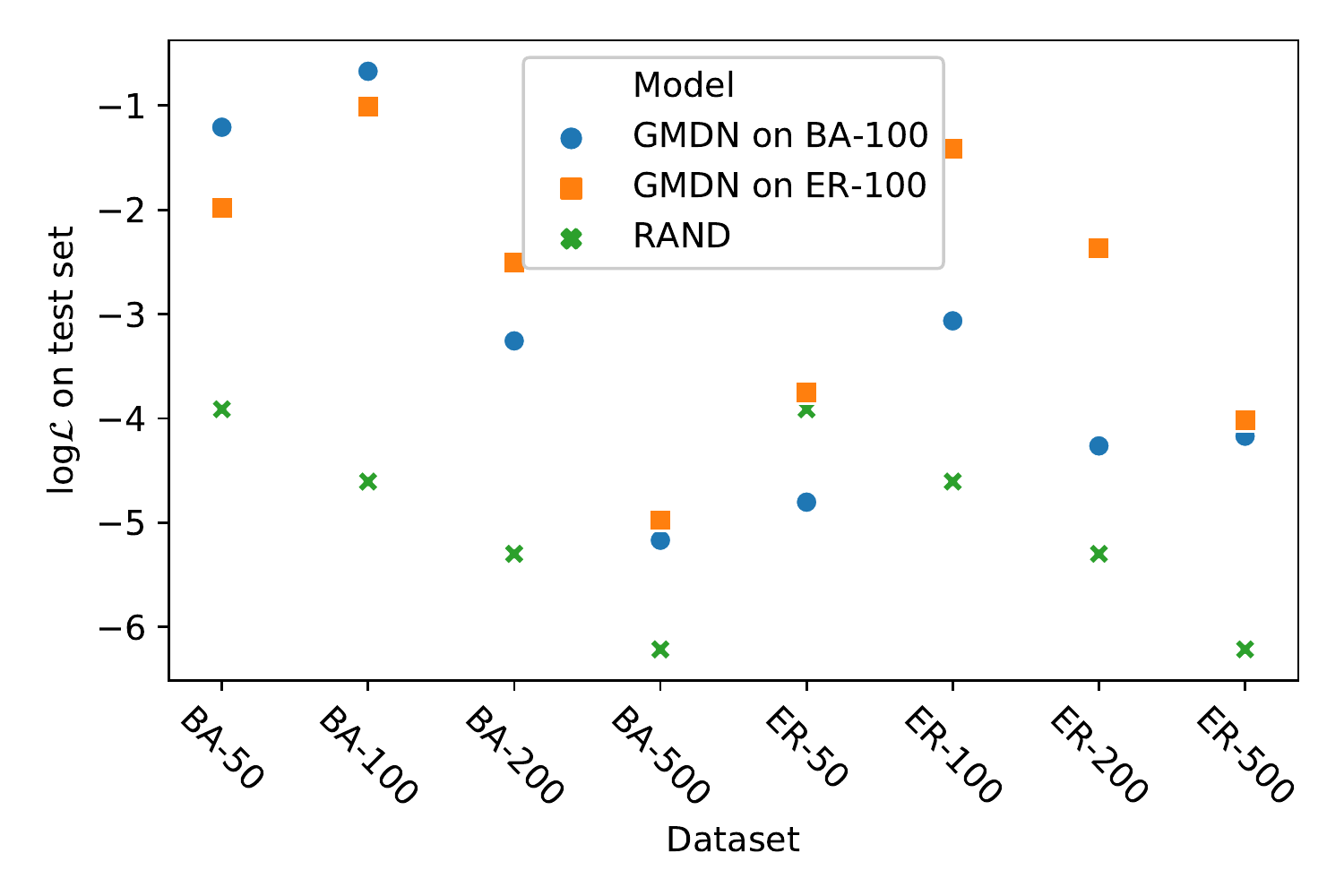}
    \caption{Transfer learning effect of the trained \GMDN{}s are shown as blue dots and orange squares. Higher scores are better. \GMDN{} trained on ER-100 exhibits better transfer on larger BA-datasets, which might be explained by the difficulty of the source task.}
    \label{fig:transfer-effect}
\end{figure}

To tell whether \GMDN{} can transfer knowledge to a random graph of different size and/or family (\ie with different structural properties), we evaluate the trained models on the six additional datasets described in Section \ref{sec:gmdn-experiments}. Results are shown in Figure \ref{fig:transfer-effect}, where the RAND score acts as the reference baseline. The general trend is that the \GMDN{} trained on ER-100 has better performances than its counterpart trained on BA-100; this is true for all ER datasets, BA-200 and BA-500. This observation suggests that training on ER-100, which we assumed to be a \quotes{harder} task than BA-100 as discussed above, allows the model to better learn the dynamics of SIR and transfer them to completely different graphs. Since the structural properties of the random graphs vary across the datasets, obtaining a transfer effect is therefore far from being a trivial task.

\subsection{Chemical Benchmarks}
\begin{figure}[ht]
    \centering
    \includegraphics[width=0.8\columnwidth]{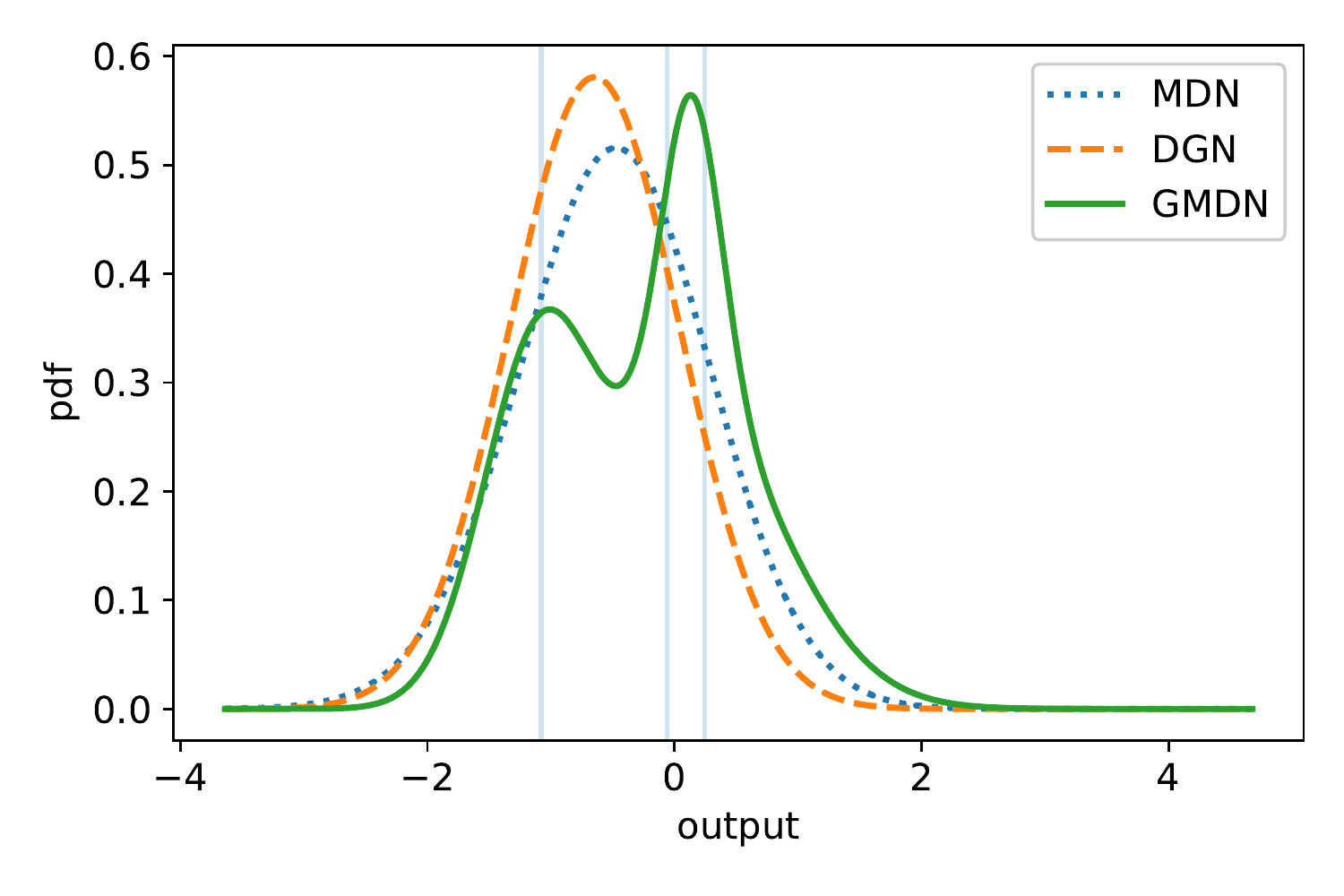}
    \caption{We illustrate the output distributions on the first component, \ie dipole moment, of an alchemy\_full graph. As noted in the text, DGN places high confidence in between the two modes of \GMDN{}. On the contrary, \GMDN{} is able to express uncertainty about the possible output values (vertical lines) associated with isomorphic graphs, which can be found if 3D attributes are not considered. The existence of the two modes suggests that 3D attributes are nonetheless ignored by the three models. See the discussion of Section \ref{sec:gmdn-experiments} for a more in-depth explanation of the phenomenon.}
    \label{fig:alchemy}
\end{figure}

We move to the results on the real-world chemical benchmarks, which are summarized in Table \ref{tab:gmdn-results-chemical}. We observe a log-likelihood trend similar to that in Table \ref{tab:gmdn-results}, with the notable difference that DGN performs much worse than MDN on alchemy\_full. Following the discussion in Section \ref{sec:gmdn-experiments}, we evaluate how models deal with the uncertainty in the prediction by analyzing one of the output components of alchemy\_full. Figure \ref{fig:alchemy} shows such an example for the first component (dipole moment).
The two modes of the \GMDN{} suggest that, for some input graphs, it may not be clear which output value is more appropriate. This is confirmed by the vertical lines representing output values of isomorphic graphs (as discussed in Section \ref{sec:gmdn-experiments}). Similarly to Figure \ref{fig:erdos-renyi}, the DGN tries to cover all possible outcomes with a single Gaussian in between the \GMDN{} modes. Although this choice may well minimize the MAE score over the dataset, the DGN fails to model the data we have.

\begin{table}[ht]
\centering
\begin{tabular}{lcccc}
\toprule
 & \multicolumn{2}{c}{\textbf{alchemy\_full}} & \multicolumn{2}{c}{\textbf{ZINC\_full}} \\
\multicolumn{1}{c}{}                                & $\log \mathcal{L}$     & MAE              & $\log \mathcal{L}$   & MAE             \\ \midrule
RAND                                                & -27.12                 & -                &  -4.20               & -               \\
HIST                                                & -21.91                 & -                &  -1.28               & -               \\
MDN                                                 & -1.36(.90)             & 0.62(.01)        & -1.14(.01)           & 0.67(.00)     \\
DGN                                                 & -7.19(1.3)             & 0.62(.01)        & -0.90(.10)           & 0.49(.03)      \\ \midrule
GMDN                                                & \textbf{-0.57}(1.4)    & \textbf{0.61}(.02)       & \textbf{-0.75}(.10)           & \textbf{0.49}(.04)     \\ \bottomrule
\end{tabular}
\caption{Results on the chemical tasks show how \GMDN{} consistently reaches better log-likelihood values than the baselines. We also report the MAE as secondary metric for future reference, using the weighted mean of the sub-networks as the prediction (see \cite{bishop_mixture_1994} for alternatives). Clearly, the MAE does not reflect the amount of uncertainty in a model's prediction, whereas the log-likelihood is the natural metric for that matter. Results are averaged over 10 training runs with standard deviation in brackets.}
\label{tab:gmdn-results-chemical}
\end{table}

\subsection{Distributional Distances for Link Prediction}
We conclude the chapter with an investigation into the ability of \GMDN{} to implicitly transfer structural information into the vertex representations by computing the distance between vertex distributions. We report our structure-reconstruction results in Table \ref{tab:link-pred}. When looking at the results, it is important to remark that the objective is \textit{not} to perform better than \textsc{GAE} but rather to assess whether distribution distances can retain a good amount of structural information.

\begin{table}[ht]
\centering
\begin{tabular}{lllll}
\toprule
                                                    & \multicolumn{2}{c}{\textbf{Cora}}                & \multicolumn{2}{c}{\textbf{Pubmed}}              \\
\multicolumn{1}{c}{}                                & \multicolumn{1}{c}{AUC} & \multicolumn{1}{c}{AP} & \multicolumn{1}{c}{AUC} & \multicolumn{1}{c}{AP} \\
\midrule
GAE                                                 &      $91.9(0.6)$        &     $92.0(0.6)$        &      $96.9(0.2)$        &       $97.0(0.2)$      \\
GMDN-B                                              &      $86.8(1.4)$        &     $87.0(1.4)$        &      $89.8(1.2)$   &       $86.2(1.7)$      \\
GMDN-L2                                             &      $81.6(3.5)$        &     $82.9(3.4)$        &      $91.8(1.4)$ &      $90.2(1.3)$       \\
GMDN-J                                              &      $87.8(1.2)$        &     $89.0(1.6)$        &      $94.9(0.4)$   &       $94.5(0.4)$      \\ \bottomrule
\end{tabular}
\caption{Results for the structure reconstruction tasks. 
}
\label{tab:link-pred}
\end{table}

In general, all the distance functions perform properly, though the weighted Jeffrey distance is the one with the best results (even close to the ones of GAE).
This distance is also more efficient than the L2, which elegantly takes into account Gaussian mixtures but has a quadratic cost in the number of sub-networks. Also, \GMDN{}-L2 was the slowest converging model, possibly due to the complex dependencies between pairs of mixtures. 

From these preliminary results, it seems clear that distributional distances can be used to approximately reconstruct most of the adjacency information without necessarily imposing explicit constraints on internal vertex representations. Therefore, future work will further investigate the potential of this technique as a regularization strategy when solving common graph regression/classification tasks, as well as finding new distributional distances that provide even better link prediction accuracies.

\section{Summary}
\label{sec:gmdn-conclusions}
With the Graph Mixture Density Networks, we have introduced a new family of models that combine the benefits of Deep Graph Networks and Mixture Density Networks. \GMDN{} can solve challenging tasks where the input is a graph and the conditional output distribution is multimodal. In this respect, we have introduced a novel benchmark application for graph conditional density estimation founded on stochastic epidemiological simulations. The effectiveness of GMDM has also been demonstrated on real-world chemical regression tasks and as a promising tool to address link prediction.

In the future, we plan to further study the impact of \GMDN{} on more biologically plausible synthetic datasets and find new application domains. We believe there are plenty of directions in which we can extend \GMDN{}, for instance by using a recurrent encoder in order to model dynamically varying graphs \cite{wu_graph_2019,wu_connecting_2020}. Moreover, we foresee that graph-based reinforcement learning \cite{kahn_uncertainty_2017,choi_uncertainty_2018} may benefit from the added degree of uncertainty over continuous outputs that \GMDN{} provides.

Overall, we hope that this general framework will play an important role in the approximation of structure-dependent phenomena that exhibit non-trivial conditional output distributions.

\chapter{Conclusions}
\label{chapter:conclusions}
\epigraph{\textit{E quindi uscimmo a riveder le stelle.}}{\textit{Inferno - Canto XXXIV}}


Over the last fifteen years, we have witnessed a race to digitalization throughout all public and private sectors. As a by-product of this ongoing modernization, the amount of data produced and stored has swiftly increased, so much so that it is now considered a new asset for businesses. Despite the ethical implications of this being debated by experts as well as by the general public, we could argue that the availability of data samples has fostered the research and application of \ML{} techniques that provide the community with valuable services. Real-time translation, breast cancer detection, and hate-speech recognition are just a few instances of what can be seen as a direct or indirect attempt to make the world a safer and more inclusive place.

Oftentimes, however, these services make use of relatively simple realizations of structured data in the form of vectors or sequences, with the consequent inability to process more complex relationships that may exist; this is the case of molecular predictions and generation, discovery of unknown protein-protein interactions, and detection of malicious activities in a social network or software, where the data is naturally encoded as a graph. Indeed, the classical algorithmic extraction of actionable information from such structures is rarely an easy task, due to the fact that any topological variability of the input graphs must be accounted for. This dissertation discussed what is Deep Learning for Graphs and how it can help in the automatic discovery of a mapping between a graph-structured input and a flat output. In Section \ref{sec:dl4g}, we tried to give a broad but systematic perspective of the basic principles that characterize this field \cite{bacciu_gentle_2020}. The top-down approach we followed was meant to be accessible even to the noninitiated, and it allowed us to see our and others' contributions through the same lens. Moreover, our review paid attention to the foundational approaches that shaped the field both to give a historical perspective and to prevent a wave of re-discovery of ideas.

Throughout the entire manuscript, we were also conscious of the fact that experimental reproducibility on some graph benchmarks had been slightly overlooked, possibly because of the tremendous stream of works produced in the last years. For this reason, before commencing any methodological chapter, we did our best to ensure that a fair, robust, and reproducible comparison on graph classification benchmarks was available to compare our models against state of the art DGNs (Section \ref{sec:scholarship-issues} \cite{errica_fair_2020}). In the process, we discovered the importance of setting up proper structure-agnostic baselines, and we showed how an incorrect evaluation setup can result in over-optimistic estimates of the models' generalization performances.

To show that Deep Learning for Graphs is actually useful in practice, we integrated a real-world example from the field of molecular biosciences \cite{errica_deep_2021} in Section \ref{sec:application-molecular}. Specifically, we showed that Deep Graph Networks are able to fairly well approximate a very complex process that takes a given protein and returns an information-theoretic quantity of interest. The real advantage of doing so is that said approximation can be done in a minuscule fraction of the time required by the original method, thus allowing a quasi-exhaustive study of the protein under consideration. That said, it is still unclear how to transfer the learned knowledge to a different family of proteins, which could radically change the way we approach the problem.

Moving to our main contribution, the design of Deep Bayesian Graph Networks has been guided by the principles of local and iterative processing of information as well as by the classical building blocks of Bayesian learning. The goal was to show that it is possible to implement an effective, deep, and fully probabilistic learning approach for one of the most unconstrained data structures. Inspired by pioneering methods, we have proposed a probabilistic framework for learning on graphs, founded on an incremental construction that facilitates information propagation through deeper architectures with respect to most neural counterparts. The Contextual Graph Markov Model of Section \ref{sec:cgmm} can be seen as the simplest realization of such a framework, in which the neighborhood aggregation is defined in probabilistic terms and can deal with discrete edge labels \cite{bacciu_contextual_2018,bacciu_probabilistic_2020}. Surprisingly enough, the unsupervised nature of the model did not come at the price of significantly worse performances in the supervised tasks considered; rather, the model reached a very competitive accuracy on a number of classification benchmarks.

Later on, in Section \ref{sec:ecgmm}, we continued to find ways to make our framework more general. One of the issues was that the presence of non-discrete edge features could not be modeled by CGMM. The solution lied in acting at the architectural level rather than on the definition of the neighborhood aggregation mechanism \cite{atzeni_modeling_2021}; the addition of a second Bayesian network, responsible for the generation of edge features, made possible to adaptively discretize (in an unsupervised fashion) edge information, so that we could apply again the original CGMM model. An unexpected outcome of this extension was that classification performances improved even on those benchmarks where edge features were missing. We attributed this phenomenon to the dynamic neighbor aggregation that arises from the discretization of edges at each layer of the constructive architecture.

There are still many interesting open problems that regard the generality of the framework. For instance, the neighborhood aggregation scheme is based on the mean operator, but the sum is a theoretically more expressive operator over multi-sets (under appropriate conditions). Hence, the investigation of a fully-probabilistic formulation for the sum aggregation would certainly enhance the representational capabilities of the framework, which could then capture richer patterns in the graph structure. Nonetheless, in the interest of a broad exploration of the research space, we instead focused on the automatic selection of some hyper-parameters of the framework, in particular the number of latent states of the vertices' categorical variables. Section \ref{sec:icgmm} bridges ideas from Deep Learning for Graphs and Bayesian Nonparametric methods to create the first DBGN whose complexity grows with the data. What we named Infinite Contextual Graph Markov Model is a deep architecture where each layer is a possibly infinite conditional mixture model, implemented as a Gibbs sampling-based Hierarchical Dirichlet Process. Empirically, the model performed similarly to CGMM, but it chose a number of latent states much smaller than the best configuration of a model selection procedure, thus saving disk space where to store the embeddings. The drawbacks of the approach are the slowness of the sampler and the inability to consider edge features, both of which will be subject of future works, for instance via variational derivations of the HDP.

To apply the DBGNs developed in this thesis to a real-world application, we considered the problem of robust malware classification in Section \ref{sec:application-security} \cite{errica_robust_2021}. In particular, by considering graph representations of programs that are unaware of the intra-procedural code changes, we managed to successfully classify a substantial number of malware families by just looking at the topology of the input graph. This, in turn, allowed the proposed procedure to be robust to a particular subset of code obfuscation techniques, and the learned unsupervised embeddings were rich enough to distinguish the peculiar structural variations in the data distribution.

The common thread of the entire thesis has been the cross-fertilization of ideas belonging to different research fields. In keeping with this spirit, the last methodological contribution of Chapter \ref{chapter:hybrid} has been a hybrid framework that combines a generic encoding transduction, realized by a DGN, and the probabilistic capabilities of Bayesian networks \cite{errica_graph_2021}. This was necessary to model multimodal output distributions conditioned on topologically varying input graphs. The resulting Graph Mixture Density Network takes the best of both neural and probabilistic worlds to solve regression problems like the prediction of an epidemic's outcome on synthetic graphs, which is inherently stochastic. This model can also be used to add a degree of trustworthiness to the predictive process, since it can express the uncertainty about the possible outcomes using a mixture of simple distributions. Being very general, GMDN is also amenable to further extensions, for example by incorporating a recurrent DGN encoder for graphs' time-series prediction. Moreover, the experiments showed that a DGN itself cannot capture any multimodality in the output distribution, due to the implicit assumptions that are usually made about the regression problem. In contrast, GMDN was able to correctly predict different outcomes alongside their likelihood.

\section{Future Directions}
There are a number of potential directions to be investigated in the future, which are mostly methodological and related to the Bayesian nature of the proposed models. One of these concerns the incorporation of supervision into the embeddings' generation process, which could make the presence of a final neural predictor unnecessary in the overall architecture. Another possibility would be to tackle unsupervised learning of temporal graphs, by making the DBGNs accept sequences of observable variables rather than static information. Moreover, the use of Bayesian networks and multimodal distributions may ease the inspection of the inner workings of our models, allowing us to detect whether or not a topological change in the input graph produces lower likelihoods or significantly different multimodal output distributions. Therefore, further studies on the interpretability properties of DBGNs and GMDN are needed to understand the extent to which these models can provide humans with actionable feedback. Finally, in terms of applications, we foresee the use of our models in contexts of scarce supervision and large amounts of raw graphs, as it may happen in chemistry, biology, or social networks' analysis, where pre-trained embeddings can play a decisive role in making the most of the available data.

These final thoughts conclude the dissertation. All in all, we hope that the few contributions made in this thesis will inspire further studies and applications of both generative and predictive approaches for the adaptive processing of structured data.

\clearpage

\appendix

\chapter{List of Publications with Code}

\begin{itemize}
    \item \bibentry{bacciu_contextual_2018} \\
    {\small URL: \url{https://github.com/diningphil/CGMM}}
    \item \bibentry{errica_fair_2020} \\
    {\small URL: \url{https://github.com/diningphil/gnn-comparison}}
    \item \bibentry{bacciu_probabilistic_2020} \\
    {\small URL: \url{https://github.com/diningphil/CGMM}}
    \item \bibentry{errica_theoretically_2020}
    \item \bibentry{bacciu_gentle_2020} \\
    {\small URL: \url{https://github.com/diningphil/PyDGN}}
    \item \bibentry{errica_deep_2021} \\
    {\small URL: \url{https://github.com/CIML-VARIAMOLS/GRAWL}}
    \item \bibentry{carta_catastrophic_2021} \\
    {\small URL: \url{https://github.com/diningphil/continual_learning_for_graphs}}
    \item \bibentry{errica_graph_2021} \\
    {\small URL: \url{https://github.com/diningphil/graph-mixture-density-networks}}
    \item \bibentry{atzeni_modeling_2021} \\
    {\small URL: \url{https://github.com/diningphil/E-CGMM}}
    \item \bibentry{errica_concept_2021} \\
    {\small URL: \url{https://github.com/facebookresearch/parcus}}
    \item \bibentry{errica_robust_2021} \\
    {\small URL: \url{https://github.com/diningphil/robust-call-graph-malware-detection}}
\end{itemize}

\chapter{List of Talks and Posters}

\begin{itemize}
    \item Oral presentation of our ICML 2018 paper \cite{bacciu_contextual_2018}.
    \item Oral presentation of our ICLR 2020 paper \cite{errica_fair_2020}.
    \item Poster presentation of our ESANN 2020 paper \cite{errica_theoretically_2020}.
    \item Spotlight presentation of our WWW 2021 workshop paper \cite{errica_fair_2020}.
    \item Oral presentation of our ICML 2021 paper \cite{errica_graph_2021}.
    \item Oral presentations of our IJCNN 2021 papers \cite{atzeni_modeling_2021,errica_concept_2021}.
    \item Poster presentation of our ESANN 2021 paper \cite{errica_robust_2021}.
    \item Invited talk at IBM Research Zurich, 2021
    \item Invited talk at ContinualAI, 2021
    \item Invited talk at NEC Labs Europe, 2021
    
\end{itemize}


\addtocontents{toc}{\vspace{2em}} 





\addtocontents{toc}{\vspace{2em}}  
\backmatter

\label{Bibliography}
\lhead{\emph{Bibliography}}  
\bibliographystyle{unsrtnat}  
\bibliography{Bibliography,Frontiers,parallel_sampling,malware_detection,coates_biblio}  

\begin{thebibliography}{288}
\providecommand{\natexlab}[1]{#1}
\providecommand{\url}[1]{\texttt{#1}}
\expandafter\ifx\csname urlstyle\endcsname\relax
  \providecommand{\doi}[1]{doi: #1}\else
  \providecommand{\doi}{doi: \begingroup \urlstyle{rm}\Url}\fi

\bibitem[Bacciu et~al.(2020{\natexlab{a}})Bacciu, Errica, Micheli, and
  Podda]{bacciu_gentle_2020}
Davide Bacciu, Federico Errica, Alessio Micheli, and Marco Podda.
\newblock A gentle introduction to deep learning for graphs.
\newblock \emph{Neural Networks}, 129:\penalty0 203--221, 9 2020{\natexlab{a}}.

\bibitem[Sperduti and Starita(1997)]{sperduti_supervised_1997}
Alessandro Sperduti and Antonina Starita.
\newblock Supervised neural networks for the classification of structures.
\newblock \emph{IEEE Transactions on Neural Networks}, 8\penalty0 (3):\penalty0
  714--735, 1997.

\bibitem[Hochreiter and Schmidhuber(1997)]{hochreiter_long_1997}
Sepp Hochreiter and Jürgen Schmidhuber.
\newblock Long short-term memory.
\newblock \emph{Neural computation}, 9\penalty0 (8):\penalty0 1735--1780, 1997.

\bibitem[Frasconi et~al.(1998)Frasconi, Gori, and
  Sperduti]{frasconi_general_1998}
Paolo Frasconi, Marco Gori, and Alessandro Sperduti.
\newblock A general framework for adaptive processing of data structures.
\newblock \emph{IEEE Transactions on Neural Networks}, 9\penalty0 (5):\penalty0
  768--786, 1998.

\bibitem[Errica et~al.(2020{\natexlab{a}})Errica, Podda, Bacciu, and
  Micheli]{errica_fair_2020}
Federico Errica, Marco Podda, Davide Bacciu, and Alessio Micheli.
\newblock A fair comparison of graph neural networks for graph classification.
\newblock In \emph{8th {International} {Conference} on {Learning}
  {Representations} ({ICLR})}, 2020{\natexlab{a}}.

\bibitem[Bacciu et~al.(2018{\natexlab{a}})Bacciu, Errica, and
  Micheli]{bacciu_contextual_2018}
Davide Bacciu, Federico Errica, and Alessio Micheli.
\newblock Contextual {Graph} {Markov} {Model}: {A} deep and generative approach
  to graph processing.
\newblock In \emph{Proceedings of the 35th {International} {Conference} on
  {Machine} {Learning} ({ICML})}, volume~80, pages 294--303,
  2018{\natexlab{a}}.

\bibitem[Bacciu et~al.(2020{\natexlab{b}})Bacciu, Errica, and
  Micheli]{bacciu_probabilistic_2020}
Davide Bacciu, Federico Errica, and Alessio Micheli.
\newblock Probabilistic learning on graphs via contextual architectures.
\newblock \emph{Journal of Machine Learning Research}, 21\penalty0
  (134):\penalty0 1--39, 2020{\natexlab{b}}.

\bibitem[Daniele~Atzeni and Micheli(2021)]{atzeni_modeling_2021}
Federico~Errica Daniele~Atzeni, Davide~Bacciu and Alessio Micheli.
\newblock Modeling edge features with deep bayesian graph networks.
\newblock In \emph{Proceedings of the International Joint Conference on Neural
  Networks (IJCNN)}, pages 1--8, 2021.

\bibitem[Errica et~al.(2021{\natexlab{a}})Errica, Bacciu, and
  Micheli]{errica_graph_2021}
Federico Errica, Davide Bacciu, and Alessio Micheli.
\newblock Graph mixture density networks.
\newblock In \emph{Proceedings of the 38th International Conference on Machine
  Learning (ICML)}, pages 3025--3035, 2021{\natexlab{a}}.

\bibitem[Errica et~al.(2021{\natexlab{b}})Errica, Giulini, Bacciu, Menichetti,
  Micheli, and Potestio]{errica_deep_2021}
Federico Errica, Marco Giulini, Davide Bacciu, Roberto Menichetti, Alessio
  Micheli, and Raffaello Potestio.
\newblock A deep graph network–enhanced sampling approach to efficiently
  explore the space of reduced representations of proteins.
\newblock \emph{Frontiers in Molecular Biosciences}, 8:\penalty0 136--150,
  2021{\natexlab{b}}.

\bibitem[Errica et~al.(2021{\natexlab{c}})Errica, Iadarola, Martinelli,
  Mercaldo, and Micheli]{errica_robust_2021}
Federico Errica, Giacomo Iadarola, Fabio Martinelli, Francesco Mercaldo, and
  Alessio Micheli.
\newblock Robust malware classification via deep graph networks on call graph
  topologies.
\newblock In \emph{Proceedings of the 29th European Symposium on Artificial
  Neural Networks, Computational Intelligence and Machine Learning (ESANN)},
  2021{\natexlab{c}}.

\bibitem[Bishop(2006)]{bishop_pattern_2006}
Christopher~M Bishop.
\newblock \emph{Pattern recognition and machine learning}.
\newblock Springer, 2006.

\bibitem[Barber(2012)]{barber_bayesian_2012}
David Barber.
\newblock \emph{Bayesian reasoning and machine learning}.
\newblock Cambridge University Press, 2012.

\bibitem[Bruni and Montanari(2017)]{bruni_models_2017}
Roberto Bruni and Ugo Montanari.
\newblock \emph{Models of computation}.
\newblock Springer, 2017.

\bibitem[Dempster et~al.(1977)Dempster, Laird, and
  Rubin]{dempster_maximum_1977}
Arthur~P Dempster, Nan~M Laird, and Donald~B Rubin.
\newblock Maximum likelihood from incomplete data via the em algorithm.
\newblock \emph{Journal of the Royal Statistical Society: Series B
  (Methodological)}, 39\penalty0 (1):\penalty0 1--22, 1977.

\bibitem[Robbins and Monro(1951)]{robbins_stochastic_1951}
Herbert Robbins and Sutton Monro.
\newblock A stochastic approximation method.
\newblock \emph{The annals of mathematical statistics}, pages 400--407, 1951.

\bibitem[Geman and Geman(1984)]{geman_stochastic_1984}
Stuart Geman and Donald Geman.
\newblock Stochastic relaxation, gibbs distributions, and the bayesian
  restoration of images.
\newblock \emph{IEEE Transactions on pattern analysis and machine
  intelligence}, PAMI-6\penalty0 (6):\penalty0 721--741, 1984.

\bibitem[Bishop(1994)]{bishop_mixture_1994}
Christopher~M Bishop.
\newblock {Mixture Density Networks}.
\newblock Technical report, Aston University, 1994.

\bibitem[Rosenblatt(1957)]{rosenblatt_perceptron_1957}
Frank Rosenblatt.
\newblock \emph{The perceptron, a perceiving and recognizing automaton
  {Project} {Para}}.
\newblock Cornell Aeronautical Laboratory, 1957.

\bibitem[Rumelhart et~al.(1986)Rumelhart, Hinton, and
  Williams]{rumelhart_learning_1986}
Do~E Rumelhart, GE~Hinton, and RJ~Williams.
\newblock Learning internal representations by error propagation, parallel
  distributed processing, vol. 1.
\newblock \emph{Foundations. MIT Press, Cambridge}, 1986.

\bibitem[Hoppe(1984)]{hoppe_polya_1984}
Fred~M Hoppe.
\newblock P{\'o}lya-like urns and the ewens' sampling formula.
\newblock \emph{Journal of Mathematical Biology}, 20\penalty0 (1):\penalty0
  91--94, 1984.

\bibitem[Sethuraman(1994)]{sethuraman_constructive_1994}
Jayaram Sethuraman.
\newblock A constructive definition of dirichlet priors.
\newblock \emph{Statistica sinica}, pages 639--650, 1994.

\bibitem[Neal(2000)]{neal_markov_2000}
Radford~M Neal.
\newblock Markov chain sampling methods for dirichlet process mixture models.
\newblock \emph{Journal of computational and graphical statistics}, 9\penalty0
  (2):\penalty0 249--265, 2000.

\bibitem[Teh et~al.(2006)Teh, Jordan, Beal, and Blei]{teh_hdp_2006}
Yee~Whye Teh, Michael~I Jordan, Matthew~J Beal, and David~M Blei.
\newblock Hierarchical dirichlet processes.
\newblock \emph{Journal of the american statistical association}, 101\penalty0
  (476):\penalty0 1566--1581, 2006.

\bibitem[Orbanz and Teh(2010)]{orbanz_bayesian_2010}
Peter Orbanz and Yee~Whye Teh.
\newblock Bayesian nonparametric models.
\newblock \emph{Encyclopedia of machine learning}, 1, 2010.

\bibitem[Teh(2010)]{teh_dirichlet_2010}
Yee~Whye Teh.
\newblock \emph{Dirichlet Process}, pages 280--287.
\newblock Springer US, 2010.

\bibitem[Gershman and Blei(2012)]{gershman_tutorial_2012}
Samuel~J Gershman and David~M Blei.
\newblock A tutorial on bayesian nonparametric models.
\newblock \emph{Journal of Mathematical Psychology}, 56\penalty0 (1):\penalty0
  1--12, 2012.

\bibitem[{de Finetti}(1931)]{funzione_finetti_1931}
Bruno {de Finetti}.
\newblock Funzione caratteristica di un fenomeno aleatorio.
\newblock \emph{Atti della R. Accademia Nazionale dei Lincei, Ser. 6. Memorie,
  Classe di Scienze Fisiche, Matematiche e Naturali 4}, pages 251--299, 1931.

\bibitem[Ferguson(1973)]{ferguson_bayesian_1973}
Thomas~S Ferguson.
\newblock A bayesian analysis of some nonparametric problems.
\newblock \emph{The annals of statistics}, pages 209--230, 1973.

\bibitem[Aldous(1985)]{aldous_exchangeability_1985}
David~J Aldous.
\newblock Exchangeability and related topics.
\newblock In \emph{{\'E}cole d'{\'E}t{\'e} de Probabilit{\'e}s de Saint-Flour
  XIII—1983}, pages 1--198. Springer, 1985.

\bibitem[MacQueen et~al.(1967)]{macqueen_some_1967}
James MacQueen et~al.
\newblock Some methods for classification and analysis of multivariate
  observations.
\newblock In \emph{Proceedings of the 5th Berkeley symposium on mathematical
  statistics and probability}, pages 281--297. Oakland, CA, USA, 1967.

\bibitem[Bondy et~al.(1976)Bondy, Murty, et~al.]{bondy_graph_1976}
John~Adrian Bondy, Uppaluri Siva~Ramachandra Murty, et~al.
\newblock \emph{Graph theory with applications}, volume 290.
\newblock Macmillan London, 1976.

\bibitem[Cormen et~al.(2009)Cormen, Leiserson, Rivest, and
  Stein]{cormen_introduction_2009}
Thomas~H Cormen, Charles~E Leiserson, Ronald~L Rivest, and Clifford Stein.
\newblock \emph{Introduction to algorithms}.
\newblock MIT press, 2009.

\bibitem[Douglas(2011)]{douglas_weisfeiler-lehman_2011}
Brendan~L Douglas.
\newblock The {Weisfeiler}-{Lehman} method and graph isomorphism testing.
\newblock \emph{arXiv preprint arXiv:1101.5211}, 2011.

\bibitem[Bollob{\'a}s and B{\'e}la(2001)]{bollobas_random_2001}
B{\'e}la Bollob{\'a}s and Bollob{\'a}s B{\'e}la.
\newblock \emph{Random graphs}.
\newblock Cambridge university press, 2001.

\bibitem[Gilbert(1959)]{gilbert_random_1959}
Edgar~N Gilbert.
\newblock Random graphs.
\newblock \emph{The Annals of Mathematical Statistics}, 30\penalty0
  (4):\penalty0 1141--1144, 1959.

\bibitem[Erdos et~al.(1960)Erdos, R{\'e}nyi, et~al.]{erdos_evolution_1960}
Paul Erdos, Alfr{\'e}d R{\'e}nyi, et~al.
\newblock On the evolution of random graphs.
\newblock \emph{Publ. Math. Inst. Hung. Acad. Sci}, 5\penalty0 (1):\penalty0
  17--60, 1960.

\bibitem[Barab{\'a}si and Albert(1999)]{barabasi_emergence_1999}
Albert-L{\'a}szl{\'o} Barab{\'a}si and R{\'e}ka Albert.
\newblock Emergence of scaling in random networks.
\newblock \emph{Science}, 286\penalty0 (5439):\penalty0 509--512, 1999.

\bibitem[Bronstein et~al.(2017)Bronstein, Bruna, LeCun, Szlam, and
  Vandergheynst]{bronstein_geometric_2017}
Michael~M. Bronstein, Joan Bruna, Yann LeCun, Arthur Szlam, and Pierre
  Vandergheynst.
\newblock Geometric deep learning: going beyond {Euclidean} data.
\newblock \emph{IEEE Signal Processing Magazine}, 34\penalty0 (4):\penalty0 25.
  18--42, 2017.

\bibitem[Battaglia et~al.(2018)Battaglia, Hamrick, Bapst, Sanchez-Gonzalez,
  Zambaldi, Malinowski, Tacchetti, Raposo, Santoro, Faulkner, and
  {others}]{battaglia_relational_2018}
Peter~W Battaglia, Jessica~B Hamrick, Victor Bapst, Alvaro Sanchez-Gonzalez,
  Vinicius Zambaldi, Mateusz Malinowski, Andrea Tacchetti, David Raposo, Adam
  Santoro, Ryan Faulkner, and {others}.
\newblock Relational inductive biases, deep learning, and graph networks.
\newblock \emph{arXiv preprint arXiv:1806.01261}, 2018.

\bibitem[Wu et~al.(2020{\natexlab{a}})Wu, Pan, Chen, Long, Zhang, and
  Philip]{wu_comprehensive_2020}
Zonghan Wu, Shirui Pan, Fengwen Chen, Guodong Long, Chengqi Zhang, and S~Yu
  Philip.
\newblock A comprehensive survey on graph neural networks.
\newblock \emph{IEEE Transactions on Neural Networks and Learning Systems},
  2020{\natexlab{a}}.

\bibitem[Ji et~al.(2021)Ji, Pan, Cambria, Marttinen, and Yu]{ji_survey_2021}
Shaoxiong Ji, Shirui Pan, Erik Cambria, Pekka Marttinen, and Philip~S. Yu.
\newblock A survey on knowledge graphs: Representation, acquisition, and
  applications.
\newblock \emph{IEEE Transactions on Neural Networks and Learning Systems},
  pages 1--21, 2021.
\newblock \doi{10.1109/TNNLS.2021.3070843}.

\bibitem[Ralaivola et~al.(2005)Ralaivola, Swamidass, Saigo, and
  Baldi]{ralaivola_graph_2005}
Liva Ralaivola, Sanjay~J Swamidass, Hiroto Saigo, and Pierre Baldi.
\newblock Graph kernels for chemical informatics.
\newblock \emph{Neural Networks}, 18\penalty0 (8):\penalty0 1093--1110, 2005.

\bibitem[Vishwanathan et~al.(2010)Vishwanathan, Schraudolph, Kondor, and
  Borgwardt]{vishwanathan_graph_2010}
S~Vichy~N Vishwanathan, Nicol~N Schraudolph, Risi Kondor, and Karsten~M
  Borgwardt.
\newblock Graph kernels.
\newblock \emph{Journal of Machine Learning Research}, 11\penalty0
  (Apr):\penalty0 1201--1242, 2010.

\bibitem[Shervashidze et~al.(2011)Shervashidze, Schweitzer, Leeuwen, Mehlhorn,
  and Borgwardt]{shervashidze_weisfeiler-lehman_2011}
Nino Shervashidze, Pascal Schweitzer, Erik Jan~van Leeuwen, Kurt Mehlhorn, and
  Karsten~M Borgwardt.
\newblock Weisfeiler-lehman graph kernels.
\newblock \emph{Journal of Machine Learning Research}, 12\penalty0
  (Sep):\penalty0 2539--2561, 2011.

\bibitem[Frasconi et~al.(2014)Frasconi, Costa, De~Raedt, and
  De~Grave]{frasconi_klog_2014}
Paolo Frasconi, Fabrizio Costa, Luc De~Raedt, and Kurt De~Grave.
\newblock klog: {A} language for logical and relational learning with kernels.
\newblock \emph{Artificial Intelligence}, 217:\penalty0 117--143, 2014.

\bibitem[Yanardag and Vishwanathan(2015)]{yanardag_deep_2015}
Pinar Yanardag and SVN Vishwanathan.
\newblock Deep graph kernels.
\newblock In \emph{Proceedings of the 21th {International} {Conference} on
  {Knowledge} {Discovery} and {Data} {Mining} ({SIGKDD}}, pages 1365--1374.
  ACM, 2015.

\bibitem[Kriege et~al.(2020)Kriege, Johansson, and Morris]{kriege_survey_2020}
Nils~M Kriege, Fredrik~D Johansson, and Christopher Morris.
\newblock A survey on graph kernels.
\newblock \emph{Applied Network Science}, 5\penalty0 (1):\penalty0 1--42, 2020.

\bibitem[Morris et~al.(2019)Morris, Ritzert, Fey, Hamilton, Lenssen, Rattan,
  and Grohe]{morris_weisfeiler_2019}
Christopher Morris, Martin Ritzert, Matthias Fey, William~L Hamilton, Jan~Eric
  Lenssen, Gaurav Rattan, and Martin Grohe.
\newblock Weisfeiler and leman go neural: {Higher}-order graph neural networks.
\newblock In \emph{Proceedings of the 33rd {AAAI} {Conference} on {Artificial}
  {Intelligence} ({AAAI})}, volume~33, pages 4602--4609, 2019.

\bibitem[Da~San~Martino et~al.(2012)Da~San~Martino, Navarin, and
  Sperduti]{da_san_martino_tree-based_2012}
Giovanni Da~San~Martino, Nicolo Navarin, and Alessandro Sperduti.
\newblock A tree-based kernel for graphs.
\newblock In \emph{Proceedings of the 12th {International} {Conference} on
  {Data} {Mining} ({ICDM})}, pages 975--986. SIAM, 2012.

\bibitem[Da~San~Martino et~al.(2016)Da~San~Martino, Navarin, and
  Sperduti]{da_san_martino_ordered_2016}
Giovanni Da~San~Martino, Nicolò Navarin, and Alessandro Sperduti.
\newblock Ordered decompositional {DAG} kernels enhancements.
\newblock \emph{Neurocomputing}, 192:\penalty0 92--103, 2016.

\bibitem[Cortes and Vapnik(1995)]{cortes_support-vector_1995}
Corinna Cortes and Vladimir Vapnik.
\newblock Support-vector networks.
\newblock \emph{Machine Learning}, 20\penalty0 (3):\penalty0 273--297, 1995.

\bibitem[Trentin and Di~Iorio(2018)]{trentin_nonparametric_2018}
Edmondo Trentin and Ernesto Di~Iorio.
\newblock Nonparametric small random networks for graph-structured pattern
  recognition.
\newblock \emph{Neurocomputing}, 313:\penalty0 14--24, 2018.

\bibitem[Shervashidze et~al.(2009)Shervashidze, Vishwanathan, Petri, Mehlhorn,
  and Borgwardt]{shervashidze_efficient_2009}
Nino Shervashidze, SVN Vishwanathan, Tobias Petri, Kurt Mehlhorn, and Karsten
  Borgwardt.
\newblock Efficient graphlet kernels for large graph comparison.
\newblock In \emph{Proceedings of the 12th {Conference} on {Artificial}
  {Intelligence} and {Statistics} ({AISTATS})}, pages 488--495, 2009.

\bibitem[Koller et~al.(2007)Koller, Friedman, D{\v{z}}eroski, Sutton, McCallum,
  Pfeffer, Abbeel, Wong, Meek, Neville, et~al.]{koller_introduction_2007}
Daphne Koller, Nir Friedman, Sa{\v{s}}o D{\v{z}}eroski, Charles Sutton, Andrew
  McCallum, Avi Pfeffer, Pieter Abbeel, Ming-Fai Wong, Chris Meek, Jennifer
  Neville, et~al.
\newblock \emph{Introduction to statistical relational learning}.
\newblock MIT press, 2007.

\bibitem[Raedt and Kersting(2017)]{de_statistical_2010}
Luc~De Raedt and Kristian Kersting.
\newblock Statistical relational learning.
\newblock In \emph{Encyclopedia of Machine Learning and Data Mining}, pages
  1177--1187. Springer, 2017.

\bibitem[Richardson and Domingos(2006)]{richardson_markov_2006}
Matthew Richardson and Pedro Domingos.
\newblock Markov logic networks.
\newblock \emph{Machine learning}, 62\penalty0 (1-2):\penalty0 107--136, 2006.

\bibitem[Clifford(1990)]{clifford_markov_1990}
Peter Clifford.
\newblock Markov random fields in statistics.
\newblock \emph{Disorder in physical systems: A volume in honour of John M.
  Hammersley}, pages 19--32, 1990.

\bibitem[Lafferty et~al.(2001)Lafferty, McCallum, and
  Pereira]{lafferty_conditional_2001}
John~D. Lafferty, Andrew McCallum, and Fernando C.~N. Pereira.
\newblock Conditional random fields: Probabilistic models for segmenting and
  labeling sequence data.
\newblock In \emph{Proceedings of the 18th International Conference on Machine
  Learning {(ICML)}}, 2001.

\bibitem[Qu et~al.(2019)Qu, Bengio, and Tang]{qu_gmnn_2019}
Meng Qu, Yoshua Bengio, and Jian Tang.
\newblock {GMNN}: {Graph} {Markov} {Neural} {Networks}.
\newblock In \emph{Proceedings of the 36th {International} {Conference} on
  {Machine} {Learning} ({ICML})}, pages 5241--5250, 2019.

\bibitem[Sadhanala et~al.(2016)Sadhanala, Wang, and
  Tibshirani]{sadhanala_graph_2016}
Veeru Sadhanala, Yu-Xiang Wang, and Ryan Tibshirani.
\newblock Graph sparsification approaches for laplacian smoothing.
\newblock In \emph{Artificial {Intelligence} and {Statistics}}, pages
  1250--1259, 2016.

\bibitem[Chapelle et~al.(2006)Chapelle, Sch{\"o}lkopf, and
  Zien]{chapelle_semi-supervised_2006}
Olivier Chapelle, Bernhard Sch{\"o}lkopf, and Alexander Zien.
\newblock Semi-supervised learning.
\newblock \emph{{IEEE} {Transactions} on {Neural} {Networks}}, 20\penalty0
  (3):\penalty0 542--542, 2006.

\bibitem[Calandriello et~al.(2018)Calandriello, Koutis, Lazaric, and
  Valko]{calandriello_improved_2018}
Daniele Calandriello, Ioannis Koutis, Alessandro Lazaric, and Michal Valko.
\newblock Improved large-scale graph learning through ridge spectral
  sparsification.
\newblock In \emph{Proceedings of the 35th {International} {Conference} on
  {Machine} {Learning} ({ICML})}, pages 687--696, 2018.

\bibitem[Von~Luxburg(2007)]{von_luxburg_tutorial_2007}
Ulrike Von~Luxburg.
\newblock A tutorial on spectral clustering.
\newblock \emph{Statistics and Computing}, 17\penalty0 (4):\penalty0 395--416,
  2007.

\bibitem[Hammond et~al.(2011)Hammond, Vandergheynst, and
  Gribonval]{hammond_wavelets_2011}
David~K Hammond, Pierre Vandergheynst, and Rémi Gribonval.
\newblock Wavelets on graphs via spectral graph theory.
\newblock \emph{Applied and Computational Harmonic Analysis}, 30\penalty0
  (2):\penalty0 129--150, 2011.

\bibitem[Bruna et~al.(2014)Bruna, Zaremba, Szlam, and
  LeCun]{bruna_spectral_2014}
Joan Bruna, Wojciech Zaremba, Arthur Szlam, and Yann LeCun.
\newblock Spectral networks and locally connected networks on graphs.
\newblock \emph{2nd International Conference on Learning Representations
  (ICLR)}, 2014.

\bibitem[Blackledge(2005)]{blackledge_chapter_2005}
Jonathan~M. Blackledge.
\newblock Chapter 2 - 2d fourier theory.
\newblock In \emph{Digital Image Processing}, pages 30--49. Woodhead
  Publishing, 2005.

\bibitem[Shuman et~al.(2016)Shuman, Ricaud, and
  Vandergheynst]{shuman_vertex_2016}
David~I Shuman, Benjamin Ricaud, and Pierre Vandergheynst.
\newblock Vertex-frequency analysis on graphs.
\newblock \emph{Applied and Computational Harmonic Analysis}, 40\penalty0
  (2):\penalty0 260--291, 2016.

\bibitem[Defferrard et~al.(2016)Defferrard, Bresson, and
  Vandergheynst]{defferrard_convolutional_2016}
Michaël Defferrard, Xavier Bresson, and Pierre Vandergheynst.
\newblock Convolutional neural networks on graphs with fast localized spectral
  filtering.
\newblock In \emph{Proceedings of the 30th {Conference} on {Neural}
  {Information} {Processing} {Systems} ({NIPS})}, pages 3844--3852, 2016.

\bibitem[Kipf and Welling(2017)]{kipf_semi-supervised_2017}
Thomas~N Kipf and Max Welling.
\newblock Semi-supervised classification with graph convolutional networks.
\newblock In \emph{5th {International} {Conference} on {Learning}
  {Representations} ({ICLR})}, 2017.

\bibitem[Lovász and {others}(1993)]{lovasz_random_1993}
László Lovász and {others}.
\newblock Random walks on graphs: {A} survey.
\newblock \emph{Combinatorics, Paul Erdos is eighty}, 2\penalty0 (1):\penalty0
  1--46, 1993.

\bibitem[Ribeiro et~al.(2017)Ribeiro, Saverese, and
  Figueiredo]{ribeiro_struc2vec_2017}
Leonardo~FR Ribeiro, Pedro~HP Saverese, and Daniel~R Figueiredo.
\newblock struc2vec: {Learning} node representations from structural identity.
\newblock In \emph{Proceedings of the 23rd {International} {Conference} on
  {Knowledge} {Discovery} and {Data} {Mining} ({SIGKDD})}, pages 385--394. ACM,
  2017.

\bibitem[Ivanov and Burnaev(2018)]{ivanov_anonymous_2018}
Sergey Ivanov and Evgeny Burnaev.
\newblock Anonymous walk embeddings.
\newblock In \emph{Proceedings of the 35th {International} {Conference} on
  {Machine} {Learning} ({ICML})}, pages 2191--2200, 2018.

\bibitem[Grover and Leskovec(2016)]{grover_node2vec_2016}
Aditya Grover and Jure Leskovec.
\newblock node2vec: {Scalable} feature learning for networks.
\newblock In \emph{Proceedings of the 22nd {International} {Conference} on
  {Knowledge} {Discovery} and {Data} {Mining} ({SIGKDD})}, pages 855--864. ACM,
  2016.

\bibitem[Perozzi et~al.(2014)Perozzi, Al-Rfou, and
  Skiena]{perozzi_deepwalk_2014}
Bryan Perozzi, Rami Al-Rfou, and Steven Skiena.
\newblock Deepwalk: {Online} learning of social representations.
\newblock In \emph{Proceedings of the 20th {International} {Conference} on
  {Knowledge} {Discovery} and {Data} {Mining} ({SIGKDD})}, pages 701--710. ACM,
  2014.

\bibitem[Bojchevski et~al.(2018)Bojchevski, Shchur, Zügner, and
  Günnemann]{bojchevski_netgan_2018}
Aleksandar Bojchevski, Oleksandr Shchur, Daniel Zügner, and Stephan
  Günnemann.
\newblock {NetGAN}: {Generating} graphs via random walks.
\newblock In \emph{Proceedings of the 35th {International} {Conference} on
  {Machine} {Learning} ({ICML})}, pages 609--618, 2018.

\bibitem[Xu et~al.(2018)Xu, Li, Tian, Sonobe, Kawarabayashi, and
  Jegelka]{xu_representation_2018}
Keyulu Xu, Chengtao Li, Yonglong Tian, Tomohiro Sonobe, Ken-ichi Kawarabayashi,
  and Stefanie Jegelka.
\newblock Representation learning on graphs with jumping knowledge networks.
\newblock \emph{Proceedings of the 35th International Conference on Machine
  Learning (ICML)}, pages 5453--5462, 2018.

\bibitem[Jang et~al.(2017)Jang, Gu, and Poole]{jang_categorical_2017}
Eric Jang, Shixiang Gu, and Ben Poole.
\newblock Categorical reparametrization with gumbel-softmax.
\newblock In \emph{5th {International} {Conference} on {Learning}
  {Representations} ({ICLR})}, 2017.

\bibitem[Simonovsky and Komodakis(2018)]{simonovsky_graphvae_2018}
Martin Simonovsky and Nikos Komodakis.
\newblock {GraphVAE}: {Towards} generation of small graphs using variational
  autoencoders.
\newblock In \emph{Proceedings of the 27th {International} {Conference} on
  {Artificial} {Neural} {Networks} ({ICANN})}, pages 412--422, 2018.

\bibitem[De~Cao and Kipf(2018)]{de_cao_molgan_2018}
Nicola De~Cao and Thomas Kipf.
\newblock {MolGAN}: {An} implicit generative model for small molecular graphs.
\newblock \emph{Workshop on Theoretical Foundations and Applications of Deep
  Generative Models, International Conference on Machine Learning (ICML)},
  2018.

\bibitem[Kipf and Welling(2016)]{kipf_variational_2016}
Thomas~N Kipf and Max Welling.
\newblock Variational graph auto-encoders.
\newblock In \emph{Workshop on {Bayesian} {Deep} {Learning}, {Neural}
  {Information} {Processing} {System} ({NIPS})}, 2016.

\bibitem[Grover et~al.(2019)Grover, Zweig, and Ermon]{grover_graphite_2019}
Aditya Grover, Aaron Zweig, and Stefano Ermon.
\newblock Graphite: {Iterative} generative modeling of graphs.
\newblock In \emph{Proceedings of the 36th {International} {Conference} on
  {Machine} {Learning} ({ICML})}, pages 2434--2444, 2019.

\bibitem[Kingma and Welling(2014)]{kingma_auto-encoding_2014}
Diederik~P Kingma and Max Welling.
\newblock Auto-encoding variational {Bayes}.
\newblock \emph{2nd International Conference on Learning Representations
  (ICLR)}, 2014.

\bibitem[Tolstikhin et~al.(2018)Tolstikhin, Bousquet, Gelly, and
  Schoelkopf]{tolstikhin_wasserstein_2018}
Ilya Tolstikhin, Olivier Bousquet, Sylvain Gelly, and Bernhard Schoelkopf.
\newblock Wasserstein auto-encoders.
\newblock In \emph{6th {International} {Conference} on {Learning}
  {Representations} ({ICLR})}, 2018.

\bibitem[Liu et~al.(2018)Liu, Allamanis, Brockschmidt, and
  Gaunt]{liu_constrained_2018}
Qi~Liu, Miltiadis Allamanis, Marc Brockschmidt, and Alexander Gaunt.
\newblock Constrained graph variational autoencoders for molecule design.
\newblock In \emph{Proceedings of the 32nd {Conference} on {Neural}
  {Information} {Processing} {Systems} ({NeurIPS})}, pages 7795--7804, 2018.

\bibitem[Samanta et~al.(2019)Samanta, De, Jana, Chattaraj, Ganguly, and
  Rodriguez]{samanta_nevae_2019}
Bidisha Samanta, Abir De, Gourhari Jana, Pratim~Kumar Chattaraj, Niloy Ganguly,
  and Manuel~Gomez Rodriguez.
\newblock {NeVAE}: {A} deep generative model for molecular graphs.
\newblock In \emph{Proceedings of the 33rd {AAAI} {Conference} on {Artificial}
  {Intelligence} ({AAAI})}, pages 1110--1117, 2019.

\bibitem[Bradshaw et~al.(2019)Bradshaw, Paige, Kusner, Segler, and
  Hernández-Lobato]{bradshaw_model_2019}
John Bradshaw, Brooks Paige, Matt~J Kusner, Marwin Segler, and José~Miguel
  Hernández-Lobato.
\newblock A model to search for synthesizable molecules.
\newblock In \emph{Proceedings of the 33rd {Conference} on {Neural}
  {Information} {Processing} {Systems} ({NeurIPS})}, pages 7935--7947, 2019.

\bibitem[Goodfellow et~al.(2014)Goodfellow, Pouget-Abadie, Mirza, Xu,
  Warde-Farley, Ozair, Courville, and Bengio]{goodfellow_generative_2014}
Ian Goodfellow, Jean Pouget-Abadie, Mehdi Mirza, Bing Xu, David Warde-Farley,
  Sherjil Ozair, Aaron Courville, and Yoshua Bengio.
\newblock Generative {Adversarial} {Nets}.
\newblock In \emph{Proceedings of the 28th {Conference} on {Neural}
  {Information} {Processing} {Systems} ({NIPS})}, pages 2672--2680, 2014.

\bibitem[Fan and Huang(2019)]{fan_conditional_2019}
S.~Fan and B.~Huang.
\newblock Conditional labeled graph generation with {GANs}.
\newblock In \emph{{Representation} {Learning} on {Graphs} and {Manifolds}
  Workshop, {International} {Conference} on {Learning} {Representations}
  ({ICLR})}, 2019.

\bibitem[Wang et~al.(2018)Wang, Wang, Wang, Zhao, Zhang, Zhang, Xie, and
  Guo]{wang_graphgan_2018}
Hongwei Wang, Jia Wang, Jialin Wang, Miao Zhao, Weinan Zhang, Fuzheng Zhang,
  Xing Xie, and Minyi Guo.
\newblock {GraphGAN}: {Graph} representation learning with generative
  adversarial nets.
\newblock In \emph{Proceedings of the 32nd {AAAI} {Conference} on {Artificial}
  {Intelligence} ({AAAI})}, pages 2508--2515, 2018.

\bibitem[Pan et~al.(2018)Pan, Hu, Long, Jiang, Yao, and
  Zhang]{pan_adversarially_2018}
Shirui Pan, Ruiqi Hu, Guodong Long, Jing Jiang, Lina Yao, and Chengqi Zhang.
\newblock Adversarially regularized graph autoencoder for graph embedding.
\newblock In \emph{Proceedings of the 27th International Joint Conference on
  Artificial Intelligence {(IJCAI)}}, pages 2609--2615, 2018.

\bibitem[Li et~al.(2018{\natexlab{a}})Li, Vinyals, Dyer, Pascanu, and
  Battaglia]{li_learning_2018}
Yujia Li, Oriol Vinyals, Chris Dyer, Razvan Pascanu, and Peter~W. Battaglia.
\newblock Learning deep generative models of graphs.
\newblock \emph{CoRR}, abs/1803.03324, 2018{\natexlab{a}}.

\bibitem[You et~al.(2018)You, Ying, Ren, Hamilton, and
  Leskovec]{you_graphrnn_2018}
Jiaxuan You, Rex Ying, Xiang Ren, William~L. Hamilton, and Jure Leskovec.
\newblock {GraphRNN}: {Generating} realistic graphs with deep auto-regressive
  models.
\newblock In \emph{Proceedings of the 35th {International} {Conference} on
  {Machine} {Learning} ({ICML})}, 2018.

\bibitem[Bacciu et~al.(2019{\natexlab{a}})Bacciu, Micheli, and
  Podda]{bacciu_graph_2019}
Davide Bacciu, Alessio Micheli, and Marco Podda.
\newblock Graph generation by sequential edge prediction.
\newblock In \emph{Proceedings of the 27th {European} {Symposium} on
  {Artificial} {Neural} {Networks}, {Computational} {Intelligence} and
  {Machine} {Learning} ({ESANN})}, 2019{\natexlab{a}}.

\bibitem[Bacciu et~al.(2019{\natexlab{b}})Bacciu, Micheli, and
  Podda]{bacciu_edge-based_2019}
Davide Bacciu, Alessio Micheli, and Marco Podda.
\newblock Edge-based sequential graph generation with recurrent neural
  networks.
\newblock \emph{Neurocomputing. Accepted}, 2019{\natexlab{b}}.

\bibitem[Podda et~al.(2020)Podda, Bacciu, and Micheli]{podda_deep_2020}
Marco Podda, Davide Bacciu, and Alessio Micheli.
\newblock A deep generative model for fragment-based molecule generation.
\newblock In \emph{Proceedings of the 23rd Conference on Artificial
  Intelligence and Statistics (AISTATS)}, pages 2240--2250, 2020.

\bibitem[Podda(2021)]{podda_thesis_2021}
Marco Podda.
\newblock \emph{Deep Learning on Graphs with Applications to the Life
  Sciences}.
\newblock PhD thesis, University of Pisa, 2021.

\bibitem[Zügner et~al.(2018)Zügner, Akbarnejad, and
  Günnemann]{zugner_adversarial_2018}
Daniel Zügner, Amir Akbarnejad, and Stephan Günnemann.
\newblock Adversarial attacks on neural networks for graph data.
\newblock In \emph{Proceedings of the 24th {International} {Conference} on
  {Knowledge} {Discovery} and {Data} {Mining} ({SIGKDD})}, pages 2847--2856.
  ACM, 2018.

\bibitem[Feng et~al.(2019)Feng, He, Tang, and Chua]{feng_graph_2019}
Fuli Feng, Xiangnan He, Jie Tang, and Tat-Seng Chua.
\newblock Graph adversarial training: {Dynamically} regularizing based on graph
  structure.
\newblock \emph{IEEE Transactions on Knowledge and Data Engineering}, 2019.

\bibitem[Z{\"u}gner and G{\"u}nnemann(2019)]{zugner_certifiable_2019}
Daniel Z{\"u}gner and Stephan G{\"u}nnemann.
\newblock Certifiable robustness and robust training for graph convolutional
  networks.
\newblock In \emph{Proceedings of the 25th ACM International Conference on
  Knowledge Discovery and Data Mining (SIGKDD)}, pages 246--256, 2019.

\bibitem[Bojchevski and G\"{u}nnemann(2019)]{bojchevski_certifiable_2019}
Aleksandar Bojchevski and Stephan G\"{u}nnemann.
\newblock Certifiable robustness to graph perturbations.
\newblock In \emph{Proceedings of the 33rd {Conference} on {Neural}
  {Information} {Processing} {Systems} ({NeurIPS})}, 2019.

\bibitem[Yang et~al.(2019)Yang, Kang, Cao, Jin, Yang, and
  Guo]{yang_topology_2019}
Liang Yang, Zesheng Kang, Xiaochun Cao, Di~Jin, Bo~Yang, and Yuanfang Guo.
\newblock Topology optimization based graph convolutional network.
\newblock In \emph{Proceedings of the 28th {International} {Joint} {Conference}
  on {Artificial} {Intelligence} ({IJCAI})}, pages 4054--4061, 2019.

\bibitem[Jin and Zhang(2019)]{jin_latent_2019}
H.~Jin and X.~Zhang.
\newblock Latent adversarial training of graph convolution networks.
\newblock In \emph{Workshop on {Learning} and {Reasoning} with
  {Graph}-{Structured} {Representations}, {International} {Conference} on
  {Machine} {Learning} ({ICML})}, 2019.

\bibitem[Z{\"u}gner and G{\"u}nnemann(2020)]{zugner_certifiable_2020}
Daniel Z{\"u}gner and Stephan G{\"u}nnemann.
\newblock Certifiable robustness of graph convolutional networks under
  structure perturbations.
\newblock In \emph{Proceedings of the 26th ACM International Conference on
  Knowledge Discovery and Data Mining (SIGKDD)}, pages 1656--1665, 2020.

\bibitem[Zhang et~al.(2021)Zhang, Regol, Pal, Khan, Ma, and
  Coates]{zhang_detection_2021}
Yingxue Zhang, Florence Regol, Soumyasundar Pal, Sakif Khan, Liheng Ma, and
  Mark Coates.
\newblock Detection and defense of topological adversarial attacks on graphs.
\newblock In \emph{International Conference on Artificial Intelligence and
  Statistics}, pages 2989--2997. PMLR, 2021.

\bibitem[Bose and Hamilton(2019)]{bose_compositional_2019}
Avishek Bose and William Hamilton.
\newblock Compositional fairness constraints for graph embeddings.
\newblock In \emph{Proceedings of the 36th International Conference on Machine
  Learning (ICML)}, pages 715--724, 2019.

\bibitem[Micheli(2009)]{micheli_neural_2009}
Alessio Micheli.
\newblock Neural network for graphs: {A} contextual constructive approach.
\newblock \emph{IEEE Transactions on Neural Networks}, 20\penalty0
  (3):\penalty0 498--511, 2009.

\bibitem[Loukas(2020{\natexlab{a}})]{loukas_what_2020}
Andreas Loukas.
\newblock What graph neural networks cannot learn: depth vs width.
\newblock In \emph{8th {International} {Conference} on {Learning}
  {Representations} ({ICLR})}, 2020{\natexlab{a}}.

\bibitem[Xu et~al.(2019)Xu, Hu, Leskovec, and Jegelka]{xu_how_2019}
Keyulu Xu, Weihua Hu, Jure Leskovec, and Stefanie Jegelka.
\newblock How powerful are graph neural networks?
\newblock In \emph{7th {International} {Conference} on {Learning}
  {Representations} ({ICLR})}, 2019.

\bibitem[Vignac et~al.(2020)Vignac, Loukas, and Frossard]{vignac_building_2020}
Cl\'{e}ment Vignac, Andreas Loukas, and Pascal Frossard.
\newblock Building powerful and equivariant graph neural networks with
  structural message-passing.
\newblock In \emph{Proceedings of the 34th {Conference} on {Neural}
  {Information} {Processing} {Systems} ({NeurIPS})}, 2020.

\bibitem[Loukas(2020{\natexlab{b}})]{loukas_how_2020}
Andreas Loukas.
\newblock How hard is to distinguish graphs with graph neural networks?
\newblock In \emph{Proceedings of the 34th {Conference} on {Neural}
  {Information} {Processing} {Systems} ({NeurIPS})}, pages 3465--3476,
  2020{\natexlab{b}}.

\bibitem[Maron et~al.(2019)Maron, Ben-Hamu, Serviansky, and
  Lipman]{maron_provably_2019}
Haggai Maron, Heli Ben-Hamu, Hadar Serviansky, and Yaron Lipman.
\newblock Provably powerful graph networks.
\newblock In H.~Wallach, H.~Larochelle, A.~Beygelzimer, F.~d\textquotesingle
  Alch\'{e}-Buc, E.~Fox, and R.~Garnett, editors, \emph{Proceedings of the 33rd
  {Conference} on {Neural} {Information} {Processing} {Systems} ({NeurIPS})},
  2019.

\bibitem[Morris et~al.(2020)Morris, Rattan, and Mutzel]{morris_weisfeiler_2020}
Christopher Morris, Gaurav Rattan, and Petra Mutzel.
\newblock Weisfeiler and leman go sparse: Towards scalable higher-order graph
  embeddings.
\newblock In \emph{Proceedings of the 34th {Conference} on {Neural}
  {Information} {Processing} {Systems} ({NeurIPS})}, 2020.

\bibitem[Geerts et~al.(2021)Geerts, Mazowiecki, and Perez]{geerts_let_2021}
Floris Geerts, Filip Mazowiecki, and Guillermo Perez.
\newblock Let’s agree to degree: Comparing graph convolutional networks in
  the message-passing framework.
\newblock In \emph{Proceedings of the 38th {International} {Conference} on
  {Machine} {Learning} ({ICML})}, pages 3640--3649, 2021.

\bibitem[Hammer et~al.(2005)Hammer, Micheli, and
  Sperduti]{hammer_universal_2005}
Barbara Hammer, Alessio Micheli, and Alessandro Sperduti.
\newblock Universal approximation capability of cascade correlation for
  structures.
\newblock \emph{Neural Computation}, 17\penalty0 (5):\penalty0 1109--1159,
  2005.

\bibitem[Zaheer et~al.(2017)Zaheer, Kottur, Ravanbakhsh, Poczos, Salakhutdinov,
  and Smola]{zaheer_deep_2017}
Manzil Zaheer, Satwik Kottur, Siamak Ravanbakhsh, Barnabas Poczos, Ruslan~R
  Salakhutdinov, and Alexander~J Smola.
\newblock Deep sets.
\newblock In \emph{Proceedings of the 31st {Conference} on {Neural}
  {Information} {Processing} {Systems} ({NIPS})}, pages 3391--3401, 2017.

\bibitem[Wagstaff et~al.(2019)Wagstaff, Fuchs, Engelcke, Posner, and
  Osborne]{wagstaff_limitations_2019}
Edward Wagstaff, Fabian~B Fuchs, Martin Engelcke, Ingmar Posner, and Michael
  Osborne.
\newblock On the limitations of representing functions on sets.
\newblock In \emph{Proceedings of the 36th {International} {Conference} on
  {Machine} {Learning} ({ICML})}, pages 6487--6494, 2019.

\bibitem[Bacciu et~al.(2013)Bacciu, Micheli, and
  Sperduti]{bacciu_inputoutput_2013}
Davide Bacciu, Alessio Micheli, and Alessandro Sperduti.
\newblock An input–output hidden {Markov} model for tree transductions.
\newblock \emph{Neurocomputing}, 112:\penalty0 34--46, 2013.

\bibitem[Bengio et~al.(2013)Bengio, Courville, and
  Vincent]{bengio_representation_2013}
Yoshua Bengio, Aaron Courville, and Pascal Vincent.
\newblock Representation learning: {A} review and new perspectives.
\newblock \emph{IEEE Transactions on Pattern Analysis and Machine
  Intelligence}, 35\penalty0 (8):\penalty0 1798--1828, 2013.

\bibitem[Goodfellow et~al.(2016)Goodfellow, Bengio, and
  Courville]{goodfellow_deep_2016}
Ian Goodfellow, Yoshua Bengio, and Aaron Courville.
\newblock \emph{Deep learning}.
\newblock MIT press, 2016.

\bibitem[Bianucci et~al.(2000)Bianucci, Micheli, Sperduti, and
  Starita]{bianucci_application_2000}
Anna~Maria Bianucci, Alessio Micheli, Alessandro Sperduti, and Antonina
  Starita.
\newblock Application of cascade correlation networks for structures to
  chemistry.
\newblock \emph{Applied Intelligence}, 12\penalty0 (1-2):\penalty0 117--147,
  2000.

\bibitem[Diligenti et~al.(2003)Diligenti, Frasconi, and
  Gori]{diligenti_hidden_2003}
Michelangelo Diligenti, Paolo Frasconi, and Marco Gori.
\newblock Hidden tree {Markov} models for document image classification.
\newblock \emph{IEEE Transactions on Pattern Analysis and Machine
  Intelligence}, 25\penalty0 (4):\penalty0 519--523, 2003.

\bibitem[Micheli et~al.(2004)Micheli, Sona, and
  Sperduti]{micheli_contextual_2004}
Alessio Micheli, Diego Sona, and Alessandro Sperduti.
\newblock Contextual processing of structured data by recursive cascade
  correlation.
\newblock \emph{IEEE Transactions on Neural Networks}, 15\penalty0
  (6):\penalty0 1396--1410, 2004.

\bibitem[Bacciu et~al.(2012{\natexlab{a}})Bacciu, Micheli, and
  Sperduti]{bacciu_compositional_2012}
Davide Bacciu, Alessio Micheli, and Alessandro Sperduti.
\newblock Compositional generative mapping for tree-structured data - part {I}:
  {Bottom}-up probabilistic modeling of trees.
\newblock \emph{IEEE Transactions on Neural Networks and Learning Systems},
  23\penalty0 (12):\penalty0 1987--2002, 2012{\natexlab{a}}.

\bibitem[Bacciu et~al.(2014)Bacciu, Micheli, and
  Sperduti]{bacciu_modeling_2014}
Davide Bacciu, Alessio Micheli, and Alessandro Sperduti.
\newblock Modeling bi-directional tree contexts by generative transductions.
\newblock In \emph{Proceedings of the 21st {International} {Conference} on
  {Neural} {Information} {Processing} ({ICONIP})}, pages 543--550. Springer,
  2014.

\bibitem[Baum and Petrie(1966)]{baum_statistical_1966}
Leonard~E Baum and Ted Petrie.
\newblock Statistical inference for probabilistic functions of finite state
  markov chains.
\newblock \emph{The annals of mathematical statistics}, 37\penalty0
  (6):\penalty0 1554--1563, 1966.

\bibitem[Werbos(1990)]{werbos_backpropagation_1990}
Paul~J Werbos.
\newblock Backpropagation through time: what it does and how to do it.
\newblock \emph{Proceedings of the IEEE}, 78\penalty0 (10):\penalty0
  1550--1560, 1990.

\bibitem[Goller and Kuchler(1996)]{goller_learning_1996}
Christoph Goller and Andreas Kuchler.
\newblock Learning task-dependent distributed representations by
  backpropagation through structure.
\newblock In \emph{Proceedings of the {International} {Conference} on {Neural}
  {Networks} ({ICNN})}, volume~1, pages 347--352. IEEE, 1996.

\bibitem[Bacciu et~al.(2012{\natexlab{b}})Bacciu, Micheli, and
  Sperduti]{bacciu_generative_2012}
Davide Bacciu, Alessio Micheli, and Alessandro Sperduti.
\newblock A generative multiset kernel for structured data.
\newblock In \emph{Proceedings of the {International} {Conference} on
  {Artificial} {Neural} {Networks} ({ICANN})}, pages 57--64. Springer,
  2012{\natexlab{b}}.

\bibitem[Scarselli et~al.(2009)Scarselli, Gori, Tsoi, Hagenbuchner, and
  Monfardini]{scarselli_graph_2009}
Franco Scarselli, Marco Gori, Ah~Chung Tsoi, Markus Hagenbuchner, and Gabriele
  Monfardini.
\newblock The graph neural network model.
\newblock \emph{IEEE Transactions on Neural Networks}, 20\penalty0
  (1):\penalty0 61--80, 2009.

\bibitem[Kipf et~al.(2018)Kipf, Fetaya, Wang, Welling, and
  Zemel]{kipf_neural_2018}
Thomas Kipf, Ethan Fetaya, Kuan-Chieh Wang, Max Welling, and Richard Zemel.
\newblock Neural relational inference for interacting systems.
\newblock In \emph{Proceedings of the 35th {International} {Conference} on
  {Machine} {Learning} ({ICML})}, pages 2688--2697, 2018.

\bibitem[Cybenko(1989)]{cybenko_approximation_1989}
George Cybenko.
\newblock Approximation by superpositions of a sigmoidal function.
\newblock \emph{Mathematics of control, signals and systems}, 2\penalty0
  (4):\penalty0 303--314, 1989.

\bibitem[Angluin(1980)]{angluin_local_1980}
Dana Angluin.
\newblock Local and global properties in networks of processors.
\newblock In \emph{Proceedings of the 12th annual {ACM} symposium on {Theory}
  of computing}, pages 82--93, 1980.

\bibitem[Linial(1992)]{linial_locality_1992}
Nathan Linial.
\newblock Locality in distributed graph algorithms.
\newblock \emph{SIAM {Journal} on {Computing}}, 21\penalty0 (1):\penalty0
  193--201, 1992.

\bibitem[Naor and Stockmeyer(1995)]{naor_can_1995}
Moni Naor and Larry Stockmeyer.
\newblock What can be computed locally?
\newblock \emph{SIAM {Journal} on {Computing}}, 24\penalty0 (6):\penalty0
  1259--1277, 1995.

\bibitem[Peleg(2000)]{peleg_distributed_2000}
David Peleg.
\newblock Distributed computing.
\newblock \emph{SIAM {Monographs} on discrete mathematics and applications},
  5:\penalty0 1--1, 2000.

\bibitem[Gilmer et~al.(2017)Gilmer, Schoenholz, Riley, Vinyals, and
  Dahl]{gilmer_neural_2017}
Justin Gilmer, Samuel~S Schoenholz, Patrick~F Riley, Oriol Vinyals, and
  George~E Dahl.
\newblock Neural message passing for quantum chemistry.
\newblock In \emph{Proceedings of the 34th {International} {Conference} on
  {Machine} {Learning} ({ICML})}, pages 1263--1272, 2017.

\bibitem[LeCun et~al.(1995)LeCun, Bengio, and
  {others}]{lecun_convolutional_1995}
Yann LeCun, Yoshua Bengio, and {others}.
\newblock Convolutional networks for images, speech, and time series.
\newblock \emph{The Handbook of Brain Theory and Neural Networks},
  3361\penalty0 (10):\penalty0 1995, 1995.

\bibitem[Li et~al.(2016)Li, Tarlow, Brockschmidt, and Zemel]{li_gated_2016}
Yujia Li, Daniel Tarlow, Marc Brockschmidt, and Richard~S. Zemel.
\newblock Gated {Graph} {Sequence} {Neural} {Networks}.
\newblock In \emph{4th {International} {Conference} on {Learning}
  {Representations}, ({ICLR})}, 2016.

\bibitem[Gallicchio and Micheli(2020)]{gallicchio_fast_2020}
Claudio Gallicchio and Alessio Micheli.
\newblock Fast and deep graph neural networks.
\newblock In \emph{Proceedings of the 34th {AAAI} {Conference} on {Artificial}
  {Intelligence} ({AAAI})}, 2020.

\bibitem[Velickovic et~al.(2018)Velickovic, Cucurull, Casanova, Romero, Lio,
  and Bengio]{velickovic_graph_2018}
Petar Velickovic, Guillem Cucurull, Arantxa Casanova, Adriana Romero, Pietro
  Lio, and Yoshua Bengio.
\newblock Graph attention networks.
\newblock In \emph{6th {International} {Conference} on {Learning}
  {Representations} ({ICLR})}, 2018.

\bibitem[Bengio et~al.(1994)Bengio, Simard, and Frasconi]{bengio_learning_1994}
Yoshua Bengio, Patrice Simard, and Paolo Frasconi.
\newblock Learning long-term dependencies with gradient descent is difficult.
\newblock \emph{IEEE Transactions on Neural Networks}, 5\penalty0 (2):\penalty0
  157--166, 1994.

\bibitem[Li et~al.(2018{\natexlab{b}})Li, Han, and Wu]{li_deeper_2018}
Qimai Li, Zhichao Han, and Xiao-Ming Wu.
\newblock Deeper insights into graph convolutional networks for semi-supervised
  learning.
\newblock In \emph{Proceedings of the 32nd {AAAI} {Conference} on {Artificial}
  {Intelligence} ({AAAI})}, 2018{\natexlab{b}}.

\bibitem[Fey and Lenssen(2019)]{fey_fast_2019}
Matthias Fey and Jan~Eric Lenssen.
\newblock Fast graph representation learning with {PyTorch} {Geometric}.
\newblock \emph{Representation Learning on Graphs and Manifolds Workshop,
  International Conference on Learning Representations (ICLR)}, 2019.

\bibitem[Paszke et~al.(2019)Paszke, Gross, Massa, Lerer, Bradbury, Chanan,
  Killeen, Lin, Gimelshein, Antiga, Desmaison, Kopf, Yang, DeVito, Raison,
  Tejani, Chilamkurthy, Steiner, Fang, Bai, and Chintala]{paszke_pytorch_2019}
Adam Paszke, Sam Gross, Francisco Massa, Adam Lerer, James Bradbury, Gregory
  Chanan, Trevor Killeen, Zeming Lin, Natalia Gimelshein, Luca Antiga, Alban
  Desmaison, Andreas Kopf, Edward Yang, Zachary DeVito, Martin Raison, Alykhan
  Tejani, Sasank Chilamkurthy, Benoit Steiner, Lu~Fang, Junjie Bai, and Soumith
  Chintala.
\newblock Pytorch: An imperative style, high-performance deep learning library.
\newblock In \emph{Proceedings of the 33rd {Conference} on {Neural}
  {Information} {Processing} {Systems} ({NeurIPS})}, pages 8026--8037, 2019.

\bibitem[Fahlman and Lebiere(1990)]{fahlman_cascade-correlation_1990}
Scott~E. Fahlman and Christian Lebiere.
\newblock The {Cascade}-{Correlation} learning architecture.
\newblock In \emph{Proceedings of the 3rd {Conference} on {Neural}
  {Information} {Processing} {Systems} ({NIPS})}, pages 524--532, 1990.

\bibitem[Schlichtkrull et~al.(2018)Schlichtkrull, Kipf, Bloem, van~den Berg,
  Titov, and Welling]{schlichtkrull_modeling_2018}
Michael Schlichtkrull, Thomas~N Kipf, Peter Bloem, Rianne van~den Berg, Ivan
  Titov, and Max Welling.
\newblock Modeling relational data with graph convolutional networks.
\newblock In \emph{Proceedings of the 15th {European} {Semantic} {Web}
  {Conference} ({ESWC})}, pages 593--607. Springer, 2018.

\bibitem[Simonovsky and Komodakis(2017)]{simonovsky_dynamic_2017}
Martin Simonovsky and Nikos Komodakis.
\newblock Dynamic edge-conditioned filters in convolutional neural networks on
  graphs.
\newblock In \emph{Proceedings of the {IEEE} {Conference} on {Computer}
  {Vision} and {Pattern} {Recognition} ({CVPR})}, pages 3693--3702, 2017.

\bibitem[Vaswani et~al.(2017)Vaswani, Shazeer, Parmar, Uszkoreit, Jones, Gomez,
  Kaiser, and Polosukhin]{vaswani_attention_2017}
Ashish Vaswani, Noam Shazeer, Niki Parmar, Jakob Uszkoreit, Llion Jones,
  Aidan~N Gomez, Lukasz Kaiser, and Illia Polosukhin.
\newblock Attention is all you need.
\newblock In \emph{Proceedings of the 31st {Conference} on {Neural}
  {Information} {Processing} {Systems} ({NIPS})}, pages 5998--6008, 2017.

\bibitem[Chen et~al.(2018)Chen, Ma, and Xiao]{chen_fastgcn_2018}
Jie Chen, Tengfei Ma, and Cao Xiao.
\newblock {FastGCN}: {Fast} learning with graph convolutional networks via
  importance sampling.
\newblock In \emph{6th {International} {Conference} on {Learning}
  {Representations} ({ICLR})}, 2018.

\bibitem[Hamilton et~al.(2017)Hamilton, Ying, and
  Leskovec]{hamilton_inductive_2017}
Will Hamilton, Zhitao Ying, and Jure Leskovec.
\newblock Inductive representation learning on large graphs.
\newblock In \emph{Proceedings of the 31st {Conference} on {Neural}
  {Information} {Processing} {Systems} ({NIPS})}, pages 1024--1034, 2017.

\bibitem[Corso et~al.(2020)Corso, Cavalleri, Beaini, Li\`{o}, and
  Veli\v{c}kovi\'{c}]{corso_principal_2020}
Gabriele Corso, Luca Cavalleri, Dominique Beaini, Pietro Li\`{o}, and Petar
  Veli\v{c}kovi\'{c}.
\newblock Principal neighbourhood aggregation for graph nets.
\newblock In \emph{Proceedings of the 34th Conference on Neural Information
  Processing Systems (NeurIPS)}, pages 13260--13271, 2020.

\bibitem[Gallicchio and Micheli(2010)]{gallicchio_graph_2010}
Claudio Gallicchio and Alessio Micheli.
\newblock Graph echo state networks.
\newblock In \emph{Proceedings of the {International} {Joint} {Conference} on
  {Neural} {Networks} ({IJCNN})}, pages 1--8. IEEE, 2010.

\bibitem[Ying et~al.(2018)Ying, You, Morris, Ren, Hamilton, and
  Leskovec]{ying_hierarchical_2018}
Zhitao Ying, Jiaxuan You, Christopher Morris, Xiang Ren, Will Hamilton, and
  Jure Leskovec.
\newblock Hierarchical graph representation learning with differentiable
  pooling.
\newblock In \emph{Proceedings of the 32nd {Conference} on {Neural}
  {Information} {Processing} {Systems} ({NeurIPS})}, 2018.

\bibitem[Gao and Ji(2019)]{gao_graph_2019}
Hongyang Gao and Shuiwang Ji.
\newblock Graph {U}-nets.
\newblock In \emph{Proceedings of the 36th {International} {Conference} on
  {Machine} {Learning} ({ICML})}, pages 2083--2092, 2019.

\bibitem[Frederik~Diehl and Knoll(2019)]{frederik_diehl_towards_2019}
Michael Truong~Le Frederik~Diehl, Thomas~Brunner and Alois Knoll.
\newblock Towards graph pooling by edge contraction.
\newblock In \emph{Workshop on learning and reasoning with graph-structured
  data, {International} {Conference} on {Machine} {Learning} ({ICML})}, 2019.

\bibitem[Dhillon et~al.(2007)Dhillon, Guan, and Kulis]{dhillon_weighted_2007}
Inderjit~S Dhillon, Yuqiang Guan, and Brian Kulis.
\newblock Weighted graph cuts without eigenvectors a multilevel approach.
\newblock \emph{IEEE Transactions on Pattern Analysis and Machine
  Intelligence}, 29\penalty0 (11):\penalty0 1944--1957, 2007.

\bibitem[Bianchi et~al.(2021)Bianchi, Grattarola, Livi, and
  Alippi]{bianchi_graph_2021}
Filippo~Maria Bianchi, Daniele Grattarola, Lorenzo Livi, and Cesare Alippi.
\newblock Graph neural networks with convolutional arma filters.
\newblock \emph{IEEE Transactions on Pattern Analysis and Machine
  Intelligence}, 2021.

\bibitem[Bacciu and Di~Sotto(2019)]{bacciu_non-negative_2019}
Davide Bacciu and Luigi Di~Sotto.
\newblock A non-negative factorization approach to node pooling in graph
  convolutional neural networks.
\newblock In \emph{AI*IA 2019 -- Advances in Artificial Intelligence}, pages
  294--306. Springer, 2019.

\bibitem[Bacciu et~al.(2020{\natexlab{c}})Bacciu, Conte, Grossi, Landolfi, and
  Marino]{bacciu_k_2020}
Davide Bacciu, Alessio Conte, Roberto Grossi, Francesco Landolfi, and Andrea
  Marino.
\newblock K-plex cover pooling for graph neural networks.
\newblock In \emph{Workshop on Learning Meets Combinatorial Algorithms, Neural
  Information Processing Systems (NeurIPS)}, 2020{\natexlab{c}}.

\bibitem[Mesquita et~al.(2020)Mesquita, Souza, and
  Kaski]{mesquita_2020_rethinking}
Diego Mesquita, Amauri Souza, and Samuel Kaski.
\newblock Rethinking pooling in graph neural networks.
\newblock In H.~Larochelle, M.~Ranzato, R.~Hadsell, M.~F. Balcan, and H.~Lin,
  editors, \emph{Proceedings of the 34th Conference on Neural Information
  Processing Systems (NeurIPS)}, pages 2220--2231, 2020.

\bibitem[Zhang et~al.(2018)Zhang, Cui, Neumann, and Chen]{zhang_end--end_2018}
Muhan Zhang, Zhicheng Cui, Marion Neumann, and Yixin Chen.
\newblock An end-to-end deep learning architecture for graph classification.
\newblock In \emph{Proceedings of the 32nd {AAAI} {Conference} on {Artificial}
  {Intelligence} ({AAAI})}, 2018.

\bibitem[Trentin and Rigutini(2009)]{trentin_maximum_2009}
Edmondo Trentin and Leonardo Rigutini.
\newblock A maximum-likelihood connectionist model for unsupervised learning
  over graphical domains.
\newblock In \emph{Proceedings of the 12th International Conference on
  Artificial Neural Networks (ICANN)}, pages 40--49. Springer, 2009.

\bibitem[Bongini et~al.(2018)Bongini, Rigutini, and
  Trentin]{bongini_recursive_2018}
Marco Bongini, Leonardo Rigutini, and Edmondo Trentin.
\newblock Recursive neural networks for density estimation over generalized
  random graphs.
\newblock \emph{IEEE Transactions on Neural Networks and Learning Systems},
  29\penalty0 (11):\penalty0 5441--5458, 2018.

\bibitem[Ma et~al.(2020)Ma, Ma, Zhang, Sun, Liu, and Coates]{ma_memory_2020}
Chen Ma, Liheng Ma, Yingxue Zhang, Jianing Sun, Xue Liu, and Mark Coates.
\newblock Memory augmented graph neural networks for sequential recommendation.
\newblock In \emph{Proceedings of the AAAI Conference on Artificial
  Intelligence}, volume~34, pages 5045--5052, 2020.

\bibitem[Nguyen et~al.(2020)Nguyen, Le, Quinn, Nguyen, Le, and
  Venkatesh]{nguyen_graphdta_2020}
Thin Nguyen, Hang Le, Thomas~P Quinn, Tri Nguyen, Thuc~Duy Le, and Svetha
  Venkatesh.
\newblock {GraphDTA: Predicting drug–target binding affinity with graph
  neural networks}.
\newblock \emph{Bioinformatics}, 10 2020.

\bibitem[Xu et~al.(2020)Xu, Zhang, Guo, Guo, Tang, and
  Coates]{xu_graphsail_2020}
Yishi Xu, Yingxue Zhang, Wei Guo, Huifeng Guo, Ruiming Tang, and Mark Coates.
\newblock Graphsail: Graph structure aware incremental learning for recommender
  systems.
\newblock In \emph{Proceedings of the 29th ACM International Conference on
  Information \& Knowledge Management}, pages 2861--2868, 2020.

\bibitem[Macskassy and Provost(2007)]{macskassy_classification_2007}
Sofus~A Macskassy and Foster Provost.
\newblock Classification in networked data: {A} toolkit and a univariate case
  study.
\newblock \emph{Journal of Machine Learning Research}, 8\penalty0
  (May):\penalty0 935--983, 2007.

\bibitem[Kohonen(1990)]{kohonen_self_1990}
Teuvo Kohonen.
\newblock The self-organizing map.
\newblock \emph{Proceedings of the IEEE}, 78\penalty0 (9):\penalty0 1464--1480,
  1990.

\bibitem[Hagenbuchner et~al.(2003)Hagenbuchner, Sperduti, and
  Tsoi]{hagenbuchner_self_2003}
Markus Hagenbuchner, Alessandro Sperduti, and Ah~Chung Tsoi.
\newblock A self-organizing map for adaptive processing of structured data.
\newblock \emph{IEEE Transactions on Neural Networks}, 14\penalty0
  (3):\penalty0 491--505, 2003.

\bibitem[Hammer et~al.(2004{\natexlab{a}})Hammer, Micheli, Sperduti, and
  Strickert]{hammer_general_2004}
Barbara Hammer, Alessio Micheli, Alessandro Sperduti, and Marc Strickert.
\newblock A general framework for unsupervised processing of structured data.
\newblock \emph{Neurocomputing}, 57:\penalty0 3--35, 2004{\natexlab{a}}.

\bibitem[Hammer et~al.(2004{\natexlab{b}})Hammer, Micheli, Sperduti, and
  Strickert]{hammer_recursive_2004}
Barbara Hammer, Alessio Micheli, Alessandro Sperduti, and Marc Strickert.
\newblock Recursive self-organizing network models.
\newblock \emph{Neural Networks}, 17\penalty0 (8-9):\penalty0 1061--1085,
  2004{\natexlab{b}}.

\bibitem[Neuhaus and Bunke(2005)]{neuhaus_self_2005}
Michel Neuhaus and Horst Bunke.
\newblock Self-organizing maps for learning the edit costs in graph matching.
\newblock \emph{IEEE Transactions on Systems, Man, and Cybernetics, Part B
  (Cybernetics)}, 35\penalty0 (3):\penalty0 503--514, 2005.

\bibitem[Hagenbuchner et~al.(2009)Hagenbuchner, Sperduti, and
  Tsoi]{hagenbuchner_graph_2009}
Markus Hagenbuchner, Alessandro Sperduti, and Ah~Chung Tsoi.
\newblock Graph self-organizing maps for cyclic and unbounded graphs.
\newblock \emph{Neurocomputing}, 72\penalty0 (7-9):\penalty0 1419--1430, 2009.

\bibitem[Wang et~al.(2017)Wang, Pan, Long, Zhu, and Jiang]{wang_mgae_2017}
Chun Wang, Shirui Pan, Guodong Long, Xingquan Zhu, and Jing Jiang.
\newblock Mgae: Marginalized graph autoencoder for graph clustering.
\newblock In \emph{Proceedings of the ACM Conference on Information and
  Knowledge Management (CIKM)}, pages 889--898, 2017.

\bibitem[Velickovic et~al.(2019)Velickovic, Fedus, Hamilton, Liò, Bengio, and
  Hjelm]{velickovic_deep_2019}
Petar Velickovic, William Fedus, William~L. Hamilton, Pietro Liò, Yoshua
  Bengio, and R.~Devon Hjelm.
\newblock Deep {Graph} {Infomax}.
\newblock In \emph{7th {International} {Conference} on {Learning}
  {Representations} ({ICLR}), {New} {Orleans}, {LA}, {USA}, {May} 6-9, 2019},
  2019.

\bibitem[Shchur et~al.(2018)Shchur, Mumme, Bojchevski, and
  Günnemann]{shchur_pitfalls_2018}
Oleksandr Shchur, Maximilian Mumme, Aleksandar Bojchevski, and Stephan
  Günnemann.
\newblock Pitfalls of graph neural network evaluation.
\newblock \emph{Workshop on Relational Representation Learning, Neural
  Information Processing Systems (NeurIPS)}, 2018.

\bibitem[Zhang et~al.(2019)Zhang, Pal, Coates, and
  Ustebay]{zhang_bayesian_2019}
Yingxue Zhang, Soumyasundar Pal, Mark Coates, and Deniz Ustebay.
\newblock Bayesian graph convolutional neural networks for semi-supervised
  classification.
\newblock In \emph{Proceedings of the AAAI Conference on Artificial
  Intelligence}, volume~33, pages 5829--5836, 2019.

\bibitem[Pal et~al.(2020)Pal, Malekmohammadi, Regol, Zhang, Xu, and
  Coates]{pal_non_2020}
Soumyasundar Pal, Saber Malekmohammadi, Florence Regol, Yingxue Zhang, Yishi
  Xu, and Mark Coates.
\newblock Non parametric graph learning for bayesian graph neural networks.
\newblock In \emph{Conference on Uncertainty in Artificial Intelligence}, pages
  1318--1327. PMLR, 2020.

\bibitem[Hu et~al.(2020{\natexlab{a}})Hu, Liu, Gomes, Zitnik, Liang, Pande, and
  Leskovec]{hu_strategies_2020}
Weihua Hu, Bowen Liu, Joseph Gomes, Marinka Zitnik, Percy Liang, Vijay Pande,
  and Jure Leskovec.
\newblock Strategies for pre-training graph neural networks.
\newblock In \emph{8th International Conference on Learning Representations
  (ICLR)}, 2020{\natexlab{a}}.

\bibitem[Devlin et~al.(2019)Devlin, Chang, Lee, and
  Toutanova]{devlin_bert_2019}
Jacob Devlin, Ming-Wei Chang, Kenton Lee, and Kristina Toutanova.
\newblock Bert: Pre-training of deep bidirectional transformers for language
  understanding.
\newblock In \emph{Proceedings of the 2019 Conference of the North American
  Chapter of the Association for Computational Linguistics (NAACL)}, pages
  4171–--4186, 2019.

\bibitem[Liu et~al.(2021)Liu, Pan, Jin, Zhou, Xia, and Yu]{liu_graph_2021}
Yixin Liu, Shirui Pan, Ming Jin, Chuan Zhou, Feng Xia, and Philip~S Yu.
\newblock Graph self-supervised learning: A survey.
\newblock \emph{arXiv preprint arXiv:2103.00111}, 2021.

\bibitem[Lee et~al.(2019)Lee, Lee, and Kang]{lee_self-attention_2019}
Junhyun Lee, Inyeop Lee, and Jaewoo Kang.
\newblock Self-attention graph pooling.
\newblock In \emph{Proceedings of the 36th {International} {Conference} on
  {Machine} {Learning} ({ICML})}, pages 3734--3743, 2019.

\bibitem[McDermott(1976)]{mcdermott_artificial_1976}
Drew McDermott.
\newblock Artificial intelligence meets natural stupidity.
\newblock \emph{ACM SIGART Bulletin}, 57:\penalty0 4--9, 1976.

\bibitem[Lipton and Steinhardt(2019)]{lipton_troubling_2018}
Zachary~C. Lipton and Jacob Steinhardt.
\newblock Troubling trends in machine learning scholarship.
\newblock \emph{{ACM} Queue}, 17\penalty0 (1):\penalty0 80, 2019.

\bibitem[National Academies~of Sciences and
  Medicine(2019)]{NAS_reproducibility_2019}
Engineering National Academies~of Sciences and Medicine.
\newblock \emph{Reproducibility and replicability in science}.
\newblock National Academies Press, 2019.

\bibitem[Dacrema et~al.(2019)Dacrema, Cremonesi, and Jannach]{dacrema_are_2019}
Maurizio~Ferrari Dacrema, Paolo Cremonesi, and Dietmar Jannach.
\newblock Are we really making much progress? {A} worrying analysis of recent
  neural recommendation approaches.
\newblock In \emph{Proceedings of the 13th {ACM} Conference on Recommender
  Systems (RecSys)}, pages 101--109, 2019.

\bibitem[Stone(1974)]{stone_cross_1974}
Mervyn Stone.
\newblock Cross-validatory choice and assessment of statistical predictions.
\newblock \emph{Journal of the royal statistical society. Series B
  (Methodological)}, pages 111--147, 1974.

\bibitem[Varma and Simon(2006)]{varma_bias_2006}
Sudhir Varma and Richard Simon.
\newblock Bias in error estimation when using cross-validation for model
  selection.
\newblock \emph{BMC bioinformatics}, 7:\penalty0 91--98, 02 2006.

\bibitem[Cawley and Talbot(2010)]{cawley_overfitting_2010}
Gavin~C. Cawley and Nicola L.~C. Talbot.
\newblock On over-fitting in model selection and subsequent selection bias in
  performance evaluation.
\newblock \emph{Journal of Machine Learning Research}, 11:\penalty0 2079--2107,
  2010.

\bibitem[Wolpert(1996)]{wolpert_lack_1996}
David~H Wolpert.
\newblock The lack of a priori distinctions between learning algorithms.
\newblock \emph{Neural computation}, 8\penalty0 (7):\penalty0 1341--1390, 1996.

\bibitem[Dobson and Doig(2003)]{dd}
Paul~D Dobson and Andrew~J Doig.
\newblock Distinguishing enzyme structures from non-enzymes without alignments.
\newblock \emph{Journal of Molecular Biology}, 330\penalty0 (4):\penalty0
  771--783, 2003.

\bibitem[Borgwardt et~al.(2005)Borgwardt, Ong, Schönauer, Vishwanathan, Smola,
  and Kriegel]{proteins}
Karsten~M Borgwardt, Cheng~Soon Ong, Stefan Schönauer, SVN Vishwanathan,
  Alex~J Smola, and Hans-Peter Kriegel.
\newblock Protein function prediction via graph kernels.
\newblock \emph{Bioinformatics}, 21\penalty0 (suppl\_1):\penalty0 i47--i56,
  2005.

\bibitem[Wale et~al.(2008)Wale, Watson, and Karypis]{nci1}
Nikil Wale, Ian~A Watson, and George Karypis.
\newblock Comparison of descriptor spaces for chemical compound retrieval and
  classification.
\newblock \emph{Knowledge and Information Systems}, 14\penalty0 (3):\penalty0
  347--375, 2008.

\bibitem[Schomburg et~al.(2004)Schomburg, Chang, Ebeling, Gremse, Heldt, Huhn,
  and Schomburg]{enzymes}
Ida Schomburg, Antje Chang, Christian Ebeling, Marion Gremse, Christian Heldt,
  Gregor Huhn, and Dietmar Schomburg.
\newblock {BRENDA}, the enzyme database: updates and major new developments.
\newblock \emph{Nucleic Acids Research}, 32\penalty0 (suppl\_1), 2004.

\bibitem[Prechelt(1998)]{prechelt_early_1998}
Lutz Prechelt.
\newblock Early stopping-but when?
\newblock In \emph{Neural {Networks}: {Tricks} of the trade}, pages 55--69.
  Springer, 1998.

\bibitem[van Rossum(1963)]{vanrossum_relation_1963}
Jacques van Rossum.
\newblock {The Relation Between Chemical Structure and Biological Activity}.
\newblock \emph{Journal of Pharmacy and Pharmacology}, 15\penalty0
  (1):\penalty0 285--316, 1963.

\bibitem[Hu et~al.(2020{\natexlab{b}})Hu, Fey, Zitnik, Dong, Ren, Liu, Catasta,
  and Leskovec]{hu_open_2020}
Weihua Hu, Matthias Fey, Marinka Zitnik, Yuxiao Dong, Hongyu Ren, Bowen Liu,
  Michele Catasta, and Jure Leskovec.
\newblock Open graph benchmark: Datasets for machine learning on graphs.
\newblock In \emph{Proceedings of the 34th Conference on Neural Information
  Processing Systems (NeurIPS)}, pages 22118--22133, 2020{\natexlab{b}}.

\bibitem[Dwivedi et~al.(2020)Dwivedi, Joshi, Laurent, Bengio, and
  Bresson]{dwivedi_benchmarking_2020}
Vijay~Prakash Dwivedi, Chaitanya~K Joshi, Thomas Laurent, Yoshua Bengio, and
  Xavier Bresson.
\newblock Benchmarking graph neural networks.
\newblock \emph{arXiv preprint arXiv:2003.00982}, 2020.

\bibitem[Alder and Wainwright(1959)]{md_general_method}
B.~J. Alder and T.~E. Wainwright.
\newblock Studies in molecular dynamics. i. general method.
\newblock \emph{The Journal of Chemical Physics}, 31\penalty0 (2):\penalty0
  459--466, 1959.

\bibitem[Karplus(2002)]{md_sim_biomol}
Martin Karplus.
\newblock Molecular dynamics simulations of biomolecules.
\newblock \emph{Accounts of Chemical Research}, 35\penalty0 (6):\penalty0
  321--323, 2002.

\bibitem[Marrink et~al.(2007)Marrink, Risselada, Yefimov, Tieleman, and
  De~Vries]{marrink2007martini}
Siewert~J Marrink, H~Jelger Risselada, Serge Yefimov, D~Peter Tieleman, and
  Alex~H De~Vries.
\newblock The martini force field: coarse grained model for biomolecular
  simulations.
\newblock \emph{The journal of physical chemistry B}, 111\penalty0
  (27):\penalty0 7812--7824, 2007.

\bibitem[Takada(2012)]{Takada2012}
Shoji Takada.
\newblock {Coarse-grained molecular simulations of large biomolecules}.
\newblock \emph{Curr. Opin. Struct. Biol.}, 22\penalty0 (2):\penalty0 130--137,
  2012.

\bibitem[Potestio et~al.(2014)Potestio, Peter, and Kremer]{Potestio2014}
Raffaello Potestio, Christine Peter, and Kurt Kremer.
\newblock Computer simulations of soft matter: Linking the scales.
\newblock \emph{Entropy}, 16\penalty0 (8):\penalty0 4199--4245, 2014.

\bibitem[Saunders and Voth(2013)]{Saunders2013}
Marissa~G Saunders and Gregory~A Voth.
\newblock Coarse-graining methods for computational biology.
\newblock \emph{Annu. Rev. Biophys.}, 42:\penalty0 73--93, 2013.

\bibitem[Noid et~al.(2008)Noid, Chu, Ayton, Krishna, Izvekov, Voth, Das, and
  Andersen]{noid2008multiscale}
William~George Noid, Jhih-Wei Chu, Gary~S Ayton, Vinod Krishna, Sergei Izvekov,
  Gregory~A Voth, Avisek Das, and Hans~C Andersen.
\newblock The multiscale coarse-graining method. i. a rigorous bridge between
  atomistic and coarse-grained models.
\newblock \emph{The Journal of chemical physics}, 128\penalty0 (24):\penalty0
  244114, 2008.

\bibitem[Noid(2013)]{noid_mapping}
William~George Noid.
\newblock Systematic methods for structurally consistent coarse-grained models.
\newblock In \emph{Biomolecular Simulations}, pages 487--531. Springer, 2013.

\bibitem[Shell(2008)]{Shell2008}
M.~Scott Shell.
\newblock The relative entropy is fundamental to multiscale and inverse
  thermodynamic problems.
\newblock \emph{J. Chem. Phys.}, 129\penalty0 (14):\penalty0 144108, 2008.

\bibitem[Kmiecik et~al.(2016)Kmiecik, Gront, Kolinski, Wieteska, Dawid, and
  Kolinski]{kmiecik2016coarse}
Sebastian Kmiecik, Dominik Gront, Michal Kolinski, Lukasz Wieteska,
  Aleksandra~Elzbieta Dawid, and Andrzej Kolinski.
\newblock Coarse-grained protein models and their applications.
\newblock \emph{Chemical reviews}, 116\penalty0 (14):\penalty0 7898--7936,
  2016.

\bibitem[Webb et~al.(2019)Webb, Delannoy, and de~Pablo]{depabloJCTC2019}
Michael~A. Webb, Jean-Yves Delannoy, and Juan~J. de~Pablo.
\newblock Graph-based approach to systematic molecular coarse-graining.
\newblock \emph{Journal of Chemical Theory and Computation}, 15\penalty0
  (2):\penalty0 1199--1208, 2019.

\bibitem[Bereau and Kremer(2015)]{bereau2015automated}
Tristan Bereau and Kurt Kremer.
\newblock Automated parametrization of the coarse-grained martini force field
  for small organic molecules.
\newblock \emph{Journal of chemical theory and computation}, 11\penalty0
  (6):\penalty0 2783--2791, 2015.

\bibitem[Murtola et~al.(2007)Murtola, Kupiainen, Falck, and
  Vattulainen]{murtola2007conformational}
Teemu Murtola, Mikko Kupiainen, Emma Falck, and Ilpo Vattulainen.
\newblock Conformational analysis of lipid molecules by self-organizing maps.
\newblock \emph{The Journal of chemical physics}, 126\penalty0 (5):\penalty0
  054707, 2007.

\bibitem[Wang and G{\'o}mez-Bombarelli(2019)]{wang2019coarse}
Wujie Wang and Rafael G{\'o}mez-Bombarelli.
\newblock Coarse-graining auto-encoders for molecular dynamics.
\newblock \emph{npj Computational Materials}, 5\penalty0 (1):\penalty0 1--9,
  2019.

\bibitem[Li et~al.(2020)Li, Wellawatte, Chakraborty, Gandhi, Xu, and
  White]{li2020graph}
Zhiheng Li, Geemi~P Wellawatte, Maghesree Chakraborty, Heta~A Gandhi, Chenliang
  Xu, and Andrew~D White.
\newblock Graph neural network based coarse-grained mapping prediction.
\newblock \emph{Chemical Science}, 11\penalty0 (35):\penalty0 9524--9531, 2020.

\bibitem[Giulini et~al.(2020)Giulini, Menichetti, Shell, and
  Potestio]{giulini2020information}
Marco Giulini, Roberto Menichetti, M.~Scott Shell, and Raffaello Potestio.
\newblock An information-theory-based approach for optimal model reduction of
  biomolecules.
\newblock \emph{Journal of Chemical Theory and Computation}, 16\penalty0
  (11):\penalty0 6795--6813, 2020.

\bibitem[Foley et~al.(2015)Foley, Shell, and Noid]{foley2015impact}
Thomas~T Foley, M~Scott Shell, and William~George Noid.
\newblock The impact of resolution upon entropy and information in
  coarse-grained models.
\newblock \emph{The Journal of chemical physics}, 143\penalty0 (24):\penalty0
  243104, 2015.

\bibitem[Rudzinski and Noid(2011)]{rudzinski_2011}
Joseph~F. Rudzinski and W.~G. Noid.
\newblock Coarse-graining entropy, forces, and structures.
\newblock \emph{The Journal of Chemical Physics}, 135\penalty0 (21):\penalty0
  214101, 2011.

\bibitem[Shell(2012)]{Shell2012}
M.~Scott Shell.
\newblock Systematic coarse-graining of potential energy landscapes and
  dynamics in liquids.
\newblock \emph{J. Chem. Phys.}, 137\penalty0 (8):\penalty0 084503, 2012.

\bibitem[Wang and Landau(2001{\natexlab{a}})]{wang2001determining}
Fugao Wang and DP~Landau.
\newblock Determining the density of states for classical statistical models: A
  random walk algorithm to produce a flat histogram.
\newblock \emph{Physical Review E}, 64\penalty0 (5):\penalty0 056101,
  2001{\natexlab{a}}.

\bibitem[Wang and Landau(2001{\natexlab{b}})]{wang2001efficient}
Fugao Wang and David~P Landau.
\newblock Efficient, multiple-range random walk algorithm to calculate the
  density of states.
\newblock \emph{Physical review letters}, 86\penalty0 (10):\penalty0 2050,
  2001{\natexlab{b}}.

\bibitem[Shell et~al.(2002)Shell, Debenedetti, and
  Panagiotopoulos]{shell2002generalization}
M~Scott Shell, Pablo~G Debenedetti, and Athanassios~Z Panagiotopoulos.
\newblock Generalization of the wang-landau method for off-lattice simulations.
\newblock \emph{Physical review E}, 66\penalty0 (5):\penalty0 056703, 2002.

\bibitem[Barash et~al.(2017)Barash, Fadeeva, and Shchur]{barash2017control}
L~Yu Barash, MA~Fadeeva, and LN~Shchur.
\newblock Control of accuracy in the wang-landau algorithm.
\newblock \emph{Physical Review E}, 96\penalty0 (4):\penalty0 043307, 2017.

\bibitem[Glorot et~al.(2011)Glorot, Bordes, and Bengio]{glorot_deep_2011}
Xavier Glorot, Antoine Bordes, and Yoshua Bengio.
\newblock Deep sparse rectifier neural networks.
\newblock In \emph{Proceedings of the fourteenth international conference on
  artificial intelligence and statistics}, pages 315--323, 2011.

\bibitem[Kingma and Ba(2015)]{kingma_adam_2015}
Diederik~P. Kingma and Jimmy Ba.
\newblock Adam: {A} method for stochastic optimization.
\newblock In \emph{3rd {International} {Conference} on {Learning}
  {Representations} ({ICLR})}, 2015.

\bibitem[Bacciu et~al.(2010)Bacciu, Micheli, and Sperduti]{bacciu_bottom_2010}
Davide Bacciu, Alessio Micheli, and Alessandro Sperduti.
\newblock Bottom-up generative modeling of tree-structured data.
\newblock In \emph{Proceedings of the 17th International Conference on Neural
  Information Processing (ICONIP)}, 2010.

\bibitem[He et~al.(2016)He, Zhang, Ren, and Sun]{he_deep_2016}
Kaiming He, Xiangyu Zhang, Shaoqing Ren, and Jian Sun.
\newblock Deep residual learning for image recognition.
\newblock In \emph{Proceedings of the IEEE conference on Computer Vision and
  Pattern Recognition (CVPR)}, 2016.

\bibitem[Saul and Jordan(1999)]{saul_mixed_1999}
Lawrence~K Saul and Michael~I Jordan.
\newblock Mixed memory {Markov} models: {Decomposing} complex stochastic
  processes as mixtures of simpler ones.
\newblock \emph{Machine Learning}, 37\penalty0 (1):\penalty0 75--87, 1999.

\bibitem[Bacciu et~al.(2018{\natexlab{b}})Bacciu, Micheli, and
  Sperduti]{bacciu_generative_2018}
Davide Bacciu, Alessio Micheli, and Alessandro Sperduti.
\newblock Generative kernels for tree-structured data.
\newblock \emph{IEEE Transactions on Neural Networks and Learning Systems},
  29\penalty0 (10):\penalty0 4932--4946, 2018{\natexlab{b}}.

\bibitem[Niepert et~al.(2016)Niepert, Ahmed, and
  Kutzkov]{niepert_learning_2016}
Mathias Niepert, Mohamed Ahmed, and Konstantin Kutzkov.
\newblock Learning convolutional neural networks for graphs.
\newblock In \emph{Proceedings of the 33rd {International} {Conference} on
  {Machine} {Learning} ({ICML})}, pages 2014--2023, 2016.

\bibitem[Neumann et~al.(2012)Neumann, Patricia, Garnett, and
  Kersting]{neumann_efficient_2012}
Marion Neumann, Novi Patricia, Roman Garnett, and Kristian Kersting.
\newblock Efficient graph kernels by randomization.
\newblock In \emph{Proceedings of the {Joint} {European} {Conference} on
  {Machine} {Learning} and {Knowledge} {Discovery} in {Databases} ({ECML}
  {PKDD})}, pages 378--393. Springer, 2012.

\bibitem[Atwood and Towsley(2016)]{atwood_diffusion-convolutional_2016}
James Atwood and Don Towsley.
\newblock Diffusion-convolutional neural networks.
\newblock In \emph{Proceedings of the 30th {Conference} on {Neural}
  {Information} {Processing} {Systems} ({NIPS})}, pages 1993--2001, 2016.

\bibitem[Tran et~al.(2018)Tran, Navarin, and Sperduti]{tran_filter_2018}
Dinh~V Tran, Nicolò Navarin, and Alessandro Sperduti.
\newblock On filter size in graph convolutional networks.
\newblock In \emph{{IEEE} {Symposium} {Series} on {Computational}
  {Intelligence} ({SSCI})}, pages 1534--1541. IEEE, 2018.

\bibitem[Marquez et~al.(2018)Marquez, Hare, and Niranjan]{marquez_deep_2018}
Enrique~S Marquez, Jonathon~S Hare, and Mahesan Niranjan.
\newblock Deep cascade learning.
\newblock \emph{IEEE Transactions on Neural Networks and Learning Systems},
  29\penalty0 (11):\penalty0 5475--5485, 2018.

\bibitem[Blum and Reymond(2009)]{qm7b}
Lorenz~C Blum and Jean-Louis Reymond.
\newblock 970 million druglike small molecules for virtual screening in the
  chemical universe database gdb-13.
\newblock \emph{Journal of the American Chemical Society}, 131\penalty0
  (25):\penalty0 8732--8733, 2009.

\bibitem[Rupp et~al.(2012)Rupp, Tkatchenko, M{\"u}ller, and
  Von~Lilienfeld]{qm7b2}
Matthias Rupp, Alexandre Tkatchenko, Klaus~Robert M{\"u}ller, and O~Anatole
  Von~Lilienfeld.
\newblock Fast and accurate modeling of molecular atomization energies with
  machine learning.
\newblock \emph{Physical review letters}, 108\penalty0 (5):\penalty0 058301,
  2012.

\bibitem[Bergstra and Bengio(2012)]{bergstra_random_2012}
James Bergstra and Yoshua Bengio.
\newblock Random search for hyper-parameter optimization.
\newblock \emph{Journal of Machine Learning Research}, 13\penalty0
  (2):\penalty0 281--305, 2012.

\bibitem[Wang and Blei(2012)]{wang_truncation_2012}
Chong Wang and David Blei.
\newblock Truncation-free stochastic variational inference for bayesian
  nonparametric models.
\newblock \emph{Proceedings of the 26th {Conference} on {Neural} {Information}
  {Processing} {Systems} ({NIPS})}, pages 422--430, 2012.

\bibitem[Bryant and Sudderth(2012)]{bryant_truly_2012}
Michael Bryant and Erik Sudderth.
\newblock Truly nonparametric online variational inference for hierarchical
  dirichlet processes.
\newblock \emph{Proceedings of the 26th {Conference} on {Neural} {Information}
  {Processing} {Systems} ({NIPS})}, pages 2699--2707, 2012.

\bibitem[Hoffman et~al.(2013)Hoffman, Blei, Wang, and
  Paisley]{hoffman_stochastic_2013}
Matthew~D Hoffman, David~M Blei, Chong Wang, and John Paisley.
\newblock Stochastic variational inference.
\newblock \emph{Journal of Machine Learning Research}, 14\penalty0 (5), 2013.

\bibitem[Hughes et~al.(2015)Hughes, Kim, and Sudderth]{hughes_reliable_2015}
Michael Hughes, Dae~Il Kim, and Erik Sudderth.
\newblock Reliable and scalable variational inference for the hierarchical
  dirichlet process.
\newblock In \emph{Proceedings of the 18th Conference on Artificial
  Intelligence and Statistics (AISTATS)}, pages 370--378. PMLR, 2015.

\bibitem[Fox et~al.(2007)Fox, Sudderth, Jordan, and Willsky]{fox_sticky_2007}
Emily~B Fox, Erik~B Sudderth, Michael~I Jordan, and Alan~S Willsky.
\newblock The sticky hdp-hmm: Bayesian nonparametric hidden markov models with
  persistent states.
\newblock \emph{Preprint}, 2007.

\bibitem[Fox et~al.(2008)Fox, Sudderth, Jordan, and
  Willsky]{fox_stickyhdp_2008}
Emily~B Fox, Erik~B Sudderth, Michael~I Jordan, and Alan~S Willsky.
\newblock An hdp-hmm for systems with state persistence.
\newblock In \emph{Proceedings of the 25th {International} {Conference} on
  {Machine} {Learning} ({ICML})}, pages 312--319, 2008.

\bibitem[Lovell et~al.(2012)Lovell, Adams, and
  Mansinghka]{Lovell12parallelmarkov}
Dan Lovell, Ryan~P. Adams, and Vikash~K. Mansinghka.
\newblock Parallel markov chain monte carlo for dirichlet process mixtures.
\newblock In \emph{Workshop on Big Learning, Advances in Neural Information
  Processing Systems (NIPS)}, 2012.

\bibitem[Williamson et~al.(2013)Williamson, Dubey, and Xing]{Williamson2014}
Sinead Williamson, Avinava Dubey, and Eric Xing.
\newblock Parallel {M}arkov chain {M}onte {C}arlo for nonparametric mixture
  models.
\newblock In \emph{Proceedings of the 30th International Conference on Machine
  Learning (ICML)}, volume~28, pages 98--106. PMLR, 2013.

\bibitem[Chang and Fisher~III(2014)]{Chang2014}
Jason Chang and John~W Fisher~III.
\newblock Parallel sampling of hdps using sub-cluster splits.
\newblock In \emph{Advances in Neural Information Processing Systems (NIPS)},
  volume~27. Curran Associates, Inc., 2014.

\bibitem[Ge et~al.(2015)Ge, Chen, Wan, and Ghahramani]{Ge2015}
Hong Ge, Yutian Chen, Moquan Wan, and Zoubin Ghahramani.
\newblock Distributed inference for dirichlet process mixture models.
\newblock In \emph{Proceedings of the 32nd International Conference on Machine
  Learning}, volume~37, pages 2276--2284. PMLR, 2015.

\bibitem[Gal and Ghahramani(2014)]{Gal2014pitfalls}
Yarin Gal and Zoubin Ghahramani.
\newblock Pitfalls in the use of parallel inference for the dirichlet process.
\newblock In \emph{Proceedings of the 31st International Conference on Machine
  Learning (ICML)}, volume~32, pages 208--216. PMLR, 22--24 Jun 2014.

\bibitem[Bernardo and Smith(2009)]{bernardo_bayesian_2009}
Jos{\'e}~M Bernardo and Adrian~FM Smith.
\newblock \emph{Bayesian theory}, volume 405.
\newblock John Wiley \& Sons, 2009.

\bibitem[Goel and Degroot(1981)]{goel_information_1981}
Prem~K. Goel and Morris~H. Degroot.
\newblock Information about hyperparamters in hierarchical models.
\newblock \emph{Journal of the American Statistical Association}, 76\penalty0
  (373):\penalty0 140--147, 1981.
\newblock URL \url{http://www.jstor.org/stable/2287059}.

\bibitem[Bacci et~al.(2018)Bacci, Bartoli, Martinelli, Medvet, and
  Mercaldo]{bacci2018detection}
Alessandro Bacci, Alberto Bartoli, Fabio Martinelli, Eric Medvet, and Francesco
  Mercaldo.
\newblock Detection of obfuscation techniques in android applications.
\newblock In \emph{Proceedings of the 13th International Conference on
  Availability, Reliability and Security (ARES)}, 2018.

\bibitem[Suarez-Tangil et~al.(2017)Suarez-Tangil, Dash, Ahmadi, Kinder,
  Giacinto, and Cavallaro]{suarez2017droidsieve}
Guillermo Suarez-Tangil, Santanu~Kumar Dash, Mansour Ahmadi, Johannes Kinder,
  Giorgio Giacinto, and Lorenzo Cavallaro.
\newblock Droidsieve: Fast and accurate classification of obfuscated android
  malware.
\newblock In \emph{Proceedings of the 7th ACM on Conference on Data and
  Application Security and Privacy (CODASPY)}, pages 309--320, 2017.

\bibitem[Maiorca et~al.(2015)Maiorca, Ariu, Corona, Aresu, and
  Giacinto]{maiorca2015stealth}
Davide Maiorca, Davide Ariu, Igino Corona, Marco Aresu, and Giorgio Giacinto.
\newblock Stealth attacks: An extended insight into the obfuscation effects on
  android malware.
\newblock \emph{Computers \& Security}, 51:\penalty0 16--31, 2015.

\bibitem[Shang et~al.(2010)Shang, Zheng, Xu, Xu, and Zhang]{shang2010detecting}
Shanhu Shang, Ning Zheng, Jian Xu, Ming Xu, and Haiping Zhang.
\newblock Detecting malware variants via function-call graph similarity.
\newblock In \emph{Proceedings of the 5th International Conference on Malicious
  and Unwanted Software (MALWARE)}. IEEE, 2010.

\bibitem[Elhadi et~al.(2014)Elhadi, Maarof, Barry, and
  Hamza]{elhadi2014enhancing}
Ammar Ahmed~E Elhadi, Mohd~Aizaini Maarof, Bazara~IA Barry, and Hentabli Hamza.
\newblock Enhancing the detection of metamorphic malware using call graphs.
\newblock \emph{computers \& security}, 46:\penalty0 62--78, 2014.

\bibitem[Gascon et~al.(2013)Gascon, Yamaguchi, Arp, and
  Rieck]{gascon_structural_2013}
Hugo Gascon, Fabian Yamaguchi, Daniel Arp, and Konrad Rieck.
\newblock Structural detection of android malware using embedded call graphs.
\newblock In \emph{Proceedings of the 2013 ACM workshop on Artificial
  intelligence and security (AISec)}, 2013.

\bibitem[Iadarola et~al.(2020)Iadarola, Martinelli, Mercaldo, and
  Santone]{iadarola2020call}
Giacomo Iadarola, Fabio Martinelli, Francesco Mercaldo, and Antonella Santone.
\newblock Call graph and model checking for fine-grained android malicious
  behaviour detection.
\newblock \emph{Applied Sciences}, 10\penalty0 (22):\penalty0 7975--7994, 2020.

\bibitem[Canfora et~al.(2015)Canfora, De~Lorenzo, Medvet, Mercaldo, and
  Visaggio]{canfora2015effectiveness}
Gerardo Canfora, Andrea De~Lorenzo, Eric Medvet, Francesco Mercaldo, and
  Corrado~Aaron Visaggio.
\newblock Effectiveness of opcode ngrams for detection of multi family android
  malware.
\newblock In \emph{Proceedings of the 10th International Conference on
  Availability, Reliability and Security (ARES)}. IEEE, 2015.

\bibitem[Kapoor and Dhavale(2016)]{kapoor2016control}
Akshay Kapoor and Sunita Dhavale.
\newblock Control flow graph based multiclass malware detection using bi-normal
  separation.
\newblock \emph{Defence Science Journal}, 66\penalty0 (2):\penalty0 138--145,
  2016.

\bibitem[Vall{\'{e}}e{-}Rai et~al.(1999)Vall{\'{e}}e{-}Rai, Co, Gagnon,
  Hendren, Lam, and Sundaresan]{vallee2010soot}
Raja Vall{\'{e}}e{-}Rai, Phong Co, Etienne Gagnon, Laurie~J. Hendren, Patrick
  Lam, and Vijay Sundaresan.
\newblock Soot: A java bytecode optimization framework.
\newblock In \emph{Proceedings of the Conference of the Centre for Advanced
  Studies on Collaborative Research (CASCON)}, 1999.

\bibitem[Iadarola et~al.(2021)Iadarola, Martinelli, Mercaldo, and
  Santone]{iadarola2021towards}
Giacomo Iadarola, Fabio Martinelli, Francesco Mercaldo, and Antonella Santone.
\newblock Towards an interpretable deep learning model for mobile malware
  detection and family identification.
\newblock \emph{Computers \& Security}, 105:\penalty0 102198--103012, 2021.

\bibitem[Davis et~al.(2020)Davis, Hollingsworth, Caudron, and
  Irvine]{davis_use_2020}
Christopher~N Davis, T~Deirdre Hollingsworth, Quentin Caudron, and Michael~A
  Irvine.
\newblock The use of mixture density networks in the emulation of complex
  epidemiological individual-based models.
\newblock \emph{PLoS computational biology}, 16\penalty0 (3):\penalty0
  e1006869, 2020.

\bibitem[Kermack and McKendrick(1927)]{kermack_contribution_1927}
William~Ogilvy Kermack and Anderson~G McKendrick.
\newblock A contribution to the mathematical theory of epidemics.
\newblock \emph{Proceedings of the royal society of london. Series A,
  Containing papers of a mathematical and physical character}, 115\penalty0
  (772):\penalty0 700--721, 1927.

\bibitem[Pilania et~al.(2013)Pilania, Wang, Jiang, Rajasekaran, and
  Ramprasad]{pilania_accelerating_2013}
Ghanshyam Pilania, Chenchen Wang, Xun Jiang, Sanguthevar Rajasekaran, and
  Ramamurthy Ramprasad.
\newblock Accelerating materials property predictions using machine learning.
\newblock \emph{Scientific reports}, 3\penalty0 (1):\penalty0 1--6, 2013.

\bibitem[Opuszko and Ruhland(2013)]{opuszko_impact_2013}
Marek Opuszko and Johannes Ruhland.
\newblock Impact of the network structure on the {SIR} model spreading
  phenomena in online networks.
\newblock In \emph{Proceedings of the 8th International Multi-Conference on
  Computing in the Global Information Technology (ICCGI’13)}, 2013.

\bibitem[Valenchon and Coates(2019)]{valenchon_multiple_2019}
Juliette Valenchon and Mark Coates.
\newblock Multiple-graph recurrent graph convolutional neural network
  architectures for predicting disease outcomes.
\newblock In \emph{ICASSP 2019-2019 IEEE International Conference on Acoustics,
  Speech and Signal Processing (ICASSP)}, pages 3157--3161. IEEE, 2019.

\bibitem[Nowicki and Snijders(2001)]{nowicki_estimation_2001}
Krzysztof Nowicki and Tom A~B Snijders.
\newblock Estimation and prediction for stochastic blockstructures.
\newblock \emph{Journal of the American statistical association}, 96\penalty0
  (455):\penalty0 1077--1087, 2001.

\bibitem[Jacobs et~al.(1991)Jacobs, Jordan, Nowlan, and
  Hinton]{jacobs_adaptive_1991}
Robert~A Jacobs, Michael~I Jordan, Steven~J Nowlan, and Geoffrey~E Hinton.
\newblock Adaptive mixtures of local experts.
\newblock \emph{Neural computation}, 3\penalty0 (1):\penalty0 79--87, 1991.

\bibitem[Jordan and Jacobs(1994)]{jordan_hierarchical_1994}
Michael~I Jordan and Robert~A Jacobs.
\newblock Hierarchical mixtures of experts and the em algorithm.
\newblock \emph{Neural computation}, 6\penalty0 (2):\penalty0 181--214, 1994.

\bibitem[Corduneanu and Bishop(2001)]{corduneanu_variational_2001}
Adrian Corduneanu and Christopher~M Bishop.
\newblock Variational bayesian model selection for mixture distributions.
\newblock In \emph{Artificial intelligence and Statistics}, volume 2001, pages
  27--34. Morgan Kaufmann Waltham, MA, 2001.

\bibitem[Eigen et~al.(2013)Eigen, Ranzato, and Sutskever]{eigen_learning_2013}
David Eigen, Marc'Aurelio Ranzato, and Ilya Sutskever.
\newblock Learning factored representations in a deep mixture of experts.
\newblock \emph{International Conference on Learning Representations (ICLR)
  Workshop}, 2013.

\bibitem[Pereyra et~al.(2017)Pereyra, Tucker, Chorowski, Kaiser, and
  Hinton]{pereyra_regularizing_2017}
Gabriel Pereyra, George Tucker, Jan Chorowski, {\L}ukasz Kaiser, and Geoffrey
  Hinton.
\newblock Regularizing neural networks by penalizing confident output
  distributions.
\newblock \emph{International Conference on Learning Representations (ICLR)
  Workshop}, 2017.

\bibitem[Cover(1999)]{cover_elements_1999}
Thomas~M Cover.
\newblock \emph{Elements of information theory}.
\newblock John Wiley \& Sons, 1999.

\bibitem[Hel{\'e}n and Virtanen(2007)]{helen_query_2007}
Marko Hel{\'e}n and Tuomas Virtanen.
\newblock Query by example of audio signals using euclidean distance between
  gaussian mixture models.
\newblock In \emph{2007 IEEE International Conference on Acoustics, Speech and
  Signal Processing-ICASSP'07}, volume~1, pages 1--225. IEEE, 2007.

\bibitem[Hinton and Roweis(2002)]{hinton_stochastic_2003}
Geoffrey~E Hinton and Sam~T Roweis.
\newblock Stochastic neighbor embedding.
\newblock In \emph{Proceedings of the 16th {Conference} on {Neural}
  {Information} {Processing} {Systems} ({NIPS})}, pages 857--864, 2002.

\bibitem[Maaten and Hinton(2008)]{maaten_visualizing_2008}
Laurens van~der Maaten and Geoffrey Hinton.
\newblock Visualizing data using t-sne.
\newblock \emph{Journal of machine learning research}, 9\penalty0
  (Nov):\penalty0 2579--2605, 2008.

\bibitem[You(2021)]{you_l2_2019}
Kisung You.
\newblock $l_2$ distance between two gaussian mixture models.
\newblock 2021.
\newblock URL \url{https://kisungyou.com/notes/note004/main004.pdf}.

\bibitem[Petersen and Pedersen(2012)]{petersen_matrix_2012}
KB~Petersen and MS~Pedersen.
\newblock The matrix cookbook, version 20121115.
\newblock \emph{Technical Univ. Denmark, Kongens Lyngby, Denmark, Tech. Rep},
  3274, 2012.

\bibitem[Seshadhri et~al.(2012)Seshadhri, Kolda, and
  Pinar]{seshadhri_community_2012}
Comandur Seshadhri, Tamara~G Kolda, and Ali Pinar.
\newblock Community structure and scale-free collections of
  erd{\H{o}}s-r{\'e}nyi graphs.
\newblock \emph{Physical Review E}, 85\penalty0 (5):\penalty0 056109, 2012.

\bibitem[Chen et~al.(2019)Chen, Chen, Hsieh, Lee, Liao, Liao, Liu, Qiu, Sun,
  Tang, et~al.]{chen_alchemy_2019}
Guangyong Chen, Pengfei Chen, Chang-Yu Hsieh, Chee-Kong Lee, Benben Liao,
  Renjie Liao, Weiwen Liu, Jiezhong Qiu, Qiming Sun, Jie Tang, et~al.
\newblock Alchemy: A quantum chemistry dataset for benchmarking ai models.
\newblock \emph{arXiv preprint arXiv:1906.09427}, 2019.

\bibitem[Irwin et~al.(2012)Irwin, Sterling, Mysinger, Bolstad, and
  Coleman]{irwin_zinc_2012}
John~J Irwin, Teague Sterling, Michael~M Mysinger, Erin~S Bolstad, and Ryan~G
  Coleman.
\newblock Zinc: a free tool to discover chemistry for biology.
\newblock \emph{Journal of chemical information and modeling}, 52\penalty0
  (7):\penalty0 1757--1768, 2012.

\bibitem[Bresson and Laurent(2019)]{bresson_two_2019}
Xavier Bresson and Thomas Laurent.
\newblock A two-step graph convolutional decoder for molecule generation.
\newblock In \emph{Workshop on Machine Learning and the Physical Sciences,
  Neural Information Processing Systems (NeurIPS)}, 2019.

\bibitem[Wu et~al.(2019)Wu, Pan, Long, Jiang, and Zhang]{wu_graph_2019}
Zonghan Wu, Shirui Pan, Guodong Long, Jing Jiang, and Chengqi Zhang.
\newblock Graph wavenet for deep spatial-temporal graph modeling.
\newblock In \emph{Proceedings of the 28th International Joint Conference on
  Artificial Intelligence {(IJCAI)}}, pages 1907--1913. International Joint
  Conferences on Artificial Intelligence Organization, 7 2019.
\newblock \doi{10.24963/ijcai.2019/264}.

\bibitem[Wu et~al.(2020{\natexlab{b}})Wu, Pan, Long, Jiang, Chang, and
  Zhang]{wu_connecting_2020}
Zonghan Wu, Shirui Pan, Guodong Long, Jing Jiang, Xiaojun Chang, and Chengqi
  Zhang.
\newblock Connecting the dots: Multivariate time series forecasting with graph
  neural networks.
\newblock In \emph{Proceedings of the 26th ACM International Conference on
  Knowledge Discovery \& Data Mining (SIGKDD)}, pages 753--763,
  2020{\natexlab{b}}.

\bibitem[Kahn et~al.(2017)Kahn, Villaflor, Pong, Abbeel, and
  Levine]{kahn_uncertainty_2017}
Gregory Kahn, Adam Villaflor, Vitchyr Pong, Pieter Abbeel, and Sergey Levine.
\newblock Uncertainty-aware reinforcement learning for collision avoidance.
\newblock \emph{arXiv preprint arXiv:1702.01182}, 2017.

\bibitem[Choi et~al.(2018)Choi, Lee, Lim, and Oh]{choi_uncertainty_2018}
Sungjoon Choi, Kyungjae Lee, Sungbin Lim, and Songhwai Oh.
\newblock Uncertainty-aware learning from demonstration using mixture density
  networks with sampling-free variance modeling.
\newblock In \emph{2018 IEEE International Conference on Robotics and
  Automation (ICRA)}, pages 6915--6922. IEEE, 2018.

\bibitem[Errica et~al.(2020{\natexlab{b}})Errica, Bacciu, and
  Micheli]{errica_theoretically_2020}
Federico Errica, Davide Bacciu, and Alessio Micheli.
\newblock Theoretically expressive and edge-aware graph learning.
\newblock In \emph{Proceedings of the 28th European Symposium on Artificial
  Neural Networks, Computational Intelligence and Machine Learning (ESANN)},
  2020{\natexlab{b}}.

\bibitem[Carta et~al.(2021)Carta, Cossu, Errica, and
  Bacciu]{carta_catastrophic_2021}
Antonio Carta, Andrea Cossu, Federico Errica, and Davide Bacciu.
\newblock Catastrophic forgetting in deep graph networks: an introductory
  benchmark for graph classification.
\newblock In \emph{Graph Learning Benchmark Workshop, The Web Conference
  (WWW)}, 2021.

\bibitem[Errica et~al.(2021{\natexlab{d}})Errica, Silvestri, Edizel, Denoyer,
  Petroni, Plachouras, and Riedel]{errica_concept_2021}
Federico Errica, Fabrizio Silvestri, Bora Edizel, Ludovic Denoyer, Fabio
  Petroni, Vassilis Plachouras, and Sebastian Riedel.
\newblock Concept matching for low-resource classification.
\newblock In \emph{Proceedings of the International Joint Conference on Neural
  Networks (IJCNN)}, pages 1--8, 2021{\natexlab{d}}.

\end{thebibliography}

\end{document}